\documentclass{article}

\usepackage[english]{babel}
\usepackage[fixlanguage]{babelbib}

\usepackage[letterpaper,top=2cm,bottom=2cm,left=3cm,right=3cm,marginparwidth=1.75cm]{geometry}

\usepackage{amssymb}

\usepackage[dvipsnames]{xcolor}
\usepackage{amsmath,amsthm,amsfonts,graphicx,hyperref,xurl,soul,color,booktabs,caption,float,multirow,array,longtable,enumitem,makecell,mwe,subcaption,authblk,hyperref,rotating}

\theoremstyle{definition}
\newtheorem{definition}{Definition}

\newcommand{\giornale}{\textit}
\newcommand{\conferenza}{}

\AtBeginEnvironment{longtable}{\footnotesize}
\AtBeginEnvironment{table}{\footnotesize}
\AtBeginEnvironment{tabular}{\footnotesize}

\usepackage{csquotes}
\usepackage{array}
\newcolumntype{R}[1]{>{\raggedright}p{#1}}

\title{A Systematic Literature Review of Spatio-Temporal Graph Neural Network Models for Time Series Forecasting and Classification}

\author[1]{Flavio Corradini}
\author[1]{Flavio Gerosa}
\author[2]{Marco Gori}
\author[1]{Carlo Lucheroni}
\author[1]{Marco Piangerelli}
\author[1,3]{Martina Zannotti\thanks{Corresponding author: \href{martina.zannotti@unicam.it}{martina.zannotti@unicam.it}}}

\affil[1]{School of Science and Technology, University of Camerino, via Madonna delle Carceri 9, Camerino, 62032, Italy}
\affil[2]{DIISM, University of Siena, Via Roma 56, Siena, 53100, Italy}
\affil[3]{Syeew s.r.l., Via Cavour 2, Jesi, 60035, Italy}

\date{}

\begin{document}

\maketitle

\begin{abstract}
In recent years, spatio-temporal graph neural networks (GNNs) have attracted considerable interest in the field of time series analysis, due to their ability to capture, at once, dependencies among variables and across time points. The objective of this systematic literature review is hence to provide a comprehensive overview of the various modeling approaches and application domains of GNNs for time series classification and forecasting. A database search was conducted, and 366 papers were selected for a detailed examination of the current state-of-the-art in the field. This examination is intended to offer to the reader a comprehensive review of proposed models, links to related source code, available datasets, benchmark models, and fitting results. All this information is hoped to assist researchers in their studies. To the best of our knowledge, this is the first and broadest systematic literature review presenting a detailed comparison of results from current spatio-temporal GNN models applied to different domains. In its final part, this review discusses current limitations and challenges in the application of spatio-temporal GNNs, such as comparability, reproducibility, explainability, poor information capacity, and scalability. This paper is complemented by a GitHub repository at \url{https://github.com/FlaGer99/SLR-Spatio-Temporal-GNN.git} providing additional interactive tools to further explore the presented findings.
\end{abstract}

\thispagestyle{plain} 
\begingroup
\renewcommand\thefootnote{}\footnotetext{
\hspace{-0.3cm}{Accepted for publication in \textit{Neural Networks}}
}

\section{Introduction}
\label{section:Introduction}
In recent years, graph neural networks (GNNs) have emerged as a powerful class of artificial neural network models aimed at processing data that can be represented as graphs \cite{Zheng2024}. GNNs are in fact particularly well-suited for a wide range of practical and engineering applications where data naturally lend themselves to be represented as graph structures, such as applications to transportation networks \cite{review_traffico_jiang}, image analysis \cite{GNN_image}, classification of molecular structure \cite{GNN_molecule}, and natural language processing \cite{GNN_NLP}. The intuitive concept behind the GNN approach is that the nodes of a graph can be put in correspondence to objects or concepts, whereas the edges can be assumed to represent their mutual relationships \cite{Scarselli2009}. Data for these quantities can then be processed at once by a GNN model. GNNs can be used at three different levels, related to three distinct classes of problems: at graph-level, at edge-level, and at node-level \cite{review_traffico_con_codici}. In graph-level problems, the goal is to predict a property of a graph based on its entire structure, rather than on individual nodes or edges. For example, a molecule from a sample can be represented as a graph, and global properties of this molecule can be inferred from data coming from its entire structure in relation to the sample.
In edge-level problems, the goal is to predict the presence or absence of edges between pairs of nodes. As an example, this task can be used in recommendation systems to predict potential connections between users and items on the basis of past interactions.
In node-level problems, the goal is to predict the role or value of each node within a graph, in order to solve either a step-wise classification or a regression task. Noticeably, regression tasks can sometimes be put in the form of temporal \enquote{autoregressions}, that is, they can include time. When introduced, GNNs were at first applied to static datasets. The inclusion of time in GNN modeling has proven to have far-reaching consequences. Spatio-temporal GNNs have been applied to a wide range of time series tasks, including forecasting, classification, anomaly detection, and data imputation.

This review examines the available literature on the use of spatio-temporal GNNs for time series step-wise classification and future data forecasting, as these are the tasks most commonly addressed in the literature and best illustrate the current developments in spatio-temporal GNNs. The idea behind the use of GNNs for time series problems is that GNNs can be made capable of capturing complex relationships, both inter-variable (connections between different variables within a multivariate series) but also inter-temporal (dependencies between different points in time) at once \cite{review_Alippi}. The inclusion of this capability in GNN models results in a characteristic \textit{spatio-temporal} approach. Within this approach, the spatial dimension is related to the multivariate framework, where nodes typically correspond to different features of a multivariate time series at a given time step $t$, and edges to their relationships, and the temporal dimension is related to the temporal nature of the data. Spatio-temporal GNNs can thus be seen as a class of GNN models especially designed to handle data in both spatial and temporal dimensions at once. In a spatio-temporal graph, data in each node or edge can evolve over time, and the challenge is then to model their related dynamic interactions over time.

In the broader landscape of time series modeling, a variety of other approaches have been developed alongside spatio-temporal GNNs. Early statistical frameworks, such as ARIMA, and, more prominently, neural networks designed for sequential data, including RNNs, LSTMs, and GRUs, have been for a long time the standard solution for modeling temporal data. These architectures effectively capture short and medium term time dependencies, but struggle with very long sequences and with explicitly modeling interactions among multiple variables. The attention mechanism and the Transformer architecture, originally developed for handling long, static word sequences in Large Language Models (LLMs), were later adopted to better address long term dependencies in time series \cite{LLM_review}. Building on these architectural advances, and supported by the availability of modern GPU hardware, many experiments have been conducted. However, it remains unclear whether these highly parallelizable architectures are truly suitable for time series tasks \cite{tan2024language}. The development of LLMs has also stimulated a further line of research in time series modeling. LLMs, trained on massive text datasets for capturing general patterns in language, can then be adapted to a variety of downstream task, that is, as foundational models. Inspired by this approach, researchers have begun developing time series foundation models (TSFMs), designed to learn general representations from diverse temporal datasets \cite{TSFM_review}. These models, which do not necessarily use LLMs, are trained on large collections of time series to enable them to generalize effectively to new sequences.

All these approaches share the ability to model the dynamics of temporal data, and are inherently sequential models. What differentiates spatio-temporal GNNs from conventional machine learning, LLMs applied to dynamics, and TSFMs is how they handle the mutual relationships among features in multivariate time series. In TSFMs, these relationships are supposed to be learned end-to-end. In contrast, spatio-temporal GNNs explicitly prioritize these relationships. In the case of GNN models based on pre-defined static adjacency matrices, the inter-variable relationships are effectively \enquote{forced} into the model, like handcrafted features (which TSFMs could not be able to learn). As for GNNs using self-learned adjacency matrices, these more recent models are naturally biased toward explicitly identifying these relationships, offering a clearly interpretable representation of inter-dependencies.

Notwithstanding all the potential interest for such important features, to the best of our knowledge there is a lack of systematic literature reviews (SLRs) on spatio-temporal GNN models for time series applications. Most of the existing surveys are not SLRs, or they are not recent, or only focus on a limited number of application domains and specific issues not necessarily related to time series. A much broader view on the subject is nowadays necessary. For example, recent reviews focus on algorithmic features \cite{cit_rev2, cit_rev3, survey_Lucheroni, survey_Chikwendu}, GNN characterization \cite{cit_rev8, review_Alippi}, specific fields \cite{cit_rev4, Fan2025}, and distributed training \cite{cit_rev5}. Only two SLR on spatio-temporal GNNs were actually recently published, in \cite{survey_spatiotemporal} and \cite{Fan2025}. However, Ref.~\cite{survey_spatiotemporal} uses a restrictive search query that limits the SLR to only 52 publications, whereas Ref.~\cite{Fan2025} examines a broader collection of papers, but it is only focused on traffic flow prediction. In addition, \cite{survey_kumar_spatiotemporal_data} is a recent survey on spatio-temporal predictive modeling techniques for different domains. However, the authors approach the topic from a more general perspective, analyzing different machine learning models without placing too much emphasis on spatio-temporal GNNs. As an organized summary of these contributions, a list with a short description of the main reviews on spatio-temporal GNN models is provided in Tab.~\ref{t:tabella altre review}.

\begin{longtable}{R{3.8cm}p{11cm}}
    \caption{Short description of main reviews on spatio-temporal GNN models. At the time of writing this SLR, articles marked with a $\diamond$ were only available as preprints.} \label{t:tabella altre review}
    \endfirsthead
    \endhead
    \toprule
    Review & Description \\
    \midrule
    Wu et al.~(2020) \cite{cit_rev3} & One of the first reviews on the use of GNNs in machine learning and data mining. It introduces a taxonomy that classifies GNN architectures into four categories: recurrent, convolutional, graph autoencoders, and spatial-temporal GNNs. The focus of this review is primarily on early GNN models, rather than applications on time series data. \\
    
    Zhou et al.~(2020) \cite{review_traffico_con_codici} & The authors outline a general design pipeline for GNN models, covering key aspects such as the graph structure, pooling techniques, loss function design, and components of the propagation module (including recurrent and convolutional operators, and skip connections). They present some applications of GNNs, including physical systems modeling, chemical analysis, text processing, image classification, recommendation systems, and analysis of other types of data. \\
    
    Jin et al.~(2023) \cite{cit_rev4} & Review on the applications of spatio-temporal GNNs to predictive learning in urban computing. In the specific, the authors describe four methods for the construction of pre-defined graphs: topology-based, distance-based, similarity-based and interaction-based approaches. Then, they discuss various design strategies for spatio-temporal GNNs and their combination with some advanced learning frameworks, such as adversarial learning, continuous learning, and physics informed learning. \\
    
    Li et al.~(2023) \cite{li2023graphneuralnetworkspatiotemporal}$^\diamond$ & The authors aim to provide a systematic review of spatio-temporal mining using GNNs. After providing an overview of the graph structure and the spatial and temporal components of GNNs, the authors describe some applications in fields such as transportation, disaster management, environment, power systems, and neuroscience. However, unlike our review, they do not outline the selection process of the papers or include detailed and comprehensive tables that cover datasets, benchmarks and results. \\
    
    Lin et al.~(2023) \cite{cit_rev5} & This review is specialized on distributed training of GNNs. It investigates optimization techniques used in this context, including parallel mini-batch generation, dynamic mini-batch allocation, mini-batch transmission pipelining, and parallel aggregation with edge partitioning. \\
    
    Longa et al.~(2023) \cite{cit_rev2} & The authors offer an overview of GNNs for temporal graphs. They describe supervised and unsupervised learning tasks on temporal graphs, and present a taxonomy of temporal GNNs that groups the model in snapshot-based and event-based approaches. \\
    
    Chen et al.~(2024) \cite{survey_Lucheroni} & One of the first reviews on the use of GNNs for time series modeling. The authors thoroughly describe a wide range of models for time series classification, forecasting, and anomaly detection, based either on recurrent/convolutional GNNs and attentional GNNs. Moreover, they provide a brief comparison of the models' performances in traffic forecasting and anomaly detection in water treatment. \\
    
    Chikwendu et al.~(2024) \cite{survey_Chikwendu} & This survey provides an overview of graph neural architecture search, and explores methods for representing nodes, edges, and sub-graphs, which add context and semantics to graphs. The authors also describe some applications, including computer vision, brain networks, recommendation system, biomedical application, natural language processing, and traffic. \\
    
    Feng et al.~(2024) \cite{feng2024comprehensivesurveydynamicgraph}$^\diamond$ & This survey examines 81 dynamic GNN models. It introduces a novel taxonomy that categorizes GNNs into discrete-time dynamic graphs and continuous-time dynamic graphs based on their structures, features, and dynamic modeling methods. In addition, the authors perform numerical experiments which evaluate 12 significant models on 6 graph datasets, which are not inherently related to time series. \\
    
    Jin et al.~(2024) \cite{review_Alippi} & The authors present a comprehensive review of GNNs for time series analysis, covering forecasting, classification, anomaly detection, and imputation. They categorize spatio-temporal GNNs according to their three main components: the spatial module, the temporal module, and the overall model architecture. Although they focus on the same topic as our SLR, they do not provide a detailed comparison of the results of existing models in the literature. \\
    
    Khemani et al.~(2024) \cite{Khemani2024} & The review discusses general concepts of GNNs, including architectures, techniques, datasets, applications, and challenges, offering a broad analysis that emphasizes these aspects without focusing specifically on time series analysis. It provides a brief description of papers in the literature, detailing application areas, datasets used, models applied, and performance evaluation. \\
    
    Kumar et al.~(2024) \cite{survey_kumar_spatiotemporal_data} & This SLR includes 200 papers from high-impact journals and prominent conferences on spatio-temporal predictive modeling techniques across different domains. Although its structure is similar to that of our review, its focus is broader and not specifically on spatio-temporal GNNs, as it also covers and discusses statistical and non-GNN machine learning models. \\
    
    Zeghina et al.~(2024) \cite{survey_spatiotemporal} & This SLR includes 52 papers on spatio-temporal GNNs across various application domains. Compared to our SLR, it covers a limited set of papers due to a restricted search query and it does not provide detailed and comprehensive tables of datasets, benchmarks, and model results. \\

    Capone et al.~(2025) \cite{CAPONE2025130400} & This review provides an overview of recent spatio-temporal GNN models and classifies them as time-then-space (TTS), space-then-time (STT), and combined space-and-time (S\&T) models. Compared to our SLR, this review is not systematic, it includes a limited number of papers, and it does not provide detailed quantitative comparisons of model performance across datasets. \\

    Fan et al.~(2025) \cite{Fan2025} & This SLR covers 86 studies on the use of GNNs for traffic flow prediction. It examines data processing, model deployment, graph construction methods, and network architectures, discussing their respective strengths and weaknesses. However, unlike our review, its focus is on a single application domain. \\

    Wu et al.~(2025) \cite{cit_rev8} & This review offers a comprehensive survey of GNN models from the perspective of computation theory. In particular, the authors compare existing GNN models through various perspectives: time, memory, communication, parallelism, and execution. \\
    
    \bottomrule
\end{longtable}

While most existing reviews focus on specific subfields, the present work takes a unique broader perspective. Its aim is to systematically collect and synthesize a wide range of publications on spatio-temporal GNNs for time series tasks, covering multiple application domains and methodological approaches. As a result, the number of articles included in this survey is larger than in more narrowly focused reviews. Although this approach results in a longer text, we hope it is able to provide a unique overview of the field, capturing both the diversity of applications and recent developments that cannot be fully appreciated in more specialized surveys.

The intention of the SLR presented here is therefore to: (i) offer a wide-ranging overview of spatio-temporal GNNs for time series classification and forecasting in different domains, (ii) assess whether their popularity stems from actual effectiveness, (iii) evaluate their efficacy and accuracy in different fields, (iv) collect results and benchmarks from the literature on GNN applications to support researchers in their work. In addition, to the best of our knowledge, this is the first review presenting comprehensive tables with all the results from the various models and benchmarks proposed in a highly fragmented literature. The tables presented in the paper are also available in the accompanying GitHub repository in an interactive format, enabling users to sort, filter, and explore the results of the various models. This synthesis and comparison of the results is performed by gathering data from the existing literature, without conducting any numerical experiment on data. Moreover, given the large number of selected papers, it was not feasible to include a description of all the proposed models. To compensate this limitation, we provide a description of the most commonly used benchmarks and the most accurate models within each group. In case, for more detailed descriptions of experimental setups and data pre-processing techniques, the reader is referred to the original papers. It is hoped that this collected and distilled knowledge can anyway become a valuable resource and reference for researchers.

During the development of this review, we formulated two sets of questions, one general and one specific, which are outlined below. The answers to the first set want to provide a general overview of the publications in the field, whereas the answers to the second set focus on specific aspects of the proposed spatio-temporal GNN models. \\
The general overview questions (GQs) are:
\begin{enumerate}[label=GQ\arabic*), left=1.5em]
    \item \textbf{Trend}: What is the temporal trend of the number of publications on spatio-temporal GNN models? Is there a growing interest in GNNs?
    \item \textbf{Fields}: In which fields are GNN models most commonly applied?
    \item \textbf{Journals}: In which journals are the papers on GNNs published? Are these journals chosen because of the specific domain of application, or not?
    \item \textbf{Conferences}: What are the most prominent conferences where GNN research is presented? 
    \item \textbf{Research groups}: Which are the most active research groups on GNNs? 
    \item \textbf{Tools}: What tools (programming languages, libraries, frameworks) are used to implement the GNN models?
    \item \textbf{Fundings}: Did the authors of the papers receive public or private funding for their research?
\end{enumerate}
The specific questions (SQs) are:
\begin{enumerate}[label=SQ\arabic*), left=1.5em]
    \item \textbf{Applications}: Which are the most studied applications of spatio-temporal GNNs? Are there differences in approaches and results across different application areas?
    \item \textbf{Graph construction}: Is the graph structure predetermined? If not, how do researchers define it?
    \item \textbf{Taxonomy}: Among the various taxonomy classes for GNNs, which are the most common? Are there some recurring mechanisms?
    \item \textbf{Benchmark models}: Which are the most commonly used benchmark models? Are they non-GNN machine learning models or also GNN models?
    \item \textbf{Benchmark datasets}: Which are the most commonly used benchmark datasets?
    \item \textbf{Modeling paradigms}: Which is the most common modeling paradigm? Is it modeling complex interacting systems, or modeling a system with multiple interacting quantities? Specifically, do graphs typically condense the relationships among multiple entities, or they describe different aspects of the same entity? In other words, is the used graph structure homogeneous or heterogeneous?
    \item \textbf{Metrics}: Which are the most common metrics used to assess the accuracy of a given model?
\end{enumerate}
As the discussion will progress, the answers to both the general and the specific questions will be integrated in it. This will hopefully provide a comprehensive overview of the state-of-the-art of spatio-temporal GNN models. At the same time, a general overview of such GNN models will be assembled by highlighting similarities and differences between the approaches across different fields. In case of need of an even more detailed analysis of a specific discipline, or other aspects of GNN models, the reader is invited to consult the specialized reviews mentioned before \cite{cit_rev1, cit_rev2, cit_rev3, survey_Lucheroni, cit_rev8, cit_rev4, cit_rev5, cit_rev6, Fan2025, review_traffico_con_codici, li2023graphneuralnetworkspatiotemporal, survey_Chikwendu, feng2024comprehensivesurveydynamicgraph, review_Alippi, Khemani2024, survey_kumar_spatiotemporal_data, survey_spatiotemporal, CAPONE2025130400}.

The paper is organized as follows. After this Introduction, Sec.~\ref{section:Methodology for the Systematic Literature Review} illustrates the methodology used to collect the papers. Sec.~\ref{section:Overview of the publications} provides a general overview of the selected papers. Sec.~\ref{section:Graph neural networks} introduces some fundamental definitions and notions at the basis of spatio-temporal GNN models, and discusses their taxonomy and the determination of the graph structure. Sec.~\ref{section:Thematic analysis} is the core of this review. It explores the diverse domains of application of the spatio-temporal GNN models proposed in the selected papers for the identified domains. A discussion of the findings and the answers to the research questions mentioned above are provided in Sec.~\ref{section:Discussion}. Finally, limits, challenges and future research directions are presented in Sec.~\ref{section:Limits, challenges, and future research directions}. Sec.~\ref{section:Conclusions} concludes the review. Appendix \ref{a:appendix_papers} provides a list of all papers included in this SLR, together with the year of publication, group they belong to, case study, and nature of the task (e.g., classification or forecasting). All links mentioned in the papers are included in this review. For the reader's convenience, links that no longer work at the time of submission are marked with a $^\dagger$. The overall structure of the paper is visually summarized in Fig.~\ref{f:paper_structure}.
\begin{figure}[ht]
    \centering
    \includegraphics[width=0.9\linewidth]{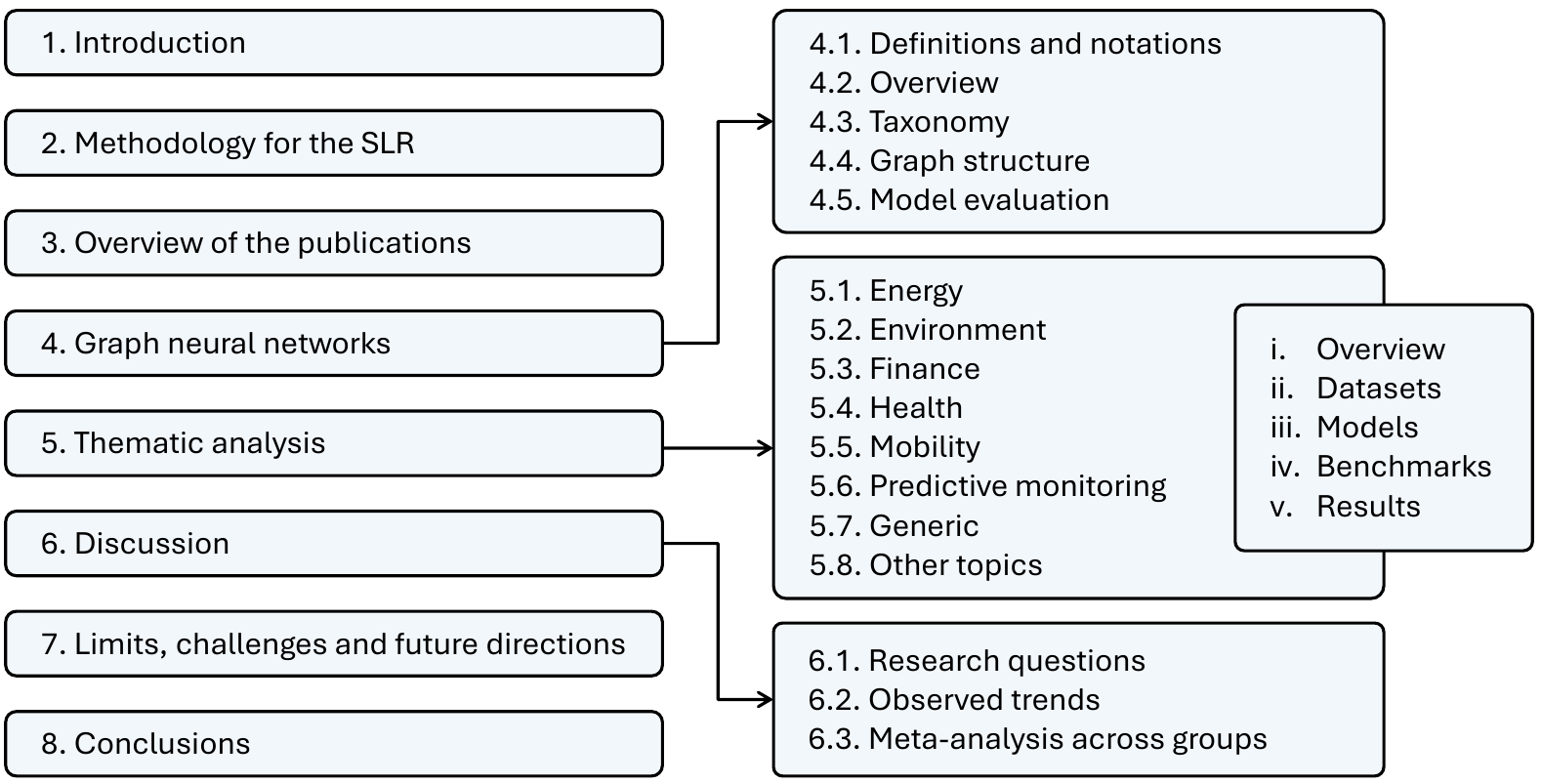}
    \caption{Overview of the structure of the SLR.}
    \label{f:paper_structure}
\end{figure}

To maximize the accessibility and usability of the collected information, the paper is supported by a GitHub repository, available at \url{https://github.com/FlaGer99/SLR-Spatio-Temporal-GNN.git}, which complements the review with interactive resources. Beyond listing the 366 reviewed papers organized by groups, the repository offers interactive tables for sorting and filtering models by any metric, dynamic plots to visualize performance and trends, and tools to explore top-performing models and most common benchmarks. These features make it easy for readers to analyze, compare, and gain deeper insights from the collected data. We strongly encourage exploring the repository as an interactive complement to this paper, especially when examining the extensive and complex tables comparing model results, as it allows readers to easily sort and visualize the data.

\section{Methodology for the systematic literature review}
\label{section:Methodology for the Systematic Literature Review}
In order to conduct this SLR, four databases of primary importance were consulted: Scopus, IEEE Xplore, Web of Science, and ACM. These databases were chosen for their comprehensive coverage of academic and scientific literature, ensuring a broad and inclusive search across multiple disciplines.

Prior to commencing the literature search, some exclusion criteria were defined. First, papers from the current year 2025 were excluded because it has only just begun, hence the review will only cover the literature up to the end of year 2024. Second, only journal articles published in Q1 and Q2 journals, as ranked by the Scimago Journal Rank, and conference papers from A$^\star$ and A venues, according to the ICORE conference rankings, were included. This selection was made to ensure that the analysis focused exclusively on high-quality, peer-reviewed research published in top-tier venues. Additionally, book chapters, workshops, and tutorials were excluded.

An advanced search query to be submitted to the databases was designed to capture the breadth of the topic while maintaining specificity. Its general form is
\texttt{(("graph neural network" OR gnn) AND "time series") AND (classification OR forecasting)} \footnote{Specific search queries for the different databases:
\begin{itemize}
    \item Scopus: \texttt{TITLE-ABS-KEY((("graph neural network" OR gnn) AND "time series") AND (classification OR forecasting)) AND PUBYEAR < 2025}.
    \item IEEE Xplore: \texttt{(("graph neural network" OR gnn) AND "time series") AND (classification OR forecasting)}. Filters applied: \texttt{2000 - 2024}.
    \item Web of Science: \texttt{(("graph neural network" OR gnn) AND "time series") AND (classification OR forecasting)}. Filters applied: \texttt{2000 - 2024}.
    \item ACM: \texttt{[[All: "graph neural network"] OR [All: gnn]] AND [All: "time series"] AND [[All: classification] OR [All: forecasting]] AND [E-Publication Date: (01/01/2000 TO 12/31/2024)]}.
\end{itemize}}.
The first two elements of the query aim to restrict the pool of GNN models to the time series domain, and the second part focuses the search on classification and forecasting applications.

Overall, the selection process of the papers followed the multi-stage approach described below:
\begin{enumerate}
    \item \textbf{Database search}: the search query was submitted to each of the four databases, and all records were imported into a csv file.
    \item \textbf{Duplicate removal and venue filtering}: duplicate records were removed, along with publications from journals outside Q1–Q2 and conferences not ranked A$^\star$ or A.
    \item \textbf{Title and abstract screening}: the remaining records were screened by reading their titles and abstracts. Studies that at first sight did not meet the criteria were discarded.
    \item \textbf{Full-text screening}: the full texts of the remaining papers were analyzed one by one, which led to a further assessment of  eligibility.
    \item \textbf{Data extraction}: relevant data and features were extracted from the included studies, and analyzed in order to better discuss the approach and present the most common datasets and models for each identified field.
\end{enumerate}

A graphical representation of this selection process is provided in Fig.~\ref{f:selection_flowchart}. A total of 2663 records were identified, 766 from Scopus, 258 from IEEE Xplore, 292 from Web of Science, and 1347 from ACM. After that, 1618 duplicates and publications from non top-tier venues were removed. Then, the titles and abstracts of the remaining 1045 records were manually evaluated to select the pertinent papers. Among these, 667 were discarded because they were considered outside the scope of this review, and 12 were discarded because they were not in English. As a result, 366 records were included in the review, comprising 262 journal papers and 104 conference papers. The complete list of selected paper can be found in Appendix \ref{a:appendix_papers} and in the accompanying GitHub repository. Next, Sec.~\ref{section:Overview of the publications} provides an overview of the selected publications.
\begin{figure}[ht]
    \centering
    \includegraphics[width=0.8\linewidth]{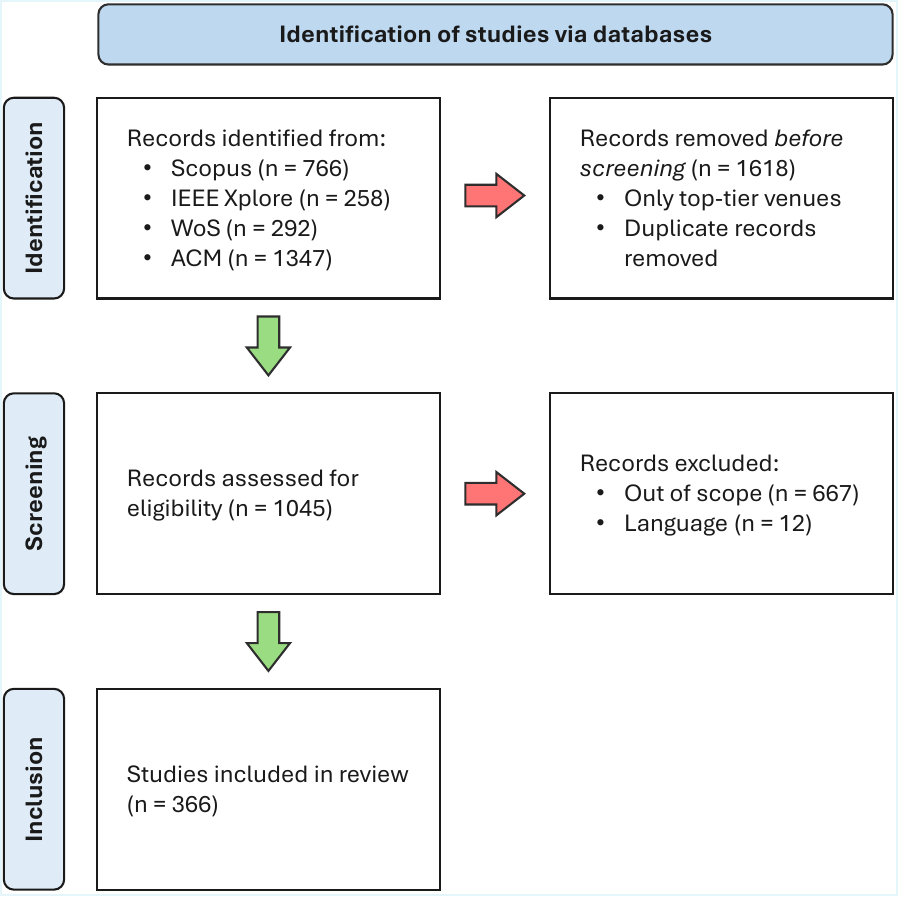}
    \caption{PRISMA flowchart summarizing the identification and selection process, adapted from \cite{PRISMA_statement}. The flowchart is structured vertically into three key phases: identification (collection of records from databases, and removal of duplicates and non-top-tier venues), screening (assessment for eligibility and exclusion of out-of-scope studies), and inclusion (inclusion of papers in the review). The boxes on the right indicate the number of studies excluded at each stage.}
    \label{f:selection_flowchart}
\end{figure}

\section{Overview of the publications}
\label{section:Overview of the publications}
This section provides a comprehensive overview of the 366 selected papers. After their selection, the papers were grouped according to their domain of application, as better explained in the thematic analysis of Sec.~\ref{section:Thematic analysis}. In the following, some overall bibliographic information about the reviewed articles is presented.

Fig.~\ref{f:grafico_anno_ambito} presents a stacked column chart showing the number of journal and conference publications over time, broken down by research group. The height of each column represents the total number of publications per year, while the colored segments represent contributions from each group. A clear upward trend can be observed over time, especially in journal papers, reflecting an increasing interest in spatio-temporal GNN models across various fields, since 2020. The groups that experienced the most significant increase are \enquote{Environment}, \enquote{Generic}, \enquote{Mobility}, and \enquote{Predictive monitoring}, among which \enquote{Generic} and \enquote{Mobility} stand out in terms of both number of publications and the novelty of their contributions. For more details, see also the group numerosity reported in Tab.~\ref{t:tabella pivot}.
\begin{figure}[ht]
    \centering
    \includegraphics[width=0.49\linewidth]{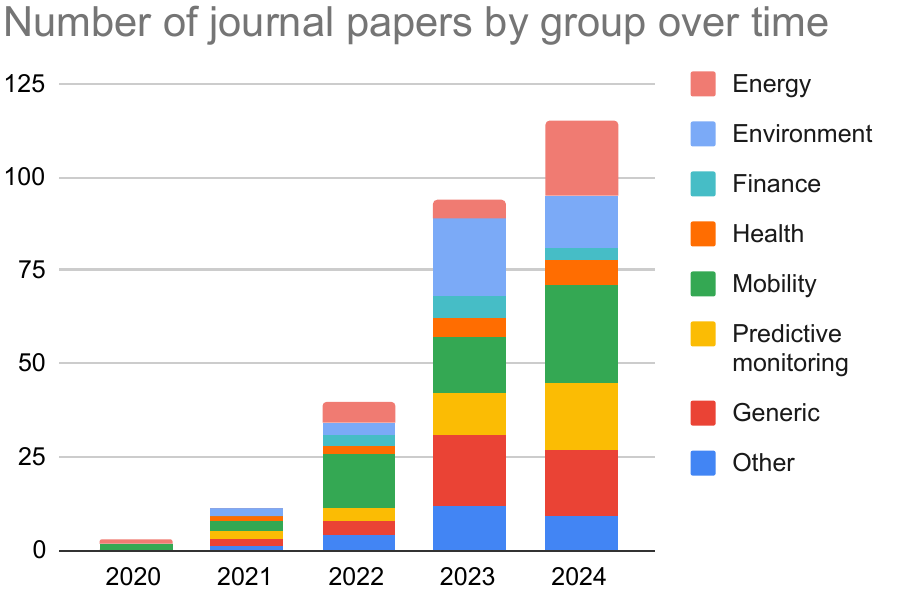}
    \includegraphics[width=0.49\linewidth]{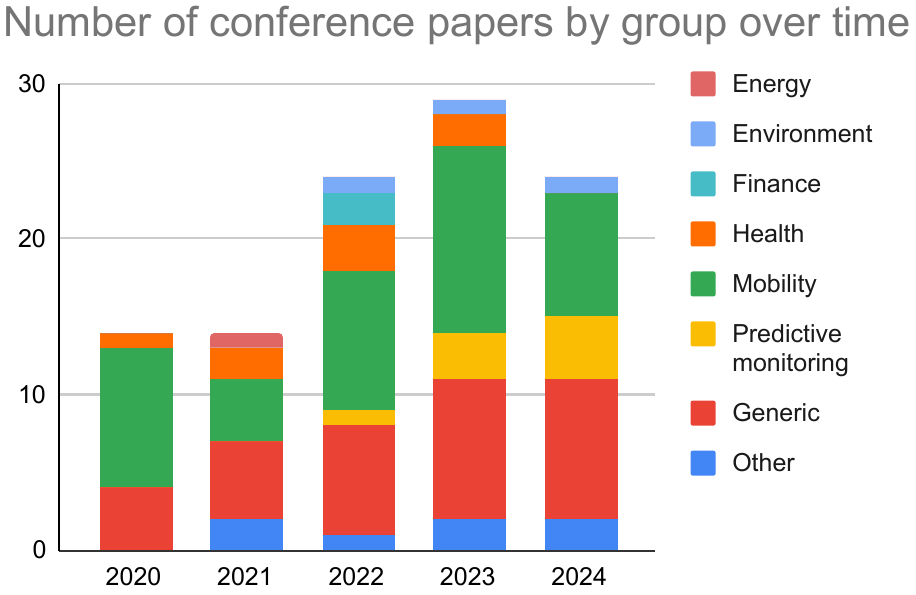}
    \caption{Number of journal and conference papers over time across the different groups.}
    \label{f:grafico_anno_ambito}
\end{figure}
\begin{table}[ht]
    \caption{Pivot table illustrating the number of publications by year and group (journal and conference papers combined).} \label{t:tabella pivot}
    \centering
    \begin{tabular}{lcccccc}
        \toprule
        Field / Year & 2020 & 2021 & 2022 & 2023 & 2024 & \textbf{Total} \\
        \midrule
        Energy & 1 & 1 & 6 & 5 & 20 & \textbf{33} \\
        Environment & 0 & 2 & 4 & 22 & 15 & \textbf{43} \\
        Finance & 0 & 0 & 5 & 6 & 3 & \textbf{14} \\
        Health & 1 & 3 & 5 & 7 & 7 & \textbf{23} \\
        Mobility & 11 & 6 & 24 & 27 & 34 & \textbf{102} \\
        Predictive monitoring & 0 & 2 & 4 & 12 & 22 & \textbf{40} \\
        Generic & 4 & 7 & 11 & 28 & 27 & \textbf{77} \\
        Other & 0 & 4 & 5 & 14 & 11 & \textbf{34} \\
        \textbf{Total} & \textbf{17} & \textbf{25} & \textbf{64} & \textbf{121} & \textbf{139} & \textbf{366} \\
        \bottomrule
    \end{tabular}
\end{table}

The catalog of the publication sources indicates that, in the studies period, 131 different Q1 and Q2 journals have published articles on GNN models for time series forecasting or classification. Tab.~\ref{t:tabella giornali} lists the journals with more than 5 published papers.
\begin{table}[ht]
    \centering
    \caption{Journals with more than 5 published paper.} \label{t:tabella giornali}
    \begin{tabular}{lccccc}
        \toprule
        Journal & Papers \\
        \midrule
        \giornale{IEEE Access} & 14 \\
        \giornale{ACM Transactions on Knowledge Discovery from Data} & 9 \\
        \giornale{Knowledge-Based Systems} & 9 \\
        \giornale{Applied Intelligence} & 8 \\
        \giornale{IEEE Transactions on Instrumentation and Measurement} & 8 \\
        \giornale{ACM Transactions on Intelligent Systems and Technology} & 7 \\
        \giornale{Expert Systems with Applications} & 7 \\
        \giornale{IEEE Transactions on Knowledge and Data Engineering} & 7 \\
        \giornale{Applied Energy} & 6 \\
        \giornale{IEEE Transactions on Industrial Informatics} & 6 \\
        \giornale{IEEE Transactions on Intelligent Transportation Systems} & 6 \\
        \giornale{Sensors} & 6 \\
        \bottomrule
    \end{tabular}
\end{table}
The diversity of the journals indicates a vast range of potential applications for spatio-temporal GNN models in different fields. This suggests that researchers often select journals based on the specific domain of application. This allows them to reach targeted audiences, and contribute within their respective fields. Such diversity also underlines the interdisciplinary nature of GNNs research, clearly spanning multiple domains. 
As for ranked A* and A conference venues, 18 different venues have hosted publications on GNN models for time series forecasting or classification. The conference venues with 5 or more papers are listed in Tab.~\ref{t:tabella conferenze}.
{\tabcolsep=1pt
\begin{table}[ht]
    \centering
    \caption{Conference venues with more than 5 published paper.} \label{t:tabella conferenze}
    \begin{tabular}{lccccc}
        \toprule
        Conference venue & Papers \\
        \midrule
        \conferenza{ACM Int.~Conf.~on Information and Knowledge Management (CIKM)} & 30 \\
        \conferenza{ACM Int.~Conf.~on Knowledge Discovery and Data Mining (KDD)} & 18 \\
        \conferenza{ACM Int.~Conf.~on Advances in Geographic Information Systems (SIGSPATIAL)} & 11 \\
        \conferenza{Advances in Neural Information Processing Systems (NeurIPS)} & 7 \\
        \conferenza{Int.~Conf.~on Machine Learning (ICML)} & 5 \\
        \bottomrule
    \end{tabular}
\end{table}
}
\newpage
Fig.~\ref{f:mappa_provenienza} shows a pie chart of how corresponding authors are distributed among different countries. This visualization provides an overview of the global distribution of research activities. Nearly 70\% of the authors are affiliated with institutions based in China.
\begin{figure}[ht]
    \centering
    \includegraphics[width=0.5\linewidth]{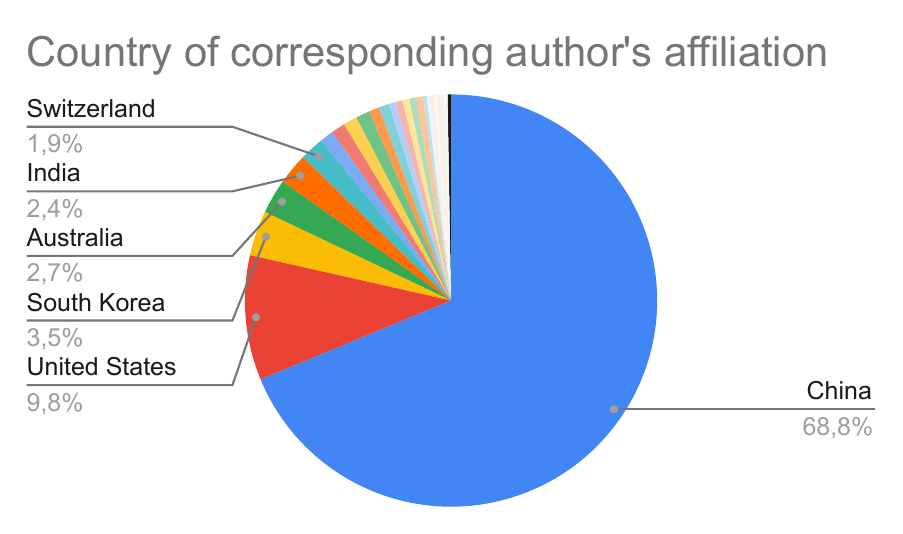}
    \caption{Pie chart of the distribution of corresponding authors in different countries.} 
    \label{f:mappa_provenienza}
\end{figure}

In order to examine the degree of collaboration among researchers, a co-authorship network was generated by using the VOSviewer software \cite{VOSviewer}. In this network, authors are represented as nodes, and co-authorships as connections, revealing clusters of researchers who frequently collaborate. Fig.~\ref{f:grafico_gruppi} shows the complete network, while Fig.~\ref{f:grafico_gruppi_B} provides an enlarged view of two selected clusters.
\begin{figure}[ht]
    \centering
    \begin{minipage}{\textwidth}
        \centering
        \includegraphics[width=0.9\textwidth]{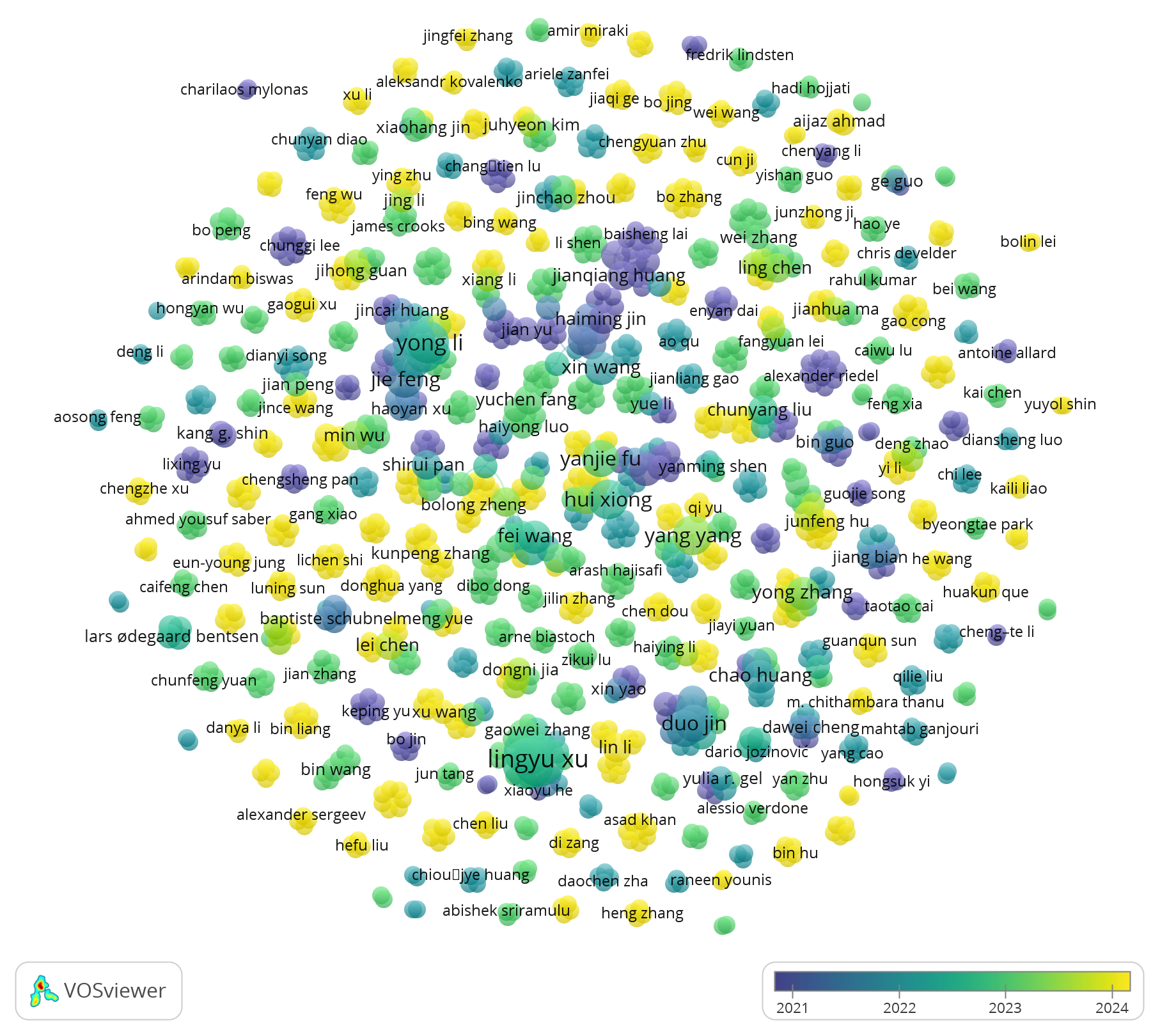}
        \subcaption{Map of the complete collaboration network.\\ ~ \\}
        \label{f:grafico_gruppi_A}
    \end{minipage}
    \begin{minipage}{\textwidth}
        \centering
        \includegraphics[height=6cm]{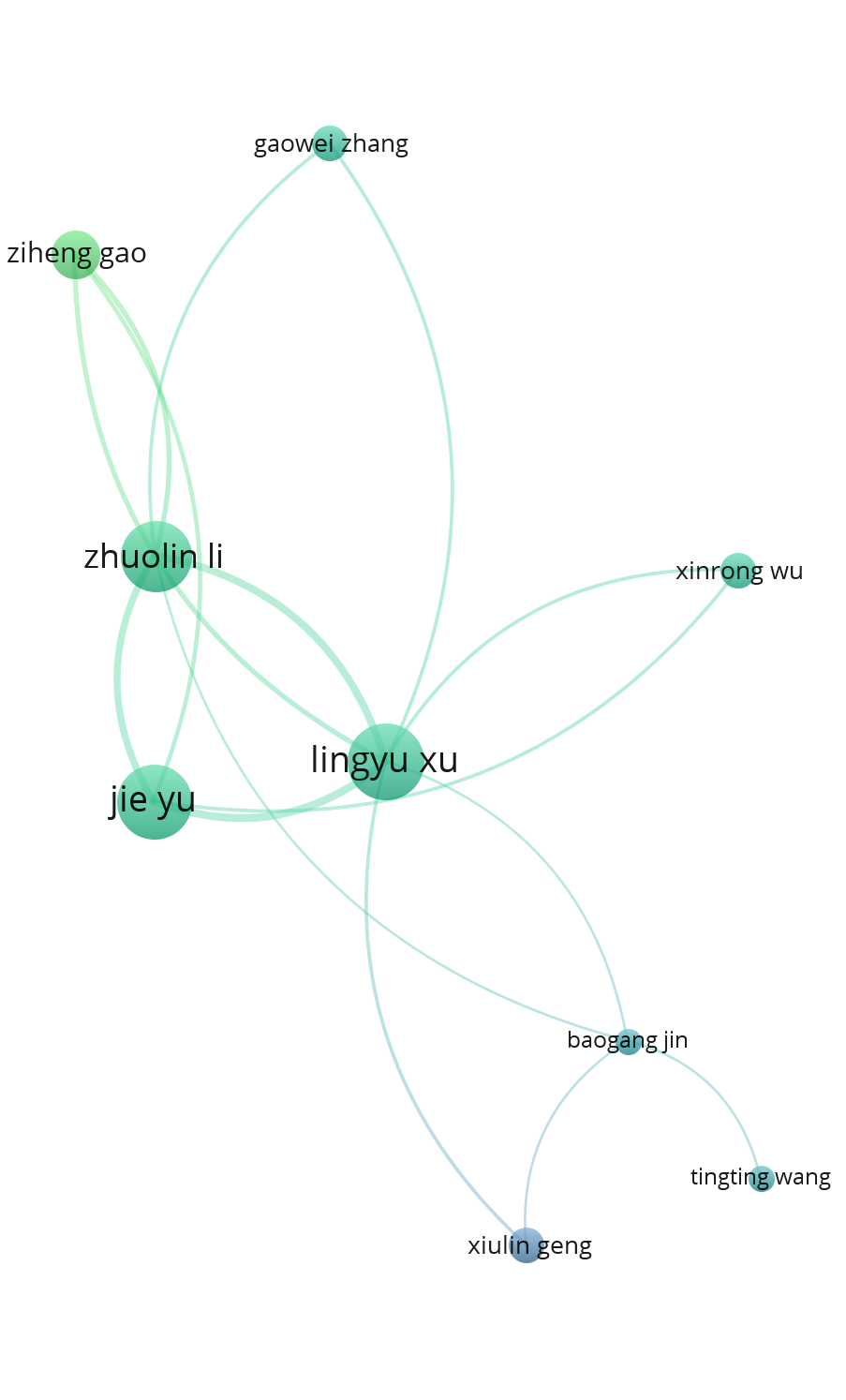}
        \hspace{2cm}
        \includegraphics[height=6cm]{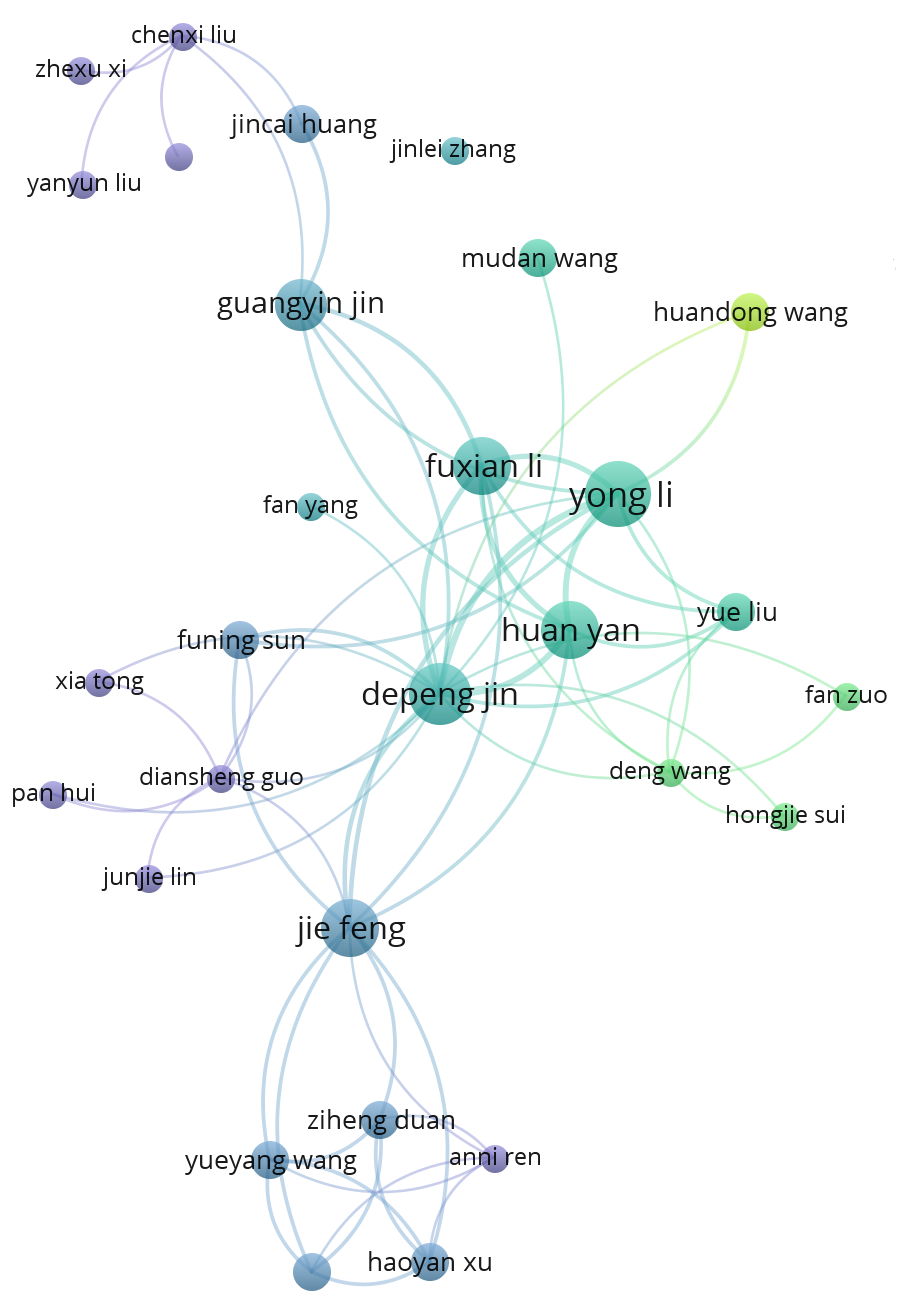}
        \subcaption{Enlargement on two different clusters.}
        \label{f:grafico_gruppi_B}
    \end{minipage}

    \caption{Network of collaboration between researchers, created using VOSviewer software.}
    \label{f:grafico_gruppi}
\end{figure}
In the figures, the color of each node corresponds to the author's average publication year, allowing the temporal evolution of research groups to be visualized. From the complete network in Fig.~\ref{f:grafico_gruppi}, two main insights emerge. First, the large number of clusters indicates a highly fragmented research community, with many small, relatively independent groups. Second, the color of the nodes indicates that some groups were active primarily in earlier years and have since faded, while others have emerged more recently, as indicated by the yellow color. Additionally, several authors appear in multiple clusters, reflecting collaborations that span different groups over time, as also shown in Fig.~\ref{f:grafico_gruppi_B}.

Additional figures and analyses of trends in the collected papers are available on our GitHub repository.

\clearpage

\section{Graph neural networks}
\label{section:Graph neural networks}
This section introduces some fundamental definitions and notions at the basis of the study of time series with spatio-temporal GNNs.

\subsection{Definitions and notations}
\begin{definition}[Time series]
A time series is a sequence of data points indexed by timestamps, that here are assumed to be equispaced. An equispaced univariate time series of length $T$ is a sequence of scalar observations collected over time, denoted as $\vec{x}_t 
\in \mathbb{R}^T$. An equispaced multivariate time series of length $T$ is a sequence of $D$-dimensional vector observations collected over time, denoted as $\vec{x}_t \in \mathbb{R}^{D \times T}$.
\end{definition}

\begin{definition}[Graph]
A graph $\mathcal{G}$ is a pair of finite sets $\mathcal{G} = (\mathcal{V},\mathcal{E})$, where $\mathcal{V} = \{ v_1,\dots,v_n \}$ is a set of $n$ nodes (also called vertices) and $\mathcal{E} \subseteq \mathcal{V} \times \mathcal{V}$ is the set of connecting edges. In an undirected graph each edge is an unordered pair of nodes $\{v_i,v_j\}$, while in a directed graph the edges have an orientation, and correspond to ordered pairs $(v_i,v_j)$. When an edge $(v_i,v_j)$ or $\{v_i,v_j\}$ exists, the nodes $v_i$ and $v_j$ are called adjacent.
\end{definition}

\begin{definition}[Spatio-temporal graph]
A spatio-temporal graph is a 4-tuple $\mathcal{G} = (\mathcal{V}_t,\mathcal{E}_t,\mathcal{X}_t,\mathcal{T})$, where $\mathcal{T} = \{t_1,\dots,t_T\}$ is a set of $T$ timestamps $t$, $\mathcal{V}_t = \{v_1(t),\dots,v_{n_t}(t)\}$ is a set of $n_t$ nodes representing spatial entities at time $t$, $\mathcal{E}_t = \{e_1(t),\dots,e_{m_t}(t)\}$ is a set of $m_t$ edges representing the relationships between nodes at time $t$, $\mathcal{X}_t = \{x_{v_1}(t),\dots,x_{v_{n_t}}(t)\} \cup \{r_{e_1}(t),\dots,r_{e_{m_t}}(t)\}$ is a set of attributes associated with nodes and edges ($x_{v_i}$ and $r_{e_i}$ are respectively the attributes of node $v_i$ and edge $e_i$, and can be either scalars of vectors) at time $t$. Notice that the sets of nodes and edges can change over time, as well as the set of attributes $\mathcal{X}_t$.
\end{definition}

Hereafter, the time index will be dropped to indicate graphs not necessarily changing in time, i.e., not necessarily spatio-temporal.

\begin{definition}[Adjacency matrix]
The adjacency matrix $\mathcal{A}$ of a graph $\mathcal{G}$ with $|\mathcal{V}| = n$ nodes is a $n \times n$ square matrix, with $\mathcal{A}_{ij}$ specifying the number of connections from node $v_i$ to node $v_j$ for $i,j = 1,\dots,n$. Here it is assumed that a graph cannot have more than one edge between any two nodes, so $\mathcal{A}_{ij}$ can only be 0 or 1 depending on whether there is a connection or not.
\end{definition}

\begin{definition}[Degree matrix]
The degree matrix $\mathcal{D}$ of a graph with $|\mathcal{V}| = n$ nodes is a diagonal matrix whose entries are given by the degree of each node, i.e., the number of edges attached to the node. In formulas, its diagonal elements are given by
\begin{equation}
    \mathcal{D}_{ii} = \sum_j \mathcal{A}_{ij} \:.
\end{equation}

\end{definition}

\subsection{Overview of graph neural networks}
\label{subsection:Overview of graph neural networks}
Given a graph $\mathcal{G}=(\mathcal{V},\mathcal{E})$, a GNN model (not necessarily spatio-temporal) aims to generate an embedding (i.e., a vector of real entries) for each node in the graph. An overall graph embedding consists of representing a graph in a space of chosen dimension, while preserving its structural information.

The notion of GNN was initially proposed by Gori et al.~in 2005 in Ref.~\cite{gori_aggiunto} and by Scarselli et al.~in 2009 in Ref.~\cite{Scarselli2009} as an architecture designed to implement a function that maps a graph $\mathcal{G}$ with $n$ nodes into an $m$-dimensional Euclidean space. Noticeably, this definition was intended in general for graph data, not specifically time-dependent data, which is the focus of this SLR. In Ref.~\cite{Scarselli2009} the target node's embedding was learned in two steps. First, neighbor information is iteratively propagated with fixed weights within the graph until a stable fixed point is reached (whose existence is guaranteed under certain assumptions by Banach's fixed point theorem). Second, errors on targets are propagated back across the graph. Notice that in this setup, propagation is performed in \enquote{internal time}, not in \enquote{clock time}. More in general, the calculation of node embeddings is quite complex, and over time many other methods have been developed. These methods are based on different architectures and sophisticated aggregation mechanisms. Based on these aggregating mechanisms, a taxonomy of GNN models with different classes has been defined. In this review, we adopt a  taxonomy for the discussed spatio-temporal GNNs similar to the one proposed by Chen et al.~in Ref.~\cite{survey_Lucheroni}. More details about that will be given in Subsec.~\ref{subsection:Taxonomy of graph neural networks}.

\subsection{Spatio-temporal GNNs and their taxonomy}
\label{subsection:Taxonomy of graph neural networks}
A large part of the research on spatio-temporal GNNs is focused on multivariate time series, as they can be naturally abstracted into spatio-temporal graphs, as depicted in Fig.~\ref{f:spatiotemporal graph}. In this context, nodes usually correspond to different variables at a given time step, and edges represent relationships between variables.
\begin{figure}[ht]
    \centering
    \includegraphics[width=0.3\textwidth]{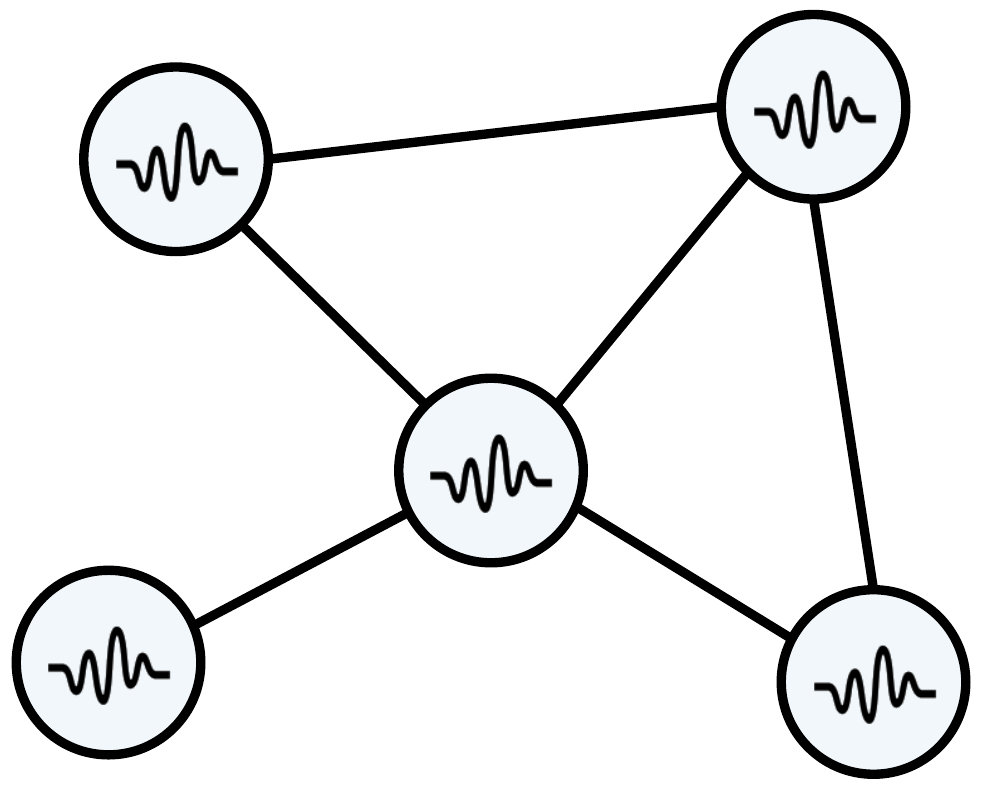}
    \caption{Representation of a spatio-temporal graph. Here, each node has a dynamic feature, given by the points of the time series.}
    \label{f:spatiotemporal graph}
\end{figure}
This specific form of spatio-temporal modeling assumes that the feature information of a node depends on both its own historical values and on the historical data from its neighboring nodes \cite{survey_spatiotemporal}. Notice that once the the graph structure is given, information is propagated (in \enquote{internal time}) within it until the output is obtained, as better explained in the following. Our discussion will build up from this perspective.

When it comes to the development of a spatio-temporal GNN model, there are two primary approaches: one that handles
spatial and temporal substructures in separate blocks, and another that integrates and processes these two substructures at once \cite{survey_spatiotemporal}. These approaches are implemented by means of modules that are purely spatial, purely temporal, or hybrid combinations of the two. These modules can then be organized into a series of blocks, thus creating the sought spatio-temporal GNN model.

In the taxonomy of spatio-temporal GNNs used in this review, the focus is exclusively on the description of the spatial module that propagates the information within the nodes at fixed time $t$ (not explicitly included in the following equations for simplicity), which, in the reviewed literature, is a fundamental aspect that differentiates one GNN model from another. Based on the mechanism at the basis of the information propagation in the graph at a given time step, spatio-temporal GNNs can be further classified as recurrent GNNs, convolutional GNNs, and attentional GNNs, as represented in Fig.~\ref{f:taxonomy} and discussed in the following. These classes are not mutually exclusive, as some models may fit into hybrid categories. In addition, these classes are not fully exhaustive of the variety of available models, as there may be models that do not align with any of the existing categories. However, for the purpose of this review, the taxonomy discussed above will be considered sufficient.
\begin{figure}[ht]
    \centering
    \includegraphics[width=\textwidth]{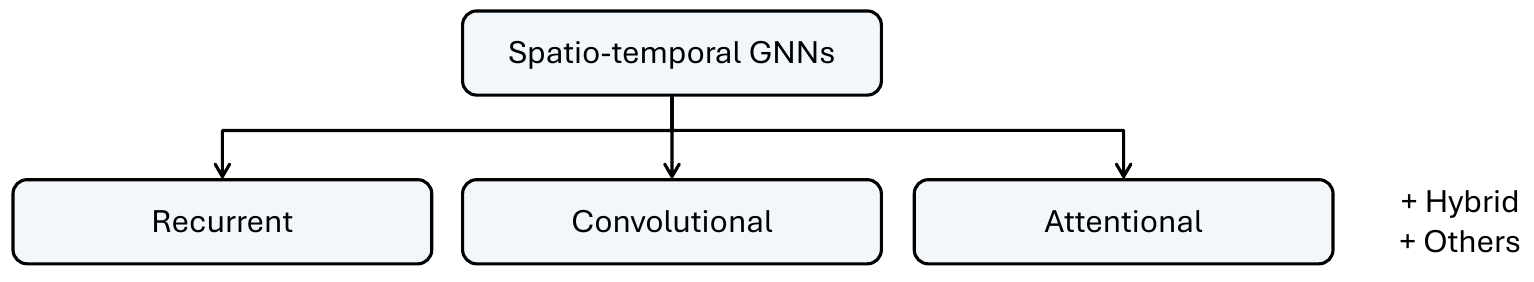}
    \caption{Taxonomy of the spatial component of spatio-temporal GNNs, categorizing models into recurrent, convolutional, and attentional.}
    \label{f:taxonomy}
\end{figure}

\textbf{Recurrent GNNs.} Recurrent GNN modeling mainly proceeds from the results of early studies on GNNs, where the node embeddings are generated by iteratively propagating neighbor information until a stable fixed point is reached. The iterative equation for the calculation of the embedding (or hidden state) of node $u$ has the form
\begin{equation}
    h_u^{(k)} = f\left(x_u, \{x_v, r_{(u,v)}, h_v^{(k-1)} \mid v \in \mathcal{N}(u) \}\right) \:,
    \label{eq:ricorrenza_RecGNN}
\end{equation}
where $k$ is the index related to iteration (hence running on \enquote{internal time}), $f$ is a parametric trainable function, $x_u$ is the attribute associated to node $u$, $r_{(u,v)}$ is the weight of the edge $(u,v)$ between the nodes $u$ and $v$, $\mathcal{N}(v)$ denotes the neighborhood of node $v$, and the initial hidden states $h^{(0)}$ are usually randomly initialized. As detailed in the seminal Ref.~\cite{Scarselli2009}, which introduced recurrent GNNs, in order to ensure the convergence of Eq.~\eqref{eq:ricorrenza_RecGNN} it is necessary that the iterative function $f$ is a contraction mapping.

\textbf{Convolutional GNNs.} Convolutional GNNs are closely related to recurrent GNNs. They were introduced by Kipf and Welling in 2017 in Ref.~\cite{graph_convolution_kipf}. Unlike recurrent GNNs, which apply the same iterative contraction mapping $f$ until an equilibrium is reached, convolutional GNNs can use different parameters at each updating step, by stacking multiple graph convolutional layers to extract the node embeddings \cite{cit_rev3}. Noticeably, as the number of convolutional layers increases, the convolutional perceptual field expands, and a more abstract representation of information is thus obtained \cite{art_461}.

A basic form for the equation for the $k^{\text{th}}$ layer of a convolutional GNN with $n$ nodes is
\begin{equation}
    \begin{gathered}
        H^{(k)} = f_k\left( \tilde{\mathcal{A}} H^{(k-1)} \Theta_{k-1} \right) \:, \quad
        \tilde{\mathcal{A}} = \hat{\mathcal{D}}^{-\frac{1}{2}} \hat{\mathcal{A}} \hat{\mathcal{D}}^{-\frac{1}{2}} \:, \quad
        \hat{\mathcal{A}} = \mathcal{A} + I \:, \quad \\
        \hat{\mathcal{D}}_{ii} = \sum_j \hat{\mathcal{A}}_{ij} \:, \quad
        \quad H^{(0)} = X \:,
        \label{eq:ConvGNN}
    \end{gathered}
\end{equation}
where $f_k$ represents the convolutional operation performed at layer $k \in \{1,\dots,K\}$ (with $K$ typically kept small to prevent issues like over-smoothing \cite{oversmoothing}), $\mathcal{A} \in \mathbb{R}^{n \times n}$ is the adjacency matrix, $I \in \mathbb{R}^{n \times n}$ is the identity matrix (used to add self-loops in the adjacency matrix $\mathcal{A}$), $\hat{\mathcal{D}} \in \mathbb{R}^{n \times n}$ is the diagonal degree matrix associated to $\hat{\mathcal{A}}$, $H^{(k)} \in \mathbb{R}^{n \times d_k}$ are the $d_k$-dimensional node embeddings $h^{(k)}_v$ of the entire graph produced by the $k^{\text{th}}$ layer, $X \in \mathbb{R}^{n \times d_0}$ is the collection of all node attributes $x_v$ of the entire graph, and $\Theta_k \in \mathbb{R}^{d_{k-1} \times d_{k}}$ is a trainable weight matrix for the $k^{\text{th}}$ layer \cite{survey_Lucheroni}. The mechanism of node embedding generation from Eq.~\eqref{eq:ConvGNN} is illustrated in Fig.~\ref{fig:convgnn}, where it is put in contrast with the mechanism of a recurrent GNNs, shown in Fig.~\ref{fig:recgnn}.
\begin{figure}[htbp]
    \centering
    \begin{subfigure}[t]{0.45\textwidth}
        \centering
        \includegraphics[height=9cm]{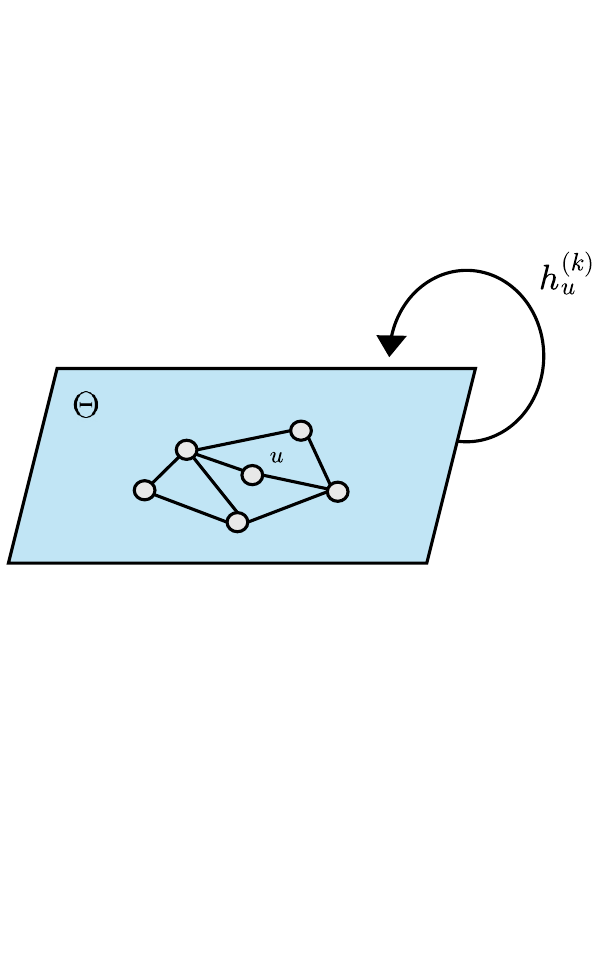}
        \caption{Recurrent GNNs use the same graph recurrent layer in the iterative generation of the node embeddings. Here $\Theta$ represents the weights of the parametric trainable function $f$.}
        \label{fig:recgnn}
    \end{subfigure}
    \hfill
    \begin{subfigure}[t]{0.45\textwidth}
        \centering
        \includegraphics[width=\textwidth]{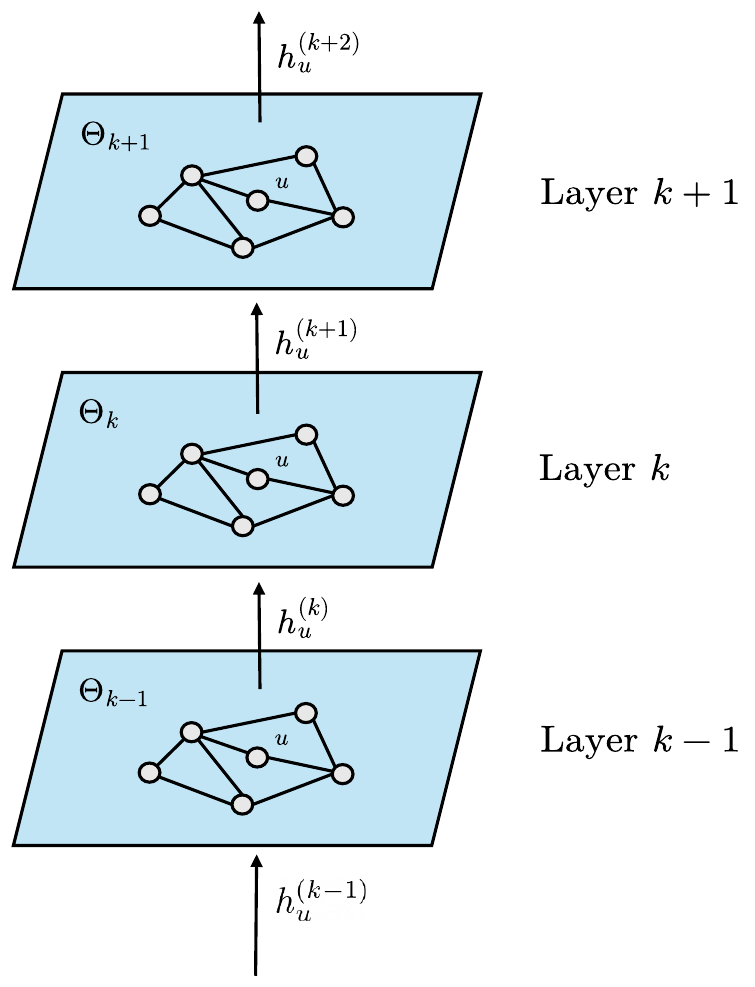}
        \caption{Convolutional GNNs use different graph convolutional layers in the generation of the node embeddings.}
        \label{fig:convgnn}
    \end{subfigure}
    \caption{Comparison of the node embedding generation mechanisms in a recurrent GNN (left panel) and a convolutional GNN (right panel).}
    \label{fig:recgnn vs convgnn}
\end{figure}

\textbf{Attentional GNNs.} In attentional GNNs, the aggregation process uses the attention mechanism \cite{attention} to combine node features. The equations to compute the node embedding $h_u$ of layer $k+1$ from the embeddings of layer $k$ for each node $u$ are
\begin{subequations}
    \begin{align}
        z_u^{(k)} & = W_{k} h_u^{(k)} \:, \label{eq:att1} \\
        \epsilon_{uv}^{(k)} & = \text{LeakyReLU} \left(\vec a^{\,(k)^\top} \left(z_u^{(k)} \, \vert\vert \, z_v^{(k)}\right)\right) \:, \label{eq:att2} \\
        \alpha_{uv}^{(k)} & = \frac{\exp\left(\epsilon_{uv}^{(k)}\right)}{\displaystyle\sum\limits_{w\in \mathcal{N}(u)}^{}\exp\left(\epsilon_{uw}^{(k)}\right)} \:, \label{eq:att3} \\
        h_u^{(k+1)} & = f\left(\sum_{v\in \mathcal{N}(u)} {\alpha^{(k)}_{uv} z^{(k)}_v }\right) \:. \label{eq:att4} 
    \end{align}
\end{subequations}
Eq.~\eqref{eq:att1} represents a learnable linear transformation of the node embedding $h_u^{(k)}$ using the trainable weight matrix $W_k$ of layer $k$. Eq.~\eqref{eq:att2} computes a pairwise masked attention score between nodes $u$ and $v$. This is done by applying the concatenation operation $||$ to the vectors $z_u^{(k)}$ and $z_v^{(k)}$, then calculating a dot product of the concatenation with a learnable weight vector $\vec a^{\,(k)}$, and finally applying the LeakyReLU nonlinearity. The quantity $\epsilon_{uv}$ indicates the importance of the features of node $v$ to node $u$. The term \enquote{masked} refers to the fact that $\epsilon_{uv}$ are computed for neighboring nodes only. Eq.~\eqref{eq:att3} normalizes the attention score through the softmax function in order to make the $\alpha_{uv}$ comparable across different nodes. Finally, Eq.~\eqref{eq:att4} computes the embedding $h_u^{(k+1)}$ of the next layer by aggregating the information in neighboring nodes, weighted by the attention scores \cite{GAT}. Graph attention networks can in addition use a multi-head attention mechanism to stabilize the learning process of self-attention. In practice, the node updating operation in graph attention networks is a generalization of the traditional averaging or max-pooling of neighboring nodes, which allows each node to compute a weighted average of its neighbors and identify the most relevant ones \cite{Brody2021HowAA}.

As anticipated, the taxonomy discussed so far focuses only on the propagation mechanisms of the spatial component. However, spatio-temporal GNNs typically consist of a stack of spatial, temporal, or also hybrid modules, each one with distinct roles. The function of these modules is detailed hereafter.

The spatial module is responsible for propagating information between nodes, thereby enabling the analysis of cross-sectional inter-dependencies between different variables. As previously discussed in the description of the mechanisms underlying the assumed taxonomy, the spatial module aggregates information starting from immediate neighboring nodes and extends it outward, thereby capturing also the influence of distant nodes. This spatial operation cannot directly account for temporal information. In contrast, the temporal module, often based on architectures such as LSTM, GRU, or self-attention, focuses on the evolution of data over time, independently from cross-sectional node interactions. When these different modules are stacked together, spatial and temporal information end up to be used simultaneously, although they are processed in separate modules. This occurs because the computation of each node state is influenced by both spatial and temporal information. This combined approach allows the spatio-temporal GNN model to be a powerful architecture able to effectively capture at the same time both spatial relationships and temporal dynamics.

\subsection{Determination of the graph structure}
\label{subsection:Determination of the graph structure}
A crucial issue when dealing with spatio-temporal GNNs for time series tasks is the determination of the graph structure, i.e., the connectivity of the nodes. The type of spatial information captured by the graph can vary widely across domains. For instance, in traffic networks, nodes often represent sensors or intersections, and edges correspond to actual roads or physical connections. In contrast, in domains such as healthcare or finance, nodes may represent patients, brain regions, or financial instruments, and edges encode correlations, interactions, or other relational information. This variability across datasets highlights the importance of carefully designing the graph structure to reflect the spatial relationships relevant to each domain. The specific choices made in different application areas are discussed in the corresponding subsections of this review.

Namely, some time series datasets come by their own nature with a pre-defined graph structure (e.g., a road network for traffic datasets), whereas some others do not. When a natural graph structure is available, it helps the model capture the underlying dynamics more effectively. When it is not directly available, it must be defined by the user in some way (based on domain knowledge or some metrics), or determined by some algorithm (such as visibility graphs \cite{visibility_Gori, visibility_graph, art_14}), or learned by the model itself.

Once the graph structure is determined, it is necessary to define the weighted graph adjacency matrix, which is a generalization of the graph adjacency matrix $\mathcal{A}$ whose entries are given by the edge weights of the graph. The edge weights can be defined \textit{a priori} by the user, or again, learned by the model itself based on a pre-defined architecture. In the first case, the edge weights are assigned based on some pre-defined metrics or criteria chosen by the user, for example by the spatial distance between the sites, or by similarity measures. In the latter case, the weights are continuously adjusted throughout training as the model keeps learning from the data.

In the following, we present a concise overview of the most common techniques for defining graph structures.

Pre-defined graphs are typically generated based on prior knowledge and specific metrics. These methods for the construction of the graphs typically rely on distance or similarity information. As for location-based techniques, they utilize geographical coordinates (latitude and the longitude) of the nodes to calculate the distance between them (e.g., using haversine or geodesic distance). This distance information is then used to determine the edge weights with techniques such as the thresholded Gaussian kernel (which applies a negative exponential function to the distance), weighted $k$-NN, or reversed MinMax. As for similarity-based techniques instead, they rely on similarity (or dissimilarity) of signal data. These techniques include methods such as correlation, cosine similarity, dynamic time warping (DTW), maximum information coefficient (MIM), or transfer entropy. In general, no single approach stands out as a clear winner, and performance should be evaluated on each dataset. For references and a more detailed explanation of these and other techniques for constructing pre-defined structures, see Ref.~\cite{art_160}.

As for automatic graph structure learning techniques, a distinction can be made between models that rely on pre-defined graph structures (sometimes referred to as supplementary graph structure learning models), and those that learn the graph structure entirely from scratch. Two notable examples from Wu et al.~illustrate these approaches.

The first example is Graph WaveNet (GWN) \cite{GWN}, which constructs a self-adaptive adjacency matrix $A$ using two randomly initialized learnable node embedding matrices $E_1$ and $E_2$ as:
\begin{equation}
    A = \text{softmax}\left(\text{ReLU}\left(E_1 \, E_2^\top\right)\right) \:,
    \label{eq:A_GWN}
\end{equation}
where $E_1$ and $E_2$ correspond to the source and target node embeddings, respectively. Their multiplication represents the spatial dependency weights between source nodes and target nodes. In the original work, these learned hidden graph dependencies are combined with a pre-defined graph structure. The authors note that when such a structure is unavailable, the self-adaptive adjacency matrix alone can be used to model hidden spatial dependencies.

As a second example, which overcomes the reliance on a pre-defined graph, the same authors proposed the MTGNN model \cite{art_655}, which extracts uni-directional relationships from node embeddings as follows:
\begin{equation}
    \begin{aligned}
        & M_1 = \tanh(\eta \, E_1 \Phi_1) \\
        & M_2 = \tanh(\eta \, E_2 \Phi_2) \\
        & A = \text{ReLU}\left(\tanh\left(\eta\left(M_1M_2^\top - M_2M_1^\top\right)\right)\right) \\
        & \text{idx} = \text{argtopk}(A[i,:]) \quad \forall i = 1,\ldots,n \\
        & A[i,-\text{idx}] = 0 \:, \\
    \end{aligned}
    \label{eq:A_MTGNN}
\end{equation}
where $E_1$ and $E_2$ are randomly initialized learnable node embedding matrices, $\Phi_1$ and $\Phi_2$ are model parameters, $\eta$ is a hyperparameter controlling the saturation rate of the activation function, and $\text{argtopk}(\cdot)$ is a function that returns the index of the top-k largest values of a vector. Unlike GWN, which can be used either standalone or to augment a pre-defined graph, MTGNN learns the graph structure fully end-to-end. These two models are among the first to explore self-learned graph structures. Although other methods have since been proposed, GWN and MTGNN continue to serve as prototypical models for self-learned graph structures.

When the graph structure is learned, some authors introduce a graph regularization term to the training loss function to optimize the learned graph adjacency matrix. This regularization term is typically designed to control the sparsity of the matrix, and to promote graph proximity, encouraging larger entries when the series in the corresponding nodes are \enquote{close enough}. Some researchers argue that models which generate graphs without prior knowledge may overlook local conditions when constructing adjacency matrices \cite{art_397}. Therefore, it is important to conduct further research in this area to identify the most effective techniques for generating informative graphs. For a more detailed explanation of graph learning models, see Refs.~\cite{art_396, art_405}.

Another key aspect in the determination of the graph structure is the choice between using a static or dynamic adjacency matrix, i.e. whether to use the same matrix for all graphs or not. When working with graphs defined in advance by specific metrics, it is common to use either a static matrix or a dynamic matrix generated through a sliding window approach. In contrast, when graphs are learned by the model, dynamic adjacency matrices are typically employed. The use of a dynamic adjacency matrix has the advantage of capturing the dynamic nature of the problem \cite{art_447}. However, the learned dynamic graphs often show significant differences between adjacent time steps, whereas real-world graph structures typically evolve smoothly over time. Therefore, this approach may not accurately capture the stable evolution observed in the real world.

Consequently, some researchers are exploring an intermediate approach, assuming that adjacent graph structures follow a basic pattern. Instead of defining static or dynamic adjacency matrices, they fuse a static backbone graph with a dynamic adjacency matrix. The static network captures the common graph structure over time, while the dynamic network focuses on learning the changes over time. An example of this intermediate approach can be found in Ref.~\cite{art_415}, which also provides a comparison between this intermediate approach and the classical ones.

In general, there is no rule that can determine in advance the best approach to follow. In fact, it depends on both the application domain and the specific dataset. For this reason, we do not explore this aspect further. As an example, for a detailed discussion of graph construction techniques in traffic flow prediction models, the reader is referred to Ref.~\cite{Fan2025}. Each subsection of the thematic analysis in Sec.~\ref{section:Thematic analysis} discusses the most prevalent approach for each group of papers.

\subsection{Model evaluation}
Evaluating the performance of a time series forecasting or classification model is essential to assess its reliability. Moreover, experimental results often show that even when datasets involve the same forecasting task, the transferability of model architectures is limited \cite{art_408}. Therefore, a proper evaluation of the model is a crucial aspect. A first key aspect to consider is its accuracy. The accuracy of a model can be measured by many different metrics, which vary between classification and forecasting tasks. Each metric provides its own insight into different aspects of prediction quality, so multiple metrics are typically used for a more comprehensive evaluation. In addition to presenting a table of the error metrics computed for the entire graph, some authors provide a spatial visualization of these errors across the various nodes, as for example in Ref.~\cite{art_28}.

As for forecasting problems, commonly used evaluation metrics include the mean absolute error (MAE), mean squared error (MSE), root mean squared error (RMSE), mean absolute percentage error (MAPE), root relative squared error (RRSE), the coefficient of determination (R$^2$), and the correlation coefficient (CORR). Let $N$ denote the number of observations, $y_i$ the true value associated with observation $i$, $\hat{y}_i$ the corresponding output of the model, and the bar denote the average. The formulas for these metrics (for a single-step forecasting horizon) are provided in Tab.~\ref{t:tabella metriche}.
\begin{table}[ht]
    \caption{Evaluation metrics for single-horizon forecasting tasks and their formulas.} \label{t:tabella metriche}
    \centering
    \begin{tabular}{ll}
        \toprule
        Metrics & Formula \\
        \midrule
        Mean absolute error (MAE) & $\frac{1}{N} \sum_{i=1}^{N} |y_i - \hat{y}_i|$ \\ \addlinespace[5pt]
        Mean squared error (MSE) & $\frac{1}{N} \sum_{i=1}^{N} (y_i - \hat{y}_i)^2$ \\ \addlinespace[5pt]
        Root mean squared error (RMSE) & $\sqrt{\frac{1}{N} \sum_{i=1}^{N} (y_i - \hat{y}_i)^2}$ \\ \addlinespace[5pt]
        Mean absolute percentage error (MAPE) & $\frac{1}{N} \sum_{i=1}^{N} \left| \frac{y_i - \hat{y}_i}{y_i} \right| \cdot 100$ \\ \addlinespace[5pt]
        Root relative squared error (RRSE) & $\sqrt{\frac{\sum_{i=1}^{N} (y_i - \hat{y}_i)^2}{\sum_{i=1}^{N} (y_i - \bar{y})^2}}$ \\ \addlinespace[5pt]
        Coefficient of determination (R$^2$) & $1-\frac{\sum_{i=1}^{N}(y_i-\hat{y}_i)^2}{\sum_{i=1}^{N}(y_i-\bar{y})^2}$ \\ \addlinespace[5pt]
        Correlation coefficient (CORR) & $\frac{\sum_{i=1}^{N} (y_i - \bar{y})(\hat{y}_i - \bar{\hat{y}})}{\sqrt{\sum_{i=1}^{N} (y_i - \bar{y})^2 \sum_{i=1}^{N} (\hat{y}_i - \bar{\hat{y}})^2}}$ \\
        \bottomrule
    \end{tabular}
\end{table}
Similar formulas can be applied to multi-horizon forecasting problems. All the metrics listed, except for the last one, are considered better when their values are lower, as they indicate smaller discrepancies between the predicted and actual values. In contrast, CORR quantifies the linear correlation between actual values and forecasts, and a high correlation is preferable.

As for classification problems, common evaluation metrics include accuracy, precision, recall, and F1 score. Given $TP$, $TN$, $FP$, and $FN$ the number of true positives, true negatives, false positives, and false negatives, respectively, the formulas for the binary classification problem are reported in Tab.~\ref{t:tabella metriche classificatione}. Similar formulas can be applied for the multi-class classification problem.
\begin{table}[ht]
    \caption{Evaluation metrics for binary classification tasks and their formulas.} \label{t:tabella metriche classificatione}
    \centering
    \begin{tabular}{ll}
        \toprule
        Metrics & Formula \\
        \midrule
        Accuracy & $\frac{TP + TN}{TP + TN + FP + FN}$ \\ \addlinespace[5pt]
        Precision & $\frac{TP}{TP + FP}$ \\ \addlinespace[5pt]
        Recall & $\frac{TP}{TP + FN}$ \\ \addlinespace[5pt]
        F1 score & $\frac{2 \cdot \text{Precision} \cdot \text{Recall}}{\text{Precision} + \text{Recall}}$ \\
        \bottomrule
    \end{tabular}
\end{table}

Besides accuracy, another important aspect to consider is the complexity of the model. Finding an optimal balance between accuracy and complexity is crucial for developing models that are not only accurate but also efficient. For this reason, some authors include the training and inference computational times, as well as the number of parameters of their models, allowing for a comparison of these aspects across different models. However, not all authors do so, and even if they do it, the characteristics of the hardware and other settings used by each author must be taken into account when comparing computational times. Consequently, it was not possible to provide an accurate and detailed comparison of the relationship between complexity and performance across the different models in this review. Although creating a plot to compare accuracy and computational times of the different models would be very useful (as in Refs.~\cite{art_407} and \cite{art_671}), it is not possible to provide such a comparison here. The authors of the various papers should be encouraged to incorporate such comparisons in their work.

\section{Thematic analysis}
\label{section:Thematic analysis}
This section explores the applications of the collected GNN models across different thematic groups. The goal is to provide an overview of available empirical studies, the most common approaches, and the results obtained concerning the discussed themes.

In each following subsection, one per theme, a list of the benchmark models that appear in the selected papers will be provided. Within each theme, they will be classified as: mathematical and statistical models coming from Econometrics, non-GNN machine learning models, and GNN-based models. Mathematical and statistical models are characterized by the fact that they generate forecasts by analyzing a limited amount of data, and they usually struggle with cross-sectional large-scale and temporal long-term dependencies. Non-GNN machine learning models use more complex and highly nonlinear architectures, mainly neural networks, but they are not necessarily able to capture intricate spatio-temporal dependencies within the data. GNN-based models, conversely, attempt to represent these complex dependencies using various techniques, resulting in better, though not perfect, accuracy. In addition, information on the most common datasets will be given, along with the best results reported in the literature. All results presented in the tables of this SLR are taken directly from the tables of the selected papers, as we did not conduct any further experiments to obtain additional results. When examining the results tables, the reader is invited to consult the GitHub repository associated with this review, which provides interactive functionality for sorting and filtering models by any metric and generating plots, making it easy to analyze and compare the results. Note that in the tables which list benchmark models we have included references to any GNN model found. In the results tables, instead, we have explicitly referenced only those models proposed in the selected papers, in order to highlight them. Further information about the models used as benchmarks can be found either in the benchmark tables or in the papers where they are used as benchmarks. The most common benchmarks used across all groups of papers are also listed in the GitHub repository.

The thematic groups of papers are presented in alphabetical order, with the exception of the \enquote{Generic} applications group (which cannot be directly attributed to a specific case study) and the \enquote{Other topics} group (focusing on specific problems of other disciplines). In subsections with a large number of papers and high homogeneity of the datasets, more precise comparisons of the results and in-depth analyses will be conducted.

\subsection{Energy}
\label{subsection:Energy}
The first thematic group is \enquote{Energy} (with 33 out of 366 papers), which includes all the studies related to energy systems. In particular, it includes research on electricity load, energy consumption, and power generation forecasting. These topics are somehow connected to those of the \enquote{Environment} subsection discussed later (Subsec.~\ref{subsection:Environment}). For example, green energy sources are associated with environmental phenomena such as wind speed and solar radiation inflow. Our rationale for this selection is that here we focus on energy applications, whereas Subsec.~\ref{subsection:Environment} will refer to the study of environmental data that may or may not be related to energy applications. Wind speed forecasting, for instance, is connected to wind power generation due to the cubic relationship between wind speed and wind power \cite{art_29}. However, wind speed forecasting may also be of interest for other reasons, and thus it is included in the more general Subsec.~\ref{subsection:Environment}.

\subsubsection{Overview}
Energy forecasting is crucial for electricity power grid stability and operational efficiency. Accurate load and generation forecasts ensure that supply meets real-time demand, preventing outages and instability, and maintaining a reliable supply of power. In addition, effective forecasting allows for optimal scheduling of power plants and energy resources, reducing operational costs and improving overall system efficiency. In the papers within this theme, the majority of econometric and non-GNN machine learning models tend to focus on time series from a single site or consider multiple sites without explicitly capturing the relationships between them. GNN models may be an innovative and effective approach to addressing energy related problems.

An overview of the selected papers reveals a relevant interest in the application of GNNs in the energy field starting from 2022. The most frequently studied fields include forecasting of wind \cite{art_20, art_24, art_27, art_332, art_333} and solar power \cite{art_21, art_22, art_26, art_349}, which are renewable energy sources characterized by their intermittency. The most popular journal among the selected papers is \giornale{Applied Energy} by Elsevier, with 5 published papers, and the only \enquote{Energy} conference paper was presented at \conferenza{IJCAI}. 

\subsubsection{Datasets}
The study of the selected papers in the \enquote{Energy} group suggests that there are no particularly common or widely recognized benchmark datasets within this field. Indeed, researchers often focus on specific datasets, typically collected within the university campus or in the country where they are located. For the same reason, the links to these datasets are not always available. In Tab.~\ref{t:link dataset energy} are listed the links to the public datasets mentioned in the papers.

\begin{longtable}{R{3.7cm}R{3cm}p{7.5cm}}
    \caption{List of public datasets in the \enquote{Energy} group and their corresponding links.}
    \label{t:link dataset energy}
    \endfirsthead
    \endhead
    \toprule
    Dataset & Used by & Link \\
    \midrule
    Commercial and residential load & \cite{art_348} & \url{https://www.osti.gov/biblio/1788456} \\
    Danish smart heat meters & \cite{art_15} & \url{https://doi.org/10.5281/zenodo.6563114} \\
    IEEE bus systems & \cite{art_23} & \url{https://doi.org/10.1109/UPEC.2015.7339813} (source paper) \\
    Individual household electric power consumption & \cite{art_344} & \url{https://archive.ics.uci.edu/ml/datasets/individual+household+electric+power+consumption} \\
    JERICHO-E-usage & \cite{art_15} & \url{https://doi.org/10.6084/m9.figshare.c.5245457.v1} \\
    NREL & \cite{art_20}, \cite{art_22}, \cite{art_24} & \url{https://www.nrel.gov/} \\
    OpenEI residential load & \cite{art_365} & \url{https://openei.org/datasets/files/961/pub/RESIDENTIAL_LOAD_DATA_E_PLUS_OUTPUT/HIGH/} \\
    OPSD & \cite{art_16} & \url{https://open-power-system-data.org/} \\
    PJM hourly energy consumption & \cite{art_390} & \url{https://www.pjm.com/markets-and-operations/data-dictionary} \\
    PVGIS & \cite{art_349} & \url{http://re.jrc.ec.europa.eu/pvgis/apps4/pvest.php} \\
    PV Switzerland & \cite{art_21}, \cite{art_22} & \url{https://doi.org/10.1109/IJCNN48605.2020.9207573} (source paper) \\
    Residential heat and electricity demand & \cite{art_341} & \url{https://ieee-dataport.org/open-access/8-years-hourly-heat-and-electricity-demand-residential-building} \\
    SGCC dataset & \cite{art_342} & \url{http://www.sgcc.com.cn/} \\
    Southern power grid & \cite{art_344} & \url{https://www.csg.cn/} \\
    UMass smart* & \cite{art_17} & \url{https://traces.cs.umass.edu/docs/traces/smartstar/} \\
    \bottomrule
\end{longtable}

Three of the selected papers \cite{art_20, art_22, art_24} use data from the National Renewable Energy Laboratory (NREL), a national laboratory of the U.S. Department of Energy (\url{https://www.nrel.gov/}).
The data time granularity ranges from 1 minute to 1 day, and the forecasting horizon from tens of minutes to days. These quantities depend on the dataset, the domain of application, and the purpose of the forecast. Most datasets require pre-processing, including missing data interpolation, normalization, and satellite image processing.

\subsubsection{Types of models}
As for the taxonomy of the proposed GNN models, 17 out of 33 papers use a convolutional GNN, and 8 use an attentional GNN. Seven papers use both convolutional and attentional approaches, and one paper \cite{art_338} uses the original recurrent approach on a graph defined via the visibility algorithm. As to the graph structure used, the papers typically provide only a brief description. For example, only few papers highlight whether the graph structure is static or dynamic, and the way in which it is constructed is often described summarily. In some papers, the graph structure reflects the geographic location of the elements (e.g., wind or solar power plants, as in Refs.~\cite{art_24, art_26, art_349}), whereas in others it is based on the correlation or similarity between time series (as in Ref.~\cite{art_337}). In some studies, a pre-defined structure is imposed and the weighted adjacency matrix is learned directly from the model, often using the attention mechanism as in Ref.~\cite{art_22}. The majority of papers in this group rely on a static graph.

The most common loss function used to train these models is the MSE. Many papers mention the use of the programming languages Python and Matlab, and few of them mention the use of the Python libraries TensorFlow \cite{tensorflow2015-whitepaper}, PyTorch \cite{PyTorch}, and PyTorch Geometric \cite{PyTorch_Geometric} for the implementation of the GNN model. In this group, only Ref.~\cite{art_17} provides the link to the source code of the proposed residential load forecasting with multiple correlation temporal graph neural networks (RLF-MGNN) model (available at \url{https://codeocean.com/capsule/9294192/tree/v1}).

\subsubsection{Benchmark models}
Tab.~\ref{t:benchmark models energy} lists the benchmark models used in the \enquote{Energy} group by at least two papers included in this review.

\begin{longtable}{p{2.5cm}p{6.5cm}p{5.2cm}}
    \caption{List of benchmark models in the \enquote{Energy} group divided per category.}
    \label{t:benchmark models energy}
    \endfirsthead
    \endhead
    \toprule
    Category & Model & Used by \\
    \midrule
    \multirow[t]{3}{=}{Mathematical and statistical methods} & Autoregressive integrated moving average (ARIMA) & \cite{art_15}, \cite{art_16}, \cite{art_22}, \cite{art_25}, \cite{art_333}, \cite{art_337}, \cite{art_338}, \cite{art_341}, \cite{art_343}, \cite{art_349}, \cite{art_635} \\
    & Historical average (HA) & \cite{art_350}, \cite{art_390} \\
    & Vector ARIMA & \cite{art_332}, \cite{art_333}, \cite{art_343}, \cite{art_350}, \cite{art_390} \\
    \midrule 
    \multirow[t]{4}{=}{Non-GNN machine learning methods} & Autoformer \cite{Autoformer} & \cite{art_333}, \cite{art_341} \\
    & Bidirectional long-short term memory network (BiLSTM) \cite{BiLSTM} & \cite{art_15}, \cite{art_339} \\
    & Convolutional neural network (CNN) & \cite{art_15}, \cite{art_18}, \cite{art_21}, \cite{art_22}, \cite{art_343}, \cite{art_347} \\
    & Hybrid convolutional neural network and long-short term memory network (CNN-LSTM) & \cite{art_15}, \cite{art_17}, \cite{art_351} \\
    & Encoder-decoder architectures (ED) & \cite{art_15}, \cite{art_21}, \cite{art_22} \\
    & Feed-forward neural network (FNN) & \cite{art_15}, \cite{art_18}, \cite{art_25}, \cite{art_27}, \cite{art_351}, \cite{art_635} \\
    & Gated recurrent unit (GRU) & \cite{art_15}, \cite{art_25}, \cite{art_334}, \cite{art_338}, \cite{art_343}, \cite{art_347} \\
    & Informer \cite{Informer} & \cite{art_15}, \cite{art_16} \\
    & $k$-nearest neighbors (KNN) & \cite{art_18}, \cite{art_24}, \cite{art_342}, \cite{art_343} \\
    & Long-short term memory network (LSTM) & \cite{art_15}, \cite{art_16}, \cite{art_20}, \cite{art_23}, \cite{art_24}, \cite{art_333}, \cite{art_334}, \cite{art_335}, \cite{art_340}, \cite{art_341}, \cite{art_343}, \cite{art_344}, \cite{art_347}, \cite{art_348}, \cite{art_349}, \cite{art_353}, \cite{art_635} \\
    & Long-short term time series network (LSTNet) \cite{LSTNet} & \cite{art_15}, \cite{art_332}, \cite{art_334} \\
    & Recurrent neural network (RNN) & \cite{art_16}, \cite{art_17}, \cite{art_20} \\
    & Seq2Seq \cite{Seq2seq} & \cite{art_335}, \cite{art_343} \\
    & Support vector regression (SVR) & \cite{art_21}, \cite{art_22}, \cite{art_24}, \cite{art_27}, \cite{art_338}, \cite{art_342}, \cite{art_343}, \cite{art_344}, \cite{art_348}, \cite{art_635} \\
    & Transformer \cite{attention} & \cite{art_16}, \cite{art_335}, \cite{art_343} \\
    & Extreme gradient boosting (XGBoost) & \cite{art_25}, \cite{art_334}, \cite{art_335}, \cite{art_341} \\
    \midrule 
    \multirow[t]{1}{=}{GNN methods} & Diffusion convolutional recurrent neural network (DCRNN) \cite{DCRNN} & \cite{art_349}, \cite{art_350}, \cite{art_390} \\
    & Multivariate time series forecasting with graph neural network (MTGNN) \cite{art_655} & \cite{art_332}, \cite{art_340}, \cite{art_341} \\
    & Spatio-temporal autoregressive model (ST-AR) \cite{ST-AR} & \cite{art_21}, \cite{art_22} \\
    & Spatio-temporal graph convolutional network (STGCN) \cite{STGCN} & \cite{art_339}, \cite{art_341}, \cite{art_348}, \cite{art_351}, \cite{art_353} \\
    \bottomrule
\end{longtable}

The majority of the papers only use non-GNN machine learning benchmarks. Some papers use simple mathematical and statistical models, such as the ARIMA model and the historical average. Only few papers use GNN-based models as benchmarks, and the most popular benchmark among them is the spatio-temporal graph convolutional network (STGCN). Before 2024, the only benchmark used by two papers was the spatio-temporal autoregressive model (ST-AR). This is due to the fact that this thematic field has not been extensively explored in the past, and only a very limited number of GNN models have been specifically developed for energy forecasting.

\subsubsection{Results}
In general, the authors of the papers in this group claim that their proposed GNN model outperforms the benchmarks. However, in some cases, it turns out that basic statistical methods may achieve good results as well \cite{art_390}. The most common error metrics used are the MAE, the RMSE, and the Normalized Root mean squared Error (NRMSE). Some papers calculate these error metrics across different seasons to evaluate the model's performance under different conditions. An example is provided in Ref.~\cite{art_349}, where the authors also use the GNNExplainer \cite{interpret_2} to assess the influence of different variables.

It is not possible to compare the accuracy of the models included in this group, because no common benchmark datasets were used. The only two papers that study the same dataset are written by the same authors (Refs.~\cite{art_21, art_22}), and their most recent paper includes a comparison with the models proposed in the first one.

\subsection{Environment}
\label{subsection:Environment}
The thematic group \enquote{Environment} includes a relevant number of selected papers (43 out of 366). This suggests a considerable research interest on the application of GNNs in this field. However, this group is highly fragmented, as it includes a large number of subdomains and applications, all related to the study of environmental data.

\subsubsection{Overview}
The group of papers related to environmental topics is broad, and includes papers on disciplines such as physics, climatology, oceanography, and atmospheric science. The breadth of the group is reflected in the number of case studies, which cover air quality prediction (in terms of PM$_{2.5}$ concentration), sea temperature, wind speed, and other applications. Forecasting these quantities can be useful for several reasons. For instance, predicting wind speed or rainfall can assist in anticipating some extreme weather phenomena, and is also related to the prediction of green energy outputs. For example, in Refs.~\cite{art_29} and \cite{art_34} the authors estimate the wind power forecasting error obtained by using wind speed forecasting models. As another example, predicting sea surface temperature can be useful for weather forecasting, fishing directions, and disaster warnings \cite{art_28}. Moreover, monitoring and forecasting environmental data are crucial for evaluating the impact of human activities on the environment and tracking the progression of climate change.

The methods for forecasting environmental data are typically categorized as numerical methods, statistical methods, and machine learning methods. Numerical methods are based on atmospheric models, and are designed to quantify the interaction of different atmospheric variables. The accuracy of these models depends on the availability of data, and the mathematical simulations needed to get the forecasts may take days or even weeks, thus limiting the ability to make good short-term forecasts. Linear statistical methods leverage long-term dependencies in the recorded data by using regression-like models. However, they cannot capture non-linear relationships in the dataset, which limits their power. Machine learning methods, on the other hand, are often advanced generalizations of linear statistical methods that can capture more complex and non-linear relationships in the data.

An overview of the selected papers reveals that there has been a rapid increase in interest on this theme among the GNN research community too, and more than 35 of the selected papers were published from 2023. The most investigated data are related to sea temperature (Refs.~\cite{art_28, art_33, art_35, art_36, art_39, art_41, art_47, art_636}), wind speed (Refs.~\cite{art_29, art_31, art_34, art_40, art_44, art_46, art_366}), and PM$_{2.5}$ concentration forecasting (Refs.~\cite{art_2, art_3, art_4, art_6, art_7, art_8, art_354, art_361, art_371}). Almost every selected paper is published in a different journal, with the exception of \giornale{Energy}, \giornale{Engineering Applications of Artificial Intelligence}, \giornale{Journal of Cleaner Production}, \giornale{Knowledge-Based Systems}, \giornale{Science of the Total Environment} by Elsevier, each with two published papers. As for conference papers, the three \enquote{Environment} papers were presented at \conferenza{KDD}, \conferenza{IJCAI} and \conferenza{CIKM}. 

\subsubsection{Datasets}
The analysis of the selected papers shows that in this thematic group there is no standardized benchmark dataset. The publicly available datasets cited in the papers are listed in Tab.~\ref{t:link dataset environment}.

\begin{longtable}{R{3.7cm}R{3cm}p{7.5cm}}
    \caption{List of public datasets in the \enquote{Environment} group and their corresponding links.}
    \label{t:link dataset environment}
    \endfirsthead
    \endhead
    \toprule
    Dataset & Used by & Link \\
    \midrule 
    AgroData-10 & \cite{art_28} & \url{https://www.aoml.noaa.gov/phod/argo} \\
    Carbon monitor & \cite{art_360} & \url{https://carbonmonitor.org/} \\
    CCMP wind data & \cite{art_31}, \cite{art_40}, \cite{art_44} & \url{http://data.remss.com} \\
    China environmental monitoring station PM2.5 & \cite{art_2}, \cite{art_8}, \cite{art_371} & \url{https://www.cnemc.cn/} \\
    China national urban air quality & \cite{art_1}, \cite{art_5} & \url{https://github.com/Friger/GAGNN} \\
    Copernicus 3D thermohaline data & \cite{art_35} & \url{https://resources.marine.copernicus.eu/products} \\
    Delaware river basin & \cite{art_636} & \url{https://waterdata.usgs.gov/nwis} \\
    ECMWF east China sea & \cite{art_45} & \url{https://github.com/Boltzxuann/GIPMN/tree/dataset} \\
    ECMWF south and east China sea & \cite{art_32} & \url{http://apps.ecmwf.int/datasets/data/interim-full-daily} $^\dagger$ \\
    HadCRUT5 & \cite{art_370} & \url{https://crudata.uea.ac.uk/cru/data/temperature/} \\
    HUC12 &  \cite{art_637} & \url{https://hawqs.tamu.edu/} \\
    Jinan PM2.5 & \cite{art_354} & \url{https://github.com/liaohaibing/MGST_GNN/tree/master} \\
    KnowAir & \cite{art_6} & \url{https://github.com/shuowang-ai/PM2.5-GNN} \\
    Korean peninsula sea data & \cite{art_33} & \url{https://www.nifs.go.kr/risa/main.risa} \\
    NARR weather conditions & \cite{art_691} & \url{https://psl.noaa.gov/data/gridded/data.narr.html} \\
    NOAA sea datasets & \cite{art_28}, \cite{art_36}, \cite{art_39}, \cite{art_46}, \cite{art_47}, \cite{art_365} & \url{https://www.ndbc.noaa.gov/} \\
    Norwegian offshore wind & \cite{art_29}, \cite{art_34} & \url{https://frost.met.no/index.html} \\
    Overbetuwe groundwater data & \cite{art_359} &\url{https://github.com/mlttac/GroundwaterFlowGNN} \\
    Taiwan air quality dataset & \cite{art_3} & \url{https://data.gov.tw/en} \\
    UrbanAir & \cite{art_6} & \url{https://doi.org/10.1016/j.chemosphere.2018.12.128} (source paper) \\
    US EPA PM$_{2.5}$ & \cite{art_1}, \cite{art_4} & \url{https://aqs.epa.gov/aqsweb/airdata/download_files.html} \\
    WeatherBench & \cite{art_691} & \url{https://github.com/pangeo-data/WeatherBench} \\
    Yangtze river PM2.5 & \cite{art_354} & \url{https://github.com/liaohaibing/MGST_GNN/tree/master} \\
    \bottomrule
\end{longtable}

Six of the selected papers \cite{art_28, art_36, art_39, art_46, art_47, art_365} use National Oceanic and Atmospheric Administration (NOAA) sea datasets (\url{https://www.noaa.gov/}) from the U.S. Department of Commerce (US DOC). The US DOC website permits users to download daily, weekly and monthly mean optimum interpolation sea surface temperature (OISST) data from September 1981 for many geographical locations. In over half of the papers, exogenous variables are incorporated into the model. These include weather data (e.g., humidity, rainfall, pressure, temperature, wind) and, in the case of PM$_{2.5}$ concentration forecasting, concentration of pollutants (like CO, NO$_2$, O$_3$, PM$_{10}$ and SO$_2$).

Overall, many papers also consider exogenous variables, in particular weather data, as they have the greatest influence on environmental data.

The data time granularity ranges from 1 minute to 1 month, depending on the case study, and the forecasting horizon ranges from 10 minutes to more than one year. Most datasets require some type of pre-processing, including interpolation of missing data, removal of outliers and normalization.

\subsubsection{Types of models}
Regarding the taxonomy of the proposed GNN models, the majority of papers present convolutional GNNs (28 out of 43 papers), followed by 8 attentional GNNs, and by 3 hybrid architectures. The description in 4 of the papers (Refs.~\cite{art_5, art_354, art_361, art_636}), however, does not allow us to specify the classification of the model within the proposed taxonomy. As for the graph structure, not all of the papers describe it with sufficient precision, and some of them are vague about the graph construction and its static or dynamic nature. In almost half of the papers, the edge weights are learned directly from the model, often using the graph attention mechanism. When the graph structure is defined manually, the most popular criterion used is the spatial distance between the sites ($d$), even in its exponential ($e^{-d}$) or inverse proportion ($1/d$) formulation, or similarity measures, such as cosine similarity and correlation of the time series. Finally, there are a few papers where more than one graph structure is fused together in order to capture more complex dynamics, as in Ref.~\cite{art_2}. Almost all the papers in this thematic group deal with multivariate time series, with the exception of Ref.~\cite{art_370}, which studies a univariate time series of global average temperature. In this work, the authors treat each time step $t$ as a node $v_t$ in the graph, and the adjacency matrix is constructed in such a way that $v_t$ is influenced by $v_{t-1}$ and $v_{t+1}$.

Not all papers openly write down the loss function used to train the models. Among those that do, the most popular loss functions are MSE and MAE. Many papers mention the use of the programming language Python and of the library PyTorch for the implementation of the models. Few papers also provide a link to the source code for the proposed model, as shown in Tab.~\ref{t:link codici environment}.

\begin{longtable}{p{7cm}p{7.5cm}}
    \caption{List of source codes of the \enquote{Environment} models in the review.}
    \label{t:link codici environment}
    \endfirsthead
    \endhead
    \toprule
    Model & Link \\
    \midrule
    Domain inspired temporal graph convolution neural network \cite{art_637} & \url{https://github.com/IBM/tgcn-soil-moisture} \\
    Dynamic spatial-temporal graph convolutional recurrent network (DSTGCRN) \cite{art_360} & \url{https://github.com/PalaceTony/DSTGCRN} \\
    Dynamic spatiotemporal graph neural network (DST-GNN) \cite{art_354} & \url{https://github.com/liaohaibing/MGST_GNN/tree/master} \\
    Group-aware graph neural network (GAGNN) \cite{art_1} & \url{https://github.com/Friger/GAGNN} \\
    Gridded information propagation and mixing network (GIPMN) \cite{art_45} & \url{https://github.com/Boltzxuann/GIPMN/tree/dataset} \\
    GroundwaterFlowGNN \cite{art_359} & \url{https://github.com/mlttac/GroundwaterFlowGNN} \\
    Hierarchical graph recurrent network (HiGRN) \cite{art_28} & 	\url{https://github.com/Neoyanghc/HiGRN} \\
    Physics-aware graph networks (PaGN) \cite{art_365} & \url{https://github.com/eunviho/novel_pagn} \\
    Spatio-temporal FFTransformer (ST-FFTransformer) \cite{art_29} & 	\url{https://github.com/LarsBentsen/FFTransformer} \\
    \bottomrule
\end{longtable}

\subsubsection{Benchmark models}
Tab.~\ref{t:benchmark models environment} presents the benchmark models employed in the \enquote{Environment} group by at least two articles examined in this review.

\begin{longtable}{p{2.5cm}p{6.5cm}p{5.2cm}}
    \caption{List of benchmark models in the \enquote{Environment} group divided per category.}
    \label{t:benchmark models environment}
    \endfirsthead
    \endhead
    \toprule
    Category & Model & Used by \\
    \midrule
    \multirow[t]{3}{=}{Mathematical and statistical methods} & Autoregressive integrated moving average (ARIMA) & \cite{art_28}, \cite{art_33}, \cite{art_34}, \cite{art_38}, \cite{art_363}, \cite{art_370}, \cite{art_371}, \cite{art_691} \\
    & Historical average (HA) & \cite{art_3}, \cite{art_6}, \cite{art_28}, \cite{art_38}, \cite{art_691} \\
    & Naive & \cite{art_29}, \cite{art_34}, \cite{art_366} \\
    & Vector autoregression (VAR) & \cite{art_28}, \cite{art_34}, \cite{art_38}, \cite{art_360} \\
    \midrule 
    \multirow[t]{4}{=}{Non-GNN machine learning methods} & Bidirectional long-short term memory network (BiLSTM) \cite{BiLSTM} & \cite{art_33}, \cite{art_46} \\
    & Combined FC-LSTM and convolution neural network (CFCC-LSTM) \cite{CFCC-LSTM} & \cite{art_35}, \cite{art_39}, \cite{art_370} \\
    & DLinear \cite{DLinear} & \cite{art_31}, \cite{art_364} \\
    & Hybrid convolutional neural network and gated recurrent unit (CGRU) \cite{CGRU} & \cite{art_31}, \cite{art_40}, \cite{art_44} \\
    & Convolutional neural network (CNN) & \cite{art_30}, \cite{art_46} \\
    & Hybrid convolutional neural network and long-short term memory network (CNN-LSTM) & \cite{art_4}, \cite{art_41}, \cite{art_368}, \cite{art_371} \\
    & Convolutional long-short term memory network (ConvLSTM) \cite{ConvLSTM} & \cite{art_28}, \cite{art_41}, \cite{art_355} \\
    & Feed-forward neural network (FNN) & \cite{art_2}, \cite{art_7}, \cite{art_38}, \cite{art_360}, \cite{art_365}, \cite{art_370} \\
    & Gated recurrent unit encoder–decoder (GED) \cite{GED} & \cite{art_36}, \cite{art_39}, \cite{art_47} \\
    & Gated recurrent unit (GRU) & \cite{art_2}, \cite{art_6}, \cite{art_7}, \cite{art_8}, \cite{art_38}, \cite{art_43}, \cite{art_46}, \cite{art_361}, \cite{art_364}, \cite{art_365}, \cite{art_370}, \cite{art_691} \\
    & Informer \cite{Informer} & \cite{art_30}, \cite{art_47}, \cite{art_354}, \cite{art_364}, \cite{art_371}, \cite{art_691} \\
    & Long-short term memory network (LSTM) & \cite{art_1}, \cite{art_2}, \cite{art_3}, \cite{art_4}, \cite{art_5}, \cite{art_6}, \cite{art_7}, \cite{art_8}, \cite{art_28}, \cite{art_30}, \cite{art_32}, \cite{art_33}, \cite{art_35}, \cite{art_36}, \cite{art_38}, \cite{art_39}, \cite{art_43}, \cite{art_46}, \cite{art_47}, \cite{art_357}, \cite{art_361}, \cite{art_364}, \cite{art_365}, \cite{art_368}, \cite{art_370}, \cite{art_371}, \cite{art_373}, \cite{art_636}, \cite{art_637}, \cite{art_691} \\
    & Long-short term time series network (LSTNet) \cite{LSTNet} & \cite{art_355}, \cite{art_373} \\
    & Predictive deep convolutional neural network (PDCNN) \cite{PDCNN} & \cite{art_31}, \cite{art_40} \\
    & Random forest (RF) & \cite{art_6}, \cite{art_364}, \cite{art_370} \\
    & Recurrent neural network (RNN) & \cite{art_46}, \cite{art_365} \\
    & Support vector regression (SVR) & \cite{art_3}, \cite{art_6}, \cite{art_35}, \cite{art_36}, \cite{art_38}, \cite{art_360}, \cite{art_370}, \cite{art_371} \\
    & Temporal convolutional network (TCN) \cite{TCN} & \cite{art_6}, \cite{art_30}, \cite{art_355}, \cite{art_368}, \cite{art_371}, \cite{art_373} \\
    & Temporal pattern attention long-short term memory network (TPA-LSTM) \cite{TPA-LSTM} & \cite{art_6}, \cite{art_31}, \cite{art_32}, \cite{art_35}, \cite{art_40}, \cite{art_44} \\
    & Transformer \cite{attention} & \cite{art_30}, \cite{art_33}\\
    & Hybrid wavelet random forest and deep belief network (Wavelet-DBN-RF) \cite{Wavelet-DBN-RF} & \cite{art_31}, \cite{art_40}, \cite{art_44} \\
    & Extreme gradient boosting (XGBoost) & \cite{art_1}, \cite{art_2}, \cite{art_5}, \cite{art_46}, \cite{art_364} \\
    \midrule 
    \multirow[t]{1}{=}{GNN methods} & Adaptive graph convolutional recurrent network (AGCRN) \cite{art_683} & \cite{art_6}, \cite{art_360}, \cite{art_365}, \cite{art_368} \\
    & Attention-based spatial-temporal graph convolutional network (ASTGCN) \cite{ASTGCN} & \cite{art_363}, \cite{art_368}, \cite{art_691} \\
    & Diffusion convolutional recurrent neural network (DCRNN) \cite{DCRNN} & \cite{art_363}, \cite{art_691} \\
    & Graph convolutional neural network and long-short term memory (GC-LSTM) \cite{GC-LSTM} & \cite{art_1}, \cite{art_2}, \cite{art_5}, \cite{art_6}, \cite{art_7}, \cite{art_354}, \cite{art_361} \\
    & Graph convolutional network (GCN) & \cite{art_38}, \cite{art_41}, \cite{art_357} \\
    & Graph WaveNet (GWN) \cite{GWN} & \cite{art_1}, \cite{art_6}, \cite{art_44}, \cite{art_47}, \cite{art_363}, \cite{art_368} \\
    & Hierarchical graph convolution network (HGCN) \cite{HGCN} & \cite{art_1}, \cite{art_28} \\
    & Multivariate time series forecasting with graph neural network (MTGNN) \cite{art_655} & \cite{art_6}, \cite{art_32}, \cite{art_44}, \cite{art_355}, \cite{art_360}, \cite{art_373}, \cite{art_691} \\
    & Multi long-short term memory network (Multi-LSTM) \cite{Multi-LSTM} & \cite{art_31}, \cite{art_40}, \cite{art_44} \\
    & Superposition graph neural network (SGNN) \cite{art_24} & \cite{art_31}, \cite{art_40} \\
    & Spatio-temporal graph convolutional network (STGCN) \cite{STGCN} & \cite{art_3}, \cite{art_6}, \cite{art_28}, \cite{art_35}, \cite{art_47}, \cite{art_363}, \cite{art_368}, \cite{art_371}, \cite{art_691} \\
    & Spatio-temporal correlation graph neural network (STGN) \cite{art_40} & \cite{art_31}, \cite{art_44} \\
    & Spatio-temporal synchronous graph convolutional network (STSGCN) \cite{STSGCN} & \cite{art_368}, \cite{art_371} \\
    & Temporal graph convolutional network (T-GCN) \cite{T-GCN} & \cite{art_6}, \cite{art_30}, \cite{art_35}, \cite{art_637}, \cite{art_691} \\
    \bottomrule
\end{longtable}

The majority of the benchmarks used in this thematic group fall within the category of non-GNN machine learning methods, although there are several GNN benchmarks as well. The two most common benchmark models are LSTMs and GRUs. These models are widely used by environmental data forecasters due to their ability to capture temporal dependencies, and are considered a reliable benchmark in the field. As for GNN models, the most common benchmarks are the graph convolutional neural network and long-short term memory network (GC-LSTM) \cite{GC-LSTM} (which appears only in quality air forecasting problems), and the spatio-temporal graph convolutional network (STGCN) \cite{STGCN}. The GC-LSTM network is a model that integrates graph convolutional networks and long-short-term memory networks. It was proposed by Qi et al.~in 2019 to forecast the spatio-temporal variation of PM$_{2.5}$ concentration. In the original paper, the GC-LSTM model works with a pre-defined graph based on the geographical distance between air quality stations. The input layer of the GC-LSTM model receives historical air quality observations, meteorological variables, spatial predictors, and temporal predictors. This information is processed through a graph convolution operation to extract spatial features. After that, the temporal feature extraction is performed using a LSTM network, which takes as input the graph convolutional features concatenated with the original signals. The STGCN model, on its part, was proposed in the traffic domain by Yu et al.~in 2018, and works with pre-defined adjacency matrices computed based on the distances among stations in the traffic network. The STGCN model is composed of two spatio-temporal convolutional blocks and a fully-connected output layer. Each convolutional block contains two temporal gated convolution layers and one spatial graph convolution layer in the middle. The code for the STGCN model is available at \url{https://github.com/VeritasYin/STGCN_IJCAI-18}.
Finally, it is important to point out that the so-called multivariate time series forecasting with graph neural network (MTGNN) model proposed by Wu et al.~in 2020 in Ref.~\cite{art_655} is often reported in the papers using it as the second-best performing model, thus it probably deserves special attention and further investigation in future research.

\subsubsection{Results}
In each paper of this group, it is stated that the proposed model is significantly better than the benchmarks. The accuracy of the models is measured in terms of MAE, RMSE, and MAPE. Among the papers, there is a high diversity in the choice of the proposed datasets. However, there are two sets of papers using the same dataset, whose results are compared hereafter.

The first dataset is the CCMP wind speed dataset (available at \url{http://data.remss.com}), which contains 3,942 geographical grid points from the east and southeast sea areas of China. The three papers using it (Refs.~\cite{art_31, art_40, art_44}) select 120 points from the area and use data records from January 2010 to April 2019 at 6 hours intervals, with a total of 13,624 samples at each node, for 10 different forecasting horizons. All papers present results for the complete set of 120 nodes and a subset of nodes. Since these results are similar, only the average results for all the 120 nodes are reported in Tab.~\ref{t:risultati ccmp} for the sake of simplicity.

{\small\tabcolsep=3pt
\begin{longtable}{l*{11}{c}}
    \caption{Comparison of the average accuracy of the different models on the CCMP wind speed dataset for different time horizons, expressed in terms of MAE and RMSE. The smallest errors are underlined. Numbers from the original papers.}
    \label{t:risultati ccmp}
    \endfirsthead
    \endhead
    \toprule
    Model & Metrics & 6 h & 12 h & 18 h & 1 d & 2 d & 3 d & 4 d & 5 d & 6 d & 7 d \\
    \midrule
    \multirow{2}{*}{CGRU} & MAE & 1.658 & 1.795 & 1.937 & 2.038 & 2.290 & 2.401 & 2.478 & 2.525 & 2.556 & 2.594 \\
     & RMSE & 2.100 & 2.277 & 2.461 & 2.594 & 2.911 & 3.042 & 3.141 & 3.198 & 3.245 & 3.285 \\
    \multirow{2}{*}{DAGLN \cite{art_31}} & MAE & 1.068 & 1.219 & 1.346 & 1.443 & 1.716 & 1.852 & 1.935 & 1.984 & 2.019 & 2.067 \\
     & RMSE & 1.423 & 1.625 & 1.793 & 1.917 & 2.255 & 2.428 & 2.512 & 2.570 & 2.607 & 2.656 \\
    \multirow{2}{*}{DASTGN \cite{art_44}} & MAE & \underline{1.039} & \underline{1.184} & \underline{1.297} & \underline{1.387} & \underline{1.662} & \underline{1.806} & \underline{1.900} & \underline{1.945} & \underline{1.995} & \underline{2.028} \\
     & RMSE & \underline{1.384} & \underline{1.572} & \underline{1.721} & \underline{1.829} & \underline{2.170} & \underline{2.338} & \underline{2.449} & \underline{2.498} & \underline{2.554} & \underline{2.600} \\
    \multirow{2}{*}{DLinear} & MAE & 1.145 & 1.306 & 1.432 & 1.528 & 1.798 & 1.937 & 2.022 & 2.080 & 2.124 & 2.155 \\
     & RMSE & 1.528 & 1.745 & 1.913 & 2.037 & 2.366 & 2.512 & 2.604 & 2.666 & 2.713 & 2.748 \\
    \multirow{2}{*}{\shortstack[l]{EMD-\\SVRCKH}} & MAE & 1.111 & 1.311 & 1.432 & 1.526 & 1.788 & 1.919 & 1.998 & 2.051 & 2.090 & 2.119 \\
     & RMSE & 1.489 & 1.755 & 1.909 & 2.027 & 2.344 & 2.495 & 2.587 & 2.650 & 2.695 & 2.730 \\
    \multirow{2}{*}{\shortstack[l]{Graph\\WaveNet}} & MAE & 1.101 & 1.265 & 1.386 & 1.487 & 1.762 & 1.901 & 1.951 & 1.992 & 2.089 & 2.124 \\
     & RMSE & 1.467 & 1.685 & 1.852 & 1.982 & 2.304 & 2.466 & 2.554 & 2.582 & 2.684 & 2.717 \\
    \multirow{2}{*}{\shortstack[l]{Historical\\Inertia}} & MAE & 1.139 & 1.301 & 1.405 & 1.511 & 1.773 & 1.912 & 2.007 & 2.081 & 2.109 & 2.114 \\
     & RMSE & 1.494 & 1.704 & 1.843 & 1.977 & 2.304 & 2.467 & 2.571 & 2.637 & 2.678 & 2.687 \\
    \multirow{2}{*}{MTGNN} & MAE & 1.091 & 1.237 & 1.351 & 1.454 & 1.724 & 1.864 & 1.935 & 1.990 & 2.036 & 2.059 \\
     & RMSE & 1.454 & 1.650 & 1.801 & 1.936 & 2.267 & 2.436 & 2.515 & 2.579 & 2.632 & 2.657 \\
    \multirow{2}{*}{Multi LSTMs} & MAE & 1.312 & 1.645 & 1.914 & 2.129 & 2.610 & 2.748 & 2.793 & 2.818 & 2.842 & 2.867 \\
     & RMSE & 1.663 & 2.079 & 2.415 & 2.679 & 3.257 & 3.420 & 3.477 & 3.510 & 3.540 & 3.573 \\
    \multirow{2}{*}{PDCNN} & MAE & 2.025 & 2.026 & 2.385 & 2.551 & 2.959 & 3.011 & 3.012 & 2.971 & 2.918 & 2.877 \\
     & RMSE & 2.561 & 2.562 & 3.033 & 3.243 & 3.750 & 3.824 & 3.833 & 3.789 & 3.727 & 3.675 \\
    \multirow{2}{*}{SGNN} & MAE & 2.542 & 2.562 & 2.573 & 2.578 & 2.583 & 2.618 & 2.666 & 2.695 & 2.712 & 2.716 \\
     & RMSE & 3.222 & 3.246 & 3.264 & 3.270 & 3.275 & 3.328 & 3.392 & 3.431 & 3.453 & 3.458 \\
    \multirow{2}{*}{STGN \cite{art_40}} & MAE & 1.084 & 1.281 & 1.407 & 1.523 & 1.809 & 1.942 & 2.046 & 2.073 & 2.137 & 2.189 \\
     & RMSE & 1.430 & 1.697 & 1.863 & 2.013 & 2.358 & 2.514 & 2.631 & 2.649 & 2.717 & 2.771 \\
    \multirow{2}{*}{TPA-LSTM} & MAE & 1.203 & 1.552 & 1.803 & 1.984 & 2.432 & 2.667 & 2.836 & 2.959 & 3.027 & 3.045 \\
     & RMSE & 1.562 & 1.997 & 2.319 & 2.544 & 3.082 & 3.383 & 3.598 & 3.745 & 3.825 & 3.842 \\
    \multirow{2}{*}{\shortstack[l]{Wavelet-\\DBN-RF}} & MAE & 1.451 & 1.597 & 1.749 & 1.836 & 2.087 & 2.162 & 2.200 & 2.217 & 2.232 & 2.243 \\
     & RMSE & 1.855 & 2.038 & 2.226 & 2.328 & 2.617 & 2.703 & 2.744 & 2.766 & 2.788 & 2.798 \\
    \bottomrule   
\end{longtable}
}

For all the time horizons considered in the papers, the dynamic adaptive spatio-temporal graph neural network (DASTGN) model \cite{art_44} proposed by Gao et al.~in 2023 has the highest accuracy. The DASTGN model is composed of a graph generation module and a spatio-temporal convolution module. The dynamic adaptive graph generation module generates a static graph to model long-term stable dependencies, and a dynamic graph to model the short-term dynamics. These graphs are then merged into an optimal dynamic graph. The spatio-temporal convolution module consists of an input module, a stack of spatio-temporal layers connected with residual and skip connections, and an output module that generates wind speed forecasts. Other two well performing models are the data-based adaptive graph learning network (DAGLN) \cite{art_31}, and the multivariate time series forecasting with graph neural network (MTGNN) \cite{art_655}. The DAGLN model was proposed by Ren at al.~in 2023, and uses an encoder-decodes architecture. The encoder includes an adaptive graph learning module that learns the optimal adjacency matrix starting from a pre-defined one obtained by using the DTW algorithm, a graph convolution with node adaptive parameter learning algorithm for personalized learning of parameter space, and a GRU for temporal modeling. Then, this information is fed into the decoder that operates a convolutional and a linear operation to produce the wind speed forecast. The MTGNN model, proposed by Wu et al.~in 2020, is a well-known state-of-the-art model for multivariate time series forecasting, here included in the \enquote{Generic} group of papers. It incorporates a graph learning layer that learns a graph adjacency matrix from scratch, followed by a stack of graph convolution and temporal convolution modules to capture spatial and temporal dependencies, and an output module. To avoid the problem of vanishing gradients, residual and skip connections are added between spatial and temporal convolutional modules.

The second dataset shared by two papers (Refs.~\cite{art_1, art_5}) is called China National Urban Air Quality dataset (available at \url{https://github.com/Friger/GAGNN}). It contains air quality index data (AQI) of 209 cities, collected from January 2017 to April 2019 at 1 hour granularity, for a total of 20,370 samples. Both these papers integrate exogenous weather variables (humidity, wind direction, wind speed, rainfall, air pressure, temperature, and visibility) collected from Envicloud (\url{http://www.envicloud.cn/}). The average accuracy of the models used in the two papers is compared in Tab.~\ref{t:risultati chinese aqi}.

\begin{longtable}{l*{7}{c}}
    \caption{Comparison of the average accuracy of the different models on the Chinese AQI dataset for time horizons from 1 to 6 hours ahead, expressed in terms of MAE and RMSE. The smallest errors are underlined. Numbers from the original papers.}
    \label{t:risultati chinese aqi}
    \endfirsthead
    \endhead
    \toprule
    Model & Metrics & 1 h & 2 h & 3 h & 4 h & 5 h & 6 h \\
    \midrule
    \multirow{2}{*}{AirFormer} & MAE & 5.95 & 9.31 & 11.87 & 13.87 & 15.52 & 16.95 \\
     & RMSE & 11.49 & 17.23 & 21.32 & 24.31 & 26.72 & 28.71 \\
    \multirow{2}{*}{FGA} & MAE & 5.87 & 9.14 & 11.71 & 13.75 & 15.42 & 16.80 \\
     & RMSE & 11.36 & 17.01 & 21.05 & 24.09 & 26.55 & 28.52 \\
    \multirow{2}{*}{GAGNN \cite{art_1}} & MAE & 5.56 & 8.59 & 10.80 & 12.52 & 13.91 & 15.10 \\
     & RMSE & 10.81 & 16.17 & 19.84 & 22.51 & 24.63 & 26.37 \\
    \multirow{2}{*}{GC-LSTM} & MAE & 5.95 & 9.16 & 11.58 & 13.46 & 15.00 & 16.31 \\
     & RMSE & 11.91 & 16.98 & 20.82 & 23.69 & 25.97 & 27.82 \\
    \multirow{2}{*}{GWNet} & MAE & 5.76 & 9.64 & 12.79 & 15.30 & 17.28 & 18.81 \\
     & RMSE & 11.27 & 17.57 & 22.31 & 25.75 & 28.52 & 30.48 \\
    \multirow{2}{*}{HGCN} & MAE & 5.70 & 9.09 & 11.73 & 13.84 & 15.55 & 16.95 \\
     & RMSE & 11.18 & 17.09 & 21.33 & 24.52 & 26.99 & 28.94 \\
    \multirow{2}{*}{HighAir} & MAE & 5.50 & 8.52 & 10.81 & 12.50 & 14.00 & 15.09 \\
     & RMSE & 10.80 & 16.10 & 19.85 & 22.70 & 24.91 & 26.40 \\
    \multirow{2}{*}{INNGNN \cite{art_5}} & MAE & 5.48 & \underline{8.49} & \underline{10.67} & \underline{12.34} & \underline{13.72} & \underline{14.91} \\
     & RMSE & 10.70 & \underline{16.03} & \underline{19.66} & \underline{22.29} & \underline{24.37} & \underline{26.11} \\
    \multirow{2}{*}{LSTM} & MAE & 6.50 & 10.26 & 13.18 & 15.52 & 17.40 & 18.91 \\
     & RMSE & 13.85 & 19.26 & 23.52 & 26.83 & 29.46 & 31.55 \\
    \multirow{2}{*}{MegaCRN} & MAE & \underline{5.38} & 8.76 & 10.80 & 12.73 & 14.46 & 16.03 \\
     & RMSE & \underline{10.64} & 16.46 & 19.92 & 22.82 & 25.45 & 27.60 \\
    \multirow{2}{*}{SHARE} & MAE & 5.84 & 9.07 & 11.49 & 13.35 & 14.74 & 15.79 \\
     & RMSE & 11.27 & 16.84 & 20.77 & 23.60 & 25.80 & 27.38 \\
    \multirow{2}{*}{XGBoost} & MAE & 6.85 & 10.89 & 13.99 & 16.27 & 18.14 & 19.56 \\
     & RMSE & 14.25 & 19.80 & 24.72 & 28.14 & 30.63 & 33.44 \\
    \bottomrule
\end{longtable}

As from Tab.~\ref{t:risultati chinese aqi}, the hybrid interpretable neural network and graph neural network model (INNGNN) \cite{art_5} is the most accurate model for all time horizons except the 1 hour horizon, where it is beaten by the meta-graph convolutional recurrent network model (MegaCRN) \cite{MegaCRN}. The authors attribute this result to the fact that in the short term the AQI is strongly influenced by nearby cities, whereas the MegaCRN model is more focused on local features. The INNGNN model, proposed by Ding et al.~in 2023, consists of a temporal and a spatial dependency module. The temporal module employs self-attention and interpretable neural networks (INN) to extract local and global features of time series data, effectively identifying periodicity and trend of the AQI time series. After that, the spatial module has an encoder-decoder architecture where GNN modules extract spatial dependencies across geographic locations, represented by an adjacency matrix based on spatial distance. The MegaCRN model, introduced by Jiang et al.~in 2023 for traffic forecasting, includes a meta-graph learner module and an encoder-decoder spatial module built using a series of convolutional GNNs. The source code for the MegaCRN model can be accessed at \url{https://github.com/deepkashiwa20/MegaCRN}.

\subsection{Finance}
\label{subsection:Finance}
\enquote{Finance} is another emerging theme of study, though not yet widely explored, with just 14 out of 366 selected papers. The difficulty of predicting financial data is a well-established fact. This is due to the inherently complex nature of markets, the influence of geopolitical events on them, and the unpredictable behavior of humans which is often irrational and difficult to predict. This complexity makes financial time series highly volatile, and accurate forecasts and data modeling are essential to develop effective trading strategies. Spatio-temporal GNNs are taken into consideration to model the inter-dependencies between variables together with series dynamics.

\subsubsection{Overview}
The \enquote{Finance} thematic group is relatively narrow in terms of sub-themes. This is in part due to the limited number of published papers. Half of the studies are concentrated on stock price forecasting \cite{art_59, art_60, art_62, art_63, art_65, art_382, art_693}, with the aim of forecasting the future trend of the price of a company. Sometimes the stock prediction problem is seen as a classification task. This is because, due to the stochastic nature and volatility of financial markets, predicting just the direction of price movements (\enquote{up}, \enquote{neutral}, \enquote{down}) can be sometimes more feasible than accurately forecasting price values. In this case, assuming a daily granularity, a rise (\enquote{up}) can be considered to occur when the next trading day's closing price ($p_t$) is at least 0.55\% higher than the previous day's closing price ($p_{t-1}$), whereas if $p_t$ is more than 0.50\% lower than $p_{t-1}$, the price movement can be classified as a \enquote{down}. Otherwise, it is classified as \enquote{neutral}. 
All the selected GNN papers addressing stock prices indeed approach the price forecasting problem from a classification perspective. Another distinctive aspect in this group of papers is that researchers usually evaluate, among others, how their predictions affect the stability of investment returns over a specified period. This is commonly done through trading simulations in the test set, evaluated with metrics such as the Sharpe ratio and overall returns \cite{art_59, art_60, art_65, art_66, art_693}. These simulations require modeling rather than forecasting.

An examination of the selected papers indicates an interest in the application of spatio-temporal GNNs in Finance which starts from 2022. Two of the papers were published in \giornale{IEEE Access} by IEEE, while the remaining ones appeared in various other journals. The only two conference papers of this group were presented at \conferenza{CIKM}. 
    
\subsubsection{Datasets}
The number of datasets in this group is limited. The links to the public datasets referenced in the selected papers are listed in Tab.~\ref{t:link dataset finance}.

\begin{longtable}{R{3cm}R{3cm}p{4.7cm}}
    \caption{List of public datasets in the \enquote{Finance} group and their corresponding links.}
    \label{t:link dataset finance}
    \endfirsthead
    \endhead
    \toprule
    Dataset & Used by & Link \\
    \midrule
    China A-Shares & \cite{art_63}, \cite{art_65}, \cite{art_66}, \cite{art_378} & \url{https://finance.sina.com.cn/stock} \\
    CSI 100 index & \cite{art_62} & \url{https://tushare.pro/} \\
    CSI 300 index & \cite{art_59}, \cite{art_62}, \cite{art_382}, \cite{art_693} & \url{https://global.csmar.com/} \\
    Nordpool dataset & \cite{art_692} & \url{https://www.nordpoolgroup.com/} \\
    S\&P 100 index & \cite{art_62} & \url{https://finance.yahoo.com/} \\
    S\&P 500 index & \cite{art_59}, \cite{art_693} & \url{https://finance.yahoo.com/} \\
    Russell 1K index & \cite{art_62} & \url{https://finance.yahoo.com/} \\
    Russell 3K index & \cite{art_62} & \url{https://finance.yahoo.com/} \\
    \bottomrule
\end{longtable}

All the listed datasets represent collections of stock indices. The most common datasets include the China Securities Index (CSI), the 100 and 500 Standard and Poor's (S\&P) indices, and the 100 and 300 China A-Shares indices. Many studies also include exogenous data, such as other market data, events, and news. News can be processed by some language model like BERT (as in Refs.~\cite{art_59, art_66, art_378}). Most of the data have a granularity of 1 day, and the forecasting horizon is also generally 1 day. One of the selected papers (Ref.~\cite{art_62}) analyses longer time horizons for stock forecasting. As for pre-processing techniques, half of the selected papers use normalization techniques, including min-max normalization, Z-score normalization, and logarithmic normalization.

\subsubsection{Types of models}
In terms of the taxonomy of the proposed GNN models, 7 out of 14 papers utilize attentional GNNs, 6 papers employ convolutional GNNs, and another one (Ref.~\cite{art_692}) does not allow us to specify the classification of the model within the proposed taxonomy. As for the graph structure, some papers use models which learn the graph structure themselves (e.g., Refs.~\cite{art_62, art_64, art_378, art_382}), whereas some others define it a priori, based on the relationships between companies and stocks (e.g., Refs.~\cite{art_63, art_67, art_68}). One paper (Ref.~\cite{art_649}) examines how graph structures change across different observation scales, based on the idea that variable correlations observed over short timescales may differ significantly from those seen over longer periods.

Half of the papers deal with a stock movement classification problem, so the most commonly used loss function is cross entropy. The majority of the papers explicitly indicate that Python is the programming language utilized, with nearly half employing PyTorch, and a few utilizing TensorFlow. Only Ref.~\cite{art_63} provides a link to the source code of the proposed models, which is the knowledge graph with graph convolution neural network (KG-GCN) model, available at \url{https://github.com/Gjl12321/KG_GCN_Stock_Price_Trend_Prediction_System}.

\subsubsection{Benchmark models}
Tab.~\ref{t:benchmark models finance} displays the benchmark models utilized in the \enquote{Finance} group by a minimum of two of the selected papers.

\begin{longtable}{p{2.5cm}p{6.5cm}p{5.2cm}}
    \caption{List of benchmark models in the \enquote{Finance} group divided per category.}
    \label{t:benchmark models finance}
    \endfirsthead
    \endhead
    \toprule
    Category & Model & Used by \\
    \midrule
    \multirow[t]{3}{=}{Mathematical and statistical methods} & Autoregressive integrated moving average (ARIMA) & \cite{art_68}, \cite{art_375} \\
    & Mean reversion (MR) & \cite{art_63}, \cite{art_65} \\
    \midrule 
    \multirow[t]{4}{=}{Non-GNN machine learning methods} & Convolutional neural network (CNN) & \cite{art_59}, \cite{art_64} \\
    & Dual-stage attention-based recurrent neural network (DARNN) & \cite{art_63}, \cite{art_65} \\
    & Feed-forward neural network (FNN) & \cite{art_59}, \cite{art_68}, \cite{art_375}, \cite{art_378} \\
    & Gated recurrent unit (GRU) & \cite{art_60}, \cite{art_68}, \cite{art_382}, \cite{art_693} \\
    & Long-short term memory network (LSTM) & \cite{art_59}, \cite{art_62}, \cite{art_63}, \cite{art_64}, \cite{art_65}, \cite{art_67}, \cite{art_68}, \cite{art_375}, \cite{art_693} \\
    & Support vector machine (SVM) & \cite{art_64}, \cite{art_68}, \cite{art_375}, \cite{art_378} \\
    & Transformer \cite{attention} & \cite{art_382}, \cite{art_693} \\
    \midrule 
    \multirow[t]{1}{=}{GNN methods} & Graph convolutional network (GCN) & \cite{art_59}, \cite{art_60}, \cite{art_62}, \cite{art_64}, \cite{art_68}, \cite{art_382} \\
    & Financial graph attention network (FinGAT) \cite{FinGAT} & \cite{art_60}, \cite{art_62} \\
    & Hierarchical graph attention network (HATS) \cite{HATS} & \cite{art_62}, \cite{art_63}, \cite{art_65}, \cite{art_693} \\
    & Multivariate time series forecasting with graph neural network (MTGNN) \cite{art_655} & \cite{art_62}, \cite{art_382} \\
    & Spatiotemporal hypergraph convolution network (STHGCN) \cite{STHGCN} & \cite{art_63}, \cite{art_65} \\
    & Temporal graph convolution (TGC) \cite{TGC} & \cite{art_63}, \cite{art_693} \\
    \bottomrule
\end{longtable}

It is evident that there is no consensus regarding the most appropriate benchmark model to use, especially of GNN type. The limited number of GNN-based benchmarks can be attributed to the fact that GNN models have not been extensively explored in the financial domain. The most common benchmark is the LSTM network, which is considered an effective approach for processing long term information in time series. The most commonly used GNN model is a vanilla graph convolutional network (GCN), followed by a more complex hierarchical graph attention network (HATS) proposed by Raehyun et al.~in 2019 \cite{HATS}. The HATS model was proposed for predicting stock movements using relational data. It aggregates information from different types of relationships among data and incorporates this information into each stock representation. From an architectural point of view, the model starts with employing LSTM and GRU networks as feature extraction modules. Next, a hierarchical GNN-based architecture is designed to capture the significance of neighboring nodes and relation type across different layers. Finally, the node representations are fed into a task-specific module, which can be implemented as a simple linear transformation layer. The source code for the HATS model is available at \url{https://github.com/dmis-lab/hats}.
    
\subsubsection{Results}
The accuracy of the proposed classification models is primarily evaluated in terms of accuracy, precision and F1 score. The only two papers with results which can be compared are Refs.~\cite{art_63} and \cite{art_65}, which study 758 frequently traded stocks collected from the A-share market in China from January 2013 to December 2019 (available at \url{https://finance.sina.com.cn/stock}). These two papers compare the results obtained by using 5, 10 and 20 trading days as records. For sake of simplicity, a comparison of the accuracy of the models in the two papers with 20 days only is provided in Tab.~\ref{t:risultati a-share}.

\begin{longtable}{l*{4}{c}}
    \caption{Comparison of the average accuracy of the different models on the A-share market in China with 20 training days as records, expressed in terms of accuracy, precision, recall, and F1 score. The best results are underlined.}
    \label{t:risultati a-share}
    \endfirsthead
    \endhead
    \toprule
    Model & Accuracy & Precision & Recall & F1 score \\
    \midrule 
    DARNN & 38.41\% & 37.99\% & 39.24\% & 38.60\% \\
    GCN+LSTM & 37.30\% & 39.28\% & 34.16\% & 36.54\% \\
    HATS & 38.85\% & 38.70\% & 35.06\% & 36.78\% \\
    HGTAN & 40.02\% & 41.77\% & 39.03\% & 40.32\% \\
    KG-GCN \cite{art_63} & \underline{51.38\%} & \underline{65.72\%} & 31.45\% & 39.27\% \\
    LSTM & 35.03\% & 36.43\% & 34.23\% & 35.20\% \\
    MOM & 35.73\% & 35.19\% & 32.82\% & 33.96\% \\
    MONEY \cite{art_65} & 39.90\% & 43.92\% & \underline{40.61\%} & \underline{42.20\%} \\
    MR & 35.32\% & 38.03\% & 33.60\% & 35.68\% \\
    SFM & 34.54\% & 26.93\% & 33.32\% & 29.49\% \\
    STHGCN & 38.45\% & 37.22\% & 32.82\% & 34.87\% \\
    TGC & 37.81\% & 36.96\% & 34.49\% & 35.67\% \\
    \bottomrule
\end{longtable}

The two best models are the knowledge graph and graph convolution neural network (KG-GCN) \cite{art_63} and the stock price movement prediction via convolutional network with adversarial hypergraph model (MONEY) \cite{art_65}, depending on the metrics considered. The KG-GCN model, proposed by Wang et al.~in 2023, first constructs a knowledge graph to represent the semantic relationships and quantify correlations between stocks. Then a community detection module based on GCN node embeddings and K-means clustering is used to group similar stocks together. Finally, the historical prices of similar stocks are fed into LSTM and GRU networks to predict stock price trends. The MONEY model, introduced by Sun et al.~in 2023, is composed of a convolutional GNN and an hypergraph convolution network to capture complex relations among similar stocks.

\subsection{Health}
\label{subsection:Health}
The fourth thematic group of papers is \enquote{Health}, which includes 23 out of 366 publications. These papers focus on critical aspects of health monitoring, disease modeling, and diagnostic tools, with particular emphasis on the spread and diagnosis of disease. All the papers working with functional magnetic resonance imaging (fMRI) data were excluded from the analysis, as the dataset was generated as image sequences rather than time series. This key difference in data structure makes them less relevant to the spatio-temporal focus of this review.

\subsubsection{Overview}
Applications of GNNs to the \enquote{Health} thematic group have been studied starting from 2021. The reason of this expansion may be attributed to two key factors. The first factor is the growing presence of sensors in hospitals, which enable the collection of multivariate time series signals, which can be used for the timely diagnosis of chronic diseases. The second factor is the spread of the COVID-19 pandemic, which motivated researchers to investigate and to forecast the spread of the virus. Epidemic forecasting is indeed the most common topic among the selected papers. The prediction of epidemics presents two significant challenges. First, epidemics' seasonality can vary greatly over time, and they can manifest with different intensities and durations. Second, occasional pandemics can disrupt established patterns even for years. These complexities make the forecasting task more difficult, and non-GNN machine learning, and GNN models themselves, have been studied to address this issue. Moreover, the spread of large-scale human diseases is closely related to both the mobility patterns of populations and the demographic characteristics of residents in different regions \cite{art_413}. Consequently, the application of GNN models to epidemic dynamics appears to be a suitable approach for analyzing and addressing such data. A common reference model for epidemic prediction is the SIR model \cite{modello_SIR}, an acronym referring to the state of the health of individuals. The three letters of the acronym stand for susceptible (S), infectious (I), and recovered (R). Namely, the SIR model divides the population into individuals who are susceptible to the disease, those who are infected and can spread it, and those who have recovered and are now immune. It uses a system of differential equations to describe how the disease spreads spatially and changes in time. The main advantage of this model is its inherent interpretability; it produces coefficients that are easy to interpret. However, its strong and simplistic assumptions also restrict its forecasting capability \cite{art_714}. In general, some papers use it as benchmark (e.g., Ref.~\cite{art_49, art_413, art_415}).

One study (Ref.~\cite{art_659}) also analyzes how air quality and satellite observations of temperature and humidity relate to COVID-19 mortality and severity.

Other papers in this section focus on different types of time series, such as electroencephalogram (EEG) data for epilepsy detection (Refs.~\cite{art_14, art_658, art_660}) or emotion recognition (Refs.~\cite{art_411, art_412, art_661}). While emotion recognition may not seem directly related to the \enquote{Health} topic, it is included here because it relies on EEG data. The use of EEG in emotion recognition helps to identify more accurately true emotional states, as there physiological signals cannot be consciously controlled in the same way as facial expressions, tone of voice, or body posture \cite{art_411}.

The selected papers were each published in different journals, with one paper per journal, primarily in the field of computer science. In terms of conference venues, \conferenza{KDD} and \conferenza{CIKM} are the two most popular, with three papers each. 

\subsubsection{Datasets}
Tab.~\ref{t:link dataset health} provides a list of the public datasets utilized in the selected papers, along with links for accessing them.

\begin{longtable}{R{3.7cm}R{3cm}p{7.5cm}}
    \caption{List of public datasets in the \enquote{Health} group and their corresponding links.}
    \label{t:link dataset health}
    \endfirsthead
    \endhead
    \toprule
    Dataset & Used by & Link \\
    \midrule
    African COVID-19 dataset & \cite{art_410} & \url{https://www.worldometers.info/coronavirus/} \\
    Bonn dataset & \cite{art_14} & \url{http://dx.doi.org/10.1103/PhysRevE.64.061907} (source paper) \\
    Brazilian COVID-19 data & \cite{art_52} & \url{https://github.com/lcaldeira/GrafoBrasilCovid} \\
    CDC US-State & \cite{art_414}, \cite{art_712} & \url{https://gis.cdc.gov/grasp/fluview/fluportaldashboard.html} \\
    Chickenpox & \cite{art_410}, \cite{art_662} & \url{https://www.kaggle.com/datasets/die9origephit/chickenpox-cases-hungary} \\
    CSSE COVID-19 & \cite{art_659}, \cite{art_713}, \cite{art_714} & \url{https://github.com/CSSEGISandData/COVID-19} \\
    EEG DEAP dataset & \cite{art_412} & \url{http://www.eecs.qmul.ac.uk/mmv/datasets/deap/download.html} \\
    eICU & \cite{art_189} & \url{https://doi.org/10.1038/sdata.2018.178} (source paper) \\
    European COVID-19 dataset & \cite{art_410} & \url{https://www.worldometers.info/coronavirus/} \\
    ILI IDWR Japan-Prefecture & \cite{art_414}, \cite{art_712} & \url{https://tinyurl.com/y5dt7stm} \\
    ILI US-HHS US-Regional & \cite{art_414}, \cite{art_712} & \url{https://tinyurl.com/y39tog3h} \\
    ISRUC-Sleep Subgroup III (ISRUC-S3) & \cite{art_661} & \url{https://sleeptight.isr.uc.pt/} \\
    Japan COVID-19 & \cite{art_415} & \url{https://github.com/swsoyee/2019-ncov-japan} \\
    Montreal archive of sleep studies (MASS-SS3) & \cite{art_661} & \url{http://www.ceams-carsm.ca/en/MASS} \\
    Meta COVID-19 data & \cite{art_414}, \cite{art_662} & \url{https://dataforgood.fb.com/tools/disease-prevention-maps/} \\
    MIMIC-IV & \cite{art_189} & \url{https://physionet.org/content/mimiciv/2.0/} \\
    Nytimes COVID-19 data & \cite{art_50}, \cite{art_413} & \url{https://github.com/nytimes/covid-19-data} \\
    OxCGRT dataset & \cite{art_416} & \url{https://www.bsg.ox.ac.uk/research/covid-19-government-response-tracker} \\
    Parkinson's progression markers initiative (PPMI) & \cite{art_92} & \url{http://www.ppmi-info.org/} \\
    Parkinson speech dataset (PS) & \cite{art_92} & \url{https://doi.org/10.1109/jbhi.2013.2245674} (source paper) \\
    SJTU emotion EEG dataset (SEED) & \cite{art_411}, \cite{art_412}, \cite{art_661} & \url{https://bcmi.sjtu.edu.cn/home/seed/} \\    
    Spanish COVID-19 data & \cite{art_51} & \url{https://cnecovid.isciii.es/} \\
    Spikes and slow waves (SSW) & \cite{art_14} & \url{http://dx.doi.org/10.1109/ICDM50108.2020.00067} (source paper) \\
    TUH EEG seizure corpus & \cite{art_660} & \url{https://isip.piconepress.com/projects/nedc/html/tuh_eeg/} \\
    US COVID-19 Johns Hopkins \footnote{This dataset includes multiple sources.} & \cite{art_415} & \url{https://github.com/CSSEGISandData/COVID-19} \\
    US COVID-19 tracking project & \cite{art_416} & \url{https://covidtracking.com/data/national} \\
    US COVID-19 vaccinations & \cite{art_50} & \url{https://ourworldindata.org/us-states-vaccinations} $^\dagger$ \\
    US influenza & \cite{art_48} & \url{https://github.com/aI-area/DVGSN/tree/main/dataset} \\
    \bottomrule
\end{longtable}

Only few subsets of papers in this thematic group use the same dataset, but none of the proposed models yield comparable results. Most of the datasets are related to COVID-19 and have a daily time granularity. In addition, some papers (Refs.~\cite{art_49, art_52, art_92, art_189, art_413, art_659, art_713}) include exogenous variables that can be either context-related (e.g., geography, population, health conditions, vaccinations, economic factors, air quality), or specific to a single patient (including personal information such as age, gender, and medical conditions). A paper that undertakes an in-depth study of exogenous variables is Ref.~\cite{art_416}. The authors categorize these variables as holiday variables, time categorical variables (e.g., categorical week and month, and the number of days and weeks since the start of the dataset), and policy variables (e.g., school and workplace closing, movement restrictions, face coverings, and vaccination policies). They carry on a significance analysis of these variables, and explore the correlation among the significant ones. The COVID-19 policy dataset (OxCGRT) which includes three years of 23 policy indicators across more than 180 countries is available at \url{https://www.bsg.ox.ac.uk/research/covid-19-government-response-tracker}.

\subsubsection{Types of models}
As for the taxonomy of the proposed models, 18 out of 23 papers use a convolutional GNN architecture. Three papers use a pure attentional model, while two other papers present a hybrid convolutional-attentional architecture. Different approaches are used to define the graph structure. Some papers base their graphs on the geographical distribution of the data (as in Refs.~\cite{art_49, art_52, art_92, art_659, art_662}), and others use models that learn the graph structure directly, as in Refs.~\cite{art_48, art_50, art_415, art_416, art_713}. There is also a paper that uses a variant of the visibility algorithm \cite{art_14}. In half of the cases, the graph is static, while the dynamic graph approach appears more frequently in the papers from 2024.

In terms of programming languages, the community of the \enquote{Health} thematic group uses only Python with PyTorch. As for the choice of the loss functions, MSE is typically used for prediction tasks, and cross entropy is used for classification tasks. Tab.~\ref{t:link codici health} shows the links to the related source codes.
\newpage
\begin{longtable}{p{7cm}p{7.5cm}}
    \caption{List of source codes of the \enquote{Health} models in the review.}
    \label{t:link codici health}
    \endfirsthead
    \endhead
    \toprule
    Model & Link \\
    \midrule
    Backbone-based dynamic spatio-temporal graph neural network (BDSTGNN) \cite{art_415} & \url{https://github.com/JkMaoya/BDSTGNN} \\
    DCRNN for COVID-19 forecasting \cite{art_659} & \url{https://github.com/Covid-19-papers/Recurrent-Graph-Neural-Network-on-COVID-19-and-NASA-Satellite-data} \\
    Deep learning of contagion dynamics on complex networks \cite{art_51} & \url{https://doi.org/10.5281/zenodo.4974521} \\
    Dynamic virtual graph significance networks (DVGSN) \cite{art_48} & \url{https://github.com/aI-area/DVGSN} \\
    Ordinary differential equation in graph neural networks (GNNODENet) \cite{art_413} & \url{https://github.com/xiongzhangxyq/GNNODENet} \\
    Hierarchical spatial-temporal framework for COVID-19 trend forecasting (HierST) \cite{art_714} & \url{https://github.com/shun-zheng/HierST} \\
    Recurrent graph gate fusion transformers (ReGraFT) \cite{art_416} & \url{https://github.com/mfriendly/ReGraFT} \\
    Sparse spectra graph convolutional network (SSGCNet) \cite{art_14} & \url{https://github.com/anonymous2020-source-code/WNFG-SSGCNet-ADMM} \\
    Spatial-temporal graph convolutional network (STGCN) \cite{art_52} & \url{https://github.com/lcaldeira/CovidPandemicForecasting} \\
    Time-aware context-gated graph attention network (T-ContextGGAN) \cite{art_189} & \url{https://github.com/OwlCitizen/TContext-GGAN} \\
    Temporal multiresolution graph neural networks (TMGNN) \cite{art_662} & \url{https://github.com/bachnguyenTE/temporal-mgn} \\
    \bottomrule
\end{longtable}

\subsubsection{Benchmark models}
Tab.~\ref{t:benchmark models health} lists the benchmark models in the \enquote{Health} group that are discussed in at least two papers.

\begin{longtable}{p{2.5cm}p{6.5cm}p{5.2cm}}
    \caption{List of benchmark models in the \enquote{Health} group divided per category.}
    \label{t:benchmark models health}
    \endfirsthead
    \endhead
    \toprule
    Category & Model & Used by \\
    \midrule
    \multirow[t]{3}{=}{Mathematical and statistical methods} & Autoregressive integrated moving average (ARIMA) & \cite{art_48}, \cite{art_49}, \cite{art_52}, \cite{art_410}, \cite{art_414}, \cite{art_415}, \cite{art_662}, \cite{art_712}, \cite{art_713} \\
    & Historical average (HA) & \cite{art_662}, \cite{art_713} \\
    & Naive & \cite{art_52}, \cite{art_662} \\
    & SIR model \cite{modello_SIR} & \cite{art_49}, \cite{art_413}, \cite{art_415} \\
    \midrule 
    \multirow[t]{4}{=}{Non-GNN machine learning methods} & Feed-forward neural network (FNN) & \cite{art_14}, \cite{art_48} \\
    & Bi-hemisphere domain adversarial neural network (Bi-DANN) \cite{Bi-DANN} & \cite{art_411}, \cite{art_412} \\
    & A novel bi-hemispheric discrepancy model for EEG emotion recognition (BiHDM) \cite{BiHDM} & \cite{art_411}, \cite{art_412} \\
    & Convolutional neural network and recurrent neural network with residual module (CNNRNN-Res) \cite{CNNRNN-Res} & \cite{art_414}, \cite{art_712} \\
    & Gated recurrent unit (GRU) & \cite{art_410}, \cite{art_413}, \cite{art_415}, \cite{art_416} \\
    & Long-short term memory network (LSTM) & \cite{art_49}, \cite{art_50}, \cite{art_410}, \cite{art_413}, \cite{art_658}, \cite{art_661}, \cite{art_662}, \cite{art_712}, \cite{art_713} \\
    & Long-short term time series network (LSTNet) \cite{LSTNet} & \cite{art_414}, \cite{art_712} \\
    & MiniRocket \cite{MiniRocket} & \cite{art_658}, \cite{art_660} \\
    & Random forest (RF) & \cite{art_48}, \cite{art_661} \\
    & Recurrent neural network (RNN) & \cite{art_189}, \cite{art_413}, \cite{art_414}, \cite{art_712} \\
    & Recurrent neural network with attention (RNN+Att) & \cite{art_414}, \cite{art_712} \\
    & Support vector machine (SVM) & \cite{art_189}, \cite{art_411}, \cite{art_661} \\
    & Time-Series representation learning framework via Temporal and Contextual Contrasting (TS-TCC) \cite{TS-TCC} & \cite{art_658}, \cite{art_660} \\
    & Time series transformer (TST) \cite{TST} & \cite{art_50}, \cite{art_660} \\
    \midrule 
    \multirow[t]{1}{=}{GNN methods} & Cross-location attention based graph neural network (ColaGNN) \cite{art_712} & \cite{art_414}, \cite{art_415}, \cite{art_416} \\
    & CovidGNN \cite{CovidGNN} & \cite{art_415}, \cite{art_713} \\
    & Diffusion convolutional recurrent neural network (DCRNN) \cite{DCRNN} & \cite{art_49}, \cite{art_413}, \cite{art_414}, \cite{art_416}, \cite{art_712} \\
    & Dynamical graph convolutional neural networks (DGCNN) \cite{DGCNN} & \cite{art_411}, \cite{art_412}, \cite{art_661} \\    
    & Graph attention network (GAT) \cite{GAT} & \cite{art_48}, \cite{art_413} \\
    & Graph convolutional network (GCN) & \cite{art_49}, \cite{art_413} \\
    & Regularized graph neural networks (RGNN) \cite{RGNN} & \cite{art_411}, \cite{art_412} \\
    & Spatio-temporal attention network for pandemic prediction (STAN) \cite{STAN} & \cite{art_413}, \cite{art_415}, \cite{art_713} \\
    & Spatio-temporal graph convolutional network (STGCN) \cite{STGCN} & \cite{art_413}, \cite{art_712} \\
    & Spatial-temporal graph ODE networks (STGODE) \cite{art_676} & \cite{art_413}, \cite{art_415} \\
    \bottomrule
\end{longtable}

There are no particularly outstanding benchmark models, and the ones categorized as GNNs appear only in papers from 2024. This is due to the fact that in the past years GNNs have not been extensively explored in the context of health. The most widely used GNN-based benchmark is the diffusion convolutional recurrent neural network (DCRNN), proposed by Li et al.~in 2017 for traffic forecasting. The model aims to capture spatial dependencies using bidirectional random walks on the graph (whose adjacency matrix is based on node proximity), and temporal dependencies using an encoder-decoder architecture with scheduled sampling. The source code for the DCRNN model is available at \url{https://github.com/liyaguang/DCRNN}.
    
\subsubsection{Results}
In each paper, the authors claim that the proposed GNN-based model outperforms the benchmarks. The most common error metrics are the MAE and the MSE for forecasting problems and accuracy, sensitivity and specificity for classification problems. The forecasting horizon spans from 1 day to more than 10 weeks ahead. A comparison of the accuracy of the different models is not possible due to the lack of common benchmark datasets.

\subsection{Mobility}
\label{subsection:Mobility}
The thematic group that includes the largest number of papers identified by our query (102 out of 366) is \enquote{Mobility}, which regards the movement of people. It includes urban traffic, air travel, and bicycle demand, among other applications. Given the large number of articles in this group, a more detailed comparison of benchmarks, models, and results can be provided.

\subsubsection{Overview}
Early traffic prediction is essential for improving the efficiency of transportation systems, helping drivers plan their trips more effectively, and, in the context of cities, preventing urban congestion. The advent of smart cities infrastructures and transportation systems has contributed to the collection of rich data from road sensors. Making accurate traffic forecasts is challenging, due to constantly changing traffic patterns over time and space, as well as the influence of external factors such as weather conditions and special events. Unlike statistical and non-GNN machine learning models, GNNs can be particularly effective for traffic forecasting due to their ability to model complex relationships in spatial and temporal domain. The increasing number of papers published each year in this thematic group reflects both the growing research interest in this field and the growing confidence in GNN models. The most commonly investigated areas regard urban traffic (approximately $1/4$ of the papers), especially urban traffic flow and urban traffic speed. Among the journals where the selected papers were published the most common are \giornale{Applied Intelligence} by Springer, \giornale{ACM Transactions on Knowledge Discovery from Data}, both by the Association for Computing Machinery, and \giornale{IEEE Transactions on Intelligent Transportation Systems} by the Institute of Electrical and Electronics Engineers, with 6 out of 66 papers each. The three conference venues with the highest number of papers are \conferenza{CIKM} with 13 papers, \conferenza{SIGSPATIAL} with 9 papers, and \conferenza{KDD} with 8 papers. 

\subsubsection{Datasets}
Tab.~\ref{t:link dataset traffico} provides a list of the public datasets mentioned in the selected papers, along with the links to access them.

\begin{longtable}{R{3.7cm}R{3cm}p{7.5cm}}
    \caption{List of public datasets in the \enquote{Mobility} group and their corresponding links.}
    \label{t:link dataset traffico}
    \endfirsthead
    \endhead
    \toprule
    Dataset & Used by & Link \\
    \midrule
    Baoding dataset & \cite{art_666} & \url{https://github.com/usail-hkust/ASeer} \\
    BRN+ & \cite{art_721} & \url{https://www.microsoft.com/en-us/research/publication/t-drive-trajectory-data-sample/} \\
    Bruxelles transport data & \cite{art_729} & \url{https://www.kaggle.com/datasets/giobbu/belgium-obu} \\
    Bureau of transportation statistics (BTS) & \cite{art_434}, \cite{art_443} & \url{https://www.transtats.bts.gov/} \\
    CAAC flights & \cite{art_129} & \url{http://www.caac.gov.cn/index.html} \\
    Capital bike share & \cite{art_446} & \url{https://ride.capitalbikeshare.com/system-data} $^\dagger$ \\
    CarPark1918 & \cite{art_653} & \url{https://github.com/JIANGYUE61610306/SAGDFN/tree/main/data/model} \\
    CD-HW & \cite{art_126} & \url{https://doi.org/10.1109/ACCESS.2020.3027375} (source paper) \\
    DiDi Beijing & \cite{art_102}, \cite{art_447} & \url{https://github.com/didi/TrafficIndex} \\
    DiDi Chengdu & \cite{art_93}, \cite{art_98}, \cite{art_667}, \cite{art_694}, \cite{art_730}, \cite{art_736} & \url{https://gaia.didichuxing.com/} $^\dagger$ \\
    Didi Shenzhen & \cite{art_667}, \cite{art_694} & \url{https://outreach.didichuxing.com/app-outreach/CTIE} \\
    Divvy bike Chicago & \cite{art_446}, \cite{art_668} & \url{https://ride.divvybikes.com/system-data} $^\dagger$ \\
    Highways England & \cite{art_663} & \url{http://tris.highwaysengland.co.uk/detail/trafficflowdata} \\
    Los-loop & \cite{art_103} & \url{https://github.com/lehaifeng/T-GCN/tree/master/data} \\
    Lyft bike dataset & \cite{art_446} & \url{https://www.lyft.com/bikes/bay-wheels/system-data} \\
    METR-LA & \cite{art_93}, \cite{art_99}, \cite{art_100}, \cite{art_109}, \cite{art_110}, \cite{art_113}, \cite{art_116}, \cite{art_120}, \cite{art_122}, \cite{art_126}, \cite{art_128}, \cite{art_408}, \cite{art_427}, \cite{art_430}, \cite{art_433}, \cite{art_435}, \cite{art_438}, \cite{art_441}, \cite{art_442}, \cite{art_444}, \cite{art_652}, \cite{art_653}, \cite{art_667}, \cite{art_669}, \cite{art_670}, \cite{art_672}, \cite{art_673}, \cite{art_694}, \cite{art_719}, \cite{art_720}, \cite{art_723}, \cite{art_724}, \cite{art_726}, \cite{art_733}, \cite{art_738}, \cite{art_740} & \url{https://www.kaggle.com/datasets/annnnguyen/metr-la-dataset} \\
    NE-BJ & \cite{art_100} & \url{https://github.com/tsinghua-fib-lab/Traffic-Benchmark} \\
    NYC-Bike & \cite{art_94}, \cite{art_96}, \cite{art_114}, \cite{art_446}, \cite{art_451}, \cite{art_682}, \cite{art_715}, \cite{art_718} & \url{https://citibikenyc.com/system-data} \\
    NYC-Taxi & \cite{art_96}, \cite{art_682}, \cite{art_715}, \cite{art_718}, \cite{art_723}, \cite{art_734} & \url{https://www1.nyc.gov/site/tlc/about/tlc-trip-record-data.page} \\
    PeMS03 & \cite{art_108}, \cite{art_113}, \cite{art_115}, \cite{art_117}, \cite{art_125}, \cite{art_130}, \cite{art_419}, \cite{art_435}, \cite{art_445}, \cite{art_453}, \cite{art_671}, \cite{art_675}, \cite{art_676}, \cite{art_682}, \cite{art_707}, \cite{art_715}, \cite{art_720}, \cite{art_721}, \cite{art_732}, \cite{art_735} & \url{https://www.kaggle.com/datasets/elmahy/pems-dataset} \\
    PeMS04 (PeMSD4) & \cite{art_97}, \cite{art_99}, \cite{art_104}, \cite{art_108}, \cite{art_113}, \cite{art_115}, \cite{art_117}, \cite{art_122}, \cite{art_125}, \cite{art_130}, \cite{art_131}, \cite{art_132}, \cite{art_406}, \cite{art_408}, \cite{art_419}, \cite{art_424}, \cite{art_433}, \cite{art_435}, \cite{art_441}, \cite{art_445}, \cite{art_453}, \cite{art_652}, \cite{art_663}, \cite{art_671}, \cite{art_672}, \cite{art_675}, \cite{art_676}, \cite{art_683}, \cite{art_707}, \cite{art_715}, \cite{art_716}, \cite{art_719}, \cite{art_720}, \cite{art_721}, \cite{art_725}, \cite{art_732}, \cite{art_735} & \url{https://www.kaggle.com/datasets/elmahy/pems-dataset} \\
    PeMS07 (PeMSD7) & \cite{art_108}, \cite{art_110}, \cite{art_113}, \cite{art_117}, \cite{art_125}, \cite{art_130}, \cite{art_406}, \cite{art_422}, \cite{art_424}, \cite{art_427}, \cite{art_432}, \cite{art_435}, \cite{art_438}, \cite{art_441}, \cite{art_444}, \cite{art_445}, \cite{art_453}, \cite{art_670}, \cite{art_671}, \cite{art_675}, \cite{art_676}, \cite{art_707}, \cite{art_715}, \cite{art_719}, \cite{art_720}, \cite{art_725}, \cite{art_732}, \cite{art_735} & \url{https://www.kaggle.com/datasets/elmahy/pems-dataset} \\
    PeMS08 (PeMSD8) & \cite{art_97}, \cite{art_99}, \cite{art_108}, \cite{art_115}, \cite{art_117}, \cite{art_122}, \cite{art_125}, \cite{art_130}, \cite{art_131}, \cite{art_132}, \cite{art_406}, \cite{art_408}, \cite{art_424}, \cite{art_432}, \cite{art_433}, \cite{art_435}, \cite{art_441}, \cite{art_444}, \cite{art_445}, \cite{art_453}, \cite{art_663}, \cite{art_671}, \cite{art_672}, \cite{art_675}, \cite{art_676}, \cite{art_682}, \cite{art_683}, \cite{art_707}, \cite{art_715}, \cite{art_716}, \cite{art_720}, \cite{art_721}, \cite{art_725}, \cite{art_730}, \cite{art_732}, \cite{art_733}, \cite{art_735} & \url{https://www.kaggle.com/datasets/elmahy/pems-dataset} \\
    PEMS-BAY & \cite{art_100}, \cite{art_109}, \cite{art_110}, \cite{art_116}, \cite{art_120}, \cite{art_122}, \cite{art_406}, \cite{art_408}, \cite{art_419}, \cite{art_420}, \cite{art_422}, \cite{art_430}, \cite{art_433}, \cite{art_435}, \cite{art_441}, \cite{art_442}, \cite{art_444}, \cite{art_652}, \cite{art_667}, \cite{art_669}, \cite{art_670}, \cite{art_672}, \cite{art_673}, \cite{art_694}, \cite{art_719}, \cite{art_720}, \cite{art_724}, \cite{art_733}, \cite{art_738}, \cite{art_740} & \url{https://zenodo.org/records/4263971} \\
    Q-Traffic & \cite{art_124} & \url{https://github.com/JingqingZ/BaiduTraffic\#Dataset} \\
    SafeGraph & \cite{art_728} & \url{https://www.safegraph.com/} \\
    Seattle loop & \cite{art_441}, \cite{art_726} & \url{https://github.com/zhiyongc/Seattle-Loop-Data} \\
    UVDS & \cite{art_128} & \url{https://doi.org/10.1007/978-3-030-73280-6_6} (source paper) \\
    TaxiBJ & \cite{art_126}, \cite{art_451}, \cite{art_701} & \url{https://www.microsoft.com/en-us/research/publication/forecasting-citywide-crowd-flows-based-big-data/} \\
    Taxi Shenzhen & \cite{art_103}, \cite{art_135} & \url{https://github.com/lehaifeng/T-GCN/tree/master/data} \\ 
    TLC Taxi & \cite{art_119} & \url{https://www.nyc.gov/site/tlc/about/tlc-trip-record-data.page} \\
    Urban-core & \cite{art_441} & \url{https://topis.seoul.go.kr/} \\
    Xiecheng delay & \cite{art_443} & \url{https://pan.baidu.com/s/1dEPyMGh#list/path$=$%2F} \\
    Zhuzhou dataset & \cite{art_666} & \url{https://github.com/usail-hkust/ASeer} \\
    \bottomrule
\end{longtable}

The most popular datasets used in this group are those  provided by the California Transportation Agency Performance Measurement System (PeMS, available at \url{http://pems.dot.ca.gov/}), which include various versions differing in time range, spatial coverage, and number of sensors, and the one called METR-LA, collected from loop detectors along the Los Angeles County highway. The number of sensors is in the hundreds, and such is also the number of nodes in the graph. For a more detailed overview of the benchmark datasets in the traffic domain, see Ref.~\cite{review_traffico_con_dataset}. Other common public traffic datasets can be found at \url{https://web.archive.org/web/20241212092408/https://paperswithcode.com/task/traffic-prediction}. As to external data, the most commonly used are data on weather conditions, time information (time of day and day of week) and events (e.g., holidays), as well as information about the road network, like points of interest.

The majority of the papers utilize data with a 5-minute granularity, and provide forecasts for horizons ranging from 5 minutes to 1 hour ahead. In addition to imputation of missing values, in almost half of the cases the pre-processing of the data includes data normalization (min-max or Z-score normalization) and the selection of time windows regarding recent, daily and weekly data. 

\subsubsection{Types of models}
As for the proposed models, 53 out of 102 belong to the pure convolutional GNN paradigm, followed by 21 hybrid convolutional-attention architectures, 20 attentional GNNs, and 1 recurrent GNN. Two papers use both convolutional and attentional models, and other five (Refs.~\cite{art_438, art_451, art_676, art_725, art_740}) do not fall into a specific class. Regarding the definition of the graph structure, most of the papers discuss it in detail. Almost half of the papers published before 2024 are based on a pre-defined adjacency matrix, while the majority of papers since 2024 propose models that learn the adjacency matrix on their own, either from scratch or based on a pre-initialization (e.g., Refs.~\cite{art_108, art_444}). Here, more than in any other thematic group, there are multiple ways to define the adjacency matrix. This can be made by taking into account the structure of the road network, the actual flow of people, or a combination of different sources of information. More in detail, a first group of papers uses road connectivity (e.g., Refs.~\cite{art_104, art_120, art_135, art_431, art_664, art_707}); second group uses spatial distance (e.g., Refs.~\cite{art_93, art_96, art_99, art_110, art_128, art_132, art_434, art_454}); a third group incorporates extra information about the so-called points of interest (such as Refs.~\cite{art_94, art_96}); a fourth group considers some similarity measures between the time series (e.g., Refs.~\cite{art_97, art_117, art_122, art_435, art_445}), such as the correlation coefficient or cosine similarity; a fifth group directly uses the number of people traveling on the road at a given time step (e.g., Refs.~\cite{art_94, art_102, art_129, art_133}), or a combination of different aspects (e.g., Refs.~\cite{art_420, art_444}). For a more detailed overview of graph construction techniques in traffic flow prediction models, the reader is referred to Ref.~\cite{Fan2025}.

The two most common loss functions used to train the models are the MAE and the MSE. Few papers consider a combination of loss functions, especially when dealing with a model that has to learn the graph structure by itself (Refs.~\cite{art_126, art_130, art_446, art_652, art_667}). Not all papers specify the language or library used for writing the code. Among those that do that, Python is the most common language, with the majority using PyTorch, whereas only a few researchers use TensorFlow. In addition, a few papers provide a link to the source code for the proposed model. The papers that provide such links are listed in Tab.~\ref{t:link codici traffico}.

\begin{longtable}{p{7cm}p{7.5cm}}
    \caption{List of source codes of the \enquote{Mobility} models in the review.}
    \label{t:link codici traffico}
    \endfirsthead
    \endhead
    \toprule
    Model & Link \\
    \midrule 
    3-Dimensional graph convolution network (3DGCN) \cite{art_94} & \url{https://github.com/FIBLAB/3D-DGCN} \\
    Adaptive graph convolutional recurrent network (AGCRN) \cite{art_683} & \url{https://github.com/LeiBAI/AGCRN} \\
    Adaptive generalized PageRank graph neural network (AGP-GNN) \cite{art_116} & \url{https://github.com/guoxiaoyuatbjtu/agp-gnn} \\
    Adaptive multigraph convolutional networks (AMGCN) \cite{art_444} & \url{https://github.com/hfimmortal/AMGCN} \\
    Asynchronous spatio-temporal graph convolutional network (ASeer) \cite{art_666} & \url{https://github.com/usail-hkust/ASeer} \\
    Automated dilated spatio-temporal synchronous graph network (Auto-DSTSGN) \cite{art_108} & \url{https://github.com/jinguangyin/Auto-DSTSGN} \\
    BigST \cite{art_447} & \url{https://github.com/usail-hkust/BigST} \\
    Dynamic multi-view graph neural network for citywide traffic inference (CTVI+) \cite{art_97} & \url{https://github.com/dsj96/TKDD} \\
    Decoupled dynamic spatial-temporal graph neural network (D2STGNN) \cite{art_672} & \url{https://github.com/zezhishao/D2STGNN} \\
    Domain adversarial spatial-temporal network (DastNet) \cite{art_725} & \url{https://github.com/YihongT/DASTNet} \\
    Dynamic graph convolutional recurrent network (DGCRN) \cite{art_100} & \url{https://github.com/tsinghua-fib-lab/Traffic-Benchmark} \\
    Deep Kalman filtering network (DKFN) \cite{art_726} & \url{https://github.com/Fanglanc/DKFN} \\
    Dynamic and multi-faceted spatio-temporal graph convolution network (DMSTGCN) \cite{art_663} & \url{https://github.com/liangzhehan/DMSTGCN} \\
    Dynamic spatial-temporal aware graph neural network (DSTAGNN) \cite{art_675} & \url{https://github.com/SYLan2019/DSTAGNN} \\
    Graph bootstrap (Gboot) \cite{art_715} & \url{https://github.com/wangzz-yyzz/Gboot} \\
    Hierarchical information enhanced spatio-temporal (HIEST) \cite{art_733} & \url{https://github.com/VAN-QIAN/CIKM23-HIEST} \\
    Multi-step dependency relation (MSDR) \cite{art_682} & \url{https://github.com/dcliu99/MSDR} \\
    Multi-range spatial-temporal transformer network (MultiSPANS) \cite{art_716} & \url{https://github.com/SELGroup/MultiSPANS} \\
    Progressive graph convolutional network (PGCN) \cite{art_441} & \url{https://github.com/yuyolshin/PGCN} \\
    Scalable adaptive graph diffusion forecasting network (SAGDFN) \cite{art_653} & \url{https://github.com/JIANGYUE61610306/SAGDFN} \\
    Scale-aware neural architecture search framework for MTS forecasting (SNAS4MTF) \cite{art_408} & \url{https://github.com/shangzongjiang/SNAS4MTF} \\
    Spatio-temporal adaptive gated graph convolution network (STAG-GCN) \cite{art_736} & \url{https://github.com/RobinLu1209/STAG-GCN} \\
    Spatial-temporal convolutional graph attention network (ST-CGA) \cite{art_734} & \url{https://github.com/jbdj-star/cga} \\
    Spatio-temporal causal graph attention network (STCGAT) \cite{art_125} & \url{https://github.com/zsqZZU/STCGAT/tree/v1.0.0} \\
    STGNN enhanced by a scalable time series pre-training model (STEP) \cite{art_652} & \url{https://github.com/zezhishao/STEP} \\
    Spatial-temporal graph ordinary differential equation networks (STGODE) \cite{art_676} & \url{https://github.com/square-coder/STGODE} \\
    Spatio-temporal graph mixformer (STGM) \cite{art_110} & \url{https://github.com/mouradost/stgm} \\
    Spatio-temporal graph prompting (STGP) \cite{art_694} & \url{https://github.com/hjf1997/STGP} \\
    Space-time separable graph convolutional network (STSGCN) \cite{art_443} & \url{https://github.com/FraLuca/STSGCN} \\
    Spatio-temporal network with discrete wavelet transform (STWave) \cite{art_671} & \url{https://github.com/LMissher/STWave} \\
    Sparse unstructured spatio temporal reconstruction (SUSTeR) \cite{art_740} & \url{https://github.com/ywoelker/SUSTeR} \\
    Transferable graphs for traffics (TransGTR) \cite{art_670} & \url{https://github.com/KL4805/TransGTR/} \\
    Two-level resolution neural network (TwoResNet) \cite{art_430} & \url{https://github.com/semink/TwoResNet} \\
    Wavelet dynamic spatiotemporal aware graph neural network (W-DSTAGNN) \cite{art_419} & \url{https://github.com/yash-jakhmola/w-dstagnn} \\
    \bottomrule
\end{longtable}

\subsubsection{Benchmark models}
Tab.~\ref{t:benchmark models traffico} lists the benchmark models used within the papers in the \enquote{Mobility} thematic group. These benchmarks are categorized into mathematical and statistical methods, non-GNN machine learning models, and GNN models, and only those appearing in more than two papers are included in the table.

\begin{longtable}{p{2.5cm}p{6.5cm}p{5.2cm}}
    \caption{List of benchmark models in the \enquote{Mobility} group divided per category.}
    \label{t:benchmark models traffico}
    \endfirsthead
    \endhead
    \toprule
    Category & Model & Used by \\
    \midrule
    \multirow[t]{6}{=}{Mathematical and statistical methods} & Autoregressive integrated moving average (ARIMA) 
    & \cite{art_93}, \cite{art_94}, \cite{art_96}, \cite{art_98}, \cite{art_100}, \cite{art_103}, \cite{art_106},  \cite{art_109}, \cite{art_116}, \cite{art_126}, \cite{art_129}, \cite{art_132}, \cite{art_135}, \cite{art_406}, \cite{art_408}, \cite{art_419}, \cite{art_421}, \cite{art_424}, \cite{art_430}, \cite{art_433}, \cite{art_438}, \cite{art_444}, \cite{art_446}, \cite{art_451}, \cite{art_453}, \cite{art_653}, \cite{art_663}, \cite{art_667}, \cite{art_669}, \cite{art_670}, \cite{art_671}, \cite{art_673}, \cite{art_676}, \cite{art_694}, \cite{art_719}, \cite{art_720}, \cite{art_724}, \cite{art_726}, \cite{art_729}, \cite{art_730}, \cite{art_734}, \cite{art_735}, \cite{art_736} \\
    & Historical average (HA) & \cite{art_93}, \cite{art_94}, \cite{art_96}, \cite{art_98}, \cite{art_100}, \cite{art_102}, \cite{art_103}, \cite{art_109}, \cite{art_110}, \cite{art_120}, \cite{art_125}, \cite{art_131}, \cite{art_132}, \cite{art_135}, \cite{art_406}, \cite{art_408}, \cite{art_420}, \cite{art_424}, \cite{art_433}, \cite{art_435}, \cite{art_438}, \cite{art_441}, \cite{art_443}, \cite{art_446}, \cite{art_447}, \cite{art_451}, \cite{art_453}, \cite{art_652}, \cite{art_663}, \cite{art_664}, \cite{art_666}, \cite{art_667}, \cite{art_668}, \cite{art_669}, \cite{art_671}, \cite{art_672}, \cite{art_673}, \cite{art_682}, \cite{art_683}, \cite{art_694}, \cite{art_724}, \cite{art_725}, \cite{art_726}, \cite{art_728}, \cite{art_729}, \cite{art_730}, \cite{art_736}, \cite{art_739} \\
    & $k$-nearest neighbors (KNN) & \cite{art_97}, \cite{art_694}, \cite{art_736} \\
    & Naive & \cite{art_666}, \cite{art_728}, \cite{art_739} \\
    & Vector autoregression (VAR) 
    & \cite{art_99}, \cite{art_100}, \cite{art_126}, \cite{art_131}, \cite{art_132}, \cite{art_406}, \cite{art_419}, \cite{art_424}, \cite{art_435}, \cite{art_443}, \cite{art_445}, \cite{art_447}, \cite{art_453}, \cite{art_652}, \cite{art_653}, \cite{art_663}, \cite{art_669}, \cite{art_671}, \cite{art_672}, \cite{art_683}, \cite{art_716}, \cite{art_720}, \cite{art_732}, \cite{art_735}, \cite{art_739} \\
    \midrule 
    \multirow[t]{13}{=}{Non-GNN machine learning methods} & Adaptive recurrent neural network (AdaRNN) \cite{AdaRNN} & \cite{art_667}, \cite{art_694} \\
    & Convolutional long-short term memory network (ConvLSTM) \cite{ConvLSTM} & \cite{art_98}, \cite{art_446}, \cite{art_728}, \cite{art_739} \\
    & Deep multi-view spatial-temporal network (DMVST-Net) \cite{DMVST-Net} & \cite{art_96}, \cite{art_451}, \cite{art_734} \\
    & Feed-forward neural network (FNN) & \cite{art_93}, \cite{art_97}, \cite{art_100}, \cite{art_105}, \cite{art_114}, \cite{art_120}, \cite{art_121}, \cite{art_124}, \cite{art_126}, \cite{art_408}, \cite{art_420}, \cite{art_431}, \cite{art_438}, \cite{art_720}, \cite{art_731}, \cite{art_736}, \cite{art_739} \\
    & Gated recurrent unit (GRU) & \cite{art_98}, \cite{art_99}, \cite{art_106}, \cite{art_121}, \cite{art_124}, \cite{art_131}, \cite{art_132}, \cite{art_135}, \cite{art_425}, \cite{art_666}, \cite{art_683}, \cite{art_725}, \cite{art_730} \\
    & Long-short term memory network (LSTM) & \cite{art_93}, \cite{art_96}, \cite{art_98}, \cite{art_99}, \cite{art_100}, \cite{art_106}, \cite{art_108}, \cite{art_109}, \cite{art_113}, \cite{art_115}, \cite{art_116}, \cite{art_120}, \cite{art_121}, \cite{art_122}, \cite{art_125}, \cite{art_126}, \cite{art_130}, \cite{art_131}, \cite{art_132}, \cite{art_406}, \cite{art_408}, \cite{art_419}, \cite{art_420}, \cite{art_421}, \cite{art_424}, \cite{art_425}, \cite{art_430}, \cite{art_433}, \cite{art_435}, \cite{art_438}, \cite{art_441}, \cite{art_442}, \cite{art_443}, \cite{art_444}, \cite{art_445}, \cite{art_453}, \cite{art_652}, \cite{art_653}, \cite{art_664}, \cite{art_666}, \cite{art_668}, \cite{art_669}, \cite{art_671}, \cite{art_672}, \cite{art_673}, \cite{art_675}, \cite{art_682}, \cite{art_707}, \cite{art_716}, \cite{art_719}, \cite{art_720}, \cite{art_722}, \cite{art_724}, \cite{art_726}, \cite{art_730}, \cite{art_731}, \cite{art_732}, \cite{art_734}, \cite{art_735}, \cite{art_736}, \cite{art_739} \\
    & Long-short term time series network (LSTNet) \cite{LSTNet} & \cite{art_102}, \cite{art_110} \\
    & Multitask deep-learning (MDL) \cite{MDL} & \cite{art_451}, \cite{art_723} \\
    & Recurrent neural network (RNN) & \cite{art_451}, \cite{art_644}, \cite{art_668}, \cite{art_669} \\
    & Seq2Seq \cite{Seq2seq} & \cite{art_105}, \cite{art_124}, \cite{art_451} \\
    & Spatial-temporal dynamic network (STDN) \cite{STDN} & \cite{art_451}, \cite{art_734} \\
    & Spatial-temporal identity (STID) \cite{STID} & \cite{art_116}, \cite{art_422}, \cite{art_433}, \cite{art_434}, \cite{art_673}, \cite{art_718}, \cite{art_722} \\
    & Spatial and temporal normalization for multivariate time series forecasting (ST-Norm) \cite{ST-Norm} & \cite{art_116}, \cite{art_117}, \cite{art_406}, \cite{art_422} \\
    & Spatio-temporal residual networks (ST-ResNet) \cite{ST-ResNet} & \cite{art_451}, \cite{art_734} \\
    & Support vector regression (SVR) & \cite{art_99}, \cite{art_100}, \cite{art_103}, \cite{art_109}, \cite{art_119}, \cite{art_129}, \cite{art_130}, \cite{art_135}, \cite{art_408}, \cite{art_419}, \cite{art_421}, \cite{art_435}, \cite{art_444}, \cite{art_445}, \cite{art_453}, \cite{art_652}, \cite{art_653}, \cite{art_669}, \cite{art_671}, \cite{art_672}, \cite{art_716}, \cite{art_718}, \cite{art_720}, \cite{art_725}, \cite{art_726}, \cite{art_729}, \cite{art_732}, \cite{art_733}, \cite{art_734} \\
    & Temporal convolutional network (TCN) \cite{TCN} & \cite{art_446}, \cite{art_453}, \cite{art_666}, \cite{art_671}, \cite{art_675}, \cite{art_707}, \cite{art_721} \\
    & Temporal 2D-variation modeling for general time series analysis (TimesNet) \cite{TimesNet} & \cite{art_434}, \cite{art_435} \\
    & Transformer \cite{attention} & \cite{art_424}, \cite{art_454}, \cite{art_723} \\
    & Extreme gradient boosting (XGBoost) \cite{XGBoost} & \cite{art_97}, \cite{art_114}, \cite{art_663} \\
    \midrule 
    \multirow[t]{44}{=}{GNN methods} & Adaptive graph convolutional recurrent network (AGCRN) \cite{art_683} & \cite{art_98}, \cite{art_100}, \cite{art_110}, \cite{art_117}, \cite{art_124}, \cite{art_125}, \cite{art_130}, \cite{art_131}, \cite{art_406}, \cite{art_419}, \cite{art_435}, \cite{art_441}, \cite{art_447}, \cite{art_453}, \cite{art_454}, \cite{art_653}, \cite{art_671}, \cite{art_673}, \cite{art_675}, \cite{art_707}, \cite{art_715}, \cite{art_720}, \cite{art_721}, \cite{art_725}, \cite{art_726}, \cite{art_731}, \cite{art_735} \\
    & Attention-based spatial-temporal graph convolutional network (ASTGCN) \cite{ASTGCN} & \cite{art_93}, \cite{art_98}, \cite{art_99}, \cite{art_100}, \cite{art_108}, \cite{art_110}, \cite{art_113}, \cite{art_115}, \cite{art_117}, \cite{art_121}, \cite{art_125}, \cite{art_126}, \cite{art_130}, \cite{art_131}, \cite{art_132}, \cite{art_406}, \cite{art_421}, \cite{art_424}, \cite{art_427}, \cite{art_432}, \cite{art_433}, \cite{art_435}, \cite{art_442}, \cite{art_445}, \cite{art_446}, \cite{art_447}, \cite{art_453}, \cite{art_454}, \cite{art_652}, \cite{art_653}, \cite{art_671}, \cite{art_672}, \cite{art_673}, \cite{art_675}, \cite{art_676}, \cite{art_682}, \cite{art_683}, \cite{art_707}, \cite{art_715}, \cite{art_716}, \cite{art_718}, \cite{art_719}, \cite{art_720}, \cite{art_721}, \cite{art_726}, \cite{art_728}, \cite{art_729}, \cite{art_731}, \cite{art_732}, \cite{art_735}, \cite{art_736} \\
    & Attention based spatial-temporal graph neural network (ASTGNN) \cite{ASTGNN} & \cite{art_422}, \cite{art_435}, \cite{art_446} \\
    & Automated spatio-temporal graph (AutoSTG) \cite{AutoSTG} & \cite{art_108}, \cite{art_408}, \cite{art_720}, \cite{art_726}, \cite{art_731} \\
    & Coupled layer-wise convolutional recurrent neural network (CCRNN) \cite{CCRNN} & \cite{art_682}, \cite{art_718} \\
    & Decoupled dynamic spatial-temporal graph neural network (D2STGNN) \cite{art_672} & \cite{art_653}, \cite{art_740} \\
    & Dual-stage attention-based recurrent neural network (DA-RNN) \cite{DA-RNN} & \cite{art_96}, \cite{art_105} \\
    & Diffusion convolutional recurrent neural network (DCRNN) \cite{DCRNN} & \cite{art_93}, \cite{art_94}, \cite{art_96}, \cite{art_100}, \cite{art_103}, \cite{art_108}, \cite{art_109}, \cite{art_110}, \cite{art_113}, \cite{art_115}, \cite{art_116}, \cite{art_117}, \cite{art_120}, \cite{art_122}, \cite{art_124}, \cite{art_125}, \cite{art_126}, \cite{art_128}, \cite{art_129}, \cite{art_130}, \cite{art_131}, \cite{art_132}, \cite{art_406}, \cite{art_408}, \cite{art_419}, \cite{art_420}, \cite{art_421}, \cite{art_422}, \cite{art_424}, \cite{art_425}, \cite{art_427}, \cite{art_430}, \cite{art_432}, \cite{art_433}, \cite{art_435}, \cite{art_438}, \cite{art_441}, \cite{art_442}, \cite{art_444}, \cite{art_445}, \cite{art_447}, \cite{art_451}, \cite{art_453}, \cite{art_652}, \cite{art_653}, \cite{art_663}, \cite{art_666}, \cite{art_669}, \cite{art_671}, \cite{art_672}, \cite{art_673}, \cite{art_675}, \cite{art_676}, \cite{art_681}, \cite{art_682}, \cite{art_683}, \cite{art_694}, \cite{art_707}, \cite{art_715}, \cite{art_716}, \cite{art_718}, \cite{art_719}, \cite{art_720}, \cite{art_722}, \cite{art_724}, \cite{art_725}, \cite{art_726}, \cite{art_728}, \cite{art_731}, \cite{art_732}, \cite{art_733}, \cite{art_735}, \cite{art_736}, \cite{art_738}, \cite{art_739} \\
    & Deep spatial-temporal network (DeepST) \cite{DeepST} & \cite{art_451}, \cite{art_734} \\
    & Dynamic graph convolutional recurrent network (DGCRN) \cite{art_100} & \cite{art_110}, \cite{art_406}, \cite{art_444}, \cite{art_672} \\
    & Dynamic and multi-faceted spatiotemporal graph convolution network (DMSTGCN) \cite{art_663} & \cite{art_441}, \cite{art_444} \\
    & Dynamic spatial-temporal aware graph neural network (DSTAGNN) \cite{art_675} & \cite{art_125}, \cite{art_419}, \cite{art_424}, \cite{art_447}, \cite{art_694}, \cite{art_707}, \cite{art_715}, \cite{art_721}, \cite{art_732}, \cite{art_735} \\
    & Fully connected gated graph architecture (FC-GAGA) \cite{FC-GAGA} & \cite{art_116}, \cite{art_408}, \cite{art_427} \\
    & Gated attention networks for learning on large and spatiotemporal graphs (GaAN) \cite{GaAN} & \cite{art_669}, \cite{art_738} \\
    & Graph convolutional network (GCN) & \cite{art_104}, \cite{art_135}, \cite{art_431}, \cite{art_668}, \cite{art_725} \\
    & GeoMAN \cite{GeoMAN} & \cite{art_96}, \cite{art_132} \\
    & Graph multi-attention network (GMAN) \cite{GMAN} & \cite{art_100}, \cite{art_109}, \cite{art_110}, \cite{art_116}, \cite{art_122}, \cite{art_128}, \cite{art_408}, \cite{art_419}, \cite{art_422}, \cite{art_427}, \cite{art_430}, \cite{art_444}, \cite{art_451}, \cite{art_652}, \cite{art_653}, \cite{art_663}, \cite{art_672}, \cite{art_673}, \cite{art_716}, \cite{art_719}, \cite{art_720}, \cite{art_724}, \cite{art_726}, \cite{art_731}, \cite{art_732}, \cite{art_733}, \cite{art_735}, \cite{art_738} \\
    & Discrete graph structure learning for time series (GTS) \cite{art_641} & \cite{art_420}, \cite{art_433}, \cite{art_652}, \cite{art_653}, \cite{art_670}, \cite{art_673}, \cite{art_722}, \cite{art_733} \\
    & Graph WaveNet (GWN) \cite{GWN} & \cite{art_93}, \cite{art_100}, \cite{art_108}, \cite{art_109}, \cite{art_110}, \cite{art_113}, \cite{art_115}, \cite{art_116}, \cite{art_117}, \cite{art_122}, \cite{art_124}, \cite{art_128}, \cite{art_406}, \cite{art_408}, \cite{art_419}, \cite{art_420}, \cite{art_422}, \cite{art_424}, \cite{art_430}, \cite{art_432}, \cite{art_433}, \cite{art_434}, \cite{art_435}, \cite{art_441}, \cite{art_444}, \cite{art_447},  \cite{art_453}, \cite{art_454}, \cite{art_652}, \cite{art_653}, \cite{art_663}, \cite{art_666}, \cite{art_670}, \cite{art_671}, \cite{art_672}, \cite{art_673}, \cite{art_676}, \cite{art_681}, \cite{art_682}, \cite{art_694}, \cite{art_715}, \cite{art_716}, \cite{art_718}, \cite{art_719}, \cite{art_720}, \cite{art_722}, \cite{art_723}, \cite{art_724}, \cite{art_726}, \cite{art_731}, \cite{art_732}, \cite{art_733}, \cite{art_735}, \cite{art_738}, \cite{art_739} \\
    & Hierarchical graph convolution network (HGCN) \cite{HGCN} & \cite{art_421}, \cite{art_733} \\
    & Long short-term graph convolutional networks (LSGCN) \cite{LSGCN} & \cite{art_99}, \cite{art_432}, \cite{art_446}, \cite{art_453}, \cite{art_671} \\
    & Multi-graph convolutional networks (MGCN) \cite{MGCN} & \cite{art_114}, \cite{art_664} \\
    & Multi-range attentive bicomponent graph convolutional network (MRA-BGCN) \cite{MRA-BGCN} & \cite{art_109}, \cite{art_116}, \cite{art_408}, \cite{art_724} \\
    & Multivariate time series forecasting with graph neural network (MTGNN) \cite{art_655} & \cite{art_100}, \cite{art_109}, \cite{art_110}, \cite{art_113}, \cite{art_116}, \cite{art_117}, \cite{art_122}, \cite{art_124}, \cite{art_128}, \cite{art_406}, \cite{art_408}, \cite{art_420}, \cite{art_422}, \cite{art_444}, \cite{art_445}, \cite{art_652}, \cite{art_653}, \cite{art_663}, \cite{art_672}, \cite{art_673}, \cite{art_716}, \cite{art_720}, \cite{art_724}, \cite{art_726}, \cite{art_731}, \cite{art_733} \\
    & Multi-view graph convolutional network (MVGCN) \cite{MVGCN} & \cite{art_94}, \cite{art_96} \\
    & Propagation delay-aware dynamic long-range transformer (PDFormer) \cite{PDFormer} & \cite{art_433}, \cite{art_666}, \cite{art_715}, \cite{art_732} \\
    & a Spatio-temporal adaptive embedding transformer (STAEformer) \cite{STAEformer} & \cite{art_666}, \cite{art_721} \\
    & Spectral temporal graph neural network for multivariate time-series forecasting (StemGNN) \cite{art_684} & \cite{art_406}, \cite{art_422}, \cite{art_722}, \cite{art_728} \\
    & STGNN enhanced by a scalable time series Pre-training model (STEP) \cite{art_652} & \cite{art_653}, \cite{art_694}, \cite{art_722} \\
    & Spatial-temporal fusion graph neural network (STFGNN) \cite{STFGNN} & \cite{art_98}, \cite{art_108}, \cite{art_115}, \cite{art_117}, \cite{art_125}, \cite{art_130}, \cite{art_406}, \cite{art_435}, \cite{art_445}, \cite{art_453}, \cite{art_671}, \cite{art_675}, \cite{art_682}, \cite{art_707}, \cite{art_715}, \cite{art_720}, \cite{art_726}, \cite{art_731}, \cite{art_735} \\
    & Spatial-temporal graph to sequence model (STG2Seq) \cite{STG2Seq} & \cite{art_99}, \cite{art_424}, \cite{art_682}, \cite{art_707}, \cite{art_718}, \cite{art_723} \\
    & Spatio-temporal graph convolutional network (STGCN) \cite{STGCN} & \cite{art_93}, \cite{art_94}, \cite{art_96}, \cite{art_99}, \cite{art_100}, \cite{art_106}, \cite{art_108}, \cite{art_109}, \cite{art_110}, \cite{art_113}, \cite{art_115}, \cite{art_116}, \cite{art_117}, \cite{art_120}, \cite{art_121}, \cite{art_122}, \cite{art_124}, \cite{art_126}, \cite{art_128}, \cite{art_129}, \cite{art_130}, \cite{art_131}, \cite{art_132}, \cite{art_406}, \cite{art_408}, \cite{art_419}, \cite{art_420}, \cite{art_421}, \cite{art_422}, \cite{art_424}, \cite{art_425}, \cite{art_433}, \cite{art_435}, \cite{art_438}, \cite{art_442}, \cite{art_443}, \cite{art_444}, \cite{art_445}, \cite{art_451}, \cite{art_453}, \cite{art_454}, \cite{art_653}, \cite{art_671}, \cite{art_675}, \cite{art_676}, \cite{art_681}, \cite{art_682}, \cite{art_683}, \cite{art_707}, \cite{art_715}, \cite{art_716}, \cite{art_719}, \cite{art_720}, \cite{art_722}, \cite{art_724}, \cite{art_725}, \cite{art_726}, \cite{art_731}, \cite{art_732}, \cite{art_733}, \cite{art_734}, \cite{art_735}, \cite{art_736}, \cite{art_738}, \cite{art_739}, \cite{art_740} \\
    & Spatio-temporal graph few shot learning (ST-GFSL) \cite{art_667} & \cite{art_670}, \cite{art_694} \\
    & Spatial temporal graph neural network (STGNN) \cite{art_669} & \cite{art_94}, \cite{art_103}, \cite{art_669} \\
    & Spatial-temporal graph ODE networks (STGODE) \cite{art_676} & \cite{art_99}, \cite{art_108}, \cite{art_117}, \cite{art_125}, \cite{art_406}, \cite{art_424}, \cite{art_447}, \cite{art_453}, \cite{art_671}, \cite{art_675}, \cite{art_681}, \cite{art_707}, \cite{art_715}, \cite{art_718}, \cite{art_720}, \cite{art_725}, \cite{art_726} \\
    & Spatio-temporal graph attention network (ST-GRAT) \cite{art_738} & \cite{art_116}, \cite{art_444} \\
    & Deep-meta-learning based model (ST-MetaNet) \cite{ST-MetaNet} & \cite{art_96}, \cite{art_100}, \cite{art_109}, \cite{art_116}, \cite{art_408}, \cite{art_451}, \cite{art_673}, \cite{art_720}, \cite{art_724}, \cite{art_726}, \cite{art_731}, \cite{art_734} \\
    & Spatiotemporal multi-graph convolution network (ST-MGCN) \cite{ST-MGCN} & \cite{art_451}, \cite{art_734} \\
    & Spatio-temporal synchronous graph convolutional network (STSGCN) \cite{STSGCN} & \cite{art_98}, \cite{art_99}, \cite{art_100}, \cite{art_108}, \cite{art_115}, \cite{art_117}, \cite{art_125}, \cite{art_130}, \cite{art_131}, \cite{art_406}, \cite{art_419}, \cite{art_424}, \cite{art_435}, \cite{art_441}, \cite{art_442}, \cite{art_443}, \cite{art_453}, \cite{art_454}, \cite{art_652}, \cite{art_653}, \cite{art_671}, \cite{art_672}, \cite{art_673}, \cite{art_675}, \cite{art_676}, \cite{art_682}, \cite{art_683}, \cite{art_707}, \cite{art_715}, \cite{art_720}, \cite{art_726}, \cite{art_731}, \cite{art_732}, \cite{art_735} \\
    & Temporal graph convolutional network (T-GCN) \cite{T-GCN} & \cite{art_103}, \cite{art_113}, \cite{art_121}, \cite{art_421}, \cite{art_433}, \cite{art_669}, \cite{art_716}, \cite{art_718}, \cite{art_725}, \cite{art_729}, \cite{art_733} \\
    & Time zigzags at graph convolutional networks for time series forecasting (Z-GCNETs) \cite{art_657} & \cite{art_125}, \cite{art_424}, \cite{art_441}, \cite{art_675}, \cite{art_707}, \cite{art_735} \\
    \bottomrule
\end{longtable}

As for the basic mathematical and statistical methods, two widely used benchmarks are the ARIMA model and the historical average model (HA), which uses the average of past observations as the forecast for future values. Despite their apparent simplicity, these benchmarks are sometimes shown to be very powerful, with an accuracy comparable to that of more complex models. As for the non-GNN machine learning models, LSTM networks are widely recognized as a type of recurrent neural network capable of handling long-term dependencies in traffic data. As for GNN-based models instead, the most used models are the spatio-temporal graph convolutional network (STGCN) \cite{STGCN}, the diffusion convolutional recurrent neural network (DCRNN) \cite{DCRNN}, Graph WaveNet (GWN) \cite{GWN}, and the attention-based spatial-temporal graph convolutional network (ASTGCN) \cite{ASTGCN}, all developed specifically for the traffic forecasting problem.

The STGCN model proposed by Yu et al.~in 2018 consists of two spatio-temporal convolutional blocks (with two temporal gated convolutional layers, one spatial graph convolutional layer in the middle, and residual connections) and a fully-connected output layer in the end. In its original implementation, it processes graphs a with weighted adjacency matrix based on spatial distance between sensors in the road network. The source code for the STGCN model is available at \url{https://github.com/VeritasYin/STGCN_IJCAI-18}. The DCRNN model proposed by Li et al.~in 2017 aims to capture the spatial dependency by using bidirectional random walks on the graph (whose adjacency matrix is based on node proximity), and to capture the temporal dependencies by using an encoder-decoder architecture with scheduled sampling. The code can be accessed at \url{https://github.com/liyaguang/DCRNN}. The GWN model was introduced by Wu et al.~in 2021 as a novel approach to spatio-temporal forecasting, particularly in traffic tasks. It is known for its ability to learn an adaptive dependency matrix from node embeddings. The model combines graph convolutional and temporal convolutional layers to effectively capture spatio-temporal dependencies. The source code is available at \url{https://github.com/nnzhan/Graph-WaveNet}. The ASTGCN model proposed by Guo et al.~in 2019 models at once recent, daily-periodic and weekly-periodic dependencies as three independent components, with a spatio-temporal attention mechanism for capturing the dynamic spatio-temporal correlations and a spatio-temporal convolution which simultaneously employs graph convolutions to capture the spatial patterns. The source code for the ASTGCN model is available at \url{https://github.com/Davidham3/ASTGCN-2019-mxnet}. For a more comprehensive list of source codes for the most dated benchmark GNN models in the traffic domain see Ref.~\cite{review_traffico_jiang}.

\subsubsection{Results}
\label{s:Mobility_results}
In all papers mentioned, the authors claim that the proposed GNN-based models are able to outperform the benchmarks. However, there are a limited but interesting number of situations in which some simple statistical and mathematical models demonstrate comparable accuracy, as it can be observed in Tabs.~\ref{t:risultati metr-la} and \ref{t:risultati pems-bay}. 

For the comparison and evaluation of these models, the most commonly used error metrics are MAE, RMSE, and MAPE. A few papers in addition use further metrics such as the coefficient of determination $R^2$. However, despite the use of the same metrics and the availability of widely used datasets, comparison among these models is not trivial, because different papers often focus on different time windows of a dataset.

In Tab.~\ref{t:risultati metr-la} it is shown the accuracy of the different models applied to the METR-LA dataset, with a 5-minute granularity from 1$^{\text{st}}$ March 2012 to 30$^{\text{th}}$ June 2012 for 15, 30 and 60 minute time horizons, expressed in terms of MAE, MAPE and RMSE.

{\tabcolsep=2.5pt
\begin{longtable}{lccccccccc}
    \caption{Comparison of the average accuracy of the different models on the METR-LA dataset for 15, 30 and 60 minute time horizons, expressed in terms of MAE, MAPE and RMSE. The results of the models marked with $^\star$ are taken from papers in the \enquote{Generic} group. The smallest errors are underlined. Numbers from the original papers.}
    \label{t:risultati metr-la}
    \endfirsthead
    \endhead
    \toprule
    Time horizon & \multicolumn{3}{c}{15 min} & \multicolumn{3}{c}{30 min} & \multicolumn{3}{c}{60 min} \\
    Metrics & MAE & MAPE & RMSE & MAE & MAPE & RMSE & MAE & MAPE & RMSE \\
    \midrule 
    AGCRN & 2.87 & 7.70 & 5.58 & 3.23 & 9.00 & 6.58 & 3.62 & 10.38 & 7.51 \\
    AGP-GNN \cite{art_116} & 2.71 & 7.21 & 5.25 & 3.04 & 8.15 & 6.20 & 3.41 & 9.70 & 7.20 \\
    AMGCN \cite{art_444} & 2.52 & 6.13 & \underline{4.48} & 2.97 & 7.70 & 5.35 & 3.38 & 8.93 & \underline{6.21} \\
    ARIMA & 3.99 & 9.60 & 8.12 & 5.15 & 12.70 & 10.45 & 6.90 & 17.40 & 13.23 \\
    ASTGCN & 4.86 & 9.21 & 9.27 & 5.43 & 10.13 & 10.61 & 6.51 & 11.64 & 12.52 \\
    ASTGNN & 2.66 & 6.96 & 5.09 & 3.06 & 8.47 & 6.26 & 3.52 & 9.56 & 7.21 \\
    AutoSTG & 2.70 & - & 5.16 & 3.06 & - & 6.17 & 3.47 & - & 7.27 \\
    AutoSTS \cite{art_720} & 2.57 & 6.55 & 4.93 & 2.89 & 7.85 & 5.87 & 3.28 & 9.43 & 6.86 \\
    CASTGNN \cite{art_442} & 2.88 & 6.67 & 6.44 & 3.25 & 7.91 & 7.56 & 3.70 & 9.76 & 8.92 \\
    D$^2$STGNN \cite{art_672} & 2.56 & 6.48 & 4.88 & 2.90 & 7.78 & 5.89 & 3.35 & 9.40 & 7.03 \\
    DCRNN & 2.77 & 7.30 & 5.38 & 3.15 & 8.80 & 6.45 & 3.60 & 10.50 & 7.60 \\
    DetectorNet \cite{art_724} & - & - & - & 3.06 & 8.12 & 6.08 & 3.40 & 9.60 & 6.98\\
    DGCRN \cite{art_100} & 2.62 & 6.63 & 5.01 & 2.99 & 8.02 & 6.05 & 3.44 & 9.73 & 7.19 \\
    DMSTGCN & 2.85 & 7.54 & 5.54 & 3.26 & 9.19 & 6.56 & 3.72 & 10.96 & 7.55 \\
    DST-GCNN \cite{art_126} & 2.68 & 7.20 & 5.35 & 3.01 & 8.50 & 6.23 & 3.41 & 10.30 & 7.47 \\
    DualGraph \cite{art_438} & 2.62 & 7.10 & 5.36 & 2.83 & 7.90 & 5.89 & 3.15 & 9.10 & 6.67 \\
    DyGCN-LSTM \cite{art_113} & 2.64 & 6.35 & 4.64 & 3.24 & 8.04 & \underline{5.24} & 3.53 & 9.30 & 6.53 \\
    DyHSL & 2.58 & 6.47 & 4.98 & 2.95 & 8.05 & 6.13 & 3.41 & 9.41 & 7.03 \\
    FC-GAGA & 2.75 & 7.25 & 5.34 & 3.10 & 8.57 & 6.30 & 3.51 & 10.14 & 7.31 \\
    FNN & 3.99 & 9.90 & 7.94 & 4.23 & 12.90 & 8.17 & 4.49 & 14.00 & 8.69 \\
    GA-LSTM \cite{art_120} & \underline{2.30} & 7.01 & 4.97 & 2.98 & 8.50 & 6.00 & 3.55 & 9.97 & 7.23 \\
    GaAN & 2.71 & 6.99 & 5.24 & 3.12 & 8.56 & 6.36 & 3.64 & 10.62 & 7.65 \\
    GGRU & 2.71 & 6.99 & 5.24 & 3.12 & 8.56 & 6.36 & 3.64 & 10.62 & 7.65 \\
    GL-STGTN \cite{art_433} & 2.61 & 6.94 & 5.08 & 2.96 & 8.23 & 5.97 & 3.37 & 9.81 & 6.99 \\
    GMAN & 2.77 & 7.25 & 5.48 & 3.07 & 8.35 & 6.34 & 3.40 & 9.72 & 7.22 \\
    GTS$^\star$ \cite{art_641} & 2.64 & 6.80 & 4.95 & 3.01 & 8.20 & 5.85 & 3.41 & 9.90 & 6.74 \\
    GWN & 2.69 & 6.90 & 5.15 & 3.07 & 8.37 & 6.22 & 3.53 & 10.01 & 7.37 \\
    HA & 4.16 & 13.00 & 7.80 & 4.16 & 13.00 & 7.80 & 4.16 & 13.00 & 7.80 \\
    HighResNet & 2.68 & 6.88 & 5.10 & 3.03 & 8.34 & 6.11 & 3.47 & 10.24 & 7.27 \\
    LSTM & 3.44 & 9.60 & 6.30 & 3.77 & 10.90 & 7.23 & 4.37 & 13.20 & 8.69 \\
    MCFGNN$^\star$ \cite{art_86} & 2.31 & 5.86 & 5.29 & \underline{2.66} & \underline{6.89} & 6.27 & \underline{3.05} & \underline{7.65} & 7.20 \\
    MRA-BGCN & 2.67 & 6.80 & 5.12 & 3.06 & 8.30 & 6.17 & 3.49 & 10.00 & 7.30 \\
    MTGNN$^\star$  \cite{art_655} & 2.69 & 6.86 & 5.18 & 3.05 & 8.19 & 6.17 & 3.49 & 9.87 & 7.23 \\
    MTGODE$^\star$ \cite{art_75} & 2.66 & 6.87 & 5.10 & 3.00 & 8.19 & 6.05 & 3.39 & 9.80 & 7.05 \\
    MVST-GNN \cite{art_109} & 2.70 & \underline{5.19} & 6.97 & 3.06 & 8.12 & 6.08 & 3.40 & 9.60 & 6.98 \\
    PAHNN \cite{art_435} & 2.51 & 6.34 & 4.82 & 2.87 & 7.63 & 5.73 & 3.25 & 9.32 & 6.67 \\
    PDFormer & 2.83 & 7.77 & 5.45 & 3.20 & 9.19 & 6.46 & 3.62 & 10.91 & 7.47 \\
    PGCN \cite{art_441} & 2.70 & 6.98 & 5.16 & 3.08 & 8.38 & 6.22 & 3.54 & 9.94 & 7.36 \\
    SAGDFN \cite{art_653} & 2.56 & 6.50 & 5.00 & 2.94 & 7.90 & 6.05 & 3.37 & 9.50 & 7.17 \\
    SDGCN & 2.76 & 7.16 & 5.38 & 3.15 & 8.53 & 6.46 & 3.61 & 9.87 & 7.43 \\
    SLCNN & 2.53 & 6.70 & 5.18 & 2.88 & 8.00 & 6.15 & 3.30 & 9.70 & 7.20 \\
    SNAS4MTF \cite{art_408} & 2.68 & 6.90 & 5.16 & 3.03 & 8.25 & 6.12 & 3.43 & 9.82 & 7.15 \\
    STAG-GCN \cite{art_93} & 2.67 & 7.00 & 5.23 & 3.07 & 8.26 & 6.15 & 3.50 & 9.93 & 7.24 \\
    STAEFormer & 2.65 & 6.85 & 5.11 & 2.97 & 8.13 & 6.00 & 3.34 & 9.70 & 7.02 \\
    STEP \cite{art_652} & 2.61 & 6.60 & 4.98 & 2.96 & 7.96 & 5.97 & 3.37 & 9.61 & 6.99 \\
    STID & 2.82 & 7.75 & 5.53 & 3.19 & 9.39 & 6.57 & 3.55 & 10.95 & 7.55 \\
    STFGNN$^\star$ & 2.57 & 6.51 & 4.73 & 2.83 & 7.46 & 5.46 & 3.18 & 8.81 & 6.40 \\
    STGCN & 2.88 & 7.62 & 5.74 & 3.47 & 9.57 & 7.24 & 4.59 & 12.70 & 9.40 \\
    STGODE$^\star$ & 3.47 & 8.76 & 6.76 & 4.36 & 11.14 & 8.47 & 5.50 & 14.32 & 10.33 \\
    STG-NCDE$^\star$ & 3.77 & 8.54 & 9.47 & 4.84 & 10.63 & 12.04 & 6.35 & 13.49 & 14.94 \\
    STGRAT & 2.60 & 6.61 & 5.07 & 3.01 & 8.15 & 6.21 & 3.49 & 10.01 & 7.42 \\
    ST-MetaNet & 2.69 & 6.91 & 5.17 & 3.10 & 8.57 & 6.28 & 3.59 & 10.63 & 7.52 \\
    STSD \cite{art_673} & 2.60 & 6.55 & 4.96 & 2.96 & 7.93 & 5.98 & 3.37 & 9.51 & 7.03 \\
    STSGCN & 3.31 & 8.06 & 7.62 & 4.13 & 10.29 & 9.77 & 5.06 & 12.91 & 11.66 \\
    SVR & 3.99 & 9.30 & 8.45 & 5.05 & 12.10 & 10.87 & 6.72 & 16.70 & 13.76 \\
    TF-GAN \cite{art_122} & 2.63 & 6.55 & 4.94 & 3.06 & 8.36 & 6.20 & 3.32 & 9.48 & 7.11 \\
    T-GCN & 3.03 & 7.81 & 5.26 & 3.52 & 9.45 & 6.12 & 4.30 & 11.80 & 7.31 \\
    TGC-GRU & 5.25 & 12.45 & 8.56 & 5.99 & 14.18 & 10.37 & 7.32 & 17.11 & 13.47 \\
    VAR & 4.42 & 10.20 & 7.89 & 5.41 & 12.70 & 9.13 & 6.52 & 15.80 & 10.11 \\
    TimesNet & 2.78 & 7.52 & 5.49 & 3.36 & 9.43 & 6.41 & 4.52 & 11.37 & 9.66 \\
    TwoResNet \cite{art_430} & 2.65 & 6.78 & 5.08 & 3.01 & 8.14 & 6.07 & 3.39 & 9.71 & 7.08 \\
    WaveNet & 2.99 & 8.04 & 5.89 & 3.59 & 10.25 & 7.28 & 4.45 & 13.62 & 8.93 \\
    Z-GCNETs & 3.23 & 7.87 & 7.48 & 3.93 & 9.75 & 9.40 & 4.83 & 12.04 & 11.57 \\
    \bottomrule
\end{longtable}
}

Clearly, the results on the METR-LA dataset are not uniform, and the best performing models vary depending on the metric and the time horizon. However, the multichannel fusion graph neural network (MCFGNN) \cite{art_86} from the \enquote{Generic} group consistently exhibits low errors across different time horizons. The MCFGNN model, proposed by Chen et al.~in 2022, is composed of three main components: a graph construction layer, a graph convolution layer, and a temporal convolution layer. The input data are first processed by the graph construction layer, which incorporates three channels to capture meta-features, time-varying patterns, and globally stabilized graphs. Spatial and temporal convolutions are then applied in pairs to capture spatio-temporal dependencies in the data. To mitigate the issue of vanishing gradients, residual connections are added to each spatio-temporal block.

Tab.~\ref{t:risultati pems-bay} displays the accuracy of various models on the PEMS-BAY dataset with a 5-minute granularity and data spanning from 1$^{\text{st}}$ January 2017 to 30$^{\text{th}}$ June 2017 for 15, 30 and 60 minute time horizons, with results expressed in terms of MAE, MAPE and RMSE.

{\tabcolsep=3pt
\begin{longtable}{lccccccccc}
    \caption{Comparison of the average accuracy of the different models on the PEMS-BAY dataset for 15, 30 and 60 minute time horizons, expressed in terms of MAE, MAPE and RMSE. The smallest errors are underlined. Numbers from the original papers.}
    \label{t:risultati pems-bay}
    \endfirsthead
    \endhead
    \toprule
    Time horizon & \multicolumn{3}{c}{15 min} & \multicolumn{3}{c}{30 min} & \multicolumn{3}{c}{60 min} \\
    Metrics & MAE & MAPE & RMSE & MAE & MAPE & RMSE & MAE & MAPE & RMSE \\
    \midrule
    AGCRN & 1.37 & 2.94 & 2.87 & 1.69 & 3.87 & 3.85 & 1.96 & 4.64 & 4.54 \\
    AGP-GNN \cite{art_116} & 1.33 & 2.79 & 2.86 & 1.61 & 3.66 & 3.67 & 1.86 & 4.30 & 4.34 \\
    AMGCN \cite{art_444} & \underline{1.22} & 2.63 & \underline{2.34} & 1.53 & 3.55 & \underline{3.13} & 1.93 & 4.66 & \underline{3.95} \\
    ARIMA & 1.62 & 3.50 & 3.30 & 2.33 & 5.40 & 4.76 & 3.38 & 8.30 & 6.50 \\
    ASTGCN & 1.52 & 3.22 & 3.13 & 2.01 & 4.48 & 4.27 & 2.61 & 6.00 & 5.42 \\
    ASTGNN & 1.34 & 3.84 & 2.75 & 1.65 & 3.72 & 3.74 & 1.94 & 4.59 & 4.51 \\
    AutoSTG & 1.33 & 2.79 & 2.78 & 1.63 & 3.66 & 3.62 & 1.91 & 4.57 & 4.42 \\
    AutoSTS \cite{art_720} & 1.28 & 2.66 & 2.63 & 1.57 & 3.54 & 3.50 & 1.83 & 4.41 & 4.27 \\
    D$^2$STGNN \cite{art_672} & 1.24 & 2.58 & 2.60 & 1.55 & 3.49 & 3.52 & 1.85 & 4.37 & 4.30 \\
    DCRNN & 1.38 & 2.90 & 2.95 & 1.74 & 3.90 & 3.97 & 2.07 & 4.90 & 4.74 \\
    DetectorNet \cite{art_724} & - & - & - & 1.57 & 3.56 & 3.54 & 1.80 & 4.19 & 4.26 \\
    DGCRN \cite{art_100} & 1.28 & 2.66 & 2.69 & 1.59 & 3.55 & 3.63 & 1.89 & 4.43 & 4.42 \\
    DMSTG & - & - & - & - & - & - & 1.91 & 4.51 & 4.43 \\
    DMSTGCN & 1.33 & 2.80 & 2.83 & 1.67 & 3.81 & 3.79 & 1.99 & 4.78 & 4.54 \\
    DSTCGCN & - & - & - & - & - & - & 1.95 & 4.66 & 4.51 \\
    DyHSL & 1.30 & 2.73 & 2.71 & 1.60 & 3.57 & 3.67 & 1.95 & 4.62 & 4.53 \\
    FC-GAGA & 1.36 & 2.87 & 2.86 & 1.68 & 3.80 & 3.80 & 1.97 & 4.67 & 4.52 \\
    FNN & 2.20 & 5.19 & 4.42 & 2.30 & 5.43 & 4.63 & 2.46 & 5.89 & 4.98 \\
    GL-STGTN \cite{art_433} & 1.26 & 2.79 & 2.73 & 1.55 & 3.63 & 3.58 & 1.80 & 4.38 & 4.30 \\
    GMAN & 1.34 & 2.81 & 2.82 & 1.62 & 3.62 & 3.72 & 1.86 & 4.31 & 4.32 \\
    GSLNN \cite{art_420} & 1.29 & 2.71 & 2.66 & 1.64 & 3.67 & 3.74 & 1.93 & 4.51 & 4.46 \\
    GTS$^\star$ \cite{art_641} & 1.32 & 2.80 & 2.62 & 1.64 & 3.60 & 3.41 & 1.91 & 4.40 & 3.97 \\
    GWN & 1.30 & 2.73 & 2.74 & 1.63 & 3.67 & 3.70 & 1.95 & 4.63 & 4.52 \\
    Informer$^\star$ & 2.30 & 5.02 & 4.21 & 2.40 & 5.32 & 4.49 & 2.55 & 5.73 & 4.85 \\
    LDS$^\star$ & 1.33 & 2.80 & 2.81 & 1.67 & 3.80 & 3.80 & 1.99 & 4.80 & 4.59 \\
    HA & 1.89 & 4.16 & 4.30 & 2.50 & 5.62 & 5.82 & 3.31 & 7.65 & 7.54 \\
    LSTM & 2.05 & 4.80 & 4.19 & 2.20 & 5.20 & 4.55 & 2.37 & 5.70 & 4.96 \\
    LSVR & 1.85 & 3.80 & 3.59 & 2.48 & 5.50 & 5.18 & 3.28 & 8.00 & 7.08 \\
    HighResNet & 1.31 & 2.74 & 2.75 & 1.64 & 3.68 & 3.75 & 1.95 & 4.61 & 4.56 \\
    MegaCRN & 1.28 & 2.67 & 2.72 & 1.60 & 3.57 & 3.68 & 1.88 & 4.41 & 4.42 \\
    MRA-BGCN & 1.29 & 2.90 & 2.72 & 1.61 & 3.80 & 3.67 & 1.91 & 4.60 & 4.46 \\
    MTGNN$^\star$ \cite{art_655} & 1.32 & 2.77 & 2.79 & 1.65 & 3.69 & 3.74 & 1.94 & 4.53 & 4.49 \\
    MTGODE$^\star$ \cite{art_75} & 1.29 & 2.72 & 2.73 & 1.61 & 3.61 & 3.66 & 1.88 & 4.39 & 4.31 \\
    MVST-GNN \cite{art_109} & 1.30 & 2.73 & 2.74 & 1.57 & 3.56 & 3.54 & 1.80 & 4.19 & 4.26 \\
    PAHNN \cite{art_435} & 1.27 & 2.60 & 2.65 & \underline{1.52} & 3.48 & 3.54 & 1.83 & 4.35 & 4.27 \\
    PDFormer & 1.32 & 2.78 & 2.83 & 1.64 & 3.71 & 3.79 & 1.91 & 4.51 & 4.43 \\
    PGCN \cite{art_441} & 1.30 & 2.72 & 2.73 & 1.62 & 3.63 & 3.67 & 1.92 & 4.45 & 4.45 \\
    RGSL & - & - & - & - & - & - & 2.03 & 4.77 & 4.64 \\
    RST-LTG \cite{art_406} & - & - & - & - & - & - & 1.88 & 4.32 & 4.31 \\
    SDGCN & 1.35 & 2.89 & 2.88 & 1.69 & 3.87 & 3.91 & 1.99 & 4.69 & 4.59 \\
    SLCNN & 1.44 & 3.00 & 2.90 & 1.72 & 3.90 & 3.81 & 2.03 & 4.80 & 4.53 \\
    SNAS4MTF \cite{art_408} & 1.30 & 2.72 & 2.77 & 1.61 & 3.63 & 3.68 & 1.89 & 4.44 & 4.36 \\
    STAEFormer & 1.31 & 2.76 & 2.78 & 1.62 & 3.62 & 3.68 & 1.88 & 4.41 & 4.34 \\
    StemGNN & 1.44 & 3.08 & 3.12 & 1.93 & 4.54 & 4.38 & 2.57 & 6.55 & 5.88 \\
    STEP \cite{art_652} & 1.26 & 2.59 & 2.73 & 1.55 & \underline{3.43} & 3.58 & \underline{1.79} & \underline{4.18} & 4.20 \\
    STID & 1.30 & 2.73 & 2.81 & 1.62 & 3.68 & 3.72 & 1.89 & 4.47 & 4.40 \\
    STFGNN & 1.42 & 2.92 & 2.97 & 1.67 & 3.92 & 3.94 & 1.98 & 4.83 & 4.74 \\
    STGCN & 1.36 & 2.90 & 2.96 & 1.81 & 4.17 & 4.27 & 2.49 & 5.79 & 5.69 \\
    STG-NCDE$^\star$ & 1.38 & 2.91 & 2.93 & 1.71 & 3.91 & 3.84 & 2.03 & 4.82 & 4.58 \\
    STGODE$^\star$ & 1.43 & 2.99 & 2.88 & 1.84 & 3.84 & 3.90 & 2.30 & 4.61 & 4.89 \\
    STGRAT \cite{art_738} & 1.29 & 2.67 & 2.71 & 1.61 & 3.63 & 3.69 & 1.95 & 4.64 & 4.54 \\
    ST-MetaNet & 1.36 & 2.82 & 2.90 & 1.76 & 4.00 & 4.02 & 2.20 & 5.45 & 5.06 \\
    ST-Norm & 1.34 & 2.82 & 2.88 & 1.67 & 3.75 & 3.83 & 1.96 & 4.62 & 4.52 \\
    STSD \cite{art_673} & 1.28 & 2.64 & 2.69 & 1.60 & 3.54 & 3.63 & 1.88 & 4.35 & 4.36 \\
    STSGCN & 1.44 & 3.04 & 3.01 & 1.83 & 4.17 & 4.18 & 2.26 & 5.40 & 5.21 \\
    SVR & 1.85 & 3.80 & 3.59 & 2.48 & 5.50 & 5.18 & 3.28 & 8.00 & 7.08 \\
    TD2MG2NN \cite{art_422} & 1.25 & 2.62 & 2.64 & 1.55 & 3.47 & 3.50 & 1.85 & 4.28 & 4.22 \\
    TGC-GRU & 2.34 & 5.08 & 4.30 & 2.44 & 5.40 & 4.64 & 2.57 & 5.78 & 5.01 \\
    T-GCN & 1.50 & 3.14 & 2.83 & 1.73 & 3.76 & 3.40 & 2.18 & 4.94 & 4.35 \\
    TimesNet & 1.32 & 2.94 & 2.72 & 1.76 & 3.74 & 4.16 & 2.18 & 4.84 & 5.20 \\
    TPGNN$^\star$ \cite{art_685} & 1.26 & \underline{2.56} & 2.64 & 1.65 & 3.47 & 3.65 & 2.05 & 4.40 & 4.58 \\ 
    TwoResNet \cite{art_430} & 1.30 & 2.70 & 2.73 & 1.61 & 3.59 & 3.69 & 1.89 & 4.40 & 4.41 \\
    VAR & 1.74 & 3.60 & 3.16 & 2.32 & 5.00 & 4.25 & 2.93 & 6.50 & 5.44 \\
    WaveNet & 1.39 & 2.91 & 3.01 & 1.83 & 4.16 & 4.21 & 2.35 & 5.87 & 5.43 \\
    Z-GCNETs & 1.36 & 2.88 & 2.86 & 1.68 & 3.79 & 3.78 & 1.98 & 4.60 & 4.53 \\
    \bottomrule
\end{longtable}
}

The results on the the PEMS-BAY dataset vary depending on the metrics and the time horizon. The most accurate models are the adaptive multigraph convolutional networks (AMGCN) \cite{art_444}, and the spatio-temporal GNN enhanced by a scalable time series pre-training model (STEP) \cite{art_652}. The AMGCN model was proposed by Li et al.~in 2024, and includes a graph fusion block, an adaptive graph generator block, a stack of spatio-temporal convolution blocks, and an output module. The graph fusion block integrates information from five pre-defined graph structures, based on spatial distance, reachability, correlation, and distributions. Then, an independent adaptive graph generator learns an evolving graph structure to complement the pre-defined graph. Spatio-temporal features are finally extracted through a stack of graph convolutional modules with residual connections, and a time-mixing strategy. The STEP model was proposed by Shao et a.~in 2021 as a framework to enhance spatio-temporal GNNs through a pre-training approach. Specifically, the pre-training model is designed to learn temporal patterns from long-term historical time series data and generate segment-level representations. These representations, which capture rich contextual information, are then integrated into a spatio-temporal GNN to improve its performance on short-term forecasting tasks. To demonstrate the effectiveness of the framework, the authors use Graph WaveNet as a representative backbone model within STEP.

Other models that, overall, achieve good results are the multi view spatial-temporal graph neural network (MVST-GNN) \cite{art_109}, and the temporal decomposition based dynamic multi granularity graph neural network (TD2MG2NN) \cite{art_422}. The MVST-GNN model, proposed by Li et al.~in 2022, includes a multi-view temporal transformer module and a multi-view graph convolution module. The multi-view temporal transformer module is designed to capture dynamic temporal features from recent, daily, and weekly patterns using self-attention and a fusion module. Then, the graph convolution module operates on a self-learned adjacency matrix and passes its output through two fully connected linear layers to generate the final forecast. The TD2MG2NN model, instead, was introduced by Zang et al.~in 2024 for evaluating the resilience of road networks in real time during heavy rain events. The process starts by decomposing the time series into seasonal and trend components. These components are processed individually through a season-temporal encoder layer with temporal convolutions, and a trend-temporal encoder layer with regression, respectively. The model then learns two feature matrices, one for daily and one for weekly patterns, which are fed into a dynamic multi-granularity graph convolution module. This module processes the feature matrices in parallel, and then passes them through a spatial decoder layer to produce the final output.

As for the other PeMS datasets (PeMS03, PeMS04, PeMS07 and PeMS08), the results are largely consistent with an indication that the periodicity aware spatial-temporal adaptive hypergraph neural network (PAHNN) \cite{art_435} is the most accurate model, as shown in Tabs.~\ref{t:risultati pems03}, \ref{t:risultati pems04}, \ref{t:risultati pems07}, and \ref{t:risultati pems08}. Tab.~\ref{t:risultati pems03} reports the accuracy of the different models applied to the PeMS03 dataset, which includes data from 1$^{\text{st}}$ September 2018 to 30$^{\text{th}}$ November 2018 at a 5-minute granularity for a 60-minute time horizon, with results given in terms of MAE, MAPE and RMSE.

\begin{longtable}{lccc}
    \caption{Comparison of the average accuracy of the different models on the PeMS03 dataset for a 60-minute time horizon, expressed in terms of MAE, MAPE and RMSE. The results of the models marked with $^\star$ are taken from papers in the \enquote{Generic} group. The smallest errors are underlined. Numbers from the original papers.}
    \label{t:risultati pems03}
    \endfirsthead
    \endhead
    \toprule
    Time horizon & \multicolumn{3}{c}{60 min} \\
    Metrics & MAE & MAPE & RMSE\\
    \midrule 
    AdaSTNet & 15.49 & 14.84 & 27.57 \\
    AGCRN & 15.98 & 15.23 & 28.25 \\
    ARIMA & 35.41 & 33.78 & 47.59 \\
    ASTGCN & 17.69 & 19.40 & 29.66 \\
    ASTGNN & 14.78 & 14.79 & 25.00 \\
    Auto-DSTSGN \cite{art_108} & 14.59 & 14.22 & 25.17 \\
    AutoSTG & 16.27 & 16.10 & 27.63 \\
    AutoSTS \cite{art_720} & 14.61 & 14.18 & 24.71 \\
    DCGCN \cite{art_117} & 15.29 & - & 25.98 \\
    DCRNN & 18.18 & 18.91 & 30.31 \\
    DeepGLO$^\star$ & 17.25 & 19.27 & 23.25 \\
    DeepState$^\star$ & 15.59 & 18.69 & \underline{20.21} \\
    DSTAGNN \cite{art_675} & 15.57 & 14.68 & 27.21 \\
    DyHSL & 15.49 & 14.38 & 27.06 \\
    FourierGNN & 17.27 & 15.88 & 27.20 \\
    Gboot \cite{art_715} & 15.43 & 14.51 & 26.42 \\
    GMAN & 15.52 & 15.19 & 26.53 \\
    GMSDR \cite{art_682} & 15.78 & 15.33 & 26.82 \\
    GWN & 19.85 & 19.31 & 32.94 \\
    HA & 31.58 & 33.78 & 52.39 \\
    LSGCN & 17.94 & - & 29.85 \\
    LSTM & 21.33 & 23.33 & 35.11 \\
    LSTNet$^\star$ & 19.07 & 17.73 & 29.67 \\
    N-BEATS$^\star$ & 18.45 & 18.35 & 31.23 \\
    PAHNN \cite{art_435} & 14.02 & \underline{13.19} & 23.67 \\
    PDFormer & 21.82 & 21.47 & 36.75 \\
    PYNet \cite{art_735} & 14.94 & 14.94 & 25.27 \\
    SFM$^\star$ & 17.67 & 18.33 & 30.01 \\
    SPGCL & 23.31 & 21.88 & 37.37 \\
    STAGCN \cite{art_130} & 15.40 & 14.48 & 26.23 \\
    STCGNN \cite{art_115} & 15.81 & 14.59 & 27.23 \\
    StemGNN$^\star$ & 14.32 & 16.24 & 21.64 \\
    STFGCN & 16.77 & 16.30 & 28.34 \\
    STGBN \cite{art_707} & 15.22 & 14.78 & 27.09 \\
    STGCN & 17.49 & 17.15 & 30.12 \\
    STG2SEC & 19.03 & 21.55 & 29.73 \\
    STGODE & 16.53 & 16.68 & 27.79 \\
    STSGCN & 17.48 & 16.78 & 29.21 \\
    STWave \cite{art_671} & 14.93 & 15.05 & 26.50 \\
    STWave+ \cite{art_453} & 14.71 & 14.88 & 26.31 \\
    SVR & 21.97 & 21.51 & 35.29 \\
    TAMP-S2GCNets$^\star$ \cite{art_674} & \underline{13.91} & 13.40 & 23.77 \\
    TCN & 19.32 & 19.93 & 33.55 \\
    TimesNet & 16.41 & 15.17 & 26.72 \\
    VAR & 23.65 & 24.51 & 38.26 \\
    W-DSTAGNN \cite{art_419} & 15.31 & 14.49 & 26.59 \\
    Z-GCNETs & 14.20 & 13.88 & 25.29 \\
    \bottomrule
\end{longtable}

In Tab.~\ref{t:risultati pems04} is shown the accuracy of the different models on the PeMS04 dataset with a 5-minute granularity and data from 1$^{\text{st}}$ January 2018 to 28$^{\text{th}}$ February 2018 a 60-minute time horizon, expressed in terms of MAE, MAPE and RMSE.

\begin{longtable}{lccc}
    \caption{Comparison of the average accuracy of the different models on the PeMS04 dataset for a 60-minute time horizon, expressed in terms of MAE, MAPE and RMSE. The results of the models marked with $^\star$ are taken from papers in the \enquote{Generic} group. The smallest errors are underlined. Numbers from the original papers.}
    \label{t:risultati pems04}
    \endfirsthead
    \endhead
    \toprule
    Time horizon & \multicolumn{3}{c}{60 min} \\
    Metrics & MAE & MAPE & RMSE\\
    \midrule 
    AdaSTNet & 19.25 & 12.77 & 31.26 \\
    AGCRN \cite{art_683} & 19.83 & 12.97 & 32.26 \\
    ARIMA & 33.73 & 24.18 & 48.80 \\
    ASTGCN & 22.93 & 16.56 & 35.22 \\
    ASTGNN & 18.60 & 12.36 & 30.91 \\
    Auto-DSTSGN \cite{art_108} & 18.85 & 13.21 & 30.48 \\
    Autoformer$^\star$ & 23.76 & 18.01 & 36.59 \\
    AutoSTG & 20.38 & 14.12 & 32.51 \\
    AutoSTS \cite{art_720} & 18.76 & 12.84 & 30.31 \\
    CL4ST$^\star$ \cite{art_709} & 18.49 & 12.00 & 30.17 \\
    Crossformer$^\star$ & 20.40 & 14.62 & 32.79 \\
    DCGCN \cite{art_117} & 20.28 & - & 31.65 \\
    DCRNN & 24.70 & 17.12 & 38.12 \\
    DGCRN$^\star$ & 19.04 & 12.80 & 30.82 \\
    DMGF-Net \cite{art_99} & 20.59 & 13.63 & 32.43 \\
    DSA-NET & 22.79 & 16.03 & 35.77 \\
    DSTAGNN \cite{art_675} & 19.30 & 12.70 & 31.46 \\
    DSTGN$^\star$ \cite{art_73} & 18.61 & 12.31 & 30.79 \\
    DSTIGNN$^\star$ \cite{art_74} & 18.41 & 12.45 & 29.97 \\
    DyHSL & 17.66 & 12.42 & 29.46 \\
    EnhanceNet & 20.44 & 13.58 & 32.37 \\
    FEDFormer$^\star$ & 22.86 & 16.04 & 35.07 \\
    FOGS$^\star$ & 19.74 & 13.05 & 31.66 \\
    FourierGNN & 22.98 & 15.14 & 36.23 \\
    Gboot \cite{art_715} & 19.28 & 12.58 & 31.02 \\
    GL-STGTN \cite{art_433} & 18.20 & 12.48 & 29.71 \\
    GMAN$^\star$ & 20.93 & 14.06 & 33.34 \\
    GMSDR$^\star$ & 20.49 & 14.15 & 32.13 \\
    GODERN-FS$^\star$ \cite{art_88} & 19.17 & 12.74 & 30.96 \\
    GRU & 23.68 & 16.44 & 39.27 \\
    GWN & 19.36 & 13.31 & 31.72 \\
    HA & 38.03 & 27.88 & 59.24 \\
    LSGCN & 21.53 & 13.18 & 33.86 \\
    LSTM & 23.81 & 18.12 & 36.62 \\
    MTGNN$^\star$ & 24.89 & 17.29 & 39.66 \\
    MTGODE$^\star$ & 19.55 & 13.08 & 32.99 \\
    PAHNN \cite{art_435} & \underline{16.28} & 11.98 & \underline{27.62} \\
    PDFormer & 25.75 & 17.55 & 42.09 \\
    SCINet$^\star$ & 19.29 & \underline{11.89} & 31.28 \\
    SDGL$^\star$ \cite{art_78} & 18.65 & 12.38 & 31.30 \\
    STAGCN \cite{art_130} & 19.02 & 12.46 & 30.75 \\
    STCGNN \cite{art_115} & 19.39 & 12.71 & 31.17 \\
    StemGNN$^\star$ & 21.61 & 16.10 & 33.80 \\
    STExplainer$^\star$ \cite{art_700} & 18.57 & 12.13 & 30.14 \\
    STFGCN & 19.83 & 13.02 & 31.88 \\
    STG2Seq & 25.20 & 18.77 & 38.48 \\
    STGCN & 21.16 & 13.83 & 34.89 \\
    STGDN \cite{art_707} & 19.10 & 12.60 & 30.83 \\
    STGNCDE$^\star$ & 19.21 & 12.76 & 31.09 \\
    STGODE & 20.84 & 13.76 & 32.84 \\
    ST-GTNN & 20.08 & 13.31 & 32.18 \\
    STIDGCN$^\star$ & 18.47 & 12.42 & 29.90 \\
    STNorm & 19.21 & 13.05 & 32.30 \\
    STSGCN & 21.19 & 13.90 & 33.65 \\
    STWave \cite{art_671} & 18.50 & 12.43 & 30.39 \\
    STWawe+ \cite{art_453} & 18.25 & 12.21 & 30.14 \\
    SVR & 28.70 & 19.20 & 44.56 \\
    TAMP-S2GCNets$^\star$ & 19.74 & 13.22 & 31.74 \\
    TCN & 23.22 & 15.59 & 37.26 \\
    TimesNet & 21.63 & 13.15 & 34.90 \\
    Transformer & 21.10 & 15.13 & 31.46 \\
    TSGDC \cite{art_424} & 18.80 & 12.67 & 31.08 \\
    VAR & 23.75 & 18.09 & 36.66 \\
    W-DSTAGNN \cite{art_419} & 19.30 & 12.70 & 31.28 \\
    Z-GCNETS \cite{art_657} & 19.50 & 12.78 & 31.61 \\
    \bottomrule
\end{longtable}

The MAE, MAPE and RMSE of the different models on the PeMS07 dataset, with data from 1$^{\text{st}}$ May 2017 to 31$^{\text{st}}$ August 2017, a 5-minute granularity and a 60-minute time horizon, are presented in Tab.~\ref{t:risultati pems07}.

\begin{longtable}{lccc}
    \caption{Comparison of the average accuracy of the different models on the PeMS07 dataset for a 60-minute time horizon, expressed in terms of MAE, MAPE and RMSE. The smallest errors are underlined. Numbers from the original papers.}
    \label{t:risultati pems07}
    \endfirsthead
    \endhead
    \toprule
    Time horizon & \multicolumn{3}{c}{60 min} \\
    Metrics & MAE & MAPE & RMSE\\
    \midrule
    AdaSTNet & 21.33 & 8.94 & 35.04 \\
    AGCRN & 22.37 & 9.12 & 36.55 \\
    AMGCN \cite{art_444} & 19.83 & 8.19 & 31.03 \\
    ARIMA & 38.17 & 19.46 & 59.27 \\
    ASTGCN & 24.01 & 10.73 & 37.87 \\
    Auto-DSTSGN \cite{art_108} & 20.08 & 8.57 & 33.02 \\
    AutoSTG & 23.22 & 9.95 & 36.47 \\
    AutoSTS \cite{art_720} & 20.26 & 8.54 & 33.09 \\
    CL4ST$^\star$ \cite{art_709} & 20.20 & 8.53 & 34.06 \\
    DCGCN \cite{art_117} & 22.06 & - & 34.66 \\
    DCRNN & 25.22 & 11.82 & 38.61 \\
    DGCRN & 19.91 & 8.32 & 31.47 \\
    DMSTG & 19.98 & 8.56 & 33.08 \\
    DMSTGCN & 20.77 & 8.94 & 33.67 \\
    DSTAGNN \cite{art_675} & 21.42 & 9.01 & 34.51 \\
    DSTCGCN & 20.37 & 8.64 & 33.47 \\
    DyHSL & 18.84 & 8.11 & 31.65 \\
    Gboot & 21.35 & 9.02 & 34.43 \\
    GMAN & 21.56 & 9.51 & 34.97 \\
    GMSDR$^\star$ &22.27 & 9.86 & 34.94 \\
    GWN & 26.85 & 12.12 & 42.78 \\
    FOGS$^\star$ & 21.28 & 8.95 & 34.88 \\
    FourierGNN & 25.47 & 10.76 & 39.69 \\
    HA & 45.12 & 24.51 & 65.64 \\
    LSGCN & 27.31 & 11.98 & 41.46 \\
    LSTM & 29.98 & 13.20 & 45.94 \\
    MTGNN & 20.57 & 9.12 & 33.54 \\
    PAHNN \cite{art_435} & \underline{17.35} & \underline{7.64} & \underline{29.78} \\
    PDFormer & 23.92 & 11.62 & 36.76 \\
    PGCN & 20.03 & 8.78 & 32.54 \\
    PYNet \cite{art_735} & 19.61 & 8.36 & 32.85 \\
    RGSL & 20.73 & 8.71 & 34.48 \\
    RST-LTG \cite{art_406} & 19.49 & 8.26 & 32.91 \\
    SLCNN & 20.38 & 8.85 & 33.27 \\
    SPGCL & 31.35 & 18.32 & 46.34 \\
    STAGCN \cite{art_130} & 21.10 & 8.92 & 34.10 \\
    StemGNN & 22.23 & 9.20 & 36.46 \\
    STExplainer$^\star$ \cite{art_700} & 20.00 & 8.51 & 33.45 \\
    STFGCN & 22.07 & 9.21 & 35.80 \\
    STFGNN & 23.46 & 9.21 & 36.60 \\
    STG2Seq & 32.77 & 20.16 & 47.16 \\
    STGBN \cite{art_707} & 21.33 & 8.94 & 35.04 \\
    STGCN & 25.38 & 11.08 & 38.78 \\
    STGNCDE$^\star$ & 20.53 & 8.80 & 33.84 \\
    STGODE & 22.59 & 10.14 & 37.54 \\
    ST-GRAT & 20.69 & 9.06 & 33.01 \\
    STNorm & 20.59 & 8.61 & 34.86 \\
    STSGCN & 24.26 & 10.21 & 39.03 \\
    STWave \cite{art_671} & 19.94 & 8.38 & 33.88 \\
    STWave+ \cite{art_453} & 19.59 & 8.17 & 33.53 \\
    SVR & 32.49 & 14.26 & 50.22 \\
    TAMP-S2GCNets$^\star$ & 21.84 & 9.24 & 35.42 \\
    TCN & 32.72 & 14.26 & 42.23 \\
    TimesNet & 25.12 & 10.60 & 40.71 \\
    Transformer & 22.07 & 10.12 & 34.21 \\
    TSGDC \cite{art_424} & 19.96 & 8.54 & 33.25 \\
    VAR & 50.22 & 32.22 & 75.63 \\
    Z-GCNETs & 21.77 & 9.25 & 35.17 \\
    \bottomrule
\end{longtable}

Finally, Tab.~\ref{t:risultati pems08} displays the accuracy of the various models on the PeMS08 dataset with a 5-minute granularity and data spanning from 1$^{\text{st}}$ July 2016 to 31$^{\text{st}}$ August 2016 for a 60-minute time horizon, expressed in terms of MAE, MAPE and RMSE.

\begin{longtable}{lccc}
    \caption{Comparison of the average accuracy of the different models on the PeMS08 dataset for a 60-minute time horizon, expressed in terms of MAE, MAPE and RMSE. The results of the models marked with $^\star$ are taken from papers in the \enquote{Generic} group. The smallest errors are underlined. Numbers from the original papers.}
    \label{t:risultati pems08}
    \endfirsthead
    \endhead
    \toprule
    Time horizon & \multicolumn{3}{c}{60 min} \\
    Metrics & MAE & MAPE & RMSE\\
    \midrule
    AdaSTNet & 15.72 & 9.98 & 24.72 \\
    AGCRN \cite{art_683} & 15.95 & 10.09 & 25.22 \\
    ARIMA & 31.09 & 22.73 & 44.32 \\
    ASTGCN & 18.25 & 11.64 & 28.06 \\
    ASTGNN & 15.00 & 9.50 & 24.70 \\
    Auto-DSTSGN \cite{art_108} & 14.74 & 9.45 & 23.76 \\
    Autoformer$^\star$ & 21.04 & 15.98 & 31.33 \\
    AutoSTG & 16.37 & 10.36 & 25.46 \\
    AutoSTS \cite{art_720} & 14.65 & 9.49 & 23.52 \\
    CL4ST$^\star$ \cite{art_709} & 14.74 & 9.61 & 24.17 \\
    Crossformer$^\star$ & 16.25 & 11.14 & 26.14 \\
    DCGCN \cite{art_117} & 15.68 & - & 24.39 \\
    DCRNN & 16.82 & 10.92 & 26.36 \\
    DHSL$^\star$ \cite{art_400} & 15.19 & 9.86 & 24.61 \\
    DGCRN & 16.22 & 12.06 & 26.10 \\
    DMGF-Net \cite{art_99} & 16.48 & 10.56 & 25.73 \\
    DMSTG & 13.85 & 9.26 & 23.30 \\
    DSA-NET & 17.14 & 11.32 & 26.96 \\
    DSTAGNN \cite{art_675} & 15.67 & - & 25.37 \\
    DSTCGCN & 15.18 & 9.68 & 24.49 \\
    DSTIGNN$^\star$ \cite{art_74} & 13.88 & 9.04 & 23.64 \\
    DyHSL & 14.01 & 8.87 & 22.91 \\
    EnhanceNet & 16.33 & 10.39 & 25.56 \\
    FC-GAGA & 19.41 & 13.53 & 30.22 \\
    FEDformer$^\star$ & 19.55 & 13.58 & 29.30 \\
    FNN & 24.24 & 16.21 & 37.57 \\
    FOGS$^\star$ & 15.73 & 9.88 & 24.92 \\
    Gboot \cite{art_715} & 15.54 & 9.76 & 24.53 \\
    GL-STGTN \cite{art_433} & 14.00 & 9.50 & 23.41 \\
    GMAN & 15.31 & 10.13 & 24.92 \\
    GMSDR \cite{art_682} & 16.36 & 10.28 & 25.58 \\
    GODERN-FS$^\star$ \cite{art_88} & 15.59 & 9.95 & 24.79 \\ 
    GRU & 22.00 & 13.33 & 36.24 \\
    GWN & 15.07 & 9.51 & 23.85 \\
    HA & 34.86 & 24.07 & 52.04 \\
    iDCGCN & 15.56 & - & 25.02 \\
    LSGCN & 17.73 & 11.20 & 26.76 \\
    LSTM & 22.20 & 14.20 & 34.06 \\
    MRA-BGCN & 18.87 & 12.96 & 28.56 \\
    MTGNN & 15.71 & 10.03 & 24.62 \\
    MTGODE$^\star$ & 15.61 & 10.15 & 25.96 \\
    PAHNN \cite{art_435} & 13.58 & \underline{8.35} & \underline{22.11} \\
    PYNet \cite{art_735} & 14.03 & 9.39 & 23.84 \\
    RGSL & 15.49 & 9.96 & 24.80 \\
    RST-LTG \cite{art_406} & 13.61 & 9.05 & 23.15 \\
    SCINet$^\star$ & 15.78 & 9.97 & 24.60 \\
    SDGL$^\star$ \cite{art_78} & 14.93 & 9.61 & 24.13 \\ 
    SNAS4MTF \cite{art_408} & 15.07 & 9.67 & 24.00 \\
    STAGCN \cite{art_130} & 15.36 & 9.80 & 24.32 \\
    STCGNN \cite{art_115} & 15.55 & 10.02 & 24.61 \\
    StemGNN & 15.91 & 10.90 & 25.44 \\
    STExplainer$^\star$ \cite{art_700} & 14.59 & 9.80 & 23.91 \\
    STFGNN & 16.64 & 10.60 & 26.25 \\
    STGNCDE$^\star$ & 15.45 & 9.92 & 24.81 \\
    STG2Seq & 20.17 & 17.32 & 30.71 \\
    STGBN \cite{art_707} & 15.35 & 9.71 & 24.18 \\
    STGCN & 17.50 & 11.29 & 27.09 \\
    STGMN & 16.04 & - & 24.18 \\
    STGODE & 16.79 & 10.58 & 26.01 \\
    ST-GTNN & 16.51 & 10.45 & 25.19 \\
    STIDGCN$^\star$ & 14.10 & 9.15 & 23.72 \\
    STMetaNet & 18.75 & 14.49 & 27.50 \\
    STNorm & 15.39 & 9.91 & 24.80 \\
    STSGCN & 17.13 & 10.96 & 26.80 \\
    STWave \cite{art_671} & 13.42 & 8.90 & 23.40 \\
    STWave+ \cite{art_453} & \underline{13.21} & 8.63 & 23.04 \\
    SVR & 23.25 & 14.64 & 36.16 \\
    TAMP-S2GCNets$^\star$ & 13.36 & 10.15 & 25.98 \\
    TCN & 22.72 & 14.03 & 35.79 \\
    TimesNet & 19.01 & 11.83 & 30.65 \\
    VAR & 19.19 & 13.10 & 29.81 \\
    Z-GCNETs$^\star$ \cite{art_657} & 15.76 & 10.01 & 25.11 \\
    \bottomrule
\end{longtable}

As stated above, the results almost consistently indicate that the periodicity aware spatial-temporal adaptive hypergraph neural network (PAHNN) \cite{art_435} proposed by Zhao et al.~in 2024 is the most accurate model. The PAHNN model includes a temporal multi-periodic to time dependencies and periodic features of traffic time series, and a spatial adaptive hypergraph neural network to capture higher-order correlations among nodes.

\subsection{Predictive monitoring}
\label{subsection:Predictive monitoring}
The theme of predictive monitoring is related to sensor-based monitoring of processes, which has become essential in modern industry. The evolution of monitoring systems is concomitant with the rapid growth of the Internet of Things (IoT), which aims at a smarter management across various applications \cite{art_9}. Hence, in this context, multi-sensor systems can be used to detect anomalies in complex scenarios and to monitor the overall status of a system. In the specific, predictive monitoring is a task that involves the continuous observation and analysis of a system in order to forecast its future states and verify whether the predicted outcomes meet certain standards. The predictive monitoring task is important because it can significantly improve efficiency, reliability, and cost of industrial operations. Related data-based approaches include techniques such as anomaly detection, fault diagnosis, and estimation of the equipment's remaining useful life.

Despite the specificity of the field, 40 out of 366 papers address the topic of predictive monitoring. This includes anomaly detection \cite{art_9, art_10, art_477, art_479, art_481, art_483, art_484, art_489, art_648, art_704, art_742, art_747, art_748, art_751}, fault diagnosis \cite{art_53, art_54, art_55, art_57, art_58, art_192, art_469, art_474, art_486, art_487, art_491} and RUL estimation \cite{art_152, art_153, art_154, art_155, art_156, art_157, art_158, art_478, art_488}. The use of GNNs in these analyses is justified by the necessity of using a model capable of capturing the spatio-temporal correlations among the many sensors present in the system.

\subsubsection{Overview}
As said, predictive monitoring is a research field which focuses on several techniques, such as anomaly detection, fault diagnosis and remaining useful life estimation. Anomaly detection is a technique that aims to identify abnormal data that are not due to random deviations from some regular pattern, but are generated by a different underlying mechanism \cite{ref_anomaly_detection_1}. It is a prerequisite part of fault diagnosis \cite{ref_anomaly_detection_2}.
Fault diagnosis is the process of determining whether a fault has occurred in a machine, including the identification of the time, location, type, and severity of the fault. It is usually considered to be a classification problem. As for remaining useful life (RUL) estimation, it aims to determine how long a machine will operate before it needs to be repaired or replaced, and it is useful for scheduling maintenance interventions.

The selected papers implement these three techniques, with anomaly detection being less studied than the other two. The main applications developed are industrial (e.g., bearings fault diagnosis \cite{art_54, art_55, art_57, art_155, art_156}) and mechanical (e.g, engines RUL estimation \cite{art_152, art_153, art_154, art_157, art_478, art_488}). There are also some papers dealing with energy related tasks, specifically fault diagnosis and degradation analysis of photovoltaic systems \cite{art_53, art_690} and of energy networks \cite{art_487}. They are included in this thematic group because of their specific focus on fault classification rather than on energy-related time series forecasting. It should also be underlined that the majority of these papers aim at solving a classification task, which is typical of anomaly detection and fault diagnosis. As for RUL estimation, the quantity to be predicted is the RUL itself. For this reason, we do not make specific reference to forecasting time horizons, unlike in the other subsections, since the goal here is to estimate the RUL rather than to forecast data over future time intervals.

The majority of the collected papers related to this theme were published in 2024, and the most popular journal for these topics is \giornale{IEEE Transactions on Instrumentation and Measurement} by IEEE, with 6 out of 35 papers. It can be observed that the majority of the articles are published in IEEE journals. As for conference venues, 4 out of 7 conference papers were presented at \conferenza{CIKM}.
    
\subsubsection{Datasets}
Tab.~\ref{t:link dataset predictive monitoring} contains a list of the public datasets used in the selected papers, with the links to access them.

\begin{longtable}{R{3.7cm}R{3cm}p{7.5cm}}
    \caption{List of public datasets in the \enquote{Predictive monitoring} group and their corresponding links.}
    \label{t:link dataset predictive monitoring}
    \endfirsthead
    \endhead
    \toprule
    Dataset & Used by & Link \\
    \midrule 
    Air quality China & \cite{art_9} & \url{https://www.aqistudy.cn/historydata/} \\
    Air quality - TPS & \cite{art_9} & \url{https://www.kaggle.com/amritpal333/tps-july-2021-original-dataset-clean} \\
    C-MAPSS & \cite{art_152}, \cite{art_153}, \cite{art_154}, \cite{art_157}, \cite{art_478}, \cite{art_488} & \url{https://ntrs.nasa.gov/api/citations/20070034949/downloads/20070034949.pdf} \\
    CWRU bearing dataset & \cite{art_55}, \cite{art_192}, \cite{art_491} & \url{http://csegroups.case.edu/bearingdatacenter/home} \\
    HAI security dataset & \cite{art_648} & \url{https://github.com/icsdataset/hai} \\
    Herbert river water level & \cite{art_483} & \url{https://zenodo.org/records/8053359} \\
    Intelligent maintenance system (IMS) & \cite{art_475} & \url{https://www.nasa.gov/intelligent-systems-division} \\
    JNU bearing dataset & \cite{art_55} & \url{http://mad-net.org:8765/explore.html?id=9&t=0.9355271549540183} $^\dagger$ \\
    MFPT & \cite{art_192} & \url{https://api.semanticscholar.org/CorpusID:26870683} (source paper) \\
    Mars science laboratory rover (MSL) & \cite{art_481}, \cite{art_489}, \cite{art_704}, \cite{art_748} & \url{https://doi.org/10.1145/3219819.3219845} (source paper) \\
    NASA battery dataset & \cite{art_158} & \url{http://ti.arc.nasa.gov/project/prognostic-data-repository} \\
    NASA prognostics data & \cite{art_58} & \url{https://ti.arc.nasa.gov/tech/dash/groups/pcoe/prognostic-data-repository/#bearing} \\
    N-CMAPSS & \cite{art_152}, \cite{art_154} & \url{https://ti.arc.nasa.gov/tech/dash/groups/pcoe/prognostic-data-repository/} \\
    NREL PV & \cite{art_53} & \url{https://www.nrel.gov/} \\
    PRONOSTIA & \cite{art_155}, \cite{art_156} & \url{https://hal.science/hal-00719503/document} (source paper) \\
    Pooled server metrics (PSM) & \cite{art_479}, \cite{art_481} & \url{https://doi.org/10.1145/3447548.3467174} (source paper) \\
    PV degradation & \cite{art_690} & \url{https://github.com/Yangxin666/ST-GTrend/tree/main/Datasets} \\
    SEU bearing dataset & \cite{art_55} & \url{https://github.com/cathysiyu/Mechanical-datasets} \\
    Soil moisture active passive satellite (SMAP) & \cite{art_479}, \cite{art_481}, \cite{art_484}, \cite{art_489}, \cite{art_748} & \url{https://doi.org/10.1145/3219819.3219845} (source paper) \\
    Server machine dataset (SMD) & \cite{art_479}, \cite{art_484}, \cite{art_704} & \url{https://github.com/NetManAIOps/OmniAnomaly} \\
    Secure water treatment (SWaT) & \cite{art_477}, \cite{art_479}, \cite{art_481}, \cite{art_484}, \cite{art_489}, \cite{art_648}, \cite{art_704}, \cite{art_742}, \cite{art_747}, \cite{art_748} & \url{https://itrust.sutd.edu.sg/itrust-labs_datasets/dataset_info/} \\
    Water distribution (WaDi) & \cite{art_477}, \cite{art_479}, \cite{art_484}, \cite{art_489}, \cite{art_648}, \cite{art_704}, \cite{art_742}, \cite{art_747}, \cite{art_748} & \url{https://itrust.sutd.edu.sg/itrust-labs_datasets/dataset_info/} \\
    Wireless capsule endoscopy images & \cite{art_475} & \url{https://doi.org/10.1055/s-0043-105488} (source paper) \\
    XJTUGearbox and XJTUSpurgear & \cite{art_192}, \cite{art_474} & \url{http://dx.doi.org/10.1016/j.ymssp.2021.108653} (source paper) \\
    \bottomrule
\end{longtable}

The two popular datasets are C-MAPSS and N-CMAPSS, two well-known datasets for RUL prediction provided by NASA. C-MAPSS describes the deterioration of aircraft engines by recording temperature, pressure, and fan speed during the cruise phase (\url{https://ntrs.nasa.gov/api/citations/20070034949/downloads/20070034949.pdf}). N-CMAPSS is an extended version of C-MAPSS, whose records cover climbing, cruising, and descending flight conditions (\url{https://ti.arc.nasa.gov/tech/dash/groups/pcoe/prognostic-data-repository/}). Almost no papers use exogenous variables, with the exception of Ref.~\cite{art_53}, which evaluates the impact of meteorological variables on fault diagnosis of photovoltaic systems. Other popular datasets are SWaT and WaDi. The secure water treatment dataset (SWaT) is a cybersecurity dataset that includes data collected from 51 sensors and actuators over a continuous period of 11 days. During this time, the system operated under normal conditions for 7 days and was subjected to attack scenarios for 4 days. The dataset contains both network traffic and physical process data recorded from all devices. The water distribution (WaDi) dataset is an extension of SWaT involving a larger setup of 127 sensors and actuators monitored for 16 consecutive days. 

The time granularity of the studied time series is highly variable, ranging from 1 hour to fractions of seconds. Many papers report the frequency of the studied sensors, with values ranging from 1 Hz to 50 kHz. Most datasets are normalized, by either using min-max normalization or Z-score normalization. In some papers, a data augmentation procedure is employed to mitigate the natural imbalance between the number of normal and anomalous conditions.

\subsubsection{Types of models}
Almost half of the proposed spatio-temporal GNN models (19 out of 40 papers) fall under the category of pure convolutional GNNs, followed by 15 attentional GNNs, 2 hybrid models, and one recurrent model (Ref.~\cite{art_470}). One paper (Ref.~\cite{art_57}) uses both convolutional and attentional GNNs, one paper (Ref.~\cite{art_475}) uses a variant of the convolutional GNN, and the description of another one (Ref.~\cite{art_155}) does not clarify which class of the taxonomy it belongs to. Almost all the models have a graph learning module, so that the graph structure is rarely defined a priori by the researchers. Three papers \cite{art_55, art_57, art_154} use the visibility graph algorithm proposed by Lacasa et al.~in Ref.~\cite{visibility_graph} for converting the time series into a graph.

The most common loss functions used within this thematic group are the cross entropy function for classification problems, and MSE or RMSE for RUL forecasting problems. Most of the papers mention the Python language and the PyTorch library, but only the papers listed in Tab.~\ref{t:link codici predictive monitoring} provide the links to the related source codes.

\begin{longtable}{p{7cm}p{6.5cm}}
    \caption{List of source codes of the \enquote{Health} models in the review.}
    \label{t:link codici predictive monitoring}
    \endfirsthead
    \endhead
    \toprule
    Model & Link \\
    \midrule
    Causality enhanced global-local graph neural network (CEGLo-GNN) \cite{art_482} & \url{https://github.com/YueYueXia/CEGLo-GNN} \\
    Dual time-oriented graph attention networks (DuoGAT) \cite{art_748} & \url{https://github.com/ByeongtaePark/DuoGAT} \\
    Graph deviation network+ (GDN+) \cite{art_483} & \url{https://github.com/KatieBuc/gnnad/releases} \\
    Graph network with causal connectivity and temporal convolutional neural network feature extractor \cite{art_155} & \url{https://github.com/mylonasc/gnn-tcnn} \\
    Spectral multi-view causal graph fusion based anomaly detection (SMV-CGAD) \cite{art_742} & \url{https://github.com/arunvignesh28/SMV-CGAD} $^\dagger$ \\
    Spatio-temporal graph neural network empowered long-term trend analysis system (ST-GTrend) \cite{art_690} & \url{https://github.com/Yangxin666/ST-GTrend} \\
    Graph attention networks with topological analysis (TopoGDN) \cite{art_704} & \url{https://github.com/ljj-cyber/TopoGDN} \\
    \bottomrule
\end{longtable}

\subsubsection{Benchmark models}
Tab.~\ref{t:benchmark models predictive monitoring} presents the benchmark models employed in the \enquote{Predictive monitoring} thematic group by at least two articles examined.

\begin{longtable}{p{2.5cm}p{6.5cm}p{5.2cm}}
    \caption{List of benchmark models in the \enquote{Predictive monitoring} group divided per category.}
    \label{t:benchmark models predictive monitoring}
    \endfirsthead
    \endhead
    \toprule
    Category & Model & Used by \\
    \midrule
    \multirow[t]{3}{=}{Mathematical and statistical methods} & Autoregressive integrated moving average (ARIMA) & \cite{art_483}, \cite{art_487} \\
    & & \\
    \midrule
    \multirow[t]{4}{=}{Non-GNN machine learning methods} & Auto-encoder (AE) & \cite{art_484}, \cite{art_648}, \cite{art_748}, \cite{art_751} \\
    & Feature-attention based bidirectional GRU and CNN model (AGCNN) \cite{AGCNN} & \cite{art_153}, \cite{art_154}, \cite{art_157}, \cite{art_478}, \cite{art_488} \\
    & Attention-based long-short term memory network & \cite{art_153}, \cite{art_157}, \cite{art_488} \\
    & Anomalous rhythm detection using adversarially generated time series (BeatGAN) \cite{BeatGAN} & \cite{art_481}, \cite{art_484} \\
    & Bidirectional long-short term memory network (BiLSTM) \cite{BiLSTM} & \cite{art_152}, \cite{art_153}, \cite{art_154}, \cite{art_157}, \cite{art_470}, \cite{art_488} \\
    & Convolutional neural network (CNN) & \cite{art_9}, \cite{art_152}, \cite{art_153}, \cite{art_154}, \cite{art_156}, \cite{art_157}, \cite{art_461}, \cite{art_469}, \cite{art_486}, \cite{art_487}, \cite{art_488} \\
    & Convolutional neural network with long-short term memory network (CNN-LSTM) & \cite{art_154}, \cite{art_470} \\
    & Crossformer \cite{Crossformer} & \cite{art_152}, \cite{art_482} \\
    & Deep autoencoding gaussian model (DAGMM) \cite{DAGMM} & \cite{art_481}, \cite{art_484}, \cite{art_648}, \cite{art_747}, \cite{art_748} \\
    & Deep belief network (DBN) \cite{DBN} & \cite{art_157}, \cite{art_488} \\
    & Decision tree (DT) & \cite{art_9}, \cite{art_487} \\
    & Feed-forward neural network (FNN) & \cite{art_55}, \cite{art_155}, \cite{art_469}, \cite{art_478} \\
    & Genetic algorithm approach on restricted Boltzmann-based long-short term memory network (GA+RBM+LSTM) \cite{GA+RBM+LSTM} & \cite{art_157}, \cite{art_488} \\
    & Gradient boosting (GB) & \cite{art_154}, \cite{art_487} \\
    & Gated recurrent unit (GRU) & \cite{art_53}, \cite{art_156}, \cite{art_469}, \cite{art_487} \\
    & Integrated multi-head dual sparse self-attention network (IMDSSN) \cite{IMDSSN} & \cite{art_154}, \cite{art_478}, \cite{art_488} \\
    & Isolation forest & \cite{art_10}, \cite{art_481}, \cite{art_742} \\
    & $k$-nearest neighbors (KNN) & 
    \cite{art_475}, \cite{art_487}, \cite{art_648}, \cite{art_748} \\
    & Local outlier factor (LOF) \cite{LOF} & \cite{art_475}, \cite{art_481} \\
    & Long-short term memory network (LSTM) & \cite{art_55}, \cite{art_154}, \cite{art_157}, \cite{art_158}, \cite{art_461}, \cite{art_470}, \cite{art_487}, \cite{art_488} \\
    & Long-short term memory-based variational auto-encoder (LSTM-VAE) & \cite{art_10}, \cite{art_477}, \cite{art_479}, \cite{art_484}, \cite{art_489}, \cite{art_648}, \cite{art_704}, \cite{art_751} \\
    & Multivariate anomaly detection with GAN (MAD-GAN) \cite{MAD-GAN} & \cite{art_10}, \cite{art_648}, \cite{art_704}, \cite{art_748} \\
    & OmniAnomaly \cite{OmniAnomaly} & \cite{art_477}, \cite{art_479}, \cite{art_481}, \cite{art_484}, \cite{art_489}, \cite{art_704} \\
    & Random forest (RF) & \cite{art_154}, \cite{art_478}, \cite{art_487} \\
    & Support vector machine (SVM) & \cite{art_9}, \cite{art_648}, \cite{art_478}, \cite{art_486}, \cite{art_487}, \cite{art_742}, \cite{art_748}, \cite{art_751} \\
    & Deep transformer networks for anomaly detection (TranAD) \cite{TranAD} & \cite{art_484}, \cite{art_489}, \cite{art_704}, \cite{art_747} \\
    & Transformer \cite{attention} & \cite{art_152}, \cite{art_154}, \cite{art_470}, \cite{art_478},  \cite{art_479}, \cite{art_488}, \cite{art_489} \\
    & Unsupervised anomaly detection on multivariate time series (USAD) \cite{USAD} & \cite{art_479}, \cite{art_484}, \cite{art_489}, \cite{art_648}, \cite{art_748}, \cite{art_751} \\
    \midrule 
    \multirow[t]{1}{=}{GNN methods} & Adaptive graph convolutional recurrent network (AGCRN) \cite{art_683} & \cite{art_58}, \cite{art_482} \\
    & Diffusion convolutional recurrent neural network (DCRNN) \cite{DCRNN} & \cite{art_480}, \cite{art_482} \\
    & Fused sparse autoencoder and graph network (FuSAGNet) \cite{art_648} & \cite{art_10}, \cite{art_479}, \cite{art_747} \\
    & Graph attention network (GAT) \cite{GAT} & \cite{art_54}, \cite{art_192}, \cite{art_470}, \cite{art_487} \\
    & Graph convolutional network (GCN) & \cite{art_54}, \cite{art_55}, \cite{art_156}, \cite{art_192}, \cite{art_461}, \cite{art_474}, \cite{art_486} \\
    & Graph deviation network (GDN) \cite{GDN} & \cite{art_10}, \cite{art_477}, \cite{art_479}, \cite{art_481}, \cite{art_484}, \cite{art_648}, \cite{art_704}, \cite{art_747}, \cite{art_748}, \cite{art_751} \\
    & Gated graph convolutional network (GGCN) \cite{GGCN} & \cite{art_153}, \cite{art_478} \\
    & Graph WaveNet (GWN) \cite{GWN} & \cite{art_58}, \cite{art_480}, \cite{art_482} \\
    & Hierarchical attention graph convolutional network (HAGCN) \cite{HAGCN} & \cite{art_152}, \cite{art_153}, \cite{art_478} \\
    & Multivariate time-series anomaly detection via graph attention network (MTAD-GAT) \cite{MTAD-GAT} & \cite{art_479}, \cite{art_648}, \cite{art_747}, \cite{art_748} \\
    & Multivariate time series forecasting with graph neural network (MTGNN) \cite{art_655} & \cite{art_470}, \cite{art_483} \\
    & Spatio-temporal fusion attention (STFA) \cite{art_157} & \cite{art_153}, \cite{art_154}, \cite{art_478}, \cite{art_488} \\
    & Spatio-temporal graph convolutional network (STGCN) \cite{STGCN} & \cite{art_58}, \cite{art_152}, \cite{art_470}, \cite{art_478}, \cite{art_480} \\
    \bottomrule
\end{longtable}

In this thematic group there is a high variety of benchmarks used, and before 2024 only few papers share a benchmark model in the GNN category. The most popular benchmarks are two classic machine learning models, the CNN and the LSTM network. Not surprisingly, almost no statistical or mathematical benchmark models were used, since they are more frequently used in forecasting problems than in classification problems.

Among the GNN benchmarks, some of them have been specifically developed for remaining useful life prediction or predictive monitoring tasks, such as the hierarchical attention graph convolutional network (HAGCN) \cite{HAGCN}, the spatio-temporal fusion attention (STFA) \cite{art_157}, and the graph deviation network (GCN) \cite{GDN}. In the HAGCN model, proposed by Li et al.~in 2021, the spatial dependencies are modeled by a hierarchical graph representation layer, and a bidirectional long-short term memory network is used for modeling temporal dependencies of sensor measurements. As for the STFA model, proposed by Kong et al.~in 2022, it integrates some \textit{a priori} knowledge about the equipment's structure with a spatio-temporal deep learning architecture which employs LSTM cells and an attention mechanism. Finally, the GDN model, proposed by Deng et al.~in 2021, learns a graph structure that represents the relationships between sensors, and aims to detect deviations from these patterns by using an attentional GNN. The source code for the GDN model is available at \url{https://github.com/d-ailin/GDN}.

\subsubsection{Results}
The reported results show that the proposed GNN models clearly outperform the benchmarks. The accuracy of the models is evaluated using different metrics, depending on whether the context is classification or forecasting. For classification tasks, several studies present confusion matrices and calculate accuracy percentages. For RUL forecasting problems instead, the most commonly used metrics are the RMSE and the score function.

The only results that can be compared across many papers are those for the C-MAPSS dataset. Tab.~\ref{t:risultati c-mapss} shows the accuracy of the different models, expressed in terms of RMSE and score function.

{\tabcolsep=1pt
\begin{longtable}{lcccccccc}
    \caption{Comparison of the accuracy of the different models on the C-MAPSS dataset on sub-datasets FD001–FD004, expressed in terms of RMSE and score function. The smallest errors are underlined. Numbers from the original papers.}
    \label{t:risultati c-mapss}
    \endfirsthead
    \endhead
    \toprule
    Sub-dataset & \multicolumn{2}{c}{FD001} & \multicolumn{2}{c}{FD002} & \multicolumn{2}{c}{FD003} & \multicolumn{2}{c}{FD004} \\
    Metrics & RMSE & Score & RMSE & Score & RMSE & Score & RMSE & Score \\
    \midrule 
    1D CNN & 12.61 & 273.70 & 22.36 & 10412.00 & 12.64 & 284.10 & 23.31 & 12466.00 \\
    2D CNN & 18.45 & 1286.70 & 30.29 & 13570.00 & 19.82 & 1596.20 & 29.16 & 7886.40 \\
    AGCNN & 12.42 & 225.51 & 19.43 & 1492.76 & 13.39 & 227.09 & 21.50 & 3392.60 \\
    ATT-LSTM & 13.95 & 320.00 & 17.65 & 2102.00 & 12.72 & 223.00 & 20.21 & 3100.00 \\
    BiGRU-AS & 13.68 & 284.00 & 20.81 & 2454.00 & 15.53 & 428.00 & 27.31 & 4708.00 \\
    Bi-LSTM & 13.65 & 295.00 & 23.18 & 4130.00 & 13.74 & 317.00 & 24.86 & 5430.00 \\
    BiLSTM-ED & 14.74 & 273.00 & 22.07 & 3099.00 & 17.48 & 574.00 & 23.49 & 3202.00 \\
    BLCNN & 13.18 & 302.28 & 19.09 & 1557.56 & 13.75 & 381.37 & 20.97 & 3858.78 \\
    CDSG \cite{art_154} & 11.26 & 188.00 & 18.13 & 1740.00 & 12.03 & 218.00 & 19.73 & 2332.00 \\
    ChebyNet-EdgePool & 15.21 & 450.22 & 16.28 & 1229.22 & 14.67 & 421.02 & 16.26 & 1159.74 \\
    CNN-LSTM & 14.40 & 290.00 & 27.23 & 9869.00 & 14.32 & 316.00 & 26.69 & 6594.00 \\
    CNN-Transformer & 12.25 & 198.00 & 17.08 & 1575.00 & 13.39 & 290.00 & 19.86 & 1741.00 \\
    ConvGAT \cite{art_153} & 11.34 & 197.43 & 14.12 & \underline{771.61} & 10.97 & 235.26 & \underline{15.51} & 1231.17 \\
    C-Transformer & 13.79 & 475.46 & 16.11 & 2214.59 & 17.10 & 939.10 & 19.77 & 3237.37 \\
    DBN & 15.21 & 417.59 & 27.12 & 9031.64 & 14.71 & 442.43 & 29.88 & 7954.51 \\
    DCNN & 12.61 & 273.70 & 22.36 & 10412.00 & 12.64 & 284.10 & 23.31 & 12466.00 \\
    DSAN & 13.40 & 242.00 & 22.06 & 2869.00 & 15.12 & 497.00 & 21.03 & 2677.00 \\
    DS-STFN \cite{art_478} & \underline{10.92} & \underline{161.35} & \underline{13.77} & 946.34 & \underline{10.01} & \underline{150.42} & 15.53 & \underline{1079.91} \\
    Earlier CNN & 18.45 & 1287.00 & 30.29 & 13570.00 & 19.82 & 1596.00 & 29.16 & 7886.00 \\
    ELM & 17.27 & 523.00 & 37.28 & 498149.97 & 18.90 & 573.78 & 38.43 & 121414.00 \\
    GA+RBM+LSTM & 12.56 & 231.00 & 22.73 & 3366.00 & 12.10 & 251.00 & 22.66 & 2840.00 \\
    GA-Transformer & 11.63 & 215.00 & 15.99 & 1133.00 & 11.35 & 228.00 & 20.15 & 2672.00 \\
    GAT-DAT & 13.83 & 318.60 & 14.81 & 1163.80 & 14.85 & 438.50 & 16.80 & 1928.60 \\
    GAT-EdgePool & 13.53 & 309.59 & 15.07 & 1380.41 & 14.66 & 470.74 & 17.60 & 1726.53 \\
    GAT-TopkPool & 13.21 & 303.18 & 17.25 & 5338.80 & 15.36 & 507.52 & 21.44 & 2971.93 \\
    GB & 15.67 & 474.01 & 29.09 & 87280.06 & 16.84 & 576.72 & 29.01 & 17817.92 \\
    GGCN & 11.82 & 186.70 & 17.24 & 1493.70 & 12.21 & 245.19 & 17.36 & 1371.50 \\
    HAGCN & 11.93 & 222.30 & 15.05 & 1144.10 & 11.53 & 240.30 & 15.74 & 1218.60 \\
    HDNN & 13.02 & 245.00 & 15.24 & 1282.42 & 12.22 & 287.72 & 18.16 & 1527.42 \\
    HGNN-ACGF & 12.58 & 218.04 & 21.67 & 4584.97 & 12.40 & 248.47 & 22.43 & 2737.86 \\
    IMDSSN & 12.14 & 206.11 & 17.40 & 1775.15 & 12.35 & 229.54 & 19.78 & 2852.81 \\
    LSTM & 16.14 & 338.00 & 24.49 & 4450.00 & 16.18 & 852.00 & 28.17 & 5550.00 \\
    LSTMBS & 14.89 & 481.00 & 26.86 & 7982.00 & 15.11 & 493.00 & 27.11 & 5200.00 \\
    MODBN & 15.04 & 334.23 & 25.05 & 5585.00 & 12.51 & 421.91 & 28.66 & 6557.00 \\
    MS-CNN & 11.44 & 196.22 & 19.35 & 3747.00 & 11.67 & 241.89 & 22.22 & 4844.00 \\
    RF & 17.91 & 479.75 & 29.59 & 70456.86 & 20.27 & 711.13 & 31.12 & 46567.63 \\
    SBI & 13.58 & 228.00 & 19.59 & 2650.00 & 19.16 & 1727.00 & 22.15 & 2901.00 \\
    SMDN & 13.40 & 272.00 & - & - & - & - & 23.40 & 4302.00 \\
    STFA \cite{art_157} & 11.35 & 194.44 & 19.17 & 2493.09 & 11.64 & 224.53 & 21.41 & 2760.13 \\
    STP-GNN \cite{art_488} & 11.38 & 192.17 & 15.81 & 1094.81 & 11.02 & 199.58 & 17.77 & 1692.19 \\
    TATFA-Transformer & 12.21 & 261.50 & 15.07 & 1359.70 & 11.23 & 210.21 & 18.81 & 2506.35 \\
    Transformer & 12.92 & 276.27 & 22.64 & 1960.25 & 13.04 & 284.41 & 24.81 & 3892.24 \\
    \bottomrule
\end{longtable}
}

In general, the best results on the different sub-datasets are achieved by the convolution-graph attention network (ConvGAT) \cite{art_153} and the the dual-stream spatio-temporal fusion network (DS-STFN) \cite{art_478} models. The ConvGAT model, proposed by Chen et al.~in 2023, applies independent convolutional kernels to each feature of the input time series, and passes the resulting values to a graph attention network, with graph structure defined using cosine similarity. Finally, a fully connected layer outputs the RUL prediction. The DS-STFN model, proposed by Zhang et al.~in 2024, organizes the data into grid-structured and graph-structured formats, with the graph structure defined by cosine similarity. A temporal convolutional network with self-attention extracts temporal features from the grid data, and, in parallel, a convolutional graph transformer module captures spatial features from the graph data. The spatial and temporal features are then fused to predict the RUL.

Other pairs of papers with comparable results are Refs.~\cite{art_648} and \cite{art_748} for SWaT and WaDi datasets and Refs.~\cite{art_481} and \cite{art_704} for the MSL dataset. However, due to space limitations, tables comparing just two studies are not included here.

\subsection{Generic}
\label{subsection:Generic}
This subsection includes all the papers that do not specifically address a problem in the specific themes discussed so far. These papers present results on benchmark case studies, for which the related datasets span different thematic groups at once. In particular, the focus is on papers that do not discuss or mention any specific case study prior to the experimental section, or analyze datasets from multiple fields. The reason for this choice is that they are considered to explore broader methodologies or frameworks that are applied to diverse datasets without a declared focus on a particular case.

For instance, some subsets of papers within the \enquote{Generic} group focus on forecasting in presence of missing data (e.g., Refs.~\cite{art_642, art_644}), or of irregularly sampled time series (e.g., Refs.~\cite{art_643, art_645, art_646, art_666, art_710}). Notably, the majority of these studies have been published in conferences rather than in journals.

Moreover, some papers in this section also tackle some theoretical issues. For example, Ref.~\cite{art_685} proves a theorem that provides a theoretical bound on the approximation error of the module that learns the dynamic variable dependence. Another example can be found in Ref.~\cite{art_703}, where the authors  theoretically establish that the generalization error bound of their model is related to various components of the GAT model, such as the number of attention heads, the number of neighboring nodes, the Frobenius norm of the weight matrices, and the norm of the input features.

\subsubsection{Overview}
A significant number of papers (77 out of 366) have been categorized under the \enquote{Generic} thematic group. The majority of these works commit themselves to address a multivariate time series forecasting problem not necessarily limited to a specific field. This \enquote{Generic} set includes 4 papers from 2002, 7 from 2021, 11 from 2022, 28 from 2023, and 27 from 2024, highlighting the recent interest in applying GNNs in cross-disciplinary contexts. These papers are scattered across various journals, with the exception of \giornale{Expert Systems with Applications} by Elsevier, \giornale{IEEE Transactions on Knowledge and Data Engineering} by IEEE, and \giornale{Information Sciences} by Elsevier, which published 4 and 3 of the selected papers, respectively. The three conference venues with the highest number of papers are \conferenza{NeurIPS} with 6 papers, \conferenza{CIKM} with 5 papers, and \conferenza{KDD} with 4 papers. 

\subsubsection{Datasets}
The selected papers make use of the datasets listed in Tab.~\ref{t:link dataset generic} in their experiments.

\begin{longtable}{R{3.7cm}R{3cm}p{7.5cm}}
    \caption{List of public datasets in the \enquote{Generic} papers and their corresponding links.}
    \label{t:link dataset generic}
    \endfirsthead
    \endhead
    \toprule
    Dataset & Used by & Link \\
    \midrule 
    Beijing Traffic & \cite{art_84} & \url{https://github.com/BuaaPercy/Traffic-DataSet-of-Beijing-Road} \\
    Bytom & \cite{art_657}, \cite{art_674}, \cite{art_686} & \url{https://etherscan.io/} \\
    CCMP wind data & \cite{art_73} & \url{http://data.remss.com} \\
    CDC US States & \cite{art_401} & \url{https://gis.cdc.gov/grasp/fluview/fluportaldashboard.html} \\
    CER-E & \cite{art_689} & \url{https://www.ucd.ie/issda/data/commissionforenergyregulationcer/} $^\dagger$ \\
    Chickenpox & \cite{art_160}, \cite{art_705} & \url{https://www.kaggle.com/datasets/die9origephit/chickenpox-cases-hungary} \\
    China AQI & \cite{art_642} & \url{https://quotsoft.net/air/} \\
    CI earthquakes & \cite{art_160} & \url{https://doi.org/10.5194/adgeo-43-31-2016} (source paper) \\
    C-MAPSS & \cite{art_403} & \url{https://ntrs.nasa.gov/api/citations/20070034949/downloads/20070034949.pdf} \\
    CORAv2.0 & \cite{art_396} & \url{https://mds.nmdis.org.cn/pages/home.html} \\
    COVID-19 CA & \cite{art_674}, \cite{art_686}, \cite{art_688}, \cite{art_701} & \url{https://www.dropbox.com/sh/n0ajd5l0tdeyb80/AABGn-ejfV1YtRwjf_L0AOsNa?dl=0} \\
    COVID-19 England & \cite{art_705} & \url{https://pytorch-geometric-temporal.readthedocs.io/en/latest/notes/introduction.html#benchmark-datasets} \\
    COVID-19 TX & \cite{art_674}, \cite{art_686} & \url{https://www.dropbox.com/sh/n0ajd5l0tdeyb80/AABGn-ejfV1YtRwjf_L0AOsNa?dl=0} \\
    CSSE COVID-19 & \cite{art_640}, \cite{art_688} & \url{https://github.com/CSSEGISandData/COVID-19} \\
    CW earthquakes & \cite{art_89} & \url{https://doi.org/10.5194/adgeo-43-31-2016} (source paper) \\
    Decentraland & \cite{art_657}, \cite{art_674} & \url{https://etherscan.io/} \\
    ECG5000 & \cite{art_397}, \cite{art_640}, \cite{art_684}, \cite{art_688} & \url{https://www.cs.ucr.edu/\~eamonn/time_series_data/} \\
    Electricity consumption & \cite{art_73}, \cite{art_74}, \cite{art_75}, \cite{art_78}, \cite{art_79}, \cite{art_80}, \cite{art_81}, \cite{art_83}, \cite{art_88}, \cite{art_90}, \cite{art_405}, \cite{art_647}, \cite{art_649}, \cite{art_650}, \cite{art_655}, \cite{art_656}, \cite{art_684}, \cite{art_685}, \cite{art_687}, \cite{art_688}, \cite{art_702} & \url{https://github.com/laiguokun/multivariate-time-series-data} \\
    Electricity consuming load (ECL) & \cite{art_82}, \cite{art_85}, \cite{art_696} & \url{https://archive.ics.uci.edu/ml/datasets/ElectricityLoadDiagrams20112014} \\
    Electricity transformer temperature (ETT) & \cite{art_76}, \cite{art_82}, \cite{art_85}, \cite{art_393}, \cite{art_417}, \cite{art_687}, \cite{art_696} & \url{https://github.com/zhouhaoyi/ETDataset} \\
    Energy & \cite{art_77}, \cite{art_388} & \url{http://dx.doi.org/10.1016/j.enbuild.2017.01.083} (source paper) \\
    ENG-RAD & \cite{art_644} & \url{https://open-meteo.com/} \\
    Enron  & \cite{art_202} & \url{https://doi.org/10.1073/pnas.1800683115} (source paper) \\
    Eu-Core  & \cite{art_202} & \url{https://doi.org/10.1145/3018661.3018731} (source paper) \\
    European wind generation & \cite{art_649} & \url{https://www.kaggle.com/datasets/sohier/30-years-of-european-wind-generation} \\
    Exchange-rate & \cite{art_73}, \cite{art_74}, \cite{art_76}, \cite{art_77}, \cite{art_78}, \cite{art_79}, \cite{art_80}, \cite{art_81}, \cite{art_85}, \cite{art_87}, \cite{art_89}, \cite{art_90}, \cite{art_393}, \cite{art_405}, \cite{art_649}, \cite{art_655}, \cite{art_656}, \cite{art_685}, \cite{art_687}, \cite{art_702} & \url{https://github.com/laiguokun/multivariate-time-series-data} \\
    Facebook & \cite{art_202} & \url{http://networkrepository.com/socfb} \\
    GEFCOM 2012  & \cite{art_89} & \url{https://doi.org/10.1016/j.ijforecast.2013.07.001} (source paper) \\
    Golem & \cite{art_674} & \url{https://www.dropbox.com/sh/n0ajd5l0tdeyb80/AABGn-ejfV1YtRwjf_L0AOsNa?dl=0} \\
    GraphMSO & \cite{art_644} & \url{https://github.com/marshka/hdtts/tree/main/config/dataset} \\
    Hypertext  & \cite{art_202} & \url{http://networkrepository.com/ia-infect-hyper} (source paper) \\
    HZMetro & \cite{art_650} & \url{https://github.com/HCPLab-SYSU/PVCGN/tree/master/data} \\
    IDWR Japan prefectures & \cite{art_401} & \url{https://www.niid.go.jp/niid/en/idwr-e.html} \\
    ILI  & \cite{art_76}, \cite{art_85}, \cite{art_89} & \url{https://gis.cdc.gov/grasp/fluview/fluportaldashboard.html} \\
    ISRUC-SLEEP & \cite{art_403} & \url{https://sleeptight.isr.uc.pt/ISRUC_Sleep/} \\
    Jena weather & \cite{art_76}, \cite{art_687} & \url{https://www.bgc-jena.mpg.de/wetter/} \\
    KnowAir & \cite{art_400} & \url{https://github.com/shuowang-ai/PM2.5-GNN} \\
    LA Traffic & \cite{art_404} & \url{https://ladot.lacity.gov/residents/traffic-counts} \\
    Loan & \cite{art_202} & \url{https://ai.ppdai.com/mirror/showCompetitionRisk} \\
    Los-loop & \cite{art_383} & \url{https://github.com/lehaifeng/T-GCN/tree/master/data} \\
    Large-scale traffic and weather events (LSTW) & \cite{art_404} & \url{https://smoosavi.org/datasets/lstw} \\
    METR-LA & \cite{art_72}, \cite{art_75}, \cite{art_79}, \cite{art_80}, \cite{art_86}, \cite{art_160}, \cite{art_387}, \cite{art_397}, \cite{art_401}, \cite{art_417}, \cite{art_641}, \cite{art_642}, \cite{art_647}, \cite{art_654}, \cite{art_655}, \cite{art_684}, \cite{art_688}, \cite{art_689}, \cite{art_710} & \url{https://www.kaggle.com/datasets/annnnguyen/metr-la-dataset} \\
    Metro & \cite{art_400} & \url{https://lbs.amap.com/} \\
    MIMIC & \cite{art_71}, \cite{art_643}, \cite{art_646} & \url{https://doi.org/10.1038/sdata.2016.35} (source paper) \\
    Montevideo buses & \cite{art_705} & \url{https://pytorch-geometric-temporal.readthedocs.io/en/latest/notes/introduction.html#benchmark-datasets} \\
    MOOC network & \cite{art_70} & \url{http://snap.stanford.edu/jodie/mooc.csv} \\
    MTM-1 hand motions & \cite{art_705} & \url{https://pytorch-geometric-temporal.readthedocs.io/en/latest/notes/introduction.html#benchmark-datasets} \\
    Nasdaq & \cite{art_77}, \cite{art_80} & \url{https://dl.acm.org/doi/10.5555/3172077.3172254} (source paper) \\
    Neonatal EEG & \cite{art_647} & \url{https://zenodo.org/records/2547147} \\
    NOAA weather & \cite{art_82}, \cite{art_393} & \url{https://www.ncei.noaa.gov/data/local-climatological-data/} \\
    NREL & \cite{art_640}, \cite{art_644}, \cite{art_688} & \url{https://www.nrel.gov/} \\
    NYC-Bike & \cite{art_649}, \cite{art_650}, \cite{art_654}, \cite{art_701} & \url{https://citibikenyc.com/system-data} \\
    NYC-Taxi & \cite{art_398}, \cite{art_639}, \cite{art_649}, \cite{art_650}, \cite{art_701} & \url{https://www1.nyc.gov/site/tlc/about/tlc-trip-record-data.page} \\
    PAMAP2 & \cite{art_159}, \cite{art_645}, \cite{art_680} & \url{https://archive.ics.uci.edu/dataset/231/pamap2+physical+activity+monitoring} \\
    Pedal Me deliveries & \cite{art_705} & \url{https://pytorch-geometric-temporal.readthedocs.io/en/latest/notes/introduction.html#benchmark-datasets} \\
    PeMS03 (PeMSD3) & \cite{art_74}, \cite{art_674}, \cite{art_684}, \cite{art_695} & \url{https://www.kaggle.com/datasets/elmahy/pems-dataset} \\
    PeMS04 (PeMSD4) & \cite{art_73}, \cite{art_74}, \cite{art_78}, \cite{art_79}, \cite{art_88}, \cite{art_396}, \cite{art_397}, \cite{art_400}, \cite{art_405}, \cite{art_642}, \cite{art_657}, \cite{art_674}, \cite{art_684}, \cite{art_686}, \cite{art_700}, \cite{art_709} & \url{https://www.kaggle.com/datasets/elmahy/pems-dataset} \\
    PeMS07 (PeMSD7) & \cite{art_684}, \cite{art_685}, \cite{art_700}, \cite{art_709} & \url{https://www.kaggle.com/datasets/elmahy/pems-dataset} \\
    PeMS08 (PeMSD8) & \cite{art_73}, \cite{art_74}, \cite{art_78}, \cite{art_79}, \cite{art_88}, \cite{art_383}, \cite{art_396}, \cite{art_397}, \cite{art_400}, \cite{art_405}, \cite{art_642}, \cite{art_657}, \cite{art_674}, \cite{art_684}, \cite{art_686}, \cite{art_698}, \cite{art_700}, \cite{art_709} & \url{https://www.kaggle.com/datasets/elmahy/pems-dataset} \\
    PEMS-BAY & \cite{art_72}, \cite{art_75}, \cite{art_160}, \cite{art_401}, \cite{art_641}, \cite{art_642}, \cite{art_647}, \cite{art_655}, \cite{art_684}, \cite{art_685}, \cite{art_689}, \cite{art_695}, \cite{art_710} & \url{https://zenodo.org/records/4263971} \\
    PhysioNet & \cite{art_71}, \cite{art_643}, \cite{art_645}, \cite{art_646} & \url{https://archive.physionet.org/challenge/2012/papers/} \\
    Reddit network & \cite{art_70} & \url{http://snap.stanford.edu/jodie/reddit.csv} \\
    Seattle-loop & \cite{art_401} & \url{https://github.com/zhiyongc/Seattle-Loop-Data} \\
    Sleep-EDF-20 & \cite{art_395} & \url{https://gist.github.com/emadeldeen24/a22691e36759934e53984289a94cb09b} \\
    Solar-Energy & \cite{art_73}, \cite{art_74}, \cite{art_75}, \cite{art_78}, \cite{art_79}, \cite{art_80}, \cite{art_81}, \cite{art_83}, \cite{art_87}, \cite{art_88}, \cite{art_90}, \cite{art_387}, \cite{art_405}, \cite{art_647}, \cite{art_649}, \cite{art_655}, \cite{art_656}, \cite{art_684}, \cite{art_685}, \cite{art_702} & \url{https://github.com/laiguokun/multivariate-time-series-data} \\
    Secure water treatment (SWaT) & \cite{art_387} & \url{https://itrust.sutd.edu.sg/itrust-labs_datasets/} \\
    SHMetro & \cite{art_650} & \url{https://github.com/HCPLab-SYSU/PVCGN/tree/master/data} \\
    Stock market data & \cite{art_703} & \url{https://www.kaggle.com/datasets/paultimothymooney/stock-market-data} \\
    Stocktwits & \cite{art_398} & \url{https://stocktwits.com/} \\
    Synthetic periodic data	synthetic & \cite{art_710} & \url{https://github.com/joeloskarsson/tgnn4i} \\
    TAIEX & \cite{art_84} & \url{https://finance.yahoo.com/} \\
    Taiwan PM2.5 & \cite{art_654} & \url{https://taqm.epa.gov.tw/taqm/en/default.aspx} $^\dagger$ \\
    TaoBao & \cite{art_202} & \url{https://tianchi.aliyun.com/competition/entrance/} \\
    T-drive & \cite{art_383} & \url{https://www.kaggle.com/datasets/arashnic/tdriver} \\
    Tiingo & \cite{art_398} & \url{https://www.tiingo.com/} \\
    Traffic & \cite{art_75}, \cite{art_78}, \cite{art_79}, \cite{art_80}, \cite{art_81}, \cite{art_83}, \cite{art_85}, \cite{art_87}, \cite{art_90}, \cite{art_388}, \cite{art_655}, \cite{art_656}, \cite{art_685}, \cite{art_687}, \cite{art_688}, \cite{art_702} & \url{https://github.com/laiguokun/multivariate-time-series-data} \\
    UCI Beijing PM2.5 & \cite{art_703} & \url{https://archive.ics.uci.edu/ml/datasets/Beijing+PM2.5+Data} $^\dagger$ \\
    UCI human activity recognition & \cite{art_403}, \cite{art_680} & \url{https://archive.ics.uci.edu/dataset/240/human+activity+recognition+using+smartphones} \\
    UCR datasets & \cite{art_711} & \url{https://www.cs.ucr.edu/~eamonn/time_series_data_2018/} \\
    University of East Anglia (UEA) datasets & \cite{art_385}, \cite{art_386}, \cite{art_389}, \cite{art_407}, \cite{art_680} & \url{https://timeseriesclassification.com/} \\
    US historical climatology network (USHCN) & \cite{art_643}, \cite{art_646}, \cite{art_710} & \url{https://www.osti.gov/biblio/1389426} \\
    US stock market price & \cite{art_89} & \url{http://dx.doi.org/10.24963/ijcai.2019/810} (source paper) \\
    Water distribution (WaDi) & \cite{art_387} & \url{https://itrust.sutd.edu.sg/itrust-labs_datasets/} \\
    WESAD dataset & \cite{art_404} & \url{https://ubi29.informatik.uni-siegen.de/usi/data_wesad.html} \\
    Wiki & \cite{art_688} & \url{https://www.kaggle.com/c/web-traffic-time-series-forecasting/data} \\
    Wikipedia network & \cite{art_70} & \url{http://snap.stanford.edu/jodie/wikipedia.csv} \\
    Windmill & \cite{art_705} & \url{https://pytorch-geometric-temporal.readthedocs.io/en/latest/notes/introduction.html#benchmark-datasets} \\
    Wind-speed & \cite{art_76} & \url{https://www.kaggle.com/datasets/fedesoriano/wind-speed-prediction-dataset} \\
    Wind generation & \cite{art_405} & \url{https://www.kaggle.com/datasets/sohier/30-years-of-european-wind-generation} \\
    WISDM & \cite{art_395}, \cite{art_159} & \url{https://archive.ics.uci.edu/dataset/507/wisdm+smartphone+and+smartwatch+activity+and+biometrics+dataset} \\
    \bottomrule
\end{longtable}

It should be noted that multiple recurring datasets can be considered as benchmarks for this community. Some of these were discussed in the previous subsections (e.g., METR-LA, PeMS datasets). Hence, if the results for these datasets were already shown in previous subsections, they will not be discussed again here. Instead, they will be just marked with the symbol $^\star$ in the tables of the previous subsections.

The four most common datasets are Electricity consumption, Exchange-rate, Solar-Energy and Traffic, all available at \url{https://github.com/laiguokun/multivariate-time-series-data}. The Electricity consumption dataset contains the hourly electricity consumption in kWh for 321 customers, recorded from 2012 to 2014. The Traffic dataset describes the road occupancy rates measured by different sensors on the San Francisco Bay area road network from 2015 to 2016, with an hourly granularity. In Solar energy it is recorded the solar power production from 137 photovoltaic plants in Alabama State in 2006, with a 10 minutes granularity. Finally, Exchange rate contains daily exchange rates of 8 countries including Australia, British, Canada, Switzerland, China, Japan, New Zealand and Singapore, from 1990 to 2016.

The University of East Anglia (UEA) dataset includes 30 multivariate time series from many real-world applications, including heart records, electroencephalogram signals, motion data, audio spectra, written characters. Since a large portion of the dataset contains data not related to the focus of this review, the details of the results and models used will not be discussed here. Interested readers are referred to Refs.~\cite{art_385, art_386, art_389, art_407, art_680}.

In general, the data granularity varies significantly, ranging from fractions of seconds to one month, indicating that there is no particular focus on a specific granularity. Individual papers often cover multiple granularities, and many papers take into account datasets with granularity from 5 minutes to 1 day. As for the forecasting horizon, almost all papers focus on multi-step forecasting, usually considering 12 different forecasting horizons at the same time. Only a few papers mention data pre-processing, and usually utilize min-max or Z-score normalization.

\subsubsection{Types of models}
Among the 77 papers in the \enquote{Generic} thematic group, 56 use a purely convolutional GNN approach, and 10 employ a purely attentional model. Five paper lack sufficient detail to be classified within the taxonomy chosen (Refs.~\cite{art_85, art_385, art_386, art_389, art_407}), and another one declares that it uses an aggregating approach that is simpler than either convolutional or attentional approaches \cite{art_81}. Moreover, in Ref.~\cite{art_689} the authors use an approach that can be proved to be equivalent to a convolutional model; Ref.~\cite{art_689} perform a general analysis that extends beyond a single model; Ref.~\cite{art_705} explores various models, all based on convolutional architectures; Ref.~\cite{art_710} proposes an architecture that can adopt either a convolutional or an attentional aggregation. Another paper proposes a so-called multivariate time series with dynamic graph neural ordinary differential equations (MTGODE) model, where the continuous dynamics of simplified graph propagation is described by an ordinary differential equation (ODE) \cite{art_75}. This is not the only paper in this section using differential equations. In Ref.~\cite{art_401}, two domain specific ODEs (one for traffic speed evolution, and the other for disease propagation) are integrated into the convolutional GNN model. The authors argue that this dynamics-informed model approach enhances robustness, allow easier generalization when the equation of the dynamics is known, and reduces the number of model parameters. Additionally, during inference, the model only requires a single data point, as the ODEs explicitly incorporate the immediate dynamics.

As for the definition of the graph structure, the majority of these models use self-learning techniques. This is a straightforward choice, since all the papers in this group want to ensure the applicability of the model in different contexts, and the use of pre-defined rules for the definition of the graph limits the model's universal applicability and negatively impacts knowledge transfer. More than half of the papers employ a dynamic graph structure, which adapts to the varying temporal patterns observed in the data.

In Ref.~\cite{art_688} the authors introduce a novel data structure called hypervariate graph, where each time serie values is represented as a graph node, and slicing windows are modeled as fully connected space-time graphs. Unlike traditional approaches that treat spatial and temporal aspects separately, according to the authors, this approach captures spatiotemporal correlations in an integrated manner.

The two most commonly utilized loss functions are the MSE and the MAE. Since the graph structure is typically learned by the models themselves, many papers include a regularization term in the loss function for the optimization of the graph structure (e.g., Refs.~\cite{art_73, art_78, art_79, art_86, art_641}).

The majority of the papers specify the language and libraries used for the code, namely Python with PyTorch, PyTorch Geometric \cite{PyTorch_Geometric}, and Torch Spatiotemporal \cite{Cini_Torch_Spatiotemporal_2022}. However, only slightly more than half of the papers provide a link to the source code. The links to the code repositories are listed in Tab.~\ref{t:link codici generic}. Ref.~\cite{art_689} is a study that does not introduce a new model but rather investigates the relationship between global and local patterns in graph-based spatiotemporal forecasting. The authors provide access to their source code at \url{https://github.com/Graph-Machine-Learning-Group/taming-local-effects-stgnns}. Additionally, this group includes the work that introduced the PyTorch Geometric Temporal library \cite{art_705}, a framework for dynamic graph learning. The corresponding source code is publicly available at \url{https://github.com/benedekrozemberczki/pytorch_geometric_temporal}. In addition, the authors compare various open-source libraries for geometric deep learning.

\begin{longtable}{p{7cm}p{7.5cm}}
    \caption{List of source codes of the \enquote{Generic} models in the review.}
    \label{t:link codici generic}
    \endfirsthead
    \endhead
    \toprule
    Model & Link \\
    \midrule
    Adaptive dependency learning neural network (ADLNN) \cite{art_83} & \url{https://github.com/AbishekSriramulu/ADLGNN.git} $^\dagger$ \\
    Attention-adjusted graph spatio-temporal network (AGSTN) \cite{art_654} & \url{https://github.com/l852888/AGSTN} \\
    Cross-attention temporal dynamic graph neural network (CATodyNet) \cite{art_386} & \url{https://github.com/HQYWY/CATodyNet} \\
    Graph neural network with neural Granger causality (CauGNN) \cite{art_77} & \url{https://github.com/RRRussell/CauGNN} \\
    Spatio-temporal contrastive learning (CL4ST) \cite{art_709} & \url{https://github.com/HKUDS/CL4ST} \\
    CrossGNN \cite{art_687} & \url{https://github.com/hqh0728/CrossGNN} \\
    Dynamic diffusion-variational GNN (DVGNN) \cite{art_383} & \url{https://github.com/gorgen2020/DVGNN} \\
    Evolving multi-scale graph neural network (ESG) \cite{art_649} & \url{https://github.com/LiuZH-19/ESG} \\
    Fourier graph neural network (FourierGNN) \cite{art_688} & \url{https://github.com/aikunyi/FourierGNN} \\
    Graph interpolation attention recursive network (GinAR) \cite{art_642} & \url{https://github.com/ChengqingYu/GinAR} \\
    Gformer \cite{art_82} & \url{https://github.com/wxh453751461/Gformer} \\
    Graphs for forecasting irregularly sampled time series with missing values (GraFITi) \cite{art_643} & \url{https://github.com/yalavarthivk/GraFITi} \\
    GraphSensor \cite{art_395} & \url{https://github.com/jldx-gjq/Graphnet/tree/main/graphsensor-code-main} \\
    Graph for time series (GTS) \cite{art_641} & \url{https://github.com/chaoshangcs/GTS} \\
    HAR-Sensor \cite{art_159} & \url{https://github.com/riktimmondal/HAR-Sensor} $^\dagger$ \\
    Hierarchical downsampling time-then-space (HD-TTS) \cite{art_644} & \url{https://github.com/marshka/hdtts} \\
    HyperTime \cite{art_680} & \url{https://gitfront.io/r/user-9180183/WXPgc7aQjSKE/HyperTime/} $^\dagger$ \\
    Multi-scale adaptive graph neural network (MAGNN) \cite{art_80} & \url{https://github.com/shangzongjiang/MAGNN} \\
    Multi-task dynamic graph generation framework for multivariate time series forecasting (MDG) \cite{art_397} & \url{https://github.com/gortwwh/MDG} \\
    Multivariate time series deep spatiotemporal forecasting model with a graph neural network (MDST-GNN) \cite{art_90} & \url{https://github.com/yiminghzc/MDST-GNN} \\
    Multivariate time series classification based on fusion features (MSTC\_FF) \cite{art_385} & \url{https://github.com/dumingsen/MTSC_FF} \\
    Multivariate time series forecasting with graph neural networks (MTGNN) \cite{art_655} & \url{https://github.com/nnzhan/MTGNN} \\
    Multivariate time series with dynamic graph neural ordinary differential equations (MTGODE) \cite{art_75} & \url{https://github.com/GRAND-Lab/MTGODE} \\
    RAINDROP \cite{art_645} & \url{https://github.com/mims-harvard/Raindrop} \\
    Static and dynamic graph learning network (SDGL) \cite{art_78} & \url{https://github.com/ZhuoLinLi-shu/SDGL} \\
    Similarity-aware time-series classification (SimTSC) \cite{art_711} & \url{https://github.com/daochenzha/SimTSC} \\
    Sparse graph learning from spatiotemporal time series \cite{art_72} & \url{https://github.com/andreacini/sparse-graph-learning} \\
    Spatial relation decomposition (SRD) \cite{art_647} & \url{https://github.com/Arthur-Null/SRD/tree/main} \\
    Spatio-temporal graph neural networks (STExplainer) \cite{art_700} & \url{https://github.com/HKUDS/STExplainer} \\
    Time-aware multipersistence spatio-supra graph convolutional network (TAMP-S2GCNet) \cite{art_674} & \url{https://www.dropbox.com/sh/n0ajd5l0tdeyb80/AABGn-ejfV1YtRwjf_L0AOsNa?dl=0} \\
    Temporal decomposition enhanced graph neural network for multivariate time series forecasting (TDG4MSF) \cite{art_81} & \url{https://github.com/TYUT-Theta/MHZN.git} \\
    Temporal graph convolution and attention (T-GAN) \cite{art_202} & \url{https://github.com/malei666666/T_GAN} \\
    Temporal graph neural networks for irregular data (TGNN4I) \cite{art_710} & \url{https://github.com/joeloskarsson/tgnn4i} \\
    Temporal dynamic graph neural network (TodyNet) \cite{art_389} & \url{https://github.com/liuxz1011/TodyNet} \\
    Transformable patching graph neural network (T-PATCHGNN) \cite{art_646} & \url{https://github.com/usail-hkust/t-PatchGNN} \\
    Temporal polynomial graph neural network (TPGNN) \cite{art_685} & \url{https://github.com/zyplanet/TPGNN} \\
    Two GNN models with different graph structures \cite{art_160} & 
    \url{https://github.com/StefanBloemheuvel/graph_comparison} \\
    Unified Spatio-Temporal Diffusion Models (USTD) \cite{art_695} & \url{https://github.com/hjf1997/USTD} \\
    Weight-bounded-GAT \cite{art_703} & \url{https://github.com/zzh237/Weight-bounded-GAT} \\
    Zigzag filtration curve based supra-hodge convolution networks (ZFC-SHCN) \cite{art_686} & \url{https://github.com/zfcshcn/ZFC-SHCN} \\
    Time zigzags at graph convolutional network (Z-GCNET) \cite{art_657} & \url{https://github.com/Z-GCNETs/Z-GCNETs} \\
    \bottomrule
\end{longtable}

\newpage
\subsubsection{Benchmark models}
Tab.~\ref{t:benchmark models generic} displays the benchmark models utilized by at least two of the selected \enquote{Generic} papers.

\begin{longtable}{p{2.5cm}p{6.5cm}p{5.2cm}}
    \caption{List of benchmark models in the \enquote{Generic} group divided per category.}
    \label{t:benchmark models generic}
    \endfirsthead
    \endhead
    \toprule
    Category & Model & Used by \\
    \midrule
    \multirow[t]{3}{=}{Mathematical and statistical methods} & Autoregressive integrated moving average (ARIMA) & \cite{art_73}, \cite{art_74},  \cite{art_78}, \cite{art_79}, \cite{art_80}, \cite{art_81}, \cite{art_83}, \cite{art_84}, \cite{art_88}, \cite{art_89}, \cite{art_393}, \cite{art_397}, \cite{art_400}, \cite{art_405}, \cite{art_639}, \cite{art_641}, \cite{art_647}, \cite{art_649}, \cite{art_654}, \cite{art_655}, \cite{art_656}, \cite{art_685}, \cite{art_701} \\
    & Gaussian process (GP) & \cite{art_73}, \cite{art_74}, \cite{art_78}, \cite{art_80}, \cite{art_81}, \cite{art_83}, \cite{art_647}, \cite{art_649}, \cite{art_655}, \cite{art_656}, \cite{art_685} \\
    & Historical average (HA) & \cite{art_78}, \cite{art_88}, \cite{art_400}, \cite{art_641}, \cite{art_650}, \cite{art_656}, \cite{art_657}, \cite{art_700}, \cite{art_709} \\
    & Vector autoregression (VAR) & \cite{art_73}, \cite{art_74}, \cite{art_77}, \cite{art_78}, \cite{art_79}, \cite{art_81}, \cite{art_84}, \cite{art_88}, \cite{art_400}, \cite{art_405}, \cite{art_640}, \cite{art_641}, \cite{art_657}, \cite{art_688}, \cite{art_696}, \cite{art_700}, \cite{art_709} \\
    \midrule 
    \multirow[t]{4}{=}{Non-GNN machine learning methods} & Autoformer \cite{Autoformer} & \cite{art_74}, \cite{art_76}, \cite{art_85}, \cite{art_640}, \cite{art_647}, \cite{art_687}, \cite{art_688} \\
    & Crossformer \cite{Crossformer} & \cite{art_74}, \cite{art_646}, \cite{art_687} \\
    & DeepAR \cite{DeepAR} & \cite{art_69}, \cite{art_393}, \cite{art_698} \\
    & Attentive hierarchical recurrent networks for crime prediction (DeepCrime) \cite{DeepCrime} & \cite{art_700}, \cite{art_709} \\
    & DeepGLO \cite{DeepGLO} & \cite{art_640}, \cite{art_674}, \cite{art_684}, \cite{art_686}, \cite{art_688}, \cite{art_701} \\
    & DeepSleepNet \cite{DeepSleepNet} & \cite{art_395}, \cite{art_403} \\
    & DeepState \cite{DeepState} & \cite{art_674}, \cite{art_684}, \cite{art_686}, \cite{art_701} \\
    & DLinear \cite{DLinear} & \cite{art_76}, \cite{art_643}, \cite{art_646}, \cite{art_687} \\
    & Deep multi-view spatial-temporal network (DMVST-Net) \cite{DMVST-Net} & \cite{art_639}, \cite{art_701} \\
    & Dual self-attention network for multivariate time series forecasting (DSANet) \cite{DSANet} & \cite{art_73}, \cite{art_78}, \cite{art_405}, \cite{art_657}, \cite{art_709} \\
    & Exponential smoothing transformers for time-series forecasting (ETSformer) \cite{ETSformer} & \cite{art_76}, \cite{art_687} \\
    & Frequency enhanced decomposed transformer (FEDformer) \cite{FEDformer} & \cite{art_74}, \cite{art_640}, \cite{art_643}, \cite{art_687} \\
    & Feed-forward neural network (FNN) & \cite{art_84}, \cite{art_89}, \cite{art_639}, \cite{art_641}, \cite{art_711} \\
    & Gated recurrent unit (GRU) & \cite{art_73}, \cite{art_78}, \cite{art_82}, \cite{art_88}, \cite{art_90}, \cite{art_398}, \cite{art_404}, \cite{art_405}, \cite{art_644}, \cite{art_645}, \cite{art_646}, \cite{art_657}, \cite{art_643}, \cite{art_696}, \cite{art_710} \\
    & Hybrid framework based on fully Dilated CNN (HyDCNN) \cite{HyDCNN} & \cite{art_75}, \cite{art_83} \\
    & InceptionTime \cite{InceptionTime} & \cite{art_647}, \cite{art_711} \\
    & Informer \cite{Informer} & \cite{art_74}, \cite{art_76}, \cite{art_82}, \cite{art_85}, \cite{art_397}, \cite{art_640}, \cite{art_643}, \cite{art_650}, \cite{art_656}, \cite{art_685}, \cite{art_688}, \cite{art_696} \\
    & $k$-nearest neighbors (KNN) & \cite{art_84}, \cite{art_387}, \cite{art_680}, \cite{art_711} \\
    & LogTrans \cite{LogTrans} & \cite{art_82}, \cite{art_85} \\
    & Long-short term memory network (LSTM) & \cite{art_71}, \cite{art_74}, \cite{art_79}, \cite{art_82}, \cite{art_84}, \cite{art_85}, \cite{art_86}, \cite{art_88}, \cite{art_89}, \cite{art_396}, \cite{art_397}, \cite{art_398}, \cite{art_400}, \cite{art_641}, \cite{art_643}, \cite{art_650}, \cite{art_654}, \cite{art_657}, \cite{art_674}, \cite{art_684}, \cite{art_685}, \cite{art_686}, \cite{art_696}, \cite{art_701} \\
    & Long-short term time series network (LSTNet) \cite{LSTNet} & \cite{art_73}, \cite{art_74}, \cite{art_75}, \cite{art_77}, \cite{art_78}, \cite{art_80}, \cite{art_81}, \cite{art_82}, \cite{art_83}, \cite{art_85}, \cite{art_87}, \cite{art_89}, \cite{art_90}, \cite{art_387}, \cite{art_393}, \cite{art_640}, \cite{art_647}, \cite{art_649}, \cite{art_655}, \cite{art_656}, \cite{art_674}, \cite{art_684}, \cite{art_685}, \cite{art_686}, \cite{art_688}, \cite{art_696}, \cite{art_701}, \cite{art_702} \\
    & Multi-level construal neural network (MLCNN) \cite{MLCNN} & \cite{art_77}, \cite{art_87} \\
    & Multi-time attention networks (mTANs) \cite{mTANs} & \cite{art_643}, \cite{art_645}, \cite{art_646} \\
    & Neural basis expansion analysis for interpretable time series forecasting (N-BEATS) \cite{N-BEATS} & \cite{art_69}, \cite{art_85}, \cite{art_674}, \cite{art_684}, \cite{art_686}, \cite{art_701} \\
    & Channel-independent patch time series Transformer (PatchTST) \cite{PatchTST} & \cite{art_646}, \cite{art_687} \\
    & Reformer \cite{Reformer} & \cite{art_76}, \cite{art_82}, \cite{art_85}, \cite{art_640}, \cite{art_688} \\
    & Recurrent neural network (RNN) & \cite{art_89}, \cite{art_689} \\
    & Recurrent neural network with fully connected gated recurrent units (RNN-GRU) & \cite{art_73}, \cite{art_77}, \cite{art_78}, \cite{art_79}, \cite{art_80}, \cite{art_81}, \cite{art_83}, \cite{art_87}, \cite{art_405}, \cite{art_649}, \cite{art_655}, \cite{art_685} \\
    & Rocket \cite{ROCKET} & \cite{art_407}, \cite{art_647} \\
    & SCINet \cite{SCINet} & \cite{art_74}, \cite{art_90} \\
    & Set functions for time series (SeFT) \cite{SeFT} & \cite{art_645}, \cite{art_646} \\
    & State frequency memory (SFM) \cite{SFM} & \cite{art_640}, \cite{art_674}, \cite{art_684}, \cite{art_686}, \cite{art_688} \\
    & Shapelet-neural network (ShapeNet) \cite{ShapeNet} & \cite{art_386}, \cite{art_389}, \cite{art_680} \\
    & Spatial-temporal dynamic network (STDN) \cite{STDN} & \cite{art_639}, \cite{art_700}, \cite{art_701}, \cite{art_709} \\
    & Spatial-temporal identity (STID) \cite{STID} & \cite{art_74}, \cite{art_642} \\
    & Spatial-temporal transformer network (STtrans) \cite{STtrans} & \cite{art_700}, \cite{art_709} \\
    & Spatio-temporal residual networks (ST-ResNet) \cite{ST-ResNet} & \cite{art_700}, \cite{art_701}, \cite{art_709} \\
    & Support vector machine (SVM) & \cite{art_84}, \cite{art_396}, \cite{art_641}, \cite{art_654}, \cite{art_701}, \cite{art_709} \\
    & TapNet \cite{TapNet} & \cite{art_385}, \cite{art_389}, \cite{art_680}, \cite{art_711} \\
    & Temporal convolutional network (TCN) \cite{TCN} & \cite{art_88}, \cite{art_387}, \cite{art_396}, \cite{art_397}, \cite{art_640}, \cite{art_674}, \cite{art_684}, \cite{art_686}, \cite{art_688}, \cite{art_701} \\
    & Temporal 2D-variation modeling for general time series analysis (TimesNet) \cite{TimesNet} & \cite{art_76}, \cite{art_646}, \cite{art_680}, \cite{art_687} \\
    & Temporal pattern attention long-short term memory network (TPA-LSTM) \cite{TPA-LSTM} & \cite{art_73}, \cite{art_74}, \cite{art_75}, \cite{art_78}, \cite{art_79}, \cite{art_80}, \cite{art_81}, \cite{art_83}, \cite{art_90}, \cite{art_405}, \cite{art_649}, \cite{art_655}, \cite{art_656}, \cite{art_685}, \cite{art_702} \\
    & TimeGrad \cite{TimeGrad} & \cite{art_695}, \cite{art_698} \\
    & Transformer \cite{attention} & \cite{art_71}, \cite{art_77}, \cite{art_393}, \cite{art_645}, \cite{art_710} \\
    & Hybrid of the multilayer perception and autoregressive model (VAR-MLP) \cite{VAR-MLP} & \cite{art_73}, \cite{art_74}, \cite{art_78}, \cite{art_79}, \cite{art_80}, \cite{art_81}, \cite{art_83}, \cite{art_87}, \cite{art_90}, \cite{art_405}, \cite{art_649}, \cite{art_655}, \cite{art_656}, \cite{art_685}, \cite{art_702} \\
    & WEASEL + multivariate unsupervised symbols and derivatives (WEASEL+MUSE) \cite{WEASEL+MUSE} & \cite{art_385}, \cite{art_386}, \cite{art_389}, \cite{art_680} \\
    & Extreme gradient boosting (XGBoost) & \cite{art_639}, \cite{art_649}, \cite{art_650} \\
    \midrule 
    \multirow[t]{1}{=}{GNN methods} & Adaptive graph convolutional recurrent network (AGCRN) \cite{art_683} & \cite{art_73}, \cite{art_74}, \cite{art_78}, \cite{art_79}, \cite{art_80}, \cite{art_88}, \cite{art_396}, \cite{art_400}, \cite{art_640}, \cite{art_644}, \cite{art_650}, \cite{art_657}, \cite{art_674}, \cite{art_686}, \cite{art_688}, \cite{art_689}, \cite{art_700}, \cite{art_701}, \cite{art_705}, \cite{art_709} \\
    & Attention-based spatial-temporal graph convolutional network (ASTGCN) \cite{ASTGCN} & \cite{art_78}, \cite{art_79}, \cite{art_383}, \cite{art_396}, \cite{art_397}, \cite{art_400}, \cite{art_654}, \cite{art_657}, \cite{art_700}, \cite{art_703}, \cite{art_709} \\
    & Attention based spatial-temporal graph neural network (ASTGNN) \cite{ASTGNN} & \cite{art_74}, \cite{art_396} \\
    & Coupled layer-wise convolutional recurrent neural network (CCRNN) \cite{CCRNN} & \cite{art_649}, \cite{art_650} \\ 
    & CrossGNN \cite{art_687} & \cite{art_646}, \cite{art_687} \\
    & Diffusion convolutional recurrent neural network (DCRNN) \cite{DCRNN} & \cite{art_73}, \cite{art_74}, \cite{art_75}, \cite{art_78}, \cite{art_79}, \cite{art_84}, \cite{art_86}, \cite{art_87}, \cite{art_88}, \cite{art_383}, \cite{art_396}, \cite{art_397}, \cite{art_400}, \cite{art_405}, \cite{art_641}, \cite{art_644}, \cite{art_647}, \cite{art_649}, \cite{art_650}, \cite{art_654}, \cite{art_655}, \cite{art_657}, \cite{art_674}, \cite{art_685}, \cite{art_686}, \cite{art_689}, \cite{art_695}, \cite{art_698}, \cite{art_700}, \cite{art_701}, \cite{art_705}, \cite{art_709} \\
    & Dynamic graph convolutional recurrent network (DGCRN) \cite{art_100} & \cite{art_397}, \cite{art_398} \\
    & Dynamic and multi-faceted spatiotemporal graph convolution network (DMSTGCN) \cite{art_663} & \cite{art_396}, \cite{art_417}, \cite{art_647}, \cite{art_700}, \cite{art_709} \\
    & Dynamic spatial-temporal aware graph neural network (DSTAGNN) \cite{art_675} & \cite{art_383}, \cite{art_700} \\
    & Evolving multi-scale graph neural network (ESG) \cite{art_649} & \cite{art_74}, \cite{art_650} \\
    & Graph auto encoder (GAE) \cite{GAE} & \cite{art_70}, \cite{art_202} \\
    & Graph deviation network (GDN) \cite{GDN} & \cite{art_387}, \cite{art_703} \\
    & Graph multi-attention network (GMAN) \cite{GMAN} & \cite{art_73}, \cite{art_75}, \cite{art_86}, \cite{art_88}, \cite{art_655}, \cite{art_700}, \cite{art_703}, \cite{art_709} \\
    & Graph-based multi-step dependency relation (GMSDR) \cite{art_682} & \cite{art_695}, \cite{art_698}, \cite{art_700}, \cite{art_709} \\
    & Discrete graph structure learning for time series (GTS) \cite{art_641} & \cite{art_72}, \cite{art_78}, \cite{art_397}, \cite{art_401}, \cite{art_649}, \cite{art_650} \\
    & Graph WaveNet (GWN) \cite{GWN} & \cite{art_74}, \cite{art_75}, \cite{art_79}, \cite{art_80}, \cite{art_86}, \cite{art_87}, \cite{art_396}, \cite{art_397}, \cite{art_417}, \cite{art_640}, \cite{art_644}, \cite{art_646}, \cite{art_650}, \cite{art_655}, \cite{art_657}, \cite{art_674}, \cite{art_684}, \cite{art_685}, \cite{art_686}, \cite{art_688}, \cite{art_689}, \cite{art_695}, \cite{art_700}, \cite{art_701}, \cite{art_709} \\
    & Hypergraph convolutional recurrent neural network (HGC-RNN) \cite{art_639} & \cite{art_400}, \cite{art_701} \\
    & Long short-term graph convolutional networks (LSGCN) \cite{LSGCN} & \cite{art_657}, \cite{art_709} \\
    & Multi-range attentive bicomponent graph convolutional network (MRA-BGCN) \cite{MRA-BGCN} & \cite{art_86}, \cite{art_655} \\
    & Multivariate time series forecasting with graph neural network (MTGNN) \cite{art_655} & \cite{art_72}, \cite{art_73}, \cite{art_74}, \cite{art_75}, \cite{art_78}, \cite{art_79}, \cite{art_80}, \cite{art_81}, \cite{art_83}, \cite{art_86}, \cite{art_87}, \cite{art_88}, \cite{art_90}, \cite{art_393}, \cite{art_396}, \cite{art_397}, \cite{art_401}, \cite{art_405}, \cite{art_417}, \cite{art_640}, \cite{art_646}, \cite{art_647}, \cite{art_649}, \cite{art_656}, \cite{art_685}, \cite{art_687}, \cite{art_688}, \cite{art_702}, \cite{art_703} \\
    & Multivariate time series with dynamic graph neural ordinary differential equations (MTGODE) \cite{art_75} & \cite{art_88}, \cite{art_401}, \cite{art_417} \\
    & Node2vec \cite{node2vec} & \cite{art_70}, \cite{art_202} \\
    & Spectral temporal graph neural network for multivariate time-series forecasting (StemGNN) \cite{art_684} & \cite{art_397}, \cite{art_405}, \cite{art_640}, \cite{art_646}, \cite{art_649}, \cite{art_674}, \cite{art_684}, \cite{art_685}, \cite{art_686}, \cite{art_688}, \cite{art_700}, \cite{art_701}, \cite{art_709} \\
    & Spatial-temporal fusion graph neural network (STFGNN) \cite{STFGNN} & \cite{art_73}, \cite{art_86}, \cite{art_405}, \cite{art_700}, \cite{art_709} \\
    & Spatio-temporal graph convolutional network (STGCN) \cite{STGCN} & \cite{art_73}, \cite{art_75}, \cite{art_78}, \cite{art_79}, \cite{art_84}, \cite{art_86}, \cite{art_88}, \cite{art_397}, \cite{art_400}, \cite{art_401}, \cite{art_649}, \cite{art_655}, \cite{art_657}, \cite{art_674}, \cite{art_684}, \cite{art_685}, \cite{art_686}, \cite{art_695}, \cite{art_698}, \cite{art_700}, \cite{art_701}, \cite{art_703}, \cite{art_709} \\
    & Spatio-temporal graph neural controlled differential equation (STG-NCDE) \cite{STG-NCDE} & \cite{art_75}, \cite{art_695}, \cite{art_698}, \cite{art_700}, \cite{art_709} \\
    & Spatial-temporal graph ODE networks (STGODE) \cite{art_676} & \cite{art_75}, \cite{art_88}, \cite{art_383}, \cite{art_695}, \cite{art_700}, \cite{art_709} \\
    & Deep-meta-learning based model (ST-MetaNet) \cite{ST-MetaNet} & \cite{art_86}, \cite{art_655}, \cite{art_700}, \cite{art_701}, \cite{art_709} \\
    & Spatiotemporal multi-graph convolution network (ST-MGCN) \cite{ST-MGCN} & \cite{art_654}, \cite{art_701} \\
    & Spatio-temporal synchronous graph convolutional network (STSGCN) \cite{STSGCN} & \cite{art_73}, \cite{art_74}, \cite{art_78}, \cite{art_387}, \cite{art_397}, \cite{art_400}, \cite{art_405}, \cite{art_649}, \cite{art_657}, \cite{art_695}, \cite{art_700}, \cite{art_709} \\
    & Spatial-temporal sequential hypergraph network (ST-SHN) \cite{ST-SHN} & \cite{art_700}, \cite{art_709} \\
    & Time-aware multipersistence spatio-supra graph convolutional network (TAMP-S2GCNet) \cite{art_674} & \cite{art_700}, \cite{art_709} \\
    & Temporal graph convolutional network (T-GCN) \cite{T-GCN} & \cite{art_383}, \cite{art_393}, \cite{art_705} \\
    & Temporal dynamic graph neural network (TodyNet) \cite{art_389} & \cite{art_386}, \cite{art_407}, \cite{art_680} \\
    & Time zigzags at graph convolutional networks for time series forecasting (Z-GCNETs) \cite{art_657} & \cite{art_400}, \cite{art_405}, \cite{art_674}, \cite{art_684}, \cite{art_686}, \cite{art_700}, \cite{art_701}, \cite{art_709} \\
    \bottomrule
\end{longtable}

This \enquote{Generic} thematic group includes a wide variety of benchmark models, also due to the large number of papers within this group. As for mathematical and statistical models, autoregressions and ARIMA models are very popular for their simplicity and interpretability. Among non-GNN machine learning models, the most widely used models are LSTM networks, the so-called temporal pattern attention long-short term memory network (TPA-LSTM) \cite{TPA-LSTM}, and the long-short term time series network (LSTNet) \cite{LSTNet}. The TPA-LSTM model, introduced by Shih et al.~in 2019, employs a set of filters to capture time-invariant temporal patterns, and an attention mechanism to identify relevant time series for multivariate forecasting. The source code for this benchmark model is available at \url{https://github.com/gantheory/TPA-LSTM}. The LSTNet model, proposed by Lai et al.~in 2018, combines a convolution neural network and a recurrent neural network to capture short-term local dependencies among variables and identify long-term patterns. Additionally, it incorporates an autoregressive model that enhances the robustness of the deep learning approach to time series with significant scale fluctuations. The source code for the model is accessible at \url{https://github.com/laiguokun/LSTNet}. Among GNN models, the most common ones are DCRNN \cite{DCRNN}, MTGNN \cite{art_655}, and GWN \cite{GWN} models, already discussed in the previous subsections.

\subsubsection{Results}
The most common error metrics used in the evaluation of the models in the \enquote{Generic} group are, in order, the mean absolute error (MAE), correlation coefficient (CORR), root mean squared error (RMSE), root relative squared error (RRSE), mean absolute percentage error (MAPE), and mean squared error (MSE). Differently from other metrics, where lower values indicate higher accuracy, the correlation coefficient (CORR) quantifies the strength of the the linear relationship between predicted and actual values, with higher values indicating a better performance. This subsection about results compares the results obtained for the Electricity consumption, Exchange rate, Solar energy and Traffic datasets. The results of the proposed models and benchmarks on the \enquote{Mobility} datasets were already reported with a $^\star$ in Subsec.~\ref{s:Mobility_results}.

Tab.~\ref{t:risultati electricity} shows the accuracy of the various models applied to the Electricity consumption dataset, which includes data from 1$^{\text{st}}$ January 2012 to 31$^{\text{st}}$ December 2014 at hourly granularity, for 3, 6, 12 and 24 steps ahead horizons. The accuracy is expressed in terms of RRSE and CORR.

{\tabcolsep=4.5pt
\begin{longtable}{lcccccccc}
    \caption{Comparison of the average accuracy of the different models on the Electricity consumption dataset for 3, 6, 12, and 24 steps time horizons, expressed in terms of RRSE and CORR. The numbers corresponding to the highest accuracy are underlined.}
    \label{t:risultati electricity}
    \endfirsthead
    \endhead
    \toprule
    Time horizon & \multicolumn{2}{c}{3 steps} & \multicolumn{2}{c}{6 steps} & \multicolumn{2}{c}{12 steps} & \multicolumn{2}{c}{24 steps} \\
    Metrics & RRSE & CORR & RRSE & CORR & RRSE & CORR & RRSE & CORR \\
    \midrule 
    ADLGNN \cite{art_83} & 0.0719 & 0.9506 & 0.0809 & 0.9386 & 0.0887 & 0.9312 & 0.0930 & 0.9294 \\
    AGCRN & 0.0766 & 0.9408 & 0.0894 & 0.9309 & 0.0921 & 0.9222 & 0.0967 & 0.9183 \\
    AGLG-GRU \cite{art_79} & 0.0738 & 0.9434 & 0.0864 & 0.9302 & 0.0912 & 0.9283 & 0.0947 & 0.9274 \\
    AR & 0.0995 & 0.8845 & 0.1035 & 0.8632 & 0.1050 & 0.8591 & 0.1054 & 0.8595 \\
    ARIMA & 0.0917 & 0.8902 & 0.1002 & 0.8834 & 0.1028 & 0.8604 & 0.1042 & 0.8487 \\
    Autoformer & 0.1258 & 0.9147 & 0.1344 & 0.9001 & 0.1357 & 0.8921 & 0.1554 & 0.8704 \\
    Crossformer & 0.0742 & 0.9452 & 0.0855 & 0.9351 & 0.0901 & 0.9280 & 0.0967 & 0.9205 \\
    DPGNN \cite{art_405} & 0.0720 & 0.9527 & 0.0810 & 0.9422 & 0.0893 & \underline{0.9357} & 0.0935 & 0.9291 \\
    DSTGN \cite{art_73} & 0.0713 & 0.9518 & 0.0821 & 0.9424 & 0.0887 & \underline{0.9357} & - & - \\
    DSTIGNN \cite{art_74} & 0.0733 & 0.9515 & 0.0820 & 0.9412 & 0.0899 & 0.9338 & 0.0954 & 0.9289 \\
    ESG \cite{art_649} & 0.0718 & 0.9494 & 0.0844 & 0.9372 & 0.0898 & 0.9321 & 0.0962 & 0.9279 \\
    FEDformer & 0.0889 & 0.9321 & 0.1006 & 0.9191 & 0.1154 & 0.9080 & 0.1202 & 0.9012 \\
    GP & 0.1500 & 0.8670 & 0.1907 & 0.8334 & 0.1621 & 0.8394 & 0.1273 & 0.8818 \\
    GRU & 0.1102 & 0.8597 & 0.1144 & 0.8623 & 0.1183 & 0.8472 & - & - \\
    GTS & 0.0790 & 0.9291 & 0.0884 & 0.9187 & 0.0957 & 0.9135 & 0.0951 & 0.9098 \\
    GWN & 0.0746 & 0.9459 & 0.0922 & 0.9310 & 0.0909 & 0.9267 & 0.0962 & 0.9226 \\
    HyDCNN & 0.0832 & 0.9354 & 0.0898 & 0.9329 & 0.0921 & 0.9285 & - & - \\
    Informer & 0.1337 & 0.8903 & 0.1532 & 0.8705 & 0.1635 & 0.8527 & 0.1834 & 0.8399 \\
    LSTNet & 0.0864 & 0.9283 & 0.0931 & 0.9135 & 0.1007 & 0.9077 & 0.1007 & 0.9119 \\
    MAGL \cite{art_702} & 0.0728 & 0.9476 & 0.0824 & 0.9377 & 0.0893 & 0.9299 & 0.0951 & 0.9240 \\
    MAGNN \cite{art_80} & 0.0745 & 0.9476 & 0.0876 & 0.9323 & 0.0908 & 0.9282 & 0.0963 & 0.9217 \\
    MDST-GNN \cite{art_90} & 0.0738 & 0.9454 & 0.0833 & 0.9346 & 0.0884 & 0.9264 & \underline{0.0922} & 0.9222 \\
    MTGNN \cite{art_655} & 0.0745 & 0.9474 & 0.0878 & 0.9316 & 0.0916 & 0.9278 & 0.0953 & 0.9234 \\
    MTGODE \cite{art_75} & 0.0736 & 0.9430 & 0.0809 & 0.9340 & 0.0891 & 0.9279 & - & - \\
    MTHetGNN & 0.0749 & 0.9456 & 0.0892 & 0.9307 & 0.0959 & 0.8783 & 0.0969 & 0.8782 \\
    MTNet & 0.0840 & 0.9319 & 0.0901 & 0.9226 & 0.0934 & 0.9165 & 0.0969 & 0.9147 \\
    RNN-GRU & 0.1102 & 0.8597 & 0.1144 & 0.8623 & 0.1183 & 0.8472 & 0.1295 & 0.8651 \\
    SARIMA & 0.0906 & 0.9055 & 0.0999 & 0.8829 & 0.1026 & 0.8674 & 0.1036 & 0.8621 \\
    SCINet & 0.0740 & 0.9494 & 0.0845 & 0.9387 & 0.0929 & 0.9305 & 0.0967 & 0.9270 \\
    SDGL \cite{art_78} & 0.0698 & \underline{0.9534} & 0.0805 & \underline{0.9445} & 0.0889 & 0.9351 & 0.0935 & \underline{0.9301} \\
    SDLGNN & 0.0726 & 0.9502 & 0.0820 & 0.9384 & 0.0896 & 0.9304 & 0.0947 & 0.9257 \\
    SDLGNN-Corr & 0.0737 & 0.9475 & 0.0841 & 0.9346 & 0.0923 & 0.9263 & 0.0971 & 0.9227 \\
    StemGNN & 0.0799 & 0.9490 & 0.0909 & 0.9397 & 0.0989 & 0.9342 & 0.1019 & 0.9209 \\
    TDG4-MSF \cite{art_81} & 0.0731 & 0.9499 & 0.0828 & 0.9371 & 0.0894 & 0.9306 & 0.0969 & 0.9246 \\
    Theta & 0.0975 & 0.8906 & 0.1029 & 0.8723 & 0.1040 & 0.8576 & 0.1041 & 0.8595 \\
    TPGNN \cite{art_685} & \underline{0.0627} & 0.9417 & \underline{0.0685} & 0.9362 & \underline{0.0699} & 0.9285 & 0.0936 & 0.9293 \\
    TPA-LSTM & 0.0823 & 0.9439 & 0.0916 & 0.9337 & 0.0964 & 0.9250 & 0.1006 & 0.9133 \\
    TRMF & 0.1802 & 0.8538 & 0.2039 & 0.8424 & 0.2186 & 0.8304 & 0.3656 & 0.7471 \\
    VAR-MLP & 0.1393 & 0.8708 & 0.1620 & 0.8389 & 0.1557 & 0.8192 & 0.1274 & 0.8679 \\
    \bottomrule
\end{longtable}
}

The accuracy of the various models varies depending on the studied forecasting horizon but, overall, one can notice that the GNN models perform better than the others, with the static and dynamic graph learning network (SDGL) \cite{art_78} and the temporal polynomial graph neural network (TPGNN) \cite{art_685} being the most accurate. The SDGL model, proposed by Li et al.~in 2023, first employs a static and dynamic graph learning module that incorporates graph regularization to control the smoothness, connectivity, and sparsity of the learned graph. The graph input then passes through gated temporal convolutional networks and graph convolution modules, which are equipped with residual and skip connections that deliver the information to the output layer. The TPGNN, proposed by Liu et al.~in 2022, builds a series of matrix polynomials that share a common matrix basis. The graph learning module first captures the overall correlations using a static matrix basis. Then, a set of time-varying coefficients and the matrix basis are used to create a dynamic graph. Overall, the TPGNN model uses an encoder-decoder architecture and makes predictions in an auto-regressive manner.

Tab.~\ref{t:risultati exchange rate} presents the accuracy of different models applied to the Exchange rate dataset, with daily data from 1990 to 2016. The evaluation covers horizons of 3, 6, 12, and 24 steps ahead, with accuracy measured by RRSE and CORR.

{\tabcolsep=4.5pt
\begin{longtable}{lcccccccc}
    \caption{Comparison of the average accuracy of the different models on the Exchange rate dataset for 3, 6, 12, and 24 steps time horizons, expressed in terms of RRSE and CORR. The numbers corresponding to the highest accuracy are underlined.}
    \label{t:risultati exchange rate}
    \endfirsthead
    \endhead
    \toprule
    Time horizon & \multicolumn{2}{c}{3 steps} & \multicolumn{2}{c}{6 steps} & \multicolumn{2}{c}{12 steps} & \multicolumn{2}{c}{24 steps} \\
    Metrics & RRSE & CORR & RRSE & CORR & RRSE & CORR & RRSE & CORR \\
    \midrule 
    AGCRN & 0.0269 & 0.9717 & 0.0331 & 0.9615 & 0.0374 & 0.9531 & 0.0476 & 0.9334 \\
    AGLG-GRU \cite{art_79} & 0.0191 & 0.9792 & \underline{0.0233} & 0.9701 & \underline{0.0328} & 0.9548 & 0.0449 & 0.9372 \\
    AR & 0.0228 & 0.9734 & 0.0279 & 0.9656 & 0.0353 & 0.9526 & 0.0445 & 0.9357 \\
    ARIMA & 0.0198 & 0.9754 & 0.0261 & 0.9721 & 0.0344 & 0.9548 & 0.0445 & 0.9301 \\
    Autoformer & 0.0291 & 0.9631 & 0.0379 & 0.9397 & 0.0441 & 0.9201 & 0.0501 & 0.9012 \\
    Crossformer & 0.0226 & 0.9759 & 0.0269 & 0.9656 & 0.0358 & 0.9436 & 0.0489 & 0.9291 \\
    DPGNN \cite{art_405} & 0.0184 & 0.9797 & 0.0246 & 0.9715 & 0.0346 & 0.9564 & 0.0456 & 0.9386 \\
    DSTGN \cite{art_73} & 0.0179 & 0.9782 & 0.0250 & 0.9715 & 0.0346 & 0.9562 & - & - \\
    DSTIGNN \cite{art_74} & 0.0173 & 0.9818 & 0.0244 & 0.9723 & 0.0333 & 0.9571 & \underline{0.0430} & 0.9407 \\
    ESG \cite{art_649} & 0.0181 & 0.9792 & 0.0246 & 0.9717 & 0.0345 & 0.9564 & 0.0468 & 0.9392 \\
    FEDformer & 0.0256 & 0.9701 & 0.0287 & 0.9555 & 0.0379 & 0.9385 & 0.0487 & 0.9190 \\
    GP & 0.0239 & 0.8713 & 0.0272 & 0.8193 & 0.0394 & 0.8484 & 0.0580 & 0.8278 \\
    GRU & 0.0192 & 0.9786 & 0.0264 & 0.9712 & 0.0408 & 0.9351 & - & - \\
    GTS & 0.0180 & \underline{0.9898} & 0.0260 & \underline{0.9824} & 0.0333 & \underline{0.9701} & 0.0442 & \underline{0.9518} \\
    GWN & 0.0251 & 0.9740 & 0.0300 & 0.9640 & 0.0381 & 0.9510 & 0.0486 & 0.9294 \\
    Informer & 0.0882 & 0.9563 & 0.1081 & 0.9321 & 0.1301 & 0.8911 & 0.1521 & 0.8021 \\
    LSTNet & 0.0226 & 0.9735 & 0.0280 & 0.9658 & 0.0356 & 0.9511 & 0.0449 & 0.9354 \\
    MAGL \cite{art_702} & 0.0178 & 0.9797 & 0.0246 & 0.9716 & 0.0337 & 0.9567 & 0.0444 & 0.9361 \\
    MAGNN \cite{art_80} & 0.0183 & 0.9778 & 0.0246 & 0.9712 & 0.0343 & 0.9557 & 0.0474 & 0.9339 \\
    MDST-GNN \cite{art_90} & 0.0172 & 0.9811 & 0.0245 & 0.9727 & 0.0337 & 0.9578 & 0.0431 & 0.9392 \\
    MTGNN \cite{art_655} & 0.0194 & 0.9786 & 0.0259 & 0.9708 & 0.0349 & 0.9551 & 0.0456 & 0.9372 \\
    MTHetGNN & 0.0198 & 0.9769 & 0.0259 & 0.9701 & 0.0345 & 0.9539 & 0.0451 & 0.9360 \\
    MTNet & 0.0212 & 0.9767 & 0.0258 & 0.9703 & 0.0347 & 0.9561 & 0.0442 & 0.9388 \\
    RNN-GRU & 0.0192 & 0.9786 & 0.0264 & 0.9712 & 0.0408 & 0.9531 & 0.0626 & 0.9223 \\
    SARIMA & 0.0197 & 0.9748 & 0.0253 & 0.9643 & 0.0338 & 0.9495 & 0.0444 & 0.9370 \\
    SCINet & \underline{0.0171} & 0.9787 & 0.0240 & 0.9704 & 0.0331 & 0.9553 & 0.0436 & 0.9396 \\
    SDGL \cite{art_78} & 0.0180 & 0.9808 & 0.0249 & 0.9730 & 0.0342 & 0.9583 & 0.0455 & 0.9402 \\
    StemGNN & 0.0506 & 0.8871 & 0.0674 & 0.8703 & 0.0676 & 0.8499 & 0.0685 & 0.8738 \\
    TDG4-MSF \cite{art_81} & 0.0172 & 0.9825 & 0.0244 & 0.9718 & 0.0330 & 0.9569 & 0.0433 & 0.9386 \\
    Theta & 0.0497 & 0.9738 & 0.0257 & 0.9656 & 0.0342 & 0.9510 & 0.0441 & 0.9323 \\
    TPA-LSTM & 0.0174 & 0.9790 & 0.0241 & 0.9709 & 0.0341 & 0.9564 & 0.0444 & 0.9381 \\
    TPGNN \cite{art_685} & 0.0174 & 0.9792 & 0.0250 & 0.9687 & 0.0350 & 0.9509 & 0.0458 & 0.9306 \\
    TRMF & 0.0351 & 0.9142 & 0.0875 & 0.8123 & 0.0494 & 0.8993 & 0.0563 & 0.8678 \\
    VAR-MLP & 0.0265 & 0.8609 & 0.0394 & 0.8725 & 0.0407 & 0.8280 & 0.0578 & 0.7675 \\
    \bottomrule
\end{longtable}
}

Also in this case, GNN models are the best. The so-called discrete graph structure learning for time series model (GTS), proposed in 2021 by Shang et al.~in Ref.~\cite{art_641} for traffic and sensor network forecasting, and used as a benchmark in Ref.~\cite{art_78}, demonstrates itself to have a superior performance. It includes a probabilistic graph learning module that optimizes the mean performance over the graph distribution, and a recurrent graph convolutional module to generate the forecasts.

Tab.~\ref{t:risultati solar energy} summarizes the accuracy of the various models on the Solar energy dataset, covering the period from January 1, 2016, to December 31, 2016, at 10-minutes intervals. The analysis considers forecasting horizons of 3, 6, 12, and 24 steps ahead, and the results are expressed in terms of RRSE and CORR.

{\tabcolsep=4.5pt
\begin{longtable}{lcccccccc}
    \caption{Comparison of the average accuracy of the different models on the Solar energy dataset for 3, 6, 12, and 24 steps time horizons, expressed in terms of RRSE and CORR. The numbers corresponding to the highest accuracy are underlined.}
    \label{t:risultati solar energy}
    \endfirsthead
    \endhead
    \toprule
    Time horizon & \multicolumn{2}{c}{3 steps} & \multicolumn{2}{c}{6 steps} & \multicolumn{2}{c}{12 steps} & \multicolumn{2}{c}{24 steps} \\
    Metrics & RRSE & CORR & RRSE & CORR & RRSE & CORR & RRSE & CORR \\
    \midrule
    ADLGNN \cite{art_83} & 0.1708 & 0.9866 & 0.2188 & 0.9768 & 0.2897 & 0.9551 & 0.4128 & 0.9060 \\
    AGCRN & 0.1840 & 0.9841 & 0.2432 & 0.9708 & 0.3185 & 0.9487 & 0.4141 & 0.9087 \\
    AGLG-GRU \cite{art_79} & 0.1762 & 0.9842 & 0.2302 & 0.9682 & 0.3021 & 0.9532 & 0.4130 & 0.9084 \\
    AR & 0.2435 & 0.9710 & 0.3790 & 0.9263 & 0.5911 & 0.8107 & 0.8699 & 0.5314 \\
    ARIMA & 0.2328 & 0.9739 & 0.3413 & 0.9402 & 0.4531 & 0.8886 & 0.5810 & 0.8133 \\
    Autoformer & 0.1935 & 0.9784 & 0.2604 & 0.9651 & 0.3959 & 0.9111 & 0.6064 & 0.8180 \\
    Crossformer & 0.1758 & 0.9801 & 0.2304 & 0.9679 & 0.3101 & 0.9437 & 0.4115 & 0.9001 \\
    DPGNN \cite{art_405} & 0.1794 & 0.9850 & 0.2364 & 0.9724 & 0.3003 & 0.9543 & 0.4091 & 0.9117 \\
    DSTGN \cite{art_73} & 0.1787 & 0.9867 & 0.2358 & 0.9721 & 0.3079 & 0.9513 & - & - \\
    DSTIGNN \cite{art_74} & 0.1684 & \underline{0.9869} & \underline{0.2165} & \underline{0.9773} & \underline{0.2863} & \underline{0.9584} & 0.4064 & 0.9118 \\
    ESG \cite{art_649} & 0.1708 & 0.9865 & 0.2278 & 0.9743 & 0.3073 & 0.9519 & 0.4101 & 0.9100 \\
    FEDformer & 0.1901 & 0.9827 & 0.2444 & 0.9606 & 0.3502 & 0.9275 & 0.4851 & 0.8876 \\
    GP & 0.2259 & 0.9751 & 0.3286 & 0.9448 & 0.5200 & 0.8518 & 0.7973 & 0.5971 \\
    GRU & 0.1932 & 0.9823 & 0.2628 & 0.9675 & 0.4163 & 0.9150 & - & - \\
    GTS & 0.1842 & 0.9842 & 0.2691 & 0.9645 & 0.3259 & 0.9481 & 0.4796 & 0.8678 \\
    GWN & 0.1773 & 0.9846 & 0.2279 & 0.9743 & 0.3068 & 0.9527 & 0.4206 & 0.9055 \\
    HyDCNN & 0.1806 & 0.9865 & 0.2335 & 0.9747 & 0.3094 & 0.9515 & - & - \\
    Informer & 0.2134 & 0.9715 & 0.2701 & 0.9549 & 0.4331 & 0.8985 & 0.7017 & 0.7921 \\
    LSTNet & 0.1843 & 0.9843 & 0.2559 & 0.9690 & 0.3254 & 0.9467 & 0.4643 & 0.887 \\
    MAGL \cite{art_702} & \underline{0.1678} & \underline{0.9869} & 0.2220 & 0.9759 & 0.2998 & 0.9550 & 0.4188 & 0.9057 \\
    MAGNN \cite{art_80} & 0.1771 & 0.9853 & 0.2361 & 0.9724 & 0.3015 & 0.9539 & 0.4108 & 0.9097 \\
    MDST-GNN \cite{art_90} & 0.1764 & 0.9855 & 0.2321 & 0.9735 & 0.3082 & 0.9519 & 0.4119 & 0.9103 \\
    MTGNN \cite{art_655} & 0.1778 & 0.9852 & 0.2348 & 0.9726 & 0.3109 & 0.9509 & 0.4270 & 0.9031 \\
    MTGODE \cite{art_75} & 0.1693 & 0.9868 & 0.2171 & 0.9771 & 0.2901 & 0.9577 & - & - \\
    MTHetGNN & 0.1838 & 0.9845 & 0.2600 & 0.9681 & 0.3169 & 0.9486 & 0.4231 & 0.9031 \\
    MTNet & 0.1847 & 0.9840 & 0.2398 & 0.9723 & 0.3251 & 0.9462 & 0.4285 & 0.9013 \\
    RNN-GRU & 0.1932 & 0.9823 & 0.2628 & 0.9675 & 0.4163 & 0.9150 & 0.4852 & 0.8823 \\
    SARIMA & 0.2227 & 0.9760 & 0.3228 & 0.9461 & 0.4512 & 0.8900 & 0.5714 & 0.8184 \\
    SCINet & 0.1775 & 0.9853 & 0.2301 & 0.9739 & 0.2997 & 0.9550 & 0.4081 & 0.9112 \\
    SDGL \cite{art_78} & 0.1699 & 0.9866 & 0.2222 & 0.9762 & 0.2924 & 0.9565 & 0.4047 & 0.9119 \\
    SDLGNN & 0.1720 & 0.9864 & 0.2249 & 0.9757 & 0.3024 & 0.9547 & 0.4184 & 0.9051 \\
    SDLGNN-Corr & 0.1806 & 0.9848 & 0.2378 & 0.9722 & 0.3042 & 0.9534 & 0.4173 & 0.9067 \\
    StemGNN & 0.1839 & 0.9841 & 0.2612 & 0.9679 & 0.3564 & 0.9395 & 0.4768 & 0.8740 \\
    STG-NCDE & 0.2346 & 0.9748 & 0.2908 & 0.9605 & 0.5149 & 0.8639 & - & - \\
    TDG4-MSF \cite{art_81} & 0.1746 & 0.9858 & 0.2348 & 0.9727 & 0.3082 & 0.9527 & 0.4031 & 0.9143 \\
    Theta & 0.2442 & 0.9685 & 0.3327 & 0.9445 & 0.4488 & 0.8979 & 0.6092 & 0.8139 \\
    TPA-LSTM & 0.1803 & 0.9850 & 0.2347 & 0.9742 & 0.3234 & 0.9487 & 0.4389 & 0.9081 \\
    TPGNN \cite{art_685} & 0.1850 & 0.9840 & 0.2412 & 0.9716 & 0.3059 & 0.9529 & \underline{0.3498} & \underline{0.9710} \\
    VAR-MLP & 0.1922 & 0.9829 & 0.2679 & 0.9655 & 0.4244 & 0.9058 & 0.6841 & 0.7149 \\
    \bottomrule
\end{longtable}
}

The results indicate that the dynamic spatiotemporal interactive graph neural network (DSTIGNN) model \cite{art_74} is the most accurate on the solar energy dataset. The DSTIGNN model, introduced by Gao et al.~in 2021, employs a hierarchical structure consisting of a spatio-temporal interactive learning module, a dynamic graph inference module, multiple spatio-temporal layers, and an output module. The spatio-temporal interactive learning module splits the time series into two subsequences to capture both temporal and spatial correlations. Then, the graph inference module learns a dynamic adjacency matrix for the graph. Once the graph structure is determined, a stack of graph convolutional layers, enhanced with residual connections, along with a convolutional output module, generate the final forecast.

Finally, Tab.~\ref{t:risultati traffic} illustrates the performance of different models on the Traffic dataset, with hourly data between 2015 and 2016. The accuracy of each model is assessed across 3, 6, 12, and 24 steps ahead horizons, with RRSE and CORR as the evaluation metrics.

{\tabcolsep=4.5pt
\begin{longtable}{lcccccccc}
    \caption{Comparison of the average accuracy of the different models on the Traffic dataset for 3, 6, 12, and 24 steps time horizons, expressed in terms of RRSE and CORR. The numbers corresponding to the highest accuracy are underlined.}
    \label{t:risultati traffic}
    \endfirsthead
    \endhead
    \toprule
    Time horizon & \multicolumn{2}{c}{3 steps} & \multicolumn{2}{c}{6 steps} & \multicolumn{2}{c}{12 steps} & \multicolumn{2}{c}{24 steps} \\
    Metrics & RRSE & CORR & RRSE & CORR & RRSE & CORR & RRSE & CORR \\
    \midrule
    ADLGNN \cite{art_83} & 0.4047 & 0.9028 & 0.4201 & 0.8928 & 0.4299 & 0.8876 & 0.4416 & 0.8818 \\
    AGCRN & 0.4379 & 0.8850 & 0.4635 & 0.8670 & 0.4694 & 0.8679 & 0.4707 & 0.8664 \\
    AGLG-GRU \cite{art_79} & 0.4173 & 0.8958 & 0.4722 & 0.8541 & 0.4427 & 0.8755 & 0.4526 & 0.8842 \\
    AR & 0.5991 & 0.7752 & 0.6218 & 0.7568 & 0.6252 & 0.7544 & 0.6300 & 0.7519 \\
    ARIMA & 0.5841 & 0.7959 & 0.6194 & 0.7655 & 0.6197 & 0.7652 & 0.6248 & 0.7610 \\
    GP & 0.6082 & 0.7831 & 0.6772 & 0.7406 & 0.6406 & 0.7671 & 0.5995 & 0.7909 \\
    GTS & 0.4665 & 0.8695 & 0.4779 & 0.8582 & 0.4792 & 0.8589 & 0.4766 & 0.8573 \\
    GWN & 0.4484 & 0.8801 & 0.4689 & 0.8674 & 0.4725 & 0.8646 & 0.4741 & 0.8646 \\
    HyDCNN & 0.4198 & 0.8915 & 0.4290 & 0.8855 & 0.4352 & 0.8858 & 0.4423 & 0.8819 \\
    LSNet & 0.4777 & 0.8721 & 0.4893 & 0.8690 & 0.4950 & 0.8614 & 0.4973 & 0.8588 \\
    MAGL \cite{art_702} & 0.4058 & 0.9013 & \underline{0.4189} & 0.8935 & 0.4308 & 0.8871 & 0.4417 & 0.8810 \\
    MAGNN \cite{art_80} & 0.4097 & 0.8992 & 0.4555 & 0.8753 & 0.4423 & 0.8815 & 0.4434 & 0.8813 \\
    MDST-GNN \cite{art_90} & 0.4162 & 0.8958 & 0.4461 & 0.8803 & 0.4377 & 0.8841 & 0.4452 & 0.8792 \\
    MTGNN \cite{art_655} & 0.4162 & 0.8963 & 0.4754 & 0.8667 & 0.4461 & 0.8794 & 0.4535 & 0.8810 \\
    MTGODE \cite{art_75} & 0.4127 & 0.9000 & 0.4259 & \underline{0.8945} & 0.4329 & 0.8899 & - & - \\
    MTHetGNN & 0.4826 & 0.8643 & 0.5198 & 0.8452 & 0.5147 & 0.8744 & 0.5250 & 0.8418 \\
    MTNet & 0.4764 & 0.8728 & 0.4855 & 0.8681 & 0.4877 & 0.8644 & 0.5023 & 0.8570 \\
    RNN-GRU & 0.5358 & 0.8511 & 0.5522 & 0.8405 & 0.5562 & 0.8345 & 0.5633 & 0.8300 \\
    SARIMA & 0.5823 & 0.7967 & 0.5974 & 0.7837 & 0.6002 & 0.7811 & 0.6151 & 0.7697 \\
    SCINet & 0.4216 & 0.8920 & 0.4414 & 0.8809 & 0.4495 & 0.8772 & 0.4453 & 0.8825 \\
    SDGL \cite{art_78} & 0.4142 & 0.9010 & 0.4475 & 0.8825 & 0.4584 & 0.8760 & 0.4571 & 0.8766 \\
    SDLGNN & 0.4053 & 0.9017 & 0.4209 & 0.8925 & 0.4313 & 0.8868 & 0.4444 & 0.8801 \\
    SDLGNN-Corr & 0.4227 & 0.8937 & 0.4378 & 0.8846 & 0.4576 & 0.8746 & 0.4579 & 0.8784 \\
    TDG4-MSF \cite{art_81} & 0.4029 & 0.9014 & 0.4196 & 0.8925 & \underline{0.4294} & 0.8864 & \underline{0.4366} & 0.8834 \\
    Theta & 0.6071 & 0.7768 & 0.6241 & 0.7606 & 0.6271 & 0.7591 & 0.6012 & 0.7830 \\
    TPA-LSTM & 0.4487 & 0.8812 & 0.4658 & 0.8717 & 0.4641 & 0.8717 & 0.4765 & 0.8629 \\
    TPGNN \cite{art_685} & \underline{0.3989} & \underline{0.9232} & 0.4715 & \underline{0.8945} & 0.4476 & \underline{0.9028} & 0.4696 & \underline{0.8858} \\
    TRMF & 0.6708 & 0.6964 & 0.6261 & 0.7430 & 0.5956 & 0.7748 & 0.6442 & 0.7278 \\
    VAR-MLP & 0.5582 & 0.8245 & 0.6570 & 0.7695 & 0.6023 & 0.7929 & 0.6146 & 0.7891 \\
    \bottomrule
\end{longtable}
}

The accuracy of the models varies by horizon and metric. Overall, the best models are the temporal polynomial graph neural network (TPGNN) \cite{art_685} already discussed for the Electricity consumption dataset, and the temporal decomposition enhanced graph neural network for multivariate time series forecasting (TDG4-MSF) \cite{art_81}. The TDG4-MSF model, proposed by Miao et al.~in 2021, is composed of four components: temporal decomposition enhanced representation learning for extracting short and long term periodic patterns, graph structure learning, gated graph neural network depending on linear transformation for representation learning, and MLP-based output layer.

There are also other pairs of papers whose results can be compared. However, due to space limitations, we do not include tables comparing the results of just two papers in this \enquote{Generic} subsection. Instead, the reader is referred to the studies Refs.~\cite{art_640} and \cite{art_688} for the CSSE COVID-19 and ECG 5000 datasets; Refs.~\cite{art_657} and \cite{art_674} for the Bytom dataset; Refs.~\cite{art_76} and \cite{art_687} for the Jena weather dataset.

\subsection{Other topics}
\label{subsection:Other}
This final subsection includes 34 papers that fall outside the previously discussed groups. The case studies include cellular traffic prediction, modulation classification, and other forecasting or classification problems. Given the heterogeneity of the papers, only a general overview is provided here, and no comparisons of the approaches or discussions of benchmarks are presented.

\subsubsection{Overview}
The interest in GNNs has grown so extensively that the selected themes are actually not sufficient to cover all relevant application domains. However, since only a few papers would belong to each potential additional theme, they are grouped under the label \enquote{Other topics}. A frequently discussed topic in this \enquote{Other topics} category is cellular traffic prediction \cite{art_138, art_144, art_148, art_468, art_457, art_717, art_743}, in terms of SMS messages, calls, internet connections, phone's traffic statistics or locations, which are essential for the cellular network resource management system. Another topic included is modulation classification \cite{art_12, art_13, art_303, art_459, art_463}, which is the process of determining the modulation used at the transmitter based on observations of the received signal \cite{modulation_classification_definition}. Four further areas of investigation to be added are the study of product demand and user preferences \cite{art_744, art_147} to enable personalized experiences and recommendations, anomalous IoT intrusion detection \cite{art_456, art_460}, encrypted traffic classification \cite{art_118, art_679}.

Other applications discussed in the papers of this subsection are urban spatio-temporal event prediction \cite{art_136}, prediction of the quality of operational processes \cite{art_137}, seismic intensity forecasting \cite{art_139} (with source code available at \url{https://github.com/StefanBloemheuvel/GCNTimeseriesRegression}), prediction of spatio-temporal dynamics with complex structures \cite{art_140}, tailing dam monitoring \cite{art_141}, subsurface production forecasting \cite{art_143}, time series prediction for data centers maintenance \cite{art_145}, hydraulic runoff \cite{art_146}, water demand forecasting \cite{art_149}, crime prediction \cite{art_465} (with source code available at \url{https://github.com/ZJUDataIntelligence/HDM-GNN}), polypropylene production data \cite{art_455}, tungsten flotation process \cite{art_458} (with source code available at \url{https://github.com/ywmcsu/SASGNN}), talent and demand supply for labor market \cite{art_678} (with source code available at \url{https://github.com/gzn00417/DH-GEM}), anomaly detection in microservice systems \cite{art_746} (with source code available at \url{https://github.com/ant-research/microservice_system_twin_graph_based_anomaly_detection} $^\dagger$), and emergency supply forecasting for logistics systems \cite{art_750}.
    
\subsubsection{Datasets}
Among the papers selected here, there are only four shared datasets, two for modulation classification, one for cellular traffic prediction, and one for botnet detection in IoT network traffic. However, the datasets are handled in different ways, either in terms of pre-processing techniques, selected time windows, or train/test split percentages. For this reason, the results of these papers are not directly comparable to each other.

The two datasets for modulation classification used in Refs.~\cite{art_12, art_13, art_303, art_463} are RML2016.10a (\url{https://www.kaggle.com/datasets/raindrops12/rml201610a}, source paper: \url{http://dx.doi.org/10.1007/978-3-319-44188-7_16}) and RML2016.10b (\url{https://www.kaggle.com/datasets/marwanabudeeb/rml201610b}, source paper: \url{https://pubs.gnuradio.org/index.php/grcon/article/view/11}). They include 220,000 and 1.2 million samples respectively, in many modulation types. Each modulation type has 20 levels of signal-to-noise ratios at 2 dB intervals from -20 dB to 18 dB.

As for the cellular traffic prediction dataset used in Refs.~\cite{art_123, art_138, art_144, art_468, art_717}, it was proposed by Telecom Italia and MIT Media Lab during the \enquote{Telecom Italia Big Data Challenge}, and it is available at \url{https://dataverse.harvard.edu/dataverse/bigdatachallenge}. The investigated geographical area is divided in a grid, and each cell records the number of SMS messages, calls, and wireless network traffic data in a 10 minutes interval.

Finally, the Bot-IoT dataset used in Refs.~\cite{art_456, art_460} and available at \url{https://www.unsw.adfa.edu.au/unsw-canberra-cyber/cybersecurity/ADFA-NB15-Datasets/bot_iot.php}, was created by designing a realistic network environment in the Cyber Range Lab of UNSW Canberra, and incorporates a combination of normal and botnet traffic.

\section{Discussion}
\label{section:Discussion}
As stated in the Introduction, the aim of this SLR is to provide a comprehensive but also detailed overview of spatio-temporal GNN models for time series classification and forecasting used across different fields, assessing their performance and highlighting patterns that could guide future research. Existing literature was collected and synthesized to provide both an overview and a distillation of datasets, models, and tables of results, addressing the two sets of research questions posed in the Introduction.

Although our intention was to provide a broad and general perspective on widely used spatio-temporal GNN models and identify patterns that could be generalized across different domains and datasets, our analysis revealed that such generalizability is highly elusive. This characteristics stems from the primary strength of spatio-temporal GNNs: the possibility of explicitly modeling relationships between different variables, which is inherently dataset-dependent and highly heterogeneous across different applications. For instance, models for human mobility (e.g., traffic flow, epidemic spread) share intrinsic similarities but differ in temporal scales, spatial resolution, and interaction types. Similarly, environmental applications such as weather forecasting and energy demand prediction are both influenced by climatic factors but rely on different variables and relationships.

We believe that a comprehensive perspective can be drawn after such an extensive examination of the available literature. Hence, we attempt now to discuss common patterns and insights that are not immediately apparent across different domains. In this section, we first address the research questions posed in the Introduction, then summarize prevailing research trends, and finally present a meta-analysis aimed at uncovering common insights.

\subsection{Research questions}
Most of the general overview questions (GQs) were addressed in the overview presented in Sec.~\ref{section:Overview of the publications}. As for the tools, the GNN community mostly uses Python and PyTorch, although some researchers use TensorFlow. Unfortunately only few papers provide a link to the source code of the proposed model. Almost all papers are funded by public or private entities.

As for the specific questions (SQs), the answers are the following.

\textbf{SQ1 (Applications).}
The three most investigated fields are \enquote{Mobility}, \enquote{Environment}, and \enquote{Generic}. Regarding the first two thematic groups, their datasets can be naturally translated into graphs, which explains why they have been studied more extensively over time. In contrast, the \enquote{Generic} thematic group has only recently gained a significant interest. As for the differences among the thematic groups, the main ones are related to the definition of the graph, which is not always explicit, and the benchmarks used, often related to the mindset of specific communities. It is not easy to compare results across different applications and to determine the most promising fields. The studies in the \enquote{Generic} thematic group appear promising across a wide range of domains.

\textbf{SQ2 (Graph construction).}
Most of the selected papers focus on pre-defined graph structures (when available), with the objective of extracting the maximum amount of information and enhancing the interpretability of the model. However, there is a recent growing interest in models that learn the graph structure and the edge weights themselves. While this trend toward adaptive graph learning is expected to gain momentum, the literature remains divided on whether learned or pre-defined adjacency matrices are preferable. The choice between these approaches should be based on the availability of domain knowledge and model performance requirements, requiring further research to establish clearer guidelines.

\textbf{SQ3 (Taxonomy).}
As observed in Ref.~\cite{survey_spatiotemporal}, there are two main approaches for the design of spatio-temporal GNNs: one that treats the spatial and temporal components in separate modules, and another that integrates and processes them together. The analysis of the collected papers reveals that the most common approach in literature is that of separate modules. Specifically, researchers very often address the spatial and temporal aspects of the problem independently from each other, and focus on each module separately. Hence, the proposed taxonomy of GNN models takes its main perspective from the point of view of the spatial component. The review indicates that convolutional approaches are the most prevalent, followed by attentional methods. Approximately 60\% of the models are purely convolutional, 22\% are purely attentional, and 10\% are hybrid convolutional-attentional. Additional statistics related to taxonomy distributions by year and research field are available in the GitHub repository associated with this review. With regard to the temporal component, the recurrent structures of GRUs and the attention mechanism are widely used.

\textbf{SQ4 (Benchmark models).}
As for the benchmark models, there are many options available, and their choice depends on the specific application. In thematic groups such as \enquote{Energy}, \enquote{Finance}, \enquote{Health}, and \enquote{Predictive monitoring}, the focus tends to be on simpler statistical and non-GNN machine learning benchmarks, with only a limited use of GNN benchmarks. In the \enquote{Generic} group, many recent machine learning benchmark models based on the Transformer architecture have emerged. Notably, in \enquote{Mobility}, \enquote{Environment}, and \enquote{Generic} groups (which are also the most investigated fields), there are many reference GNN benchmark models. The most widely used GNN benchmarks, spanning all thematic groups, include: the graph convolutional neural network with long-short term memory (GC-LSTM) \cite{GC-LSTM}, spatio-temporal graph convolutional network (STGCN) \cite{STGCN}, attention-based spatial-temporal graph convolutional network (ASTGCN) \cite{ASTGCN}, diffusion convolutional recurrent neural network (DCRNN) \cite{DCRNN}, and multivariate time series forecasting with graph neural network (MTGNN) \cite{art_655}. These GNN benchmarks, mostly published between 2018 and 2020, represent some of the earliest and most cited works in spatio-temporal GNN literature. Almost all of them employ graph convolutional mechanisms, often Chebyshev-based or similar variants, reflecting the state-of-the-art at the time. In most cases, the adjacency matrix is pre-defined, often based on distances between measurement sites, whereas just one approach incorporates a learned adjacency matrix (i.e.,  MTGNN, which is one of the first models to use self-learned graph structures). In terms of application domains, four out of five of these benchmarks were originally developed to focus on \enquote{Mobility} or \enquote{Generic} tasks, with only GC-LSTM originally addressing environmental applications. Among these benchmarks, ASTGCN also explores attention mechanisms alongside graph convolutions and constitutes one of the earliest hybrid architectures. Overall, these benchmarks illustrate the early development of GNNs: graph convolution dominates, adjacency matrices are usually static, and mobility-related tasks are the most commonly studied.

\textbf{SQ5 (Benchmark datasets).}
The datasets mentioned in the selected papers are closely related to the specific case of study. Even though some benchmark datasets are listed in the \enquote{Generic} thematic group, there is no common standard dataset for the entire spatio-temporal GNN research community. Hence, it would be beneficial for the community to agree on, and start adopting, some of the most commonly used datasets, with the goal of developing a shared benchmark dataset. Some possible common benchmark could be the ones presented in the \enquote{Mobility} and the \enquote{Generic} thematic group. Hopefully, this review could serve this purpose by providing comprehensive tables of the results of the models on different datasets to facilitate comparison between spatio-temporal GNN models.

\textbf{SQ6 (Modeling paradigms).}
As for the modeling paradigm used in each paper, most of the selected papers work with a homogeneous graph, which models the relationships among multiple entities of the same nature. This is due to the fact that many basic GNN algorithms were originally developed for homogeneous graphs, and because it is easier to identify relationships between quantities of the same type. Also, in many cases, the focus is on multivariate series where the variables of interest and the target quantities are inherently of the same nature.

\textbf{SQ7 (Metrics).}
Regarding the error metrics used in the papers, their choice is highly dependent on the specific case study. However, the most common ones are mean absolute error (MAE), mean squared error (MSE), and mean absolute percentage error (MAPE) for forecasting problems, and accuracy for classification problems.

\subsection{Observed trends}
Our analysis of papers from different domains indicates that some thematic groups have seen more intense research on GNN models than others. For instance, the papers in the \enquote{Mobility} and \enquote{Generic} groups generally present more complex models, likely reflecting the greater research effort in these fields, as evidenced by the larger number of published papers. In contrast, the \enquote{Finance} and \enquote{Health} thematic groups have fewer papers and often offer only a limited number of benchmarks for comparison. Another notable example is the \enquote{Energy} group, which has a relatively large number of papers, particularly in 2024. However, the related community tends to rely more on traditional econometrics or non-GNN machine learning approaches, and the GNN models proposed are generally less complex than those in other domains. Moreover, many papers in the \enquote{Energy} thematic group test their models on proprietary datasets, which further complicates model development across the group. Overall, this analysis suggests that some fields are more prone to the development of innovative and complex models, whereas others focus on adapting existing models to specific real-world problems. Consequently, we expect some form of cross-pollination between the communities, which is likely to reduce these differences over time. For readers interested in a more detailed view of how GNN taxonomies evolve across fields and over time, interactive charts are available in the GitHub repository accompanying this paper.

Another important aspect related to research trends is the focus on reproducibility and comparability of studies, which, more in general, is essential for advancing scientific knowledge. Some fields pay greater attention to reproducibility, facilitated by the availability of datasets, a larger number of benchmarks, and more detailed model descriptions. Tab.~\ref{t:tabella finale} shows the average number of datasets and benchmark models used, as well as the percentage of papers reporting the source code for each group.
\begin{table}[ht]
    \caption{Average number of datasets and benchmark models used, and percentage of papers with the link to the source code for each group.}
    \label{t:tabella finale}
    \centering
    \begin{tabular}{lccc}
        \toprule
        Group & Datasets (avg.) & Benchmarks (avg.) & Code availability \\
        \midrule
        Energy & 1.6 & 5.7 & 3\% \\
        Environment & 1.3 & 6.6 & 23\% \\
        Finance & 1.5 & 6.3 & 7\% \\
        Generic & 4.5 & 10.6 & 56\% \\
        Health & 1.7 & 8.3 & 48\% \\
        Mobility & 2.8 & 9.4 & 35\% \\
        Other & 1.9 & 6.3 & 15\% \\
        Predictive monitoring & 2.2 & 7.9 & 22\% \\
        \bottomrule
    \end{tabular}
\end{table}

\subsection{Meta-analysis across groups}
Here, we conduct a meta-analysis of the reviewed spatio-temporal GNNs models across multiple application domains. By definition, all models incorporate both spatial and temporal components, either arranged sequentially (space-then-time) or jointly (space-and-time). Beyond this common feature, designs vary widely, and our meta-analysis focuses on the key aspects that characterize these models, namely graph construction and spatial aggregation, aiming to reveal patterns where possible.

However, the highly fragmented research landscape makes identifying these trends challenging. We observed that many models are tailored to specific case studies, and differences in domain focus, limited reproducibility, and inconsistent experimental setups further complicate cross-domain comparisons and the extraction of generalizable insights. For this reason, rather than attempting a meta-analysis across the entire fragmented landscape of reviewed papers, we focused on the best-performing models within each domain to identify shared characteristics and emerging trends. In particular, we selected those models that, in the results tables, ranked among the top performers for the associated datasets and that were briefly discussed below each table. These models were gathered across all groups and are available in the GitHub repository associated with this work.

In this selected subset, models from the \enquote{Generic} and \enquote{Mobility} thematic groups are more numerous. This is not because they are inherently more important, but because these models offer greater comparability, having been evaluated across multiple common datasets in their original studies. For this reason, they are particularly suitable for this broader meta-analysis, which does not focus on highly specific use cases.

As for the determination of the graph structure, most models in the selected subset have a self-learning module. Pre-defined adjacency matrices based on physical distances, correlations, cosine similarities, or similar techniques remain more common in studies focused on specific tasks. From a historical perspective, early models relied more heavily on pre-defined adjacency matrices, whereas more recent works increasingly favor learned or hybrid adjacency matrices, often combining learned and pre-defined structures to enhance adaptivity and robustness. This shift toward learned or hybrid adjacency matrices, often implemented via self-learning modules, supports broader applicability across diverse domains. Among self-learning modules in spatio-temporal GNNs, two predominant approaches have emerged with comparable prevalence in the selected subset of papers: the method of the GWN model \cite{GWN} and that of the MTGNN model \cite{art_655}, both developed by Wu et al, and discussed in Subsec.~\ref{subsection:Determination of the graph structure}.

As for the spatial aggregation module, the subset reveals that graph convolution remains the dominant spatial aggregation mechanism in high-performing spatio-temporal GNNs, while attention-based spatial aggregation is less common. This suggests that the practical advantages of convolutional aggregation, in terms of efficiency and training stability, continue to make it preferable to attention mechanisms. The aggregation mechanism of recurrent GNNs, which defined the original GNN formulation, is instead almost entirely unused.

\section{Limits, challenges, and future research directions}
\label{section:Limits, challenges, and future research directions}
In this section the limitations and challenges of spatio-temporal GNN modeling will be discussed, together with directions for further research.

\textbf{Comparability.}
As highlighted in Ref.~\cite{cit_rev2}, the evaluation of GNN models has improved thanks to the introduction of the Open Graph Benchmark (OGB) \cite{NEURIPS2020_fb60d411}, which provides a standardized evaluation framework and a variety of benchmark graph datasets. However, there is currently no standardized benchmark for spatio-temporal GNNs. As a result, each model is evaluated on its own selection of datasets, and this fragmentation makes it difficult to compare the results of different studies. To alleviate this problem, this review presented all the information gathered from the selected papers, including datasets, benchmarks, codes, and tables of results. All that is intended to serve as a detailed overview and a foundation for further exploration, in the hope that researchers will begin to examine the data collected in the presented tables and, over time, identify relevant datasets and benchmark models. Moreover, we encourage the authors to discuss the relationship between complexity and performance of the proposed models to facilitate a more comprehensive comparison.

\textbf{Reproducibility.}
The limited availability of links to repositories, source code, and datasets makes it difficult to evaluate the progress of knowledge in spatio-temporal GNNs. Moreover, in many papers the authors do not give detailed information about the model and the graph structure, like definition of the graph, number of nodes, calculation of edge weights. This issue complicates the verification of results and the assessment of the reproducibility of experiments, which are essential to drive further research in this field. Without access to this fundamental information, it is extremely difficult to assess the accuracy and significance of the proposed models.

\textbf{Poor information capacity.}
Another limitation of GNNs lies in the limited amount of information used in the design of the graph structure and the model definition. A poorly constructed graph with unrepresentative or overly sparse connections, as well as a lack of physical constraints in the equations of the model, can significantly affect the performance. Although efforts have been made to address these issues, such as integrating physical constraints through differential equations, these approaches are not always effective. Therefore, it is essential to develop new techniques to overcome these limitations.

\textbf{Scalability.}
Spatio-temporal GNNs are widely used for modeling and analyzing large and complex time series-based network structures. However, GNN models often scale badly, and require a significant amount of memory and computing resources to calculate the adjacency matrix and node embeddings, especially in the case of dynamic graph structures. This challenge can result in high computational costs, and in many cases in the necessity of using intensively GPU devices. Therefore, the development of more scalable GNN models is crucial to facilitate time series analysis even in environments with limited computing power.

\textbf{Heterogeneity.}
The majority of current GNN models are designed to deal with homogeneous graphs, where nodes and edges are all of the same nature. As a result, it would be challenging to use these GNNs with heterogeneous graphs which have different types of nodes and edges or different inputs. Therefore, further research is necessary to develop GNN models that can effectively capture the interactions between various types of nodes and edges.

\textbf{Explainability.}
A crucial aspect for GNN models is explainability, intended as the ability to interpret and understand the decision-making processes implemented in them. Despite their attempt to explicitly model spatial relationships between series, GNNs are often considered as \enquote{black box} models. Remarkably, none of the selected papers adequately tackle the concept of explainability. Some models are claimed to be explainable because of the integration of non-linear physical laws (e.g., Ref.~\cite{art_401}), while other papers claim explainability on the basis of the possibility to visualize a graph (e.g., Ref.~\cite{art_360}) or because they present a feature aggregation form that can be intuitively understood. In addition, other approaches focus on determining feature importance through tools such as Captum \cite{Captum}, which enables the interpretation and understanding of PyTorch model predictions by exploring the features that contribute to a specific outcome (e.g., Ref.~\cite{art_361}). This highlights a research gap in this area. However, recent literature is beginning to investigate this important topic, as also discussed in \cite{art_700}. For instance, few papers studied perturbation-based explanation methods. 
Other approaches, such as GNNExplainer \cite{interpret_2}, aim to identify a small subset of node features that have a crucial role in GNN's predictions, as in Ref.~\cite{art_349, art_357}. Recently, some researchers have started to explore graph counterfactuals as a means of generating explanations, as reviewed in Ref.~\cite{interpret_1}. For a recent survey on GNNs for explainable AI see Ref.~\cite{Nandan2025}.

\textbf{Quasi-equilibrium learning.}
As pointed out in Subsec.~\ref{subsection:Overview of graph neural networks}, GNNs were originally introduced in Ref.~\cite{Scarselli2009} by promoting an extension of the backpropagation algorithm which exploits a fixed-point principle. The subsequent introduction of the notion of convolutional graph has fundamentally changed such computational scheme. It is worth mentioning that in contemporary GNNs there is no need to wait for the system to reach equilibrium. For example, the deep neural network scheme no longer requires weight-sharing. These improvements have led to successful applications in many real-world problems, where modern features contribute significantly to both accuracy and computational efficiency. However, when analyzing the processing of temporal information, GNNs are often implemented within a single graph domain where a temporal function is defined. This is a case where the original computational approach proposed in Ref.~\cite{Scarselli2009} seems to be more adequate, especially when considering that the rate of weight updates is generally much slower than the rate of external temporal signals. This suggests the necessity of revisiting the process of convergence to the fixed point, and of exploring computational schemes which take place at quasi-equilibrium points. We expect that this approach can significantly alleviate the computational burden that is typical of the convergence to fixed-points required for spatio-temporal graph collections.

\textbf{Broader Directions.}
The field of time series modeling by means of GNNs is expected to expand in a set of further directions. While most of the reviewed studies focus on point forecasting with deterministic models, relatively few address probabilistic forecasting (e.g., Refs.~\cite{art_698, art_695}), which is crucial for assessing uncertainty in predictions. Another potential direction could be the integration of spatio-temporal GNNs with transformer-based architectures or large language models. Although studies are still limited, such models could help capture long term dependencies and complex patterns in time series data, yet not forgetting all possible limitations of LLMs in time series applications \cite{tan2024language}. Exploring these directions may help overcome current limitations in modeling uncertainty and complex temporal relationships, providing more informative and reliable results.

\section{Conclusions}
\label{section:Conclusions}
This paper presented the results of a SLR on the application of spatio-temporal GNN models to time series classification and forecasting problems in different fields. The interest on this subject comes from the fact that lately, GNNs have gained significant popularity due to their ability to process graph-structured data. This has led in more recent years to the development of spatio-temporal GNNs and their application to time series analysis, due to their ability to model dependencies between variables and across time points at once.

This SLR has brought forward two sets of research questions: generic questions which can be answered at the level of a bibliographic overview, and more specific questions regarding particular aspects of the proposed spatio-temporal GNN models. The answers to these questions provide a list of interesting considerations.

It has emerged that the majority of the models in the selected papers is characterized by a convolutional spatial aggregation. However, graph attentional models are also emerging as important tools. In addition, while the majority of the selected papers focus on models with a pre-defined graph structure, an increasing number of studies, particularly in the \enquote{Generic} thematic group, are beginning to develop models that learn the graph structure autonomously.

Another key point which results from the presented overview is that the current literature on spatio-temporal GNNs is very fragmented, as it was pictorially reported in Fig.~\ref{f:grafico_gruppi}. This can be attributed to the fact that the involved researchers come from different communities which focus on distinct application domains. As a result, there seems to be a lack of standardized datasets or benchmarks in the studies. The objective of this review was also to collect information on datasets, proposed models, links to source codes, benchmarks, and results, in order to provide common background information for future studies. To support this goal, the GitHub repository associated with this review offers interactive tools for sorting, filtering, and visualizing the results achieved by the different models, making it easier for researchers to explore and compare existing work.

A further major key point resulting from our review, which we believe should be emphasized in these conclusions, is the need for the spatio-temporal GNN research community to work on common datasets and develop comparable and reproducible models. That would enhance transparency, and it would make it easier for all researchers to reciprocally evaluate advancements in the field and to share them more easily.

\section*{Supplementary material}
Link to the GitHub repository containing information on the 366 reviewed papers organized by groups, along with interactive tables and charts: \url{https://github.com/FlaGer99/SLR-Spatio-Temporal-GNN.git}.

\section*{CRediT author statement}
\textbf{Flavio Corradini:} Project administration, Funding acquisition.
\textbf{Flavio Gerosa:} Investigation, Writing - Review \& Editing, Visualization.
\textbf{Marco Gori:} Writing - Review \& Editing, Supervision.
\textbf{Carlo Lucheroni:} Writing - Review \& Editing, Supervision.
\textbf{Marco Piangerelli:} Conceptualization, Methodology, Writing - Original Draft, Writing - Review \& Editing, Supervision.
\textbf{Martina Zannotti:} Conceptualization, Methodology, Validation, Formal analysis, Investigation, Data Curation, Writing - Original Draft, Writing - Review \& Editing, Visualization. \\

Author names appear in alphabetical order and do not reflect relative contributions.



\section*{Acknowledgments}
This work has been funded by the European Union - NextGenerationEU, Mission 4, Component 2, under the Italian Ministry of University and Research (MUR) National Innovation Ecosystem grant ECS00000041 - VITALITY - CUPJ13C22000430001; 
and the NextGenerationEU - National Recovery and Resilience Plan (PNRR) grant CUP J11J23001450006, with the contribution of Syeew srl. 


\appendix

\section{List of selected papers}
\label{a:appendix_papers}
Tab.~\ref{t:selected papers} lists all the papers included in this SLR, together with the year of publication, group they belong to, case study, and nature of the task (classification or forecasting).

{\tabcolsep=5pt
\begin{longtable}{cclll}
    \caption{List of the selected papers, with year, group they belong to, case study, and nature of the task.}
    \label{t:selected papers}
    \endfirsthead
    \endhead
    \toprule
    Reference & Year & Group & Case study & Task \\
    \midrule
    \cite{art_24} & 2020 & Energy & Wind power & Forecasting \\
    \cite{art_635} & 2021 & Energy & Electric load & Forecasting \\
    \cite{art_20} & 2022 & Energy & Wind power & Forecasting \\
    \cite{art_21} & 2022 & Energy & Photovoltaic power & Forecasting \\
    \cite{art_22} & 2022 & Energy & Photovoltaic power & Forecasting \\
    \cite{art_23} & 2022 & Energy & Power system transient dynamics & Forecasting \\
    \cite{art_25} & 2022 & Energy & Urban energy consumption & Forecasting \\
    \cite{art_27} & 2022 & Energy & Wind power & Forecasting \\
    \cite{art_15} & 2023 & Energy & Heat load & Forecasting \\
    \cite{art_16} & 2023 & Energy & Electricity load & Forecasting \\
    \cite{art_17} & 2023 & Energy & Residential load & Forecasting \\
    \cite{art_18} & 2023 & Energy & Electricity load & Forecasting \\
    \cite{art_26} & 2023 & Energy & Solar power & Forecasting \\
    \cite{art_332} & 2024 & Energy & Wind power & Forecasting \\
    \cite{art_333} & 2024 & Energy & Wind power & Forecasting \\
    \cite{art_334} & 2024 & Energy & Air conditioning control & Forecasting \\
    \cite{art_335} & 2024 & Energy & Air conditioning control & Forecasting \\
    \cite{art_337} & 2024 & Energy & Electrical demand and prices & Forecasting \\
    \cite{art_338} & 2024 & Energy & Electric load & Forecasting \\
    \cite{art_339} & 2024 & Energy & Electric load & Forecasting \\
    \cite{art_340} & 2024 & Energy & Chiller energy consumption & Forecasting \\
    \cite{art_341} & 2024 & Energy & Electricity demand & Forecasting \\
    \cite{art_342} & 2024 & Energy & Electric power anomalies & Classification \\
    \cite{art_343} & 2024 & Energy & Electricity demand & Forecasting \\
    \cite{art_344} & 2024 & Energy & Electricity consumption & Forecasting \\
    \cite{art_345} & 2024 & Energy & District heating networks & Forecasting \\
    \cite{art_347} & 2024 & Energy & Power quality disturbances & Classification \\
    \cite{art_348} & 2024 & Energy & Residential load & Forecasting \\
    \cite{art_349} & 2024 & Energy & Photovoltaic energy & Forecasting \\
    \cite{art_350} & 2024 & Energy & Electric vehicle charging stations & Forecasting \\
    \cite{art_351} & 2024 & Energy & Natural gas consumption & Forecasting \\
    \cite{art_353} & 2024 & Energy & Networked microgrids & Forecasting \\
    \cite{art_390} & 2024 & Energy & Multivariate time series & Forecasting \\
    \cite{art_40} & 2021 & Environment & Wind speed & Forecasting \\
    \cite{art_41} & 2021 & Environment & Sea temperature & Forecasting \\
    \cite{art_636} & 2022 & Environment & Water temperature, stream flow & Forecasting \\
    \cite{art_8} & 2022 & Environment & PM 2.5 concentration & Forecasting \\
    \cite{art_38} & 2022 & Environment & Frost & Both \\
    \cite{art_39} & 2022 & Environment & Sea temperature & Forecasting \\
    \cite{art_637} & 2023 & Environment & Soil Moisture & Forecasting \\
    \cite{art_1} & 2023 & Environment & Air quality index & Forecasting \\
    \cite{art_2} & 2023 & Environment & PM 2.5 concentration & Forecasting \\
    \cite{art_3} & 2023 & Environment & PM 2.5 concentration & Forecasting \\
    \cite{art_4} & 2023 & Environment & PM 2.5 concentration & Forecasting \\
    \cite{art_5} & 2023 & Environment & Air quality index & Forecasting \\
    \cite{art_6} & 2023 & Environment & PM 2.5 concentration & Forecasting \\
    \cite{art_7} & 2023 & Environment & PM 2.5 concentration & Forecasting \\
    \cite{art_28} & 2023 & Environment & Sea temperature & Forecasting \\
    \cite{art_29} & 2023 & Environment & Wind speed & Forecasting \\
    \cite{art_30} & 2023 & Environment & Water quality & Forecasting \\
    \cite{art_31} & 2023 & Environment & Wind speed & Forecasting \\
    \cite{art_32} & 2023 & Environment & Significant wave height & Forecasting \\
    \cite{art_33} & 2023 & Environment & Sea temperature & Forecasting \\
    \cite{art_34} & 2023 & Environment & Wind speed & Forecasting \\
    \cite{art_35} & 2023 & Environment & Sea temperature & Forecasting \\
    \cite{art_36} & 2023 & Environment & Sea temperature & Forecasting \\
    \cite{art_43} & 2023 & Environment & Groundwater level & Forecasting \\
    \cite{art_44} & 2023 & Environment & Wind speed & Forecasting \\
    \cite{art_45} & 2023 & Environment & Rainfall & Forecasting \\
    \cite{art_46} & 2023 & Environment & Wind speed & Forecasting \\
    \cite{art_47} & 2023 & Environment & Sea temperature & Forecasting \\
    \cite{art_691} & 2024 & Environment & Weather & Forecasting \\
    \cite{art_354} & 2024 & Environment & PM2.5 prediction & Forecasting \\
    \cite{art_355} & 2024 & Environment & Sea clutter & Forecasting \\
    \cite{art_357} & 2024 & Environment & Plankton dynamics & Forecasting \\
    \cite{art_359} & 2024 & Environment & Groundwater data & Forecasting \\
    \cite{art_360} & 2024 & Environment & Carbon emission & Forecasting \\
    \cite{art_361} & 2024 & Environment & PM2.5 prediction & Forecasting \\
    \cite{art_363} & 2024 & Environment & Water demand & Forecasting \\
    \cite{art_364} & 2024 & Environment & Water quality and flow rate & Forecasting \\
    \cite{art_365} & 2024 & Environment & Weather & Forecasting \\
    \cite{art_366} & 2024 & Environment & Wind speed scenario & Forecasting \\
    \cite{art_368} & 2024 & Environment & Air quality & Forecasting \\
    \cite{art_370} & 2024 & Environment & Temperature & Forecasting \\
    \cite{art_371} & 2024 & Environment & Air quality & Forecasting \\
    \cite{art_373} & 2024 & Environment & Greenhouse gas concentrations & Forecasting \\
    \cite{art_692} & 2022 & Finance & Energy markets & Forecasting \\
    \cite{art_693} & 2022 & Finance & Stock prices & Classification \\
    \cite{art_59} & 2022 & Finance & Stock prediction & Classification \\
    \cite{art_64} & 2022 & Finance & Investment prediction & Forecasting \\
    \cite{art_66} & 2022 & Finance & Financial prediction & Classification \\
    \cite{art_60} & 2023 & Finance & Stock prediction & Classification \\
    \cite{art_62} & 2023 & Finance & Stock prediction & Classification \\
    \cite{art_63} & 2023 & Finance & Stock prediction & Classification \\
    \cite{art_65} & 2023 & Finance & Stock prediction & Classification \\
    \cite{art_67} & 2023 & Finance & Trading & Classification \\
    \cite{art_68} & 2023 & Finance & Sale prediction & Forecasting \\
    \cite{art_375} & 2024 & Finance & Supply chain demand & Forecasting \\
    \cite{art_378} & 2024 & Finance & Stock market volatility & Forecasting \\
    \cite{art_382} & 2024 & Finance & Stock trend & Classification \\
    \cite{art_712} & 2020 & Health & Epidemics & Forecasting \\
    \cite{art_659} & 2021 & Health & Covid 19 severity & Forecasting \\
    \cite{art_714} & 2021 & Health & Epidemics & Forecasting \\
    \cite{art_51} & 2021 & Health & Epidemic prediction & Forecasting \\
    \cite{art_658} & 2022 & Health & Epilepsy diagnosis & Classification \\
    \cite{art_662} & 2022 & Health & Epidemics & Forecasting \\
    \cite{art_713} & 2022 & Health & Epidemics & Forecasting \\
    \cite{art_50} & 2022 & Health & Epidemic prediction & Forecasting \\
    \cite{art_52} & 2022 & Health & Epidemic prediction & Forecasting \\
    \cite{art_660} & 2023 & Health & Epilepsy diagnosis & Classification \\
    \cite{art_661} & 2023 & Health & Sleep stage, emotion recognition & Classification \\
    \cite{art_14} & 2023 & Health & Epilepsy diagnosis & Classification \\
    \cite{art_48} & 2023 & Health & Epidemic prediction & Forecasting \\
    \cite{art_49} & 2023 & Health & Epidemic prediction & Forecasting \\
    \cite{art_92} & 2023 & Health & Disease diagnosis & Classification \\
    \cite{art_189} & 2023 & Health & Clinical risk classification & Classification \\
    \cite{art_410} & 2024 & Health & Infectious disease & Forecasting \\
    \cite{art_411} & 2024 & Health & EEG emotion recognition & Classification \\
    \cite{art_412} & 2024 & Health & EEG emotion recognition & Classification \\
    \cite{art_413} & 2024 & Health & Epidemic & Forecasting \\
    \cite{art_414} & 2024 & Health & Epidemic & Forecasting \\
    \cite{art_415} & 2024 & Health & Epidemic & Forecasting \\
    \cite{art_416} & 2024 & Health & Epidemic & Forecasting \\
    \cite{art_664} & 2020 & Mobility & E-Scooter distribution & Forecasting \\
    \cite{art_668} & 2020 & Mobility & Bike sharing & Forecasting \\
    \cite{art_669} & 2020 & Mobility & Urban traffic & Forecasting \\
    \cite{art_683} & 2020 & Mobility & Urban traffic & Forecasting \\
    \cite{art_726} & 2020 & Mobility & Urban traffic & Forecasting \\
    \cite{art_734} & 2020 & Mobility & Urban traffic & Forecasting \\
    \cite{art_736} & 2020 & Mobility & Urban traffic & Forecasting \\
    \cite{art_738} & 2020 & Mobility & Urban traffic & Forecasting \\
    \cite{art_739} & 2020 & Mobility & Metro flow & Forecasting \\
    \cite{art_105} & 2020 & Mobility & Urban traffic & Forecasting \\
    \cite{art_126} & 2020 & Mobility & Urban traffic & Forecasting \\
    \cite{art_663} & 2021 & Mobility & Urban traffic & Forecasting \\
    \cite{art_676} & 2021 & Mobility & Urban traffic & Forecasting \\
    \cite{art_724} & 2021 & Mobility & Urban traffic & Forecasting \\
    \cite{art_730} & 2021 & Mobility & Urban traffic & Forecasting \\
    \cite{art_94} & 2021 & Mobility & Crowd flow & Forecasting \\
    \cite{art_102} & 2021 & Mobility & Passenger demand & Forecasting \\
    \cite{art_652} & 2022 & Mobility & Urban traffic & Forecasting \\
    \cite{art_667} & 2022 & Mobility & Urban traffic & Forecasting \\
    \cite{art_672} & 2022 & Mobility & Urban traffic & Forecasting \\
    \cite{art_675} & 2022 & Mobility & Urban traffic & Forecasting \\
    \cite{art_682} & 2022 & Mobility & Urban traffic & Forecasting \\
    \cite{art_719} & 2022 & Mobility & Urban traffic & Forecasting \\
    \cite{art_720} & 2022 & Mobility & Urban traffic & Forecasting \\
    \cite{art_725} & 2022 & Mobility & Urban traffic & Forecasting \\
    \cite{art_729} & 2022 & Mobility & Urban traffic & Forecasting \\
    \cite{art_93} & 2022 & Mobility & Urban traffic & Forecasting \\
    \cite{art_96} & 2022 & Mobility & Crowd flow & Forecasting \\
    \cite{art_98} & 2022 & Mobility & Urban traffic & Forecasting \\
    \cite{art_104} & 2022 & Mobility & Connected internet of vehicles & Forecasting \\
    \cite{art_119} & 2022 & Mobility & Taxi demand & Forecasting \\
    \cite{art_120} & 2022 & Mobility & Urban traffic & Forecasting \\
    \cite{art_121} & 2022 & Mobility & Metro passengers & Forecasting \\
    \cite{art_122} & 2022 & Mobility & Urban traffic & Forecasting \\
    \cite{art_124} & 2022 & Mobility & Urban traffic & Forecasting \\
    \cite{art_128} & 2022 & Mobility & Urban traffic & Forecasting \\
    \cite{art_129} & 2022 & Mobility & Flight delay & Forecasting \\
    \cite{art_130} & 2022 & Mobility & Urban traffic & Forecasting \\
    \cite{art_132} & 2022 & Mobility & Urban traffic & Forecasting \\
    \cite{art_133} & 2022 & Mobility & Bicycle demand & Forecasting \\
    \cite{art_135} & 2022 & Mobility & Urban traffic & Forecasting \\
    \cite{art_670} & 2023 & Mobility & Urban traffic & Forecasting \\
    \cite{art_671} & 2023 & Mobility & Urban traffic & Forecasting \\
    \cite{art_673} & 2023 & Mobility & Urban traffic & Forecasting \\
    \cite{art_681} & 2023 & Mobility & Urban traffic & Forecasting \\
    \cite{art_707} & 2023 & Mobility & Urban traffic & Forecasting \\
    \cite{art_718} & 2023 & Mobility & Urban traffic & Forecasting \\
    \cite{art_722} & 2023 & Mobility & Urban traffic congestion & Forecasting \\
    \cite{art_728} & 2023 & Mobility & Visitation patterns & Forecasting \\
    \cite{art_731} & 2023 & Mobility & Urban traffic congestion & Classification \\
    \cite{art_733} & 2023 & Mobility & Urban traffic & Forecasting \\
    \cite{art_735} & 2023 & Mobility & Urban traffic & Forecasting \\
    \cite{art_740} & 2023 & Mobility & Urban traffic & Forecasting \\
    \cite{art_97} & 2023 & Mobility & Urban traffic & Forecasting \\
    \cite{art_99} & 2023 & Mobility & Urban traffic & Forecasting \\
    \cite{art_100} & 2023 & Mobility & Urban traffic & Forecasting \\
    \cite{art_103} & 2023 & Mobility & Urban traffic & Forecasting \\
    \cite{art_106} & 2023 & Mobility & Urban traffic & Forecasting \\
    \cite{art_108} & 2023 & Mobility & Urban traffic & Forecasting \\
    \cite{art_109} & 2023 & Mobility & Urban traffic & Forecasting \\
    \cite{art_110} & 2023 & Mobility & Urban traffic & Forecasting \\
    \cite{art_113} & 2023 & Mobility & Urban traffic & Forecasting \\
    \cite{art_114} & 2023 & Mobility & Bike sharing demand & Forecasting \\
    \cite{art_115} & 2023 & Mobility & Urban traffic & Forecasting \\
    \cite{art_116} & 2023 & Mobility & Urban traffic & Forecasting \\
    \cite{art_117} & 2023 & Mobility & Urban traffic & Forecasting \\
    \cite{art_125} & 2023 & Mobility & Urban traffic & Forecasting \\
    \cite{art_131} & 2023 & Mobility & Urban traffic & Forecasting \\
    \cite{art_653} & 2024 & Mobility & Urban traffic & Forecasting \\
    \cite{art_666} & 2024 & Mobility & Urban traffic & Forecasting \\
    \cite{art_694} & 2024 & Mobility & Urban traffic & Forecasting \\
    \cite{art_715} & 2024 & Mobility & Urban traffic & Forecasting \\
    \cite{art_716} & 2024 & Mobility & Urban traffic & Forecasting \\
    \cite{art_721} & 2024 & Mobility & Urban traffic & Forecasting \\
    \cite{art_723} & 2024 & Mobility & Urban traffic & Forecasting \\
    \cite{art_732} & 2024 & Mobility & Urban traffic & Forecasting \\
    \cite{art_406} & 2024 & Mobility & Urban traffic & Forecasting \\
    \cite{art_408} & 2024 & Mobility & Urban traffic & Forecasting \\
    \cite{art_419} & 2024 & Mobility & Urban traffic & Forecasting \\
    \cite{art_420} & 2024 & Mobility & Urban traffic & Forecasting \\
    \cite{art_421} & 2024 & Mobility & Metro flow traffic & Forecasting \\
    \cite{art_422} & 2024 & Mobility & Urban traffic & Forecasting \\
    \cite{art_424} & 2024 & Mobility & Urban traffic & Forecasting \\
    \cite{art_425} & 2024 & Mobility & Network traffic volume & Forecasting \\
    \cite{art_427} & 2024 & Mobility & Traffic speed & Forecasting \\
    \cite{art_430} & 2024 & Mobility & Traffic speed & Forecasting \\
    \cite{art_431} & 2024 & Mobility & Urban traffic & Forecasting \\
    \cite{art_432} & 2024 & Mobility & Urban traffic & Forecasting \\
    \cite{art_433} & 2024 & Mobility & Urban traffic & Forecasting \\
    \cite{art_434} & 2024 & Mobility & Airport performances & Forecasting \\
    \cite{art_435} & 2024 & Mobility & Urban traffic & Forecasting \\
    \cite{art_438} & 2024 & Mobility & Urban traffic & Forecasting \\
    \cite{art_441} & 2024 & Mobility & Urban traffic & Forecasting \\
    \cite{art_442} & 2024 & Mobility & Urban traffic & Forecasting \\
    \cite{art_443} & 2024 & Mobility & Flight delay prediction & Forecasting \\
    \cite{art_444} & 2024 & Mobility & Traffic speed & Forecasting \\
    \cite{art_445} & 2024 & Mobility & Urban traffic & Forecasting \\
    \cite{art_446} & 2024 & Mobility & Bike sharing demand & Forecasting \\
    \cite{art_447} & 2024 & Mobility & Urban traffic & Forecasting \\
    \cite{art_451} & 2024 & Mobility & Urban traffic & Forecasting \\
    \cite{art_453} & 2024 & Mobility & Urban traffic & Forecasting \\
    \cite{art_454} & 2024 & Mobility & Bus stop mobility pattern & Forecasting \\
    \cite{art_55} & 2021 & Predictive monitoring & Fault diagnosis & Classification \\
    \cite{art_155} & 2021 & Predictive monitoring & Remaining useful life & Forecasting \\
    \cite{art_648} & 2022 & Predictive monitoring & Anomaly detections & Forecasting \\
    \cite{art_9} & 2022 & Predictive monitoring & Anomaly detection & Classification \\
    \cite{art_156} & 2022 & Predictive monitoring & Remaining useful life) & Forecasting \\
    \cite{art_157} & 2022 & Predictive monitoring & Remaining useful life & Forecasting \\
    \cite{art_748} & 2023 & Predictive monitoring & Anomaly detection & Classification \\
    \cite{art_751} & 2023 & Predictive monitoring & Anomaly detection & Classification \\
    \cite{art_10} & 2023 & Predictive monitoring & Anomaly detection & Forecasting \\
    \cite{art_53} & 2023 & Predictive monitoring & Fault diagnosis & Classification \\
    \cite{art_54} & 2023 & Predictive monitoring & Fault diagnosis & Classification \\
    \cite{art_57} & 2023 & Predictive monitoring & Fault diagnosis & Classification \\
    \cite{art_58} & 2023 & Predictive monitoring & Fault prediction & Forecasting \\
    \cite{art_152} & 2023 & Predictive monitoring & Remaining useful life & Forecasting \\
    \cite{art_153} & 2023 & Predictive monitoring & Remaining useful life & Forecasting \\
    \cite{art_154} & 2023 & Predictive monitoring & Remaining useful life & Forecasting \\
    \cite{art_158} & 2023 & Predictive monitoring & Remaining useful life & Forecasting \\
    \cite{art_192} & 2023 & Predictive monitoring & Fault diagnosis & Classification \\
    \cite{art_690} & 2024 & Predictive monitoring & Photovoltaic degradation & Forecasting \\
    \cite{art_704} & 2024 & Predictive monitoring & Anomaly detection & Classification \\
    \cite{art_742} & 2024 & Predictive monitoring & Anomaly detection & Classification \\
    \cite{art_747} & 2024 & Predictive monitoring & Anomaly detection & Classification \\
    \cite{art_461} & 2024 & Predictive monitoring & Motor temperature & Forecasting \\
    \cite{art_469} & 2024 & Predictive monitoring & Fault diagnosis and detection & Classification \\
    \cite{art_470} & 2024 & Predictive monitoring & Root cause localization & Forecasting \\
    \cite{art_474} & 2024 & Predictive monitoring & Fault diagnosis & Classification \\
    \cite{art_475} & 2024 & Predictive monitoring & Outlier detection & Classification \\
    \cite{art_477} & 2024 & Predictive monitoring & Anomaly detection & Classification \\
    \cite{art_478} & 2024 & Predictive monitoring & Remaining useful life & Forecasting \\
    \cite{art_479} & 2024 & Predictive monitoring & Anomaly detection & Classification \\
    \cite{art_480} & 2024 & Predictive monitoring & Sensor data & Forecasting \\
    \cite{art_481} & 2024 & Predictive monitoring & Anomaly detection & Classification \\
    \cite{art_482} & 2024 & Predictive monitoring & Process factors & Forecasting \\
    \cite{art_483} & 2024 & Predictive monitoring & Anomaly detection & Classification \\
    \cite{art_484} & 2024 & Predictive monitoring & Anomaly detection & Classification \\
    \cite{art_486} & 2024 & Predictive monitoring & Fault diagnosis & Classification \\
    \cite{art_487} & 2024 & Predictive monitoring & Fault diagnosis & Classification \\
    \cite{art_488} & 2024 & Predictive monitoring & Remaining useful life & Forecasting \\
    \cite{art_489} & 2024 & Predictive monitoring & Anomaly detection & Classification \\
    \cite{art_491} & 2024 & Predictive monitoring & Fault diagnosis & Classification \\
    \cite{art_743} & 2021 & Other & Cellular traffic & Forecasting \\
    \cite{art_744} & 2021 & Other & Product demand & Forecasting \\
    \cite{art_123} & 2021 & Other & Cellular traffic & Forecasting \\
    \cite{art_148} & 2021 & Other & Cellular Network Traffic & Forecasting \\
    \cite{art_678} & 2022 & Other & Talent demand and supply & Forecasting \\
    \cite{art_13} & 2022 & Other & Modulation classification & Classification \\
    \cite{art_136} & 2022 & Other & Spatial temporal event   prediction & Forecasting \\
    \cite{art_147} & 2022 & Other & App popularity & Forecasting \\
    \cite{art_149} & 2022 & Other & Water demand & Forecasting \\
    \cite{art_679} & 2023 & Other & Encrypted traffic & Classification \\
    \cite{art_717} & 2023 & Other & Cellular traffic & Forecasting \\
    \cite{art_12} & 2023 & Other & Modulation classification & Classification \\
    \cite{art_118} & 2023 & Other & Encrypted traffic classification & Classification \\
    \cite{art_137} & 2023 & Other & Quality indicators prediction & Forecasting \\
    \cite{art_138} & 2023 & Other & Cellular traffic prediction & Forecasting \\
    \cite{art_139} & 2023 & Other & Seismic data & Forecasting \\
    \cite{art_140} & 2023 & Other & Structural dynamics  & Forecasting \\
    \cite{art_141} & 2023 & Other & Tailing dam monitoring & Forecasting \\
    \cite{art_143} & 2023 & Other & Subsurface production & Forecasting \\
    \cite{art_144} & 2023 & Other & Cellular Network Traffic & Forecasting \\
    \cite{art_145} & 2023 & Other & Data centers maintenance & Forecasting \\
    \cite{art_146} & 2023 & Other & Hydraulic runoff & Forecasting \\
    \cite{art_303} & 2023 & Other & Modulation classification & Classification \\
    \cite{art_746} & 2024 & Other & Anomaly in microservice system & Classification \\
    \cite{art_750} & 2024 & Other & Emergency supply & Forecasting \\
    \cite{art_455} & 2024 & Other & Polypropylene production & Forecasting \\
    \cite{art_456} & 2024 & Other & Botnet detection & Classification \\
    \cite{art_457} & 2024 & Other & Cellular KPI prediction & Forecasting \\
    \cite{art_458} & 2024 & Other & Tungsten floatation plant & Forecasting \\
    \cite{art_459} & 2024 & Other & Modulation recognition & Classification \\
    \cite{art_460} & 2024 & Other & IoT intrusion detection & Classification \\
    \cite{art_463} & 2024 & Other & Modulation recognition & Classification \\
    \cite{art_465} & 2024 & Other & Crime prediction & Forecasting \\
    \cite{art_468} & 2024 & Other & Mobile traffic prediction & Forecasting \\
    \cite{art_639} & 2020 & Generic & Multivariate time series & Forecasting \\
    \cite{art_654} & 2020 & Generic & Multivariate time series & Forecasting \\
    \cite{art_655} & 2020 & Generic & Multivariate time series & Forecasting \\
    \cite{art_684} & 2020 & Generic & Multivariate time series & Forecasting \\
    \cite{art_641} & 2021 & Generic & Multivariate time series & Forecasting \\
    \cite{art_656} & 2021 & Generic & Multivariate time series & Forecasting \\
    \cite{art_657} & 2021 & Generic & Multivariate time series & Forecasting \\
    \cite{art_696} & 2021 & Generic & Multivariate time series & Forecasting \\
    \cite{art_705} & 2021 & Generic & Multivariate time series & Forecasting \\
    \cite{art_159} & 2021 & Generic & Multivariate time series & Classification \\
    \cite{art_202} & 2021 & Generic & Generic & Both \\
    \cite{art_645} & 2022 & Generic & Multivariate time series & Forecasting \\
    \cite{art_649} & 2022 & Generic & Multivariate time series & Forecasting \\
    \cite{art_674} & 2022 & Generic & Time series & Forecasting \\
    \cite{art_685} & 2022 & Generic & Multivariate time series & Forecasting \\
    \cite{art_686} & 2022 & Generic & Multivariate time series & Forecasting \\
    \cite{art_702} & 2022 & Generic & Multivariate time series & Forecasting \\
    \cite{art_711} & 2022 & Generic & Multivariate time series & Classification \\
    \cite{art_84} & 2022 & Generic & Multivariate time series & Forecasting \\
    \cite{art_86} & 2022 & Generic & Multivariate time series & Forecasting \\
    \cite{art_87} & 2022 & Generic & Multivariate time series & Forecasting \\
    \cite{art_90} & 2022 & Generic & Multivariate time series & Forecasting \\
    \cite{art_647} & 2023 & Generic & Multivariate time series & Both \\
    \cite{art_687} & 2023 & Generic & Multivariate time series & Forecasting \\
    \cite{art_688} & 2023 & Generic & Multivariate time series & Forecasting \\
    \cite{art_689} & 2023 & Generic & Multivariate time series & Forecasting \\
    \cite{art_698} & 2023 & Generic & Multivariate time series & Forecasting \\
    \cite{art_700} & 2023 & Generic & Multivariate time series & Forecasting \\
    \cite{art_701} & 2023 & Generic & Multivariate time series & Forecasting \\
    \cite{art_709} & 2023 & Generic & Multivariate time series & Forecasting \\
    \cite{art_710} & 2023 & Generic & Multivariate time series  & Forecasting \\
    \cite{art_69} & 2023 & Generic & Batch workloads & Forecasting \\
    \cite{art_70} & 2023 & Generic & Node classification for websites & Classification \\
    \cite{art_71} & 2023 & Generic & Clinical risk classification & Classification \\
    \cite{art_72} & 2023 & Generic & Sensor network & Forecasting \\
    \cite{art_73} & 2023 & Generic & Multivariate time series & Forecasting \\
    \cite{art_74} & 2023 & Generic & Multivariate time series & Forecasting \\
    \cite{art_75} & 2023 & Generic & Multivariate time series & Forecasting \\
    \cite{art_76} & 2023 & Generic & Long term series forecasting & Forecasting \\
    \cite{art_77} & 2023 & Generic & Multivariate time series & Forecasting \\
    \cite{art_78} & 2023 & Generic & Multivariate time series & Forecasting \\
    \cite{art_79} & 2023 & Generic & Multivariate time series & Forecasting \\
    \cite{art_80} & 2023 & Generic & Multivariate time series & Forecasting \\
    \cite{art_81} & 2023 & Generic & Multivariate time series & Forecasting \\
    \cite{art_82} & 2023 & Generic & Multivariate time series & Forecasting \\
    \cite{art_83} & 2023 & Generic & Multivariate time series & Forecasting \\
    \cite{art_85} & 2023 & Generic & Multivariate time series & Forecasting \\
    \cite{art_88} & 2023 & Generic & Multivariate time series & Forecasting \\
    \cite{art_89} & 2023 & Generic & Multivariate time series & Forecasting \\
    \cite{art_160} & 2023 & Generic & Impact of graph construction & Forecasting \\
    \cite{art_640} & 2024 & Generic & Multivariate time series & Forecasting \\
    \cite{art_642} & 2024 & Generic & Multivariate time series & Forecasting \\
    \cite{art_643} & 2024 & Generic & Multivariate time series & Forecasting \\
    \cite{art_644} & 2024 & Generic & Multivariate time series & Forecasting \\
    \cite{art_646} & 2024 & Generic & Multivariate time series & Forecasting  \\
    \cite{art_650} & 2024 & Generic & Multivariate time series & Forecasting \\
    \cite{art_680} & 2024 & Generic & Multivariate time series & Classification \\
    \cite{art_695} & 2024 & Generic & Multivariate time series & Forecasting \\
    \cite{art_703} & 2024 & Generic & Multivariate time series & Forecasting \\
    \cite{art_383} & 2024 & Generic & Multivariate time series & Forecasting \\
    \cite{art_385} & 2024 & Generic & Multivariate time series & Classification \\
    \cite{art_386} & 2024 & Generic & Multivariate time series & Classification \\
    \cite{art_387} & 2024 & Generic & Multivariate time series & Both \\
    \cite{art_388} & 2024 & Generic & Multivariate time series & Forecasting \\
    \cite{art_389} & 2024 & Generic & Multivariate time series & Classification \\
    \cite{art_393} & 2024 & Generic & Multivariate time series & Forecasting \\
    \cite{art_395} & 2024 & Generic & Multivariate time series & Classification \\
    \cite{art_396} & 2024 & Generic & Multivariate time series & Forecasting \\
    \cite{art_397} & 2024 & Generic & Multivariate time series & Forecasting \\
    \cite{art_398} & 2024 & Generic & Multivariate time series & Forecasting \\
    \cite{art_400} & 2024 & Generic & Multivariate time series & Forecasting \\
    \cite{art_401} & 2024 & Generic & Multivariate time series & Forecasting \\
    \cite{art_403} & 2024 & Generic & Multivariate time series & Both \\
    \cite{art_404} & 2024 & Generic & Multivariate time series & Forecasting \\
    \cite{art_405} & 2024 & Generic & Multivariate time series & Forecasting \\
    \cite{art_407} & 2024 & Generic & Multivariate time series & Classification \\
    \cite{art_417} & 2024 & Generic & Multivariate time series & Forecasting \\
    \bottomrule
\end{longtable}
}


\bibliographystyle{bababbrv-fl}
\nocite{*}
\bibliography{references}

\begin{thebibliography}{100}
  \providebibliographyfont{name}{}%
  \providebibliographyfont{lastname}{}%
  \providebibliographyfont{title}{\emph}%
  \providebibliographyfont{jtitle}{\btxtitlefont}%
  \providebibliographyfont{etal}{\emph}%
  \providebibliographyfont{journal}{}%
  \providebibliographyfont{volume}{}%
  \providebibliographyfont{ISBN}{\MakeUppercase}%
  \providebibliographyfont{ISSN}{\MakeUppercase}%
  \providebibliographyfont{url}{\url}%
  \providebibliographyfont{numeral}{}%
  \expandafter\btxselectlanguage\expandafter {\btxfallbacklanguage}

\expandafter\btxselectlanguage\expandafter {\btxfallbacklanguage}
\bibitem {tensorflow2015-whitepaper}
\btxnamefont {M.~\btxlastnamefont {Abadi}}, \btxnamefont {A.~\btxlastnamefont
  {Agarwal}}, \btxnamefont {P.~\btxlastnamefont {Barham}}, \btxnamefont
  {E.~\btxlastnamefont {Brevdo}}, \btxnamefont {Z.~\btxlastnamefont {Chen}},
  \btxnamefont {C.~\btxlastnamefont {Citro}}, \btxnamefont
  {G.\btxfnamespaceshort S. \btxlastnamefont {Corrado}}, \btxnamefont
  {A.~\btxlastnamefont {Davis}}, \btxnamefont {J.~\btxlastnamefont {Dean}},
  \btxnamefont {M.~\btxlastnamefont {Devin}}, \btxnamefont {S.~\btxlastnamefont
  {Ghemawat}}, \btxnamefont {I.~\btxlastnamefont {Goodfellow}}, \btxnamefont
  {A.~\btxlastnamefont {Harp}}, \btxnamefont {G.~\btxlastnamefont {Irving}},
  \btxnamefont {M.~\btxlastnamefont {Isard}}, \btxnamefont {Y.~\btxlastnamefont
  {Jia}}, \btxnamefont {R.~\btxlastnamefont {Jozefowicz}}, \btxnamefont
  {L.~\btxlastnamefont {Kaiser}}, \btxnamefont {M.~\btxlastnamefont {Kudlur}},
  \btxnamefont {J.~\btxlastnamefont {Levenberg}}, \btxnamefont
  {D.~\btxlastnamefont {Man\'{e}}}, \btxnamefont {R.~\btxlastnamefont {Monga}},
  \btxnamefont {S.~\btxlastnamefont {Moore}}, \btxnamefont {D.~\btxlastnamefont
  {Murray}}, \btxnamefont {C.~\btxlastnamefont {Olah}}, \btxnamefont
  {M.~\btxlastnamefont {Schuster}}, \btxnamefont {J.~\btxlastnamefont
  {Shlens}}, \btxnamefont {B.~\btxlastnamefont {Steiner}}, \btxnamefont
  {I.~\btxlastnamefont {Sutskever}}, \btxnamefont {K.~\btxlastnamefont
  {Talwar}}, \btxnamefont {P.~\btxlastnamefont {Tucker}}, \btxnamefont
  {V.~\btxlastnamefont {Vanhoucke}}, \btxnamefont {V.~\btxlastnamefont
  {Vasudevan}}, \btxnamefont {F.~\btxlastnamefont {Vi\'{e}gas}}, \btxnamefont
  {O.~\btxlastnamefont {Vinyals}}, \btxnamefont {P.~\btxlastnamefont {Warden}},
  \btxnamefont {M.~\btxlastnamefont {Wattenberg}}, \btxnamefont
  {M.~\btxlastnamefont {Wicke}}, \btxnamefont {Y.~\btxlastnamefont
  {Yu}}\btxandcomma {} \btxandshort {.}\ \btxnamefont {X.~\btxlastnamefont
  {Zheng}}\btxauthorcolon\ \btxtitlefont {\btxifchangecase {{TensorFlow}:
  Large-scale machine learning on heterogeneous systems}{{TensorFlow}:
  Large-Scale Machine Learning on Heterogeneous Systems}}, 2015.
\newblock {\latintext \btxurlfont{https://www.tensorflow.org/}}, Software
  available from tensorflow.org.

\bibitem {art_456}
\btxnamefont {T.~\btxlastnamefont {Altaf}}, \btxnamefont {X.~\btxlastnamefont
  {Wang}}, \btxnamefont {W.~\btxlastnamefont {Ni}}, \btxnamefont
  {G.~\btxlastnamefont {Yu}}, \btxnamefont {R.\btxfnamespaceshort P.
  \btxlastnamefont {Liu}}\btxandcomma {} \btxandshort {.}\ \btxnamefont
  {R.~\btxlastnamefont {Braun}}\btxauthorcolon\ \btxjtitlefont
  {\btxifchangecase {{GNN-based} network traffic analysis for the detection of
  sequential attacks in {IoT}}{{GNN-based} network traffic analysis for the
  detection of sequential attacks in {IoT}}}.
\newblock \btxjournalfont {Electronics (Basel)}, 13(12):2274,
  \btxprintmonthyear{.}{6}{2024}{short}.

\bibitem {USAD}
\btxnamefont {J.~\btxlastnamefont {Audibert}}, \btxnamefont
  {P.~\btxlastnamefont {Michiardi}}, \btxnamefont {F.~\btxlastnamefont
  {Guyard}}, \btxnamefont {S.~\btxlastnamefont {Marti}}\btxandcomma {}
  \btxandshort {.}\ \btxnamefont {M.\btxfnamespaceshort A. \btxlastnamefont
  {Zuluaga}}\btxauthorcolon\ \btxtitlefont {\btxifchangecase {Usad:
  Unsupervised anomaly detection on multivariate time series}{USAD:
  UnSupervised Anomaly Detection on Multivariate Time Series}}.
\newblock \Btxinshort {.}\ \btxtitlefont {Proceedings of the 26th ACM SIGKDD
  International Conference on Knowledge Discovery \& Data Mining}, KDD '20,
  \btxpageshort {.}\ 3395–3404, New York, NY, USA, 2020. \btxpublisherfont
  {Association for Computing Machinery}\ifbtxprintISBN {,
  \mbox{\btxISBN~\btxISBNfont {9781450379984}}}.
\newblock {\latintext \btxurlfont{https://doi.org/10.1145/3394486.3403392}}.

\bibitem {art_637}
\btxnamefont {M.~\btxlastnamefont {Azmat}}, \btxnamefont {M.~\btxlastnamefont
  {Madondo}}, \btxnamefont {A.~\btxlastnamefont {Bawa}}, \btxnamefont
  {K.~\btxlastnamefont {Dipietro}}, \btxnamefont {R.~\btxlastnamefont
  {Horesh}}, \btxnamefont {M.~\btxlastnamefont {Jacobs}}, \btxnamefont
  {R.~\btxlastnamefont {Srinivasan}}\btxandcomma {} \btxandshort {.}\
  \btxnamefont {F.~\btxlastnamefont {O'Donncha}}\btxauthorcolon\ \btxtitlefont
  {\btxifchangecase {Forecasting soil moisture using domain inspired temporal
  graph convolution neural networks to guide sustainable crop
  management}{Forecasting soil moisture using domain inspired temporal graph
  convolution neural networks to guide sustainable crop management}}.
\newblock \Btxinshort {.}\ \btxtitlefont {Proceedings of the {Thirty-Second}
  International Joint Conference on Artificial Intelligence}, \btxpagesshort
  {.}\ 5897--5905, California, \btxprintmonthyear{.}{8}{2023}{short}.
  \btxpublisherfont {International Joint Conferences on Artificial Intelligence
  Organization}.

\bibitem {art_303}
\btxnamefont {H.~\btxlastnamefont {Bai}}, \btxnamefont {J.~\btxlastnamefont
  {Yang}}, \btxnamefont {M.~\btxlastnamefont {Huang}}\btxandcomma {}
  \btxandshort {.}\ \btxnamefont {W.~\btxlastnamefont {Li}}\btxauthorcolon\
  \btxjtitlefont {\btxifchangecase {A symmetric adaptive visibility graph
  classification method of orthogonal signals for automatic modulation
  classification}{A symmetric adaptive visibility graph classification method
  of orthogonal signals for automatic modulation classification}}.
\newblock \btxjournalfont {IET Communications}, 17(10):1208–1219,
  \btxprintmonthyear{.}{4}{2023}{short}\ifbtxprintISSN {,
  \mbox{\btxISSN~\btxISSNfont {1751-8636}}}.
\newblock {\latintext \btxurlfont{http://dx.doi.org/10.1049/cmu2.12608}}.

\bibitem {STG2Seq}
\btxnamefont {L.~\btxlastnamefont {Bai}}, \btxnamefont {L.~\btxlastnamefont
  {Yao}}, \btxnamefont {S.\btxfnamespaceshort S. \btxlastnamefont {Kanhere}},
  \btxnamefont {X.~\btxlastnamefont {Wang}}\btxandcomma {} \btxandshort {.}\
  \btxnamefont {Q.\btxfnamespaceshort Z. \btxlastnamefont
  {Sheng}}\btxauthorcolon\ \btxtitlefont {\btxifchangecase {Stg2seq:
  spatial-temporal graph to sequence model for multi-step passenger demand
  forecasting}{STG2seq: spatial-temporal graph to sequence model for multi-step
  passenger demand forecasting}}.
\newblock \Btxinshort {.}\ \btxtitlefont {Proceedings of the 28th International
  Joint Conference on Artificial Intelligence}, IJCAI'19, \btxpageshort {.}\
  1981–1987. \btxpublisherfont {AAAI Press}, 2019\ifbtxprintISBN {,
  \mbox{\btxISBN~\btxISBNfont {9780999241141}}}.

\bibitem {art_683}
\btxnamefont {L.~\btxlastnamefont {Bai}}, \btxnamefont {L.~\btxlastnamefont
  {Yao}}, \btxnamefont {C.~\btxlastnamefont {Li}}, \btxnamefont
  {X.~\btxlastnamefont {Wang}}\btxandcomma {} \btxandshort {.}\ \btxnamefont
  {C.~\btxlastnamefont {Wang}}\btxauthorcolon\ \btxtitlefont {\btxifchangecase
  {Adaptive graph convolutional recurrent network for traffic
  forecasting}{Adaptive graph convolutional recurrent network for traffic
  forecasting}}.
\newblock \Btxinshort {.}\ \btxtitlefont {Proceedings of the 34th International
  Conference on Neural Information Processing Systems}, NIPS '20, Red Hook, NY,
  USA, 2020. \btxpublisherfont {Curran Associates Inc.}\ifbtxprintISBN {,
  \mbox{\btxISBN~\btxISBNfont {9781713829546}}}.

\bibitem {TCN}
\btxnamefont {S.~\btxlastnamefont {Bai}}, \btxnamefont {J.\btxfnamespaceshort
  Z. \btxlastnamefont {Kolter}}\btxandcomma {} \btxandshort {.}\ \btxnamefont
  {V.~\btxlastnamefont {Koltun}}\btxauthorcolon\ \btxjtitlefont
  {\btxifchangecase {An empirical evaluation of generic convolutional and
  recurrent networks for sequence modeling}{An Empirical Evaluation of Generic
  Convolutional and Recurrent Networks for Sequence Modeling}}.
\newblock \btxjournalfont {ArXiv}, abs/1803.01271, 2018.
\newblock {\latintext
  \btxurlfont{https://api.semanticscholar.org/CorpusID:4747877}}.

\bibitem {art_43}
\btxnamefont {T.~\btxlastnamefont {Bai}} \btxandshort {.}\ \btxnamefont
  {P.~\btxlastnamefont {Tahmasebi}}\btxauthorcolon\ \btxjtitlefont
  {\btxifchangecase {Graph neural network for groundwater level
  forecasting}{Graph Neural Network for Groundwater Level Forecasting}}.
\newblock \btxjournalfont {Journal of Hydrology}, 616:128792,
  \btxprintmonthyear{.}{11}{2022}{short}.

\bibitem {art_29}
\btxnamefont {L.\btxfnamespaceshort {\O}. \btxlastnamefont {Bentsen}},
  \btxnamefont {N.\btxfnamespaceshort D. \btxlastnamefont {Warakagoda}},
  \btxnamefont {R.~\btxlastnamefont {Stenbro}}\btxandcomma {} \btxandshort {.}\
  \btxnamefont {P.~\btxlastnamefont {Engelstad}}\btxauthorcolon\ \btxjtitlefont
  {\btxifchangecase {Spatio-temporal wind speed forecasting using graph
  networks and novel transformer architectures}{Spatio-temporal wind speed
  forecasting using graph networks and novel Transformer architectures}}.
\newblock \btxjournalfont {Applied Energy}, 333, 2023.
\newblock {\latintext
  \btxurlfont{https://www.scopus.com/inward/record.uri?eid=2-s2.0-85144917861&doi=10.1016

\bibitem {art_34}
\btxnamefont {L.\btxfnamespaceshort {\O}. \btxlastnamefont {Bentsen}},
  \btxnamefont {N.\btxfnamespaceshort D. \btxlastnamefont {Warakagoda}},
  \btxnamefont {R.~\btxlastnamefont {Stenbro}}\btxandcomma {} \btxandshort {.}\
  \btxnamefont {P.~\btxlastnamefont {Engelstad}}\btxauthorcolon\ \btxjtitlefont
  {\btxifchangecase {A unified graph formulation for spatio-temporal wind
  forecasting}{A Unified Graph Formulation for Spatio-Temporal Wind
  Forecasting}}.
\newblock \btxjournalfont {Energies}, 16(20), 2023.
\newblock {\latintext
  \btxurlfont{https://www.scopus.com/inward/record.uri?eid=2-s2.0-85175042962&doi=10.3390

\bibitem {art_427}
\btxnamefont {P.~\btxlastnamefont {Bikram}}, \btxnamefont {S.~\btxlastnamefont
  {Das}}\btxandcomma {} \btxandshort {.}\ \btxnamefont {A.~\btxlastnamefont
  {Biswas}}\btxauthorcolon\ \btxjtitlefont {\btxifchangecase {Dynamic attention
  aggregated missing spatial--temporal data imputation for traffic speed
  prediction}{Dynamic attention aggregated missing spatial--temporal data
  imputation for traffic speed prediction}}.
\newblock \btxjournalfont {Neurocomputing}, 607(128441):128441,
  \btxprintmonthyear{.}{11}{2024}{short}.

\bibitem {cit_rev1}
\btxnamefont {R.~\btxlastnamefont {Bing}}, \btxnamefont {G.~\btxlastnamefont
  {Yuan}}, \btxnamefont {M.~\btxlastnamefont {Zhu}}, \btxnamefont
  {F.~\btxlastnamefont {Meng}}, \btxnamefont {H.~\btxlastnamefont
  {Ma}}\btxandcomma {} \btxandshort {.}\ \btxnamefont {S.~\btxlastnamefont
  {Qiao}}\btxauthorcolon\ \btxjtitlefont {\btxifchangecase {Heterogeneous graph
  neural networks analysis: a survey of techniques, evaluations and
  applications}{Heterogeneous graph neural networks analysis: a survey of
  techniques, evaluations and applications}}.
\newblock \btxjournalfont {Artif. Intell. Rev.}, 56(8):8003–8042,
  \btxprintmonthyear{.}{dec}{2022}{short}\ifbtxprintISSN {,
  \mbox{\btxISSN~\btxISSNfont {0269-2821}}}.
\newblock {\latintext \btxurlfont{https://doi.org/10.1007/s10462-022-10375-2}}.

\bibitem {art_160}
\btxnamefont {S.~\btxlastnamefont {Bloemheuvel}}, \btxnamefont {J.~van~den
  \btxlastnamefont {Hoogen}}\btxandcomma {} \btxandshort {.}\ \btxnamefont
  {M.~\btxlastnamefont {Atzmueller}}\btxauthorcolon\ \btxjtitlefont
  {\btxifchangecase {Graph construction on complex spatiotemporal data for
  enhancing graph neural network-based approaches}{Graph construction on
  complex spatiotemporal data for enhancing graph neural network-based
  approaches}}.
\newblock \btxjournalfont {International Journal of Data Science and
  Analytics}, 2023.
\newblock {\latintext
  \btxurlfont{https://www.scopus.com/inward/record.uri?eid=2-s2.0-85172328971&doi=10.1007

\bibitem {art_139}
\btxnamefont {S.~\btxlastnamefont {Bloemheuvel}}, \btxnamefont {J.~van~den
  \btxlastnamefont {Hoogen}}, \btxnamefont {D.~\btxlastnamefont {Jozinović}},
  \btxnamefont {A.~\btxlastnamefont {Michelini}}\btxandcomma {} \btxandshort
  {.}\ \btxnamefont {M.~\btxlastnamefont {Atzmueller}}\btxauthorcolon\
  \btxjtitlefont {\btxifchangecase {Graph neural networks for multivariate time
  series regression with application to seismic data}{Graph neural networks for
  multivariate time series regression with application to seismic data}}.
\newblock \btxjournalfont {International Journal of Data Science and
  Analytics}, 16(3):317 – 332, 2023.
\newblock {\latintext
  \btxurlfont{https://www.scopus.com/inward/record.uri?eid=2-s2.0-85137122128&doi=10.1007

\bibitem {art_373}
\btxnamefont {V.~\btxlastnamefont {Bobakov}}, \btxnamefont {S.~\btxlastnamefont
  {Kuzmin}}, \btxnamefont {A.~\btxlastnamefont {Butorova}}\btxandcomma {}
  \btxandshort {.}\ \btxnamefont {A.~\btxlastnamefont
  {Sergeev}}\btxauthorcolon\ \btxjtitlefont {\btxifchangecase {Application of
  graph-structured data for forecasting the dynamics of time series of natural
  origin}{Application of graph-structured data for forecasting the dynamics of
  time series of natural origin}}.
\newblock \btxjournalfont {Eur. Phys. J. Spec. Top.},
  \btxprintmonthyear{.}{10}{2024}{short}.

\bibitem {art_345}
\btxnamefont {T.~\btxlastnamefont {Boussaid}}, \btxnamefont
  {F.~\btxlastnamefont {Rousset}}, \btxnamefont {V.\btxfnamespaceshort M.
  \btxlastnamefont {Scuturici}}\btxandcomma {} \btxandshort {.}\ \btxnamefont
  {M.~\btxlastnamefont {Clausse}}\btxauthorcolon\ \btxjtitlefont
  {\btxifchangecase {Enabling fast prediction of district heating networks
  transients via a physics-guided graph neural network}{Enabling fast
  prediction of district heating networks transients via a physics-guided graph
  neural network}}.
\newblock \btxjournalfont {Appl. Energy}, 370(123634):123634,
  \btxprintmonthyear{.}{9}{2024}{short}.

\bibitem {LOF}
\btxnamefont {M.\btxfnamespaceshort M. \btxlastnamefont {Breunig}},
  \btxnamefont {H.\btxfnamespaceshort P. \btxlastnamefont {Kriegel}},
  \btxnamefont {R.\btxfnamespaceshort T. \btxlastnamefont {Ng}}\btxandcomma {}
  \btxandshort {.}\ \btxnamefont {J.~\btxlastnamefont {Sander}}\btxauthorcolon\
  \btxjtitlefont {\btxifchangecase {Lof: identifying density-based local
  outliers}{LOF: identifying density-based local outliers}}.
\newblock \btxjournalfont {SIGMOD Rec.}, 29(2):93–104,
  \btxprintmonthyear{.}{5}{2000}{short}\ifbtxprintISSN {,
  \mbox{\btxISSN~\btxISSNfont {0163-5808}}}.
\newblock {\latintext \btxurlfont{https://doi.org/10.1145/335191.335388}}.

\bibitem {Brody2021HowAA}
\btxnamefont {S.~\btxlastnamefont {Brody}}, \btxnamefont {U.~\btxlastnamefont
  {Alon}}\btxandcomma {} \btxandshort {.}\ \btxnamefont {E.~\btxlastnamefont
  {Yahav}}\btxauthorcolon\ \btxtitlefont {\btxifchangecase {How attentive are
  graph attention networks?}{How Attentive are Graph Attention Networks?}}
\newblock \Btxinshort {.}\ \btxtitlefont {International Conference on Learning
  Representations}, 2022.
\newblock {\latintext
  \btxurlfont{https://openreview.net/forum?id=F72ximsx7C1}}.

\bibitem {art_483}
\btxnamefont {K.~\btxlastnamefont {Buchhorn}}, \btxnamefont
  {E.~\btxlastnamefont {Santos-Fernandez}}, \btxnamefont {K.~\btxlastnamefont
  {Mengersen}}\btxandcomma {} \btxandshort {.}\ \btxnamefont
  {R.~\btxlastnamefont {Salomone}}\btxauthorcolon\ \btxjtitlefont
  {\btxifchangecase {Graph neural network-based anomaly detection for river
  network systems}{Graph neural network-based anomaly detection for river
  network systems}}.
\newblock \btxjournalfont {F1000Res.}, 12:991, 2023.

\bibitem {art_128}
\btxnamefont {K.\btxfnamespaceshort H.\btxfnamespaceshort N. \btxlastnamefont
  {Bui}}, \btxnamefont {J.~\btxlastnamefont {Cho}}\btxandcomma {} \btxandshort
  {.}\ \btxnamefont {H.~\btxlastnamefont {Yi}}\btxauthorcolon\ \btxjtitlefont
  {\btxifchangecase {Spatial-temporal graph neural network for traffic
  forecasting: An overview and open research issues}{Spatial-temporal graph
  neural network for traffic forecasting: An overview and open research
  issues}}.
\newblock \btxjournalfont {Applied Intelligence}, 52(3):2763 – 2774, 2022.
\newblock {\latintext
  \btxurlfont{https://www.scopus.com/inward/record.uri?eid=2-s2.0-85124804317&doi=10.1007

\bibitem {art_660}
\btxnamefont {D.~\btxlastnamefont {Cai}}, \btxnamefont {J.~\btxlastnamefont
  {Chen}}, \btxnamefont {Y.~\btxlastnamefont {Yang}}, \btxnamefont
  {T.~\btxlastnamefont {Liu}}\btxandcomma {} \btxandshort {.}\ \btxnamefont
  {Y.~\btxlastnamefont {Li}}\btxauthorcolon\ \btxtitlefont {\btxifchangecase
  {{MBrain}: A multi-channel self-supervised learning framework for brain
  signals}{{MBrain}: A multi-channel self-supervised learning framework for
  brain signals}}.
\newblock \Btxinshort {.}\ \btxtitlefont {Proceedings of the 29th {ACM}
  {SIGKDD} Conference on Knowledge Discovery and Data Mining}, \btxpagesshort
  {.}\ 130--141, New York, NY, USA, \btxprintmonthyear{.}{8}{2023}{short}.
  \btxpublisherfont {ACM}.

\bibitem {art_347}
\btxnamefont {J.~\btxlastnamefont {Cai}}, \btxnamefont {H.~\btxlastnamefont
  {Wang}}\btxandcomma {} \btxandshort {.}\ \btxnamefont {H.~\btxlastnamefont
  {Jiang}}\btxauthorcolon\ \btxjtitlefont {\btxifchangecase {A temporal
  ensembling based semi-supervised graph convolutional network for power
  quality disturbances classification}{A temporal ensembling based
  semi-supervised graph convolutional network for power quality disturbances
  classification}}.
\newblock \btxjournalfont {IEEE Access}, 12:75249--75261, 2024.

\bibitem {art_129}
\btxnamefont {K.~\btxlastnamefont {Cai}}, \btxnamefont {Y.~\btxlastnamefont
  {Li}}, \btxnamefont {Y.\btxfnamespaceshort P. \btxlastnamefont
  {Fang}}\btxandcomma {} \btxandshort {.}\ \btxnamefont {Y.~\btxlastnamefont
  {Zhu}}\btxauthorcolon\ \btxjtitlefont {\btxifchangecase {A deep learning
  approach for flight delay prediction through time-evolving graphs}{A Deep
  Learning Approach for Flight Delay Prediction Through Time-Evolving Graphs}}.
\newblock \btxjournalfont {IEEE Transactions on Intelligent Transportation
  Systems}, 23(8):11397 – 11407, 2022.
\newblock {\latintext
  \btxurlfont{https://www.scopus.com/inward/record.uri?eid=2-s2.0-85136088717&doi=10.1109

\bibitem {art_684}
\btxnamefont {D.~\btxlastnamefont {Cao}}, \btxnamefont {Y.~\btxlastnamefont
  {Wang}}, \btxnamefont {J.~\btxlastnamefont {Duan}}, \btxnamefont
  {C.~\btxlastnamefont {Zhang}}, \btxnamefont {X.~\btxlastnamefont {Zhu}},
  \btxnamefont {C.~\btxlastnamefont {Huang}}, \btxnamefont {Y.~\btxlastnamefont
  {Tong}}, \btxnamefont {B.~\btxlastnamefont {Xu}}, \btxnamefont
  {J.~\btxlastnamefont {Bai}}, \btxnamefont {J.~\btxlastnamefont
  {Tong}}\btxandcomma {} \btxandshort {.}\ \btxnamefont {Q.~\btxlastnamefont
  {Zhang}}\btxauthorcolon\ \btxtitlefont {\btxifchangecase {Spectral temporal
  graph neural network for multivariate time-series forecasting}{Spectral
  temporal graph neural network for multivariate time-series forecasting}}.
\newblock \Btxinshort {.}\ \btxtitlefont {Proceedings of the 34th International
  Conference on Neural Information Processing Systems}, NIPS '20, Red Hook, NY,
  USA, 2020. \btxpublisherfont {Curran Associates Inc.}\ifbtxprintISBN {,
  \mbox{\btxISBN~\btxISBNfont {9781713829546}}}.

\bibitem {art_132}
\btxnamefont {Y.~\btxlastnamefont {Cao}}, \btxnamefont {D.~\btxlastnamefont
  {Liu}}, \btxnamefont {Q.~\btxlastnamefont {Yin}}, \btxnamefont
  {F.~\btxlastnamefont {Xue}}\btxandcomma {} \btxandshort {.}\ \btxnamefont
  {H.~\btxlastnamefont {Tang}}\btxauthorcolon\ \btxjtitlefont {\btxifchangecase
  {Msasgcn: Multi-head self-attention spatiotemporal graph convolutional
  network for traffic flow forecasting}{MSASGCN: Multi-Head Self-Attention
  Spatiotemporal Graph Convolutional Network for Traffic Flow Forecasting}}.
\newblock \btxjournalfont {Journal of Advanced Transportation}, 2022, 2022.
\newblock {\latintext
  \btxurlfont{https://www.scopus.com/inward/record.uri?eid=2-s2.0-85133003671&doi=10.1155

\bibitem {CAPONE2025130400}
\btxnamefont {V.~\btxlastnamefont {Capone}}, \btxnamefont {A.~\btxlastnamefont
  {Casolaro}}\btxandcomma {} \btxandshort {.}\ \btxnamefont
  {F.~\btxlastnamefont {Camastra}}\btxauthorcolon\ \btxjtitlefont
  {\btxifchangecase {Spatio-temporal prediction using graph neural networks: A
  survey}{Spatio-temporal prediction using graph neural networks: A survey}}.
\newblock \btxjournalfont {Neurocomputing}, 643:130400, 2025\ifbtxprintISSN {,
  \mbox{\btxISSN~\btxISSNfont {0925-2312}}}.
\newblock {\latintext
  \btxurlfont{https://www.sciencedirect.com/science/article/pii/S0925231225010720}}.

\bibitem {ST-AR}
\btxnamefont {R.\btxfnamespaceshort E. \btxlastnamefont {Carrillo}},
  \btxnamefont {M.~\btxlastnamefont {Leblanc}}, \btxnamefont
  {B.~\btxlastnamefont {Schubnel}}, \btxnamefont {R.~\btxlastnamefont
  {Langou}}, \btxnamefont {C.~\btxlastnamefont {Topfel}}\btxandcomma {}
  \btxandshort {.}\ \btxnamefont {P.\btxfnamespaceshort J. \btxlastnamefont
  {Alet}}\btxauthorcolon\ \btxjtitlefont {\btxifchangecase {High-resolution pv
  forecasting from imperfect data: A graph-based solution}{High-Resolution PV
  Forecasting from Imperfect Data: A Graph-Based Solution}}.
\newblock \btxjournalfont {Energies}, 13(21), 2020\ifbtxprintISSN {,
  \mbox{\btxISSN~\btxISSNfont {1996-1073}}}.
\newblock {\latintext \btxurlfont{https://www.mdpi.com/1996-1073/13/21/5763}}.

\bibitem {MGCN}
\btxnamefont {D.~\btxlastnamefont {Chai}}, \btxnamefont {L.~\btxlastnamefont
  {Wang}}\btxandcomma {} \btxandshort {.}\ \btxnamefont {Q.~\btxlastnamefont
  {Yang}}\btxauthorcolon\ \btxtitlefont {\btxifchangecase {Bike flow prediction
  with multi-graph convolutional networks}{Bike flow prediction with
  multi-graph convolutional networks}}.
\newblock \Btxinshort {.}\ \btxtitlefont {Proceedings of the 26th ACM
  SIGSPATIAL International Conference on Advances in Geographic Information
  Systems}, SIGSPATIAL '18, \btxpageshort {.}\ 397–400, New York, NY, USA,
  2018. \btxpublisherfont {Association for Computing Machinery}\ifbtxprintISBN
  {, \mbox{\btxISBN~\btxISBNfont {9781450358897}}}.
\newblock {\latintext \btxurlfont{https://doi.org/10.1145/3274895.3274896}}.

\bibitem {art_461}
\btxnamefont {C.~\btxlastnamefont {Chen}}, \btxnamefont {Y.~\btxlastnamefont
  {Yuan}}, \btxnamefont {W.~\btxlastnamefont {Sun}}\btxandcomma {} \btxandshort
  {.}\ \btxnamefont {F.~\btxlastnamefont {Zhao}}\btxauthorcolon\ \btxjtitlefont
  {\btxifchangecase {Multivariate multi-step time series prediction of
  induction motor situation based on fused temporal-spatial
  features}{Multivariate multi-step time series prediction of induction motor
  situation based on fused temporal-spatial features}}.
\newblock \btxjournalfont {Int. J. Hydrogen Energy}, 50:1386--1394,
  \btxprintmonthyear{.}{1}{2024}{short}.

\bibitem {art_408}
\btxnamefont {D.~\btxlastnamefont {Chen}}, \btxnamefont {L.~\btxlastnamefont
  {Chen}}, \btxnamefont {Z.~\btxlastnamefont {Shang}}, \btxnamefont
  {Y.~\btxlastnamefont {Zhang}}, \btxnamefont {B.~\btxlastnamefont
  {Wen}}\btxandcomma {} \btxandshort {.}\ \btxnamefont {C.~\btxlastnamefont
  {Yang}}\btxauthorcolon\ \btxjtitlefont {\btxifchangecase {Scale-aware neural
  architecture search for multivariate time series forecasting}{Scale-aware
  neural architecture search for multivariate time series forecasting}}.
\newblock \btxjournalfont {ACM Trans. Knowl. Discov. Data}, 19(1):1--23,
  \btxprintmonthyear{.}{1}{2025}{short}.

\bibitem {art_726}
\btxnamefont {F.~\btxlastnamefont {Chen}}, \btxnamefont {Z.~\btxlastnamefont
  {Chen}}, \btxnamefont {S.~\btxlastnamefont {Biswas}}, \btxnamefont
  {S.~\btxlastnamefont {Lei}}, \btxnamefont {N.~\btxlastnamefont
  {Ramakrishnan}}\btxandcomma {} \btxandshort {.}\ \btxnamefont
  {C.\btxfnamespaceshort T. \btxlastnamefont {Lu}}\btxauthorcolon\
  \btxtitlefont {\btxifchangecase {Graph convolutional networks with kalman
  filtering for traffic prediction}{Graph convolutional networks with Kalman
  filtering for traffic prediction}}.
\newblock \Btxinshort {.}\ \btxtitlefont {Proceedings of the 28th International
  Conference on Advances in Geographic Information Systems}, New York, NY, USA,
  \btxprintmonthyear{.}{11}{2020}{short}. \btxpublisherfont {ACM}.

\bibitem {art_144}
\btxnamefont {G.~\btxlastnamefont {Chen}}, \btxnamefont {Y.~\btxlastnamefont
  {Guo}}, \btxnamefont {Q.~\btxlastnamefont {Zeng}}\btxandcomma {} \btxandshort
  {.}\ \btxnamefont {Y.~\btxlastnamefont {Zhang}}\btxauthorcolon\
  \btxjtitlefont {\btxifchangecase {A novel cellular network traffic prediction
  algorithm based on graph convolution neural networks and long short-term
  memory through extraction of spatial-temporal characteristics}{A Novel
  Cellular Network Traffic Prediction Algorithm Based on Graph Convolution
  Neural Networks and Long Short-Term Memory through Extraction of
  Spatial-Temporal Characteristics}}.
\newblock \btxjournalfont {Processes}, 11(8), 2023.
\newblock {\latintext
  \btxurlfont{https://www.scopus.com/inward/record.uri?eid=2-s2.0-85169154539&doi=10.3390

\bibitem {survey_Lucheroni}
\btxnamefont {H.~\btxlastnamefont {Chen}} \btxandshort {.}\ \btxnamefont
  {H.~\btxlastnamefont {Eldardiry}}\btxauthorcolon\ \btxjtitlefont
  {\btxifchangecase {Graph time-series modeling in deep learning: A
  survey}{Graph Time-series Modeling in Deep Learning: A Survey}}.
\newblock \btxjournalfont {ACM Trans. Knowl. Discov. Data}, 18(5),
  \btxprintmonthyear{.}{feb}{2024}{short}\ifbtxprintISSN {,
  \mbox{\btxISSN~\btxISSNfont {1556-4681}}}.
\newblock {\latintext \btxurlfont{https://doi.org/10.1145/3638534}}.

\bibitem {art_69}
\btxnamefont {H.~\btxlastnamefont {Chen}}, \btxnamefont {R.\btxfnamespaceshort
  A. \btxlastnamefont {Rossi}}, \btxnamefont {K.~\btxlastnamefont {Mahadik}},
  \btxnamefont {S.~\btxlastnamefont {Kim}}\btxandcomma {} \btxandshort {.}\
  \btxnamefont {H.~\btxlastnamefont {Eldardiry}}\btxauthorcolon\ \btxjtitlefont
  {\btxifchangecase {Graph deep factors for probabilistic time-series
  forecasting}{Graph Deep Factors for Probabilistic Time-series Forecasting}}.
\newblock \btxjournalfont {ACM Trans. Knowl. Discov. Data}, 17(2),
  \btxprintmonthyear{.}{feb}{2023}{short}\ifbtxprintISSN {,
  \mbox{\btxISSN~\btxISSNfont {1556-4681}}}.
\newblock {\latintext \btxurlfont{https://doi.org/10.1145/3543511}}.

\bibitem {art_382}
\btxnamefont {J.~\btxlastnamefont {Chen}}, \btxnamefont {L.~\btxlastnamefont
  {Xie}}, \btxnamefont {W.~\btxlastnamefont {Lin}}, \btxnamefont
  {Y.~\btxlastnamefont {Wu}}\btxandcomma {} \btxandshort {.}\ \btxnamefont
  {H.~\btxlastnamefont {Xu}}\btxauthorcolon\ \btxjtitlefont {\btxifchangecase
  {Multi-granularity spatio-temporal correlation networks for stock trend
  prediction}{Multi-granularity spatio-temporal correlation networks for stock
  trend prediction}}.
\newblock \btxjournalfont {IEEE Access}, 12:67219--67232, 2024.

\bibitem {art_658}
\btxnamefont {J.~\btxlastnamefont {Chen}}, \btxnamefont {Y.~\btxlastnamefont
  {Yang}}, \btxnamefont {T.~\btxlastnamefont {Yu}}, \btxnamefont
  {Y.~\btxlastnamefont {Fan}}, \btxnamefont {X.~\btxlastnamefont
  {Mo}}\btxandcomma {} \btxandshort {.}\ \btxnamefont {C.~\btxlastnamefont
  {Yang}}\btxauthorcolon\ \btxtitlefont {\btxifchangecase {Brainnet: Epileptic
  wave detection from seeg with hierarchical graph diffusion
  learning}{BrainNet: Epileptic Wave Detection from SEEG with Hierarchical
  Graph Diffusion Learning}}.
\newblock \Btxinshort {.}\ \btxtitlefont {Proceedings of the 28th ACM SIGKDD
  Conference on Knowledge Discovery and Data Mining}, KDD '22, \btxpageshort
  {.}\ 2741–2751, New York, NY, USA, 2022. \btxpublisherfont {Association for
  Computing Machinery}\ifbtxprintISBN {, \mbox{\btxISBN~\btxISBNfont
  {9781450393850}}}.
\newblock {\latintext \btxurlfont{https://doi.org/10.1145/3534678.3539178}}.

\bibitem {art_126}
\btxnamefont {K.~\btxlastnamefont {Chen}}, \btxnamefont {F.~\btxlastnamefont
  {Chen}}, \btxnamefont {B.~\btxlastnamefont {Lai}}, \btxnamefont
  {Z.~\btxlastnamefont {Jin}}, \btxnamefont {Y.~\btxlastnamefont {Liu}},
  \btxnamefont {K.~\btxlastnamefont {Li}}, \btxnamefont {L.~\btxlastnamefont
  {Wei}}, \btxnamefont {P.~\btxlastnamefont {Wang}}, \btxnamefont
  {Y.~\btxlastnamefont {Tang}}, \btxnamefont {J.~\btxlastnamefont
  {Huang}}\btxandcomma {} \btxandshort {.}\ \btxnamefont {X.\btxfnamespaceshort
  S. \btxlastnamefont {Hua}}\btxauthorcolon\ \btxjtitlefont {\btxifchangecase
  {Dynamic spatio-temporal graph-based cnns for traffic flow
  prediction}{Dynamic spatio-temporal graph-based CNNs for traffic flow
  prediction}}.
\newblock \btxjournalfont {IEEE Access}, 8:185136 – 185145, 2020.
\newblock {\latintext
  \btxurlfont{https://www.scopus.com/inward/record.uri?eid=2-s2.0-85098607394&doi=10.1109

\bibitem {art_80}
\btxnamefont {L.~\btxlastnamefont {Chen}}, \btxnamefont {D.~\btxlastnamefont
  {Chen}}, \btxnamefont {Z.~\btxlastnamefont {Shang}}, \btxnamefont
  {B.~\btxlastnamefont {Wu}}, \btxnamefont {C.~\btxlastnamefont {Zheng}},
  \btxnamefont {B.~\btxlastnamefont {Wen}}\btxandcomma {} \btxandshort {.}\
  \btxnamefont {W.~\btxlastnamefont {Zhang}}\btxauthorcolon\ \btxjtitlefont
  {\btxifchangecase {Multi-scale adaptive graph neural network for multivariate
  time series forecasting}{Multi-Scale Adaptive Graph Neural Network for
  Multivariate Time Series Forecasting}}.
\newblock \btxjournalfont {IEEE Transactions on Knowledge and Data
  Engineering}, 35(10):10748--10761, 2023.

\bibitem {art_1}
\btxnamefont {L.~\btxlastnamefont {Chen}}, \btxnamefont {J.~\btxlastnamefont
  {Xu}}, \btxnamefont {B.~\btxlastnamefont {Wu}}\btxandcomma {} \btxandshort
  {.}\ \btxnamefont {J.~\btxlastnamefont {Huang}}\btxauthorcolon\
  \btxjtitlefont {\btxifchangecase {Group-aware graph neural network for
  nationwide city air quality forecasting}{Group-Aware Graph Neural Network for
  Nationwide City Air Quality Forecasting}}.
\newblock \btxjournalfont {ACM Trans. Knowl. Discov. Data}, 18(3),
  \btxprintmonthyear{.}{dec}{2023}{short}\ifbtxprintISSN {,
  \mbox{\btxISSN~\btxISSNfont {1556-4681}}}.
\newblock {\latintext \btxurlfont{https://doi.org/10.1145/3631713}}.

\bibitem {art_636}
\btxnamefont {S.~\btxlastnamefont {Chen}}, \btxnamefont {J.\btxfnamespaceshort
  A. \btxlastnamefont {Zwart}}\btxandcomma {} \btxandshort {.}\ \btxnamefont
  {X.~\btxlastnamefont {Jia}}\btxauthorcolon\ \btxtitlefont {\btxifchangecase
  {Physics-guided graph meta learning for predicting water temperature and
  streamflow in stream networks}{Physics-guided graph meta learning for
  predicting water temperature and streamflow in stream networks}}.
\newblock \Btxinshort {.}\ \btxtitlefont {Proceedings of the 28th {ACM}
  {SIGKDD} Conference on Knowledge Discovery and Data Mining}, \btxpagesshort
  {.}\ 2752--2761, New York, NY, USA, \btxprintmonthyear{.}{8}{2022}{short}.
  \btxpublisherfont {ACM}.

\bibitem {XGBoost}
\btxnamefont {T.~\btxlastnamefont {Chen}} \btxandshort {.}\ \btxnamefont
  {C.~\btxlastnamefont {Guestrin}}\btxauthorcolon\ \btxtitlefont
  {\btxifchangecase {Xgboost: A scalable tree boosting system}{XGBoost: A
  Scalable Tree Boosting System}}.
\newblock \Btxinshort {.}\ \btxtitlefont {Proceedings of the 22nd ACM SIGKDD
  International Conference on Knowledge Discovery and Data Mining}, KDD '16,
  \btxpageshort {.}\ 785–794, New York, NY, USA, 2016. \btxpublisherfont
  {Association for Computing Machinery}\ifbtxprintISBN {,
  \mbox{\btxISBN~\btxISBNfont {9781450342322}}}.
\newblock {\latintext \btxurlfont{https://doi.org/10.1145/2939672.2939785}}.

\bibitem {MRA-BGCN}
\btxnamefont {W.~\btxlastnamefont {Chen}}, \btxnamefont {L.~\btxlastnamefont
  {Chen}}, \btxnamefont {Y.~\btxlastnamefont {Xie}}, \btxnamefont
  {W.~\btxlastnamefont {Cao}}, \btxnamefont {Y.~\btxlastnamefont
  {Gao}}\btxandcomma {} \btxandshort {.}\ \btxnamefont {X.~\btxlastnamefont
  {Feng}}\btxauthorcolon\ \btxjtitlefont {\btxifchangecase {Multi-range
  attentive bicomponent graph convolutional network for traffic
  forecasting}{Multi-Range Attentive Bicomponent Graph Convolutional Network
  for Traffic Forecasting}}.
\newblock \btxjournalfont {Proceedings of the AAAI Conference on Artificial
  Intelligence}, 34(04):3529--3536, \btxprintmonthyear{.}{Apr.}{2020}{short}.
\newblock {\latintext
  \btxurlfont{https://ojs.aaai.org/index.php/AAAI/article/view/5758}}.

\bibitem {art_368}
\btxnamefont {X.~\btxlastnamefont {Chen}}, \btxnamefont {H.~\btxlastnamefont
  {Xia}}, \btxnamefont {M.~\btxlastnamefont {Wu}}, \btxnamefont
  {Y.~\btxlastnamefont {Hu}}\btxandcomma {} \btxandshort {.}\ \btxnamefont
  {Z.~\btxlastnamefont {Wang}}\btxauthorcolon\ \btxjtitlefont {\btxifchangecase
  {Spatiotemporal hierarchical transmit neural network for regional-level
  air-quality prediction}{Spatiotemporal hierarchical transmit neural network
  for regional-level air-quality prediction}}.
\newblock \btxjournalfont {Knowl. Based Syst.}, 289(111555):111555,
  \btxprintmonthyear{.}{4}{2024}{short}.

\bibitem {art_153}
\btxnamefont {X.~\btxlastnamefont {Chen}} \btxandshort {.}\ \btxnamefont
  {M.~\btxlastnamefont {Zeng}}\btxauthorcolon\ \btxjtitlefont {\btxifchangecase
  {Convolution-graph attention network with sensor embeddings for remaining
  useful life prediction of turbofan engines}{Convolution-Graph Attention
  Network with Sensor Embeddings for Remaining Useful Life Prediction of
  Turbofan Engines}}.
\newblock \btxjournalfont {IEEE Sensors Journal}, 23(14):15786 – 15794, 2023.
\newblock {\latintext
  \btxurlfont{https://www.scopus.com/inward/record.uri?eid=2-s2.0-85162688641&doi=10.1109

\bibitem {art_84}
\btxnamefont {Y.~\btxlastnamefont {Chen}}, \btxnamefont {F.~\btxlastnamefont
  {Ding}}\btxandcomma {} \btxandshort {.}\ \btxnamefont {L.~\btxlastnamefont
  {Zhai}}\btxauthorcolon\ \btxjtitlefont {\btxifchangecase {Multi-scale
  temporal features extraction based graph convolutional network with attention
  for multivariate time series prediction}{Multi-scale temporal features
  extraction based graph convolutional network with attention for multivariate
  time series prediction}}.
\newblock \btxjournalfont {Expert Systems with Applications}, 200, 2022.
\newblock {\latintext
  \btxurlfont{https://www.scopus.com/inward/record.uri?eid=2-s2.0-85127194643&doi=10.1016

\bibitem {art_686}
\btxnamefont {Y.~\btxlastnamefont {Chen}}, \btxnamefont {Y.\btxfnamespaceshort
  R. \btxlastnamefont {Gel}}\btxandcomma {} \btxandshort {.}\ \btxnamefont
  {H.\btxfnamespaceshort V. \btxlastnamefont {Poor}}\btxauthorcolon\
  \btxtitlefont {\btxifchangecase {Time-conditioned dances with simplicial
  complexes: zigzag filtration curve based supra-hodge convolution networks for
  time-series forecasting}{Time-conditioned dances with simplicial complexes:
  zigzag filtration curve based supra-hodge convolution networks for
  time-series forecasting}}.
\newblock \Btxinshort {.}\ \btxtitlefont {Proceedings of the 36th International
  Conference on Neural Information Processing Systems}, NIPS '22, Red Hook, NY,
  USA, 2022. \btxpublisherfont {Curran Associates Inc.}\ifbtxprintISBN {,
  \mbox{\btxISBN~\btxISBNfont {9781713871088}}}.

\bibitem {art_701}
\btxnamefont {Y.~\btxlastnamefont {Chen}}, \btxnamefont {T.~\btxlastnamefont
  {Jiang}}\btxandcomma {} \btxandshort {.}\ \btxnamefont {Y.\btxfnamespaceshort
  R. \btxlastnamefont {Gel}}\btxauthorcolon\ \btxtitlefont {\btxifchangecase
  {{H$^2$-nets}: Hyper-hodge convolutional neural networks for time-series
  forecasting}{{H$^2$-nets}: Hyper-hodge convolutional neural networks for
  time-series forecasting}}.
\newblock \Btxinshort {.}\ \btxtitlefont {Lecture Notes in Computer Science},
  Lecture notes in computer science, \btxpagesshort {.}\ 271--289.
  \btxpublisherfont {Springer Nature Switzerland}, Cham, 2023.

\bibitem {art_657}
\btxnamefont {Y.~\btxlastnamefont {Chen}}, \btxnamefont {I.~\btxlastnamefont
  {Segovia}}\btxandcomma {} \btxandshort {.}\ \btxnamefont
  {Y.\btxfnamespaceshort R. \btxlastnamefont {Gel}}\btxauthorcolon\
  \btxtitlefont {\btxifchangecase {Z-gcnets: Time zigzags at graph
  convolutional networks for time series forecasting}{Z-GCNETs: Time Zigzags at
  Graph Convolutional Networks for Time Series Forecasting}}.
\newblock \Btxinshort {.}\ \btxnamefont {M.~\btxlastnamefont {Meila}}
  \btxandshort {.}\ \btxnamefont {T.~\btxlastnamefont {Zhang}}\
  (\btxeditorsshort {.}): \btxtitlefont {Proceedings of the 38th International
  Conference on Machine Learning}, \btxvolumeshort {.}\ \btxvolumefont {139}
  \btxofseriesshort {.}\ \btxtitlefont {Proceedings of Machine Learning
  Research}, \btxpagesshort {.}\ 1684--1694. \btxpublisherfont {PMLR},
  \btxprintmonthyear{.}{18--24 Jul}{2021}{short}.
\newblock {\latintext
  \btxurlfont{https://proceedings.mlr.press/v139/chen21o.html}}.

\bibitem {art_674}
\btxnamefont {Y.~\btxlastnamefont {Chen}}, \btxnamefont {I.~\btxlastnamefont
  {Segovia-Dominguez}}, \btxnamefont {B.~\btxlastnamefont
  {Coskunuzer}}\btxandcomma {} \btxandshort {.}\ \btxnamefont
  {Y.~\btxlastnamefont {Gel}}\btxauthorcolon\ \btxtitlefont {\btxifchangecase
  {{TAMP}-s2{GCN}ets: Coupling time-aware multipersistence knowledge
  representation with spatio-supra graph convolutional networks for time-series
  forecasting}{{TAMP}-S2{GCN}ets: Coupling Time-Aware Multipersistence
  Knowledge Representation with Spatio-Supra Graph Convolutional Networks for
  Time-Series Forecasting}}.
\newblock \Btxinshort {.}\ \btxtitlefont {International Conference on Learning
  Representations}, 2022.
\newblock {\latintext
  \btxurlfont{https://openreview.net/forum?id=wv6g8fWLX2q}}.

\bibitem {art_86}
\btxnamefont {Y.~\btxlastnamefont {Chen}} \btxandshort {.}\ \btxnamefont
  {Z.~\btxlastnamefont {Xie}}\btxauthorcolon\ \btxjtitlefont {\btxifchangecase
  {Multi-channel fusion graph neural network for multivariate time series
  forecasting}{Multi-channel fusion graph neural network for multivariate time
  series forecasting}}.
\newblock \btxjournalfont {Journal of Computational Science}, 64, 2022.
\newblock {\latintext
  \btxurlfont{https://www.scopus.com/inward/record.uri?eid=2-s2.0-85139025049&doi=10.1016

\bibitem {art_66}
\btxnamefont {D.~\btxlastnamefont {Cheng}}, \btxnamefont {F.~\btxlastnamefont
  {Yang}}, \btxnamefont {S.~\btxlastnamefont {Xiang}}\btxandcomma {}
  \btxandshort {.}\ \btxnamefont {J.~\btxlastnamefont {Liu}}\btxauthorcolon\
  \btxjtitlefont {\btxifchangecase {Financial time series forecasting with
  multi-modality graph neural network}{Financial time series forecasting with
  multi-modality graph neural network}}.
\newblock \btxjournalfont {Pattern Recognition}, 121, 2022.
\newblock {\latintext
  \btxurlfont{https://www.scopus.com/inward/record.uri?eid=2-s2.0-85112022477&doi=10.1016

\bibitem {MLCNN}
\btxnamefont {J.~\btxlastnamefont {Cheng}}, \btxnamefont {K.~\btxlastnamefont
  {Huang}}\btxandcomma {} \btxandshort {.}\ \btxnamefont {Z.~\btxlastnamefont
  {Zheng}}\btxauthorcolon\ \btxjtitlefont {\btxifchangecase {Towards better
  forecasting by fusing near and distant future visions}{Towards Better
  Forecasting by Fusing Near and Distant Future Visions}}.
\newblock \btxjournalfont {Proceedings of the AAAI Conference on Artificial
  Intelligence}, 34(04):3593–3600,
  \btxprintmonthyear{.}{4}{2020}{short}\ifbtxprintISSN {,
  \mbox{\btxISSN~\btxISSNfont {2159-5399}}}.
\newblock {\latintext \btxurlfont{http://dx.doi.org/10.1609/aaai.v34i04.5766}}.

\bibitem {survey_Chikwendu}
\btxnamefont {I.\btxfnamespaceshort A. \btxlastnamefont {Chikwendu}},
  \btxnamefont {X.~\btxlastnamefont {Zhang}}, \btxnamefont
  {I.\btxfnamespaceshort O. \btxlastnamefont {Agyemang}}, \btxnamefont
  {I.~\btxlastnamefont {Adjei-Mensah}}, \btxnamefont {U.\btxfnamespaceshort C.
  \btxlastnamefont {Chima}}\btxandcomma {} \btxandshort {.}\ \btxnamefont
  {C.\btxfnamespaceshort J. \btxlastnamefont {Ejiyi}}\btxauthorcolon\
  \btxjtitlefont {\btxifchangecase {A comprehensive survey on deep graph
  representation learning methods}{A Comprehensive Survey on Deep Graph
  Representation Learning Methods}}.
\newblock \btxjournalfont {J. Artif. Int. Res.}, 78,
  \btxprintmonthyear{.}{1}{2024}{short}\ifbtxprintISSN {,
  \mbox{\btxISSN~\btxISSNfont {1076-9757}}}.
\newblock {\latintext \btxurlfont{https://doi.org/10.1613/jair.1.14768}}.

\bibitem {STG-NCDE}
\btxnamefont {J.~\btxlastnamefont {Choi}}, \btxnamefont {H.~\btxlastnamefont
  {Choi}}, \btxnamefont {J.~\btxlastnamefont {Hwang}}\btxandcomma {}
  \btxandshort {.}\ \btxnamefont {N.~\btxlastnamefont {Park}}\btxauthorcolon\
  \btxjtitlefont {\btxifchangecase {Graph neural controlled differential
  equations for traffic forecasting}{Graph Neural Controlled Differential
  Equations for Traffic Forecasting}}.
\newblock \btxjournalfont {Proceedings of the AAAI Conference on Artificial
  Intelligence}, 36(6):6367–6374,
  \btxprintmonthyear{.}{6}{2022}{short}\ifbtxprintISSN {,
  \mbox{\btxISSN~\btxISSNfont {2159-5399}}}.
\newblock {\latintext \btxurlfont{http://dx.doi.org/10.1609/aaai.v36i6.20587}}.

\bibitem {art_89}
\btxnamefont {W.~\btxlastnamefont {Chung}}, \btxnamefont {J.~\btxlastnamefont
  {Moon}}, \btxnamefont {D.~\btxlastnamefont {Kim}}\btxandcomma {} \btxandshort
  {.}\ \btxnamefont {E.~\btxlastnamefont {Hwang}}\btxauthorcolon\
  \btxjtitlefont {\btxifchangecase {Graph construction method for gnn-based
  multivariate time-series forecasting}{Graph Construction Method for GNN-Based
  Multivariate Time-Series Forecasting}}.
\newblock \btxjournalfont {Computers, Materials and Continua}, 75(3):5817 –
  5836, 2023.
\newblock {\latintext
  \btxurlfont{https://www.scopus.com/inward/record.uri?eid=2-s2.0-85165540972&doi=10.32604

\bibitem {Cini_Torch_Spatiotemporal_2022}
\btxnamefont {A.~\btxlastnamefont {Cini}} \btxandshort {.}\ \btxnamefont
  {I.~\btxlastnamefont {Marisca}}\btxauthorcolon\ \btxtitlefont
  {\btxifchangecase {{Torch Spatiotemporal}}{{Torch Spatiotemporal}}},
  \btxprintmonthyear{.}{3}{2022}{short}.
\newblock {\latintext \btxurlfont{https://github.com/TorchSpatiotemporal/tsl}}.

\bibitem {art_689}
\btxnamefont {A.~\btxlastnamefont {Cini}}, \btxnamefont {I.~\btxlastnamefont
  {Marisca}}, \btxnamefont {D.~\btxlastnamefont {Zambon}}\btxandcomma {}
  \btxandshort {.}\ \btxnamefont {C.~\btxlastnamefont {Alippi}}\btxauthorcolon\
  \btxtitlefont {\btxifchangecase {Taming local effects in graph-based
  spatiotemporal forecasting}{Taming local effects in graph-based
  spatiotemporal forecasting}}.
\newblock \Btxinshort {.}\ \btxtitlefont {Proceedings of the 37th International
  Conference on Neural Information Processing Systems}, NIPS '23, Red Hook, NY,
  USA, 2023. \btxpublisherfont {Curran Associates Inc.}

\bibitem {art_72}
\btxnamefont {A.~\btxlastnamefont {Cini}}, \btxnamefont {D.~\btxlastnamefont
  {Zambon}}\btxandcomma {} \btxandshort {.}\ \btxnamefont {C.~\btxlastnamefont
  {Alippi}}\btxauthorcolon\ \btxjtitlefont {\btxifchangecase {Sparse graph
  learning from spatiotemporal time series}{Sparse Graph Learning from
  Spatiotemporal Time Series}}.
\newblock \btxjournalfont {Journal of Machine Learning Research}, 24, 2023.
\newblock {\latintext
  \btxurlfont{https://www.scopus.com/inward/record.uri?eid=2-s2.0-85175601404&partnerID=40&md5=e8cf0bc88a664da0ef171127c7a14f0d}}.

\bibitem {art_656}
\btxnamefont {Y.~\btxlastnamefont {Cui}}, \btxnamefont {K.~\btxlastnamefont
  {Zheng}}, \btxnamefont {D.~\btxlastnamefont {Cui}}, \btxnamefont
  {J.~\btxlastnamefont {Xie}}, \btxnamefont {L.~\btxlastnamefont {Deng}},
  \btxnamefont {F.~\btxlastnamefont {Huang}}\btxandcomma {} \btxandshort {.}\
  \btxnamefont {X.~\btxlastnamefont {Zhou}}\btxauthorcolon\ \btxjtitlefont
  {\btxifchangecase {Metro: a generic graph neural network framework for
  multivariate time series forecasting}{METRO: a generic graph neural network
  framework for multivariate time series forecasting}}.
\newblock \btxjournalfont {Proc. VLDB Endow.}, 15(2):224–236,
  \btxprintmonthyear{.}{10}{2021}{short}\ifbtxprintISSN {,
  \mbox{\btxISSN~\btxISSNfont {2150-8097}}}.
\newblock {\latintext \btxurlfont{https://doi.org/10.14778/3489496.3489503}}.

\bibitem {art_97}
\btxnamefont {S.~\btxlastnamefont {Dai}}, \btxnamefont {J.~\btxlastnamefont
  {Wang}}, \btxnamefont {C.~\btxlastnamefont {Huang}}, \btxnamefont
  {Y.~\btxlastnamefont {Yu}}\btxandcomma {} \btxandshort {.}\ \btxnamefont
  {J.~\btxlastnamefont {Dong}}\btxauthorcolon\ \btxjtitlefont {\btxifchangecase
  {Dynamic multi-view graph neural networks for citywide traffic
  inference}{Dynamic Multi-View Graph Neural Networks for Citywide Traffic
  Inference}}.
\newblock \btxjournalfont {ACM Trans. Knowl. Discov. Data}, 17(4),
  \btxprintmonthyear{.}{feb}{2023}{short}\ifbtxprintISSN {,
  \mbox{\btxISSN~\btxISSNfont {1556-4681}}}.
\newblock {\latintext \btxurlfont{https://doi.org/10.1145/3564754}}.

\bibitem {art_721}
\btxnamefont {T.~\btxlastnamefont {Dan}}, \btxnamefont {X.~\btxlastnamefont
  {Pan}}, \btxnamefont {B.~\btxlastnamefont {Zheng}}\btxandcomma {}
  \btxandshort {.}\ \btxnamefont {X.~\btxlastnamefont {Meng}}\btxauthorcolon\
  \btxtitlefont {\btxifchangecase {{ByGCN}: Spatial temporal byroad-aware graph
  convolution network for traffic flow prediction in road networks}{{ByGCN}:
  Spatial temporal byroad-aware graph convolution network for traffic flow
  prediction in road networks}}.
\newblock \Btxinshort {.}\ \btxtitlefont {Proceedings of the 33rd {ACM}
  International Conference on Information and Knowledge Management},
  \btxvolumeshort {.}\ \btxvolumefont {121}, \btxpagesshort {.}\ 415--424, New
  York, NY, USA, \btxprintmonthyear{.}{10}{2024}{short}. \btxpublisherfont
  {ACM}.

\bibitem {ROCKET}
\btxnamefont {A.~\btxlastnamefont {Dempster}}, \btxnamefont
  {F.~\btxlastnamefont {Petitjean}}\btxandcomma {} \btxandshort {.}\
  \btxnamefont {G.\btxfnamespaceshort I. \btxlastnamefont
  {Webb}}\btxauthorcolon\ \btxjtitlefont {\btxifchangecase {Rocket:
  exceptionally fast and accurate time series classification using random
  convolutional kernels}{ROCKET: exceptionally fast and accurate time series
  classification using random convolutional kernels}}.
\newblock \btxjournalfont {Data Mining and Knowledge Discovery},
  34(5):1454–1495, \btxprintmonthyear{.}{7}{2020}{short}\ifbtxprintISSN {,
  \mbox{\btxISSN~\btxISSNfont {1573-756X}}}.
\newblock {\latintext
  \btxurlfont{http://dx.doi.org/10.1007/s10618-020-00701-z}}.

\bibitem {MiniRocket}
\btxnamefont {A.~\btxlastnamefont {Dempster}}, \btxnamefont
  {D.\btxfnamespaceshort F. \btxlastnamefont {Schmidt}}\btxandcomma {}
  \btxandshort {.}\ \btxnamefont {G.\btxfnamespaceshort I. \btxlastnamefont
  {Webb}}\btxauthorcolon\ \btxtitlefont {\btxifchangecase {Minirocket: A very
  fast (almost) deterministic transform for time series
  classification}{MiniRocket: A Very Fast (Almost) Deterministic Transform for
  Time Series Classification}}.
\newblock \Btxinshort {.}\ \btxtitlefont {Proceedings of the 27th ACM SIGKDD
  Conference on Knowledge Discovery \& Data Mining}, KDD '21, \btxpageshort
  {.}\ 248–257, New York, NY, USA, 2021. \btxpublisherfont {Association for
  Computing Machinery}\ifbtxprintISBN {, \mbox{\btxISBN~\btxISBNfont
  {9781450383325}}}.
\newblock {\latintext \btxurlfont{https://doi.org/10.1145/3447548.3467231}}.

\bibitem {GDN}
\btxnamefont {A.~\btxlastnamefont {Deng}} \btxandshort {.}\ \btxnamefont
  {B.~\btxlastnamefont {Hooi}}\btxauthorcolon\ \btxjtitlefont {\btxifchangecase
  {Graph neural network-based anomaly detection in multivariate time
  series}{Graph Neural Network-Based Anomaly Detection in Multivariate Time
  Series}}.
\newblock \btxjournalfont {Proceedings of the AAAI Conference on Artificial
  Intelligence}, 35(5):4027–4035,
  \btxprintmonthyear{.}{5}{2021}{short}\ifbtxprintISSN {,
  \mbox{\btxISSN~\btxISSNfont {2159-5399}}}.
\newblock {\latintext \btxurlfont{http://dx.doi.org/10.1609/aaai.v35i5.16523}}.

\bibitem {ST-Norm}
\btxnamefont {J.~\btxlastnamefont {Deng}}, \btxnamefont {X.~\btxlastnamefont
  {Chen}}, \btxnamefont {R.~\btxlastnamefont {Jiang}}, \btxnamefont
  {X.~\btxlastnamefont {Song}}\btxandcomma {} \btxandshort {.}\ \btxnamefont
  {I.\btxfnamespaceshort W. \btxlastnamefont {Tsang}}\btxauthorcolon\
  \btxtitlefont {\btxifchangecase {St-norm: Spatial and temporal normalization
  for multi-variate time series forecasting}{ST-Norm: Spatial and Temporal
  Normalization for Multi-variate Time Series Forecasting}}.
\newblock \Btxinshort {.}\ \btxtitlefont {Proceedings of the 27th ACM SIGKDD
  Conference on Knowledge Discovery \& Data Mining}, KDD '21, \btxpageshort
  {.}\ 269–278, New York, NY, USA, 2021. \btxpublisherfont {Association for
  Computing Machinery}\ifbtxprintISBN {, \mbox{\btxISBN~\btxISBNfont
  {9781450383325}}}.
\newblock {\latintext \btxurlfont{https://doi.org/10.1145/3447548.3467330}}.

\bibitem {art_712}
\btxnamefont {S.~\btxlastnamefont {Deng}}, \btxnamefont {S.~\btxlastnamefont
  {Wang}}, \btxnamefont {H.~\btxlastnamefont {Rangwala}}, \btxnamefont
  {L.~\btxlastnamefont {Wang}}\btxandcomma {} \btxandshort {.}\ \btxnamefont
  {Y.~\btxlastnamefont {Ning}}\btxauthorcolon\ \btxtitlefont {\btxifchangecase
  {Cola-gnn: Cross-location attention based graph neural networks for long-term
  ili prediction}{Cola-GNN: Cross-location Attention based Graph Neural
  Networks for Long-term ILI Prediction}}.
\newblock \Btxinshort {.}\ \btxtitlefont {Proceedings of the 29th ACM
  International Conference on Information \& Knowledge Management}, CIKM '20,
  \btxpageshort {.}\ 245–254, New York, NY, USA, 2020. \btxpublisherfont
  {Association for Computing Machinery}\ifbtxprintISBN {,
  \mbox{\btxISBN~\btxISBNfont {9781450368599}}}.
\newblock {\latintext \btxurlfont{https://doi.org/10.1145/3340531.3411975}}.

\bibitem {art_131}
\btxnamefont {C.~\btxlastnamefont {Diao}}, \btxnamefont {D.~\btxlastnamefont
  {Zhang}}, \btxnamefont {W.~\btxlastnamefont {Liang}}, \btxnamefont
  {K.\btxfnamespaceshort C. \btxlastnamefont {Li}}, \btxnamefont
  {Y.~\btxlastnamefont {Hong}}\btxandcomma {} \btxandshort {.}\ \btxnamefont
  {J.\btxfnamespaceshort L. \btxlastnamefont {Gaudiot}}\btxauthorcolon\
  \btxjtitlefont {\btxifchangecase {A novel spatial-temporal multi-scale
  alignment graph neural network security model for vehicles prediction}{A
  Novel Spatial-Temporal Multi-Scale Alignment Graph Neural Network Security
  Model for Vehicles Prediction}}.
\newblock \btxjournalfont {IEEE Transactions on Intelligent Transportation
  Systems}, 24(1):904 – 914, 2023.
\newblock {\latintext
  \btxurlfont{https://www.scopus.com/inward/record.uri?eid=2-s2.0-85123379822&doi=10.1109

\bibitem {art_118}
\btxnamefont {Z.~\btxlastnamefont {Diao}}, \btxnamefont {G.~\btxlastnamefont
  {Xie}}, \btxnamefont {X.~\btxlastnamefont {Wang}}, \btxnamefont
  {R.~\btxlastnamefont {Ren}}, \btxnamefont {X.~\btxlastnamefont {Meng}},
  \btxnamefont {G.~\btxlastnamefont {Zhang}}, \btxnamefont {K.~\btxlastnamefont
  {Xie}}\btxandcomma {} \btxandshort {.}\ \btxnamefont {M.~\btxlastnamefont
  {Qiao}}\btxauthorcolon\ \btxjtitlefont {\btxifchangecase {Ec-gcn: A encrypted
  traffic classification framework based on multi-scale graph convolution
  networks}{EC-GCN: A encrypted traffic classification framework based on
  multi-scale graph convolution networks}}.
\newblock \btxjournalfont {Computer Networks}, 224, 2023.
\newblock {\latintext
  \btxurlfont{https://www.scopus.com/inward/record.uri?eid=2-s2.0-85148540984&doi=10.1016

\bibitem {art_133}
\btxnamefont {H.~\btxlastnamefont {Ding}}, \btxnamefont {Y.~\btxlastnamefont
  {Lu}}, \btxnamefont {N.~\btxlastnamefont {Sze}}\btxandcomma {} \btxandshort
  {.}\ \btxnamefont {H.~\btxlastnamefont {Li}}\btxauthorcolon\ \btxjtitlefont
  {\btxifchangecase {Effect of dockless bike-sharing scheme on the demand for
  london cycle hire at the disaggregate level using a deep learning
  approach}{Effect of dockless bike-sharing scheme on the demand for London
  Cycle Hire at the disaggregate level using a deep learning approach}}.
\newblock \btxjournalfont {Transportation Research Part A: Policy and
  Practice}, 166:150 – 163, 2022.
\newblock {\latintext
  \btxurlfont{https://www.scopus.com/inward/record.uri?eid=2-s2.0-85143767347&doi=10.1016

\bibitem {art_5}
\btxnamefont {H.~\btxlastnamefont {Ding}} \btxandshort {.}\ \btxnamefont
  {G.~\btxlastnamefont {Noh}}\btxauthorcolon\ \btxjtitlefont {\btxifchangecase
  {A hybrid model for spatiotemporal air quality prediction based on
  interpretable neural networks and a graph neural network}{A Hybrid Model for
  Spatiotemporal Air Quality Prediction Based on Interpretable Neural Networks
  and a Graph Neural Network}}.
\newblock \btxjournalfont {Atmosphere}, 14(12), 2023.
\newblock {\latintext
  \btxurlfont{https://www.scopus.com/inward/record.uri?eid=2-s2.0-85180550137&doi=10.3390

\bibitem {art_46}
\btxnamefont {D.~\btxlastnamefont {Dong}}, \btxnamefont {S.~\btxlastnamefont
  {Wang}}, \btxnamefont {Q.~\btxlastnamefont {Guo}}, \btxnamefont
  {X.~\btxlastnamefont {Li}}, \btxnamefont {W.~\btxlastnamefont
  {Zou}}\btxandcomma {} \btxandshort {.}\ \btxnamefont {Z.~\btxlastnamefont
  {You}}\btxauthorcolon\ \btxjtitlefont {\btxifchangecase {Ocean wind speed
  prediction based on the fusion of spatial clustering and an improved residual
  graph attention network}{Ocean Wind Speed Prediction Based on the Fusion of
  Spatial Clustering and an Improved Residual Graph Attention Network}}.
\newblock \btxjournalfont {Journal of Marine Science and Engineering}, 11:2350,
  \btxprintmonthyear{.}{12}{2023}{short}.

\bibitem {art_385}
\btxnamefont {M.~\btxlastnamefont {Du}}, \btxnamefont {Y.~\btxlastnamefont
  {Wei}}, \btxnamefont {Y.~\btxlastnamefont {Hu}}, \btxnamefont
  {X.~\btxlastnamefont {Zheng}}\btxandcomma {} \btxandshort {.}\ \btxnamefont
  {C.~\btxlastnamefont {Ji}}\btxauthorcolon\ \btxjtitlefont {\btxifchangecase
  {Multivariate time series classification based on fusion
  features}{Multivariate time series classification based on fusion features}}.
\newblock \btxjournalfont {Expert Syst. Appl.}, 248(123452):123452,
  \btxprintmonthyear{.}{8}{2024}{short}.

\bibitem {AdaRNN}
\btxnamefont {Y.~\btxlastnamefont {Du}}, \btxnamefont {J.~\btxlastnamefont
  {Wang}}, \btxnamefont {W.~\btxlastnamefont {Feng}}, \btxnamefont
  {S.~\btxlastnamefont {Pan}}, \btxnamefont {T.~\btxlastnamefont {Qin}},
  \btxnamefont {R.~\btxlastnamefont {Xu}}\btxandcomma {} \btxandshort {.}\
  \btxnamefont {C.~\btxlastnamefont {Wang}}\btxauthorcolon\ \btxtitlefont
  {\btxifchangecase {Adarnn: Adaptive learning and forecasting of time
  series}{AdaRNN: Adaptive Learning and Forecasting of Time Series}}.
\newblock \Btxinshort {.}\ \btxtitlefont {Proceedings of the 30th ACM
  International Conference on Information \& Knowledge Management}, CIKM '21,
  \btxpageshort {.}\ 402–411, New York, NY, USA, 2021. \btxpublisherfont
  {Association for Computing Machinery}\ifbtxprintISBN {,
  \mbox{\btxISBN~\btxISBNfont {9781450384469}}}.
\newblock {\latintext \btxurlfont{https://doi.org/10.1145/3459637.3482315}}.

\bibitem {art_77}
\btxnamefont {Z.~\btxlastnamefont {Duan}}, \btxnamefont {H.~\btxlastnamefont
  {Xu}}, \btxnamefont {Y.~\btxlastnamefont {Huang}}, \btxnamefont
  {J.~\btxlastnamefont {Feng}}\btxandcomma {} \btxandshort {.}\ \btxnamefont
  {Y.~\btxlastnamefont {Wang}}\btxauthorcolon\ \btxjtitlefont {\btxifchangecase
  {Multivariate time series forecasting with transfer entropy
  graph}{Multivariate Time Series Forecasting with Transfer Entropy Graph}}.
\newblock \btxjournalfont {Tsinghua Science and Technology}, 28(1):141 – 149,
  2023.
\newblock {\latintext
  \btxurlfont{https://www.scopus.com/inward/record.uri?eid=2-s2.0-85135310646&doi=10.26599

\bibitem {TS-TCC}
\btxnamefont {E.~\btxlastnamefont {Eldele}}, \btxnamefont {M.~\btxlastnamefont
  {Ragab}}, \btxnamefont {Z.~\btxlastnamefont {Chen}}, \btxnamefont
  {M.~\btxlastnamefont {Wu}}, \btxnamefont {C.\btxfnamespaceshort K.
  \btxlastnamefont {Kwoh}}, \btxnamefont {X.~\btxlastnamefont {Li}}\btxandcomma
  {} \btxandshort {.}\ \btxnamefont {C.~\btxlastnamefont
  {Guan}}\btxauthorcolon\ \btxtitlefont {\btxifchangecase {Time-series
  representation learning via temporal and contextual contrasting}{Time-Series
  Representation Learning via Temporal and Contextual Contrasting}}.
\newblock \Btxinshort {.}\ \btxtitlefont {Proceedings of the Thirtieth
  International Joint Conference on Artificial Intelligence, {IJCAI-21}},
  \btxpagesshort {.}\ 2352--2359, 2021.

\bibitem {Fan2025}
\btxnamefont {G.~\btxlastnamefont {Fan}}, \btxnamefont {A.\btxfnamespaceshort
  Q.\btxfnamespaceshort M. \btxlastnamefont {Sabri}}, \btxnamefont
  {S.\btxfnamespaceshort S.\btxfnamespaceshort A. \btxlastnamefont {Rahman}},
  \btxnamefont {L.~\btxlastnamefont {Pan}}\btxandcomma {} \btxandshort {.}\
  \btxnamefont {S.~\btxlastnamefont {Rahardja}}\btxauthorcolon\ \btxjtitlefont
  {\btxifchangecase {Emerging trends in graph neural networks for traffic flow
  prediction: A survey}{Emerging Trends in Graph Neural Networks for Traffic
  Flow Prediction: A Survey}}.
\newblock \btxjournalfont {Archives of Computational Methods in Engineering},
  \btxprintmonthyear{.}{4}{2025}{short}\ifbtxprintISSN {,
  \mbox{\btxISSN~\btxISSNfont {1886-1784}}}.
\newblock {\latintext
  \btxurlfont{http://dx.doi.org/10.1007/s11831-025-10286-9}}.

\bibitem {art_690}
\btxnamefont {Y.~\btxlastnamefont {Fan}}, \btxnamefont {R.~\btxlastnamefont
  {Wieser}}, \btxnamefont {L.\btxfnamespaceshort S. \btxlastnamefont
  {Bruckman}}, \btxnamefont {R.\btxfnamespaceshort H. \btxlastnamefont
  {French}}\btxandcomma {} \btxandshort {.}\ \btxnamefont {Y.~\btxlastnamefont
  {Wu}}\btxauthorcolon\ \btxtitlefont {\btxifchangecase {Parallel-friendly
  spatio-temporal graph learning for photovoltaic degradation analysis at
  scale}{Parallel-friendly spatio-temporal graph learning for photovoltaic
  degradation analysis at scale}}.
\newblock \Btxinshort {.}\ \btxtitlefont {Proceedings of the 33rd {ACM}
  International Conference on Information and Knowledge Management},
  \btxpagesshort {.}\ 4470--4478, New York, NY, USA,
  \btxprintmonthyear{.}{10}{2024}{short}. \btxpublisherfont {ACM}.

\bibitem {art_707}
\btxnamefont {Y.~\btxlastnamefont {Fan}}, \btxnamefont {C.\btxfnamespaceshort
  C.\btxfnamespaceshort M. \btxlastnamefont {Yeh}}, \btxnamefont
  {H.~\btxlastnamefont {Chen}}, \btxnamefont {Y.~\btxlastnamefont {Zheng}},
  \btxnamefont {L.~\btxlastnamefont {Wang}}, \btxnamefont {J.~\btxlastnamefont
  {Wang}}, \btxnamefont {X.~\btxlastnamefont {Dai}}, \btxnamefont
  {Z.~\btxlastnamefont {Zhuang}}\btxandcomma {} \btxandshort {.}\ \btxnamefont
  {W.~\btxlastnamefont {Zhang}}\btxauthorcolon\ \btxtitlefont {\btxifchangecase
  {Spatial-temporal graph boosting networks: Enhancing spatial-temporal graph
  neural networks via gradient boosting}{Spatial-temporal graph boosting
  networks: Enhancing spatial-temporal graph neural networks via gradient
  boosting}}.
\newblock \Btxinshort {.}\ \btxtitlefont {Proceedings of the 32nd {ACM}
  International Conference on Information and Knowledge Management}, New York,
  NY, USA, \btxprintmonthyear{.}{10}{2023}{short}. \btxpublisherfont {ACM}.

\bibitem {art_671}
\btxnamefont {Y.~\btxlastnamefont {Fang}}, \btxnamefont {Y.~\btxlastnamefont
  {Qin}}, \btxnamefont {H.~\btxlastnamefont {Luo}}, \btxnamefont
  {F.~\btxlastnamefont {Zhao}}, \btxnamefont {B.~\btxlastnamefont {Xu}},
  \btxnamefont {L.~\btxlastnamefont {Zeng}}\btxandcomma {} \btxandshort {.}\
  \btxnamefont {C.~\btxlastnamefont {Wang}}\btxauthorcolon\ \btxtitlefont
  {\btxifchangecase {When spatio-temporal meet wavelets: Disentangled traffic
  forecasting via efficient spectral graph attention networks}{When
  spatio-temporal meet wavelets: Disentangled traffic forecasting via efficient
  spectral graph attention networks}}.
\newblock \Btxinshort {.}\ \btxtitlefont {2023 {IEEE} 39th International
  Conference on Data Engineering ({ICDE})}, \btxpagesshort {.}\ 517--529.
  \btxpublisherfont {IEEE}, \btxprintmonthyear{.}{4}{2023}{short}.

\bibitem {art_453}
\btxnamefont {Y.~\btxlastnamefont {Fang}}, \btxnamefont {Y.~\btxlastnamefont
  {Qin}}, \btxnamefont {H.~\btxlastnamefont {Luo}}, \btxnamefont
  {F.~\btxlastnamefont {Zhao}}\btxandcomma {} \btxandshort {.}\ \btxnamefont
  {K.~\btxlastnamefont {Zheng}}\btxauthorcolon\ \btxjtitlefont
  {\btxifchangecase {{STWave+}: A multi-scale efficient spectral graph
  attention network with long-term trends for disentangled traffic flow
  forecasting}{{STWave+}: A multi-scale efficient spectral graph attention
  network with long-term trends for disentangled traffic flow forecasting}}.
\newblock \btxjournalfont {IEEE Trans. Knowl. Data Eng.}, 36(6):2671--2685,
  \btxprintmonthyear{.}{6}{2024}{short}.

\bibitem {art_647}
\btxnamefont {Y.~\btxlastnamefont {Fang}}, \btxnamefont {K.~\btxlastnamefont
  {Ren}}, \btxnamefont {C.~\btxlastnamefont {Shan}}, \btxnamefont
  {Y.~\btxlastnamefont {Shen}}, \btxnamefont {Y.~\btxlastnamefont {Li}},
  \btxnamefont {W.~\btxlastnamefont {Zhang}}, \btxnamefont {Y.~\btxlastnamefont
  {Yu}}\btxandcomma {} \btxandshort {.}\ \btxnamefont {D.~\btxlastnamefont
  {Li}}\btxauthorcolon\ \btxjtitlefont {\btxifchangecase {Learning decomposed
  spatial relations for multi-variate time-series modeling}{Learning decomposed
  spatial relations for multi-variate time-series modeling}}.
\newblock \btxjournalfont {Proc. Conf. AAAI Artif. Intell.}, 37(6):7530--7538,
  \btxprintmonthyear{.}{6}{2023}{short}.

\bibitem {art_676}
\btxnamefont {Z.~\btxlastnamefont {Fang}}, \btxnamefont {Q.~\btxlastnamefont
  {Long}}, \btxnamefont {G.~\btxlastnamefont {Song}}\btxandcomma {}
  \btxandshort {.}\ \btxnamefont {K.~\btxlastnamefont {Xie}}\btxauthorcolon\
  \btxtitlefont {\btxifchangecase {Spatial-temporal graph {ODE} networks for
  traffic flow forecasting}{Spatial-Temporal Graph {ODE} Networks for Traffic
  Flow Forecasting}}.
\newblock \Btxinshort {.}\ \btxtitlefont {Proceedings of the 27th ACM SIGKDD
  Conference on Knowledge Discovery \& Data Mining}, KDD '21, \btxpageshort
  {.}\ 364–373, New York, NY, USA, 2021. \btxpublisherfont {Association for
  Computing Machinery}\ifbtxprintISBN {, \mbox{\btxISBN~\btxISBNfont
  {9781450383325}}}.
\newblock {\latintext \btxurlfont{https://doi.org/10.1145/3447548.3467430}}.

\bibitem {art_719}
\btxnamefont {A.~\btxlastnamefont {Feng}} \btxandshort {.}\ \btxnamefont
  {L.~\btxlastnamefont {Tassiulas}}\btxauthorcolon\ \btxtitlefont
  {\btxifchangecase {Adaptive graph spatial-temporal transformer network for
  traffic forecasting}{Adaptive graph spatial-temporal transformer network for
  traffic forecasting}}.
\newblock \Btxinshort {.}\ \btxtitlefont {Proceedings of the 31st {ACM}
  International Conference on Information \& Knowledge Management}, New York,
  NY, USA, \btxprintmonthyear{.}{10}{2022}{short}. \btxpublisherfont {ACM}.

\bibitem {art_470}
\btxnamefont {C.~\btxlastnamefont {Feng}}, \btxnamefont {C.~\btxlastnamefont
  {Liu}}\btxandcomma {} \btxandshort {.}\ \btxnamefont {D.~\btxlastnamefont
  {Jiang}}\btxauthorcolon\ \btxjtitlefont {\btxifchangecase {Root cause
  localization for wind turbines using physics guided multivariate graphical
  modeling and fault propagation analysis}{Root cause localization for wind
  turbines using physics guided multivariate graphical modeling and fault
  propagation analysis}}.
\newblock \btxjournalfont {Knowl. Based Syst.}, 295(111838):111838,
  \btxprintmonthyear{.}{7}{2024}{short}.

\bibitem {art_348}
\btxnamefont {D.~\btxlastnamefont {Feng}}, \btxnamefont {D.~\btxlastnamefont
  {Li}}, \btxnamefont {Y.~\btxlastnamefont {Zhou}}\btxandcomma {} \btxandshort
  {.}\ \btxnamefont {W.~\btxlastnamefont {Wang}}\btxauthorcolon\ \btxjtitlefont
  {\btxifchangecase {{MLFGCN}: short-term residential load forecasting via
  graph attention temporal convolution network}{{MLFGCN}: short-term
  residential load forecasting via graph attention temporal convolution
  network}}.
\newblock \btxjournalfont {Front. Neurorobot.}, 18:1461403,
  \btxprintmonthyear{.}{9}{2024}{short}.

\bibitem {TGC}
\btxnamefont {F.~\btxlastnamefont {Feng}}, \btxnamefont {X.~\btxlastnamefont
  {He}}, \btxnamefont {X.~\btxlastnamefont {Wang}}, \btxnamefont
  {C.~\btxlastnamefont {Luo}}, \btxnamefont {Y.~\btxlastnamefont
  {Liu}}\btxandcomma {} \btxandshort {.}\ \btxnamefont {T.\btxfnamespaceshort
  S. \btxlastnamefont {Chua}}\btxauthorcolon\ \btxjtitlefont {\btxifchangecase
  {Temporal relational ranking for stock prediction}{Temporal Relational
  Ranking for Stock Prediction}}.
\newblock \btxjournalfont {ACM Trans. Inf. Syst.}, 37(2),
  \btxprintmonthyear{.}{3}{2019}{short}\ifbtxprintISSN {,
  \mbox{\btxISSN~\btxISSNfont {1046-8188}}}.
\newblock {\latintext \btxurlfont{https://doi.org/10.1145/3309547}}.

\bibitem {art_446}
\btxnamefont {J.~\btxlastnamefont {Feng}}, \btxnamefont {H.~\btxlastnamefont
  {Liu}}, \btxnamefont {J.~\btxlastnamefont {Zhou}}\btxandcomma {} \btxandshort
  {.}\ \btxnamefont {Y.~\btxlastnamefont {Zhou}}\btxauthorcolon\ \btxjtitlefont
  {\btxifchangecase {A spatial-temporal aggregated graph neural network for
  docked bike-sharing demand forecasting}{A spatial-Temporal Aggregated Graph
  Neural Network for docked bike-sharing demand forecasting}}.
\newblock \btxjournalfont {ACM Trans. Knowl. Discov. Data}, 18(9):1--27,
  \btxprintmonthyear{.}{11}{2024}{short}.

\bibitem {feng2024comprehensivesurveydynamicgraph}
\btxnamefont {Z.~\btxlastnamefont {Feng}}, \btxnamefont {R.~\btxlastnamefont
  {Wang}}, \btxnamefont {T.~\btxlastnamefont {Wang}}, \btxnamefont
  {M.~\btxlastnamefont {Song}}, \btxnamefont {S.~\btxlastnamefont
  {Wu}}\btxandcomma {} \btxandshort {.}\ \btxnamefont {S.~\btxlastnamefont
  {He}}\btxauthorcolon\ \btxtitlefont {\btxifchangecase {A comprehensive survey
  of dynamic graph neural networks: Models, frameworks, benchmarks, experiments
  and challenges}{A Comprehensive Survey of Dynamic Graph Neural Networks:
  Models, Frameworks, Benchmarks, Experiments and Challenges}}, 2024.
\newblock {\latintext \btxurlfont{https://arxiv.org/abs/2405.00476}}.

\bibitem {PyTorch_Geometric}
\btxnamefont {M.~\btxlastnamefont {Fey}} \btxandshort {.}\ \btxnamefont
  {J.\btxfnamespaceshort E. \btxlastnamefont {Lenssen}}\btxauthorcolon\
  \btxtitlefont {\btxifchangecase {Fast graph representation learning with
  {PyTorch Geometric}}{Fast Graph Representation Learning with {PyTorch
  Geometric}}}.
\newblock \Btxinshort {.}\ \btxtitlefont {ICLR Workshop on Representation
  Learning on Graphs and Manifolds}, 2019.

\bibitem {art_744}
\btxnamefont {A.~\btxlastnamefont {Gandhi}}, \btxnamefont {\btxlastnamefont
  {Aakanksha}}, \btxnamefont {S.~\btxlastnamefont {Kaveri}}\btxandcomma {}
  \btxandshort {.}\ \btxnamefont {V.~\btxlastnamefont {Chaoji}}\btxauthorcolon\
  \btxtitlefont {\btxifchangecase {Spatio-temporal multi-graph networks for
  demand forecasting in online marketplaces}{Spatio-Temporal Multi-graph
  Networks for Demand Forecasting in Online Marketplaces}}.
\newblock \Btxinshort {.}\ \btxnamefont {Y.~\btxlastnamefont {Dong}},
  \btxnamefont {N.~\btxlastnamefont {Kourtellis}}, \btxnamefont
  {B.~\btxlastnamefont {Hammer}}\btxandcomma {} \btxandshort {.}\ \btxnamefont
  {J.\btxfnamespaceshort A. \btxlastnamefont {Lozano}}\ (\btxeditorsshort {.}):
  \btxtitlefont {Machine Learning and Knowledge Discovery in Databases. Applied
  Data Science Track}, \btxpagesshort {.}\ 187--203, Cham, 2021.
  \btxpublisherfont {Springer International Publishing}\ifbtxprintISBN {,
  \mbox{\btxISBN~\btxISBNfont {978-3-030-86514-6}}}.

\bibitem {art_18}
\btxnamefont {M.~\btxlastnamefont {Ganjouri}}, \btxnamefont
  {M.~\btxlastnamefont {Moattari}}, \btxnamefont {A.~\btxlastnamefont
  {Forouzantabar}}\btxandcomma {} \btxandshort {.}\ \btxnamefont
  {M.~\btxlastnamefont {Azadi}}\btxauthorcolon\ \btxjtitlefont
  {\btxifchangecase {Spatial-temporal learning structure for short-term load
  forecasting}{Spatial-temporal learning structure for short-term load
  forecasting}}.
\newblock \btxjournalfont {IET Generation, Transmission and Distribution},
  17(2):427 – 437, 2023.
\newblock {\latintext
  \btxurlfont{https://www.scopus.com/inward/record.uri?eid=2-s2.0-85143894086&doi=10.1049

\bibitem {STAN}
\btxnamefont {J.~\btxlastnamefont {Gao}}, \btxnamefont {R.~\btxlastnamefont
  {Sharma}}, \btxnamefont {C.~\btxlastnamefont {Qian}}, \btxnamefont
  {L.\btxfnamespaceshort M. \btxlastnamefont {Glass}}, \btxnamefont
  {J.~\btxlastnamefont {Spaeder}}, \btxnamefont {J.~\btxlastnamefont
  {Romberg}}, \btxnamefont {J.~\btxlastnamefont {Sun}}\btxandcomma {}
  \btxandshort {.}\ \btxnamefont {C.~\btxlastnamefont {Xiao}}\btxauthorcolon\
  \btxjtitlefont {\btxifchangecase {Stan: spatio-temporal attention network for
  pandemic prediction using real-world evidence}{STAN: spatio-temporal
  attention network for pandemic prediction using real-world evidence}}.
\newblock \btxjournalfont {Journal of the American Medical Informatics
  Association}, 28(4):733--743,
  \btxprintmonthyear{.}{01}{2021}{short}\ifbtxprintISSN {,
  \mbox{\btxISSN~\btxISSNfont {1527-974X}}}.
\newblock {\latintext \btxurlfont{https://doi.org/10.1093/jamia/ocaa322}}.

\bibitem {art_715}
\btxnamefont {Q.~\btxlastnamefont {Gao}}, \btxnamefont {Z.~\btxlastnamefont
  {Wang}}, \btxnamefont {L.~\btxlastnamefont {Huang}}, \btxnamefont
  {G.~\btxlastnamefont {Trajcevski}}, \btxnamefont {K.~\btxlastnamefont
  {Zhang}}\btxandcomma {} \btxandshort {.}\ \btxnamefont {X.~\btxlastnamefont
  {Chen}}\btxauthorcolon\ \btxtitlefont {\btxifchangecase {Enhancing dependency
  dynamics in traffic flow forecasting via graph risk bootstrap}{Enhancing
  dependency dynamics in traffic flow forecasting via graph risk bootstrap}}.
\newblock \Btxinshort {.}\ \btxtitlefont {Proceedings of the 32nd {ACM}
  International Conference on Advances in Geographic Information Systems},
  \btxpagesshort {.}\ 147--159, New York, NY, USA,
  \btxprintmonthyear{.}{10}{2024}{short}. \btxpublisherfont {ACM}.

\bibitem {art_475}
\btxnamefont {Y.~\btxlastnamefont {Gao}}, \btxnamefont {Q.~\btxlastnamefont
  {Lin}}, \btxnamefont {S.~\btxlastnamefont {Ye}}, \btxnamefont
  {Y.~\btxlastnamefont {Cheng}}, \btxnamefont {T.~\btxlastnamefont {Zhang}},
  \btxnamefont {B.~\btxlastnamefont {Liang}}\btxandcomma {} \btxandshort {.}\
  \btxnamefont {W.~\btxlastnamefont {Lu}}\btxauthorcolon\ \btxjtitlefont
  {\btxifchangecase {Outlier detection in temporal and spatial sequences via
  correlation analysis based on graph neural networks}{Outlier detection in
  temporal and spatial sequences via correlation analysis based on graph neural
  networks}}.
\newblock \btxjournalfont {Displays}, 84(102775):102775,
  \btxprintmonthyear{.}{9}{2024}{short}.

\bibitem {art_44}
\btxnamefont {Z.~\btxlastnamefont {Gao}}, \btxnamefont {Z.~\btxlastnamefont
  {Li}}, \btxnamefont {L.~\btxlastnamefont {Xu}}\btxandcomma {} \btxandshort
  {.}\ \btxnamefont {J.~\btxlastnamefont {Yu}}\btxauthorcolon\ \btxjtitlefont
  {\btxifchangecase {Dynamic adaptive spatio-temporal graph neural network for
  multi-node offshore wind speed forecasting}{Dynamic adaptive spatio-temporal
  graph neural network for multi-node offshore wind speed forecasting}}.
\newblock \btxjournalfont {Applied Soft Computing}, 141:110294,
  \btxprintmonthyear{.}{04}{2023}{short}.

\bibitem {art_36}
\btxnamefont {Z.~\btxlastnamefont {Gao}}, \btxnamefont {Z.~\btxlastnamefont
  {Li}}, \btxnamefont {J.~\btxlastnamefont {Yu}}\btxandcomma {} \btxandshort
  {.}\ \btxnamefont {L.~\btxlastnamefont {Xu}}\btxauthorcolon\ \btxjtitlefont
  {\btxifchangecase {Global spatiotemporal graph attention network for sea
  surface temperature prediction}{Global Spatiotemporal Graph Attention Network
  for Sea Surface Temperature Prediction}}.
\newblock \btxjournalfont {IEEE Geoscience and Remote Sensing Letters}, 20,
  2023.
\newblock {\latintext
  \btxurlfont{https://www.scopus.com/inward/record.uri?eid=2-s2.0-85149383269&doi=10.1109

\bibitem {art_74}
\btxnamefont {Z.~\btxlastnamefont {Gao}}, \btxnamefont {Z.~\btxlastnamefont
  {Li}}, \btxnamefont {H.~\btxlastnamefont {Zhang}}, \btxnamefont
  {J.~\btxlastnamefont {Yu}}\btxandcomma {} \btxandshort {.}\ \btxnamefont
  {L.~\btxlastnamefont {Xu}}\btxauthorcolon\ \btxjtitlefont {\btxifchangecase
  {Dynamic spatiotemporal interactive graph neural network for multivariate
  time series forecasting}{Dynamic spatiotemporal interactive graph neural
  network for multivariate time series forecasting}}.
\newblock \btxjournalfont {Knowledge-Based Systems}, 280, 2023.
\newblock {\latintext
  \btxurlfont{https://www.scopus.com/inward/record.uri?eid=2-s2.0-85172002920&doi=10.1016

\bibitem {art_395}
\btxnamefont {J.~\btxlastnamefont {Ge}}, \btxnamefont {G.~\btxlastnamefont
  {Xu}}, \btxnamefont {J.~\btxlastnamefont {Lu}}, \btxnamefont
  {X.~\btxlastnamefont {Xu}}\btxandcomma {} \btxandshort {.}\ \btxnamefont
  {X.~\btxlastnamefont {Meng}}\btxauthorcolon\ \btxjtitlefont {\btxifchangecase
  {{GraphSensor}: A graph attention network for time-series
  sensor}{{GraphSensor}: A graph attention network for time-series sensor}}.
\newblock \btxjournalfont {Electronics (Basel)}, 13(12):2290,
  \btxprintmonthyear{.}{6}{2024}{short}.

\bibitem {ST-MGCN}
\btxnamefont {X.~\btxlastnamefont {Geng}}, \btxnamefont {Y.~\btxlastnamefont
  {Li}}, \btxnamefont {L.~\btxlastnamefont {Wang}}, \btxnamefont
  {L.~\btxlastnamefont {Zhang}}, \btxnamefont {Q.~\btxlastnamefont {Yang}},
  \btxnamefont {J.~\btxlastnamefont {Ye}}\btxandcomma {} \btxandshort {.}\
  \btxnamefont {Y.~\btxlastnamefont {Liu}}\btxauthorcolon\ \btxjtitlefont
  {\btxifchangecase {Spatiotemporal multi-graph convolution network for
  ride-hailing demand forecasting}{Spatiotemporal Multi-Graph Convolution
  Network for Ride-Hailing Demand Forecasting}}.
\newblock \btxjournalfont {Proceedings of the AAAI Conference on Artificial
  Intelligence}, 33(01):3656--3663, \btxprintmonthyear{.}{Jul.}{2019}{short}.
\newblock {\latintext
  \btxurlfont{https://ojs.aaai.org/index.php/AAAI/article/view/4247}}.

\bibitem {art_40}
\btxnamefont {X.~\btxlastnamefont {Geng}}, \btxnamefont {L.~\btxlastnamefont
  {Xu}}, \btxnamefont {X.~\btxlastnamefont {He}}\btxandcomma {} \btxandshort
  {.}\ \btxnamefont {J.~\btxlastnamefont {Yu}}\btxauthorcolon\ \btxjtitlefont
  {\btxifchangecase {Graph optimization neural network with spatio-temporal
  correlation learning for multi-node offshore wind speed forecasting}{Graph
  optimization neural network with spatio-temporal correlation learning for
  multi-node offshore wind speed forecasting}}.
\newblock \btxjournalfont {Renewable Energy}, 180:1014 – 1025, 2021.
\newblock {\latintext
  \btxurlfont{https://www.scopus.com/inward/record.uri?eid=2-s2.0-85114689825&doi=10.1016

\bibitem {Multi-LSTM}
\btxnamefont {A.~\btxlastnamefont {Ghaderi}}, \btxnamefont
  {B.\btxfnamespaceshort M. \btxlastnamefont {Sanandaji}}\btxandcomma {}
  \btxandshort {.}\ \btxnamefont {F.~\btxlastnamefont
  {Ghaderi}}\btxauthorcolon\ \btxtitlefont {\btxifchangecase {Deep forecast:
  Deep learning-based spatio-temporal forecasting}{Deep Forecast: Deep
  Learning-based Spatio-Temporal Forecasting}}.
\newblock \Btxinshort {.}\ \btxtitlefont {The 34th International Conference on
  Machine Learning (ICML), Time series Workshop}, 2017.

\bibitem {art_338}
\btxnamefont {N.~\btxlastnamefont {Giamarelos}} \btxandshort {.}\ \btxnamefont
  {E.\btxfnamespaceshort N. \btxlastnamefont {Zois}}\btxauthorcolon\
  \btxjtitlefont {\btxifchangecase {Boosting short term electric load
  forecasting of high \& medium voltage substations with visibility graphs and
  graph neural networks}{Boosting short term electric load forecasting of high
  \& medium voltage substations with visibility graphs and graph neural
  networks}}.
\newblock \btxjournalfont {Sustain. Energy Grids Netw.}, 38(101304):101304,
  \btxprintmonthyear{.}{6}{2024}{short}.

\bibitem {oversmoothing}
\btxnamefont {J.\btxfnamespaceshort H. \btxlastnamefont {Giraldo}},
  \btxnamefont {K.~\btxlastnamefont {Skianis}}, \btxnamefont
  {T.~\btxlastnamefont {Bouwmans}}\btxandcomma {} \btxandshort {.}\
  \btxnamefont {F.\btxfnamespaceshort D. \btxlastnamefont
  {Malliaros}}\btxauthorcolon\ \btxtitlefont {\btxifchangecase {On the
  trade-off between over-smoothing and over-squashing in deep graph neural
  networks}{On the Trade-off between Over-smoothing and Over-squashing in Deep
  Graph Neural Networks}}.
\newblock \Btxinshort {.}\ \btxtitlefont {Proceedings of the 32nd ACM
  International Conference on Information and Knowledge Management}, CIKM '23,
  \btxpageshort {.}\ 566–576, New York, NY, USA, 2023. \btxpublisherfont
  {Association for Computing Machinery}\ifbtxprintISBN {,
  \mbox{\btxISBN~\btxISBNfont {9798400701245}}}.
\newblock {\latintext \btxurlfont{https://doi.org/10.1145/3583780.3614997}}.

\bibitem {art_360}
\btxnamefont {M.~\btxlastnamefont {Gong}}, \btxnamefont {Y.~\btxlastnamefont
  {Zhang}}, \btxnamefont {J.~\btxlastnamefont {Li}}\btxandcomma {} \btxandshort
  {.}\ \btxnamefont {L.~\btxlastnamefont {Chen}}\btxauthorcolon\ \btxjtitlefont
  {\btxifchangecase {Dynamic spatial--temporal model for carbon emission
  forecasting}{Dynamic spatial--temporal model for carbon emission
  forecasting}}.
\newblock \btxjournalfont {J. Clean. Prod.}, 463(142581):142581,
  \btxprintmonthyear{.}{7}{2024}{short}.

\bibitem {gori_aggiunto}
\btxnamefont {M.~\btxlastnamefont {Gori}}, \btxnamefont {G.~\btxlastnamefont
  {Monfardini}}\btxandcomma {} \btxandshort {.}\ \btxnamefont
  {F.~\btxlastnamefont {Scarselli}}\btxauthorcolon\ \btxtitlefont
  {\btxifchangecase {A new model for learning in graph domains}{A new model for
  learning in graph domains}}.
\newblock \Btxinshort {.}\ \btxtitlefont {Proceedings. 2005 IEEE International
  Joint Conference on Neural Networks, 2005.}, \btxvolumeshort
  {.}~\btxvolumefont {2}, \btxpagesshort {.}\ 729--734 vol. 2, 2005.

\bibitem {art_717}
\btxnamefont {X.~\btxlastnamefont {Gou}} \btxandshort {.}\ \btxnamefont
  {X.~\btxlastnamefont {Zhang}}\btxauthorcolon\ \btxtitlefont {\btxifchangecase
  {Telecommunication traffic forecasting via multi-task
  learning}{Telecommunication traffic forecasting via multi-task learning}}.
\newblock \Btxinshort {.}\ \btxtitlefont {Proceedings of the Sixteenth {ACM}
  International Conference on Web Search and Data Mining}, New York, NY, USA,
  \btxprintmonthyear{.}{2}{2023}{short}. \btxpublisherfont {ACM}.

\bibitem {BiLSTM}
\btxnamefont {A.~\btxlastnamefont {Graves}} \btxandshort {.}\ \btxnamefont
  {J.~\btxlastnamefont {Schmidhuber}}\btxauthorcolon\ \btxtitlefont
  {\btxifchangecase {Framewise phoneme classification with bidirectional lstm
  networks}{Framewise phoneme classification with bidirectional LSTM
  networks}}.
\newblock \Btxinshort {.}\ \btxtitlefont {Proceedings. 2005 IEEE International
  Joint Conference on Neural Networks, 2005.}, \btxvolumeshort
  {.}~\btxvolumefont {4}, \btxpagesshort {.}\ 2047--2052 vol. 4, 2005.

\bibitem {node2vec}
\btxnamefont {A.~\btxlastnamefont {Grover}} \btxandshort {.}\ \btxnamefont
  {J.~\btxlastnamefont {Leskovec}}\btxauthorcolon\ \btxtitlefont
  {\btxifchangecase {node2vec: Scalable feature learning for
  networks}{node2vec: Scalable Feature Learning for Networks}}.
\newblock \Btxinshort {.}\ \btxtitlefont {Proceedings of the 22nd ACM SIGKDD
  International Conference on Knowledge Discovery and Data Mining}, KDD '16,
  \btxpageshort {.}\ 855–864, New York, NY, USA, 2016. \btxpublisherfont
  {Association for Computing Machinery}\ifbtxprintISBN {,
  \mbox{\btxISBN~\btxISBNfont {9781450342322}}}.
\newblock {\latintext \btxurlfont{https://doi.org/10.1145/2939672.2939754}}.

\bibitem {art_117}
\btxnamefont {J.~\btxlastnamefont {Gu}}, \btxnamefont {Z.~\btxlastnamefont
  {Jia}}, \btxnamefont {T.~\btxlastnamefont {Cai}}, \btxnamefont
  {X.~\btxlastnamefont {Song}}\btxandcomma {} \btxandshort {.}\ \btxnamefont
  {A.~\btxlastnamefont {Mahmood}}\btxauthorcolon\ \btxjtitlefont
  {\btxifchangecase {Dynamic correlation adjacency-matrix-based graph neural
  networks for traffic flow prediction}{Dynamic Correlation
  Adjacency-Matrix-Based Graph Neural Networks for Traffic Flow Prediction}}.
\newblock \btxjournalfont {Sensors}, 23(6), 2023.
\newblock {\latintext
  \btxurlfont{https://www.scopus.com/inward/record.uri?eid=2-s2.0-85151047271&doi=10.3390

\bibitem {art_130}
\btxnamefont {Y.~\btxlastnamefont {Gu}} \btxandshort {.}\ \btxnamefont
  {L.~\btxlastnamefont {Deng}}\btxauthorcolon\ \btxjtitlefont {\btxifchangecase
  {Stagcn: Spatial–temporal attention graph convolution network for traffic
  forecasting}{STAGCN: Spatial–Temporal Attention Graph Convolution Network
  for Traffic Forecasting}}.
\newblock \btxjournalfont {Mathematics}, 10(9), 2022.
\newblock {\latintext
  \btxurlfont{https://www.scopus.com/inward/record.uri?eid=2-s2.0-85131314499&doi=10.3390

\bibitem {art_386}
\btxnamefont {H.~\btxlastnamefont {Gui}}, \btxnamefont {G.~\btxlastnamefont
  {Li}}, \btxnamefont {X.~\btxlastnamefont {Tang}}\btxandcomma {} \btxandshort
  {.}\ \btxnamefont {J.~\btxlastnamefont {Lu}}\btxauthorcolon\ \btxjtitlefont
  {\btxifchangecase {{CATodyNet}: Cross-attention temporal dynamic graph neural
  network for multivariate time series classification}{{CATodyNet}:
  Cross-attention temporal dynamic graph neural network for multivariate time
  series classification}}.
\newblock \btxjournalfont {Knowl. Based Syst.}, 300(112210):112210,
  \btxprintmonthyear{.}{9}{2024}{short}.

\bibitem {art_481}
\btxnamefont {H.~\btxlastnamefont {Guo}}, \btxnamefont {Z.~\btxlastnamefont
  {Zhou}}, \btxnamefont {D.~\btxlastnamefont {Zhao}}\btxandcomma {}
  \btxandshort {.}\ \btxnamefont {W.~\btxlastnamefont
  {Gaaloul}}\btxauthorcolon\ \btxjtitlefont {\btxifchangecase {{EGNN}:
  Energy-efficient anomaly detection for {IoT} multivariate time series data
  using graph neural network}{{EGNN}: Energy-efficient anomaly detection for
  {IoT} multivariate time series data using graph neural network}}.
\newblock \btxjournalfont {Future Gener. Comput. Syst.}, 151:45--56,
  \btxprintmonthyear{.}{2}{2024}{short}.

\bibitem {HGCN}
\btxnamefont {K.~\btxlastnamefont {Guo}}, \btxnamefont {Y.~\btxlastnamefont
  {Hu}}, \btxnamefont {Y.~\btxlastnamefont {Sun}}, \btxnamefont
  {S.~\btxlastnamefont {Qian}}, \btxnamefont {J.~\btxlastnamefont
  {Gao}}\btxandcomma {} \btxandshort {.}\ \btxnamefont {B.~\btxlastnamefont
  {Yin}}\btxauthorcolon\ \btxjtitlefont {\btxifchangecase {Hierarchical graph
  convolution network for traffic forecasting}{Hierarchical Graph Convolution
  Network for Traffic Forecasting}}.
\newblock \btxjournalfont {Proceedings of the AAAI Conference on Artificial
  Intelligence}, 35(1):151–159,
  \btxprintmonthyear{.}{5}{2021}{short}\ifbtxprintISSN {,
  \mbox{\btxISSN~\btxISSNfont {2159-5399}}}.
\newblock {\latintext \btxurlfont{http://dx.doi.org/10.1609/aaai.v35i1.16088}}.

\bibitem {ASTGCN}
\btxnamefont {S.~\btxlastnamefont {Guo}}, \btxnamefont {Y.~\btxlastnamefont
  {Lin}}, \btxnamefont {N.~\btxlastnamefont {Feng}}, \btxnamefont
  {C.~\btxlastnamefont {Song}}\btxandcomma {} \btxandshort {.}\ \btxnamefont
  {H.~\btxlastnamefont {Wan}}\btxauthorcolon\ \btxjtitlefont {\btxifchangecase
  {Attention based spatial-temporal graph convolutional networks for traffic
  flow forecasting}{Attention Based Spatial-Temporal Graph Convolutional
  Networks for Traffic Flow Forecasting}}.
\newblock \btxjournalfont {Proceedings of the AAAI Conference on Artificial
  Intelligence}, 33(01):922–929,
  \btxprintmonthyear{.}{7}{2019}{short}\ifbtxprintISSN {,
  \mbox{\btxISSN~\btxISSNfont {2159-5399}}}.
\newblock {\latintext
  \btxurlfont{http://dx.doi.org/10.1609/aaai.v33i01.3301922}}.

\bibitem {ASTGNN}
\btxnamefont {S.~\btxlastnamefont {Guo}}, \btxnamefont {Y.~\btxlastnamefont
  {Lin}}, \btxnamefont {H.~\btxlastnamefont {Wan}}, \btxnamefont
  {X.~\btxlastnamefont {Li}}\btxandcomma {} \btxandshort {.}\ \btxnamefont
  {G.~\btxlastnamefont {Cong}}\btxauthorcolon\ \btxjtitlefont {\btxifchangecase
  {Learning dynamics and heterogeneity of spatial-temporal graph data for
  traffic forecasting}{Learning Dynamics and Heterogeneity of Spatial-Temporal
  Graph Data for Traffic Forecasting}}.
\newblock \btxjournalfont {IEEE Transactions on Knowledge and Data
  Engineering}, 34(11):5415--5428, 2022.

\bibitem {art_79}
\btxnamefont {T.~\btxlastnamefont {Guo}}, \btxnamefont {F.~\btxlastnamefont
  {Hou}}, \btxnamefont {Y.~\btxlastnamefont {Pang}}, \btxnamefont
  {X.~\btxlastnamefont {Jia}}, \btxnamefont {Z.~\btxlastnamefont
  {Wang}}\btxandcomma {} \btxandshort {.}\ \btxnamefont {R.~\btxlastnamefont
  {Wang}}\btxauthorcolon\ \btxjtitlefont {\btxifchangecase {Learning and
  integration of adaptive hybrid graph structures for multivariate time series
  forecasting}{Learning and integration of adaptive hybrid graph structures for
  multivariate time series forecasting}}.
\newblock \btxjournalfont {Information Sciences}, 648, 2023.
\newblock {\latintext
  \btxurlfont{https://www.scopus.com/inward/record.uri?eid=2-s2.0-85168803579&doi=10.1016

\bibitem {art_412}
\btxnamefont {W.~\btxlastnamefont {Guo}} \btxandshort {.}\ \btxnamefont
  {Y.~\btxlastnamefont {Wang}}\btxauthorcolon\ \btxjtitlefont {\btxifchangecase
  {Convolutional gated recurrent unit-driven multidimensional dynamic graph
  neural network for subject-independent emotion recognition}{Convolutional
  gated recurrent unit-driven multidimensional dynamic graph neural network for
  subject-independent emotion recognition}}.
\newblock \btxjournalfont {Expert Syst. Appl.}, 238(121889):121889,
  \btxprintmonthyear{.}{3}{2024}{short}.

\bibitem {art_116}
\btxnamefont {X.~\btxlastnamefont {Guo}}, \btxnamefont {X.~\btxlastnamefont
  {Kong}}, \btxnamefont {W.~\btxlastnamefont {Xing}}, \btxnamefont
  {X.~\btxlastnamefont {Wei}}, \btxnamefont {J.~\btxlastnamefont
  {Zhang}}\btxandcomma {} \btxandshort {.}\ \btxnamefont {W.~\btxlastnamefont
  {Lu}}\btxauthorcolon\ \btxjtitlefont {\btxifchangecase {Adaptive graph
  generation based on generalized pagerank graph neural network for traffic
  flow forecasting}{Adaptive graph generation based on generalized pagerank
  graph neural network for traffic flow forecasting}}.
\newblock \btxjournalfont {Applied Intelligence}, 53(24):30971 – 30986, 2023.
\newblock {\latintext
  \btxurlfont{https://www.scopus.com/inward/record.uri?eid=2-s2.0-85178201409&doi=10.1007

\bibitem {art_678}
\btxnamefont {Z.~\btxlastnamefont {Guo}}, \btxnamefont {H.~\btxlastnamefont
  {Liu}}, \btxnamefont {L.~\btxlastnamefont {Zhang}}, \btxnamefont
  {Q.~\btxlastnamefont {Zhang}}, \btxnamefont {H.~\btxlastnamefont
  {Zhu}}\btxandcomma {} \btxandshort {.}\ \btxnamefont {H.~\btxlastnamefont
  {Xiong}}\btxauthorcolon\ \btxtitlefont {\btxifchangecase {Talent
  demand-supply joint prediction with dynamic heterogeneous graph enhanced
  meta-learning}{Talent demand-supply joint prediction with dynamic
  heterogeneous graph enhanced meta-learning}}.
\newblock \Btxinshort {.}\ \btxtitlefont {Proceedings of the 28th {ACM}
  {SIGKDD} Conference on Knowledge Discovery and Data Mining}, \btxvolumeshort
  {.}~\btxvolumefont {12}, \btxpagesshort {.}\ 2957--2967, New York, NY, USA,
  \btxprintmonthyear{.}{8}{2022}{short}. \btxpublisherfont {ACM}.

\bibitem {art_681}
\btxnamefont {M.~\btxlastnamefont {Gupta}}, \btxnamefont {H.~\btxlastnamefont
  {Kodamana}}\btxandcomma {} \btxandshort {.}\ \btxnamefont
  {S.~\btxlastnamefont {Ranu}}\btxauthorcolon\ \btxtitlefont {\btxifchangecase
  {Frigate: Frugal spatio-temporal forecasting on road networks}{Frigate:
  Frugal spatio-temporal forecasting on road networks}}.
\newblock \Btxinshort {.}\ \btxtitlefont {Proceedings of the 29th {ACM}
  {SIGKDD} Conference on Knowledge Discovery and Data Mining}, \btxpagesshort
  {.}\ 649--660, New York, NY, USA, \btxprintmonthyear{.}{8}{2023}{short}.
  \btxpublisherfont {ACM}.

\bibitem {art_728}
\btxnamefont {A.~\btxlastnamefont {Hajisafi}}, \btxnamefont
  {H.~\btxlastnamefont {Lin}}, \btxnamefont {S.~\btxlastnamefont {Shaham}},
  \btxnamefont {H.~\btxlastnamefont {Hu}}, \btxnamefont {M.\btxfnamespaceshort
  D. \btxlastnamefont {Siampou}}, \btxnamefont {Y.\btxfnamespaceshort Y.
  \btxlastnamefont {Chiang}}\btxandcomma {} \btxandshort {.}\ \btxnamefont
  {C.~\btxlastnamefont {Shahabi}}\btxauthorcolon\ \btxtitlefont
  {\btxifchangecase {Learning dynamic graphs from all contextual information
  for accurate point-of-interest visit forecasting}{Learning dynamic graphs
  from all contextual information for accurate point-of-interest visit
  forecasting}}.
\newblock \Btxinshort {.}\ \btxtitlefont {Proceedings of the 31st {ACM}
  International Conference on Advances in Geographic Information Systems}, New
  York, NY, USA, \btxprintmonthyear{.}{11}{2023}{short}. \btxpublisherfont
  {ACM}.

\bibitem {modulation_classification_definition}
\btxnamefont {J.~\btxlastnamefont {Hamkins}} \btxandshort {.}\ \btxnamefont
  {M.\btxfnamespaceshort K. \btxlastnamefont {Simon}}\btxauthorcolon\
  \btxtitlefont {\btxifchangecase {Modulation classification}{Modulation
  Classification}}, \btxprintmonthyear{.}{5}{2006}{short}\ifbtxprintISBN {,
  \mbox{\btxISBN~\btxISBNfont {9780470087800}}}.
\newblock {\latintext
  \btxurlfont{http://dx.doi.org/10.1002/9780470087800.ch9}}.

\bibitem {art_388}
\btxnamefont {H.~\btxlastnamefont {Han}}, \btxnamefont {H.~\btxlastnamefont
  {Neira-Molina}}, \btxnamefont {A.~\btxlastnamefont {Khan}}, \btxnamefont
  {M.~\btxlastnamefont {Fang}}, \btxnamefont {H.\btxfnamespaceshort A.
  \btxlastnamefont {Mahmoud}}, \btxnamefont {E.\btxfnamespaceshort M.
  \btxlastnamefont {Awwad}}, \btxnamefont {B.~\btxlastnamefont
  {Ahmed}}\btxandcomma {} \btxandshort {.}\ \btxnamefont {Y.\btxfnamespaceshort
  Y. \btxlastnamefont {Ghadi}}\btxauthorcolon\ \btxjtitlefont {\btxifchangecase
  {Advanced series decomposition with a gated recurrent unit and graph
  convolutional neural network for non-stationary data patterns}{Advanced
  series decomposition with a gated recurrent unit and graph convolutional
  neural network for non-stationary data patterns}}.
\newblock \btxjournalfont {J. Cloud Comput. Adv. Syst. Appl.}, 13(1),
  \btxprintmonthyear{.}{1}{2024}{short}.

\bibitem {art_447}
\btxnamefont {J.~\btxlastnamefont {Han}}, \btxnamefont {W.~\btxlastnamefont
  {Zhang}}, \btxnamefont {H.~\btxlastnamefont {Liu}}, \btxnamefont
  {T.~\btxlastnamefont {Tao}}, \btxnamefont {N.~\btxlastnamefont
  {Tan}}\btxandcomma {} \btxandshort {.}\ \btxnamefont {H.~\btxlastnamefont
  {Xiong}}\btxauthorcolon\ \btxjtitlefont {\btxifchangecase {{BigST}: Linear
  complexity spatio-temporal graph neural network for traffic forecasting on
  large-scale road networks}{{BigST}: Linear complexity spatio-Temporal Graph
  Neural Network for traffic forecasting on large-scale road networks}}.
\newblock \btxjournalfont {Proceedings VLDB Endowment}, 17(5):1081--1090,
  \btxprintmonthyear{.}{1}{2024}{short}.

\bibitem {art_663}
\btxnamefont {L.~\btxlastnamefont {Han}}, \btxnamefont {B.~\btxlastnamefont
  {Du}}, \btxnamefont {L.~\btxlastnamefont {Sun}}, \btxnamefont
  {Y.~\btxlastnamefont {Fu}}, \btxnamefont {Y.~\btxlastnamefont
  {Lv}}\btxandcomma {} \btxandshort {.}\ \btxnamefont {H.~\btxlastnamefont
  {Xiong}}\btxauthorcolon\ \btxtitlefont {\btxifchangecase {Dynamic and
  multi-faceted spatio-temporal deep learning for traffic speed
  forecasting}{Dynamic and Multi-faceted Spatio-temporal Deep Learning for
  Traffic Speed Forecasting}}.
\newblock \Btxinshort {.}\ \btxtitlefont {Proceedings of the 27th ACM SIGKDD
  Conference on Knowledge Discovery \& Data Mining}, KDD '21, \btxpageshort
  {.}\ 547–555, New York, NY, USA, 2021. \btxpublisherfont {Association for
  Computing Machinery}\ifbtxprintISBN {, \mbox{\btxISBN~\btxISBNfont
  {9781450383325}}}.
\newblock {\latintext \btxurlfont{https://doi.org/10.1145/3447548.3467275}}.

\bibitem {art_648}
\btxnamefont {S.~\btxlastnamefont {Han}} \btxandshort {.}\ \btxnamefont
  {S.\btxfnamespaceshort S. \btxlastnamefont {Woo}}\btxauthorcolon\
  \btxtitlefont {\btxifchangecase {Learning sparse latent graph representations
  for anomaly detection in multivariate time series}{Learning Sparse Latent
  Graph Representations for Anomaly Detection in Multivariate Time Series}}.
\newblock \Btxinshort {.}\ \btxtitlefont {Proceedings of the 28th ACM SIGKDD
  Conference on Knowledge Discovery and Data Mining}, KDD '22, \btxpageshort
  {.}\ 2977–2986, New York, NY, USA, 2022. \btxpublisherfont {Association for
  Computing Machinery}\ifbtxprintISBN {, \mbox{\btxISBN~\btxISBNfont
  {9781450393850}}}.
\newblock {\latintext \btxurlfont{https://doi.org/10.1145/3534678.3539117}}.

\bibitem {art_71}
\btxnamefont {X.~\btxlastnamefont {Han}}, \btxnamefont {Y.~\btxlastnamefont
  {Huang}}, \btxnamefont {Z.~\btxlastnamefont {Pan}}, \btxnamefont
  {W.~\btxlastnamefont {Li}}, \btxnamefont {Y.~\btxlastnamefont
  {Hu}}\btxandcomma {} \btxandshort {.}\ \btxnamefont {G.~\btxlastnamefont
  {Lin}}\btxauthorcolon\ \btxjtitlefont {\btxifchangecase {Multi-task time
  series forecasting based on graph neural networks}{Multi-Task Time Series
  Forecasting Based on Graph Neural Networks}}.
\newblock \btxjournalfont {Entropy}, 25(8), 2023.
\newblock {\latintext
  \btxurlfont{https://www.scopus.com/inward/record.uri?eid=2-s2.0-85168860312&doi=10.3390

\bibitem {art_455}
\btxnamefont {Y.~\btxlastnamefont {Han}}, \btxnamefont {X.~\btxlastnamefont
  {Liu}}, \btxnamefont {C.~\btxlastnamefont {Guo}}, \btxnamefont
  {H.~\btxlastnamefont {Wu}}, \btxnamefont {M.~\btxlastnamefont
  {Liu}}\btxandcomma {} \btxandshort {.}\ \btxnamefont {Z.~\btxlastnamefont
  {Geng}}\btxauthorcolon\ \btxjtitlefont {\btxifchangecase {Improved pearson
  correlation coefficient-based graph neural network for dynamic soft sensor of
  polypropylene industries}{Improved Pearson correlation coefficient-based
  graph neural network for dynamic soft sensor of polypropylene industries}}.
\newblock \btxjournalfont {Ind. Eng. Chem. Res.},
  \btxprintmonthyear{.}{12}{2024}{short}.

\bibitem {ref_anomaly_detection_1}
\btxnamefont {D.\btxfnamespaceshort M. \btxlastnamefont
  {Hawkins}}\btxauthorcolon\ \btxtitlefont {Identification of Outliers}.
\newblock \btxpublisherfont {Springer Netherlands}, 1980\ifbtxprintISBN {,
  \mbox{\btxISBN~\btxISBNfont {9789401539944}}}.
\newblock {\latintext
  \btxurlfont{http://dx.doi.org/10.1007/978-94-015-3994-4}}.

\bibitem {art_20}
\btxnamefont {H.~\btxlastnamefont {He}}, \btxnamefont {F.~\btxlastnamefont
  {Fu}}\btxandcomma {} \btxandshort {.}\ \btxnamefont {D.~\btxlastnamefont
  {Luo}}\btxauthorcolon\ \btxjtitlefont {\btxifchangecase {Multiplex parallel
  gat-alstm: A novel spatial-temporal learning model for multi-sites wind power
  collaborative forecasting}{Multiplex parallel GAT-ALSTM: A novel
  spatial-temporal learning model for multi-sites wind power collaborative
  forecasting}}.
\newblock \btxjournalfont {Frontiers in Energy Research}, 10, 2022.
\newblock {\latintext
  \btxurlfont{https://www.scopus.com/inward/record.uri?eid=2-s2.0-85137693094&doi=10.3389

\bibitem {art_691}
\btxnamefont {J.~\btxlastnamefont {He}}, \btxnamefont {J.~\btxlastnamefont
  {Ji}}\btxandcomma {} \btxandshort {.}\ \btxnamefont {M.~\btxlastnamefont
  {Lei}}\btxauthorcolon\ \btxtitlefont {\btxifchangecase {Spatio-temporal
  transformer network with physical knowledge distillation for weather
  forecasting}{Spatio-temporal transformer network with physical knowledge
  distillation for weather forecasting}}.
\newblock \Btxinshort {.}\ \btxtitlefont {Proceedings of the 33rd {ACM}
  International Conference on Information and Knowledge Management},
  \btxpagesshort {.}\ 819--828, New York, NY, USA,
  \btxprintmonthyear{.}{10}{2024}{short}. \btxpublisherfont {ACM}.

\bibitem {art_664}
\btxnamefont {S.~\btxlastnamefont {He}} \btxandshort {.}\ \btxnamefont
  {K.\btxfnamespaceshort G. \btxlastnamefont {Shin}}\btxauthorcolon\
  \btxtitlefont {\btxifchangecase {Dynamic flow distribution prediction for
  urban dockless e-scooter sharing reconfiguration}{Dynamic flow distribution
  prediction for urban dockless E-scooter sharing reconfiguration}}.
\newblock \Btxinshort {.}\ \btxtitlefont {Proceedings of The Web Conference
  2020}, New York, NY, USA, \btxprintmonthyear{.}{4}{2020}{short}.
  \btxpublisherfont {ACM}.

\bibitem {art_668}
\btxnamefont {S.~\btxlastnamefont {He}} \btxandshort {.}\ \btxnamefont
  {K.\btxfnamespaceshort G. \btxlastnamefont {Shin}}\btxauthorcolon\
  \btxtitlefont {\btxifchangecase {Towards fine-grained flow forecasting: A
  graph attention approach for bike sharing systems}{Towards fine-grained flow
  forecasting: A graph attention approach for bike sharing systems}}.
\newblock \Btxinshort {.}\ \btxtitlefont {Proceedings of The Web Conference
  2020}, New York, NY, USA, \btxprintmonthyear{.}{4}{2020}{short}.
  \btxpublisherfont {ACM}.

\bibitem {art_489}
\btxnamefont {Y.~\btxlastnamefont {He}}, \btxnamefont {Y.~\btxlastnamefont
  {Bian}}, \btxnamefont {X.~\btxlastnamefont {Ding}}, \btxnamefont
  {B.~\btxlastnamefont {Wu}}, \btxnamefont {J.~\btxlastnamefont {Guan}},
  \btxnamefont {J.~\btxlastnamefont {Zhang}}\btxandcomma {} \btxandshort {.}\
  \btxnamefont {S.~\btxlastnamefont {Zhou}}\btxauthorcolon\ \btxjtitlefont
  {\btxifchangecase {Variate associated domain adaptation for unsupervised
  multivariate time series anomaly detection}{Variate associated domain
  adaptation for unsupervised multivariate Time Series Anomaly Detection}}.
\newblock \btxjournalfont {ACM Trans. Knowl. Discov. Data}, 18(8):1--24,
  \btxprintmonthyear{.}{9}{2024}{short}.

\bibitem {art_396}
\btxnamefont {Z.~\btxlastnamefont {He}}, \btxnamefont {L.~\btxlastnamefont
  {Xu}}, \btxnamefont {J.~\btxlastnamefont {Yu}}\btxandcomma {} \btxandshort
  {.}\ \btxnamefont {X.~\btxlastnamefont {Wu}}\btxauthorcolon\ \btxjtitlefont
  {\btxifchangecase {Dynamic multi-fusion spatio-temporal graph neural network
  for multivariate time series forecasting}{Dynamic multi-fusion
  spatio-temporal graph neural network for multivariate time series
  forecasting}}.
\newblock \btxjournalfont {Expert Syst. Appl.}, 241(122729):122729,
  \btxprintmonthyear{.}{5}{2024}{short}.

\bibitem {art_90}
\btxnamefont {Z.~\btxlastnamefont {He}}, \btxnamefont {C.~\btxlastnamefont
  {Zhao}}\btxandcomma {} \btxandshort {.}\ \btxnamefont {Y.~\btxlastnamefont
  {Huang}}\btxauthorcolon\ \btxjtitlefont {\btxifchangecase {Multivariate time
  series deep spatiotemporal forecasting with graph neural
  network}{Multivariate Time Series Deep Spatiotemporal Forecasting with Graph
  Neural Network}}.
\newblock \btxjournalfont {Applied Sciences (Switzerland)}, 12(11), 2022.
\newblock {\latintext
  \btxurlfont{https://www.scopus.com/inward/record.uri?eid=2-s2.0-85132031664&doi=10.3390

\bibitem {art_751}
\btxnamefont {H.~\btxlastnamefont {Hojjati}}, \btxnamefont {M.~\btxlastnamefont
  {Sadeghi}}\btxandcomma {} \btxandshort {.}\ \btxnamefont {N.~\btxlastnamefont
  {Armanfard}}\btxauthorcolon\ \btxtitlefont {\btxifchangecase {Multivariate
  time-series anomaly detection with temporal self-supervision and graphs:
  Application to vehicle failure prediction}{Multivariate Time-Series Anomaly
  Detection with Temporal Self-supervision and Graphs: Application to Vehicle
  Failure Prediction}}.
\newblock \Btxinshort {.}\ \btxnamefont {G.~\btxlastnamefont
  {De~Francisci~Morales}}, \btxnamefont {C.~\btxlastnamefont {Perlich}},
  \btxnamefont {N.~\btxlastnamefont {Ruchansky}}, \btxnamefont
  {N.~\btxlastnamefont {Kourtellis}}, \btxnamefont {E.~\btxlastnamefont
  {Baralis}}\btxandcomma {} \btxandshort {.}\ \btxnamefont {F.~\btxlastnamefont
  {Bonchi}}\ (\btxeditorsshort {.}): \btxtitlefont {Machine Learning and
  Knowledge Discovery in Databases: Applied Data Science and Demo Track},
  \btxpagesshort {.}\ 242--259, Cham, 2023. \btxpublisherfont {Springer Nature
  Switzerland}\ifbtxprintISBN {, \mbox{\btxISBN~\btxISBNfont
  {978-3-031-43430-3}}}.

\bibitem {art_407}
\btxnamefont {X.~\btxlastnamefont {Hong}}, \btxnamefont {J.~\btxlastnamefont
  {Hu}}, \btxnamefont {T.~\btxlastnamefont {Xu}}, \btxnamefont
  {X.~\btxlastnamefont {Ren}}, \btxnamefont {F.~\btxlastnamefont {Wu}},
  \btxnamefont {X.~\btxlastnamefont {Ma}}\btxandcomma {} \btxandshort {.}\
  \btxnamefont {W.~\btxlastnamefont {Li}}\btxauthorcolon\ \btxjtitlefont
  {\btxifchangecase {{MagNet}: Multilevel dynamic wavelet graph neural network
  for multivariate time series classification}{{MagNet}: Multilevel Dynamic
  Wavelet Graph Neural Network for multivariate Time Series Classification}}.
\newblock \btxjournalfont {ACM Trans. Knowl. Discov. Data}, 19(1):1--22,
  \btxprintmonthyear{.}{1}{2025}{short}.

\bibitem {SeFT}
\btxnamefont {M.~\btxlastnamefont {Horn}}, \btxnamefont {M.~\btxlastnamefont
  {Moor}}, \btxnamefont {C.~\btxlastnamefont {Bock}}, \btxnamefont
  {B.~\btxlastnamefont {Rieck}}\btxandcomma {} \btxandshort {.}\ \btxnamefont
  {K.~\btxlastnamefont {Borgwardt}}\btxauthorcolon\ \btxtitlefont
  {\btxifchangecase {Set functions for time series}{Set Functions for Time
  Series}}.
\newblock \Btxinshort {.}\ \btxnamefont {H.\btxfnamespaceshort D.
  \btxlastnamefont {III}} \btxandshort {.}\ \btxnamefont {A.~\btxlastnamefont
  {Singh}}\ (\btxeditorsshort {.}): \btxtitlefont {Proceedings of the 37th
  International Conference on Machine Learning}, \btxvolumeshort {.}\
  \btxvolumefont {119} \btxofseriesshort {.}\ \btxtitlefont {Proceedings of
  Machine Learning Research}, \btxpagesshort {.}\ 4353--4363. \btxpublisherfont
  {PMLR}, \btxprintmonthyear{.}{13--18 Jul}{2020}{short}.
\newblock {\latintext
  \btxurlfont{https://proceedings.mlr.press/v119/horn20a.html}}.

\bibitem {FinGAT}
\btxnamefont {Y.\btxfnamespaceshort L. \btxlastnamefont {Hsu}}, \btxnamefont
  {Y.\btxfnamespaceshort C. \btxlastnamefont {Tsai}}\btxandcomma {}
  \btxandshort {.}\ \btxnamefont {C.\btxfnamespaceshort T. \btxlastnamefont
  {Li}}\btxauthorcolon\ \btxjtitlefont {\btxifchangecase {Fingat: Financial
  graph attention networks for recommending top-k profitable stocks}{FinGAT:
  Financial Graph Attention Networks for Recommending Top-K Profitable
  Stocks}}.
\newblock \btxjournalfont {IEEE Transactions on Knowledge and Data
  Engineering}, \btxpageshort {.}\ 1–1, 2022\ifbtxprintISSN {,
  \mbox{\btxISSN~\btxISSNfont {2326-3865}}}.
\newblock {\latintext
  \btxurlfont{http://dx.doi.org/10.1109/TKDE.2021.3079496}}.

\bibitem {art_695}
\btxnamefont {J.~\btxlastnamefont {Hu}}, \btxnamefont {X.~\btxlastnamefont
  {Liu}}, \btxnamefont {Z.~\btxlastnamefont {Fan}}, \btxnamefont
  {Y.~\btxlastnamefont {Liang}}\btxandcomma {} \btxandshort {.}\ \btxnamefont
  {R.~\btxlastnamefont {Zimmermann}}\btxauthorcolon\ \btxtitlefont
  {\btxifchangecase {Towards unifying diffusion models for probabilistic
  spatio-temporal graph learning}{Towards unifying diffusion models for
  probabilistic spatio-temporal graph learning}}.
\newblock \Btxinshort {.}\ \btxtitlefont {Proceedings of the 32nd {ACM}
  International Conference on Advances in Geographic Information Systems},
  \btxpagesshort {.}\ 135--146, New York, NY, USA,
  \btxprintmonthyear{.}{10}{2024}{short}. \btxpublisherfont {ACM}.

\bibitem {art_694}
\btxnamefont {J.~\btxlastnamefont {Hu}}, \btxnamefont {X.~\btxlastnamefont
  {Liu}}, \btxnamefont {Z.~\btxlastnamefont {Fan}}, \btxnamefont
  {Y.~\btxlastnamefont {Yin}}, \btxnamefont {S.~\btxlastnamefont {Xiang}},
  \btxnamefont {S.~\btxlastnamefont {Ramasamy}}\btxandcomma {} \btxandshort
  {.}\ \btxnamefont {R.~\btxlastnamefont {Zimmermann}}\btxauthorcolon\
  \btxtitlefont {\btxifchangecase {{Prompt-Based} {Spatio-Temporal} graph
  transfer learning}{{Prompt-Based} {Spatio-Temporal} Graph Transfer
  Learning}}.
\newblock \Btxinshort {.}\ \btxtitlefont {Proceedings of the 33rd {ACM}
  International Conference on Information and Knowledge Management},
  \btxpagesshort {.}\ 890--899, New York, NY, USA,
  \btxprintmonthyear{.}{10}{2024}{short}. \btxpublisherfont {ACM}.

\bibitem {art_735}
\btxnamefont {J.~\btxlastnamefont {Hu}}, \btxnamefont {C.~\btxlastnamefont
  {Wang}}\btxandcomma {} \btxandshort {.}\ \btxnamefont {X.~\btxlastnamefont
  {Lin}}\btxauthorcolon\ \btxtitlefont {\btxifchangecase {Spatio-temporal
  pyramid networks for traffic forecasting}{Spatio-Temporal Pyramid Networks
  for Traffic Forecasting}}.
\newblock \Btxinshort {.}\ \btxnamefont {D.~\btxlastnamefont {Koutra}},
  \btxnamefont {C.~\btxlastnamefont {Plant}}, \btxnamefont {M.~\btxlastnamefont
  {Gomez~Rodriguez}}, \btxnamefont {E.~\btxlastnamefont {Baralis}}\btxandcomma
  {} \btxandshort {.}\ \btxnamefont {F.~\btxlastnamefont {Bonchi}}\
  (\btxeditorsshort {.}): \btxtitlefont {Machine Learning and Knowledge
  Discovery in Databases: Research Track}, \btxpagesshort {.}\ 339--354, Cham,
  2023. \btxpublisherfont {Springer Nature Switzerland}\ifbtxprintISBN {,
  \mbox{\btxISBN~\btxISBNfont {978-3-031-43412-9}}}.

\bibitem {NEURIPS2020_fb60d411}
\btxnamefont {W.~\btxlastnamefont {Hu}}, \btxnamefont {M.~\btxlastnamefont
  {Fey}}, \btxnamefont {M.~\btxlastnamefont {Zitnik}}, \btxnamefont
  {Y.~\btxlastnamefont {Dong}}, \btxnamefont {H.~\btxlastnamefont {Ren}},
  \btxnamefont {B.~\btxlastnamefont {Liu}}, \btxnamefont {M.~\btxlastnamefont
  {Catasta}}\btxandcomma {} \btxandshort {.}\ \btxnamefont {J.~\btxlastnamefont
  {Leskovec}}\btxauthorcolon\ \btxtitlefont {\btxifchangecase {Open graph
  benchmark: Datasets for machine learning on graphs}{Open Graph Benchmark:
  Datasets for Machine Learning on Graphs}}.
\newblock \Btxinshort {.}\ \btxnamefont {H.~\btxlastnamefont {Larochelle}},
  \btxnamefont {M.~\btxlastnamefont {Ranzato}}, \btxnamefont
  {R.~\btxlastnamefont {Hadsell}}, \btxnamefont {M.~\btxlastnamefont
  {Balcan}}\btxandcomma {} \btxandshort {.}\ \btxnamefont {H.~\btxlastnamefont
  {Lin}}\ (\btxeditorsshort {.}): \btxtitlefont {Advances in Neural Information
  Processing Systems}, \btxvolumeshort {.}~\btxvolumefont {33}, \btxpagesshort
  {.}\ 22118--22133. \btxpublisherfont {Curran Associates, Inc.}, 2020.
\newblock {\latintext
  \btxurlfont{https://proceedings.neurips.cc/paper_files/paper/2020/file/fb60d411a5c5b72b2e7d3527cfc84fd0-Paper.pdf}}.

\bibitem {art_25}
\btxnamefont {Y.~\btxlastnamefont {Hu}}, \btxnamefont {X.~\btxlastnamefont
  {Cheng}}, \btxnamefont {S.~\btxlastnamefont {Wang}}, \btxnamefont
  {J.~\btxlastnamefont {Chen}}, \btxnamefont {T.~\btxlastnamefont
  {Zhao}}\btxandcomma {} \btxandshort {.}\ \btxnamefont {E.~\btxlastnamefont
  {Dai}}\btxauthorcolon\ \btxjtitlefont {\btxifchangecase {Times series
  forecasting for urban building energy consumption based on graph
  convolutional network}{Times series forecasting for urban building energy
  consumption based on graph convolutional network}}.
\newblock \btxjournalfont {Applied Energy}, 307, 2022.
\newblock {\latintext
  \btxurlfont{https://www.scopus.com/inward/record.uri?eid=2-s2.0-85120306571&doi=10.1016

\bibitem {art_335}
\btxnamefont {Z.~\btxlastnamefont {Hu}}, \btxnamefont {Y.~\btxlastnamefont
  {Gao}}, \btxnamefont {L.~\btxlastnamefont {Sun}}, \btxnamefont
  {M.~\btxlastnamefont {Mae}}\btxandcomma {} \btxandshort {.}\ \btxnamefont
  {T.~\btxlastnamefont {Imaizumi}}\btxauthorcolon\ \btxjtitlefont
  {\btxifchangecase {Self-learning dynamic graph neural network with
  self-attention based on historical data and future data for multi-task
  multivariate residential air conditioning forecasting}{Self-learning dynamic
  graph neural network with self-attention based on historical data and future
  data for multi-task multivariate residential air conditioning forecasting}}.
\newblock \btxjournalfont {Appl. Energy}, 364(123156):123156,
  \btxprintmonthyear{.}{6}{2024}{short}.

\bibitem {DeepCrime}
\btxnamefont {C.~\btxlastnamefont {Huang}}, \btxnamefont {J.~\btxlastnamefont
  {Zhang}}, \btxnamefont {Y.~\btxlastnamefont {Zheng}}\btxandcomma {}
  \btxandshort {.}\ \btxnamefont {N.\btxfnamespaceshort V. \btxlastnamefont
  {Chawla}}\btxauthorcolon\ \btxtitlefont {\btxifchangecase {Deepcrime:
  Attentive hierarchical recurrent networks for crime prediction}{DeepCrime:
  Attentive Hierarchical Recurrent Networks for Crime Prediction}}.
\newblock \Btxinshort {.}\ \btxtitlefont {Proceedings of the 27th ACM
  International Conference on Information and Knowledge Management}, CIKM '18,
  \btxpageshort {.}\ 1423–1432, New York, NY, USA, 2018. \btxpublisherfont
  {Association for Computing Machinery}\ifbtxprintISBN {,
  \mbox{\btxISBN~\btxISBNfont {9781450360142}}}.
\newblock {\latintext \btxurlfont{https://doi.org/10.1145/3269206.3271793}}.

\bibitem {art_746}
\btxnamefont {J.~\btxlastnamefont {Huang}}, \btxnamefont {Y.~\btxlastnamefont
  {Yang}}, \btxnamefont {H.~\btxlastnamefont {Yu}}, \btxnamefont
  {J.~\btxlastnamefont {Li}}\btxandcomma {} \btxandshort {.}\ \btxnamefont
  {X.~\btxlastnamefont {Zheng}}\btxauthorcolon\ \btxtitlefont {\btxifchangecase
  {Twin graph-based anomaly detection via attentive multi-modal learning for
  microservice system}{Twin graph-based anomaly detection via attentive
  multi-modal learning for microservice system}}.
\newblock \Btxinshort {.}\ \btxtitlefont {2023 38th {IEEE/ACM} International
  Conference on Automated Software Engineering ({ASE})}. \btxpublisherfont
  {IEEE}, \btxprintmonthyear{.}{9}{2023}{short}.

\bibitem {art_59}
\btxnamefont {K.~\btxlastnamefont {Huang}}, \btxnamefont {X.~\btxlastnamefont
  {Li}}, \btxnamefont {F.~\btxlastnamefont {Liu}}, \btxnamefont
  {X.~\btxlastnamefont {Yang}}\btxandcomma {} \btxandshort {.}\ \btxnamefont
  {W.~\btxlastnamefont {Yu}}\btxauthorcolon\ \btxjtitlefont {\btxifchangecase
  {Ml-gat:a multilevel graph attention model for stock prediction}{ML-GAT:A
  Multilevel Graph Attention Model for Stock Prediction}}.
\newblock \btxjournalfont {IEEE Access}, 10:86408--86422, 2022.

\bibitem {art_397}
\btxnamefont {L.~\btxlastnamefont {Huang}}, \btxnamefont {J.~\btxlastnamefont
  {Yuan}}, \btxnamefont {S.~\btxlastnamefont {Chen}}\btxandcomma {}
  \btxandshort {.}\ \btxnamefont {X.~\btxlastnamefont {Li}}\btxauthorcolon\
  \btxjtitlefont {\btxifchangecase {{MDG}: A multi-task dynamic graph
  generation framework for multivariate time series forecasting}{{MDG}: A
  multi-task dynamic graph generation framework for multivariate time series
  forecasting}}.
\newblock \btxjournalfont {IEEE Trans. Emerg. Top. Comput. Intell.},
  8(2):1337--1349, \btxprintmonthyear{.}{4}{2024}{short}.

\bibitem {art_687}
\btxnamefont {Q.~\btxlastnamefont {Huang}}, \btxnamefont {L.~\btxlastnamefont
  {Shen}}, \btxnamefont {R.~\btxlastnamefont {Zhang}}, \btxnamefont
  {S.~\btxlastnamefont {Ding}}, \btxnamefont {B.~\btxlastnamefont {Wang}},
  \btxnamefont {Z.~\btxlastnamefont {Zhou}}\btxandcomma {} \btxandshort {.}\
  \btxnamefont {Y.~\btxlastnamefont {Wang}}\btxauthorcolon\ \btxtitlefont
  {\btxifchangecase {Crossgnn: confronting noisy multivariate time series via
  cross interaction refinement}{CrossGNN: confronting noisy multivariate time
  series via cross interaction refinement}}.
\newblock \Btxinshort {.}\ \btxtitlefont {Proceedings of the 37th International
  Conference on Neural Information Processing Systems}, NIPS '23, Red Hook, NY,
  USA, 2023. \btxpublisherfont {Curran Associates Inc.}

\bibitem {LSGCN}
\btxnamefont {R.~\btxlastnamefont {Huang}}, \btxnamefont {C.~\btxlastnamefont
  {Huang}}, \btxnamefont {Y.~\btxlastnamefont {Liu}}, \btxnamefont
  {G.~\btxlastnamefont {Dai}}\btxandcomma {} \btxandshort {.}\ \btxnamefont
  {W.~\btxlastnamefont {Kong}}\btxauthorcolon\ \btxtitlefont {\btxifchangecase
  {Lsgcn: long short-term traffic prediction with graph convolutional
  networks}{LSGCN: long short-term traffic prediction with graph convolutional
  networks}}.
\newblock \Btxinshort {.}\ \btxtitlefont {Proceedings of the Twenty-Ninth
  International Joint Conference on Artificial Intelligence}, IJCAI'20,
  2021\ifbtxprintISBN {, \mbox{\btxISBN~\btxISBNfont {9780999241165}}}.

\bibitem {art_202}
\btxnamefont {R.~\btxlastnamefont {Huang}}, \btxnamefont {L.~\btxlastnamefont
  {Ma}}, \btxnamefont {J.~\btxlastnamefont {He}}\btxandcomma {} \btxandshort
  {.}\ \btxnamefont {X.~\btxlastnamefont {Chu}}\btxauthorcolon\ \btxjtitlefont
  {\btxifchangecase {T-gan: A deep learning framework for prediction of
  temporal complex networks with adaptive graph convolution and attention
  mechanism}{T-GAN: A deep learning framework for prediction of temporal
  complex networks with adaptive graph convolution and attention mechanism}}.
\newblock \btxjournalfont {Displays}, 68:102023, 2021\ifbtxprintISSN {,
  \mbox{\btxISSN~\btxISSNfont {0141-9382}}}.
\newblock {\latintext
  \btxurlfont{https://www.sciencedirect.com/science/article/pii/S0141938221000366}}.

\bibitem {art_344}
\btxnamefont {S.~\btxlastnamefont {Huang}}, \btxnamefont {H.~\btxlastnamefont
  {Que}}, \btxnamefont {L.~\btxlastnamefont {Zeng}}, \btxnamefont
  {J.~\btxlastnamefont {Yang}}\btxandcomma {} \btxandshort {.}\ \btxnamefont
  {K.~\btxlastnamefont {Zheng}}\btxauthorcolon\ \btxjtitlefont
  {\btxifchangecase {Multi-scale graph attention network based on encoding
  decomposition for electricity consumption prediction}{Multi-scale graph
  attention network based on encoding decomposition for electricity consumption
  prediction}}.
\newblock \btxjournalfont {Energies}, 17(23):5813,
  \btxprintmonthyear{.}{11}{2024}{short}.

\bibitem {DSANet}
\btxnamefont {S.~\btxlastnamefont {Huang}}, \btxnamefont {D.~\btxlastnamefont
  {Wang}}, \btxnamefont {X.~\btxlastnamefont {Wu}}\btxandcomma {} \btxandshort
  {.}\ \btxnamefont {A.~\btxlastnamefont {Tang}}\btxauthorcolon\ \btxtitlefont
  {\btxifchangecase {Dsanet: Dual self-attention network for multivariate time
  series forecasting}{DSANet: Dual Self-Attention Network for Multivariate Time
  Series Forecasting}}.
\newblock \Btxinshort {.}\ \btxtitlefont {Proceedings of the 28th ACM
  International Conference on Information and Knowledge Management}, CIKM '19,
  \btxpageshort {.}\ 2129–2132, New York, NY, USA, 2019. \btxpublisherfont
  {Association for Computing Machinery}\ifbtxprintISBN {,
  \mbox{\btxISBN~\btxISBNfont {9781450369763}}}.
\newblock {\latintext \btxurlfont{https://doi.org/10.1145/3357384.3358132}}.

\bibitem {art_67}
\btxnamefont {W.\btxfnamespaceshort C. \btxlastnamefont {Huang}}, \btxnamefont
  {C.\btxfnamespaceshort T. \btxlastnamefont {Chen}}, \btxnamefont
  {C.~\btxlastnamefont {Lee}}, \btxnamefont {F.\btxfnamespaceshort H.
  \btxlastnamefont {Kuo}}\btxandcomma {} \btxandshort {.}\ \btxnamefont
  {S.\btxfnamespaceshort H. \btxlastnamefont {Huang}}\btxauthorcolon\
  \btxjtitlefont {\btxifchangecase {Attentive gated graph sequence neural
  network-based time-series information fusion for financial trading}{Attentive
  gated graph sequence neural network-based time-series information fusion for
  financial trading}}.
\newblock \btxjournalfont {Information Fusion}, 91:261 – 276, 2023.
\newblock {\latintext
  \btxurlfont{https://www.scopus.com/inward/record.uri?eid=2-s2.0-85140905775&doi=10.1016

\bibitem {art_420}
\btxnamefont {G.~\btxlastnamefont {Huo}}, \btxnamefont {Y.~\btxlastnamefont
  {Zhang}}, \btxnamefont {Y.~\btxlastnamefont {Lv}}, \btxnamefont
  {H.~\btxlastnamefont {Ren}}\btxandcomma {} \btxandshort {.}\ \btxnamefont
  {B.~\btxlastnamefont {Yin}}\btxauthorcolon\ \btxjtitlefont {\btxifchangecase
  {Urban traffic flow forecasting based on graph structure learning}{Urban
  traffic flow forecasting based on graph structure learning}}.
\newblock \btxjournalfont {J. Adv. Transp.}, 2024(1),
  \btxprintmonthyear{.}{1}{2024}{short}.

\bibitem {art_662}
\btxnamefont {T.\btxfnamespaceshort S. \btxlastnamefont {Hy}}, \btxnamefont
  {V.\btxfnamespaceshort B. \btxlastnamefont {Nguyen}}, \btxnamefont
  {L.~\btxlastnamefont {Tran-Thanh}}\btxandcomma {} \btxandshort {.}\
  \btxnamefont {R.~\btxlastnamefont {Kondor}}\btxauthorcolon\ \btxtitlefont
  {\btxifchangecase {Temporal multiresolution graph neural networks for
  epidemic prediction}{Temporal Multiresolution Graph Neural Networks For
  Epidemic Prediction}}.
\newblock 2022.

\bibitem {art_342}
\btxnamefont {S.~\btxlastnamefont {Ida~Evangeline}}, \btxnamefont
  {S.~\btxlastnamefont {Darwin}}, \btxnamefont {P.~\btxlastnamefont
  {Peter~Anandkumar}}\btxandcomma {} \btxandshort {.}\ \btxnamefont
  {M.~\btxlastnamefont {Chithambara~Thanu}}\btxauthorcolon\ \btxjtitlefont
  {\btxifchangecase {Anomaly detection in smart grid using a trace-based graph
  deep learning model}{Anomaly detection in smart grid using a trace-based
  graph deep learning model}}.
\newblock \btxjournalfont {Electr. Eng. (Berl., Print)}, 106(5):5851--5867,
  \btxprintmonthyear{.}{10}{2024}{short}.

\bibitem {InceptionTime}
\btxnamefont {H.~\btxlastnamefont {Ismail~Fawaz}}, \btxnamefont
  {B.~\btxlastnamefont {Lucas}}, \btxnamefont {G.~\btxlastnamefont
  {Forestier}}, \btxnamefont {C.~\btxlastnamefont {Pelletier}}, \btxnamefont
  {D.\btxfnamespaceshort F. \btxlastnamefont {Schmidt}}, \btxnamefont
  {J.~\btxlastnamefont {Weber}}, \btxnamefont {G.\btxfnamespaceshort I.
  \btxlastnamefont {Webb}}, \btxnamefont {L.~\btxlastnamefont {Idoumghar}},
  \btxnamefont {P.\btxfnamespaceshort A. \btxlastnamefont {Muller}}\btxandcomma
  {} \btxandshort {.}\ \btxnamefont {F.~\btxlastnamefont
  {Petitjean}}\btxauthorcolon\ \btxjtitlefont {\btxifchangecase {Inceptiontime:
  Finding alexnet for time series classification}{InceptionTime: Finding
  AlexNet for time series classification}}.
\newblock \btxjournalfont {Data Mining and Knowledge Discovery},
  34(6):1936–1962, \btxprintmonthyear{.}{9}{2020}{short}\ifbtxprintISSN {,
  \mbox{\btxISSN~\btxISSNfont {1573-756X}}}.
\newblock {\latintext
  \btxurlfont{http://dx.doi.org/10.1007/s10618-020-00710-y}}.

\bibitem {art_419}
\btxnamefont {Y.~\btxlastnamefont {Jakhmola}}, \btxnamefont
  {M.~\btxlastnamefont {Panja}}, \btxnamefont {N.\btxfnamespaceshort K.
  \btxlastnamefont {Mishra}}, \btxnamefont {K.~\btxlastnamefont {Ghosh}},
  \btxnamefont {U.~\btxlastnamefont {Kumar}}\btxandcomma {} \btxandshort {.}\
  \btxnamefont {T.~\btxlastnamefont {Chakraborty}}\btxauthorcolon\
  \btxjtitlefont {\btxifchangecase {Spatiotemporal forecasting of traffic flow
  using wavelet-based temporal attention}{Spatiotemporal forecasting of traffic
  flow using wavelet-based temporal attention}}.
\newblock \btxjournalfont {IEEE Access}, \btxpagesshort {.}\ 1--1, 2024.

\bibitem {Wavelet-DBN-RF}
\btxnamefont {H.~\btxlastnamefont {Jiajun}}, \btxnamefont {Y.~\btxlastnamefont
  {Chuanjin}}, \btxnamefont {L.~\btxlastnamefont {Yongle}}\btxandcomma {}
  \btxandshort {.}\ \btxnamefont {X.~\btxlastnamefont {Huoyue}}\btxauthorcolon\
  \btxjtitlefont {\btxifchangecase {Ultra-short term wind prediction with
  wavelet transform, deep belief network and ensemble learning}{Ultra-short
  term wind prediction with wavelet transform, deep belief network and ensemble
  learning}}.
\newblock \btxjournalfont {Energy Conversion and Management}, 205:112418,
  2020\ifbtxprintISSN {, \mbox{\btxISSN~\btxISSNfont {0196-8904}}}.
\newblock {\latintext
  \btxurlfont{https://www.sciencedirect.com/science/article/pii/S0196890419314256}}.

\bibitem {PDFormer}
\btxnamefont {J.~\btxlastnamefont {Jiang}}, \btxnamefont {C.~\btxlastnamefont
  {Han}}, \btxnamefont {W.\btxfnamespaceshort X. \btxlastnamefont
  {Zhao}}\btxandcomma {} \btxandshort {.}\ \btxnamefont {J.~\btxlastnamefont
  {Wang}}\btxauthorcolon\ \btxtitlefont {\btxifchangecase {Pdformer:
  propagation delay-aware dynamic long-range transformer for traffic flow
  prediction}{PDFormer: propagation delay-aware dynamic long-range transformer
  for traffic flow prediction}}.
\newblock \Btxinshort {.}\ \btxtitlefont {Proceedings of the Thirty-Seventh
  AAAI Conference on Artificial Intelligence and Thirty-Fifth Conference on
  Innovative Applications of Artificial Intelligence and Thirteenth Symposium
  on Educational Advances in Artificial Intelligence}, AAAI'23/IAAI'23/EAAI'23.
  \btxpublisherfont {AAAI Press}, 2023\ifbtxprintISBN {,
  \mbox{\btxISBN~\btxISBNfont {978-1-57735-880-0}}}.
\newblock {\latintext \btxurlfont{https://doi.org/10.1609/aaai.v37i4.25556}}.

\bibitem {MegaCRN}
\btxnamefont {R.~\btxlastnamefont {Jiang}}, \btxnamefont {Z.~\btxlastnamefont
  {Wang}}, \btxnamefont {J.~\btxlastnamefont {Yong}}, \btxnamefont
  {P.~\btxlastnamefont {Jeph}}, \btxnamefont {Q.~\btxlastnamefont {Chen}},
  \btxnamefont {Y.~\btxlastnamefont {Kobayashi}}, \btxnamefont
  {X.~\btxlastnamefont {Song}}, \btxnamefont {S.~\btxlastnamefont
  {Fukushima}}\btxandcomma {} \btxandshort {.}\ \btxnamefont
  {T.~\btxlastnamefont {Suzumura}}\btxauthorcolon\ \btxtitlefont
  {\btxifchangecase {Spatio-temporal meta-graph learning for traffic
  forecasting}{Spatio-temporal meta-graph learning for traffic forecasting}}.
\newblock \Btxinshort {.}\ \btxtitlefont {Proceedings of the Thirty-Seventh
  AAAI Conference on Artificial Intelligence and Thirty-Fifth Conference on
  Innovative Applications of Artificial Intelligence and Thirteenth Symposium
  on Educational Advances in Artificial Intelligence}, AAAI'23/IAAI'23/EAAI'23.
  \btxpublisherfont {AAAI Press}, 2023\ifbtxprintISBN {,
  \mbox{\btxISBN~\btxISBNfont {978-1-57735-880-0}}}.
\newblock {\latintext \btxurlfont{https://doi.org/10.1609/aaai.v37i7.25976}}.

\bibitem {review_traffico_con_dataset}
\btxnamefont {R.~\btxlastnamefont {Jiang}}, \btxnamefont {D.~\btxlastnamefont
  {Yin}}, \btxnamefont {Z.~\btxlastnamefont {Wang}}, \btxnamefont
  {Y.~\btxlastnamefont {Wang}}, \btxnamefont {J.~\btxlastnamefont {Deng}},
  \btxnamefont {H.~\btxlastnamefont {Liu}}, \btxnamefont {Z.~\btxlastnamefont
  {Cai}}, \btxnamefont {J.~\btxlastnamefont {Deng}}, \btxnamefont
  {X.~\btxlastnamefont {Song}}\btxandcomma {} \btxandshort {.}\ \btxnamefont
  {R.~\btxlastnamefont {Shibasaki}}\btxauthorcolon\ \btxtitlefont
  {\btxifchangecase {Dl-traff: Survey and benchmark of deep learning models for
  urban traffic prediction}{DL-Traff: Survey and Benchmark of Deep Learning
  Models for Urban Traffic Prediction}}.
\newblock \Btxinshort {.}\ \btxtitlefont {Proceedings of the 30th ACM
  International Conference on Information \& Knowledge Management}, CIKM '21,
  \btxpageshort {.}\ 4515–4525, New York, NY, USA, 2021. \btxpublisherfont
  {Association for Computing Machinery}\ifbtxprintISBN {,
  \mbox{\btxISBN~\btxISBNfont {9781450384469}}}.
\newblock {\latintext \btxurlfont{https://doi.org/10.1145/3459637.3482000}}.

\bibitem {review_traffico_jiang}
\btxnamefont {W.~\btxlastnamefont {Jiang}} \btxandshort {.}\ \btxnamefont
  {J.~\btxlastnamefont {Luo}}\btxauthorcolon\ \btxjtitlefont {\btxifchangecase
  {Graph neural network for traffic forecasting: A survey}{Graph neural network
  for traffic forecasting: A survey}}.
\newblock \btxjournalfont {Expert Systems with Applications}, 207:117921,
  2022\ifbtxprintISSN {, \mbox{\btxISSN~\btxISSNfont {0957-4174}}}.
\newblock {\latintext
  \btxurlfont{https://www.sciencedirect.com/science/article/pii/S0957417422011654}}.

\bibitem {art_135}
\btxnamefont {W.~\btxlastnamefont {Jiang}}, \btxnamefont {Y.~\btxlastnamefont
  {Xiao}}, \btxnamefont {Y.~\btxlastnamefont {Liu}}, \btxnamefont
  {Q.~\btxlastnamefont {Liu}}\btxandcomma {} \btxandshort {.}\ \btxnamefont
  {Z.~\btxlastnamefont {Li}}\btxauthorcolon\ \btxjtitlefont {\btxifchangecase
  {Bi-grcn: A spatio-temporal traffic flow prediction model based on graph
  neural network}{Bi-GRCN: A Spatio-Temporal Traffic Flow Prediction Model
  Based on Graph Neural Network}}.
\newblock \btxjournalfont {Journal of Advanced Transportation}, 2022, 2022.
\newblock {\latintext
  \btxurlfont{https://www.scopus.com/inward/record.uri?eid=2-s2.0-85124792614&doi=10.1155

\bibitem {art_468}
\btxnamefont {W.~\btxlastnamefont {Jiang}}, \btxnamefont {Y.~\btxlastnamefont
  {Zhang}}, \btxnamefont {H.~\btxlastnamefont {Han}}, \btxnamefont
  {Z.~\btxlastnamefont {Huang}}, \btxnamefont {Q.~\btxlastnamefont
  {Li}}\btxandcomma {} \btxandshort {.}\ \btxnamefont {J.~\btxlastnamefont
  {Mu}}\btxauthorcolon\ \btxjtitlefont {\btxifchangecase {Mobile traffic
  prediction in consumer applications: A multimodal deep learning
  approach}{Mobile traffic prediction in consumer applications: A multimodal
  deep learning approach}}.
\newblock \btxjournalfont {IEEE Trans. Consum. Electron.}, 70(1):3425--3435,
  \btxprintmonthyear{.}{2}{2024}{short}.

\bibitem {art_653}
\btxnamefont {Y.~\btxlastnamefont {Jiang}}, \btxnamefont {X.~\btxlastnamefont
  {Li}}, \btxnamefont {Y.~\btxlastnamefont {Chen}}, \btxnamefont
  {S.~\btxlastnamefont {Liu}}, \btxnamefont {W.~\btxlastnamefont {Kong}},
  \btxnamefont {A.\btxfnamespaceshort F. \btxlastnamefont
  {Lentzakis}}\btxandcomma {} \btxandshort {.}\ \btxnamefont
  {G.~\btxlastnamefont {Cong}}\btxauthorcolon\ \btxtitlefont {\btxifchangecase
  {{SAGDFN}: A scalable adaptive graph diffusion forecasting network for
  multivariate time series forecasting}{{SAGDFN}: A scalable adaptive graph
  diffusion forecasting network for multivariate time series forecasting}}.
\newblock \Btxinshort {.}\ \btxtitlefont {2024 {IEEE} 40th International
  Conference on Data Engineering ({ICDE})}. \btxpublisherfont {IEEE},
  \btxprintmonthyear{.}{5}{2024}{short}.

\bibitem {ref_anomaly_detection_2}
\btxnamefont {J.~\btxlastnamefont {Jiao}}, \btxnamefont {M.~\btxlastnamefont
  {Zhao}}, \btxnamefont {J.~\btxlastnamefont {Lin}}\btxandcomma {} \btxandshort
  {.}\ \btxnamefont {K.~\btxlastnamefont {Liang}}\btxauthorcolon\
  \btxjtitlefont {\btxifchangecase {Hierarchical discriminating sparse coding
  for weak fault feature extraction of rolling bearings}{Hierarchical
  Discriminating Sparse Coding for Weak Fault Feature Extraction of Rolling
  Bearings}}.
\newblock \btxjournalfont {Reliability Engineering and System Safety}, 184,
  \btxprintmonthyear{.}{02}{2018}{short}.

\bibitem {art_108}
\btxnamefont {G.~\btxlastnamefont {Jin}}, \btxnamefont {F.~\btxlastnamefont
  {Li}}, \btxnamefont {J.~\btxlastnamefont {Zhang}}, \btxnamefont
  {M.~\btxlastnamefont {Wang}}\btxandcomma {} \btxandshort {.}\ \btxnamefont
  {J.~\btxlastnamefont {Huang}}\btxauthorcolon\ \btxjtitlefont
  {\btxifchangecase {Automated dilated spatio-temporal synchronous graph
  modeling for traffic prediction}{Automated Dilated Spatio-Temporal
  Synchronous Graph Modeling for Traffic Prediction}}.
\newblock \btxjournalfont {IEEE Transactions on Intelligent Transportation
  Systems}, 24(8):8820 – 8830, 2023.
\newblock {\latintext
  \btxurlfont{https://www.scopus.com/inward/record.uri?eid=2-s2.0-85136873211&doi=10.1109

\bibitem {cit_rev4}
\btxnamefont {G.~\btxlastnamefont {Jin}}, \btxnamefont {Y.~\btxlastnamefont
  {Liang}}, \btxnamefont {Y.~\btxlastnamefont {Fang}}, \btxnamefont
  {Z.~\btxlastnamefont {Shao}}, \btxnamefont {J.~\btxlastnamefont {Huang}},
  \btxnamefont {J.~\btxlastnamefont {Zhang}}\btxandcomma {} \btxandshort {.}\
  \btxnamefont {Y.~\btxlastnamefont {Zheng}}\btxauthorcolon\ \btxjtitlefont
  {\btxifchangecase {Spatio-temporal graph neural networks for predictive
  learning in urban computing: A survey}{Spatio-Temporal Graph Neural Networks
  for Predictive Learning in Urban Computing: A Survey}}.
\newblock \btxjournalfont {IEEE Transactions on Knowledge and Data
  Engineering}, \btxpagesshort {.}\ 1--20, 2023.

\bibitem {art_136}
\btxnamefont {G.~\btxlastnamefont {Jin}}, \btxnamefont {C.~\btxlastnamefont
  {Liu}}, \btxnamefont {Z.~\btxlastnamefont {Xi}}, \btxnamefont
  {H.~\btxlastnamefont {Sha}}, \btxnamefont {Y.~\btxlastnamefont
  {Liu}}\btxandcomma {} \btxandshort {.}\ \btxnamefont {J.~\btxlastnamefont
  {Huang}}\btxauthorcolon\ \btxjtitlefont {\btxifchangecase {Adaptive dual-view
  wavenet for urban spatial–temporal event prediction}{Adaptive Dual-View
  WaveNet for urban spatial–temporal event prediction}}.
\newblock \btxjournalfont {Information Sciences}, 588:315 – 330, 2022.
\newblock {\latintext
  \btxurlfont{https://www.scopus.com/inward/record.uri?eid=2-s2.0-85122370178&doi=10.1016

\bibitem {review_Alippi}
\btxnamefont {M.~\btxlastnamefont {Jin}}, \btxnamefont {H.\btxfnamespaceshort
  Y. \btxlastnamefont {Koh}}, \btxnamefont {Q.~\btxlastnamefont {Wen}},
  \btxnamefont {D.~\btxlastnamefont {Zambon}}, \btxnamefont
  {C.~\btxlastnamefont {Alippi}}, \btxnamefont {G.\btxfnamespaceshort I.
  \btxlastnamefont {Webb}}, \btxnamefont {I.~\btxlastnamefont
  {King}}\btxandcomma {} \btxandshort {.}\ \btxnamefont {S.~\btxlastnamefont
  {Pan}}\btxauthorcolon\ \btxjtitlefont {\btxifchangecase {A survey on graph
  neural networks for time series: Forecasting, classification, imputation, and
  anomaly detection}{A Survey on Graph Neural Networks for Time Series:
  Forecasting, Classification, Imputation, and Anomaly Detection}}.
\newblock \btxjournalfont {IEEE Transactions on Pattern Analysis and Machine
  Intelligence}, 46(12):10466--10485, 2024.

\bibitem {art_75}
\btxnamefont {M.~\btxlastnamefont {Jin}}, \btxnamefont {Y.~\btxlastnamefont
  {Zheng}}, \btxnamefont {Y.\btxfnamespaceshort F. \btxlastnamefont {Li}},
  \btxnamefont {S.~\btxlastnamefont {Chen}}, \btxnamefont {B.~\btxlastnamefont
  {Yang}}\btxandcomma {} \btxandshort {.}\ \btxnamefont {S.~\btxlastnamefont
  {Pan}}\btxauthorcolon\ \btxjtitlefont {\btxifchangecase {Multivariate time
  series forecasting with dynamic graph neural odes}{Multivariate Time Series
  Forecasting With Dynamic Graph Neural ODEs}}.
\newblock \btxjournalfont {IEEE Transactions on Knowledge and Data
  Engineering}, 35(9):9168 – 9180, 2023.
\newblock {\latintext
  \btxurlfont{https://www.scopus.com/inward/record.uri?eid=2-s2.0-85142836712&doi=10.1109

\bibitem {art_670}
\btxnamefont {Y.~\btxlastnamefont {Jin}}, \btxnamefont {K.~\btxlastnamefont
  {Chen}}\btxandcomma {} \btxandshort {.}\ \btxnamefont {Q.~\btxlastnamefont
  {Yang}}\btxauthorcolon\ \btxtitlefont {\btxifchangecase {Transferable graph
  structure learning for graph-based traffic forecasting across
  cities}{Transferable graph structure learning for graph-based traffic
  forecasting across cities}}.
\newblock \Btxinshort {.}\ \btxtitlefont {Proceedings of the 29th {ACM}
  {SIGKDD} Conference on Knowledge Discovery and Data Mining}, \btxpagesshort
  {.}\ 1032--1043, New York, NY, USA, \btxprintmonthyear{.}{8}{2023}{short}.
  \btxpublisherfont {ACM}.

\bibitem {CovidGNN}
\btxnamefont {A.~\btxlastnamefont {Kapoor}}, \btxnamefont {X.~\btxlastnamefont
  {Ben}}, \btxnamefont {L.~\btxlastnamefont {Liu}}, \btxnamefont
  {B.~\btxlastnamefont {Perozzi}}, \btxnamefont {M.~\btxlastnamefont {Barnes}},
  \btxnamefont {M.~\btxlastnamefont {Blais}}\btxandcomma {} \btxandshort {.}\
  \btxnamefont {S.~\btxlastnamefont {O'Banion}}\btxauthorcolon\ \btxtitlefont
  {\btxifchangecase {Examining covid-19 forecasting using spatio-temporal graph
  neural networks}{Examining COVID-19 Forecasting using Spatio-Temporal Graph
  Neural Networks}}, 2020.
\newblock {\latintext \btxurlfont{https://arxiv.org/abs/2007.03113}}.

\bibitem {modello_SIR}
\btxnamefont {W.\btxfnamespaceshort O. \btxlastnamefont {Kermack}} \btxandshort
  {.}\ \btxnamefont {A.\btxfnamespaceshort G. \btxlastnamefont
  {McKendrick}}\btxauthorcolon\ \btxjtitlefont {\btxifchangecase {A
  contribution to the mathematical theory of epidemics}{A contribution to the
  mathematical theory of epidemics}}.
\newblock \btxjournalfont {Proceedings of the Royal Society of London. Series
  A, Containing Papers of a Mathematical and Physical Character},
  115(772):700--721, 1927\ifbtxprintISSN {, \mbox{\btxISSN~\btxISSNfont
  {2053-9150}}}.
\newblock {\latintext \btxurlfont{http://dx.doi.org/10.1098/rspa.1927.0118}}.

\bibitem {art_122}
\btxnamefont {A.~\btxlastnamefont {Khaled}}, \btxnamefont
  {A.\btxfnamespaceshort M.\btxfnamespaceshort T. \btxlastnamefont
  {Elsir}}\btxandcomma {} \btxandshort {.}\ \btxnamefont {Y.~\btxlastnamefont
  {Shen}}\btxauthorcolon\ \btxjtitlefont {\btxifchangecase {Tfgan: Traffic
  forecasting using generative adversarial network with multi-graph
  convolutional network}{TFGAN: Traffic forecasting using generative
  adversarial network with multi-graph convolutional network}}.
\newblock \btxjournalfont {Knowledge-Based Systems}, 249, 2022.
\newblock {\latintext
  \btxurlfont{https://www.scopus.com/inward/record.uri?eid=2-s2.0-85133943024&doi=10.1016

\bibitem {Khemani2024}
\btxnamefont {B.~\btxlastnamefont {Khemani}}, \btxnamefont {S.~\btxlastnamefont
  {Patil}}, \btxnamefont {K.~\btxlastnamefont {Kotecha}}\btxandcomma {}
  \btxandshort {.}\ \btxnamefont {S.~\btxlastnamefont {Tanwar}}\btxauthorcolon\
  \btxjtitlefont {\btxifchangecase {A review of graph neural networks:
  concepts, architectures, techniques, challenges, datasets, applications, and
  future directions}{A review of graph neural networks: concepts,
  architectures, techniques, challenges, datasets, applications, and future
  directions}}.
\newblock \btxjournalfont {Journal of Big Data}, 11(1),
  \btxprintmonthyear{.}{1}{2024}{short}\ifbtxprintISSN {,
  \mbox{\btxISSN~\btxISSNfont {2196-1115}}}.
\newblock {\latintext
  \btxurlfont{http://dx.doi.org/10.1186/s40537-023-00876-4}}.

\bibitem {art_7}
\btxnamefont {D.\btxfnamespaceshort Y. \btxlastnamefont {Kim}}, \btxnamefont
  {D.\btxfnamespaceshort Y. \btxlastnamefont {Jin}}\btxandcomma {} \btxandshort
  {.}\ \btxnamefont {H.\btxfnamespaceshort I. \btxlastnamefont
  {Suk}}\btxauthorcolon\ \btxjtitlefont {\btxifchangecase {Spatiotemporal graph
  neural networks for predicting mid-to-long-term pm2.5
  concentrations}{Spatiotemporal graph neural networks for predicting
  mid-to-long-term PM2.5 concentrations}}.
\newblock \btxjournalfont {Journal of Cleaner Production}, 425, 2023.
\newblock {\latintext
  \btxurlfont{https://www.scopus.com/inward/record.uri?eid=2-s2.0-85171550808&doi=10.1016

\bibitem {art_357}
\btxnamefont {H.\btxfnamespaceshort G. \btxlastnamefont {Kim}}, \btxnamefont
  {E.\btxfnamespaceshort Y. \btxlastnamefont {Jung}}, \btxnamefont
  {H.~\btxlastnamefont {Jeong}}, \btxnamefont {H.~\btxlastnamefont {Son}},
  \btxnamefont {S.\btxfnamespaceshort S. \btxlastnamefont {Baek}}\btxandcomma
  {} \btxandshort {.}\ \btxnamefont {K.\btxfnamespaceshort H. \btxlastnamefont
  {Cho}}\btxauthorcolon\ \btxjtitlefont {\btxifchangecase {Modeling freshwater
  plankton community dynamics with static and dynamic interactions using graph
  convolution embedded long short-term memory}{Modeling freshwater plankton
  community dynamics with static and dynamic interactions using graph
  convolution embedded long short-term memory}}.
\newblock \btxjournalfont {Water Res.}, 266(122401):122401,
  \btxprintmonthyear{.}{11}{2024}{short}.

\bibitem {art_33}
\btxnamefont {J.~\btxlastnamefont {Kim}}, \btxnamefont {T.~\btxlastnamefont
  {Kim}}, \btxnamefont {J.\btxfnamespaceshort G. \btxlastnamefont
  {Ryu}}\btxandcomma {} \btxandshort {.}\ \btxnamefont {J.~\btxlastnamefont
  {Kim}}\btxauthorcolon\ \btxjtitlefont {\btxifchangecase {Spatiotemporal graph
  neural network for multivariate multi-step ahead time-series forecasting of
  sea temperature}{Spatiotemporal graph neural network for multivariate
  multi-step ahead time-series forecasting of sea temperature}}.
\newblock \btxjournalfont {Engineering Applications of Artificial
  Intelligence}, 126, 2023.
\newblock {\latintext
  \btxurlfont{https://www.scopus.com/inward/record.uri?eid=2-s2.0-85166650882&doi=10.1016

\bibitem {art_85}
\btxnamefont {J.~\btxlastnamefont {Kim}}, \btxnamefont {H.~\btxlastnamefont
  {Lee}}, \btxnamefont {S.~\btxlastnamefont {Yu}}, \btxnamefont
  {U.~\btxlastnamefont {Hwang}}, \btxnamefont {W.~\btxlastnamefont
  {Jung}}\btxandcomma {} \btxandshort {.}\ \btxnamefont {K.~\btxlastnamefont
  {Yoon}}\btxauthorcolon\ \btxjtitlefont {\btxifchangecase {Hierarchical joint
  graph learning and multivariate time series forecasting}{Hierarchical Joint
  Graph Learning and Multivariate Time Series Forecasting}}.
\newblock \btxjournalfont {IEEE Access}, 11:118386 – 118394, 2023.
\newblock {\latintext
  \btxurlfont{https://www.scopus.com/inward/record.uri?eid=2-s2.0-85174798168&doi=10.1109

\bibitem {art_416}
\btxnamefont {M.~\btxlastnamefont {Kim}}, \btxnamefont {J.\btxfnamespaceshort
  H. \btxlastnamefont {Kim}}\btxandcomma {} \btxandshort {.}\ \btxnamefont
  {B.~\btxlastnamefont {Jang}}\btxauthorcolon\ \btxjtitlefont {\btxifchangecase
  {Forecasting epidemic spread with recurrent graph gate fusion
  transformers}{Forecasting epidemic spread with recurrent graph gate fusion
  Transformers}}.
\newblock \btxjournalfont {IEEE J. Biomed. Health Inform.}, PP,
  \btxprintmonthyear{.}{10}{2024}{short}.

\bibitem {HATS}
\btxnamefont {R.~\btxlastnamefont {Kim}}, \btxnamefont {C.\btxfnamespaceshort
  H. \btxlastnamefont {So}}, \btxnamefont {M.~\btxlastnamefont {Jeong}},
  \btxnamefont {S.~\btxlastnamefont {Lee}}, \btxnamefont {J.~\btxlastnamefont
  {Kim}}\btxandcomma {} \btxandshort {.}\ \btxnamefont {J.~\btxlastnamefont
  {Kang}}\btxauthorcolon\ \btxtitlefont {\btxifchangecase {Hats: A hierarchical
  graph attention network for stock movement prediction}{HATS: A Hierarchical
  Graph Attention Network for Stock Movement Prediction}}, 2019.
\newblock {\latintext \btxurlfont{https://arxiv.org/abs/1908.07999}}.

\bibitem {GAE}
\btxnamefont {T.~\btxlastnamefont {Kipf}}, \btxnamefont {E.~\btxlastnamefont
  {Fetaya}}, \btxnamefont {K.\btxfnamespaceshort C. \btxlastnamefont {Wang}},
  \btxnamefont {M.~\btxlastnamefont {Welling}}\btxandcomma {} \btxandshort {.}\
  \btxnamefont {R.~\btxlastnamefont {Zemel}}\btxauthorcolon\ \btxtitlefont
  {\btxifchangecase {Neural relational inference for interacting
  systems}{Neural Relational Inference for Interacting Systems}}.
\newblock \Btxinshort {.}\ \btxnamefont {J.~\btxlastnamefont {Dy}} \btxandshort
  {.}\ \btxnamefont {A.~\btxlastnamefont {Krause}}\ (\btxeditorsshort {.}):
  \btxtitlefont {Proceedings of the 35th International Conference on Machine
  Learning}, \btxvolumeshort {.}~\btxvolumefont {80} \btxofseriesshort {.}\
  \btxtitlefont {Proceedings of Machine Learning Research}, \btxpagesshort {.}\
  2688--2697. \btxpublisherfont {PMLR}, \btxprintmonthyear{.}{10--15
  Jul}{2018}{short}.
\newblock {\latintext
  \btxurlfont{https://proceedings.mlr.press/v80/kipf18a.html}}.

\bibitem {graph_convolution_kipf}
\btxnamefont {T.\btxfnamespaceshort N. \btxlastnamefont {Kipf}} \btxandshort
  {.}\ \btxnamefont {M.~\btxlastnamefont {Welling}}\btxauthorcolon\
  \btxtitlefont {\btxifchangecase {{Semi-Supervised Classification with Graph
  Convolutional Networks}}{{Semi-Supervised Classification with Graph
  Convolutional Networks}}}.
\newblock \Btxinshort {.}\ \btxtitlefont {Proceedings of the 5th International
  Conference on Learning Representations}, ICLR '17, 2017.
\newblock {\latintext \btxurlfont{https://openreview.net/forum?id=SJU4ayYgl}}.

\bibitem {Reformer}
\btxnamefont {N.~\btxlastnamefont {Kitaev}}, \btxnamefont {L.~\btxlastnamefont
  {Kaiser}}\btxandcomma {} \btxandshort {.}\ \btxnamefont {A.~\btxlastnamefont
  {Levskaya}}\btxauthorcolon\ \btxtitlefont {\btxifchangecase {Reformer: The
  efficient transformer}{Reformer: The Efficient Transformer}}, 2020.
\newblock {\latintext \btxurlfont{https://openreview.net/forum?id=rkgNKkHtvB}}.

\bibitem {Captum}
\btxnamefont {N.~\btxlastnamefont {Kokhlikyan}}, \btxnamefont
  {V.~\btxlastnamefont {Miglani}}, \btxnamefont {M.~\btxlastnamefont {Martin}},
  \btxnamefont {E.~\btxlastnamefont {Wang}}, \btxnamefont {B.~\btxlastnamefont
  {Alsallakh}}, \btxnamefont {J.~\btxlastnamefont {Reynolds}}, \btxnamefont
  {A.~\btxlastnamefont {Melnikov}}, \btxnamefont {N.~\btxlastnamefont
  {Kliushkina}}, \btxnamefont {C.~\btxlastnamefont {Araya}}, \btxnamefont
  {S.~\btxlastnamefont {Yan}}\btxandcomma {} \btxandshort {.}\ \btxnamefont
  {O.~\btxlastnamefont {Reblitz{-}Richardson}}\btxauthorcolon\ \btxtitlefont
  {\btxifchangecase {Captum: {A} unified and generic model interpretability
  library for pytorch}{Captum: {A} unified and generic model interpretability
  library for PyTorch}}, 2020.
\newblock {\latintext \btxurlfont{https://arxiv.org/abs/2009.07896}}.

\bibitem {art_454}
\btxnamefont {X.~\btxlastnamefont {Kong}}, \btxnamefont {Z.~\btxlastnamefont
  {Shen}}, \btxnamefont {K.~\btxlastnamefont {Wang}}, \btxnamefont
  {G.~\btxlastnamefont {Shen}}\btxandcomma {} \btxandshort {.}\ \btxnamefont
  {Y.~\btxlastnamefont {Fu}}\btxauthorcolon\ \btxjtitlefont {\btxifchangecase
  {Exploring bus stop mobility pattern: A multi-pattern deep learning
  prediction framework}{Exploring bus stop mobility pattern: A multi-pattern
  deep learning prediction framework}}.
\newblock \btxjournalfont {IEEE Trans. Intell. Transp. Syst.},
  25(7):6604--6616, \btxprintmonthyear{.}{7}{2024}{short}.

\bibitem {art_488}
\btxnamefont {Z.~\btxlastnamefont {Kong}}, \btxnamefont {X.~\btxlastnamefont
  {Jin}}, \btxnamefont {F.~\btxlastnamefont {Wang}}\btxandcomma {} \btxandshort
  {.}\ \btxnamefont {Z.~\btxlastnamefont {Xu}}\btxauthorcolon\ \btxjtitlefont
  {\btxifchangecase {Spatio-temporal propagation: An extended message passing
  graph neural network for remaining useful life prediction}{Spatio-temporal
  propagation: An extended message passing graph neural network for remaining
  useful life prediction}}.
\newblock \btxjournalfont {IEEE Sens. J.}, \btxpagesshort {.}\ 1--1, 2024.

\bibitem {art_157}
\btxnamefont {Z.~\btxlastnamefont {Kong}}, \btxnamefont {X.~\btxlastnamefont
  {Jin}}, \btxnamefont {Z.~\btxlastnamefont {Xu}}\btxandcomma {} \btxandshort
  {.}\ \btxnamefont {B.~\btxlastnamefont {Zhang}}\btxauthorcolon\
  \btxjtitlefont {\btxifchangecase {Spatio-temporal fusion attention: A novel
  approach for remaining useful life prediction based on graph neural
  network}{Spatio-Temporal Fusion Attention: A Novel Approach for Remaining
  Useful Life Prediction Based on Graph Neural Network}}.
\newblock \btxjournalfont {IEEE Transactions on Instrumentation and
  Measurement}, 71, 2022.
\newblock {\latintext
  \btxurlfont{https://www.scopus.com/inward/record.uri?eid=2-s2.0-85133611731&doi=10.1109

\bibitem {art_469}
\btxnamefont {A.~\btxlastnamefont {Kovalenko}}, \btxnamefont
  {V.~\btxlastnamefont {Pozdnyakov}}\btxandcomma {} \btxandshort {.}\
  \btxnamefont {I.~\btxlastnamefont {Makarov}}\btxauthorcolon\ \btxjtitlefont
  {\btxifchangecase {Graph neural networks with trainable adjacency matrices
  for fault diagnosis on multivariate sensor data}{Graph neural networks with
  trainable adjacency matrices for fault diagnosis on multivariate sensor
  data}}.
\newblock \btxjournalfont {IEEE Access}, 12:152860--152872, 2024.

\bibitem {survey_kumar_spatiotemporal_data}
\btxnamefont {R.~\btxlastnamefont {Kumar}}, \btxnamefont {M.~\btxlastnamefont
  {Bhanu}}, \btxnamefont {J.\btxfnamespaceshort a. \btxlastnamefont
  {Mendes-Moreira}}\btxandcomma {} \btxandshort {.}\ \btxnamefont
  {J.~\btxlastnamefont {Chandra}}\btxauthorcolon\ \btxjtitlefont
  {\btxifchangecase {Spatio-temporal predictive modeling techniques for
  different domains: a survey}{Spatio-Temporal Predictive Modeling Techniques
  for Different Domains: a Survey}}.
\newblock \btxjournalfont {ACM Comput. Surv.}, 57(2),
  \btxprintmonthyear{.}{10}{2024}{short}\ifbtxprintISSN {,
  \mbox{\btxISSN~\btxISSNfont {0360-0300}}}.
\newblock {\latintext \btxurlfont{https://doi.org/10.1145/3696661}}.

\bibitem {art_113}
\btxnamefont {R.~\btxlastnamefont {Kumar}}, \btxnamefont {J.~\btxlastnamefont
  {Mendes~Moreira}}\btxandcomma {} \btxandshort {.}\ \btxnamefont
  {J.~\btxlastnamefont {Chandra}}\btxauthorcolon\ \btxjtitlefont
  {\btxifchangecase {Dygcn-lstm: A dynamic gcn-lstm based encoder-decoder
  framework for multistep traffic prediction}{DyGCN-LSTM: A dynamic GCN-LSTM
  based encoder-decoder framework for multistep traffic prediction}}.
\newblock \btxjournalfont {Applied Intelligence}, 53(21):25388 – 25411, 2023.
\newblock {\latintext
  \btxurlfont{https://www.scopus.com/inward/record.uri?eid=2-s2.0-85167340209&doi=10.1007

\bibitem {art_430}
\btxnamefont {S.~\btxlastnamefont {Kwak}}, \btxnamefont {D.~\btxlastnamefont
  {Li}}\btxandcomma {} \btxandshort {.}\ \btxnamefont {N.~\btxlastnamefont
  {Geroliminis}}\btxauthorcolon\ \btxjtitlefont {\btxifchangecase {A two-level
  resolution neural network with enhanced interpretability for freeway traffic
  forecasting}{A two-level resolution neural network with enhanced
  interpretability for freeway traffic forecasting}}.
\newblock \btxjournalfont {Sci. Rep.}, 14(1):31624,
  \btxprintmonthyear{.}{12}{2024}{short}.

\bibitem {art_110}
\btxnamefont {M.~\btxlastnamefont {Lablack}} \btxandshort {.}\ \btxnamefont
  {Y.~\btxlastnamefont {Shen}}\btxauthorcolon\ \btxjtitlefont {\btxifchangecase
  {Spatio-temporal graph mixformer for traffic forecasting}{Spatio-temporal
  graph mixformer for traffic forecasting}}.
\newblock \btxjournalfont {Expert Systems with Applications}, 228, 2023.
\newblock {\latintext
  \btxurlfont{https://www.scopus.com/inward/record.uri?eid=2-s2.0-85159101478&doi=10.1016

\bibitem {visibility_graph}
\btxnamefont {L.~\btxlastnamefont {Lacasa}}, \btxnamefont {B.~\btxlastnamefont
  {Luque}}, \btxnamefont {F.~\btxlastnamefont {Ballesteros}}, \btxnamefont
  {J.~\btxlastnamefont {Luque}}\btxandcomma {} \btxandshort {.}\ \btxnamefont
  {J.\btxfnamespaceshort C. \btxlastnamefont {Nuño}}\btxauthorcolon\
  \btxjtitlefont {\btxifchangecase {From time series to complex networks: The
  visibility graph}{From time series to complex networks: The visibility
  graph}}.
\newblock \btxjournalfont {Proceedings of the National Academy of Sciences},
  105(13):4972–4975, \btxprintmonthyear{.}{4}{2008}{short}\ifbtxprintISSN {,
  \mbox{\btxISSN~\btxISSNfont {1091-6490}}}.
\newblock {\latintext \btxurlfont{http://dx.doi.org/10.1073/pnas.0709247105}}.

\bibitem {LSTNet}
\btxnamefont {G.~\btxlastnamefont {Lai}}, \btxnamefont {W.\btxfnamespaceshort
  C. \btxlastnamefont {Chang}}, \btxnamefont {Y.~\btxlastnamefont
  {Yang}}\btxandcomma {} \btxandshort {.}\ \btxnamefont {H.~\btxlastnamefont
  {Liu}}\btxauthorcolon\ \btxtitlefont {\btxifchangecase {Modeling long- and
  short-term temporal patterns with deep neural networks}{Modeling Long- and
  Short-Term Temporal Patterns with Deep Neural Networks}}.
\newblock \Btxinshort {.}\ \btxtitlefont {The 41st International ACM SIGIR
  Conference on Research \& Development in Information Retrieval}, SIGIR '18,
  \btxpageshort {.}\ 95–104, New York, NY, USA, 2018. \btxpublisherfont
  {Association for Computing Machinery}\ifbtxprintISBN {,
  \mbox{\btxISBN~\btxISBNfont {9781450356572}}}.
\newblock {\latintext \btxurlfont{https://doi.org/10.1145/3209978.3210006}}.

\bibitem {art_675}
\btxnamefont {S.~\btxlastnamefont {Lan}}, \btxnamefont {Y.~\btxlastnamefont
  {Ma}}, \btxnamefont {W.~\btxlastnamefont {Huang}}, \btxnamefont
  {W.~\btxlastnamefont {Wang}}, \btxnamefont {H.~\btxlastnamefont
  {Yang}}\btxandcomma {} \btxandshort {.}\ \btxnamefont {P.~\btxlastnamefont
  {Li}}\btxauthorcolon\ \btxtitlefont {\btxifchangecase {{DSTAGNN}: Dynamic
  spatial-temporal aware graph neural network for traffic flow
  forecasting}{{DSTAGNN}: Dynamic Spatial-Temporal Aware Graph Neural Network
  for Traffic Flow Forecasting}}.
\newblock \Btxinshort {.}\ \btxnamefont {K.~\btxlastnamefont {Chaudhuri}},
  \btxnamefont {S.~\btxlastnamefont {Jegelka}}, \btxnamefont
  {L.~\btxlastnamefont {Song}}, \btxnamefont {C.~\btxlastnamefont
  {Szepesvari}}, \btxnamefont {G.~\btxlastnamefont {Niu}}\btxandcomma {}
  \btxandshort {.}\ \btxnamefont {S.~\btxlastnamefont {Sabato}}\
  (\btxeditorsshort {.}): \btxtitlefont {Proceedings of the 39th International
  Conference on Machine Learning}, \btxvolumeshort {.}\ \btxvolumefont {162}
  \btxofseriesshort {.}\ \btxtitlefont {Proceedings of Machine Learning
  Research}, \btxpagesshort {.}\ 11906--11917. \btxpublisherfont {PMLR},
  \btxprintmonthyear{.}{17--23 Jul}{2022}{short}.
\newblock {\latintext
  \btxurlfont{https://proceedings.mlr.press/v162/lan22a.html}}.

\bibitem {art_748}
\btxnamefont {J.~\btxlastnamefont {Lee}}, \btxnamefont {B.~\btxlastnamefont
  {Park}}\btxandcomma {} \btxandshort {.}\ \btxnamefont {D.\btxfnamespaceshort
  K. \btxlastnamefont {Chae}}\btxauthorcolon\ \btxtitlefont {\btxifchangecase
  {{DuoGAT}: Dual time-oriented graph attention networks for accurate,
  efficient and explainable anomaly detection on time-series}{{DuoGAT}: Dual
  time-oriented graph attention networks for accurate, efficient and
  explainable anomaly detection on time-series}}.
\newblock \Btxinshort {.}\ \btxtitlefont {Proceedings of the 32nd {ACM}
  International Conference on Information and Knowledge Management}, New York,
  NY, USA, \btxprintmonthyear{.}{10}{2023}{short}. \btxpublisherfont {ACM}.

\bibitem {art_378}
\btxnamefont {B.~\btxlastnamefont {Lei}} \btxandshort {.}\ \btxnamefont
  {Y.~\btxlastnamefont {Song}}\btxauthorcolon\ \btxjtitlefont {\btxifchangecase
  {Volatility forecasting for stock market incorporating media reports,
  investors' sentiment, and attention based on {MTGNN} model}{Volatility
  forecasting for stock market incorporating media reports, investors'
  sentiment, and attention based on {MTGNN} model}}.
\newblock \btxjournalfont {J. Forecast.}, 43(5):1706--1730,
  \btxprintmonthyear{.}{8}{2024}{short}.

\bibitem {art_192}
\btxnamefont {F.~\btxlastnamefont {Lei}}, \btxnamefont {X.~\btxlastnamefont
  {Luo}}, \btxnamefont {Z.~\btxlastnamefont {Chen}}\btxandcomma {} \btxandshort
  {.}\ \btxnamefont {H.~\btxlastnamefont {Zhou}}\btxauthorcolon\ \btxjtitlefont
  {\btxifchangecase {Signal feature extract based on dual-channel wavelet
  convolutional network mixed with hypergraph convolutional network for fault
  diagnosis}{Signal Feature Extract Based on Dual-Channel Wavelet Convolutional
  Network Mixed With Hypergraph Convolutional Network for Fault Diagnosis}}.
\newblock \btxjournalfont {IEEE Sensors Journal}, 23(22):28378--28389, 2023.

\bibitem {art_55}
\btxnamefont {C.~\btxlastnamefont {Li}}, \btxnamefont {L.~\btxlastnamefont
  {Mo}}\btxandcomma {} \btxandshort {.}\ \btxnamefont {R.~\btxlastnamefont
  {Yan}}\btxauthorcolon\ \btxjtitlefont {\btxifchangecase {Fault diagnosis of
  rolling bearing based on whvg and gcn}{Fault Diagnosis of Rolling Bearing
  Based on WHVG and GCN}}.
\newblock \btxjournalfont {IEEE Transactions on Instrumentation and
  Measurement}, PP:1--1, \btxprintmonthyear{.}{06}{2021}{short}.

\bibitem {art_444}
\btxnamefont {C.~\btxlastnamefont {Li}}, \btxnamefont {Y.~\btxlastnamefont
  {Zhao}}\btxandcomma {} \btxandshort {.}\ \btxnamefont {Z.~\btxlastnamefont
  {Zhang}}\btxauthorcolon\ \btxjtitlefont {\btxifchangecase {{AMGCN}: adaptive
  multigraph convolutional networks for traffic speed forecasting}{{AMGCN}:
  adaptive multigraph convolutional networks for traffic speed forecasting}}.
\newblock \btxjournalfont {Appl. Intell.}, 54(3):2594--2613,
  \btxprintmonthyear{.}{2}{2024}{short}.

\bibitem {MAD-GAN}
\btxnamefont {D.~\btxlastnamefont {Li}}, \btxnamefont {D.~\btxlastnamefont
  {Chen}}, \btxnamefont {B.~\btxlastnamefont {Jin}}, \btxnamefont
  {L.~\btxlastnamefont {Shi}}, \btxnamefont {J.~\btxlastnamefont
  {Goh}}\btxandcomma {} \btxandshort {.}\ \btxnamefont {S.\btxfnamespaceshort
  K. \btxlastnamefont {Ng}}\btxauthorcolon\ \btxtitlefont {\btxifchangecase
  {Mad-gan: Multivariate anomaly detection for time series data with generative
  adversarial networks}{MAD-GAN: Multivariate Anomaly Detection for Time Series
  Data with Generative Adversarial Networks}}.
\newblock \Btxinshort {.}\ \btxnamefont {I.\btxfnamespaceshort V.
  \btxlastnamefont {Tetko}}, \btxnamefont {V.~\btxlastnamefont
  {K{\r{u}}rkov{\'a}}}, \btxnamefont {P.~\btxlastnamefont {Karpov}}\btxandcomma
  {} \btxandshort {.}\ \btxnamefont {F.~\btxlastnamefont {Theis}}\
  (\btxeditorsshort {.}): \btxtitlefont {Artificial Neural Networks and Machine
  Learning -- ICANN 2019: Text and Time Series}, \btxpagesshort {.}\ 703--716,
  Cham, 2019. \btxpublisherfont {Springer International
  Publishing}\ifbtxprintISBN {, \mbox{\btxISBN~\btxISBNfont
  {978-3-030-30490-4}}}.

\bibitem {art_30}
\btxnamefont {D.~\btxlastnamefont {Li}}, \btxnamefont {W.~\btxlastnamefont
  {Zhao}}, \btxnamefont {J.~\btxlastnamefont {Hu}}, \btxnamefont
  {S.~\btxlastnamefont {Zhao}}\btxandcomma {} \btxandshort {.}\ \btxnamefont
  {S.~\btxlastnamefont {Liu}}\btxauthorcolon\ \btxjtitlefont {\btxifchangecase
  {A long-term water quality prediction model for marine ranch based on
  time-graph convolutional neural network}{A long-term water quality prediction
  model for marine ranch based on time-graph convolutional neural network}}.
\newblock \btxjournalfont {Ecological Indicators}, 154, 2023.
\newblock {\latintext
  \btxurlfont{https://www.scopus.com/inward/record.uri?eid=2-s2.0-85169001765&doi=10.1016

\bibitem {art_96}
\btxnamefont {F.~\btxlastnamefont {Li}}, \btxnamefont {J.~\btxlastnamefont
  {Feng}}, \btxnamefont {H.~\btxlastnamefont {Yan}}, \btxnamefont
  {D.~\btxlastnamefont {Jin}}\btxandcomma {} \btxandshort {.}\ \btxnamefont
  {Y.~\btxlastnamefont {Li}}\btxauthorcolon\ \btxjtitlefont {\btxifchangecase
  {Crowd flow prediction for irregular regions with semantic graph attention
  network}{Crowd Flow Prediction for Irregular Regions with Semantic Graph
  Attention Network}}.
\newblock \btxjournalfont {ACM Trans. Intell. Syst. Technol.}, 13(5),
  \btxprintmonthyear{.}{jun}{2022}{short}\ifbtxprintISSN {,
  \mbox{\btxISSN~\btxISSNfont {2157-6904}}}.
\newblock {\latintext \btxurlfont{https://doi.org/10.1145/3501805}}.

\bibitem {art_100}
\btxnamefont {F.~\btxlastnamefont {Li}}, \btxnamefont {J.~\btxlastnamefont
  {Feng}}, \btxnamefont {H.~\btxlastnamefont {Yan}}, \btxnamefont
  {G.~\btxlastnamefont {Jin}}, \btxnamefont {F.~\btxlastnamefont {Yang}},
  \btxnamefont {F.~\btxlastnamefont {Sun}}, \btxnamefont {D.~\btxlastnamefont
  {Jin}}\btxandcomma {} \btxandshort {.}\ \btxnamefont {Y.~\btxlastnamefont
  {Li}}\btxauthorcolon\ \btxjtitlefont {\btxifchangecase {Dynamic graph
  convolutional recurrent network for traffic prediction: Benchmark and
  solution}{Dynamic Graph Convolutional Recurrent Network for Traffic
  Prediction: Benchmark and Solution}}.
\newblock \btxjournalfont {ACM Trans. Knowl. Discov. Data}, 17(1),
  \btxprintmonthyear{.}{feb}{2023}{short}\ifbtxprintISSN {,
  \mbox{\btxISSN~\btxISSNfont {1556-4681}}}.
\newblock {\latintext \btxurlfont{https://doi.org/10.1145/3532611}}.

\bibitem {art_720}
\btxnamefont {F.~\btxlastnamefont {Li}}, \btxnamefont {H.~\btxlastnamefont
  {Yan}}, \btxnamefont {G.~\btxlastnamefont {Jin}}, \btxnamefont
  {Y.~\btxlastnamefont {Liu}}, \btxnamefont {Y.~\btxlastnamefont
  {Li}}\btxandcomma {} \btxandshort {.}\ \btxnamefont {D.~\btxlastnamefont
  {Jin}}\btxauthorcolon\ \btxtitlefont {\btxifchangecase {Automated
  spatio-temporal synchronous modeling with multiple graphs for traffic
  prediction}{Automated spatio-temporal synchronous modeling with multiple
  graphs for traffic prediction}}.
\newblock \Btxinshort {.}\ \btxtitlefont {Proceedings of the 31st {ACM}
  International Conference on Information \& Knowledge Management}, New York,
  NY, USA, \btxprintmonthyear{.}{10}{2022}{short}. \btxpublisherfont {ACM}.

\bibitem {art_731}
\btxnamefont {F.~\btxlastnamefont {Li}}, \btxnamefont {H.~\btxlastnamefont
  {Yan}}, \btxnamefont {H.~\btxlastnamefont {Sui}}, \btxnamefont
  {D.~\btxlastnamefont {Wang}}, \btxnamefont {F.~\btxlastnamefont {Zuo}},
  \btxnamefont {Y.~\btxlastnamefont {Liu}}, \btxnamefont {Y.~\btxlastnamefont
  {Li}}\btxandcomma {} \btxandshort {.}\ \btxnamefont {D.~\btxlastnamefont
  {Jin}}\btxauthorcolon\ \btxtitlefont {\btxifchangecase {Periodic shift and
  event-aware spatio-temporal graph convolutional network for traffic
  congestion prediction}{Periodic shift and event-aware spatio-temporal graph
  convolutional network for traffic congestion prediction}}.
\newblock \Btxinshort {.}\ \btxtitlefont {Proceedings of the 31st {ACM}
  International Conference on Advances in Geographic Information Systems}, New
  York, NY, USA, \btxprintmonthyear{.}{11}{2023}{short}. \btxpublisherfont
  {ACM}.

\bibitem {ShapeNet}
\btxnamefont {G.~\btxlastnamefont {Li}}, \btxnamefont {B.~\btxlastnamefont
  {Choi}}, \btxnamefont {J.~\btxlastnamefont {Xu}}, \btxnamefont
  {S.~\btxlastnamefont {S~Bhowmick}}, \btxnamefont {K.\btxfnamespaceshort P.
  \btxlastnamefont {Chun}}\btxandcomma {} \btxandshort {.}\ \btxnamefont
  {G.\btxfnamespaceshort L.\btxfnamespaceshort H. \btxlastnamefont
  {Wong}}\btxauthorcolon\ \btxjtitlefont {\btxifchangecase {Shapenet: A
  shapelet-neural network approach for multivariate time series
  classification}{ShapeNet: A Shapelet-Neural Network Approach for Multivariate
  Time Series Classification}}.
\newblock \btxjournalfont {Proceedings of the AAAI Conference on Artificial
  Intelligence}, 35(9):8375--8383, \btxprintmonthyear{.}{May}{2021}{short}.
\newblock {\latintext
  \btxurlfont{https://ojs.aaai.org/index.php/AAAI/article/view/17018}}.

\bibitem {art_27}
\btxnamefont {H.~\btxlastnamefont {Li}}\btxauthorcolon\ \btxjtitlefont
  {\btxifchangecase {Short-term wind power prediction via spatial temporal
  analysis and deep residual networks}{Short-Term Wind Power Prediction via
  Spatial Temporal Analysis and Deep Residual Networks}}.
\newblock \btxjournalfont {Frontiers in Energy Research}, 10, 2022.
\newblock {\latintext
  \btxurlfont{https://www.scopus.com/inward/record.uri?eid=2-s2.0-85130333390&doi=10.3389

\bibitem {art_109}
\btxnamefont {H.~\btxlastnamefont {Li}}, \btxnamefont {D.~\btxlastnamefont
  {Jin}}, \btxnamefont {X.~\btxlastnamefont {Li}}, \btxnamefont
  {H.~\btxlastnamefont {Huang}}, \btxnamefont {J.~\btxlastnamefont
  {Yun}}\btxandcomma {} \btxandshort {.}\ \btxnamefont {L.~\btxlastnamefont
  {Huang}}\btxauthorcolon\ \btxjtitlefont {\btxifchangecase {Multi-view
  spatial-temporal graph neural network for traffic prediction}{Multi-View
  Spatial-Temporal Graph Neural Network for Traffic Prediction}}.
\newblock \btxjournalfont {Computer Journal}, 66(10):2393 – 2408, 2023.
\newblock {\latintext
  \btxurlfont{https://www.scopus.com/inward/record.uri?eid=2-s2.0-85150044802&doi=10.1093

\bibitem {art_99}
\btxnamefont {H.~\btxlastnamefont {Li}}, \btxnamefont {D.~\btxlastnamefont
  {Jin}}, \btxnamefont {X.~\btxlastnamefont {Li}}, \btxnamefont
  {J.~\btxlastnamefont {Huang}}, \btxnamefont {X.~\btxlastnamefont {Ma}},
  \btxnamefont {J.~\btxlastnamefont {Cui}}, \btxnamefont {D.~\btxlastnamefont
  {Huang}}, \btxnamefont {S.~\btxlastnamefont {Qiao}}\btxandcomma {}
  \btxandshort {.}\ \btxnamefont {J.~\btxlastnamefont {Yoo}}\btxauthorcolon\
  \btxjtitlefont {\btxifchangecase {Dmgf-net: An efficient dynamic multi-graph
  fusion network for traffic prediction}{DMGF-Net: An Efficient Dynamic
  Multi-Graph Fusion Network for Traffic Prediction}}.
\newblock \btxjournalfont {ACM Trans. Knowl. Discov. Data}, 17(7),
  \btxprintmonthyear{.}{apr}{2023}{short}\ifbtxprintISSN {,
  \mbox{\btxISSN~\btxISSNfont {1556-4681}}}.
\newblock {\latintext \btxurlfont{https://doi.org/10.1145/3586164}}.

\bibitem {art_730}
\btxnamefont {H.~\btxlastnamefont {Li}}, \btxnamefont {D.~\btxlastnamefont
  {Jin}}, \btxnamefont {X.~\btxlastnamefont {Li}}, \btxnamefont
  {J.~\btxlastnamefont {Huang}}\btxandcomma {} \btxandshort {.}\ \btxnamefont
  {J.~\btxlastnamefont {Yoo}}\btxauthorcolon\ \btxtitlefont {\btxifchangecase
  {Multi-task synchronous graph neural networks for traffic spatial-temporal
  prediction}{Multi-task synchronous graph neural networks for traffic
  spatial-temporal prediction}}.
\newblock \Btxinshort {.}\ \btxtitlefont {Proceedings of the 29th International
  Conference on Advances in Geographic Information Systems}, New York, NY, USA,
  \btxprintmonthyear{.}{11}{2021}{short}. \btxpublisherfont {ACM}.

\bibitem {art_98}
\btxnamefont {H.~\btxlastnamefont {Li}}, \btxnamefont {X.~\btxlastnamefont
  {Li}}, \btxnamefont {L.~\btxlastnamefont {Su}}, \btxnamefont
  {D.~\btxlastnamefont {Jin}}, \btxnamefont {J.~\btxlastnamefont
  {Huang}}\btxandcomma {} \btxandshort {.}\ \btxnamefont {D.~\btxlastnamefont
  {Huang}}\btxauthorcolon\ \btxjtitlefont {\btxifchangecase {Deep
  spatio-temporal adaptive 3d convolutional neural networks for traffic flow
  prediction}{Deep Spatio-temporal Adaptive 3D Convolutional Neural Networks
  for Traffic Flow Prediction}}.
\newblock \btxjournalfont {ACM Trans. Intell. Syst. Technol.}, 13(2),
  \btxprintmonthyear{.}{jan}{2022}{short}\ifbtxprintISSN {,
  \mbox{\btxISSN~\btxISSNfont {2157-6904}}}.
\newblock {\latintext \btxurlfont{https://doi.org/10.1145/3510829}}.

\bibitem {art_9}
\btxnamefont {H.~\btxlastnamefont {Li}}, \btxnamefont {X.~\btxlastnamefont
  {Wang}}, \btxnamefont {Z.~\btxlastnamefont {Yang}}, \btxnamefont
  {S.~\btxlastnamefont {Ali}}, \btxnamefont {N.~\btxlastnamefont
  {Tong}}\btxandcomma {} \btxandshort {.}\ \btxnamefont {S.~\btxlastnamefont
  {Baseer}}\btxauthorcolon\ \btxjtitlefont {\btxifchangecase {Correlation-based
  anomaly detection method for multi-sensor system}{Correlation-Based Anomaly
  Detection Method for Multi-sensor System}}.
\newblock \btxjournalfont {Computational Intelligence and Neuroscience}, 2022,
  2022.
\newblock {\latintext
  \btxurlfont{https://www.scopus.com/inward/record.uri?eid=2-s2.0-85131710989&doi=10.1155

\bibitem {art_724}
\btxnamefont {H.~\btxlastnamefont {Li}}, \btxnamefont {S.~\btxlastnamefont
  {Zhang}}, \btxnamefont {X.~\btxlastnamefont {Li}}, \btxnamefont
  {L.~\btxlastnamefont {Su}}, \btxnamefont {H.~\btxlastnamefont {Huang}},
  \btxnamefont {D.~\btxlastnamefont {Jin}}, \btxnamefont {L.~\btxlastnamefont
  {Chen}}, \btxnamefont {J.~\btxlastnamefont {Huang}}\btxandcomma {}
  \btxandshort {.}\ \btxnamefont {J.~\btxlastnamefont {Yoo}}\btxauthorcolon\
  \btxtitlefont {\btxifchangecase {Detectornet: Transformer-enhanced spatial
  temporal graph neural network for traffic prediction}{DetectorNet:
  Transformer-enhanced Spatial Temporal Graph Neural Network for Traffic
  Prediction}}.
\newblock \Btxinshort {.}\ \btxtitlefont {Proceedings of the 29th International
  Conference on Advances in Geographic Information Systems}, SIGSPATIAL '21,
  \btxpageshort {.}\ 133–136, New York, NY, USA, 2021. \btxpublisherfont
  {Association for Computing Machinery}\ifbtxprintISBN {,
  \mbox{\btxISBN~\btxISBNfont {9781450386647}}}.
\newblock {\latintext \btxurlfont{https://doi.org/10.1145/3474717.3483920}}.

\bibitem {art_4}
\btxnamefont {J.~\btxlastnamefont {Li}}, \btxnamefont {J.~\btxlastnamefont
  {Crooks}}, \btxnamefont {J.~\btxlastnamefont {Murdock}}, \btxnamefont
  {P.~de~\btxlastnamefont {Souza}}, \btxnamefont {K.~\btxlastnamefont
  {Hohsfield}}, \btxnamefont {B.~\btxlastnamefont {Obermann}}\btxandcomma {}
  \btxandshort {.}\ \btxnamefont {T.~\btxlastnamefont
  {Stockman}}\btxauthorcolon\ \btxjtitlefont {\btxifchangecase {A nested
  machine learning approach to short-term pm2.5 prediction in metropolitan
  areas using pm2.5 data from different sensor networks}{A nested machine
  learning approach to short-term PM2.5 prediction in metropolitan areas using
  PM2.5 data from different sensor networks}}.
\newblock \btxjournalfont {Science of the Total Environment}, 873, 2023.
\newblock {\latintext
  \btxurlfont{https://www.scopus.com/inward/record.uri?eid=2-s2.0-85148685120&doi=10.1016

\bibitem {art_480}
\btxnamefont {J.~\btxlastnamefont {Li}}, \btxnamefont {K.~\btxlastnamefont
  {Hao}}, \btxnamefont {X.~\btxlastnamefont {Shi}}, \btxnamefont
  {L.~\btxlastnamefont {Chen}}\btxandcomma {} \btxandshort {.}\ \btxnamefont
  {R.~\btxlastnamefont {Xie}}\btxauthorcolon\ \btxjtitlefont {\btxifchangecase
  {Community inspired edge specific message graph convolution network for
  predictive monitoring of large-scale polymerization processes}{Community
  inspired edge specific message graph convolution network for predictive
  monitoring of large-scale polymerization processes}}.
\newblock \btxjournalfont {Control Eng. Pract.}, 151(106020):106020,
  \btxprintmonthyear{.}{10}{2024}{short}.

\bibitem {art_64}
\btxnamefont {J.~\btxlastnamefont {Li}} \btxandshort {.}\ \btxnamefont
  {X.~\btxlastnamefont {Yao}}\btxauthorcolon\ \btxjtitlefont {\btxifchangecase
  {Corporate investment prediction using a weighted temporal graph neural
  network}{Corporate investment prediction using a weighted temporal graph
  neural network}}.
\newblock \btxjournalfont {Wiley Interdisciplinary Reviews: Data Mining and
  Knowledge Discovery}, 12(6), 2022.
\newblock {\latintext
  \btxurlfont{https://www.scopus.com/inward/record.uri?eid=2-s2.0-85133454187&doi=10.1002

\bibitem {art_10}
\btxnamefont {K.~\btxlastnamefont {Li}}, \btxnamefont {T.~\btxlastnamefont
  {Zhang}}, \btxnamefont {W.~\btxlastnamefont {Dong}}\btxandcomma {}
  \btxandshort {.}\ \btxnamefont {H.~\btxlastnamefont {Ye}}\btxauthorcolon\
  \btxjtitlefont {\btxifchangecase {Abnormality detection of blast furnace
  ironmaking process based on an improved diffusion convolutional gated
  recurrent unit network}{Abnormality Detection of Blast Furnace Ironmaking
  Process Based on an Improved Diffusion Convolutional Gated Recurrent Unit
  Network}}.
\newblock \btxjournalfont {IEEE Transactions on Instrumentation and
  Measurement}, 72:1--12, 2023.

\bibitem {art_406}
\btxnamefont {L.~\btxlastnamefont {Li}}, \btxnamefont {X.~\btxlastnamefont
  {Zhou}}, \btxnamefont {G.~\btxlastnamefont {Hu}}, \btxnamefont
  {S.~\btxlastnamefont {Li}}\btxandcomma {} \btxandshort {.}\ \btxnamefont
  {D.~\btxlastnamefont {Jia}}\btxauthorcolon\ \btxjtitlefont {\btxifchangecase
  {A recurrent spatio-temporal graph neural network based on latent time graph
  for multi-channel time series forecasting}{A recurrent spatio-temporal graph
  neural network based on latent time graph for multi-channel time series
  forecasting}}.
\newblock \btxjournalfont {IEEE Signal Process. Lett.}, 31:2875--2879, 2024.

\bibitem {STFGNN}
\btxnamefont {M.~\btxlastnamefont {Li}} \btxandshort {.}\ \btxnamefont
  {Z.~\btxlastnamefont {Zhu}}\btxauthorcolon\ \btxjtitlefont {\btxifchangecase
  {Spatial-temporal fusion graph neural networks for traffic flow
  forecasting}{Spatial-Temporal Fusion Graph Neural Networks for Traffic Flow
  Forecasting}}.
\newblock \btxjournalfont {Proceedings of the AAAI Conference on Artificial
  Intelligence}, 35(5):4189--4196, \btxprintmonthyear{.}{May}{2021}{short}.
\newblock {\latintext
  \btxurlfont{https://ojs.aaai.org/index.php/AAAI/article/view/16542}}.

\bibitem {art_355}
\btxnamefont {Q.~\btxlastnamefont {Li}}, \btxnamefont {Y.~\btxlastnamefont
  {Chen}}, \btxnamefont {X.~\btxlastnamefont {Dang}}, \btxnamefont
  {H.~\btxlastnamefont {Yin}}, \btxnamefont {G.~\btxlastnamefont
  {Xu}}\btxandcomma {} \btxandshort {.}\ \btxnamefont {X.~\btxlastnamefont
  {Chen}}\btxauthorcolon\ \btxjtitlefont {\btxifchangecase {Sea clutter
  prediction based on fusion of fourier transform and graph neural network}{Sea
  clutter prediction based on fusion of Fourier transform and graph neural
  network}}.
\newblock \btxjournalfont {Int. J. Remote Sens.}, 45(18):6544--6571,
  \btxprintmonthyear{.}{9}{2024}{short}.

\bibitem {HAGCN}
\btxnamefont {T.~\btxlastnamefont {Li}}, \btxnamefont {Z.~\btxlastnamefont
  {Zhao}}, \btxnamefont {C.~\btxlastnamefont {Sun}}, \btxnamefont
  {R.~\btxlastnamefont {Yan}}\btxandcomma {} \btxandshort {.}\ \btxnamefont
  {X.~\btxlastnamefont {Chen}}\btxauthorcolon\ \btxjtitlefont {\btxifchangecase
  {Hierarchical attention graph convolutional network to fuse multi-sensor
  signals for remaining useful life prediction}{Hierarchical attention graph
  convolutional network to fuse multi-sensor signals for remaining useful life
  prediction}}.
\newblock \btxjournalfont {Reliability Engineering and System Safety},
  215:107878, \btxprintmonthyear{.}{11}{2021}{short}\ifbtxprintISSN {,
  \mbox{\btxISSN~\btxISSNfont {0951-8320}}}.
\newblock {\latintext
  \btxurlfont{http://dx.doi.org/10.1016/j.ress.2021.107878}}.

\bibitem {art_393}
\btxnamefont {X.~\btxlastnamefont {Li}} \btxandshort {.}\ \btxnamefont
  {G.~\btxlastnamefont {Tang}}\btxauthorcolon\ \btxjtitlefont {\btxifchangecase
  {Multivariate sequence prediction for graph convolutional networks based on
  {ESMD} and transfer entropy}{Multivariate sequence prediction for graph
  convolutional networks based on {ESMD} and transfer entropy}}.
\newblock \btxjournalfont {Multimed. Tools Appl.},
  \btxprintmonthyear{.}{3}{2024}{short}.

\bibitem {art_70}
\btxnamefont {X.~\btxlastnamefont {Li}}, \btxnamefont {Z.~\btxlastnamefont
  {Wang}}, \btxnamefont {X.~\btxlastnamefont {Chen}}, \btxnamefont
  {B.~\btxlastnamefont {Guo}}\btxandcomma {} \btxandshort {.}\ \btxnamefont
  {Z.~\btxlastnamefont {Yu}}\btxauthorcolon\ \btxjtitlefont {\btxifchangecase
  {A hybrid continuous-time dynamic graph representation learning model by
  exploring both temporal and repetitive information}{A Hybrid Continuous-Time
  Dynamic Graph Representation Learning Model by Exploring Both Temporal and
  Repetitive Information}}.
\newblock \btxjournalfont {ACM Trans. Knowl. Discov. Data}, 17(9),
  \btxprintmonthyear{.}{jun}{2023}{short}\ifbtxprintISSN {,
  \mbox{\btxISSN~\btxISSNfont {1556-4681}}}.
\newblock {\latintext \btxurlfont{https://doi.org/10.1145/3596447}}.

\bibitem {HyDCNN}
\btxnamefont {Y.~\btxlastnamefont {Li}}, \btxnamefont {K.~\btxlastnamefont
  {Li}}, \btxnamefont {C.~\btxlastnamefont {Chen}}, \btxnamefont
  {X.~\btxlastnamefont {Zhou}}, \btxnamefont {Z.~\btxlastnamefont
  {Zeng}}\btxandcomma {} \btxandshort {.}\ \btxnamefont {K.~\btxlastnamefont
  {Li}}\btxauthorcolon\ \btxjtitlefont {\btxifchangecase {Modeling temporal
  patterns with dilated convolutions for time-series forecasting}{Modeling
  Temporal Patterns with Dilated Convolutions for Time-Series Forecasting}}.
\newblock \btxjournalfont {ACM Trans. Knowl. Discov. Data}, 16(1),
  \btxprintmonthyear{.}{7}{2021}{short}\ifbtxprintISSN {,
  \mbox{\btxISSN~\btxISSNfont {1556-4681}}}.
\newblock {\latintext \btxurlfont{https://doi.org/10.1145/3453724}}.

\bibitem {BiHDM}
\btxnamefont {Y.~\btxlastnamefont {Li}}, \btxnamefont {L.~\btxlastnamefont
  {Wang}}, \btxnamefont {W.~\btxlastnamefont {Zheng}}, \btxnamefont
  {Y.~\btxlastnamefont {Zong}}, \btxnamefont {L.~\btxlastnamefont {Qi}},
  \btxnamefont {Z.~\btxlastnamefont {Cui}}, \btxnamefont {T.~\btxlastnamefont
  {Zhang}}\btxandcomma {} \btxandshort {.}\ \btxnamefont {T.~\btxlastnamefont
  {Song}}\btxauthorcolon\ \btxjtitlefont {\btxifchangecase {A novel
  bi-hemispheric discrepancy model for eeg emotion recognition}{A Novel
  Bi-Hemispheric Discrepancy Model for EEG Emotion Recognition}}.
\newblock \btxjournalfont {IEEE Transactions on Cognitive and Developmental
  Systems}, 13(2):354--367, 2021.

\bibitem {art_50}
\btxnamefont {Y.~\btxlastnamefont {Li}}, \btxnamefont {Y.~\btxlastnamefont
  {Wang}}\btxandcomma {} \btxandshort {.}\ \btxnamefont {K.~\btxlastnamefont
  {Ma}}\btxauthorcolon\ \btxjtitlefont {\btxifchangecase {Integrating
  transformer and gcn for covid-19 forecasting}{Integrating Transformer and GCN
  for COVID-19 Forecasting}}.
\newblock \btxjournalfont {Sustainability (Switzerland)}, 14(16), 2022.
\newblock {\latintext
  \btxurlfont{https://www.scopus.com/inward/record.uri?eid=2-s2.0-85137680940&doi=10.3390

\bibitem {li2023graphneuralnetworkspatiotemporal}
\btxnamefont {Y.~\btxlastnamefont {Li}}, \btxnamefont {D.~\btxlastnamefont
  {Yu}}, \btxnamefont {Z.~\btxlastnamefont {Liu}}, \btxnamefont
  {M.~\btxlastnamefont {Zhang}}, \btxnamefont {X.~\btxlastnamefont
  {Gong}}\btxandcomma {} \btxandshort {.}\ \btxnamefont {L.~\btxlastnamefont
  {Zhao}}\btxauthorcolon\ \btxtitlefont {\btxifchangecase {Graph neural network
  for spatiotemporal data: methods and applications}{Graph Neural Network for
  spatiotemporal data: methods and applications}}, 2023.
\newblock {\latintext \btxurlfont{https://arxiv.org/abs/2306.00012}}.

\bibitem {DCRNN}
\btxnamefont {Y.~\btxlastnamefont {Li}}, \btxnamefont {R.~\btxlastnamefont
  {Yu}}, \btxnamefont {C.~\btxlastnamefont {Shahabi}}\btxandcomma {}
  \btxandshort {.}\ \btxnamefont {Y.~\btxlastnamefont {Liu}}\btxauthorcolon\
  \btxjtitlefont {\btxifchangecase {Diffusion convolutional recurrent neural
  network: Data-driven traffic forecasting}{Diffusion Convolutional Recurrent
  Neural Network: Data-Driven Traffic Forecasting}}.
\newblock 2018.

\bibitem {Bi-DANN}
\btxnamefont {Y.~\btxlastnamefont {Li}}, \btxnamefont {W.~\btxlastnamefont
  {Zheng}}, \btxnamefont {Y.~\btxlastnamefont {Zong}}, \btxnamefont
  {Z.~\btxlastnamefont {Cui}}, \btxnamefont {T.~\btxlastnamefont
  {Zhang}}\btxandcomma {} \btxandshort {.}\ \btxnamefont {X.~\btxlastnamefont
  {Zhou}}\btxauthorcolon\ \btxjtitlefont {\btxifchangecase {A bi-hemisphere
  domain adversarial neural network model for eeg emotion recognition}{A
  Bi-Hemisphere Domain Adversarial Neural Network Model for EEG Emotion
  Recognition}}.
\newblock \btxjournalfont {IEEE Transactions on Affective Computing},
  12(2):494--504, 2021.

\bibitem {art_747}
\btxnamefont {Y.~\btxlastnamefont {Li}}, \btxnamefont {Z.~\btxlastnamefont
  {Zhou}}, \btxnamefont {S.~\btxlastnamefont {Deng}}, \btxnamefont
  {X.~\btxlastnamefont {Sun}}, \btxnamefont {X.~\btxlastnamefont {Xue}},
  \btxnamefont {S.~\btxlastnamefont {Yangui}}\btxandcomma {} \btxandshort {.}\
  \btxnamefont {W.~\btxlastnamefont {Gaaloul}}\btxauthorcolon\ \btxtitlefont
  {\btxifchangecase {Accurate anomaly detection leveraging knowledge-enhanced
  {GAT}}{Accurate anomaly detection leveraging knowledge-enhanced {GAT}}}.
\newblock \Btxinshort {.}\ \btxtitlefont {2024 {IEEE} International Conference
  on Web Services ({ICWS})}, \btxpagesshort {.}\ 568--577. \btxpublisherfont
  {IEEE}, \btxprintmonthyear{.}{7}{2024}{short}.

\bibitem {art_405}
\btxnamefont {Z.~\btxlastnamefont {Li}}, \btxnamefont {Z.~\btxlastnamefont
  {Gao}}, \btxnamefont {G.~\btxlastnamefont {Zhang}}, \btxnamefont
  {J.~\btxlastnamefont {Liu}}\btxandcomma {} \btxandshort {.}\ \btxnamefont
  {L.~\btxlastnamefont {Xu}}\btxauthorcolon\ \btxjtitlefont {\btxifchangecase
  {Dynamic personalized graph neural network with linear complexity for
  multivariate time series forecasting}{Dynamic personalized graph neural
  network with linear complexity for multivariate time series forecasting}}.
\newblock \btxjournalfont {Eng. Appl. Artif. Intell.}, 127(107291):107291,
  \btxprintmonthyear{.}{1}{2024}{short}.

\bibitem {art_73}
\btxnamefont {Z.~\btxlastnamefont {Li}}, \btxnamefont {J.~\btxlastnamefont
  {Yu}}, \btxnamefont {G.~\btxlastnamefont {Zhang}}\btxandcomma {} \btxandshort
  {.}\ \btxnamefont {L.~\btxlastnamefont {Xu}}\btxauthorcolon\ \btxjtitlefont
  {\btxifchangecase {Dynamic spatio-temporal graph network with adaptive
  propagation mechanism for multivariate time series forecasting}{Dynamic
  spatio-temporal graph network with adaptive propagation mechanism for
  multivariate time series forecasting}}.
\newblock \btxjournalfont {Expert Systems with Applications}, 216, 2023.
\newblock {\latintext
  \btxurlfont{https://www.scopus.com/inward/record.uri?eid=2-s2.0-85144433062&doi=10.1016

\bibitem {art_433}
\btxnamefont {Z.~\btxlastnamefont {Li}}, \btxnamefont {J.~\btxlastnamefont
  {Zhou}}, \btxnamefont {Z.~\btxlastnamefont {Lin}}\btxandcomma {} \btxandshort
  {.}\ \btxnamefont {T.~\btxlastnamefont {Zhou}}\btxauthorcolon\ \btxjtitlefont
  {\btxifchangecase {Dynamic spatial aware graph transformer for spatiotemporal
  traffic flow forecasting}{Dynamic spatial aware graph transformer for
  spatiotemporal traffic flow forecasting}}.
\newblock \btxjournalfont {Knowl. Based Syst.}, 297(111946):111946,
  \btxprintmonthyear{.}{8}{2024}{short}.

\bibitem {art_78}
\btxnamefont {Z.\btxfnamespaceshort L. \btxlastnamefont {Li}}, \btxnamefont
  {G.\btxfnamespaceshort W. \btxlastnamefont {Zhang}}, \btxnamefont
  {J.~\btxlastnamefont {Yu}}\btxandcomma {} \btxandshort {.}\ \btxnamefont
  {L.\btxfnamespaceshort Y. \btxlastnamefont {Xu}}\btxauthorcolon\
  \btxjtitlefont {\btxifchangecase {Dynamic graph structure learning for
  multivariate time series forecasting}{Dynamic graph structure learning for
  multivariate time series forecasting}}.
\newblock \btxjournalfont {Pattern Recognition}, 138, 2023.
\newblock {\latintext
  \btxurlfont{https://www.scopus.com/inward/record.uri?eid=2-s2.0-85149751108&doi=10.1016

\bibitem {art_383}
\btxnamefont {G.~\btxlastnamefont {Liang}}, \btxnamefont {P.~\btxlastnamefont
  {Tiwari}}, \btxnamefont {S.~\btxlastnamefont {Nowaczyk}}, \btxnamefont
  {S.~\btxlastnamefont {Byttner}}\btxandcomma {} \btxandshort {.}\ \btxnamefont
  {F.~\btxlastnamefont {Alonso-Fernandez}}\btxauthorcolon\ \btxjtitlefont
  {\btxifchangecase {Dynamic causal explanation based diffusion-variational
  graph neural network for spatiotemporal forecasting}{Dynamic causal
  explanation based diffusion-variational graph neural network for
  spatiotemporal forecasting}}.
\newblock \btxjournalfont {IEEE Trans. Neural Netw. Learn. Syst.}, PP:1--14,
  \btxprintmonthyear{.}{7}{2024}{short}.

\bibitem {art_114}
\btxnamefont {Y.~\btxlastnamefont {Liang}}, \btxnamefont {F.~\btxlastnamefont
  {Ding}}, \btxnamefont {G.~\btxlastnamefont {Huang}}\btxandcomma {}
  \btxandshort {.}\ \btxnamefont {Z.~\btxlastnamefont {Zhao}}\btxauthorcolon\
  \btxjtitlefont {\btxifchangecase {Deep trip generation with graph neural
  networks for bike sharing system expansion}{Deep trip generation with graph
  neural networks for bike sharing system expansion}}.
\newblock \btxjournalfont {Transportation Research Part C: Emerging
  Technologies}, 154, 2023.
\newblock {\latintext
  \btxurlfont{https://www.scopus.com/inward/record.uri?eid=2-s2.0-85166619924&doi=10.1016

\bibitem {GeoMAN}
\btxnamefont {Y.~\btxlastnamefont {Liang}}, \btxnamefont {S.~\btxlastnamefont
  {Ke}}, \btxnamefont {J.~\btxlastnamefont {Zhang}}, \btxnamefont
  {X.~\btxlastnamefont {Yi}}\btxandcomma {} \btxandshort {.}\ \btxnamefont
  {Y.~\btxlastnamefont {Zheng}}\btxauthorcolon\ \btxtitlefont {\btxifchangecase
  {Geoman: multi-level attention networks for geo-sensory time series
  prediction}{GeoMAN: multi-level attention networks for geo-sensory time
  series prediction}}.
\newblock \Btxinshort {.}\ \btxtitlefont {Proceedings of the 27th International
  Joint Conference on Artificial Intelligence}, IJCAI'18, \btxpageshort {.}\
  3428–3434. \btxpublisherfont {AAAI Press}, 2018\ifbtxprintISBN {,
  \mbox{\btxISBN~\btxISBNfont {9780999241127}}}.

\bibitem {TSFM_review}
\btxnamefont {Y.~\btxlastnamefont {Liang}}, \btxnamefont {H.~\btxlastnamefont
  {Wen}}, \btxnamefont {Y.~\btxlastnamefont {Nie}}, \btxnamefont
  {Y.~\btxlastnamefont {Jiang}}, \btxnamefont {M.~\btxlastnamefont {Jin}},
  \btxnamefont {D.~\btxlastnamefont {Song}}, \btxnamefont {S.~\btxlastnamefont
  {Pan}}\btxandcomma {} \btxandshort {.}\ \btxnamefont {Q.~\btxlastnamefont
  {Wen}}\btxauthorcolon\ \btxtitlefont {\btxifchangecase {Foundation models for
  time series analysis: A tutorial and survey}{Foundation Models for Time
  Series Analysis: A Tutorial and Survey}}.
\newblock \Btxinshort {.}\ \btxtitlefont {Proceedings of the 30th ACM SIGKDD
  Conference on Knowledge Discovery and Data Mining}, KDD '24, \btxpageshort
  {.}\ 6555–6565, New York, NY, USA, 2024. \btxpublisherfont {Association for
  Computing Machinery}\ifbtxprintISBN {, \mbox{\btxISBN~\btxISBNfont
  {9798400704901}}}.
\newblock {\latintext \btxurlfont{https://doi.org/10.1145/3637528.3671451}}.

\bibitem {art_354}
\btxnamefont {H.~\btxlastnamefont {Liao}}, \btxnamefont {M.~\btxlastnamefont
  {Wu}}, \btxnamefont {L.~\btxlastnamefont {Yuan}}, \btxnamefont
  {Y.~\btxlastnamefont {Hu}}\btxandcomma {} \btxandshort {.}\ \btxnamefont
  {H.~\btxlastnamefont {Gong}}\btxauthorcolon\ \btxjtitlefont {\btxifchangecase
  {{PM2.5} prediction based on dynamic spatiotemporal graph neural
  network}{{PM2.5} prediction based on dynamic spatiotemporal graph neural
  network}}.
\newblock \btxjournalfont {Appl. Intell.},
  \btxprintmonthyear{.}{9}{2024}{short}.

\bibitem {art_479}
\btxnamefont {J.~\btxlastnamefont {Liao}}, \btxnamefont {J.~\btxlastnamefont
  {Li}}, \btxnamefont {Y.~\btxlastnamefont {Chen}}, \btxnamefont
  {R.~\btxlastnamefont {Gu}}, \btxnamefont {Y.~\btxlastnamefont
  {Zhu}}\btxandcomma {} \btxandshort {.}\ \btxnamefont {W.~\btxlastnamefont
  {Peng}}\btxauthorcolon\ \btxjtitlefont {\btxifchangecase {{DPDGAD}: A
  {Dual-Process} dynamic graph-based anomaly detection for multivariate time
  series analysis in cyber-physical systems}{{DPDGAD}: A {Dual-Process} Dynamic
  Graph-based Anomaly Detection for multivariate time series analysis in
  cyber-physical systems}}.
\newblock \btxjournalfont {Adv. Eng. Inform.}, 61(102547):102547,
  \btxprintmonthyear{.}{8}{2024}{short}.

\bibitem {art_445}
\btxnamefont {K.~\btxlastnamefont {Liao}}, \btxnamefont {W.~\btxlastnamefont
  {Zhou}}\btxandcomma {} \btxandshort {.}\ \btxnamefont {W.~\btxlastnamefont
  {Wu}}\btxauthorcolon\ \btxjtitlefont {\btxifchangecase {A self-attention
  dynamic graph convolution network model for traffic flow prediction}{A
  self-attention dynamic graph convolution network model for traffic flow
  prediction}}.
\newblock \btxjournalfont {Int. J. Mach. Learn. Cybern.},
  \btxprintmonthyear{.}{5}{2024}{short}.

\bibitem {art_119}
\btxnamefont {W.~\btxlastnamefont {Liao}}, \btxnamefont {B.~\btxlastnamefont
  {Zeng}}, \btxnamefont {J.~\btxlastnamefont {Liu}}, \btxnamefont
  {P.~\btxlastnamefont {Wei}}\btxandcomma {} \btxandshort {.}\ \btxnamefont
  {X.~\btxlastnamefont {Cheng}}\btxauthorcolon\ \btxjtitlefont
  {\btxifchangecase {Taxi demand forecasting based on the temporal multimodal
  information fusion graph neural network}{Taxi demand forecasting based on the
  temporal multimodal information fusion graph neural network}}.
\newblock \btxjournalfont {Applied Intelligence}, 52(10):12077 – 12090, 2022.
\newblock {\latintext
  \btxurlfont{https://www.scopus.com/inward/record.uri?eid=2-s2.0-85123916554&doi=10.1007

\bibitem {art_743}
\btxnamefont {C.\btxfnamespaceshort Y. \btxlastnamefont {Lin}}, \btxnamefont
  {H.\btxfnamespaceshort T. \btxlastnamefont {Su}}, \btxnamefont
  {S.\btxfnamespaceshort L. \btxlastnamefont {Tung}}\btxandcomma {}
  \btxandshort {.}\ \btxnamefont {W.\btxfnamespaceshort H. \btxlastnamefont
  {Hsu}}\btxauthorcolon\ \btxtitlefont {\btxifchangecase {Multivariate and
  propagation graph attention network for spatial-temporal prediction with
  outdoor cellular traffic}{Multivariate and propagation graph attention
  network for spatial-temporal prediction with outdoor cellular traffic}}.
\newblock \Btxinshort {.}\ \btxtitlefont {Proceedings of the 30th {ACM}
  International Conference on Information \& Knowledge Management}, New York,
  NY, USA, \btxprintmonthyear{.}{10}{2021}{short}. \btxpublisherfont {ACM}.

\bibitem {cit_rev5}
\btxnamefont {H.~\btxlastnamefont {Lin}}, \btxnamefont {M.~\btxlastnamefont
  {Yan}}, \btxnamefont {X.~\btxlastnamefont {Ye}}, \btxnamefont
  {D.~\btxlastnamefont {Fan}}, \btxnamefont {S.~\btxlastnamefont {Pan}},
  \btxnamefont {W.~\btxlastnamefont {Chen}}\btxandcomma {} \btxandshort {.}\
  \btxnamefont {Y.~\btxlastnamefont {Xie}}\btxauthorcolon\ \btxjtitlefont
  {\btxifchangecase {A comprehensive survey on distributed training of graph
  neural networks}{A Comprehensive Survey on Distributed Training of Graph
  Neural Networks}}.
\newblock \btxjournalfont {Proceedings of the IEEE}, 111(12):1572--1606, 2023.

\bibitem {art_457}
\btxnamefont {J.~\btxlastnamefont {Lin}}, \btxnamefont {T.~\btxlastnamefont
  {Lan}}, \btxnamefont {B.~\btxlastnamefont {Zhang}}, \btxnamefont
  {K.~\btxlastnamefont {Lin}}, \btxnamefont {D.~\btxlastnamefont {Miao}},
  \btxnamefont {H.~\btxlastnamefont {He}}, \btxnamefont {J.~\btxlastnamefont
  {Ye}}, \btxnamefont {C.~\btxlastnamefont {Zhang}}\btxandcomma {} \btxandshort
  {.}\ \btxnamefont {Y.\btxfnamespaceshort F. \btxlastnamefont
  {Li}}\btxauthorcolon\ \btxjtitlefont {\btxifchangecase {Multi-scenario
  cellular {KPI} prediction based on spatiotemporal graph neural
  network}{Multi-scenario cellular {KPI} prediction based on spatiotemporal
  graph neural network}}.
\newblock \btxjournalfont {IEEE Trans. Autom. Sci. Eng.}, \btxpagesshort {.}\
  1--12, 2024.

\bibitem {art_750}
\btxnamefont {L.~\btxlastnamefont {Lin}}, \btxnamefont {K.~\btxlastnamefont
  {Xia}}, \btxnamefont {A.~\btxlastnamefont {Zheng}}, \btxnamefont
  {S.~\btxlastnamefont {Hu}}\btxandcomma {} \btxandshort {.}\ \btxnamefont
  {S.~\btxlastnamefont {Wang}}\btxauthorcolon\ \btxtitlefont {\btxifchangecase
  {Hierarchical spatio-temporal graph learning based on metapath aggregation
  for emergency supply forecasting}{Hierarchical spatio-temporal graph learning
  based on metapath aggregation for emergency supply forecasting}}.
\newblock \Btxinshort {.}\ \btxtitlefont {Proceedings of the 33rd {ACM}
  International Conference on Information and Knowledge Management},
  \btxpagesshort {.}\ 1410--1419, New York, NY, USA,
  \btxprintmonthyear{.}{10}{2024}{short}. \btxpublisherfont {ACM}.

\bibitem {art_635}
\btxnamefont {W.~\btxlastnamefont {Lin}} \btxandshort {.}\ \btxnamefont
  {D.~\btxlastnamefont {Wu}}\btxauthorcolon\ \btxtitlefont {\btxifchangecase
  {Residential electric load forecasting via attentive transfer of graph neural
  networks}{Residential electric load forecasting via attentive transfer of
  graph neural networks}}.
\newblock \Btxinshort {.}\ \btxtitlefont {Proceedings of the Thirtieth
  International Joint Conference on Artificial Intelligence}, California,
  \btxprintmonthyear{.}{8}{2021}{short}. \btxpublisherfont {International Joint
  Conferences on Artificial Intelligence Organization}.

\bibitem {art_38}
\btxnamefont {H.~\btxlastnamefont {Lira}}, \btxnamefont {L.~\btxlastnamefont
  {Martí}}\btxandcomma {} \btxandshort {.}\ \btxnamefont {N.~\btxlastnamefont
  {Sanchez-Pi}}\btxauthorcolon\ \btxjtitlefont {\btxifchangecase {A graph
  neural network with spatio-temporal attention for multi-sources time series
  data: An application to frost forecast}{A Graph Neural Network with
  Spatio-Temporal Attention for Multi-Sources Time Series Data: An Application
  to Frost Forecast}}.
\newblock \btxjournalfont {Sensors}, 22(4), 2022.
\newblock {\latintext
  \btxurlfont{https://www.scopus.com/inward/record.uri?eid=2-s2.0-85124480934&doi=10.3390

\bibitem {GA+RBM+LSTM}
\btxnamefont {A.~\btxlastnamefont {{Listou Ellefsen}}}, \btxnamefont
  {E.~\btxlastnamefont {Bjørlykhaug}}, \btxnamefont {V.~\btxlastnamefont
  {Æsøy}}, \btxnamefont {S.~\btxlastnamefont {Ushakov}}\btxandcomma {}
  \btxandshort {.}\ \btxnamefont {H.~\btxlastnamefont {Zhang}}\btxauthorcolon\
  \btxjtitlefont {\btxifchangecase {Remaining useful life predictions for
  turbofan engine degradation using semi-supervised deep
  architecture}{Remaining useful life predictions for turbofan engine
  degradation using semi-supervised deep architecture}}.
\newblock \btxjournalfont {Reliability Engineering \& System Safety},
  183:240--251, 2019\ifbtxprintISSN {, \mbox{\btxISSN~\btxISSNfont
  {0951-8320}}}.
\newblock {\latintext
  \btxurlfont{https://www.sciencedirect.com/science/article/pii/S0951832018307506}}.

\bibitem {art_682}
\btxnamefont {D.~\btxlastnamefont {Liu}}, \btxnamefont {J.~\btxlastnamefont
  {Wang}}, \btxnamefont {S.~\btxlastnamefont {Shang}}\btxandcomma {}
  \btxandshort {.}\ \btxnamefont {P.~\btxlastnamefont {Han}}\btxauthorcolon\
  \btxtitlefont {\btxifchangecase {{MSDR}: Multi-step dependency relation
  networks for spatial temporal forecasting}{{MSDR}: Multi-step dependency
  relation networks for spatial temporal forecasting}}.
\newblock \Btxinshort {.}\ \btxtitlefont {Proceedings of the 28th {ACM}
  {SIGKDD} Conference on Knowledge Discovery and Data Mining}, \btxpagesshort
  {.}\ 1042--1050, New York, NY, USA, \btxprintmonthyear{.}{8}{2022}{short}.
  \btxpublisherfont {ACM}.

\bibitem {STAEformer}
\btxnamefont {H.~\btxlastnamefont {Liu}}, \btxnamefont {Z.~\btxlastnamefont
  {Dong}}, \btxnamefont {R.~\btxlastnamefont {Jiang}}, \btxnamefont
  {J.~\btxlastnamefont {Deng}}, \btxnamefont {J.~\btxlastnamefont {Deng}},
  \btxnamefont {Q.~\btxlastnamefont {Chen}}\btxandcomma {} \btxandshort {.}\
  \btxnamefont {X.~\btxlastnamefont {Song}}\btxauthorcolon\ \btxtitlefont
  {\btxifchangecase {Spatio-temporal adaptive embedding makes vanilla
  transformer sota for traffic forecasting}{Spatio-temporal adaptive embedding
  makes vanilla transformer sota for traffic forecasting}}.
\newblock \Btxinshort {.}\ \btxtitlefont {Proceedings of the 32nd ACM
  International Conference on Information and Knowledge Management},
  \btxpagesshort {.}\ 4125--4129, 2023.

\bibitem {art_389}
\btxnamefont {H.~\btxlastnamefont {Liu}}, \btxnamefont {D.~\btxlastnamefont
  {Yang}}, \btxnamefont {X.~\btxlastnamefont {Liu}}, \btxnamefont
  {X.~\btxlastnamefont {Chen}}, \btxnamefont {Z.~\btxlastnamefont {Liang}},
  \btxnamefont {H.~\btxlastnamefont {Wang}}, \btxnamefont {Y.~\btxlastnamefont
  {Cui}}\btxandcomma {} \btxandshort {.}\ \btxnamefont {J.~\btxlastnamefont
  {Gu}}\btxauthorcolon\ \btxjtitlefont {\btxifchangecase {{TodyNet}: Temporal
  dynamic graph neural network for multivariate time series
  classification}{{TodyNet}: Temporal dynamic graph neural network for
  multivariate time series classification}}.
\newblock \btxjournalfont {Inf. Sci. (Ny)}, 677(120914):120914,
  \btxprintmonthyear{.}{8}{2024}{short}.

\bibitem {art_106}
\btxnamefont {J.~\btxlastnamefont {Liu}}, \btxnamefont {G.~\btxlastnamefont
  {Guo}}\btxandcomma {} \btxandshort {.}\ \btxnamefont {X.~\btxlastnamefont
  {Jiang}}\btxauthorcolon\ \btxjtitlefont {\btxifchangecase {Bayesian
  combination approach to traffic forecasting with graph attention network and
  arima model}{Bayesian Combination Approach to Traffic Forecasting With Graph
  Attention Network and ARIMA Model}}.
\newblock \btxjournalfont {IEEE Access}, 11:94732--94741, 2023.

\bibitem {art_103}
\btxnamefont {J.~\btxlastnamefont {Liu}}, \btxnamefont {X.~\btxlastnamefont
  {Wang}}, \btxnamefont {H.~\btxlastnamefont {Lin}}\btxandcomma {} \btxandshort
  {.}\ \btxnamefont {F.~\btxlastnamefont {Yu}}\btxauthorcolon\ \btxjtitlefont
  {\btxifchangecase {Gsaa: A novel graph spatiotemporal attention algorithm for
  smart city traffic prediction}{GSAA: A Novel Graph Spatiotemporal Attention
  Algorithm for Smart City Traffic Prediction}}.
\newblock \btxjournalfont {ACM Trans. Sen. Netw.},
  \btxprintmonthyear{.}{nov}{2023}{short}\ifbtxprintISSN {,
  \mbox{\btxISSN~\btxISSNfont {1550-4859}}}.
\newblock {\latintext \btxurlfont{https://doi.org/10.1145/3631608}}, Just
  Accepted.

\bibitem {SCINet}
\btxnamefont {M.~\btxlastnamefont {Liu}}, \btxnamefont {A.~\btxlastnamefont
  {Zeng}}, \btxnamefont {M.~\btxlastnamefont {Chen}}, \btxnamefont
  {Z.~\btxlastnamefont {Xu}}, \btxnamefont {Q.~\btxlastnamefont {LAI}},
  \btxnamefont {L.~\btxlastnamefont {Ma}}\btxandcomma {} \btxandshort {.}\
  \btxnamefont {Q.~\btxlastnamefont {Xu}}\btxauthorcolon\ \btxtitlefont
  {\btxifchangecase {Scinet: Time series modeling and forecasting with sample
  convolution and interaction}{SCINet: Time Series Modeling and Forecasting
  with Sample Convolution and Interaction}}.
\newblock \Btxinshort {.}\ \btxnamefont {S.~\btxlastnamefont {Koyejo}},
  \btxnamefont {S.~\btxlastnamefont {Mohamed}}, \btxnamefont
  {A.~\btxlastnamefont {Agarwal}}, \btxnamefont {D.~\btxlastnamefont
  {Belgrave}}, \btxnamefont {K.~\btxlastnamefont {Cho}}\btxandcomma {}
  \btxandshort {.}\ \btxnamefont {A.~\btxlastnamefont {Oh}}\ (\btxeditorsshort
  {.}): \btxtitlefont {Advances in Neural Information Processing Systems},
  \btxvolumeshort {.}~\btxvolumefont {35}, \btxpagesshort {.}\ 5816--5828.
  \btxpublisherfont {Curran Associates, Inc.}, 2022.
\newblock {\latintext
  \btxurlfont{https://proceedings.neurips.cc/paper_files/paper/2022/file/266983d0949aed78a16fa4782237dea7-Paper-Conference.pdf}}.

\bibitem {art_340}
\btxnamefont {Q.~\btxlastnamefont {Liu}}, \btxnamefont {X.~\btxlastnamefont
  {Cheng}}, \btxnamefont {J.~\btxlastnamefont {Shi}}, \btxnamefont
  {Y.~\btxlastnamefont {Ma}}\btxandcomma {} \btxandshort {.}\ \btxnamefont
  {P.~\btxlastnamefont {Peng}}\btxauthorcolon\ \btxjtitlefont {\btxifchangecase
  {Modeling and predicting energy consumption of chiller based on dynamic
  spatial-temporal graph neural network}{Modeling and predicting energy
  consumption of chiller based on dynamic spatial-temporal graph neural
  network}}.
\newblock \btxjournalfont {J. Build. Eng.}, 91(109657):109657,
  \btxprintmonthyear{.}{8}{2024}{short}.

\bibitem {art_143}
\btxnamefont {W.~\btxlastnamefont {Liu}} \btxandshort {.}\ \btxnamefont
  {M.\btxfnamespaceshort J. \btxlastnamefont {Pyrcz}}\btxauthorcolon\
  \btxjtitlefont {\btxifchangecase {Physics-informed graph neural network for
  spatial-temporal production forecasting}{Physics-informed graph neural
  network for spatial-temporal production forecasting}}.
\newblock \btxjournalfont {Geoenergy Science and Engineering}, 223, 2023.
\newblock {\latintext
  \btxurlfont{https://www.scopus.com/inward/record.uri?eid=2-s2.0-85159768181&doi=10.1016

\bibitem {art_2}
\btxnamefont {X.~\btxlastnamefont {Liu}} \btxandshort {.}\ \btxnamefont
  {W.~\btxlastnamefont {Li}}\btxauthorcolon\ \btxjtitlefont {\btxifchangecase
  {Mgc-lstm: a deep learning model based on graph convolution of multiple
  graphs for pm2.5 prediction}{MGC-LSTM: a deep learning model based on graph
  convolution of multiple graphs for PM2.5 prediction}}.
\newblock \btxjournalfont {International Journal of Environmental Science and
  Technology}, 20(9):10297 – 10312, 2023.
\newblock {\latintext
  \btxurlfont{https://www.scopus.com/inward/record.uri?eid=2-s2.0-85139197916&doi=10.1007

\bibitem {art_702}
\btxnamefont {X.~\btxlastnamefont {Liu}}, \btxnamefont {X.~\btxlastnamefont
  {Lyu}}, \btxnamefont {X.~\btxlastnamefont {Zhang}}, \btxnamefont
  {J.~\btxlastnamefont {Gao}}\btxandcomma {} \btxandshort {.}\ \btxnamefont
  {J.~\btxlastnamefont {Chen}}\btxauthorcolon\ \btxtitlefont {\btxifchangecase
  {Memory augmented graph learning networks for multivariate time series
  forecasting}{Memory augmented graph learning networks for multivariate time
  series forecasting}}.
\newblock \Btxinshort {.}\ \btxtitlefont {Proceedings of the 31st {ACM}
  International Conference on Information \& Knowledge Management}, New York,
  NY, USA, \btxprintmonthyear{.}{10}{2022}{short}. \btxpublisherfont {ACM}.

\bibitem {art_366}
\btxnamefont {X.~\btxlastnamefont {Liu}}, \btxnamefont {J.~\btxlastnamefont
  {Yu}}, \btxnamefont {L.~\btxlastnamefont {Gong}}, \btxnamefont
  {M.~\btxlastnamefont {Liu}}\btxandcomma {} \btxandshort {.}\ \btxnamefont
  {X.~\btxlastnamefont {Xiang}}\btxauthorcolon\ \btxjtitlefont
  {\btxifchangecase {A {GCN-based} adaptive generative adversarial network
  model for short-term wind speed scenario prediction}{A {GCN-based} adaptive
  generative adversarial network model for short-term wind speed scenario
  prediction}}.
\newblock \btxjournalfont {Energy (Oxf.)}, 294(130931):130931,
  \btxprintmonthyear{.}{5}{2024}{short}.

\bibitem {art_410}
\btxnamefont {X.\btxfnamespaceshort D. \btxlastnamefont {Liu}}, \btxnamefont
  {B.\btxfnamespaceshort H. \btxlastnamefont {Hou}}, \btxnamefont
  {Z.\btxfnamespaceshort J. \btxlastnamefont {Xie}}, \btxnamefont
  {N.~\btxlastnamefont {Feng}}\btxandcomma {} \btxandshort {.}\ \btxnamefont
  {X.\btxfnamespaceshort P. \btxlastnamefont {Dong}}\btxauthorcolon\
  \btxjtitlefont {\btxifchangecase {Integrating gated recurrent unit in graph
  neural network to improve infectious disease prediction: an
  attempt}{Integrating gated recurrent unit in graph neural network to improve
  infectious disease prediction: an attempt}}.
\newblock \btxjournalfont {Front. Public Health}, 12:1397260,
  \btxprintmonthyear{.}{5}{2024}{short}.

\bibitem {art_685}
\btxnamefont {Y.~\btxlastnamefont {Liu}}, \btxnamefont {Q.~\btxlastnamefont
  {Liu}}, \btxnamefont {J.\btxfnamespaceshort W. \btxlastnamefont {Zhang}},
  \btxnamefont {H.~\btxlastnamefont {Feng}}, \btxnamefont {Z.~\btxlastnamefont
  {Wang}}, \btxnamefont {Z.~\btxlastnamefont {Zhou}}\btxandcomma {}
  \btxandshort {.}\ \btxnamefont {W.~\btxlastnamefont {Chen}}\btxauthorcolon\
  \btxtitlefont {\btxifchangecase {Multivariate time-series forecasting with
  temporal polynomial graph neural networks}{Multivariate time-series
  forecasting with temporal polynomial graph neural networks}}.
\newblock \Btxinshort {.}\ \btxtitlefont {Proceedings of the 36th International
  Conference on Neural Information Processing Systems}, NIPS '22, Red Hook, NY,
  USA, 2022. \btxpublisherfont {Curran Associates Inc.}\ifbtxprintISBN {,
  \mbox{\btxISBN~\btxISBNfont {9781713871088}}}.

\bibitem {art_704}
\btxnamefont {Z.~\btxlastnamefont {Liu}}, \btxnamefont {X.~\btxlastnamefont
  {Huang}}, \btxnamefont {J.~\btxlastnamefont {Zhang}}, \btxnamefont
  {Z.~\btxlastnamefont {Hao}}, \btxnamefont {L.~\btxlastnamefont
  {Sun}}\btxandcomma {} \btxandshort {.}\ \btxnamefont {H.~\btxlastnamefont
  {Peng}}\btxauthorcolon\ \btxtitlefont {\btxifchangecase {Multivariate
  time-series anomaly detection based on enhancing graph attention networks
  with topological analysis}{Multivariate Time-Series Anomaly Detection based
  on Enhancing Graph Attention Networks with Topological Analysis}}.
\newblock \Btxinshort {.}\ \btxtitlefont {Proceedings of the 33rd ACM
  International Conference on Information and Knowledge Management}, CIKM '24,
  \btxpageshort {.}\ 1555–1564, New York, NY, USA, 2024. \btxpublisherfont
  {Association for Computing Machinery}\ifbtxprintISBN {,
  \mbox{\btxISBN~\btxISBNfont {9798400704369}}}.
\newblock {\latintext \btxurlfont{https://doi.org/10.1145/3627673.3679614}}.

\bibitem {cit_rev2}
\btxnamefont {A.~\btxlastnamefont {Longa}}, \btxnamefont {V.~\btxlastnamefont
  {Lachi}}, \btxnamefont {G.~\btxlastnamefont {Santin}}, \btxnamefont
  {M.~\btxlastnamefont {Bianchini}}, \btxnamefont {B.~\btxlastnamefont
  {Lepri}}, \btxnamefont {P.~\btxlastnamefont {Lio}}, \btxnamefont {franco
  \btxlastnamefont {scarselli}}\btxandcomma {} \btxandshort {.}\ \btxnamefont
  {A.~\btxlastnamefont {Passerini}}\btxauthorcolon\ \btxjtitlefont
  {\btxifchangecase {Graph neural networks for temporal graphs: State of the
  art, open challenges, and opportunities}{Graph Neural Networks for Temporal
  Graphs: State of the Art, Open Challenges, and Opportunities}}.
\newblock \btxjournalfont {Transactions on Machine Learning Research},
  2023\ifbtxprintISSN {, \mbox{\btxISSN~\btxISSNfont {2835-8856}}}.
\newblock {\latintext \btxurlfont{https://openreview.net/forum?id=pHCdMat0gI}}.

\bibitem {art_93}
\btxnamefont {B.~\btxlastnamefont {Lu}}, \btxnamefont {X.~\btxlastnamefont
  {Gan}}, \btxnamefont {H.~\btxlastnamefont {Jin}}, \btxnamefont
  {L.~\btxlastnamefont {Fu}}, \btxnamefont {X.~\btxlastnamefont
  {Wang}}\btxandcomma {} \btxandshort {.}\ \btxnamefont {H.~\btxlastnamefont
  {Zhang}}\btxauthorcolon\ \btxjtitlefont {\btxifchangecase {Make more
  connections: Urban traffic flow forecasting with spatiotemporal adaptive
  gated graph convolution network}{Make More Connections: Urban Traffic Flow
  Forecasting with Spatiotemporal Adaptive Gated Graph Convolution Network}}.
\newblock \btxjournalfont {ACM Trans. Intell. Syst. Technol.}, 13(2),
  \btxprintmonthyear{.}{jan}{2022}{short}\ifbtxprintISSN {,
  \mbox{\btxISSN~\btxISSNfont {2157-6904}}}.
\newblock {\latintext \btxurlfont{https://doi.org/10.1145/3488902}}.

\bibitem {art_736}
\btxnamefont {B.~\btxlastnamefont {Lu}}, \btxnamefont {X.~\btxlastnamefont
  {Gan}}, \btxnamefont {H.~\btxlastnamefont {Jin}}, \btxnamefont
  {L.~\btxlastnamefont {Fu}}\btxandcomma {} \btxandshort {.}\ \btxnamefont
  {H.~\btxlastnamefont {Zhang}}\btxauthorcolon\ \btxtitlefont {\btxifchangecase
  {Spatiotemporal adaptive gated graph convolution network for urban traffic
  flow forecasting}{Spatiotemporal adaptive gated graph convolution network for
  urban traffic flow forecasting}}.
\newblock \Btxinshort {.}\ \btxtitlefont {Proceedings of the 29th {ACM}
  International Conference on Information \& Knowledge Management}, New York,
  NY, USA, \btxprintmonthyear{.}{10}{2020}{short}. \btxpublisherfont {ACM}.

\bibitem {art_667}
\btxnamefont {B.~\btxlastnamefont {Lu}}, \btxnamefont {X.~\btxlastnamefont
  {Gan}}, \btxnamefont {W.~\btxlastnamefont {Zhang}}, \btxnamefont
  {H.~\btxlastnamefont {Yao}}, \btxnamefont {L.~\btxlastnamefont
  {Fu}}\btxandcomma {} \btxandshort {.}\ \btxnamefont {X.~\btxlastnamefont
  {Wang}}\btxauthorcolon\ \btxtitlefont {\btxifchangecase {Spatio-temporal
  graph few-shot learning with {Cross-City} knowledge transfer}{Spatio-temporal
  graph few-shot learning with {Cross-City} knowledge transfer}}.
\newblock \Btxinshort {.}\ \btxtitlefont {Proceedings of the 28th {ACM}
  {SIGKDD} Conference on Knowledge Discovery and Data Mining}, \btxpagesshort
  {.}\ 1162--1172, New York, NY, USA, \btxprintmonthyear{.}{8}{2022}{short}.
  \btxpublisherfont {ACM}.

\bibitem {art_654}
\btxnamefont {Y.\btxfnamespaceshort J. \btxlastnamefont {Lu}} \btxandshort {.}\
  \btxnamefont {C.\btxfnamespaceshort T. \btxlastnamefont {Li}}\btxauthorcolon\
  \btxtitlefont {\btxifchangecase {{AGSTN}: Learning attention-adjusted graph
  spatio-temporal networks for short-term urban sensor value
  forecasting}{{AGSTN}: Learning attention-adjusted graph spatio-temporal
  networks for short-term urban sensor value forecasting}}.
\newblock \Btxinshort {.}\ \btxtitlefont {2020 {IEEE} International Conference
  on Data Mining ({ICDM})}. \btxpublisherfont {IEEE},
  \btxprintmonthyear{.}{11}{2020}{short}.

\bibitem {art_124}
\btxnamefont {Z.~\btxlastnamefont {Lu}}, \btxnamefont {W.~\btxlastnamefont
  {Lv}}, \btxnamefont {Z.~\btxlastnamefont {Xie}}, \btxnamefont
  {B.~\btxlastnamefont {Du}}, \btxnamefont {G.~\btxlastnamefont {Xiong}},
  \btxnamefont {L.~\btxlastnamefont {Sun}}\btxandcomma {} \btxandshort {.}\
  \btxnamefont {H.~\btxlastnamefont {Wang}}\btxauthorcolon\ \btxjtitlefont
  {\btxifchangecase {Graph sequence neural network with an attention mechanism
  for traffic speed prediction}{Graph Sequence Neural Network with an Attention
  Mechanism for Traffic Speed Prediction}}.
\newblock \btxjournalfont {ACM Transactions on Intelligent Systems and
  Technology}, 13(2), 2022.
\newblock {\latintext
  \btxurlfont{https://www.scopus.com/inward/record.uri?eid=2-s2.0-85129490445&doi=10.1145

\bibitem {art_650}
\btxnamefont {M.~\btxlastnamefont {Ma}}, \btxnamefont {J.~\btxlastnamefont
  {Hu}}, \btxnamefont {C.\btxfnamespaceshort S. \btxlastnamefont {Jensen}},
  \btxnamefont {F.~\btxlastnamefont {Teng}}, \btxnamefont {P.~\btxlastnamefont
  {Han}}, \btxnamefont {Z.~\btxlastnamefont {Xu}}\btxandcomma {} \btxandshort
  {.}\ \btxnamefont {T.~\btxlastnamefont {Li}}\btxauthorcolon\ \btxtitlefont
  {\btxifchangecase {Learning time-aware graph structures for spatially
  correlated time series forecasting}{Learning time-aware graph structures for
  spatially correlated time series forecasting}}.
\newblock \Btxinshort {.}\ \btxtitlefont {2024 {IEEE} 40th International
  Conference on Data Engineering ({ICDE})}. \btxpublisherfont {IEEE},
  \btxprintmonthyear{.}{5}{2024}{short}.

\bibitem {GNN_molecule}
\btxnamefont {M.~\btxlastnamefont {Ma}} \btxandshort {.}\ \btxnamefont
  {X.~\btxlastnamefont {Lei}}\btxauthorcolon\ \btxjtitlefont {\btxifchangecase
  {A dual graph neural network for drug–drug interactions prediction based on
  molecular structure and interactions}{A dual graph neural network for
  drug–drug interactions prediction based on molecular structure and
  interactions}}.
\newblock \btxjournalfont {PLOS Computational Biology}, 19(1):e1010812,
  \btxprintmonthyear{.}{1}{2023}{short}\ifbtxprintISSN {,
  \mbox{\btxISSN~\btxISSNfont {1553-7358}}}.
\newblock {\latintext
  \btxurlfont{http://dx.doi.org/10.1371/journal.pcbi.1010812}}.

\bibitem {art_733}
\btxnamefont {Q.~\btxlastnamefont {Ma}}, \btxnamefont {Z.~\btxlastnamefont
  {Zhang}}, \btxnamefont {X.~\btxlastnamefont {Zhao}}, \btxnamefont
  {H.~\btxlastnamefont {Li}}, \btxnamefont {H.~\btxlastnamefont {Zhao}},
  \btxnamefont {Y.~\btxlastnamefont {Wang}}, \btxnamefont {Z.~\btxlastnamefont
  {Liu}}\btxandcomma {} \btxandshort {.}\ \btxnamefont {W.~\btxlastnamefont
  {Wang}}\btxauthorcolon\ \btxtitlefont {\btxifchangecase {Rethinking sensors
  modeling: Hierarchical information enhanced traffic forecasting}{Rethinking
  sensors modeling: Hierarchical information enhanced traffic forecasting}}.
\newblock \Btxinshort {.}\ \btxtitlefont {Proceedings of the 32nd {ACM}
  International Conference on Information and Knowledge Management}, New York,
  NY, USA, \btxprintmonthyear{.}{10}{2023}{short}. \btxpublisherfont {ACM}.

\bibitem {art_713}
\btxnamefont {Y.~\btxlastnamefont {Ma}}, \btxnamefont {P.~\btxlastnamefont
  {Gerard}}, \btxnamefont {Y.~\btxlastnamefont {Tian}}, \btxnamefont
  {Z.~\btxlastnamefont {Guo}}\btxandcomma {} \btxandshort {.}\ \btxnamefont
  {N.\btxfnamespaceshort V. \btxlastnamefont {Chawla}}\btxauthorcolon\
  \btxtitlefont {\btxifchangecase {Hierarchical spatio-temporal graph neural
  networks for pandemic forecasting}{Hierarchical spatio-temporal graph neural
  networks for pandemic forecasting}}.
\newblock \Btxinshort {.}\ \btxtitlefont {Proceedings of the 31st {ACM}
  International Conference on Information \& Knowledge Management}, New York,
  NY, USA, \btxprintmonthyear{.}{10}{2022}{short}. \btxpublisherfont {ACM}.

\bibitem {art_742}
\btxnamefont {A.\btxfnamespaceshort V. \btxlastnamefont {Malarkkan}},
  \btxnamefont {D.~\btxlastnamefont {Wang}}\btxandcomma {} \btxandshort {.}\
  \btxnamefont {Y.~\btxlastnamefont {Fu}}\btxauthorcolon\ \btxtitlefont
  {\btxifchangecase {Multi-view causal graph fusion based anomaly detection in
  cyber-physical infrastructures}{Multi-view causal graph fusion based anomaly
  detection in cyber-physical infrastructures}}.
\newblock \Btxinshort {.}\ \btxtitlefont {Proceedings of the 33rd {ACM}
  International Conference on Information and Knowledge Management},
  \btxvolumeshort {.}\ \btxvolumefont {227}, \btxpagesshort {.}\ 4760--4767,
  New York, NY, USA, \btxprintmonthyear{.}{10}{2024}{short}. \btxpublisherfont
  {ACM}.

\bibitem {GNN_NLP}
\btxnamefont {M.~\btxlastnamefont {Malekzadeh}}, \btxnamefont
  {P.~\btxlastnamefont {Hajibabaee}}, \btxnamefont {M.~\btxlastnamefont
  {Heidari}}, \btxnamefont {S.~\btxlastnamefont {Zad}}, \btxnamefont
  {O.~\btxlastnamefont {Uzuner}}\btxandcomma {} \btxandshort {.}\ \btxnamefont
  {J.~\btxlastnamefont {Jones}}\btxauthorcolon\ \btxtitlefont {\btxifchangecase
  {Review of graph neural network in text classification}{Review of Graph
  Neural Network in Text Classification}}.
\newblock \btxpagesshort {.}\ 0084--0091,
  \btxprintmonthyear{.}{12}{2021}{short}.

\bibitem {art_415}
\btxnamefont {J.~\btxlastnamefont {Mao}}, \btxnamefont {Y.~\btxlastnamefont
  {Han}}, \btxnamefont {G.~\btxlastnamefont {Tanaka}}\btxandcomma {}
  \btxandshort {.}\ \btxnamefont {B.~\btxlastnamefont {Wang}}\btxauthorcolon\
  \btxjtitlefont {\btxifchangecase {Backbone-based dynamic {Spatio-Temporal}
  graph neural network for epidemic forecasting}{Backbone-based Dynamic
  {Spatio-Temporal} Graph Neural Network for epidemic forecasting}}.
\newblock \btxjournalfont {Knowl. Based Syst.}, 296(111952):111952,
  \btxprintmonthyear{.}{7}{2024}{short}.

\bibitem {art_644}
\btxnamefont {I.~\btxlastnamefont {Marisca}}, \btxnamefont {C.~\btxlastnamefont
  {Alippi}}\btxandcomma {} \btxandshort {.}\ \btxnamefont
  {F.\btxfnamespaceshort M. \btxlastnamefont {Bianchi}}\btxauthorcolon\
  \btxtitlefont {\btxifchangecase {Graph-based forecasting with missing data
  through spatiotemporal downsampling}{Graph-based forecasting with missing
  data through spatiotemporal downsampling}}.
\newblock ICML'24. \btxpublisherfont {JMLR.org}, 2024.

\bibitem {art_81}
\btxnamefont {H.~\btxlastnamefont {Miao}}, \btxnamefont {Y.~\btxlastnamefont
  {Zhang}}, \btxnamefont {Z.~\btxlastnamefont {Ning}}, \btxnamefont
  {Z.~\btxlastnamefont {Jiang}}\btxandcomma {} \btxandshort {.}\ \btxnamefont
  {L.~\btxlastnamefont {Wang}}\btxauthorcolon\ \btxjtitlefont {\btxifchangecase
  {Tdg4msf: A temporal decomposition enhanced graph neural network for
  multivariate time series forecasting}{TDG4MSF: A temporal decomposition
  enhanced graph neural network for multivariate time series forecasting}}.
\newblock \btxjournalfont {Applied Intelligence}, 53(23):28254 – 28267, 2023.
\newblock {\latintext
  \btxurlfont{https://www.scopus.com/inward/record.uri?eid=2-s2.0-85172877266&doi=10.1007

\bibitem {art_341}
\btxnamefont {A.~\btxlastnamefont {Miraki}}, \btxnamefont {P.~\btxlastnamefont
  {Parviainen}}\btxandcomma {} \btxandshort {.}\ \btxnamefont
  {R.~\btxlastnamefont {Arghandeh}}\btxauthorcolon\ \btxjtitlefont
  {\btxifchangecase {Electricity demand forecasting at distribution and
  household levels using explainable causal graph neural network}{Electricity
  demand forecasting at distribution and household levels using explainable
  causal graph neural network}}.
\newblock \btxjournalfont {Energy and AI}, 16(100368):100368,
  \btxprintmonthyear{.}{5}{2024}{short}.

\bibitem {art_159}
\btxnamefont {R.~\btxlastnamefont {Mondal}}, \btxnamefont {D.~\btxlastnamefont
  {Mukherjee}}, \btxnamefont {P.\btxfnamespaceshort K. \btxlastnamefont
  {Singh}}, \btxnamefont {V.~\btxlastnamefont {Bhateja}}\btxandcomma {}
  \btxandshort {.}\ \btxnamefont {R.~\btxlastnamefont {Sarkar}}\btxauthorcolon\
  \btxjtitlefont {\btxifchangecase {A new framework for smartphone sensor-based
  human activity recognition using graph neural network}{A New Framework for
  Smartphone Sensor-Based Human Activity Recognition Using Graph Neural
  Network}}.
\newblock \btxjournalfont {IEEE Sensors Journal}, 21(10):11461--11468, 2021.

\bibitem {art_51}
\btxnamefont {C.~\btxlastnamefont {Murphy}}, \btxnamefont {E.~\btxlastnamefont
  {Laurence}}\btxandcomma {} \btxandshort {.}\ \btxnamefont
  {A.~\btxlastnamefont {Allard}}\btxauthorcolon\ \btxjtitlefont
  {\btxifchangecase {Deep learning of contagion dynamics on complex
  networks}{Deep learning of contagion dynamics on complex networks}}.
\newblock \btxjournalfont {Nature Communications}, 12(1), 2021.
\newblock {\latintext
  \btxurlfont{https://www.scopus.com/inward/record.uri?eid=2-s2.0-85112016386&doi=10.1038

\bibitem {art_155}
\btxnamefont {C.~\btxlastnamefont {Mylonas}} \btxandshort {.}\ \btxnamefont
  {E.~\btxlastnamefont {Chatzi}}\btxauthorcolon\ \btxjtitlefont
  {\btxifchangecase {Remaining useful life estimation for engineered systems
  operating under uncertainty with causal graphnets}{Remaining useful life
  estimation for engineered systems operating under uncertainty with causal
  graphnets}}.
\newblock \btxjournalfont {Sensors}, 21(19), 2021.
\newblock {\latintext
  \btxurlfont{https://www.scopus.com/inward/record.uri?eid=2-s2.0-85115354292&doi=10.3390

\bibitem {Nandan2025}
\btxnamefont {M.~\btxlastnamefont {Nandan}}, \btxnamefont {S.~\btxlastnamefont
  {Mitra}}\btxandcomma {} \btxandshort {.}\ \btxnamefont {D.~\btxlastnamefont
  {De}}\btxauthorcolon\ \btxjtitlefont {\btxifchangecase {Graphxai: a survey of
  graph neural networks (gnns) for explainable ai (xai)}{GraphXAI: a survey of
  graph neural networks (GNNs) for explainable AI (XAI)}}.
\newblock \btxjournalfont {Neural Computing and Applications},
  \btxprintmonthyear{.}{3}{2025}{short}\ifbtxprintISSN {,
  \mbox{\btxISSN~\btxISSNfont {1433-3058}}}.
\newblock {\latintext
  \btxurlfont{http://dx.doi.org/10.1007/s00521-025-11054-3}}.

\bibitem {art_6}
\btxnamefont {Q.~\btxlastnamefont {Ni}}, \btxnamefont {Y.~\btxlastnamefont
  {Wang}}\btxandcomma {} \btxandshort {.}\ \btxnamefont {J.~\btxlastnamefont
  {Yuan}}\btxauthorcolon\ \btxjtitlefont {\btxifchangecase {Adaptive scalable
  spatio-temporal graph convolutional network for pm2.5 prediction}{Adaptive
  scalable spatio-temporal graph convolutional network for PM2.5 prediction}}.
\newblock \btxjournalfont {Engineering Applications of Artificial
  Intelligence}, 126, 2023.
\newblock {\latintext
  \btxurlfont{https://www.scopus.com/inward/record.uri?eid=2-s2.0-85170435593&doi=10.1016

\bibitem {LogTrans}
\btxnamefont {X.~\btxlastnamefont {Nie}}, \btxnamefont {X.~\btxlastnamefont
  {Zhou}}, \btxnamefont {Z.~\btxlastnamefont {Li}}, \btxnamefont
  {L.~\btxlastnamefont {Wang}}, \btxnamefont {X.~\btxlastnamefont
  {Lin}}\btxandcomma {} \btxandshort {.}\ \btxnamefont {T.~\btxlastnamefont
  {Tong}}\btxauthorcolon\ \btxjtitlefont {\btxifchangecase {Logtrans: Providing
  efficient local-global fusion with transformer and cnn parallel network for
  biomedical image segmentation}{LogTrans: Providing Efficient Local-Global
  Fusion with Transformer and CNN Parallel Network for Biomedical Image
  Segmentation}}.
\newblock \btxjournalfont {2022 IEEE 24th Int Conf on High Performance
  Computing \& Communications; 8th Int Conf on Data Science \& Systems; 20th
  Int Conf on Smart City; 8th Int Conf on Dependability in Sensor, Cloud \& Big
  Data Systems \& Application (HPCC/DSS/SmartCity/DependSys)}, \btxpagesshort
  {.}\ 769--776, 2022.
\newblock {\latintext
  \btxurlfont{https://api.semanticscholar.org/CorpusID:257809606}}.

\bibitem {PatchTST}
\btxnamefont {Y.~\btxlastnamefont {Nie}}, \btxnamefont {N.\btxfnamespaceshort
  H. \btxlastnamefont {Nguyen}}, \btxnamefont {P.~\btxlastnamefont
  {Sinthong}}\btxandcomma {} \btxandshort {.}\ \btxnamefont
  {J.~\btxlastnamefont {Kalagnanam}}\btxauthorcolon\ \btxtitlefont
  {\btxifchangecase {A time series is worth 64 words: Long-term forecasting
  with transformers}{A Time Series is Worth 64 Words: Long-term Forecasting
  with Transformers}}.
\newblock \Btxinshort {.}\ \btxtitlefont {The Eleventh International Conference
  on Learning Representations}, 2023.
\newblock {\latintext
  \btxurlfont{https://openreview.net/forum?id=Jbdc0vTOcol}}.

\bibitem {art_57}
\btxnamefont {S.~\btxlastnamefont {Ning}}, \btxnamefont {Y.~\btxlastnamefont
  {Ren}}\btxandcomma {} \btxandshort {.}\ \btxnamefont {Y.~\btxlastnamefont
  {Wu}}\btxauthorcolon\ \btxjtitlefont {\btxifchangecase {Intelligent fault
  diagnosis of rolling bearings based on the visibility algorithm and graph
  neural networks}{Intelligent fault diagnosis of rolling bearings based on the
  visibility algorithm and graph neural networks}}.
\newblock \btxjournalfont {Journal of the Brazilian Society of Mechanical
  Sciences and Engineering}, 45(2), 2023.
\newblock {\latintext
  \btxurlfont{https://www.scopus.com/inward/record.uri?eid=2-s2.0-85145776581&doi=10.1007

\bibitem {art_375}
\btxnamefont {T.~\btxlastnamefont {Niu}}, \btxnamefont {H.~\btxlastnamefont
  {Zhang}}, \btxnamefont {X.~\btxlastnamefont {Yan}}\btxandcomma {}
  \btxandshort {.}\ \btxnamefont {Q.~\btxlastnamefont {Miao}}\btxauthorcolon\
  \btxjtitlefont {\btxifchangecase {Intricate supply chain demand forecasting
  based on graph convolution network}{Intricate supply chain demand forecasting
  based on graph convolution network}}.
\newblock \btxjournalfont {Sustainability}, 16(21):9608,
  \btxprintmonthyear{.}{11}{2024}{short}.

\bibitem {art_52}
\btxnamefont {L.\btxfnamespaceshort C. \btxlastnamefont {Oliveira}},
  \btxnamefont {J.\btxfnamespaceshort T. \btxlastnamefont {Oliva}},
  \btxnamefont {M.\btxfnamespaceshort H.\btxfnamespaceshort D. \btxlastnamefont
  {Ribeiro}}, \btxnamefont {M.~\btxlastnamefont {Teixeira}}\btxandcomma {}
  \btxandshort {.}\ \btxnamefont {D.~\btxlastnamefont
  {Casanova}}\btxauthorcolon\ \btxjtitlefont {\btxifchangecase {Forecasting the
  covid-19 space-time dynamics in brazil with convolutional graph neural
  networks and transport modals}{Forecasting the COVID-19 Space-Time Dynamics
  in Brazil With Convolutional Graph Neural Networks and Transport Modals}}.
\newblock \btxjournalfont {IEEE Access}, 10:85064 – 85079, 2022.
\newblock {\latintext
  \btxurlfont{https://www.scopus.com/inward/record.uri?eid=2-s2.0-85135735451&doi=10.1109

\bibitem {N-BEATS}
\btxnamefont {B.\btxfnamespaceshort N. \btxlastnamefont {Oreshkin}},
  \btxnamefont {D.~\btxlastnamefont {Carpov}}, \btxnamefont
  {N.~\btxlastnamefont {Chapados}}\btxandcomma {} \btxandshort {.}\
  \btxnamefont {Y.~\btxlastnamefont {Bengio}}\btxauthorcolon\ \btxtitlefont
  {\btxifchangecase {{N-BEATS}: Neural basis expansion analysis for
  interpretable time series forecasting}{{N-BEATS}: Neural basis expansion
  analysis for interpretable time series forecasting}}, 2020.
\newblock {\latintext \btxurlfont{https://openreview.net/forum?id=r1ecqn4YwB}}.

\bibitem {art_710}
\btxnamefont {J.~\btxlastnamefont {Oskarsson}}, \btxnamefont
  {P.~\btxlastnamefont {Sid{\'e}n}}\btxandcomma {} \btxandshort {.}\
  \btxnamefont {F.~\btxlastnamefont {Lindsten}}\btxauthorcolon\ \btxtitlefont
  {\btxifchangecase {Temporal graph neural networks for irregular
  data}{Temporal Graph Neural Networks for Irregular Data}}.
\newblock 2023.

\bibitem {art_739}
\btxnamefont {J.~\btxlastnamefont {Ou}}, \btxnamefont {J.~\btxlastnamefont
  {Sun}}, \btxnamefont {Y.~\btxlastnamefont {Zhu}}, \btxnamefont
  {H.~\btxlastnamefont {Jin}}, \btxnamefont {Y.~\btxlastnamefont {Liu}},
  \btxnamefont {F.~\btxlastnamefont {Zhang}}, \btxnamefont {J.~\btxlastnamefont
  {Huang}}\btxandcomma {} \btxandshort {.}\ \btxnamefont {X.~\btxlastnamefont
  {Wang}}\btxauthorcolon\ \btxtitlefont {\btxifchangecase {{STP-TrellisNets}:
  Spatial-temporal parallel {TrellisNets} for metro station passenger flow
  prediction}{{STP-TrellisNets}: Spatial-temporal parallel {TrellisNets} for
  metro station passenger flow prediction}}.
\newblock \Btxinshort {.}\ \btxtitlefont {Proceedings of the 29th {ACM}
  International Conference on Information \& Knowledge Management}, New York,
  NY, USA, \btxprintmonthyear{.}{10}{2020}{short}. \btxpublisherfont {ACM}.

\bibitem {PRISMA_statement}
\btxnamefont {M.\btxfnamespaceshort J. \btxlastnamefont {Page}}, \btxnamefont
  {J.\btxfnamespaceshort E. \btxlastnamefont {McKenzie}}, \btxnamefont
  {P.\btxfnamespaceshort M. \btxlastnamefont {Bossuyt}}, \btxnamefont
  {I.~\btxlastnamefont {Boutron}}, \btxnamefont {T.\btxfnamespaceshort C.
  \btxlastnamefont {Hoffmann}}, \btxnamefont {C.\btxfnamespaceshort D.
  \btxlastnamefont {Mulrow}}, \btxnamefont {L.~\btxlastnamefont {Shamseer}},
  \btxnamefont {J.\btxfnamespaceshort M. \btxlastnamefont {Tetzlaff}},
  \btxnamefont {E.\btxfnamespaceshort A. \btxlastnamefont {Akl}}, \btxnamefont
  {S.\btxfnamespaceshort E. \btxlastnamefont {Brennan}}, \btxnamefont
  {R.~\btxlastnamefont {Chou}}, \btxnamefont {J.~\btxlastnamefont {Glanville}},
  \btxnamefont {J.\btxfnamespaceshort M. \btxlastnamefont {Grimshaw}},
  \btxnamefont {A.~\btxlastnamefont {Hr{\'o}bjartsson}}, \btxnamefont
  {M.\btxfnamespaceshort M. \btxlastnamefont {Lalu}}, \btxnamefont
  {T.~\btxlastnamefont {Li}}, \btxnamefont {E.\btxfnamespaceshort W.
  \btxlastnamefont {Loder}}, \btxnamefont {E.~\btxlastnamefont {Mayo-Wilson}},
  \btxnamefont {S.~\btxlastnamefont {McDonald}}, \btxnamefont
  {L.\btxfnamespaceshort A. \btxlastnamefont {McGuinness}}, \btxnamefont
  {L.\btxfnamespaceshort A. \btxlastnamefont {Stewart}}, \btxnamefont
  {J.~\btxlastnamefont {Thomas}}, \btxnamefont {A.\btxfnamespaceshort C.
  \btxlastnamefont {Tricco}}, \btxnamefont {V.\btxfnamespaceshort A.
  \btxlastnamefont {Welch}}, \btxnamefont {P.~\btxlastnamefont
  {Whiting}}\btxandcomma {} \btxandshort {.}\ \btxnamefont {D.~\btxlastnamefont
  {Moher}}\btxauthorcolon\ \btxjtitlefont {\btxifchangecase {The {PRISMA} 2020
  statement: an updated guideline for reporting systematic reviews}{The
  {PRISMA} 2020 statement: an updated guideline for reporting systematic
  reviews}}.
\newblock \btxjournalfont {Syst. Rev.}, 10(1):89,
  \btxprintmonthyear{.}{3}{2021}{short}.

\bibitem {art_123}
\btxnamefont {C.~\btxlastnamefont {Pan}}, \btxnamefont {J.~\btxlastnamefont
  {Zhu}}, \btxnamefont {Z.~\btxlastnamefont {Kong}}, \btxnamefont
  {H.~\btxlastnamefont {Shi}}\btxandcomma {} \btxandshort {.}\ \btxnamefont
  {W.~\btxlastnamefont {Yang}}\btxauthorcolon\ \btxjtitlefont {\btxifchangecase
  {Dc-stgcn: Dual-channel based graph convolutional networks for network
  traffic forecasting}{Dc-stgcn: Dual-channel based graph convolutional
  networks for network traffic forecasting}}.
\newblock \btxjournalfont {Electronics (Switzerland)}, 10(9), 2021.
\newblock {\latintext
  \btxurlfont{https://www.scopus.com/inward/record.uri?eid=2-s2.0-85104608714&doi=10.3390

\bibitem {art_35}
\btxnamefont {J.~\btxlastnamefont {Pan}}, \btxnamefont {Z.~\btxlastnamefont
  {Li}}, \btxnamefont {S.~\btxlastnamefont {Shi}}, \btxnamefont
  {L.~\btxlastnamefont {Xu}}, \btxnamefont {J.~\btxlastnamefont
  {Yu}}\btxandcomma {} \btxandshort {.}\ \btxnamefont {X.~\btxlastnamefont
  {Wu}}\btxauthorcolon\ \btxjtitlefont {\btxifchangecase {Adaptive graph neural
  network based south china sea seawater temperature prediction and
  multivariate uncertainty correlation analysis}{Adaptive graph neural network
  based South China Sea seawater temperature prediction and multivariate
  uncertainty correlation analysis}}.
\newblock \btxjournalfont {Stochastic Environmental Research and Risk
  Assessment}, 37(5):1877 – 1896, 2023.
\newblock {\latintext
  \btxurlfont{https://www.scopus.com/inward/record.uri?eid=2-s2.0-85144879589&doi=10.1007

\bibitem {AutoSTG}
\btxnamefont {Z.~\btxlastnamefont {Pan}}, \btxnamefont {S.~\btxlastnamefont
  {Ke}}, \btxnamefont {X.~\btxlastnamefont {Yang}}, \btxnamefont
  {Y.~\btxlastnamefont {Liang}}, \btxnamefont {Y.~\btxlastnamefont {Yu}},
  \btxnamefont {J.~\btxlastnamefont {Zhang}}\btxandcomma {} \btxandshort {.}\
  \btxnamefont {Y.~\btxlastnamefont {Zheng}}\btxauthorcolon\ \btxtitlefont
  {\btxifchangecase {Autostg: Neural architecture search for predictions of
  spatio-temporal graph}{AutoSTG: Neural Architecture Search for Predictions of
  Spatio-Temporal Graph}}.
\newblock \Btxinshort {.}\ \btxtitlefont {Proceedings of the Web Conference
  2021}, WWW '21, \btxpageshort {.}\ 1846–1855, New York, NY, USA, 2021.
  \btxpublisherfont {Association for Computing Machinery}\ifbtxprintISBN {,
  \mbox{\btxISBN~\btxISBNfont {9781450383127}}}.
\newblock {\latintext \btxurlfont{https://doi.org/10.1145/3442381.3449816}}.

\bibitem {ST-MetaNet}
\btxnamefont {Z.~\btxlastnamefont {Pan}}, \btxnamefont {Y.~\btxlastnamefont
  {Liang}}, \btxnamefont {W.~\btxlastnamefont {Wang}}, \btxnamefont
  {Y.~\btxlastnamefont {Yu}}, \btxnamefont {Y.~\btxlastnamefont
  {Zheng}}\btxandcomma {} \btxandshort {.}\ \btxnamefont {J.~\btxlastnamefont
  {Zhang}}\btxauthorcolon\ \btxtitlefont {\btxifchangecase {Urban traffic
  prediction from spatio-temporal data using deep meta learning}{Urban Traffic
  Prediction from Spatio-Temporal Data Using Deep Meta Learning}}.
\newblock \Btxinshort {.}\ \btxtitlefont {Proceedings of the 25th ACM SIGKDD
  International Conference on Knowledge Discovery \& Data Mining}, KDD '19,
  \btxpageshort {.}\ 1720–1730, New York, NY, USA, 2019. \btxpublisherfont
  {Association for Computing Machinery}\ifbtxprintISBN {,
  \mbox{\btxISBN~\btxISBNfont {9781450362016}}}.
\newblock {\latintext \btxurlfont{https://doi.org/10.1145/3292500.3330884}}.

\bibitem {art_60}
\btxnamefont {B.~\btxlastnamefont {Pang}}, \btxnamefont {W.~\btxlastnamefont
  {Wei}}, \btxnamefont {X.~\btxlastnamefont {Li}}, \btxnamefont
  {X.~\btxlastnamefont {Feng}}\btxandcomma {} \btxandshort {.}\ \btxnamefont
  {C.~\btxlastnamefont {Li}}\btxauthorcolon\ \btxjtitlefont {\btxifchangecase
  {A representation-learning-based approach to predict stock price trend via
  dynamic spatiotemporal feature embedding}{A representation-learning-based
  approach to predict stock price trend via dynamic spatiotemporal feature
  embedding}}.
\newblock \btxjournalfont {Engineering Applications of Artificial
  Intelligence}, 126, 2023.
\newblock {\latintext
  \btxurlfont{https://www.scopus.com/inward/record.uri?eid=2-s2.0-85166943902&doi=10.1016

\bibitem {art_738}
\btxnamefont {C.~\btxlastnamefont {Park}}, \btxnamefont {C.~\btxlastnamefont
  {Lee}}, \btxnamefont {H.~\btxlastnamefont {Bahng}}, \btxnamefont
  {Y.~\btxlastnamefont {Tae}}, \btxnamefont {S.~\btxlastnamefont {Jin}},
  \btxnamefont {K.~\btxlastnamefont {Kim}}, \btxnamefont {S.~\btxlastnamefont
  {Ko}}\btxandcomma {} \btxandshort {.}\ \btxnamefont {J.~\btxlastnamefont
  {Choo}}\btxauthorcolon\ \btxtitlefont {\btxifchangecase {{ST-GRAT}: A novel
  spatio-temporal graph attention networks for accurately forecasting
  dynamically changing road speed}{{ST-GRAT}: A novel spatio-temporal graph
  attention networks for accurately forecasting dynamically changing road
  speed}}.
\newblock \Btxinshort {.}\ \btxtitlefont {Proceedings of the 29th {ACM}
  International Conference on Information \& Knowledge Management}, New York,
  NY, USA, \btxprintmonthyear{.}{10}{2020}{short}. \btxpublisherfont {ACM}.

\bibitem {PyTorch}
\btxnamefont {A.~\btxlastnamefont {Paszke}}, \btxnamefont {S.~\btxlastnamefont
  {Gross}}, \btxnamefont {F.~\btxlastnamefont {Massa}}, \btxnamefont
  {A.~\btxlastnamefont {Lerer}}, \btxnamefont {J.~\btxlastnamefont {Bradbury}},
  \btxnamefont {G.~\btxlastnamefont {Chanan}}, \btxnamefont
  {T.~\btxlastnamefont {Killeen}}, \btxnamefont {Z.~\btxlastnamefont {Lin}},
  \btxnamefont {N.~\btxlastnamefont {Gimelshein}}, \btxnamefont
  {L.~\btxlastnamefont {Antiga}}, \btxnamefont {A.~\btxlastnamefont
  {Desmaison}}, \btxnamefont {A.~\btxlastnamefont {Kopf}}, \btxnamefont
  {E.~\btxlastnamefont {Yang}}, \btxnamefont {Z.~\btxlastnamefont {DeVito}},
  \btxnamefont {M.~\btxlastnamefont {Raison}}, \btxnamefont
  {A.~\btxlastnamefont {Tejani}}, \btxnamefont {S.~\btxlastnamefont
  {Chilamkurthy}}, \btxnamefont {B.~\btxlastnamefont {Steiner}}, \btxnamefont
  {L.~\btxlastnamefont {Fang}}, \btxnamefont {J.~\btxlastnamefont
  {Bai}}\btxandcomma {} \btxandshort {.}\ \btxnamefont {S.~\btxlastnamefont
  {Chintala}}\btxauthorcolon\ \btxtitlefont {\btxifchangecase {Pytorch: An
  imperative style, high-performance deep learning library}{PyTorch: An
  Imperative Style, High-Performance Deep Learning Library}}.
\newblock \Btxinshort {.}\ \btxtitlefont {Advances in Neural Information
  Processing Systems 32}, \btxpagesshort {.}\ 8024--8035. \btxpublisherfont
  {Curran Associates, Inc.}, 2019.
\newblock {\latintext
  \btxurlfont{http://papers.neurips.cc/paper/9015-pytorch-an-imperative-style-high-performance-deep-learning-library.pdf}}.

\bibitem {art_8}
\btxnamefont {Y.~\btxlastnamefont {Pei}}, \btxnamefont {C.\btxfnamespaceshort
  J. \btxlastnamefont {Huang}}, \btxnamefont {Y.~\btxlastnamefont
  {Shen}}\btxandcomma {} \btxandshort {.}\ \btxnamefont {Y.~\btxlastnamefont
  {Ma}}\btxauthorcolon\ \btxjtitlefont {\btxifchangecase {An ensemble model
  with adaptive variational mode decomposition and multivariate temporal graph
  neural network for pm2.5 concentration forecasting}{An Ensemble Model with
  Adaptive Variational Mode Decomposition and Multivariate Temporal Graph
  Neural Network for PM2.5 Concentration Forecasting}}.
\newblock \btxjournalfont {Sustainability (Switzerland)}, 14(20), 2022.
\newblock {\latintext
  \btxurlfont{https://www.scopus.com/inward/record.uri?eid=2-s2.0-85140791792&doi=10.3390

\bibitem {art_45}
\btxnamefont {X.~\btxlastnamefont {Peng}}, \btxnamefont {Q.~\btxlastnamefont
  {Li}}, \btxnamefont {L.~\btxlastnamefont {Chen}}, \btxnamefont
  {X.~\btxlastnamefont {Ning}}, \btxnamefont {H.~\btxlastnamefont
  {Chu}}\btxandcomma {} \btxandshort {.}\ \btxnamefont {J.~\btxlastnamefont
  {Liu}}\btxauthorcolon\ \btxjtitlefont {\btxifchangecase {A structured graph
  neural network for improving the numerical weather prediction of rainfall}{A
  Structured Graph Neural Network for Improving the Numerical Weather
  Prediction of Rainfall}}.
\newblock \btxjournalfont {Journal of Geophysical Research: Atmospheres}, 128,
  \btxprintmonthyear{.}{11}{2023}{short}.

\bibitem {art_425}
\btxnamefont {Y.~\btxlastnamefont {Peng}}, \btxnamefont {Y.~\btxlastnamefont
  {Guo}}, \btxnamefont {R.~\btxlastnamefont {Hao}}\btxandcomma {} \btxandshort
  {.}\ \btxnamefont {C.~\btxlastnamefont {Xu}}\btxauthorcolon\ \btxjtitlefont
  {\btxifchangecase {Network traffic prediction with attention-based
  {Spatial--Temporal} graph network}{Network traffic prediction with
  Attention-based {Spatial--Temporal} Graph Network}}.
\newblock \btxjournalfont {Comput. Netw.}, 243(110296):110296,
  \btxprintmonthyear{.}{4}{2024}{short}.

\bibitem {GNN_image}
\btxnamefont {P.~\btxlastnamefont {Pradhyumna}}, \btxnamefont
  {G.\btxfnamespaceshort P. \btxlastnamefont {Shreya}}\btxandcomma {}
  \btxandshort {.}\ \btxnamefont {\btxlastnamefont {Mohana}}\btxauthorcolon\
  \btxtitlefont {\btxifchangecase {Graph neural network (gnn) in image and
  video understanding using deep learning for computer vision
  applications}{Graph Neural Network (GNN) in Image and Video Understanding
  Using Deep Learning for Computer Vision Applications}}.
\newblock \Btxinshort {.}\ \btxtitlefont {2021 Second International Conference
  on Electronics and Sustainable Communication Systems (ICESC)}, \btxpagesshort
  {.}\ 1183--1189, 2021.

\bibitem {interpret_1}
\btxnamefont {M.\btxfnamespaceshort A. \btxlastnamefont {Prado-Romero}},
  \btxnamefont {B.~\btxlastnamefont {Prenkaj}}, \btxnamefont
  {G.~\btxlastnamefont {Stilo}}\btxandcomma {} \btxandshort {.}\ \btxnamefont
  {F.~\btxlastnamefont {Giannotti}}\btxauthorcolon\ \btxjtitlefont
  {\btxifchangecase {A survey on graph counterfactual explanations:
  Definitions, methods, evaluation, and research challenges}{A Survey on Graph
  Counterfactual Explanations: Definitions, Methods, Evaluation, and Research
  Challenges}}.
\newblock \btxjournalfont {ACM Comput. Surv.}, 56(7),
  \btxprintmonthyear{.}{apr}{2024}{short}\ifbtxprintISSN {,
  \mbox{\btxISSN~\btxISSNfont {0360-0300}}}.
\newblock {\latintext \btxurlfont{https://doi.org/10.1145/3618105}}.

\bibitem {art_351}
\btxnamefont {Y.~\btxlastnamefont {Pu}}, \btxnamefont {C.~\btxlastnamefont
  {Zhu}}, \btxnamefont {K.~\btxlastnamefont {Yang}}, \btxnamefont
  {Z.~\btxlastnamefont {L{\"u}}}\btxandcomma {} \btxandshort {.}\ \btxnamefont
  {Q.~\btxlastnamefont {Yang}}\btxauthorcolon\ \btxjtitlefont {\btxifchangecase
  {A novel multiscale transformer network framework for natural gas consumption
  forecasting}{A novel multiscale transformer network framework for natural gas
  consumption forecasting}}.
\newblock \btxjournalfont {IEEE Trans. Industr. Inform.}, 20(8):10040--10053,
  \btxprintmonthyear{.}{8}{2024}{short}.

\bibitem {GC-LSTM}
\btxnamefont {Y.~\btxlastnamefont {Qi}}, \btxnamefont {Q.~\btxlastnamefont
  {Li}}, \btxnamefont {H.~\btxlastnamefont {Karimian}}\btxandcomma {}
  \btxandshort {.}\ \btxnamefont {D.~\btxlastnamefont {Liu}}\btxauthorcolon\
  \btxjtitlefont {\btxifchangecase {A hybrid model for spatiotemporal
  forecasting of pm2.5 based on graph convolutional neural network and long
  short-term memory}{A hybrid model for spatiotemporal forecasting of PM2.5
  based on graph convolutional neural network and long short-term memory}}.
\newblock \btxjournalfont {Science of The Total Environment}, 664:1--10,
  2019\ifbtxprintISSN {, \mbox{\btxISSN~\btxISSNfont {0048-9697}}}.
\newblock {\latintext
  \btxurlfont{https://www.sciencedirect.com/science/article/pii/S0048969719303821}}.

\bibitem {art_26}
\btxnamefont {C.~\btxlastnamefont {Qin}}, \btxnamefont {A.\btxfnamespaceshort
  K. \btxlastnamefont {Srivastava}}, \btxnamefont {A.\btxfnamespaceshort Y.
  \btxlastnamefont {Saber}}, \btxnamefont {D.~\btxlastnamefont
  {Matthews}}\btxandcomma {} \btxandshort {.}\ \btxnamefont
  {K.~\btxlastnamefont {Davies}}\btxauthorcolon\ \btxjtitlefont
  {\btxifchangecase {Geometric deep-learning-based spatiotemporal forecasting
  for inverter-based solar power}{Geometric Deep-Learning-Based Spatiotemporal
  Forecasting for Inverter-Based Solar Power}}.
\newblock \btxjournalfont {IEEE Systems Journal}, 17(3):3425--3435, 2023.

\bibitem {DA-RNN}
\btxnamefont {Y.~\btxlastnamefont {Qin}}, \btxnamefont {D.~\btxlastnamefont
  {Song}}, \btxnamefont {H.~\btxlastnamefont {Chen}}, \btxnamefont
  {W.~\btxlastnamefont {Cheng}}, \btxnamefont {G.~\btxlastnamefont
  {Jiang}}\btxandcomma {} \btxandshort {.}\ \btxnamefont {G.\btxfnamespaceshort
  W. \btxlastnamefont {Cottrell}}\btxauthorcolon\ \btxtitlefont
  {\btxifchangecase {A dual-stage attention-based recurrent neural network for
  time series prediction}{A dual-stage attention-based recurrent neural network
  for time series prediction}}.
\newblock IJCAI'17, \btxpageshort {.}\ 2627–2633. \btxpublisherfont {AAAI
  Press}, 2017\ifbtxprintISBN {, \mbox{\btxISBN~\btxISBNfont {9780999241103}}}.

\bibitem {art_390}
\btxnamefont {X.~\btxlastnamefont {Qiu}}, \btxnamefont {J.~\btxlastnamefont
  {Qian}}, \btxnamefont {H.~\btxlastnamefont {Wang}}, \btxnamefont
  {X.~\btxlastnamefont {Tan}}\btxandcomma {} \btxandshort {.}\ \btxnamefont
  {Y.~\btxlastnamefont {Jin}}\btxauthorcolon\ \btxjtitlefont {\btxifchangecase
  {An attentive copula-based spatio-temporal graph model for multivariate
  time-series forecasting}{An attentive Copula-based spatio-temporal graph
  model for multivariate time-series forecasting}}.
\newblock \btxjournalfont {Appl. Soft Comput.}, 154(111324):111324,
  \btxprintmonthyear{.}{3}{2024}{short}.

\bibitem {art_729}
\btxnamefont {R.\btxfnamespaceshort S. \btxlastnamefont {Ramhormozi}},
  \btxnamefont {A.~\btxlastnamefont {Mozhdehi}}, \btxnamefont
  {S.~\btxlastnamefont {Kalantari}}, \btxnamefont {Y.~\btxlastnamefont {Wang}},
  \btxnamefont {S.~\btxlastnamefont {Sun}}\btxandcomma {} \btxandshort {.}\
  \btxnamefont {X.~\btxlastnamefont {Wang}}\btxauthorcolon\ \btxtitlefont
  {\btxifchangecase {Multi-task graph neural network for truck speed prediction
  under extreme weather conditions}{Multi-task graph neural network for truck
  speed prediction under extreme weather conditions}}.
\newblock \Btxinshort {.}\ \btxtitlefont {Proceedings of the 30th International
  Conference on Advances in Geographic Information Systems}, New York, NY, USA,
  \btxprintmonthyear{.}{11}{2022}{short}. \btxpublisherfont {ACM}.

\bibitem {DeepState}
\btxnamefont {S.\btxfnamespaceshort S. \btxlastnamefont {Rangapuram}},
  \btxnamefont {M.~\btxlastnamefont {Seeger}}, \btxnamefont
  {J.~\btxlastnamefont {Gasthaus}}, \btxnamefont {L.~\btxlastnamefont
  {Stella}}, \btxnamefont {Y.~\btxlastnamefont {Wang}}\btxandcomma {}
  \btxandshort {.}\ \btxnamefont {T.~\btxlastnamefont
  {Januschowski}}\btxauthorcolon\ \btxtitlefont {\btxifchangecase {Deep state
  space models for time series forecasting}{Deep state space models for time
  series forecasting}}.
\newblock \Btxinshort {.}\ \btxtitlefont {Proceedings of the 32nd International
  Conference on Neural Information Processing Systems}, NIPS'18, \btxpageshort
  {.}\ 7796–7805, Red Hook, NY, USA, 2018. \btxpublisherfont {Curran
  Associates Inc.}

\bibitem {TimeGrad}
\btxnamefont {K.~\btxlastnamefont {Rasul}}, \btxnamefont {C.~\btxlastnamefont
  {Seward}}, \btxnamefont {I.~\btxlastnamefont {Schuster}}\btxandcomma {}
  \btxandshort {.}\ \btxnamefont {R.~\btxlastnamefont
  {Vollgraf}}\btxauthorcolon\ \btxtitlefont {\btxifchangecase {Autoregressive
  denoising diffusion models for multivariate probabilistic time series
  forecasting}{Autoregressive Denoising Diffusion Models for Multivariate
  Probabilistic Time Series Forecasting}}.
\newblock \Btxinshort {.}\ \btxnamefont {M.~\btxlastnamefont {Meila}}
  \btxandshort {.}\ \btxnamefont {T.~\btxlastnamefont {Zhang}}\
  (\btxeditorsshort {.}): \btxtitlefont {Proceedings of the 38th International
  Conference on Machine Learning}, \btxvolumeshort {.}\ \btxvolumefont {139}
  \btxofseriesshort {.}\ \btxtitlefont {Proceedings of Machine Learning
  Research}, \btxpagesshort {.}\ 8857--8868. \btxpublisherfont {PMLR},
  \btxprintmonthyear{.}{18--24 Jul}{2021}{short}.
\newblock {\latintext
  \btxurlfont{https://proceedings.mlr.press/v139/rasul21a.html}}.

\bibitem {art_337}
\btxnamefont {K.~\btxlastnamefont {Rawal}} \btxandshort {.}\ \btxnamefont
  {A.~\btxlastnamefont {Ahmad}}\btxauthorcolon\ \btxjtitlefont
  {\btxifchangecase {Mining latent patterns with multi-scale decomposition for
  electricity demand and price forecasting using modified deep graph
  convolutional neural networks}{Mining latent patterns with multi-scale
  decomposition for electricity demand and price forecasting using modified
  deep graph convolutional neural networks}}.
\newblock \btxjournalfont {Sustain. Energy Grids Netw.}, 39(101436):101436,
  \btxprintmonthyear{.}{9}{2024}{short}.

\bibitem {art_343}
\btxnamefont {K.~\btxlastnamefont {Rawal}} \btxandshort {.}\ \btxnamefont
  {A.~\btxlastnamefont {Ahmad}}\btxauthorcolon\ \btxjtitlefont
  {\btxifchangecase {Towards efficient model recommendation: An innovative
  hybrid graph neural network approach integrating multisignature analysis of
  electrical time series}{Towards efficient model recommendation: An innovative
  hybrid graph neural network approach integrating multisignature analysis of
  electrical time series}}.
\newblock \btxjournalfont {e-Prime - Advances in Electrical Engineering,
  Electronics and Energy}, 8(100544):100544,
  \btxprintmonthyear{.}{6}{2024}{short}.

\bibitem {art_361}
\btxnamefont {K.~\btxlastnamefont {Ren}}, \btxnamefont {K.~\btxlastnamefont
  {Chen}}, \btxnamefont {C.~\btxlastnamefont {Jin}}, \btxnamefont
  {X.~\btxlastnamefont {Li}}, \btxnamefont {Y.~\btxlastnamefont
  {Yu}}\btxandcomma {} \btxandshort {.}\ \btxnamefont {Y.~\btxlastnamefont
  {Lin}}\btxauthorcolon\ \btxjtitlefont {\btxifchangecase {{TEMDI}: A temporal
  enhanced multisource data integration model for accurate {PM2.5}
  concentration forecasting}{{TEMDI}: A Temporal Enhanced Multisource Data
  Integration model for accurate {PM2.5} concentration forecasting}}.
\newblock \btxjournalfont {Atmos. Pollut. Res.}, 15(11):102269,
  \btxprintmonthyear{.}{11}{2024}{short}.

\bibitem {art_31}
\btxnamefont {Y.~\btxlastnamefont {Ren}}, \btxnamefont {Z.~\btxlastnamefont
  {Li}}, \btxnamefont {L.~\btxlastnamefont {Xu}}\btxandcomma {} \btxandshort
  {.}\ \btxnamefont {J.~\btxlastnamefont {Yu}}\btxauthorcolon\ \btxjtitlefont
  {\btxifchangecase {The data-based adaptive graph learning network for
  analysis and prediction of offshore wind speed}{The data-based adaptive graph
  learning network for analysis and prediction of offshore wind speed}}.
\newblock \btxjournalfont {Energy}, 267, 2023.
\newblock {\latintext
  \btxurlfont{https://www.scopus.com/inward/record.uri?eid=2-s2.0-85146061902&doi=10.1016

\bibitem {art_705}
\btxnamefont {B.~\btxlastnamefont {Rozemberczki}}, \btxnamefont
  {P.~\btxlastnamefont {Scherer}}, \btxnamefont {Y.~\btxlastnamefont {He}},
  \btxnamefont {G.~\btxlastnamefont {Panagopoulos}}, \btxnamefont
  {A.~\btxlastnamefont {Riedel}}, \btxnamefont {M.~\btxlastnamefont
  {Astefanoaei}}, \btxnamefont {O.~\btxlastnamefont {Kiss}}, \btxnamefont
  {F.~\btxlastnamefont {Beres}}, \btxnamefont {G.~\btxlastnamefont
  {L{\'o}pez}}, \btxnamefont {N.~\btxlastnamefont {Collignon}}\btxandcomma {}
  \btxandshort {.}\ \btxnamefont {R.~\btxlastnamefont {Sarkar}}\btxauthorcolon\
  \btxtitlefont {\btxifchangecase {{PyTorch} geometric temporal: Spatiotemporal
  signal processing with neural machine learning models}{{PyTorch} geometric
  temporal: Spatiotemporal signal processing with neural machine learning
  models}}.
\newblock \Btxinshort {.}\ \btxtitlefont {Proceedings of the 30th {ACM}
  International Conference on Information \& Knowledge Management},
  \btxvolumeshort {.}~\btxvolumefont {33}, \btxpagesshort {.}\ 4564--4573, New
  York, NY, USA, \btxprintmonthyear{.}{10}{2021}{short}. \btxpublisherfont
  {ACM}.

\bibitem {art_141}
\btxnamefont {S.~\btxlastnamefont {Ruan}}, \btxnamefont {S.~\btxlastnamefont
  {Han}}, \btxnamefont {C.~\btxlastnamefont {Lu}}\btxandcomma {} \btxandshort
  {.}\ \btxnamefont {Q.~\btxlastnamefont {Gu}}\btxauthorcolon\ \btxjtitlefont
  {\btxifchangecase {Proactive control model for safety prediction in tailing
  dam management: Applying graph depth learning optimization}{Proactive control
  model for safety prediction in tailing dam management: Applying graph depth
  learning optimization}}.
\newblock \btxjournalfont {Process Safety and Environmental Protection},
  172:329 – 340, 2023.
\newblock {\latintext
  \btxurlfont{https://www.scopus.com/inward/record.uri?eid=2-s2.0-85148375797&doi=10.1016

\bibitem {DeepAR}
\btxnamefont {D.~\btxlastnamefont {Salinas}}, \btxnamefont {V.~\btxlastnamefont
  {Flunkert}}, \btxnamefont {J.~\btxlastnamefont {Gasthaus}}\btxandcomma {}
  \btxandshort {.}\ \btxnamefont {T.~\btxlastnamefont
  {Januschowski}}\btxauthorcolon\ \btxjtitlefont {\btxifchangecase {Deepar:
  Probabilistic forecasting with autoregressive recurrent networks}{DeepAR:
  Probabilistic forecasting with autoregressive recurrent networks}}.
\newblock \btxjournalfont {International Journal of Forecasting},
  36(3):1181--1191, 2020\ifbtxprintISSN {, \mbox{\btxISSN~\btxISSNfont
  {0169-2070}}}.
\newblock {\latintext
  \btxurlfont{https://www.sciencedirect.com/science/article/pii/S0169207019301888}}.

\bibitem {art_431}
\btxnamefont {J.\btxfnamespaceshort F. \btxlastnamefont {S{\'a}nchez-Rada}},
  \btxnamefont {R.~\btxlastnamefont {Vila-Rodr{\'\i}guez}}, \btxnamefont
  {J.~\btxlastnamefont {Montes}}\btxandcomma {} \btxandshort {.}\ \btxnamefont
  {P.\btxfnamespaceshort J. \btxlastnamefont {Zufiria}}\btxauthorcolon\
  \btxjtitlefont {\btxifchangecase {Predicting the aggregate mobility of a
  vehicle fleet within a city graph}{Predicting the aggregate mobility of a
  vehicle fleet within a city graph}}.
\newblock \btxjournalfont {Algorithms}, 17(4):166,
  \btxprintmonthyear{.}{4}{2024}{short}.

\bibitem {STHGCN}
\btxnamefont {R.~\btxlastnamefont {Sawhney}}, \btxnamefont {S.~\btxlastnamefont
  {Agarwal}}, \btxnamefont {A.~\btxlastnamefont {Wadhwa}}\btxandcomma {}
  \btxandshort {.}\ \btxnamefont {R.\btxfnamespaceshort R. \btxlastnamefont
  {Shah}}\btxauthorcolon\ \btxtitlefont {\btxifchangecase {Spatiotemporal
  hypergraph convolution network for stock movement forecasting}{Spatiotemporal
  Hypergraph Convolution Network for Stock Movement Forecasting}}.
\newblock \Btxinshort {.}\ \btxtitlefont {2020 IEEE International Conference on
  Data Mining (ICDM)}, \btxpagesshort {.}\ 482--491, 2020.

\bibitem {Scarselli2009}
\btxnamefont {F.~\btxlastnamefont {Scarselli}}, \btxnamefont
  {M.~\btxlastnamefont {Gori}}, \btxnamefont {A.\btxfnamespaceshort C.
  \btxlastnamefont {Tsoi}}, \btxnamefont {M.~\btxlastnamefont
  {Hagenbuchner}}\btxandcomma {} \btxandshort {.}\ \btxnamefont
  {G.~\btxlastnamefont {Monfardini}}\btxauthorcolon\ \btxjtitlefont
  {\btxifchangecase {The graph neural network model}{The Graph Neural Network
  Model}}.
\newblock \btxjournalfont {IEEE Transactions on Neural Networks},
  20(1):61–80, \btxprintmonthyear{.}{1}{2009}{short}\ifbtxprintISSN {,
  \mbox{\btxISSN~\btxISSNfont {1941-0093}}}.
\newblock {\latintext \btxurlfont{http://dx.doi.org/10.1109/TNN.2008.2005605}}.

\bibitem {WEASEL+MUSE}
\btxnamefont {P.~\btxlastnamefont {Schäfer}} \btxandshort {.}\ \btxnamefont
  {U.~\btxlastnamefont {Leser}}\btxauthorcolon\ \btxjtitlefont
  {\btxifchangecase {Multivariate time series classification with
  weasel+muse}{Multivariate Time Series Classification with WEASEL+MUSE}}.
\newblock \btxprintmonthyear{.}{11}{2017}{short}.

\bibitem {art_659}
\btxnamefont {I.~\btxlastnamefont {Segovia~Dominguez}}, \btxnamefont
  {H.~\btxlastnamefont {Lee}}, \btxnamefont {Y.~\btxlastnamefont {Chen}},
  \btxnamefont {M.~\btxlastnamefont {Garay}}, \btxnamefont
  {K.\btxfnamespaceshort M. \btxlastnamefont {Gorski}}\btxandcomma {}
  \btxandshort {.}\ \btxnamefont {Y.\btxfnamespaceshort R. \btxlastnamefont
  {Gel}}\btxauthorcolon\ \btxtitlefont {\btxifchangecase {Does air quality
  really impact {COVID-19} clinical severity: Coupling {NASA} satellite
  datasets with geometric deep learning}{Does air quality really impact
  {COVID-19} clinical severity: Coupling {NASA} satellite datasets with
  geometric deep learning}}.
\newblock \Btxinshort {.}\ \btxtitlefont {Proceedings of the 27th {ACM}
  {SIGKDD} Conference on Knowledge Discovery \& Data Mining}, New York, NY,
  USA, \btxprintmonthyear{.}{8}{2021}{short}. \btxpublisherfont {ACM}.

\bibitem {DeepGLO}
\btxnamefont {R.~\btxlastnamefont {Sen}}, \btxnamefont {H.\btxfnamespaceshort
  F. \btxlastnamefont {Yu}}\btxandcomma {} \btxandshort {.}\ \btxnamefont
  {I.~\btxlastnamefont {Dhillon}}\btxauthorcolon\ \btxtitlefont {Think
  globally, act locally: a deep neural network approach to high-dimensional
  time series forecasting}.
\newblock \btxpublisherfont {Curran Associates Inc.}, Red Hook, NY, USA, 2019.

\bibitem {art_365}
\btxnamefont {Y.~\btxlastnamefont {Seol}}, \btxnamefont {S.~\btxlastnamefont
  {Kim}}, \btxnamefont {M.~\btxlastnamefont {Jung}}\btxandcomma {} \btxandshort
  {.}\ \btxnamefont {Y.~\btxlastnamefont {Hong}}\btxauthorcolon\ \btxjtitlefont
  {\btxifchangecase {A novel physics-aware graph network using high-order
  numerical methods in weather forecasting model}{A novel physics-aware graph
  network using high-order numerical methods in weather forecasting model}}.
\newblock \btxjournalfont {Knowl. Based Syst.}, 300(112158):112158,
  \btxprintmonthyear{.}{9}{2024}{short}.

\bibitem {art_404}
\btxnamefont {T.~\btxlastnamefont {Shaik}}, \btxnamefont {X.~\btxlastnamefont
  {Tao}}, \btxnamefont {H.~\btxlastnamefont {Xie}}, \btxnamefont
  {L.~\btxlastnamefont {Li}}, \btxnamefont {J.~\btxlastnamefont
  {Yong}}\btxandcomma {} \btxandshort {.}\ \btxnamefont {Y.~\btxlastnamefont
  {Li}}\btxauthorcolon\ \btxjtitlefont {\btxifchangecase {Graph-enabled
  reinforcement learning for time series forecasting with adaptive
  intelligence}{Graph-enabled reinforcement learning for time series
  forecasting with adaptive intelligence}}.
\newblock \btxjournalfont {IEEE Trans. Emerg. Top. Comput. Intell.},
  8(4):2908--2918, \btxprintmonthyear{.}{8}{2024}{short}.

\bibitem {art_641}
\btxnamefont {C.~\btxlastnamefont {Shang}}, \btxnamefont {J.~\btxlastnamefont
  {Chen}}\btxandcomma {} \btxandshort {.}\ \btxnamefont {J.~\btxlastnamefont
  {Bi}}\btxauthorcolon\ \btxtitlefont {\btxifchangecase {Discrete graph
  structure learning for forecasting multiple time series}{Discrete Graph
  Structure Learning for Forecasting Multiple Time Series}}.
\newblock \Btxinshort {.}\ \btxtitlefont {International Conference on Learning
  Representations}, 2021.
\newblock {\latintext \btxurlfont{https://openreview.net/forum?id=WEHSlH5mOk}}.

\bibitem {cit_rev6}
\btxnamefont {Y.~\btxlastnamefont {Shao}}, \btxnamefont {H.~\btxlastnamefont
  {Li}}, \btxnamefont {X.~\btxlastnamefont {Gu}}, \btxnamefont
  {H.~\btxlastnamefont {Yin}}, \btxnamefont {Y.~\btxlastnamefont {Li}},
  \btxnamefont {X.~\btxlastnamefont {Miao}}, \btxnamefont {W.~\btxlastnamefont
  {Zhang}}, \btxnamefont {B.~\btxlastnamefont {Cui}}\btxandcomma {}
  \btxandshort {.}\ \btxnamefont {L.~\btxlastnamefont {Chen}}\btxauthorcolon\
  \btxjtitlefont {\btxifchangecase {Distributed graph neural network training:
  A survey}{Distributed Graph Neural Network Training: A Survey}}.
\newblock \btxjournalfont {ACM Comput. Surv.}, 56(8),
  \btxprintmonthyear{.}{apr}{2024}{short}\ifbtxprintISSN {,
  \mbox{\btxISSN~\btxISSNfont {0360-0300}}}.
\newblock {\latintext \btxurlfont{https://doi.org/10.1145/3648358}}.

\bibitem {STID}
\btxnamefont {Z.~\btxlastnamefont {Shao}}, \btxnamefont {Z.~\btxlastnamefont
  {Zhang}}, \btxnamefont {F.~\btxlastnamefont {Wang}}, \btxnamefont
  {W.~\btxlastnamefont {Wei}}\btxandcomma {} \btxandshort {.}\ \btxnamefont
  {Y.~\btxlastnamefont {Xu}}\btxauthorcolon\ \btxtitlefont {\btxifchangecase
  {Spatial-temporal identity: A simple yet effective baseline for multivariate
  time series forecasting}{Spatial-Temporal Identity: A Simple yet Effective
  Baseline for Multivariate Time Series Forecasting}}.
\newblock \Btxinshort {.}\ \btxtitlefont {Proceedings of the 31st ACM
  International Conference on Information \& Knowledge Management}, CIKM '22,
  \btxpageshort {.}\ 4454–4458, New York, NY, USA, 2022. \btxpublisherfont
  {Association for Computing Machinery}\ifbtxprintISBN {,
  \mbox{\btxISBN~\btxISBNfont {9781450392365}}}.
\newblock {\latintext \btxurlfont{https://doi.org/10.1145/3511808.3557702}}.

\bibitem {art_652}
\btxnamefont {Z.~\btxlastnamefont {Shao}}, \btxnamefont {Z.~\btxlastnamefont
  {Zhang}}, \btxnamefont {F.~\btxlastnamefont {Wang}}\btxandcomma {}
  \btxandshort {.}\ \btxnamefont {Y.~\btxlastnamefont {Xu}}\btxauthorcolon\
  \btxtitlefont {\btxifchangecase {Pre-training enhanced spatial-temporal graph
  neural network for multivariate time series forecasting}{Pre-training
  enhanced spatial-temporal graph neural network for multivariate time series
  forecasting}}.
\newblock \Btxinshort {.}\ \btxtitlefont {Proceedings of the 28th {ACM}
  {SIGKDD} Conference on Knowledge Discovery and Data Mining}, \btxpagesshort
  {.}\ 1567--1577, New York, NY, USA, \btxprintmonthyear{.}{8}{2022}{short}.
  \btxpublisherfont {ACM}.

\bibitem {art_672}
\btxnamefont {Z.~\btxlastnamefont {Shao}}, \btxnamefont {Z.~\btxlastnamefont
  {Zhang}}, \btxnamefont {W.~\btxlastnamefont {Wei}}, \btxnamefont
  {F.~\btxlastnamefont {Wang}}, \btxnamefont {Y.~\btxlastnamefont {Xu}},
  \btxnamefont {X.~\btxlastnamefont {Cao}}\btxandcomma {} \btxandshort {.}\
  \btxnamefont {C.\btxfnamespaceshort S. \btxlastnamefont
  {Jensen}}\btxauthorcolon\ \btxjtitlefont {\btxifchangecase {Decoupled dynamic
  spatial-temporal graph neural network for traffic forecasting}{Decoupled
  dynamic spatial-temporal graph neural network for traffic forecasting}}.
\newblock \btxjournalfont {Proceedings VLDB Endowment}, 15(11):2733--2746,
  \btxprintmonthyear{.}{7}{2022}{short}.

\bibitem {art_692}
\btxnamefont {S.~\btxlastnamefont {Sharma}}, \btxnamefont {S.~\btxlastnamefont
  {Iyengar}}, \btxnamefont {S.~\btxlastnamefont {Zheng}}, \btxnamefont
  {K.~\btxlastnamefont {Kapoor}}, \btxnamefont {W.~\btxlastnamefont {Cao}},
  \btxnamefont {J.~\btxlastnamefont {Bian}}, \btxnamefont {S.~\btxlastnamefont
  {Kalyanaraman}}\btxandcomma {} \btxandshort {.}\ \btxnamefont
  {J.~\btxlastnamefont {Lemmon}}\btxauthorcolon\ \btxtitlefont
  {\btxifchangecase {A graph-based spatiotemporal model for energy markets}{A
  graph-based spatiotemporal model for energy markets}}.
\newblock \Btxinshort {.}\ \btxtitlefont {Proceedings of the 31st {ACM}
  International Conference on Information \& Knowledge Management}, New York,
  NY, USA, \btxprintmonthyear{.}{10}{2022}{short}. \btxpublisherfont {ACM}.

\bibitem {art_145}
\btxnamefont {F.~\btxlastnamefont {Shen}}, \btxnamefont {J.~\btxlastnamefont
  {Wang}}, \btxnamefont {Z.~\btxlastnamefont {Zhang}}, \btxnamefont
  {X.~\btxlastnamefont {Wang}}, \btxnamefont {Y.~\btxlastnamefont {Li}},
  \btxnamefont {Z.~\btxlastnamefont {Geng}}, \btxnamefont {B.~\btxlastnamefont
  {Pan}}, \btxnamefont {Z.~\btxlastnamefont {Lu}}, \btxnamefont
  {W.~\btxlastnamefont {Zhao}}\btxandcomma {} \btxandshort {.}\ \btxnamefont
  {W.~\btxlastnamefont {Zhu}}\btxauthorcolon\ \btxjtitlefont {\btxifchangecase
  {Long-term multivariate time series forecasting in data centers based on
  multi-factor separation evolutionary spatial–temporal graph neural
  networks}{Long-term multivariate time series forecasting in data centers
  based on multi-factor separation evolutionary spatial–temporal graph neural
  networks}}.
\newblock \btxjournalfont {Knowledge-Based Systems}, 280:110997,
  2023\ifbtxprintISSN {, \mbox{\btxISSN~\btxISSNfont {0950-7051}}}.
\newblock {\latintext
  \btxurlfont{https://www.sciencedirect.com/science/article/pii/S0950705123007475}}.

\bibitem {art_47}
\btxnamefont {L.~\btxlastnamefont {Sheng}}, \btxnamefont {L.~\btxlastnamefont
  {Xu}}, \btxnamefont {J.~\btxlastnamefont {Yu}}\btxandcomma {} \btxandshort
  {.}\ \btxnamefont {Z.~\btxlastnamefont {Li}}\btxauthorcolon\ \btxjtitlefont
  {\btxifchangecase {A graph multi-head self-attention neural network for the
  multi-point long-term prediction of sea surface temperature}{A graph
  multi-head self-attention neural network for the multi-point long-term
  prediction of sea surface temperature}}.
\newblock \btxjournalfont {Remote Sensing Letters}, 14:786--796,
  \btxprintmonthyear{.}{07}{2023}{short}.

\bibitem {ConvLSTM}
\btxnamefont {X.~\btxlastnamefont {Shi}}, \btxnamefont {Z.~\btxlastnamefont
  {Chen}}, \btxnamefont {H.~\btxlastnamefont {Wang}}, \btxnamefont
  {D.\btxfnamespaceshort Y. \btxlastnamefont {Yeung}}, \btxnamefont
  {W.\btxfnamespaceshort k. \btxlastnamefont {Wong}}\btxandcomma {}
  \btxandshort {.}\ \btxnamefont {W.\btxfnamespaceshort c. \btxlastnamefont
  {Woo}}\btxauthorcolon\ \btxtitlefont {\btxifchangecase {Convolutional lstm
  network: a machine learning approach for precipitation
  nowcasting}{Convolutional LSTM Network: a machine learning approach for
  precipitation nowcasting}}.
\newblock NIPS'15, \btxpageshort {.}\ 802–810, Cambridge, MA, USA, 2015.
  \btxpublisherfont {MIT Press}.

\bibitem {TPA-LSTM}
\btxnamefont {S.\btxfnamespaceshort Y. \btxlastnamefont {Shih}}, \btxnamefont
  {F.\btxfnamespaceshort K. \btxlastnamefont {Sun}}\btxandcomma {} \btxandshort
  {.}\ \btxnamefont {H.\btxfnamespaceshort y. \btxlastnamefont
  {Lee}}\btxauthorcolon\ \btxjtitlefont {\btxifchangecase {Temporal pattern
  attention for multivariate time series forecasting}{Temporal pattern
  attention for multivariate time series forecasting}}.
\newblock \btxjournalfont {Machine Learning}, 108(8–9):1421–1441,
  \btxprintmonthyear{.}{6}{2019}{short}\ifbtxprintISSN {,
  \mbox{\btxISSN~\btxISSNfont {1573-0565}}}.
\newblock {\latintext
  \btxurlfont{http://dx.doi.org/10.1007/s10994-019-05815-0}}.

\bibitem {art_441}
\btxnamefont {Y.~\btxlastnamefont {Shin}} \btxandshort {.}\ \btxnamefont
  {Y.~\btxlastnamefont {Yoon}}\btxauthorcolon\ \btxjtitlefont {\btxifchangecase
  {{PGCN}: Progressive graph convolutional networks for spatial--temporal
  traffic forecasting}{{PGCN}: Progressive graph convolutional networks for
  spatial--temporal traffic forecasting}}.
\newblock \btxjournalfont {IEEE Trans. Intell. Transp. Syst.},
  25(7):7633--7644, \btxprintmonthyear{.}{7}{2024}{short}.

\bibitem {mTANs}
\btxnamefont {S.\btxfnamespaceshort N. \btxlastnamefont {Shukla}} \btxandshort
  {.}\ \btxnamefont {B.~\btxlastnamefont {Marlin}}\btxauthorcolon\
  \btxtitlefont {\btxifchangecase {Multi-time attention networks for
  irregularly sampled time series}{Multi-Time Attention Networks for
  Irregularly Sampled Time Series}}.
\newblock \Btxinshort {.}\ \btxtitlefont {International Conference on Learning
  Representations}, 2021.
\newblock {\latintext \btxurlfont{https://openreview.net/forum?id=4c0J6lwQ4_}}.

\bibitem {art_21}
\btxnamefont {J.~\btxlastnamefont {Simeunović}}, \btxnamefont
  {B.~\btxlastnamefont {Schubnel}}, \btxnamefont {P.\btxfnamespaceshort J.
  \btxlastnamefont {Alet}}\btxandcomma {} \btxandshort {.}\ \btxnamefont
  {R.\btxfnamespaceshort E. \btxlastnamefont {Carrillo}}\btxauthorcolon\
  \btxjtitlefont {\btxifchangecase {Spatio-temporal graph neural networks for
  multi-site pv power forecasting}{Spatio-Temporal Graph Neural Networks for
  Multi-Site PV Power Forecasting}}.
\newblock \btxjournalfont {IEEE Transactions on Sustainable Energy},
  13(2):1210--1220, 2022.

\bibitem {art_22}
\btxnamefont {J.~\btxlastnamefont {Simeunović}}, \btxnamefont
  {B.~\btxlastnamefont {Schubnel}}, \btxnamefont {P.\btxfnamespaceshort J.
  \btxlastnamefont {Alet}}, \btxnamefont {R.\btxfnamespaceshort E.
  \btxlastnamefont {Carrillo}}\btxandcomma {} \btxandshort {.}\ \btxnamefont
  {P.~\btxlastnamefont {Frossard}}\btxauthorcolon\ \btxjtitlefont
  {\btxifchangecase {Interpretable temporal-spatial graph attention network for
  multi-site pv power forecasting}{Interpretable temporal-spatial graph
  attention network for multi-site PV power forecasting}}.
\newblock \btxjournalfont {Applied Energy}, 327, 2022.
\newblock {\latintext
  \btxurlfont{https://www.scopus.com/inward/record.uri?eid=2-s2.0-85140322468&doi=10.1016

\bibitem {art_432}
\btxnamefont {V.~\btxlastnamefont {Singh}}, \btxnamefont {S.\btxfnamespaceshort
  K. \btxlastnamefont {Sahana}}\btxandcomma {} \btxandshort {.}\ \btxnamefont
  {V.~\btxlastnamefont {Bhattacharjee}}\btxauthorcolon\ \btxjtitlefont
  {\btxifchangecase {Integrated spatio-temporal graph neural network for
  traffic forecasting}{Integrated spatio-temporal graph neural network for
  traffic forecasting}}.
\newblock \btxjournalfont {Appl. Sci. (Basel)}, 14(24):11477,
  \btxprintmonthyear{.}{12}{2024}{short}.

\bibitem {STSGCN}
\btxnamefont {C.~\btxlastnamefont {Song}}, \btxnamefont {Y.~\btxlastnamefont
  {Lin}}, \btxnamefont {S.~\btxlastnamefont {Guo}}\btxandcomma {} \btxandshort
  {.}\ \btxnamefont {H.~\btxlastnamefont {Wan}}\btxauthorcolon\ \btxjtitlefont
  {\btxifchangecase {Spatial-temporal synchronous graph convolutional networks:
  A new framework for spatial-temporal network data
  forecasting}{Spatial-Temporal Synchronous Graph Convolutional Networks: A New
  Framework for Spatial-Temporal Network Data Forecasting}}.
\newblock \btxjournalfont {Proceedings of the AAAI Conference on Artificial
  Intelligence}, 34:914--921, \btxprintmonthyear{.}{04}{2020}{short}.

\bibitem {DGCNN}
\btxnamefont {T.~\btxlastnamefont {Song}}, \btxnamefont {W.~\btxlastnamefont
  {Zheng}}, \btxnamefont {P.~\btxlastnamefont {Song}}\btxandcomma {}
  \btxandshort {.}\ \btxnamefont {Z.~\btxlastnamefont {Cui}}\btxauthorcolon\
  \btxjtitlefont {\btxifchangecase {Eeg emotion recognition using dynamical
  graph convolutional neural networks}{EEG Emotion Recognition Using Dynamical
  Graph Convolutional Neural Networks}}.
\newblock \btxjournalfont {IEEE Transactions on Affective Computing},
  11(3):532--541, 2020.

\bibitem {art_83}
\btxnamefont {A.~\btxlastnamefont {Sriramulu}}, \btxnamefont
  {N.~\btxlastnamefont {Fourrier}}\btxandcomma {} \btxandshort {.}\
  \btxnamefont {C.~\btxlastnamefont {Bergmeir}}\btxauthorcolon\ \btxjtitlefont
  {\btxifchangecase {Adaptive dependency learning graph neural
  networks}{Adaptive dependency learning graph neural networks}}.
\newblock \btxjournalfont {Information Sciences}, 625:700 – 714, 2023.
\newblock {\latintext
  \btxurlfont{https://www.scopus.com/inward/record.uri?eid=2-s2.0-85146595559&doi=10.1016

\bibitem {art_696}
\btxnamefont {H.~\btxlastnamefont {Su}}, \btxnamefont {X.~\btxlastnamefont
  {Wang}}\btxandcomma {} \btxandshort {.}\ \btxnamefont {Y.~\btxlastnamefont
  {Qin}}\btxauthorcolon\ \btxtitlefont {\btxifchangecase {Agcnt: Adaptive graph
  convolutional network for transformer-based long sequence time-series
  forecasting}{AGCNT: Adaptive Graph Convolutional Network for
  Transformer-based Long Sequence Time-Series Forecasting}}.
\newblock \Btxinshort {.}\ \btxtitlefont {Proceedings of the 30th ACM
  International Conference on Information \& Knowledge Management}, CIKM '21,
  \btxpageshort {.}\ 3439–3442, New York, NY, USA, 2021. \btxpublisherfont
  {Association for Computing Machinery}\ifbtxprintISBN {,
  \mbox{\btxISBN~\btxISBNfont {9781450384469}}}.
\newblock {\latintext \btxurlfont{https://doi.org/10.1145/3459637.3482054}}.

\bibitem {art_3}
\btxnamefont {I.\btxfnamespaceshort F. \btxlastnamefont {Su}}, \btxnamefont
  {Y.\btxfnamespaceshort C. \btxlastnamefont {Chung}}, \btxnamefont
  {C.~\btxlastnamefont {Lee}}\btxandcomma {} \btxandshort {.}\ \btxnamefont
  {P.\btxfnamespaceshort M. \btxlastnamefont {Huang}}\btxauthorcolon\
  \btxjtitlefont {\btxifchangecase {Effective pm2.5 concentration forecasting
  based on multiple spatial–temporal gnn for areas without monitoring
  stations}{Effective PM2.5 concentration forecasting based on multiple
  spatial–temporal GNN for areas without monitoring stations}}.
\newblock \btxjournalfont {Expert Systems with Applications}, 234, 2023.
\newblock {\latintext
  \btxurlfont{https://www.scopus.com/inward/record.uri?eid=2-s2.0-85172418127&doi=10.1016

\bibitem {OmniAnomaly}
\btxnamefont {Y.~\btxlastnamefont {Su}}, \btxnamefont {Y.~\btxlastnamefont
  {Zhao}}, \btxnamefont {C.~\btxlastnamefont {Niu}}, \btxnamefont
  {R.~\btxlastnamefont {Liu}}, \btxnamefont {W.~\btxlastnamefont
  {Sun}}\btxandcomma {} \btxandshort {.}\ \btxnamefont {D.~\btxlastnamefont
  {Pei}}\btxauthorcolon\ \btxtitlefont {\btxifchangecase {Robust anomaly
  detection for multivariate time series through stochastic recurrent neural
  network}{Robust Anomaly Detection for Multivariate Time Series through
  Stochastic Recurrent Neural Network}}.
\newblock \Btxinshort {.}\ \btxtitlefont {Proceedings of the 25th ACM SIGKDD
  International Conference on Knowledge Discovery \& Data Mining}, KDD '19,
  \btxpageshort {.}\ 2828–2837, New York, NY, USA, 2019. \btxpublisherfont
  {Association for Computing Machinery}\ifbtxprintISBN {,
  \mbox{\btxISBN~\btxISBNfont {9781450362016}}}.
\newblock {\latintext \btxurlfont{https://doi.org/10.1145/3292500.3330672}}.

\bibitem {MVGCN}
\btxnamefont {J.~\btxlastnamefont {Sun}}, \btxnamefont {J.~\btxlastnamefont
  {Zhang}}, \btxnamefont {Q.~\btxlastnamefont {Li}}, \btxnamefont
  {X.~\btxlastnamefont {Yi}}\btxandcomma {} \btxandshort {.}\ \btxnamefont
  {Y.~\btxlastnamefont {Zheng}}\btxauthorcolon\ \btxjtitlefont
  {\btxifchangecase {Predicting citywide crowd flows in irregular regions using
  multi-view graph convolutional networks}{Predicting Citywide Crowd Flows in
  Irregular Regions Using Multi-View Graph Convolutional Networks}}.
\newblock \btxjournalfont {IEEE Transactions on Knowledge and Data
  Engineering}, 34:2348--2359, 2019.
\newblock {\latintext
  \btxurlfont{https://api.semanticscholar.org/CorpusID:83458663}}.

\bibitem {art_401}
\btxnamefont {Y.~\btxlastnamefont {Sun}}, \btxnamefont {C.~\btxlastnamefont
  {Chen}}, \btxnamefont {Y.~\btxlastnamefont {Xu}}, \btxnamefont
  {S.~\btxlastnamefont {Xie}}, \btxnamefont {R.\btxfnamespaceshort S.
  \btxlastnamefont {Blum}}\btxandcomma {} \btxandshort {.}\ \btxnamefont
  {P.~\btxlastnamefont {Venkitasubramaniam}}\btxauthorcolon\ \btxjtitlefont
  {\btxifchangecase {On the generalization discrepancy of spatiotemporal
  dynamics-informed graph convolutional networks}{On the generalization
  discrepancy of spatiotemporal dynamics-informed graph convolutional
  networks}}.
\newblock \btxjournalfont {Front. Mech. Eng.}, 10,
  \btxprintmonthyear{.}{7}{2024}{short}.

\bibitem {art_41}
\btxnamefont {Y.~\btxlastnamefont {Sun}}, \btxnamefont {X.~\btxlastnamefont
  {Yao}}, \btxnamefont {X.~\btxlastnamefont {Bi}}, \btxnamefont
  {X.~\btxlastnamefont {Huang}}, \btxnamefont {X.~\btxlastnamefont
  {Zhao}}\btxandcomma {} \btxandshort {.}\ \btxnamefont {B.~\btxlastnamefont
  {Qiao}}\btxauthorcolon\ \btxjtitlefont {\btxifchangecase {Time-series graph
  network for sea surface temperature prediction}{Time-Series Graph Network for
  Sea Surface Temperature Prediction}}.
\newblock \btxjournalfont {Big Data Research}, 25, 2021.
\newblock {\latintext
  \btxurlfont{https://www.scopus.com/inward/record.uri?eid=2-s2.0-85108111192&doi=10.1016

\bibitem {art_65}
\btxnamefont {Z.~\btxlastnamefont {Sun}}, \btxnamefont {A.~\btxlastnamefont
  {Harit}}, \btxnamefont {A.\btxfnamespaceshort I. \btxlastnamefont {Cristea}},
  \btxnamefont {J.~\btxlastnamefont {Wang}}\btxandcomma {} \btxandshort {.}\
  \btxnamefont {P.~\btxlastnamefont {Lio}}\btxauthorcolon\ \btxjtitlefont
  {\btxifchangecase {Money: Ensemble learning for stock price movement
  prediction via a convolutional network with adversarial hypergraph
  model}{MONEY: Ensemble learning for stock price movement prediction via a
  convolutional network with adversarial hypergraph model}}.
\newblock \btxjournalfont {AI Open}, 4:165 – 174, 2023.
\newblock {\latintext
  \btxurlfont{https://www.scopus.com/inward/record.uri?eid=2-s2.0-85174744697&doi=10.1016

\bibitem {art_482}
\btxnamefont {Z.~\btxlastnamefont {Sun}}, \btxnamefont {Y.~\btxlastnamefont
  {Li}}, \btxnamefont {Q.~\btxlastnamefont {He}}, \btxnamefont
  {H.~\btxlastnamefont {Xu}}, \btxnamefont {W.~\btxlastnamefont
  {Wang}}\btxandcomma {} \btxandshort {.}\ \btxnamefont {X.~\btxlastnamefont
  {Liu}}\btxauthorcolon\ \btxjtitlefont {\btxifchangecase {Causality enhanced
  global-local graph neural network for bioprocess factor
  forecasting}{Causality enhanced global-local graph neural network for
  bioprocess factor forecasting}}.
\newblock \btxjournalfont {IEEE Trans. Industr. Inform.}, 20(10):12428--12438,
  \btxprintmonthyear{.}{10}{2024}{short}.

\bibitem {DeepSleepNet}
\btxnamefont {A.~\btxlastnamefont {Supratak}}, \btxnamefont
  {H.~\btxlastnamefont {Dong}}, \btxnamefont {C.~\btxlastnamefont
  {Wu}}\btxandcomma {} \btxandshort {.}\ \btxnamefont {Y.~\btxlastnamefont
  {Guo}}\btxauthorcolon\ \btxjtitlefont {\btxifchangecase {Deepsleepnet: A
  model for automatic sleep stage scoring based on raw single-channel
  eeg}{DeepSleepNet: A Model for Automatic Sleep Stage Scoring Based on Raw
  Single-Channel EEG}}.
\newblock \btxjournalfont {IEEE Transactions on Neural Systems and
  Rehabilitation Engineering}, 25(11):1998--2008, 2017.

\bibitem {Seq2seq}
\btxnamefont {I.~\btxlastnamefont {Sutskever}}, \btxnamefont
  {O.~\btxlastnamefont {Vinyals}}\btxandcomma {} \btxandshort {.}\ \btxnamefont
  {Q.\btxfnamespaceshort V. \btxlastnamefont {Le}}\btxauthorcolon\
  \btxtitlefont {\btxifchangecase {Sequence to sequence learning with neural
  networks}{Sequence to sequence learning with neural networks}}.
\newblock \Btxinshort {.}\ \btxtitlefont {Proceedings of the 27th International
  Conference on Neural Information Processing Systems - Volume 2}, NIPS'14,
  \btxpageshort {.}\ 3104–3112, Cambridge, MA, USA, 2014. \btxpublisherfont
  {MIT Press}.

\bibitem {art_359}
\btxnamefont {M.\btxfnamespaceshort L. \btxlastnamefont {Taccari}},
  \btxnamefont {H.~\btxlastnamefont {Wang}}, \btxnamefont {J.~\btxlastnamefont
  {Nuttall}}, \btxnamefont {X.~\btxlastnamefont {Chen}}\btxandcomma {}
  \btxandshort {.}\ \btxnamefont {P.\btxfnamespaceshort K. \btxlastnamefont
  {Jimack}}\btxauthorcolon\ \btxjtitlefont {\btxifchangecase {Spatial-temporal
  graph neural networks for groundwater data}{Spatial-temporal graph neural
  networks for groundwater data}}.
\newblock \btxjournalfont {Sci. Rep.}, 14(1):24564,
  \btxprintmonthyear{.}{10}{2024}{short}.

\bibitem {tan2024language}
\btxnamefont {M.~\btxlastnamefont {Tan}}, \btxnamefont {M.\btxfnamespaceshort
  A. \btxlastnamefont {Merrill}}, \btxnamefont {V.~\btxlastnamefont {Gupta}},
  \btxnamefont {T.~\btxlastnamefont {Althoff}}\btxandcomma {} \btxandshort {.}\
  \btxnamefont {T.~\btxlastnamefont {Hartvigsen}}\btxauthorcolon\ \btxtitlefont
  {\btxifchangecase {Are language models actually useful for time series
  forecasting?}{Are Language Models Actually Useful for Time Series
  Forecasting?}}
\newblock \Btxinshort {.}\ \btxtitlefont {Neural Information Processing Systems
  (NeurIPS)}, 2024.

\bibitem {art_709}
\btxnamefont {J.~\btxlastnamefont {Tang}}, \btxnamefont {L.~\btxlastnamefont
  {Xia}}, \btxnamefont {J.~\btxlastnamefont {Hu}}\btxandcomma {} \btxandshort
  {.}\ \btxnamefont {C.~\btxlastnamefont {Huang}}\btxauthorcolon\ \btxtitlefont
  {\btxifchangecase {{Spatio-Temporal} meta contrastive
  learning}{{Spatio-Temporal} Meta Contrastive Learning}}.
\newblock \Btxinshort {.}\ \btxtitlefont {Proceedings of the 32nd {ACM}
  International Conference on Information and Knowledge Management}, New York,
  NY, USA, \btxprintmonthyear{.}{10}{2023}{short}. \btxpublisherfont {ACM}.

\bibitem {art_700}
\btxnamefont {J.~\btxlastnamefont {Tang}}, \btxnamefont {L.~\btxlastnamefont
  {Xia}}\btxandcomma {} \btxandshort {.}\ \btxnamefont {C.~\btxlastnamefont
  {Huang}}\btxauthorcolon\ \btxtitlefont {\btxifchangecase {Explainable
  spatio-temporal graph neural networks}{Explainable spatio-temporal graph
  neural networks}}.
\newblock \Btxinshort {.}\ \btxtitlefont {Proceedings of the 32nd {ACM}
  International Conference on Information and Knowledge Management}, New York,
  NY, USA, \btxprintmonthyear{.}{10}{2023}{short}. \btxpublisherfont {ACM}.

\bibitem {art_725}
\btxnamefont {Y.~\btxlastnamefont {Tang}}, \btxnamefont {A.~\btxlastnamefont
  {Qu}}, \btxnamefont {A.\btxfnamespaceshort H.\btxfnamespaceshort F.
  \btxlastnamefont {Chow}}, \btxnamefont {W.\btxfnamespaceshort
  H.\btxfnamespaceshort K. \btxlastnamefont {Lam}}, \btxnamefont
  {S.\btxfnamespaceshort C. \btxlastnamefont {Wong}}\btxandcomma {}
  \btxandshort {.}\ \btxnamefont {W.~\btxlastnamefont {Ma}}\btxauthorcolon\
  \btxtitlefont {\btxifchangecase {Domain adversarial spatial-temporal
  network}{Domain adversarial spatial-temporal network}}.
\newblock \Btxinshort {.}\ \btxtitlefont {Proceedings of the 31st {ACM}
  International Conference on Information \& Knowledge Management}, New York,
  NY, USA, \btxprintmonthyear{.}{10}{2022}{short}. \btxpublisherfont {ACM}.

\bibitem {art_62}
\btxnamefont {H.~\btxlastnamefont {Tian}}, \btxnamefont {X.~\btxlastnamefont
  {Zhang}}, \btxnamefont {X.~\btxlastnamefont {Zheng}}\btxandcomma {}
  \btxandshort {.}\ \btxnamefont {D.\btxfnamespaceshort D. \btxlastnamefont
  {Zeng}}\btxauthorcolon\ \btxjtitlefont {\btxifchangecase {Learning dynamic
  dependencies with graph evolution recurrent unit for stock
  predictions}{Learning Dynamic Dependencies With Graph Evolution Recurrent
  Unit for Stock Predictions}}.
\newblock \btxjournalfont {IEEE Transactions on Systems, Man, and Cybernetics:
  Systems}, 53(11):6705 – 6717, 2023.
\newblock {\latintext
  \btxurlfont{https://www.scopus.com/inward/record.uri?eid=2-s2.0-85164722955&doi=10.1109

\bibitem {TranAD}
\btxnamefont {S.~\btxlastnamefont {Tuli}}, \btxnamefont {G.~\btxlastnamefont
  {Casale}}\btxandcomma {} \btxandshort {.}\ \btxnamefont
  {N.\btxfnamespaceshort R. \btxlastnamefont {Jennings}}\btxauthorcolon\
  \btxjtitlefont {\btxifchangecase {Tranad: deep transformer networks for
  anomaly detection in multivariate time series data}{TranAD: deep transformer
  networks for anomaly detection in multivariate time series data}}.
\newblock \btxjournalfont {Proc. VLDB Endow.}, 15(6):1201–1214,
  \btxprintmonthyear{.}{2}{2022}{short}\ifbtxprintISSN {,
  \mbox{\btxISSN~\btxISSNfont {2150-8097}}}.
\newblock {\latintext \btxurlfont{https://doi.org/10.14778/3514061.3514067}}.

\btxselectlanguage {english}
\bibitem {VOSviewer}
\btxnamefont {N.\btxfnamespaceshort J. \btxlastnamefont {Van~Eck}} \btxandshort
  {.}\ \btxnamefont {L.~\btxlastnamefont {Waltman}}\btxauthorcolon\
  \btxjtitlefont {\btxifchangecase {Software survey: {VOSviewer}, a computer
  program for bibliometric mapping}{Software survey: {VOSviewer}, a computer
  program for bibliometric mapping}}.
\newblock \btxjournalfont {Scientometrics}, 84(2):523--538,
  \btxprintmonthyear{.}{8}{2010}{short}.

\expandafter\btxselectlanguage\expandafter {\btxfallbacklanguage}
\bibitem {art_53}
\btxnamefont {J.~\btxlastnamefont {Van~Gompel}}, \btxnamefont
  {D.~\btxlastnamefont {Spina}}\btxandcomma {} \btxandshort {.}\ \btxnamefont
  {C.~\btxlastnamefont {Develder}}\btxauthorcolon\ \btxjtitlefont
  {\btxifchangecase {Cost-effective fault diagnosis of nearby photovoltaic
  systems using graph neural networks}{Cost-effective fault diagnosis of nearby
  photovoltaic systems using graph neural networks}}.
\newblock \btxjournalfont {Energy}, 266, 2023.
\newblock {\latintext
  \btxurlfont{https://www.scopus.com/inward/record.uri?eid=2-s2.0-85145678193&doi=10.1016

\bibitem {attention}
\btxnamefont {A.~\btxlastnamefont {Vaswani}}, \btxnamefont {N.~\btxlastnamefont
  {Shazeer}}, \btxnamefont {N.~\btxlastnamefont {Parmar}}, \btxnamefont
  {J.~\btxlastnamefont {Uszkoreit}}, \btxnamefont {L.~\btxlastnamefont
  {Jones}}, \btxnamefont {A.\btxfnamespaceshort N. \btxlastnamefont {Gomez}},
  \btxnamefont {L.~\btxlastnamefont {Kaiser}}\btxandcomma {} \btxandshort {.}\
  \btxnamefont {I.~\btxlastnamefont {Polosukhin}}\btxauthorcolon\ \btxtitlefont
  {\btxifchangecase {Attention is all you need}{Attention is all you need}}.
\newblock \Btxinshort {.}\ \btxtitlefont {Proceedings of the 31st International
  Conference on Neural Information Processing Systems}, NIPS'17, \btxpageshort
  {.}\ 6000–6010, Red Hook, NY, USA, 2017. \btxpublisherfont {Curran
  Associates Inc.}\ifbtxprintISBN {, \mbox{\btxISBN~\btxISBNfont
  {9781510860964}}}.

\bibitem {GAT}
\btxnamefont {P.~\btxlastnamefont {Veli{\v{c}}kovi{\'{c}}}}, \btxnamefont
  {G.~\btxlastnamefont {Cucurull}}, \btxnamefont {A.~\btxlastnamefont
  {Casanova}}, \btxnamefont {A.~\btxlastnamefont {Romero}}, \btxnamefont
  {P.~\btxlastnamefont {Li{\`{o}}}}\btxandcomma {} \btxandshort {.}\
  \btxnamefont {Y.~\btxlastnamefont {Bengio}}\btxauthorcolon\ \btxjtitlefont
  {\btxifchangecase {{Graph Attention Networks}}{{Graph Attention Networks}}}.
\newblock \btxjournalfont {International Conference on Learning
  Representations}, 2018.
\newblock {\latintext \btxurlfont{https://openreview.net/forum?id=rJXMpikCZ}},
  accepted as poster.

\bibitem {art_349}
\btxnamefont {A.~\btxlastnamefont {Verdone}}, \btxnamefont {S.~\btxlastnamefont
  {Scardapane}}\btxandcomma {} \btxandshort {.}\ \btxnamefont
  {M.~\btxlastnamefont {Panella}}\btxauthorcolon\ \btxjtitlefont
  {\btxifchangecase {Explainable {Spatio-Temporal} graph neural networks for
  multi-site photovoltaic energy production}{Explainable {Spatio-Temporal}
  Graph Neural Networks for multi-site photovoltaic energy production}}.
\newblock \btxjournalfont {Appl. Energy}, 353(122151):122151,
  \btxprintmonthyear{.}{1}{2024}{short}.

\bibitem {art_442}
\btxnamefont {B.~\btxlastnamefont {Wang}}, \btxnamefont {F.~\btxlastnamefont
  {Gao}}, \btxnamefont {L.~\btxlastnamefont {Tong}}, \btxnamefont
  {Q.~\btxlastnamefont {Zhang}}\btxandcomma {} \btxandshort {.}\ \btxnamefont
  {S.~\btxlastnamefont {Zhu}}\btxauthorcolon\ \btxjtitlefont {\btxifchangecase
  {Channel attention-based spatial-temporal graph neural networks for traffic
  prediction}{Channel attention-based spatial-temporal graph neural networks
  for traffic prediction}}.
\newblock \btxjournalfont {Data Technol. Appl.}, 58(1):81--94,
  \btxprintmonthyear{.}{1}{2024}{short}.

\bibitem {art_486}
\btxnamefont {B.~\btxlastnamefont {Wang}}, \btxnamefont {Y.~\btxlastnamefont
  {Xu}}, \btxnamefont {M.~\btxlastnamefont {Wang}}\btxandcomma {} \btxandshort
  {.}\ \btxnamefont {Y.~\btxlastnamefont {Li}}\btxauthorcolon\ \btxjtitlefont
  {\btxifchangecase {Gear fault diagnosis method based on the optimized graph
  neural networks}{Gear fault diagnosis method based on the optimized graph
  neural networks}}.
\newblock \btxjournalfont {IEEE Trans. Instrum. Meas.}, 73:1--11, 2024.

\bibitem {art_12}
\btxnamefont {D.~\btxlastnamefont {Wang}}, \btxnamefont {M.~\btxlastnamefont
  {Lin}}, \btxnamefont {X.~\btxlastnamefont {Zhang}}, \btxnamefont
  {Y.~\btxlastnamefont {Huang}}\btxandcomma {} \btxandshort {.}\ \btxnamefont
  {Y.~\btxlastnamefont {Zhu}}\btxauthorcolon\ \btxjtitlefont {\btxifchangecase
  {Automatic modulation classification based on cnn-transformer graph neural
  network}{Automatic Modulation Classification Based on CNN-Transformer Graph
  Neural Network}}.
\newblock \btxjournalfont {Sensors}, 23(16), 2023.
\newblock {\latintext
  \btxurlfont{https://www.scopus.com/inward/record.uri?eid=2-s2.0-85168726736&doi=10.3390

\bibitem {art_491}
\btxnamefont {H.~\btxlastnamefont {Wang}}, \btxnamefont {X.~\btxlastnamefont
  {Dai}}, \btxnamefont {L.~\btxlastnamefont {Shi}}, \btxnamefont
  {M.~\btxlastnamefont {Li}}, \btxnamefont {Z.~\btxlastnamefont {Liu}},
  \btxnamefont {R.~\btxlastnamefont {Wang}}\btxandcomma {} \btxandshort {.}\
  \btxnamefont {X.~\btxlastnamefont {Xia}}\btxauthorcolon\ \btxjtitlefont
  {\btxifchangecase {Data-augmentation based {CBAM-ResNet-GCN} method for
  unbalance fault diagnosis of rotating machinery}{Data-augmentation based
  {CBAM-ResNet-GCN} method for unbalance fault diagnosis of rotating
  machinery}}.
\newblock \btxjournalfont {IEEE Access}, 12:34785--34799, 2024.

\bibitem {art_154}
\btxnamefont {H.~\btxlastnamefont {Wang}}, \btxnamefont {Z.~\btxlastnamefont
  {Zhang}}, \btxnamefont {X.~\btxlastnamefont {Li}}, \btxnamefont
  {X.~\btxlastnamefont {Deng}}\btxandcomma {} \btxandshort {.}\ \btxnamefont
  {W.~\btxlastnamefont {Jiang}}\btxauthorcolon\ \btxjtitlefont
  {\btxifchangecase {Comprehensive dynamic structure graph neural network for
  aero-engine remaining useful life prediction}{Comprehensive Dynamic Structure
  Graph Neural Network for Aero-Engine Remaining Useful Life Prediction}}.
\newblock \btxjournalfont {IEEE Transactions on Instrumentation and
  Measurement}, 72, 2023.
\newblock {\latintext
  \btxurlfont{https://www.scopus.com/inward/record.uri?eid=2-s2.0-85174803368&doi=10.1109

\bibitem {art_14}
\btxnamefont {J.~\btxlastnamefont {Wang}}, \btxnamefont {R.~\btxlastnamefont
  {Gao}}, \btxnamefont {H.~\btxlastnamefont {Zheng}}, \btxnamefont
  {H.~\btxlastnamefont {Zhu}}\btxandcomma {} \btxandshort {.}\ \btxnamefont
  {C.\btxfnamespaceshort J.\btxfnamespaceshort R. \btxlastnamefont
  {Shi}}\btxauthorcolon\ \btxjtitlefont {\btxifchangecase {Ssgcnet: A sparse
  spectra graph convolutional network for epileptic eeg signal
  classification}{SSGCNet: A Sparse Spectra Graph Convolutional Network for
  Epileptic EEG Signal Classification}}.
\newblock \btxjournalfont {IEEE Transactions on Neural Networks and Learning
  Systems}, \btxpagesshort {.}\ 1--15, 2023.

\bibitem {art_661}
\btxnamefont {J.~\btxlastnamefont {Wang}}, \btxnamefont {X.~\btxlastnamefont
  {Ning}}, \btxnamefont {W.~\btxlastnamefont {Shi}}\btxandcomma {} \btxandshort
  {.}\ \btxnamefont {Y.~\btxlastnamefont {Lin}}\btxauthorcolon\ \btxtitlefont
  {\btxifchangecase {A bayesian graph neural network for {EEG} classification
  --- a win-win on performance and interpretability}{A Bayesian graph neural
  network for {EEG} classification --- A win-win on performance and
  interpretability}}.
\newblock \Btxinshort {.}\ \btxtitlefont {2023 {IEEE} 39th International
  Conference on Data Engineering ({ICDE})}, \btxpagesshort {.}\ 2126--2139.
  \btxpublisherfont {IEEE}, \btxprintmonthyear{.}{4}{2023}{short}.

\bibitem {art_421}
\btxnamefont {J.~\btxlastnamefont {Wang}}, \btxnamefont {Y.~\btxlastnamefont
  {Zhang}}, \btxnamefont {Y.~\btxlastnamefont {Hu}}\btxandcomma {} \btxandshort
  {.}\ \btxnamefont {B.~\btxlastnamefont {Yin}}\btxauthorcolon\ \btxjtitlefont
  {\btxifchangecase {Metro flow prediction with hierarchical hypergraph
  attention networks}{Metro flow prediction with hierarchical hypergraph
  attention networks}}.
\newblock \btxjournalfont {IEEE Trans. Artif. Intell.}, 5(6):3012--3021,
  \btxprintmonthyear{.}{6}{2024}{short}.

\bibitem {GGCN}
\btxnamefont {L.~\btxlastnamefont {Wang}}, \btxnamefont {H.~\btxlastnamefont
  {Cao}}, \btxnamefont {H.~\btxlastnamefont {Xu}}\btxandcomma {} \btxandshort
  {.}\ \btxnamefont {H.~\btxlastnamefont {Liu}}\btxauthorcolon\ \btxjtitlefont
  {\btxifchangecase {A gated graph convolutional network with multi-sensor
  signals for remaining useful life prediction}{A gated graph convolutional
  network with multi-sensor signals for remaining useful life prediction}}.
\newblock \btxjournalfont {Knowledge-Based Systems}, 252:109340,
  2022\ifbtxprintISSN {, \mbox{\btxISSN~\btxISSNfont {0950-7051}}}.
\newblock {\latintext
  \btxurlfont{https://www.sciencedirect.com/science/article/pii/S0950705122006724}}.

\bibitem {art_722}
\btxnamefont {M.~\btxlastnamefont {Wang}}, \btxnamefont {H.~\btxlastnamefont
  {Yan}}, \btxnamefont {H.~\btxlastnamefont {Wang}}, \btxnamefont
  {Y.~\btxlastnamefont {Li}}\btxandcomma {} \btxandshort {.}\ \btxnamefont
  {D.~\btxlastnamefont {Jin}}\btxauthorcolon\ \btxtitlefont {\btxifchangecase
  {Contagion process guided cross-scale spatio-temporal graph neural network
  for traffic congestion prediction}{Contagion process guided cross-scale
  spatio-temporal graph neural network for traffic congestion prediction}}.
\newblock \Btxinshort {.}\ \btxtitlefont {Proceedings of the 31st {ACM}
  International Conference on Advances in Geographic Information Systems}, New
  York, NY, USA, \btxprintmonthyear{.}{11}{2023}{short}. \btxpublisherfont
  {ACM}.

\bibitem {art_451}
\btxnamefont {P.~\btxlastnamefont {Wang}}, \btxnamefont {X.~\btxlastnamefont
  {Luo}}, \btxnamefont {W.~\btxlastnamefont {Tai}}, \btxnamefont
  {K.~\btxlastnamefont {Zhang}}, \btxnamefont {G.~\btxlastnamefont
  {Trajcevsky}}\btxandcomma {} \btxandshort {.}\ \btxnamefont
  {F.~\btxlastnamefont {Zhou}}\btxauthorcolon\ \btxjtitlefont {\btxifchangecase
  {Score-based graph learning for urban flow prediction}{Score-based graph
  learning for urban flow prediction}}.
\newblock \btxjournalfont {ACM Trans. Intell. Syst. Technol.}, 15(3):1--25,
  \btxprintmonthyear{.}{6}{2024}{short}.

\bibitem {art_474}
\btxnamefont {S.~\btxlastnamefont {Wang}}, \btxnamefont {B.~\btxlastnamefont
  {Jing}}, \btxnamefont {J.~\btxlastnamefont {Pan}}, \btxnamefont
  {X.~\btxlastnamefont {Meng}}, \btxnamefont {Y.~\btxlastnamefont
  {Huang}}\btxandcomma {} \btxandshort {.}\ \btxnamefont {X.~\btxlastnamefont
  {Jiao}}\btxauthorcolon\ \btxjtitlefont {\btxifchangecase {Coupling fault
  diagnosis based on dynamic vertex interpretable graph neural
  network}{Coupling fault diagnosis based on dynamic vertex interpretable graph
  neural network}}.
\newblock \btxjournalfont {Sensors (Basel)}, 24(13):4356,
  \btxprintmonthyear{.}{7}{2024}{short}.

\bibitem {art_400}
\btxnamefont {S.~\btxlastnamefont {Wang}}, \btxnamefont {Y.~\btxlastnamefont
  {Zhang}}, \btxnamefont {X.~\btxlastnamefont {Lin}}, \btxnamefont
  {Y.~\btxlastnamefont {Hu}}, \btxnamefont {Q.~\btxlastnamefont
  {Huang}}\btxandcomma {} \btxandshort {.}\ \btxnamefont {B.~\btxlastnamefont
  {Yin}}\btxauthorcolon\ \btxjtitlefont {\btxifchangecase {Dynamic hypergraph
  structure learning for multivariate time series forecasting}{Dynamic
  hypergraph structure learning for multivariate time series forecasting}}.
\newblock \btxjournalfont {IEEE Trans. Big Data}, 10(4):556--567,
  \btxprintmonthyear{.}{8}{2024}{short}.

\bibitem {art_63}
\btxnamefont {T.~\btxlastnamefont {Wang}}, \btxnamefont {J.~\btxlastnamefont
  {Guo}}, \btxnamefont {Y.~\btxlastnamefont {Shan}}, \btxnamefont
  {Y.~\btxlastnamefont {Zhang}}, \btxnamefont {B.~\btxlastnamefont
  {Peng}}\btxandcomma {} \btxandshort {.}\ \btxnamefont {Z.~\btxlastnamefont
  {Wu}}\btxauthorcolon\ \btxjtitlefont {\btxifchangecase {A knowledge
  graph–gcn–community detection integrated model for large-scale stock
  price prediction}{A knowledge graph–GCN–community detection integrated
  model for large-scale stock price prediction}}.
\newblock \btxjournalfont {Applied Soft Computing}, 145, 2023.
\newblock {\latintext
  \btxurlfont{https://www.scopus.com/inward/record.uri?eid=2-s2.0-85165231582&doi=10.1016

\bibitem {art_39}
\btxnamefont {T.~\btxlastnamefont {Wang}}, \btxnamefont {Z.~\btxlastnamefont
  {Li}}, \btxnamefont {X.~\btxlastnamefont {Geng}}, \btxnamefont
  {B.~\btxlastnamefont {Jin}}\btxandcomma {} \btxandshort {.}\ \btxnamefont
  {L.~\btxlastnamefont {Xu}}\btxauthorcolon\ \btxjtitlefont {\btxifchangecase
  {Time series prediction of sea surface temperature based on an adaptive graph
  learning neural model}{Time Series Prediction of Sea Surface Temperature
  Based on an Adaptive Graph Learning Neural Model}}.
\newblock \btxjournalfont {Future Internet}, 14(6), 2022.
\newblock {\latintext
  \btxurlfont{https://www.scopus.com/inward/record.uri?eid=2-s2.0-85131754722&doi=10.3390

\bibitem {art_669}
\btxnamefont {X.~\btxlastnamefont {Wang}}, \btxnamefont {Y.~\btxlastnamefont
  {Ma}}, \btxnamefont {Y.~\btxlastnamefont {Wang}}, \btxnamefont
  {W.~\btxlastnamefont {Jin}}, \btxnamefont {X.~\btxlastnamefont {Wang}},
  \btxnamefont {J.~\btxlastnamefont {Tang}}, \btxnamefont {C.~\btxlastnamefont
  {Jia}}\btxandcomma {} \btxandshort {.}\ \btxnamefont {J.~\btxlastnamefont
  {Yu}}\btxauthorcolon\ \btxtitlefont {\btxifchangecase {Traffic flow
  prediction via spatial temporal graph neural network}{Traffic Flow Prediction
  via Spatial Temporal Graph Neural Network}}.
\newblock \Btxinshort {.}\ \btxtitlefont {Proceedings of The Web Conference
  2020}, WWW '20, \btxpageshort {.}\ 1082–1092, New York, NY, USA, 2020.
  \btxpublisherfont {Association for Computing Machinery}\ifbtxprintISBN {,
  \mbox{\btxISBN~\btxISBNfont {9781450370233}}}.
\newblock {\latintext \btxurlfont{https://doi.org/10.1145/3366423.3380186}}.

\bibitem {art_460}
\btxnamefont {X.~\btxlastnamefont {Wang}}, \btxnamefont {X.~\btxlastnamefont
  {Wang}}, \btxnamefont {M.~\btxlastnamefont {He}}, \btxnamefont
  {M.~\btxlastnamefont {Zhang}}\btxandcomma {} \btxandshort {.}\ \btxnamefont
  {Z.~\btxlastnamefont {Lu}}\btxauthorcolon\ \btxjtitlefont {\btxifchangecase
  {Spatial-temporal graph model based on attention mechanism for anomalous
  {IoT} intrusion detection}{Spatial-temporal graph model based on attention
  mechanism for anomalous {IoT} intrusion detection}}.
\newblock \btxjournalfont {IEEE Trans. Industr. Inform.}, 20(3):3497--3509,
  \btxprintmonthyear{.}{3}{2024}{short}.

\bibitem {art_82}
\btxnamefont {X.~\btxlastnamefont {Wang}}, \btxnamefont {Y.~\btxlastnamefont
  {Wang}}, \btxnamefont {J.~\btxlastnamefont {Peng}}\btxandcomma {}
  \btxandshort {.}\ \btxnamefont {Z.~\btxlastnamefont {Zhang}}\btxauthorcolon\
  \btxjtitlefont {\btxifchangecase {Multivariate long sequence time-series
  forecasting using dynamic graph learning}{Multivariate long sequence
  time-series forecasting using dynamic graph learning}}.
\newblock \btxjournalfont {Journal of Ambient Intelligence and Humanized
  Computing}, 14(6):7679 – 7693, 2023.
\newblock {\latintext
  \btxurlfont{https://www.scopus.com/inward/record.uri?eid=2-s2.0-85149947973&doi=10.1007

\bibitem {art_387}
\btxnamefont {X.~\btxlastnamefont {Wang}}, \btxnamefont {Q.~\btxlastnamefont
  {Xing}}, \btxnamefont {H.~\btxlastnamefont {Xiao}}\btxandcomma {}
  \btxandshort {.}\ \btxnamefont {M.~\btxlastnamefont {Ye}}\btxauthorcolon\
  \btxjtitlefont {\btxifchangecase {Contrastive learning enhanced by graph
  neural networks for universal multivariate time series
  representation}{Contrastive learning enhanced by graph neural networks for
  Universal Multivariate Time Series Representation}}.
\newblock \btxjournalfont {Inf. Syst.}, 125(102429):102429,
  \btxprintmonthyear{.}{11}{2024}{short}.

\bibitem {art_87}
\btxnamefont {Y.~\btxlastnamefont {Wang}}, \btxnamefont {Z.~\btxlastnamefont
  {Duan}}, \btxnamefont {Y.~\btxlastnamefont {Huang}}, \btxnamefont
  {H.~\btxlastnamefont {Xu}}, \btxnamefont {J.~\btxlastnamefont
  {Feng}}\btxandcomma {} \btxandshort {.}\ \btxnamefont {A.~\btxlastnamefont
  {Ren}}\btxauthorcolon\ \btxjtitlefont {\btxifchangecase {Mthetgnn: A
  heterogeneous graph embedding framework for multivariate time series
  forecasting}{MTHetGNN: A heterogeneous graph embedding framework for
  multivariate time series forecasting}}.
\newblock \btxjournalfont {Pattern Recognition Letters}, 153:151 – 158, 2022.
\newblock {\latintext
  \btxurlfont{https://www.scopus.com/inward/record.uri?eid=2-s2.0-85122529802&doi=10.1016

\bibitem {art_17}
\btxnamefont {Y.~\btxlastnamefont {Wang}}, \btxnamefont {L.~\btxlastnamefont
  {Rui}}, \btxnamefont {J.~\btxlastnamefont {Ma}}\btxandcomma {} \btxandshort
  {.}\ \btxnamefont {Q.~\btxlastnamefont {jin}}\btxauthorcolon\ \btxjtitlefont
  {\btxifchangecase {A short-term residential load forecasting scheme based on
  the multiple correlation-temporal graph neural networks}{A short-term
  residential load forecasting scheme based on the multiple
  correlation-temporal graph neural networks}}.
\newblock \btxjournalfont {Applied Soft Computing}, 146, 2023.
\newblock {\latintext
  \btxurlfont{https://www.scopus.com/inward/record.uri?eid=2-s2.0-85166632235&doi=10.1016

\bibitem {art_458}
\btxnamefont {Y.~\btxlastnamefont {Wang}}, \btxnamefont {X.~\btxlastnamefont
  {Wang}}, \btxnamefont {J.~\btxlastnamefont {Zhou}}, \btxnamefont
  {C.~\btxlastnamefont {Yang}}\btxandcomma {} \btxandshort {.}\ \btxnamefont
  {Y.~\btxlastnamefont {Yang}}\btxauthorcolon\ \btxjtitlefont {\btxifchangecase
  {Long sequence multivariate time-series forecasting for industrial processes
  using {SASGNN}}{Long sequence multivariate time-series forecasting for
  industrial processes using {SASGNN}}}.
\newblock \btxjournalfont {IEEE Trans. Industr. Inform.}, \btxpagesshort {.}\
  1--11, 2024.

\bibitem {art_152}
\btxnamefont {Y.~\btxlastnamefont {Wang}}, \btxnamefont {M.~\btxlastnamefont
  {Wu}}, \btxnamefont {R.~\btxlastnamefont {Jin}}, \btxnamefont
  {X.~\btxlastnamefont {Li}}, \btxnamefont {L.~\btxlastnamefont
  {Xie}}\btxandcomma {} \btxandshort {.}\ \btxnamefont {Z.~\btxlastnamefont
  {Chen}}\btxauthorcolon\ \btxjtitlefont {\btxifchangecase {Local-global
  correlation fusion-based graph neural network for remaining useful life
  prediction}{Local-Global Correlation Fusion-Based Graph Neural Network for
  Remaining Useful Life Prediction}}.
\newblock \btxjournalfont {IEEE Transactions on Neural Networks and Learning
  Systems}, \btxpageshort {.}\ 1–14, 2023.
\newblock {\latintext
  \btxurlfont{https://www.scopus.com/inward/record.uri?eid=2-s2.0-85178023045&doi=10.1109

\bibitem {art_403}
\btxnamefont {Y.~\btxlastnamefont {Wang}}, \btxnamefont {M.~\btxlastnamefont
  {Wu}}, \btxnamefont {X.~\btxlastnamefont {Li}}, \btxnamefont
  {L.~\btxlastnamefont {Xie}}\btxandcomma {} \btxandshort {.}\ \btxnamefont
  {Z.~\btxlastnamefont {Chen}}\btxauthorcolon\ \btxjtitlefont {\btxifchangecase
  {Multivariate time-series representation learning via hierarchical
  correlation pooling boosted graph neural network}{Multivariate time-series
  representation learning via hierarchical correlation pooling boosted graph
  neural network}}.
\newblock \btxjournalfont {IEEE Trans. Artif. Intell.}, 5(1):321--333,
  \btxprintmonthyear{.}{1}{2024}{short}.

\bibitem {art_102}
\btxnamefont {Y.~\btxlastnamefont {Wang}}, \btxnamefont {H.~\btxlastnamefont
  {Yin}}, \btxnamefont {T.~\btxlastnamefont {Chen}}, \btxnamefont
  {C.~\btxlastnamefont {Liu}}, \btxnamefont {B.~\btxlastnamefont {Wang}},
  \btxnamefont {T.~\btxlastnamefont {Wo}}\btxandcomma {} \btxandshort {.}\
  \btxnamefont {J.~\btxlastnamefont {Xu}}\btxauthorcolon\ \btxjtitlefont
  {\btxifchangecase {Passenger mobility prediction via representation learning
  for dynamic directed and weighted graphs}{Passenger Mobility Prediction via
  Representation Learning for Dynamic Directed and Weighted Graphs}}.
\newblock \btxjournalfont {ACM Trans. Intell. Syst. Technol.}, 13(1),
  \btxprintmonthyear{.}{nov}{2021}{short}\ifbtxprintISSN {,
  \mbox{\btxISSN~\btxISSNfont {2157-6904}}}.
\newblock {\latintext \btxurlfont{https://doi.org/10.1145/3446344}}.

\bibitem {art_370}
\btxnamefont {Z.~\btxlastnamefont {Wang}}, \btxnamefont {Z.~\btxlastnamefont
  {Chen}}, \btxnamefont {Z.~\btxlastnamefont {Shi}}\btxandcomma {} \btxandshort
  {.}\ \btxnamefont {J.~\btxlastnamefont {Gao}}\btxauthorcolon\ \btxjtitlefont
  {\btxifchangecase {Predicting global average temperature time series using an
  entire graph node training approach}{Predicting global average temperature
  time series using an entire graph node training approach}}.
\newblock \btxjournalfont {IEEE Trans. Geosci. Remote Sens.}, 62:1--14, 2024.

\bibitem {art_138}
\btxnamefont {Z.~\btxlastnamefont {Wang}}, \btxnamefont {J.~\btxlastnamefont
  {Hu}}, \btxnamefont {G.~\btxlastnamefont {Min}}, \btxnamefont
  {Z.~\btxlastnamefont {Zhao}}, \btxnamefont {Z.~\btxlastnamefont
  {Chang}}\btxandcomma {} \btxandshort {.}\ \btxnamefont {Z.~\btxlastnamefont
  {Wang}}\btxauthorcolon\ \btxjtitlefont {\btxifchangecase {Spatial-temporal
  cellular traffic prediction for 5g and beyond: A graph neural networks-based
  approach}{Spatial-Temporal Cellular Traffic Prediction for 5G and Beyond: A
  Graph Neural Networks-Based Approach}}.
\newblock \btxjournalfont {IEEE Transactions on Industrial Informatics},
  19(4):5722 – 5731, 2023.
\newblock {\latintext
  \btxurlfont{https://www.scopus.com/inward/record.uri?eid=2-s2.0-85133780529&doi=10.1109

\bibitem {art_15}
\btxnamefont {Z.~\btxlastnamefont {Wang}}, \btxnamefont {X.~\btxlastnamefont
  {Liu}}, \btxnamefont {Y.~\btxlastnamefont {Huang}}, \btxnamefont
  {P.~\btxlastnamefont {Zhang}}\btxandcomma {} \btxandshort {.}\ \btxnamefont
  {Y.~\btxlastnamefont {Fu}}\btxauthorcolon\ \btxjtitlefont {\btxifchangecase
  {A multivariate time series graph neural network for district heat load
  forecasting}{A multivariate time series graph neural network for district
  heat load forecasting}}.
\newblock \btxjournalfont {Energy}, 278, 2023.
\newblock {\latintext
  \btxurlfont{https://www.scopus.com/inward/record.uri?eid=2-s2.0-85160794010&doi=10.1016

\bibitem {art_438}
\btxnamefont {L.~\btxlastnamefont {Wei}}, \btxnamefont {Z.~\btxlastnamefont
  {Yu}}, \btxnamefont {Z.~\btxlastnamefont {Jin}}, \btxnamefont
  {L.~\btxlastnamefont {Xie}}, \btxnamefont {J.~\btxlastnamefont {Huang}},
  \btxnamefont {D.~\btxlastnamefont {Cai}}, \btxnamefont {X.~\btxlastnamefont
  {He}}\btxandcomma {} \btxandshort {.}\ \btxnamefont {X.\btxfnamespaceshort S.
  \btxlastnamefont {Hua}}\btxauthorcolon\ \btxjtitlefont {\btxifchangecase
  {Dual graph for traffic forecasting}{Dual graph for traffic forecasting}}.
\newblock \btxjournalfont {IEEE Access}, \btxpagesshort {.}\ 1--1, 2024.

\bibitem {art_424}
\btxnamefont {S.~\btxlastnamefont {Wei}}, \btxnamefont {Y.~\btxlastnamefont
  {Yang}}, \btxnamefont {D.~\btxlastnamefont {Liu}}, \btxnamefont
  {K.~\btxlastnamefont {Deng}}\btxandcomma {} \btxandshort {.}\ \btxnamefont
  {C.~\btxlastnamefont {Wang}}\btxauthorcolon\ \btxjtitlefont {\btxifchangecase
  {Transformer-based spatiotemporal graph diffusion convolution network for
  traffic flow forecasting}{Transformer-based Spatiotemporal Graph Diffusion
  Convolution Network for Traffic Flow Forecasting}}.
\newblock \btxjournalfont {Electronics (Basel)}, 13(16):3151,
  \btxprintmonthyear{.}{8}{2024}{short}.

\bibitem {art_158}
\btxnamefont {Y.~\btxlastnamefont {Wei}} \btxandshort {.}\ \btxnamefont
  {D.~\btxlastnamefont {Wu}}\btxauthorcolon\ \btxjtitlefont {\btxifchangecase
  {Prediction of state of health and remaining useful life of lithium-ion
  battery using graph convolutional network with dual attention
  mechanisms}{Prediction of state of health and remaining useful life of
  lithium-ion battery using graph convolutional network with dual attention
  mechanisms}}.
\newblock \btxjournalfont {Reliability Engineering and System Safety}, 230,
  2023.
\newblock {\latintext
  \btxurlfont{https://www.scopus.com/inward/record.uri?eid=2-s2.0-85141763841&doi=10.1016

\bibitem {art_698}
\btxnamefont {H.~\btxlastnamefont {Wen}}, \btxnamefont {Y.~\btxlastnamefont
  {Lin}}, \btxnamefont {Y.~\btxlastnamefont {Xia}}, \btxnamefont
  {H.~\btxlastnamefont {Wan}}, \btxnamefont {Q.~\btxlastnamefont {Wen}},
  \btxnamefont {R.~\btxlastnamefont {Zimmermann}}\btxandcomma {} \btxandshort
  {.}\ \btxnamefont {Y.~\btxlastnamefont {Liang}}\btxauthorcolon\ \btxtitlefont
  {\btxifchangecase {{DiffSTG}: Probabilistic spatio-temporal graph forecasting
  with denoising diffusion models}{{DiffSTG}: Probabilistic spatio-temporal
  graph forecasting with denoising diffusion models}}.
\newblock \Btxinshort {.}\ \btxtitlefont {Proceedings of the 31st {ACM}
  International Conference on Advances in Geographic Information Systems}, New
  York, NY, USA, \btxprintmonthyear{.}{11}{2023}{short}. \btxpublisherfont
  {ACM}.

\bibitem {art_140}
\btxnamefont {Z.~\btxlastnamefont {Wen}}, \btxnamefont {Y.~\btxlastnamefont
  {Li}}, \btxnamefont {H.~\btxlastnamefont {Wang}}\btxandcomma {} \btxandshort
  {.}\ \btxnamefont {Y.~\btxlastnamefont {Peng}}\btxauthorcolon\ \btxjtitlefont
  {\btxifchangecase {Data-driven spatiotemporal modeling for structural
  dynamics on irregular domains by stochastic dependency neural
  estimation}{Data-driven spatiotemporal modeling for structural dynamics on
  irregular domains by stochastic dependency neural estimation}}.
\newblock \btxjournalfont {Computer Methods in Applied Mechanics and
  Engineering}, 404, 2023.
\newblock {\latintext
  \btxurlfont{https://www.scopus.com/inward/record.uri?eid=2-s2.0-85144323571&doi=10.1016

\bibitem {art_740}
\btxnamefont {Y.~\btxlastnamefont {W{\"o}lker}}, \btxnamefont
  {C.~\btxlastnamefont {Beth}}, \btxnamefont {M.~\btxlastnamefont
  {Renz}}\btxandcomma {} \btxandshort {.}\ \btxnamefont {A.~\btxlastnamefont
  {Biastoch}}\btxauthorcolon\ \btxtitlefont {\btxifchangecase {{SUSTeR}: Sparse
  unstructured spatio temporal reconstruction on traffic prediction}{{SUSTeR}:
  Sparse unstructured spatio temporal reconstruction on traffic prediction}}.
\newblock \Btxinshort {.}\ \btxtitlefont {Proceedings of the 31st {ACM}
  International Conference on Advances in Geographic Information Systems}, New
  York, NY, USA, \btxprintmonthyear{.}{11}{2023}{short}. \btxpublisherfont
  {ACM}.

\bibitem {ETSformer}
\btxnamefont {G.~\btxlastnamefont {Woo}}, \btxnamefont {C.~\btxlastnamefont
  {Liu}}, \btxnamefont {D.~\btxlastnamefont {Sahoo}}, \btxnamefont
  {A.~\btxlastnamefont {Kumar}}\btxandcomma {} \btxandshort {.}\ \btxnamefont
  {S.\btxfnamespaceshort C.\btxfnamespaceshort H. \btxlastnamefont
  {Hoi}}\btxauthorcolon\ \btxjtitlefont {\btxifchangecase {Etsformer:
  Exponential smoothing transformers for time-series forecasting}{ETSformer:
  Exponential Smoothing Transformers for Time-series Forecasting}}.
\newblock \btxjournalfont {CoRR}, abs/2202.01381, 2022.
\newblock {\latintext \btxurlfont{https://arxiv.org/abs/2202.01381}}.

\bibitem {TimesNet}
\btxnamefont {H.~\btxlastnamefont {Wu}}, \btxnamefont {T.~\btxlastnamefont
  {Hu}}, \btxnamefont {Y.~\btxlastnamefont {Liu}}, \btxnamefont
  {H.~\btxlastnamefont {Zhou}}, \btxnamefont {J.~\btxlastnamefont
  {Wang}}\btxandcomma {} \btxandshort {.}\ \btxnamefont {M.~\btxlastnamefont
  {Long}}\btxauthorcolon\ \btxtitlefont {\btxifchangecase {Timesnet: Temporal
  2d-variation modeling for general time series analysis}{TimesNet: Temporal
  2D-Variation Modeling for General Time Series Analysis}}, 2023.

\bibitem {Autoformer}
\btxnamefont {H.~\btxlastnamefont {Wu}}, \btxnamefont {J.~\btxlastnamefont
  {Xu}}, \btxnamefont {J.~\btxlastnamefont {Wang}}\btxandcomma {} \btxandshort
  {.}\ \btxnamefont {M.~\btxlastnamefont {Long}}\btxauthorcolon\ \btxtitlefont
  {\btxifchangecase {Autoformer: decomposition transformers with
  auto-correlation for long-term series forecasting}{Autoformer: decomposition
  transformers with auto-correlation for long-term series forecasting}},
  2021\ifbtxprintISBN {, \mbox{\btxISBN~\btxISBNfont {9781713845393}}}.

\bibitem {cit_rev8}
\btxnamefont {M.~\btxlastnamefont {Wu}}, \btxnamefont {M.~\btxlastnamefont
  {Yan}}, \btxnamefont {W.~\btxlastnamefont {Li}}, \btxnamefont
  {X.~\btxlastnamefont {Ye}}, \btxnamefont {D.~\btxlastnamefont
  {Fan}}\btxandcomma {} \btxandshort {.}\ \btxnamefont {Y.~\btxlastnamefont
  {Xie}}\btxauthorcolon\ \btxjtitlefont {\btxifchangecase {Survey on
  characterizing and understanding gnns from a computer architecture
  perspective}{Survey on Characterizing and Understanding GNNs From a Computer
  Architecture Perspective}}.
\newblock \btxjournalfont {IEEE Transactions on Parallel and Distributed
  Systems}, 36(3):537--552, 2025.

\bibitem {art_363}
\btxnamefont {W.~\btxlastnamefont {Wu}} \btxandshort {.}\ \btxnamefont
  {Y.~\btxlastnamefont {Kang}}\btxauthorcolon\ \btxjtitlefont {\btxifchangecase
  {Ensemble empirical mode decomposition granger causality test dynamic graph
  attention transformer network: Integrating transformer and graph neural
  network models for multi-sensor cross-temporal granularity water demand
  forecasting}{Ensemble Empirical Mode Decomposition Granger causality test
  Dynamic Graph Attention Transformer Network: Integrating Transformer and
  graph neural network models for multi-sensor cross-temporal granularity water
  demand forecasting}}.
\newblock \btxjournalfont {Appl. Sci. (Basel)}, 14(8):3428,
  \btxprintmonthyear{.}{4}{2024}{short}.

\bibitem {art_364}
\btxnamefont {X.~\btxlastnamefont {Wu}}, \btxnamefont {M.~\btxlastnamefont
  {Chen}}, \btxnamefont {T.~\btxlastnamefont {Zhu}}, \btxnamefont
  {D.~\btxlastnamefont {Chen}}\btxandcomma {} \btxandshort {.}\ \btxnamefont
  {J.~\btxlastnamefont {Xiong}}\btxauthorcolon\ \btxjtitlefont
  {\btxifchangecase {Pre-training enhanced spatio-temporal graph neural network
  for predicting influent water quality and flow rate of wastewater treatment
  plant: Improvement of forecast accuracy and analysis of related
  factors}{Pre-training enhanced spatio-temporal graph neural network for
  predicting influent water quality and flow rate of wastewater treatment
  plant: Improvement of forecast accuracy and analysis of related factors}}.
\newblock \btxjournalfont {Sci. Total Environ.}, 951(175411):175411,
  \btxprintmonthyear{.}{11}{2024}{short}.

\bibitem {STtrans}
\btxnamefont {X.~\btxlastnamefont {Wu}}, \btxnamefont {C.~\btxlastnamefont
  {Huang}}, \btxnamefont {C.~\btxlastnamefont {Zhang}}\btxandcomma {}
  \btxandshort {.}\ \btxnamefont {N.\btxfnamespaceshort V. \btxlastnamefont
  {Chawla}}\btxauthorcolon\ \btxtitlefont {\btxifchangecase {Hierarchically
  structured transformer networks for fine-grained spatial event
  forecasting}{Hierarchically Structured Transformer Networks for Fine-Grained
  Spatial Event Forecasting}}.
\newblock \Btxinshort {.}\ \btxtitlefont {Proceedings of The Web Conference
  2020}, WWW '20, \btxpageshort {.}\ 2320–2330, New York, NY, USA, 2020.
  \btxpublisherfont {Association for Computing Machinery}\ifbtxprintISBN {,
  \mbox{\btxISBN~\btxISBNfont {9781450370233}}}.
\newblock {\latintext \btxurlfont{https://doi.org/10.1145/3366423.3380296}}.

\bibitem {art_443}
\btxnamefont {Y.~\btxlastnamefont {Wu}}, \btxnamefont {H.~\btxlastnamefont
  {Yang}}, \btxnamefont {Y.~\btxlastnamefont {Lin}}\btxandcomma {} \btxandshort
  {.}\ \btxnamefont {H.~\btxlastnamefont {Liu}}\btxauthorcolon\ \btxjtitlefont
  {\btxifchangecase {Spatiotemporal propagation learning for network-wide
  flight delay prediction}{Spatiotemporal propagation learning for network-wide
  flight delay prediction}}.
\newblock \btxjournalfont {IEEE Trans. Knowl. Data Eng.}, 36(1):386--400,
  \btxprintmonthyear{.}{1}{2024}{short}.

\bibitem {art_434}
\btxnamefont {Y.~\btxlastnamefont {Wu}}, \btxnamefont {J.~\btxlastnamefont
  {Yang}}, \btxnamefont {X.~\btxlastnamefont {Chen}}, \btxnamefont
  {Y.~\btxlastnamefont {Lin}}\btxandcomma {} \btxandshort {.}\ \btxnamefont
  {H.~\btxlastnamefont {Yang}}\btxauthorcolon\ \btxjtitlefont {\btxifchangecase
  {Long-term airport network performance forecasting with linear diffusion
  graph networks}{Long-term airport network performance forecasting with linear
  diffusion graph networks}}.
\newblock \btxjournalfont {IEEE Trans. Intell. Transp. Syst.},
  25(11):18264--18278, \btxprintmonthyear{.}{11}{2024}{short}.

\bibitem {CNNRNN-Res}
\btxnamefont {Y.~\btxlastnamefont {Wu}}, \btxnamefont {Y.~\btxlastnamefont
  {Yang}}, \btxnamefont {H.~\btxlastnamefont {Nishiura}}\btxandcomma {}
  \btxandshort {.}\ \btxnamefont {M.~\btxlastnamefont {Saitoh}}\btxauthorcolon\
  \btxtitlefont {\btxifchangecase {Deep learning for epidemiological
  predictions}{Deep Learning for Epidemiological Predictions}}.
\newblock \Btxinshort {.}\ \btxtitlefont {The 41st International ACM SIGIR
  Conference on Research \& Development in Information Retrieval}, SIGIR '18,
  \btxpageshort {.}\ 1085–1088. \btxpublisherfont {ACM},
  \btxprintmonthyear{.}{6}{2018}{short}.
\newblock {\latintext \btxurlfont{http://dx.doi.org/10.1145/3209978.3210077}}.

\bibitem {art_718}
\btxnamefont {Y.~\btxlastnamefont {Wu}}, \btxnamefont {X.~\btxlastnamefont
  {Zhang}}\btxandcomma {} \btxandshort {.}\ \btxnamefont {Y.~\btxlastnamefont
  {Wang}}\btxauthorcolon\ \btxtitlefont {\btxifchangecase {Adaptive graph
  neural diffusion for traffic demand forecasting}{Adaptive graph neural
  diffusion for traffic demand forecasting}}.
\newblock \Btxinshort {.}\ \btxtitlefont {Proceedings of the 32nd {ACM}
  International Conference on Information and Knowledge Management}, New York,
  NY, USA, \btxprintmonthyear{.}{10}{2023}{short}. \btxpublisherfont {ACM}.

\bibitem {cit_rev3}
\btxnamefont {Z.~\btxlastnamefont {Wu}}, \btxnamefont {S.~\btxlastnamefont
  {Pan}}, \btxnamefont {F.~\btxlastnamefont {Chen}}, \btxnamefont
  {G.~\btxlastnamefont {Long}}, \btxnamefont {C.~\btxlastnamefont
  {Zhang}}\btxandcomma {} \btxandshort {.}\ \btxnamefont {P.~\btxlastnamefont
  {Yu}}\btxauthorcolon\ \btxjtitlefont {\btxifchangecase {A comprehensive
  survey on graph neural networks}{A Comprehensive Survey on Graph Neural
  Networks}}.
\newblock \btxjournalfont {IEEE Transactions on Neural Networks and Learning
  Systems}, PP:1--21, \btxprintmonthyear{.}{03}{2020}{short}.

\bibitem {art_655}
\btxnamefont {Z.~\btxlastnamefont {Wu}}, \btxnamefont {S.~\btxlastnamefont
  {Pan}}, \btxnamefont {G.~\btxlastnamefont {Long}}, \btxnamefont
  {J.~\btxlastnamefont {Jiang}}, \btxnamefont {X.~\btxlastnamefont
  {Chang}}\btxandcomma {} \btxandshort {.}\ \btxnamefont {C.~\btxlastnamefont
  {Zhang}}\btxauthorcolon\ \btxtitlefont {\btxifchangecase {Connecting the
  dots: Multivariate time series forecasting with graph neural
  networks}{Connecting the Dots: Multivariate Time Series Forecasting with
  Graph Neural Networks}}.
\newblock \Btxinshort {.}\ \btxtitlefont {Proceedings of the 26th ACM SIGKDD
  International Conference on Knowledge Discovery \& Data Mining}, KDD '20,
  \btxpageshort {.}\ 753–763, New York, NY, USA, 2020. \btxpublisherfont
  {Association for Computing Machinery}\ifbtxprintISBN {,
  \mbox{\btxISBN~\btxISBNfont {9781450379984}}}.
\newblock {\latintext \btxurlfont{https://doi.org/10.1145/3394486.3403118}}.

\bibitem {GWN}
\btxnamefont {Z.~\btxlastnamefont {Wu}}, \btxnamefont {S.~\btxlastnamefont
  {Pan}}, \btxnamefont {G.~\btxlastnamefont {Long}}, \btxnamefont
  {J.~\btxlastnamefont {Jiang}}\btxandcomma {} \btxandshort {.}\ \btxnamefont
  {C.~\btxlastnamefont {Zhang}}\btxauthorcolon\ \btxtitlefont {\btxifchangecase
  {Graph wavenet for deep spatial-temporal graph modeling}{Graph WaveNet for
  Deep Spatial-Temporal Graph Modeling}}.
\newblock \Btxinshort {.}\ \btxtitlefont {Proceedings of the Twenty-Eighth
  International Joint Conference on Artificial Intelligence, {IJCAI-19}},
  \btxpagesshort {.}\ 1907--1913. \btxpublisherfont {International Joint
  Conferences on Artificial Intelligence Organization},
  \btxprintmonthyear{.}{7}{2019}{short}.
\newblock {\latintext \btxurlfont{https://doi.org/10.24963/ijcai.2019/264}}.

\bibitem {art_371}
\btxnamefont {H.~\btxlastnamefont {Xia}}, \btxnamefont {X.~\btxlastnamefont
  {Chen}}, \btxnamefont {Z.~\btxlastnamefont {Wang}}, \btxnamefont
  {X.~\btxlastnamefont {Chen}}\btxandcomma {} \btxandshort {.}\ \btxnamefont
  {F.~\btxlastnamefont {Dong}}\btxauthorcolon\ \btxjtitlefont {\btxifchangecase
  {A multi-modal deep-learning air quality prediction method based on
  multi-station time-series data and remote-sensing images: Case study of
  beijing and tianjin}{A multi-modal deep-learning air quality prediction
  method based on multi-station time-series data and remote-sensing images:
  Case study of Beijing and Tianjin}}.
\newblock \btxjournalfont {Entropy (Basel)}, 26(1),
  \btxprintmonthyear{.}{1}{2024}{short}.

\bibitem {ST-SHN}
\btxnamefont {L.~\btxlastnamefont {Xia}}, \btxnamefont {C.~\btxlastnamefont
  {Huang}}, \btxnamefont {Y.~\btxlastnamefont {Xu}}, \btxnamefont
  {P.~\btxlastnamefont {Dai}}, \btxnamefont {L.~\btxlastnamefont {Bo}},
  \btxnamefont {X.~\btxlastnamefont {Zhang}}\btxandcomma {} \btxandshort {.}\
  \btxnamefont {T.~\btxlastnamefont {Chen}}\btxauthorcolon\ \btxtitlefont
  {\btxifchangecase {Spatial-temporal sequential hypergraph network for crime
  prediction with dynamic multiplex relation learning}{Spatial-Temporal
  Sequential Hypergraph Network for Crime Prediction with Dynamic Multiplex
  Relation Learning}}.
\newblock \Btxinshort {.}\ \btxnamefont {Z.\btxfnamespaceshort H.
  \btxlastnamefont {Zhou}}\ (\btxeditorshort {.}): \btxtitlefont {Proceedings
  of the Thirtieth International Joint Conference on Artificial Intelligence,
  {IJCAI-21}}, \btxpagesshort {.}\ 1631--1637. \btxpublisherfont {International
  Joint Conferences on Artificial Intelligence Organization},
  \btxprintmonthyear{.}{8}{2021}{short}.
\newblock {\latintext \btxurlfont{https://doi.org/10.24963/ijcai.2021/225}},
  Main Track.

\bibitem {art_94}
\btxnamefont {T.~\btxlastnamefont {Xia}}, \btxnamefont {J.~\btxlastnamefont
  {Lin}}, \btxnamefont {Y.~\btxlastnamefont {Li}}, \btxnamefont
  {J.~\btxlastnamefont {Feng}}, \btxnamefont {P.~\btxlastnamefont {Hui}},
  \btxnamefont {F.~\btxlastnamefont {Sun}}, \btxnamefont {D.~\btxlastnamefont
  {Guo}}\btxandcomma {} \btxandshort {.}\ \btxnamefont {D.~\btxlastnamefont
  {Jin}}\btxauthorcolon\ \btxjtitlefont {\btxifchangecase {3dgcn: 3-dimensional
  dynamic graph convolutional network for citywide crowd flow
  prediction}{3DGCN: 3-Dimensional Dynamic Graph Convolutional Network for
  Citywide Crowd Flow Prediction}}.
\newblock \btxjournalfont {ACM Trans. Knowl. Discov. Data}, 15(6),
  \btxprintmonthyear{.}{jun}{2021}{short}\ifbtxprintISSN {,
  \mbox{\btxISSN~\btxISSNfont {1556-4681}}}.
\newblock {\latintext \btxurlfont{https://doi.org/10.1145/3451394}}.

\bibitem {art_693}
\btxnamefont {S.~\btxlastnamefont {Xiang}}, \btxnamefont {D.~\btxlastnamefont
  {Cheng}}, \btxnamefont {C.~\btxlastnamefont {Shang}}, \btxnamefont
  {Y.~\btxlastnamefont {Zhang}}\btxandcomma {} \btxandshort {.}\ \btxnamefont
  {Y.~\btxlastnamefont {Liang}}\btxauthorcolon\ \btxtitlefont {\btxifchangecase
  {Temporal and heterogeneous graph neural network for financial time series
  prediction}{Temporal and heterogeneous graph neural network for financial
  time series prediction}}.
\newblock \Btxinshort {.}\ \btxtitlefont {Proceedings of the 31st {ACM}
  International Conference on Information \& Knowledge Management}, New York,
  NY, USA, \btxprintmonthyear{.}{10}{2022}{short}. \btxpublisherfont {ACM}.

\bibitem {GED}
\btxnamefont {J.~\btxlastnamefont {Xie}}, \btxnamefont {J.~\btxlastnamefont
  {Zhang}}, \btxnamefont {J.~\btxlastnamefont {Yu}}\btxandcomma {} \btxandshort
  {.}\ \btxnamefont {L.~\btxlastnamefont {Xu}}\btxauthorcolon\ \btxjtitlefont
  {\btxifchangecase {An adaptive scale sea surface temperature predicting
  method based on deep learning with attention mechanism}{An Adaptive Scale Sea
  Surface Temperature Predicting Method Based on Deep Learning With Attention
  Mechanism}}.
\newblock \btxjournalfont {IEEE Geoscience and Remote Sensing Letters},
  17(5):740--744, 2020.

\bibitem {art_105}
\btxnamefont {Z.~\btxlastnamefont {Xie}}, \btxnamefont {W.~\btxlastnamefont
  {Lv}}, \btxnamefont {S.~\btxlastnamefont {Huang}}, \btxnamefont
  {Z.~\btxlastnamefont {Lu}}, \btxnamefont {B.~\btxlastnamefont
  {Du}}\btxandcomma {} \btxandshort {.}\ \btxnamefont {R.~\btxlastnamefont
  {Huang}}\btxauthorcolon\ \btxjtitlefont {\btxifchangecase {Sequential graph
  neural network for urban road traffic speed prediction}{Sequential Graph
  Neural Network for Urban Road Traffic Speed Prediction}}.
\newblock \btxjournalfont {IEEE Access}, 8:63349--63358, 2020.

\bibitem {art_640}
\btxnamefont {H.~\btxlastnamefont {Xu}}, \btxnamefont {W.~\btxlastnamefont
  {Fan}}, \btxnamefont {K.~\btxlastnamefont {Yi}}\btxandcomma {} \btxandshort
  {.}\ \btxnamefont {P.~\btxlastnamefont {Wang}}\btxauthorcolon\ \btxtitlefont
  {\btxifchangecase {Decoupled invariant attention network for multivariate
  time-series forecasting}{Decoupled Invariant Attention Network for
  multivariate Time-series Forecasting}}.
\newblock \Btxinshort {.}\ \btxtitlefont {Proceedings of the
  {Thirty-ThirdInternational} Joint Conference on Artificial Intelligence},
  California, \btxprintmonthyear{.}{8}{2024}{short}. \btxpublisherfont
  {International Joint Conferences on Artificial Intelligence Organization}.

\bibitem {CGRU}
\btxnamefont {L.~\btxlastnamefont {Xu}}, \btxnamefont {X.~\btxlastnamefont
  {Geng}}, \btxnamefont {X.~\btxlastnamefont {He}}, \btxnamefont
  {J.~\btxlastnamefont {Li}}\btxandcomma {} \btxandshort {.}\ \btxnamefont
  {J.~\btxlastnamefont {Yu}}\btxauthorcolon\ \btxjtitlefont {\btxifchangecase
  {Prediction in autism by deep learning short-time spontaneous hemodynamic
  fluctuations}{Prediction in Autism by Deep Learning Short-Time Spontaneous
  Hemodynamic Fluctuations}}.
\newblock \btxjournalfont {Frontiers in Neuroscience}, 13,
  \btxprintmonthyear{.}{11}{2019}{short}\ifbtxprintISSN {,
  \mbox{\btxISSN~\btxISSNfont {1662-453X}}}.
\newblock {\latintext \btxurlfont{http://dx.doi.org/10.3389/fnins.2019.01120}}.

\bibitem {art_189}
\btxnamefont {Y.~\btxlastnamefont {Xu}}, \btxnamefont {H.~\btxlastnamefont
  {Ying}}, \btxnamefont {S.~\btxlastnamefont {Qian}}, \btxnamefont
  {F.~\btxlastnamefont {Zhuang}}, \btxnamefont {X.~\btxlastnamefont {Zhang}},
  \btxnamefont {D.~\btxlastnamefont {Wang}}, \btxnamefont {J.~\btxlastnamefont
  {Wu}}\btxandcomma {} \btxandshort {.}\ \btxnamefont {H.~\btxlastnamefont
  {Xiong}}\btxauthorcolon\ \btxjtitlefont {\btxifchangecase {Time-aware
  context-gated graph attention network for clinical risk
  prediction}{Time-aware Context-Gated Graph Attention Network for Clinical
  Risk Prediction}}.
\newblock \btxjournalfont {IEEE Transactions on Knowledge and Data
  Engineering}, \btxpageshort {.}\ 1–12, 2022\ifbtxprintISSN {,
  \mbox{\btxISSN~\btxISSNfont {2326-3865}}}.
\newblock {\latintext
  \btxurlfont{http://dx.doi.org/10.1109/TKDE.2022.3181780}}.

\bibitem {art_13}
\btxnamefont {Q.~\btxlastnamefont {Xuan}}, \btxnamefont {J.~\btxlastnamefont
  {Zhou}}, \btxnamefont {K.~\btxlastnamefont {Qiu}}, \btxnamefont
  {Z.~\btxlastnamefont {Chen}}, \btxnamefont {D.~\btxlastnamefont {Xu}},
  \btxnamefont {S.~\btxlastnamefont {Zheng}}\btxandcomma {} \btxandshort {.}\
  \btxnamefont {X.~\btxlastnamefont {Yang}}\btxauthorcolon\ \btxjtitlefont
  {\btxifchangecase {Avgnet: Adaptive visibility graph neural network and its
  application in modulation classification}{AvgNet: Adaptive Visibility Graph
  Neural Network and Its Application in Modulation Classification}}.
\newblock \btxjournalfont {IEEE Transactions on Network Science and
  Engineering}, 9(3):1516--1526, 2022.

\bibitem {art_643}
\btxnamefont {V.\btxfnamespaceshort K. \btxlastnamefont {Yalavarthi}},
  \btxnamefont {K.~\btxlastnamefont {Madhusudhanan}}, \btxnamefont
  {R.~\btxlastnamefont {Scholz}}, \btxnamefont {N.~\btxlastnamefont {Ahmed}},
  \btxnamefont {J.~\btxlastnamefont {Burchert}}, \btxnamefont
  {S.~\btxlastnamefont {Jawed}}, \btxnamefont {S.~\btxlastnamefont
  {Born}}\btxandcomma {} \btxandshort {.}\ \btxnamefont {L.~\btxlastnamefont
  {Schmidt-Thieme}}\btxauthorcolon\ \btxjtitlefont {\btxifchangecase
  {{GraFITi}: Graphs for forecasting irregularly sampled time
  series}{{GraFITi}: Graphs for Forecasting Irregularly Sampled Time Series}}.
\newblock \btxjournalfont {Proc. Conf. AAAI Artif. Intell.},
  38(15):16255--16263, \btxprintmonthyear{.}{3}{2024}{short}.

\bibitem {art_28}
\btxnamefont {H.~\btxlastnamefont {Yang}}, \btxnamefont {W.~\btxlastnamefont
  {Li}}, \btxnamefont {S.~\btxlastnamefont {Hou}}, \btxnamefont
  {J.~\btxlastnamefont {Guan}}\btxandcomma {} \btxandshort {.}\ \btxnamefont
  {S.~\btxlastnamefont {Zhou}}\btxauthorcolon\ \btxjtitlefont {\btxifchangecase
  {Higrn: A hierarchical graph recurrent network for global sea surface
  temperature prediction}{HiGRN: A Hierarchical Graph Recurrent Network for
  Global Sea Surface Temperature Prediction}}.
\newblock \btxjournalfont {ACM Trans. Intell. Syst. Technol.}, 14(4),
  \btxprintmonthyear{.}{jul}{2023}{short}\ifbtxprintISSN {,
  \mbox{\btxISSN~\btxISSNfont {2157-6904}}}.
\newblock {\latintext \btxurlfont{https://doi.org/10.1145/3597937}}.

\bibitem {art_115}
\btxnamefont {J.~\btxlastnamefont {Yang}}, \btxnamefont {F.~\btxlastnamefont
  {Xie}}, \btxnamefont {J.~\btxlastnamefont {Yang}}, \btxnamefont
  {J.~\btxlastnamefont {Shi}}, \btxnamefont {J.~\btxlastnamefont
  {Zhao}}\btxandcomma {} \btxandshort {.}\ \btxnamefont {R.~\btxlastnamefont
  {Zhang}}\btxauthorcolon\ \btxjtitlefont {\btxifchangecase {Spatial-temporal
  correlated graph neural networks based on neighborhood feature selection for
  traffic data prediction}{Spatial-temporal correlated graph neural networks
  based on neighborhood feature selection for traffic data prediction}}.
\newblock \btxjournalfont {Applied Intelligence}, 53(4):4717 – 4732, 2023.
\newblock {\latintext
  \btxurlfont{https://www.scopus.com/inward/record.uri?eid=2-s2.0-85132117827&doi=10.1007

\bibitem {art_332}
\btxnamefont {L.~\btxlastnamefont {Yang}}, \btxnamefont {F.~\btxlastnamefont
  {Tsung}}, \btxnamefont {K.~\btxlastnamefont {Wang}}\btxandcomma {}
  \btxandshort {.}\ \btxnamefont {J.~\btxlastnamefont {Zhou}}\btxauthorcolon\
  \btxjtitlefont {\btxifchangecase {Wind power forecasting based on a
  spatial-temporal graph convolution network with limited engineering
  knowledge}{Wind power forecasting based on a spatial-temporal graph
  convolution network with limited engineering knowledge}}.
\newblock \btxjournalfont {IEEE Trans. Instrum. Meas.}, \btxpagesshort {.}\
  1--1, 2024.

\bibitem {art_146}
\btxnamefont {S.~\btxlastnamefont {Yang}}, \btxnamefont {Y.~\btxlastnamefont
  {Zhang}}\btxandcomma {} \btxandshort {.}\ \btxnamefont {Z.~\btxlastnamefont
  {Zhang}}\btxauthorcolon\ \btxjtitlefont {\btxifchangecase {Runoff prediction
  based on dynamic spatiotemporal graph neural network}{Runoff Prediction Based
  on Dynamic Spatiotemporal Graph Neural Network}}.
\newblock \btxjournalfont {Water}, 15(13), 2023\ifbtxprintISSN {,
  \mbox{\btxISSN~\btxISSNfont {2073-4441}}}.
\newblock {\latintext \btxurlfont{https://www.mdpi.com/2073-4441/15/13/2463}}.

\bibitem {art_156}
\btxnamefont {X.~\btxlastnamefont {Yang}}, \btxnamefont {Y.~\btxlastnamefont
  {Zheng}}, \btxnamefont {Y.~\btxlastnamefont {Zhang}}, \btxnamefont
  {D.\btxfnamespaceshort S.\btxfnamespaceshort H. \btxlastnamefont
  {Wong}}\btxandcomma {} \btxandshort {.}\ \btxnamefont {W.~\btxlastnamefont
  {Yang}}\btxauthorcolon\ \btxjtitlefont {\btxifchangecase {Bearing remaining
  useful life prediction based on regression shapalet and graph neural
  network}{Bearing Remaining Useful Life Prediction Based on Regression
  Shapalet and Graph Neural Network}}.
\newblock \btxjournalfont {IEEE Transactions on Instrumentation and
  Measurement}, 71, 2022.
\newblock {\latintext
  \btxurlfont{https://www.scopus.com/inward/record.uri?eid=2-s2.0-85124758480&doi=10.1109

\bibitem {CFCC-LSTM}
\btxnamefont {Y.~\btxlastnamefont {Yang}}, \btxnamefont {J.~\btxlastnamefont
  {Dong}}, \btxnamefont {X.~\btxlastnamefont {Sun}}, \btxnamefont
  {E.~\btxlastnamefont {Lima}}, \btxnamefont {Q.~\btxlastnamefont
  {Mu}}\btxandcomma {} \btxandshort {.}\ \btxnamefont {X.~\btxlastnamefont
  {Wang}}\btxauthorcolon\ \btxjtitlefont {\btxifchangecase {A cfcc-lstm model
  for sea surface temperature prediction}{A CFCC-LSTM Model for Sea Surface
  Temperature Prediction}}.
\newblock \btxjournalfont {IEEE Geoscience and Remote Sensing Letters},
  15(2):207--211, 2018.

\bibitem {art_417}
\btxnamefont {Y.~\btxlastnamefont {Yang}}, \btxnamefont {L.~\btxlastnamefont
  {Kou}}, \btxnamefont {P.~\btxlastnamefont {Jiao}}, \btxnamefont
  {J.~\btxlastnamefont {Zhang}}, \btxnamefont {C.~\btxlastnamefont
  {Miao}}\btxandcomma {} \btxandshort {.}\ \btxnamefont {Y.~\btxlastnamefont
  {Lin}}\btxauthorcolon\ \btxjtitlefont {\btxifchangecase {{GAGNN}: Generative
  adversarial network and graph neural network for prognostic and health
  management}{{GAGNN}: Generative adversarial network and graph neural network
  for prognostic and health management}}.
\newblock \btxjournalfont {IEEE Internet Things J.}, 11(24):39448--39463,
  \btxprintmonthyear{.}{12}{2024}{short}.

\bibitem {art_68}
\btxnamefont {Z.~\btxlastnamefont {Yang}}, \btxnamefont {Z.~\btxlastnamefont
  {Zhihan}}, \btxnamefont {L.~\btxlastnamefont {Haiying}}, \btxnamefont
  {Z.~\btxlastnamefont {Weiyi}}, \btxnamefont {D.~\btxlastnamefont
  {Qian}}\btxandcomma {} \btxandshort {.}\ \btxnamefont {T.~\btxlastnamefont
  {Mingjie}}\btxauthorcolon\ \btxjtitlefont {\btxifchangecase {Research on
  commodities constraint optimization based on graph neural network
  prediction}{Research on Commodities Constraint Optimization Based on Graph
  Neural Network Prediction}}.
\newblock \btxjournalfont {IEEE Access}, 11:90131--90142, 2023.

\bibitem {art_413}
\btxnamefont {X.~\btxlastnamefont {Yanqin}}, \btxnamefont {W.~\btxlastnamefont
  {Huandong}}, \btxnamefont {L.~\btxlastnamefont {Guanghua}}, \btxnamefont
  {L.~\btxlastnamefont {Yong}}\btxandcomma {} \btxandshort {.}\ \btxnamefont
  {J.~\btxlastnamefont {Tao}}\btxauthorcolon\ \btxjtitlefont {\btxifchangecase
  {Graph neural ordinary differential equations for epidemic forecasting}{Graph
  neural ordinary differential equations for epidemic forecasting}}.
\newblock \btxjournalfont {CCF Trans. Pervasive Comp. Interact.},
  6(3):281--295, \btxprintmonthyear{.}{9}{2024}{short}.

\bibitem {STDN}
\btxnamefont {H.~\btxlastnamefont {Yao}}, \btxnamefont {X.~\btxlastnamefont
  {Tang}}, \btxnamefont {H.~\btxlastnamefont {Wei}}, \btxnamefont
  {G.~\btxlastnamefont {Zheng}}, \btxnamefont {Y.~\btxlastnamefont
  {Yu}}\btxandcomma {} \btxandshort {.}\ \btxnamefont {Z.~\btxlastnamefont
  {Li}}\btxauthorcolon\ \btxtitlefont {\btxifchangecase {Revisiting
  spatial-temporal similarity: A deep learning framework for traffic
  prediction}{Revisiting Spatial-Temporal Similarity: A Deep Learning Framework
  for Traffic Prediction}}.
\newblock \Btxinshort {.}\ \btxtitlefont {AAAI Conference on Artificial
  Intelligence, 2019}, 2019.

\bibitem {DMVST-Net}
\btxnamefont {H.~\btxlastnamefont {Yao}}, \btxnamefont {F.~\btxlastnamefont
  {Wu}}, \btxnamefont {J.~\btxlastnamefont {ke}}, \btxnamefont
  {X.~\btxlastnamefont {Tang}}, \btxnamefont {Y.~\btxlastnamefont {Jia}},
  \btxnamefont {S.~\btxlastnamefont {Lu}}, \btxnamefont {P.~\btxlastnamefont
  {Gong}}\btxandcomma {} \btxandshort {.}\ \btxnamefont {J.~\btxlastnamefont
  {Ye}}\btxauthorcolon\ \btxjtitlefont {\btxifchangecase {Deep multi-view
  spatial-temporal network for taxi demand prediction}{Deep Multi-View
  Spatial-Temporal Network for Taxi Demand Prediction}}.
\newblock \btxjournalfont {Proceedings of the AAAI Conference on Artificial
  Intelligence}, 32, \btxprintmonthyear{.}{02}{2018}{short}.

\bibitem {art_649}
\btxnamefont {J.~\btxlastnamefont {Ye}}, \btxnamefont {Z.~\btxlastnamefont
  {Liu}}, \btxnamefont {B.~\btxlastnamefont {Du}}, \btxnamefont
  {L.~\btxlastnamefont {Sun}}, \btxnamefont {W.~\btxlastnamefont {Li}},
  \btxnamefont {Y.~\btxlastnamefont {Fu}}\btxandcomma {} \btxandshort {.}\
  \btxnamefont {H.~\btxlastnamefont {Xiong}}\btxauthorcolon\ \btxtitlefont
  {\btxifchangecase {Learning the evolutionary and multi-scale graph structure
  for multivariate time series forecasting}{Learning the evolutionary and
  multi-scale graph structure for multivariate time series forecasting}}.
\newblock \Btxinshort {.}\ \btxtitlefont {Proceedings of the 28th {ACM}
  {SIGKDD} Conference on Knowledge Discovery and Data Mining}, \btxvolumeshort
  {.}~\btxvolumefont {32}, \btxpagesshort {.}\ 2296--2306, New York, NY, USA,
  \btxprintmonthyear{.}{8}{2022}{short}. \btxpublisherfont {ACM}.

\bibitem {CCRNN}
\btxnamefont {J.~\btxlastnamefont {Ye}}, \btxnamefont {L.~\btxlastnamefont
  {Sun}}, \btxnamefont {B.~\btxlastnamefont {Du}}, \btxnamefont
  {Y.~\btxlastnamefont {Fu}}\btxandcomma {} \btxandshort {.}\ \btxnamefont
  {H.~\btxlastnamefont {Xiong}}\btxauthorcolon\ \btxjtitlefont
  {\btxifchangecase {Coupled layer-wise graph convolution for transportation
  demand prediction}{Coupled Layer-wise Graph Convolution for Transportation
  Demand Prediction}}.
\newblock \btxjournalfont {Proceedings of the AAAI Conference on Artificial
  Intelligence}, 35(5):4617--4625, \btxprintmonthyear{.}{May}{2021}{short}.
\newblock {\latintext
  \btxurlfont{https://ojs.aaai.org/index.php/AAAI/article/view/16591}}.

\bibitem {art_639}
\btxnamefont {J.~\btxlastnamefont {Yi}} \btxandshort {.}\ \btxnamefont
  {J.~\btxlastnamefont {Park}}\btxauthorcolon\ \btxtitlefont {\btxifchangecase
  {Hypergraph convolutional recurrent neural network}{Hypergraph Convolutional
  Recurrent Neural Network}}.
\newblock \Btxinshort {.}\ \btxtitlefont {Proceedings of the 26th {ACM}
  {SIGKDD} International Conference on Knowledge Discovery \& Data Mining},
  \btxvolumeshort {.}\ \btxvolumefont {187}, \btxpagesshort {.}\ 3366--3376,
  New York, NY, USA, \btxprintmonthyear{.}{8}{2020}{short}. \btxpublisherfont
  {ACM}.

\bibitem {art_688}
\btxnamefont {K.~\btxlastnamefont {Yi}}, \btxnamefont {Q.~\btxlastnamefont
  {Zhang}}, \btxnamefont {W.~\btxlastnamefont {Fan}}, \btxnamefont
  {H.~\btxlastnamefont {He}}, \btxnamefont {L.~\btxlastnamefont {Hu}},
  \btxnamefont {P.~\btxlastnamefont {Wang}}, \btxnamefont {N.~\btxlastnamefont
  {An}}, \btxnamefont {L.~\btxlastnamefont {Cao}}\btxandcomma {} \btxandshort
  {.}\ \btxnamefont {Z.~\btxlastnamefont {Niu}}\btxauthorcolon\ \btxtitlefont
  {\btxifchangecase {Fouriergnn: rethinking multivariate time series
  forecasting from a pure graph perspective}{FourierGNN: rethinking
  multivariate time series forecasting from a pure graph perspective}}.
\newblock \Btxinshort {.}\ \btxtitlefont {Proceedings of the 37th International
  Conference on Neural Information Processing Systems}, NIPS '23, Red Hook, NY,
  USA, 2023. \btxpublisherfont {Curran Associates Inc.}

\bibitem {art_333}
\btxnamefont {P.~\btxlastnamefont {Yi}}, \btxnamefont {Z.~\btxlastnamefont
  {Bao}}, \btxnamefont {F.~\btxlastnamefont {Huang}}, \btxnamefont
  {J.~\btxlastnamefont {Wang}}, \btxnamefont {J.~\btxlastnamefont
  {Peng}}\btxandcomma {} \btxandshort {.}\ \btxnamefont {L.~\btxlastnamefont
  {Zhang}}\btxauthorcolon\ \btxjtitlefont {\btxifchangecase {Towards effective
  long-term wind power forecasting: A deep conditional generative
  spatio-temporal approach}{Towards effective long-term wind power forecasting:
  A deep conditional generative spatio-temporal approach}}.
\newblock \btxjournalfont {IEEE Trans. Knowl. Data Eng.}, 36(12):9403--9417,
  \btxprintmonthyear{.}{12}{2024}{short}.

\bibitem {art_398}
\btxnamefont {N.~\btxlastnamefont {Yin}}, \btxnamefont {L.~\btxlastnamefont
  {Shen}}, \btxnamefont {H.~\btxlastnamefont {Xiong}}, \btxnamefont
  {B.~\btxlastnamefont {Gu}}, \btxnamefont {C.~\btxlastnamefont {Chen}},
  \btxnamefont {X.\btxfnamespaceshort S. \btxlastnamefont {Hua}}, \btxnamefont
  {S.~\btxlastnamefont {Liu}}\btxandcomma {} \btxandshort {.}\ \btxnamefont
  {X.~\btxlastnamefont {Luo}}\btxauthorcolon\ \btxjtitlefont {\btxifchangecase
  {Messages are never propagated alone: Collaborative hypergraph neural network
  for time-series forecasting}{Messages are never propagated alone:
  Collaborative hypergraph neural network for time-series forecasting}}.
\newblock \btxjournalfont {IEEE Trans. Pattern Anal. Mach. Intell.},
  46(4):2333--2347, \btxprintmonthyear{.}{4}{2024}{short}.

\bibitem {art_463}
\btxnamefont {P.~\btxlastnamefont {Yin}}, \btxnamefont {J.~\btxlastnamefont
  {Zhou}}, \btxnamefont {Y.~\btxlastnamefont {Ge}}\btxandcomma {} \btxandshort
  {.}\ \btxnamefont {Z.~\btxlastnamefont {Chen}}\btxauthorcolon\ \btxjtitlefont
  {\btxifchangecase {{DTSG-net}: Dynamic time series graph neural network and
  it's application in modulation recognition}{{DTSG-net}: Dynamic time series
  graph neural network and it's application in modulation recognition}}.
\newblock \btxjournalfont {IEEE Internet Things J.}, \btxpagesshort {.}\ 1--1,
  2024.

\bibitem {interpret_2}
\btxnamefont {R.~\btxlastnamefont {Ying}}, \btxnamefont {D.~\btxlastnamefont
  {Bourgeois}}, \btxnamefont {J.~\btxlastnamefont {You}}, \btxnamefont
  {M.~\btxlastnamefont {Zitnik}}\btxandcomma {} \btxandshort {.}\ \btxnamefont
  {J.~\btxlastnamefont {Leskovec}}\btxauthorcolon\ \btxtitlefont {GNNExplainer:
  generating explanations for graph neural networks}.
\newblock \btxpublisherfont {Curran Associates Inc.}, Red Hook, NY, USA, 2019.

\bibitem {art_680}
\btxnamefont {R.~\btxlastnamefont {Younis}} \btxandshort {.}\ \btxnamefont
  {Z.~\btxlastnamefont {Ahmadi}}\btxauthorcolon\ \btxtitlefont
  {\btxifchangecase {{HyperTime}: A dynamic hypergraph approach for time series
  classification}{{HyperTime}: A dynamic hypergraph approach for time series
  classification}}.
\newblock \Btxinshort {.}\ \btxtitlefont {2024 {IEEE} International Conference
  on Data Mining ({ICDM})}, \btxpagesshort {.}\ 570--579. \btxpublisherfont
  {IEEE}, \btxprintmonthyear{.}{12}{2024}{short}.

\bibitem {STGCN}
\btxnamefont {B.~\btxlastnamefont {Yu}}, \btxnamefont {H.~\btxlastnamefont
  {Yin}}\btxandcomma {} \btxandshort {.}\ \btxnamefont {Z.~\btxlastnamefont
  {Zhu}}\btxauthorcolon\ \btxtitlefont {\btxifchangecase {Spatio-temporal graph
  convolutional networks: A deep learning framework for traffic
  forecasting}{Spatio-Temporal Graph Convolutional Networks: A Deep Learning
  Framework for Traffic Forecasting}}.
\newblock \Btxinshort {.}\ \btxtitlefont {Proceedings of the Twenty-Seventh
  International Joint Conference on Artificial Intelligence, {IJCAI-18}},
  \btxpagesshort {.}\ 3634--3640. \btxpublisherfont {International Joint
  Conferences on Artificial Intelligence Organization},
  \btxprintmonthyear{.}{7}{2018}{short}.
\newblock {\latintext \btxurlfont{https://doi.org/10.24963/ijcai.2018/505}}.

\bibitem {art_642}
\btxnamefont {C.~\btxlastnamefont {Yu}}, \btxnamefont {F.~\btxlastnamefont
  {Wang}}, \btxnamefont {Z.~\btxlastnamefont {Shao}}, \btxnamefont
  {T.~\btxlastnamefont {Qian}}, \btxnamefont {Z.~\btxlastnamefont {Zhang}},
  \btxnamefont {W.~\btxlastnamefont {Wei}}\btxandcomma {} \btxandshort {.}\
  \btxnamefont {Y.~\btxlastnamefont {Xu}}\btxauthorcolon\ \btxtitlefont
  {\btxifchangecase {{GinAR}: An end-to-end multivariate time series
  forecasting model suitable for variable missing}{{GinAR}: An end-to-end
  multivariate time series forecasting model suitable for variable missing}}.
\newblock \Btxinshort {.}\ \btxtitlefont {Proceedings of the 30th {ACM}
  {SIGKDD} Conference on Knowledge Discovery and Data Mining}, \btxvolumeshort
  {.}~\btxvolumefont {34}, \btxpagesshort {.}\ 3989--4000, New York, NY, USA,
  \btxprintmonthyear{.}{8}{2024}{short}. \btxpublisherfont {ACM}.

\bibitem {art_148}
\btxnamefont {L.~\btxlastnamefont {Yu}}, \btxnamefont {M.~\btxlastnamefont
  {Li}}, \btxnamefont {W.~\btxlastnamefont {Jin}}, \btxnamefont
  {Y.~\btxlastnamefont {Guo}}, \btxnamefont {Q.~\btxlastnamefont {Wang}},
  \btxnamefont {F.~\btxlastnamefont {Yan}}\btxandcomma {} \btxandshort {.}\
  \btxnamefont {P.~\btxlastnamefont {Li}}\btxauthorcolon\ \btxjtitlefont
  {\btxifchangecase {Step: A spatio-temporal fine-granular user traffic
  prediction system for cellular networks}{STEP: A Spatio-Temporal
  Fine-Granular User Traffic Prediction System for Cellular Networks}}.
\newblock \btxjournalfont {IEEE Transactions on Mobile Computing}, 20(12):3453
  – 3466, 2021.
\newblock {\latintext
  \btxurlfont{https://www.scopus.com/inward/record.uri?eid=2-s2.0-85112368991&doi=10.1109

\bibitem {art_24}
\btxnamefont {M.~\btxlastnamefont {Yu}}, \btxnamefont {Z.~\btxlastnamefont
  {Zhang}}, \btxnamefont {X.~\btxlastnamefont {Li}}, \btxnamefont
  {J.~\btxlastnamefont {Yu}}, \btxnamefont {J.~\btxlastnamefont {Gao}},
  \btxnamefont {Z.~\btxlastnamefont {Liu}}, \btxnamefont {B.~\btxlastnamefont
  {You}}, \btxnamefont {X.~\btxlastnamefont {Zheng}}\btxandcomma {}
  \btxandshort {.}\ \btxnamefont {R.~\btxlastnamefont {Yu}}\btxauthorcolon\
  \btxjtitlefont {\btxifchangecase {Superposition graph neural network for
  offshore wind power prediction}{Superposition Graph Neural Network for
  offshore wind power prediction}}.
\newblock \btxjournalfont {Future Generation Computer Systems}, 113:145 –
  157, 2020.
\newblock {\latintext
  \btxurlfont{https://www.scopus.com/inward/record.uri?eid=2-s2.0-85087721191&doi=10.1016

\bibitem {art_732}
\btxnamefont {Q.~\btxlastnamefont {Yu}}, \btxnamefont {W.~\btxlastnamefont
  {Ding}}, \btxnamefont {H.~\btxlastnamefont {Zhang}}, \btxnamefont
  {Y.~\btxlastnamefont {Yang}}\btxandcomma {} \btxandshort {.}\ \btxnamefont
  {T.~\btxlastnamefont {Zhang}}\btxauthorcolon\ \btxtitlefont {\btxifchangecase
  {Rethinking attention mechanism for spatio-temporal modeling: A decoupling
  perspective in traffic flow prediction}{Rethinking attention mechanism for
  spatio-temporal modeling: A decoupling perspective in traffic flow
  prediction}}.
\newblock \Btxinshort {.}\ \btxtitlefont {Proceedings of the 33rd {ACM}
  International Conference on Information and Knowledge Management},
  \btxvolumeshort {.}~\btxvolumefont {6}, \btxpagesshort {.}\ 3032--3041, New
  York, NY, USA, \btxprintmonthyear{.}{10}{2024}{short}. \btxpublisherfont
  {ACM}.

\bibitem {art_49}
\btxnamefont {S.~\btxlastnamefont {Yu}}, \btxnamefont {F.~\btxlastnamefont
  {Xia}}, \btxnamefont {S.~\btxlastnamefont {Li}}, \btxnamefont
  {M.~\btxlastnamefont {Hou}}\btxandcomma {} \btxandshort {.}\ \btxnamefont
  {Q.\btxfnamespaceshort Z. \btxlastnamefont {Sheng}}\btxauthorcolon\
  \btxjtitlefont {\btxifchangecase {Spatio-temporal graph learning for epidemic
  prediction}{Spatio-temporal Graph Learning for Epidemic Prediction}}.
\newblock \btxjournalfont {ACM Trans. Intell. Syst. Technol.}, 14(2),
  \btxprintmonthyear{.}{feb}{2023}{short}\ifbtxprintISSN {,
  \mbox{\btxISSN~\btxISSNfont {2157-6904}}}.
\newblock {\latintext \btxurlfont{https://doi.org/10.1145/3579815}}.

\bibitem {art_353}
\btxnamefont {Y.~\btxlastnamefont {Yu}}, \btxnamefont {X.~\btxlastnamefont
  {Jiang}}, \btxnamefont {D.~\btxlastnamefont {Huang}}, \btxnamefont
  {Y.~\btxlastnamefont {Li}}, \btxnamefont {M.~\btxlastnamefont
  {Yue}}\btxandcomma {} \btxandshort {.}\ \btxnamefont {T.~\btxlastnamefont
  {Zhao}}\btxauthorcolon\ \btxjtitlefont {\btxifchangecase {{PIDGeuN}: Graph
  neural network-enabled transient dynamics prediction of networked microgrids
  through full-field measurement}{{PIDGeuN}: Graph neural network-enabled
  transient dynamics prediction of networked microgrids through full-field
  measurement}}.
\newblock \btxjournalfont {IEEE Access}, \btxpagesshort {.}\ 1--1, 2024.

\bibitem {art_54}
\btxnamefont {Z.~\btxlastnamefont {Yuan}}, \btxnamefont {X.~\btxlastnamefont
  {Li}}, \btxnamefont {S.~\btxlastnamefont {Liu}}\btxandcomma {} \btxandshort
  {.}\ \btxnamefont {Z.~\btxlastnamefont {Ma}}\btxauthorcolon\ \btxjtitlefont
  {\btxifchangecase {A recursive multi-head graph attention residual network
  for high-speed train wheelset bearing fault diagnosis}{A recursive multi-head
  graph attention residual network for high-speed train wheelset bearing fault
  diagnosis}}.
\newblock \btxjournalfont {Measurement Science and Technology}, 34(6), 2023.
\newblock {\latintext
  \btxurlfont{https://www.scopus.com/inward/record.uri?eid=2-s2.0-85149833631&doi=10.1088

\bibitem {art_149}
\btxnamefont {A.~\btxlastnamefont {Zanfei}}, \btxnamefont
  {B.\btxfnamespaceshort M. \btxlastnamefont {Brentan}}, \btxnamefont
  {A.~\btxlastnamefont {Menapace}}, \btxnamefont {M.~\btxlastnamefont
  {Righetti}}\btxandcomma {} \btxandshort {.}\ \btxnamefont
  {M.~\btxlastnamefont {Herrera}}\btxauthorcolon\ \btxjtitlefont
  {\btxifchangecase {Graph convolutional recurrent neural networks for water
  demand forecasting}{Graph Convolutional Recurrent Neural Networks for Water
  Demand Forecasting}}.
\newblock \btxjournalfont {Water Resources Research}, 58(7), 2022.
\newblock {\latintext
  \btxurlfont{https://www.scopus.com/inward/record.uri?eid=2-s2.0-85134874398&doi=10.1029

\bibitem {art_422}
\btxnamefont {D.~\btxlastnamefont {Zang}}, \btxnamefont {Y.~\btxlastnamefont
  {Ding}}, \btxnamefont {J.~\btxlastnamefont {Zhao}}, \btxnamefont
  {K.~\btxlastnamefont {Tang}}\btxandcomma {} \btxandshort {.}\ \btxnamefont
  {H.~\btxlastnamefont {Zhu}}\btxauthorcolon\ \btxjtitlefont {\btxifchangecase
  {Predictive resilience assessment of road networks based on dynamic
  multi-granularity graph neural network}{Predictive resilience assessment of
  road networks based on dynamic multi-granularity graph neural network}}.
\newblock \btxjournalfont {Neurocomputing}, 601(128207):128207,
  \btxprintmonthyear{.}{10}{2024}{short}.

\bibitem {survey_spatiotemporal}
\btxnamefont {A.~\btxlastnamefont {Zeghina}}, \btxnamefont {A.~\btxlastnamefont
  {Leborgne}}, \btxnamefont {F.~\btxlastnamefont {{Le Ber}}}\btxandcomma {}
  \btxandshort {.}\ \btxnamefont {A.~\btxlastnamefont
  {Vacavant}}\btxauthorcolon\ \btxjtitlefont {\btxifchangecase {Deep learning
  on spatiotemporal graphs: A systematic review, methodological landscape, and
  research opportunities}{Deep learning on spatiotemporal graphs: A systematic
  review, methodological landscape, and research opportunities}}.
\newblock \btxjournalfont {Neurocomputing}, 594:127861, 2024\ifbtxprintISSN {,
  \mbox{\btxISSN~\btxISSNfont {0925-2312}}}.
\newblock {\latintext
  \btxurlfont{https://www.sciencedirect.com/science/article/pii/S0925231224006325}}.

\bibitem {visibility_Gori}
\btxnamefont {K.~\btxlastnamefont {Zeinalipour}} \btxandshort {.}\ \btxnamefont
  {M.~\btxlastnamefont {Gori}}\btxauthorcolon\ \btxtitlefont {Graph Neural
  Networks for Topological Feature Extraction in ECG Classification},
  \btxpagesshort {.}\ 17--27.
\newblock \btxpublisherfont {Springer Nature Singapore}, Singapore,
  2023\ifbtxprintISBN {, \mbox{\btxISBN~\btxISBNfont {978-981-99-3592-5}}}.
\newblock {\latintext
  \btxurlfont{https://doi.org/10.1007/978-981-99-3592-5_2}}.

\bibitem {DLinear}
\btxnamefont {A.~\btxlastnamefont {Zeng}}, \btxnamefont {M.~\btxlastnamefont
  {Chen}}, \btxnamefont {L.~\btxlastnamefont {Zhang}}\btxandcomma {}
  \btxandshort {.}\ \btxnamefont {Q.~\btxlastnamefont {Xu}}\btxauthorcolon\
  \btxtitlefont {\btxifchangecase {Are transformers effective for time series
  forecasting?}{Are transformers effective for time series forecasting?}},
  2023\ifbtxprintISBN {, \mbox{\btxISBN~\btxISBNfont {978-1-57735-880-0}}}.
\newblock {\latintext \btxurlfont{https://doi.org/10.1609/aaai.v37i9.26317}}.

\bibitem {FC-GAGA}
\btxnamefont {Z.~\btxlastnamefont {Zeng}}, \btxnamefont {W.~\btxlastnamefont
  {Zhao}}, \btxnamefont {P.~\btxlastnamefont {Qian}}, \btxnamefont
  {Y.~\btxlastnamefont {Zhou}}, \btxnamefont {Z.~\btxlastnamefont {Zhao}},
  \btxnamefont {C.~\btxlastnamefont {Chen}}\btxandcomma {} \btxandshort {.}\
  \btxnamefont {C.~\btxlastnamefont {Guan}}\btxauthorcolon\ \btxjtitlefont
  {\btxifchangecase {Robust traffic prediction from spatial–temporal data
  based on conditional distribution learning}{Robust Traffic Prediction From
  Spatial–Temporal Data Based on Conditional Distribution Learning}}.
\newblock \btxjournalfont {IEEE Transactions on Cybernetics},
  52(12):13458--13471, 2022.

\bibitem {TST}
\btxnamefont {G.~\btxlastnamefont {Zerveas}}, \btxnamefont {S.~\btxlastnamefont
  {Jayaraman}}, \btxnamefont {D.~\btxlastnamefont {Patel}}, \btxnamefont
  {A.~\btxlastnamefont {Bhamidipaty}}\btxandcomma {} \btxandshort {.}\
  \btxnamefont {C.~\btxlastnamefont {Eickhoff}}\btxauthorcolon\ \btxtitlefont
  {\btxifchangecase {A transformer-based framework for multivariate time series
  representation learning}{A Transformer-based Framework for Multivariate Time
  Series Representation Learning}}.
\newblock \Btxinshort {.}\ \btxtitlefont {Proceedings of the 27th ACM SIGKDD
  Conference on Knowledge Discovery \& Data Mining}, KDD '21, \btxpageshort
  {.}\ 2114–2124, New York, NY, USA, 2021. \btxpublisherfont {Association for
  Computing Machinery}\ifbtxprintISBN {, \mbox{\btxISBN~\btxISBNfont
  {9781450383325}}}.
\newblock {\latintext \btxurlfont{https://doi.org/10.1145/3447548.3467401}}.

\bibitem {art_711}
\btxnamefont {D.~\btxlastnamefont {Zha}}, \btxnamefont {K.\btxfnamespaceshort
  H. \btxlastnamefont {Lai}}, \btxnamefont {K.~\btxlastnamefont
  {Zhou}}\btxandcomma {} \btxandshort {.}\ \btxnamefont {X.~\btxlastnamefont
  {Hu}}\btxauthorcolon\ \btxtitlefont {Towards Similarity-Aware Time-Series
  Classification}, \btxpublisherfont {\btxpagesshort {.}\ 199--207}.

\bibitem {art_88}
\btxnamefont {F.~\btxlastnamefont {Zhan}}, \btxnamefont {X.~\btxlastnamefont
  {Zhou}}, \btxnamefont {S.~\btxlastnamefont {Li}}, \btxnamefont
  {D.~\btxlastnamefont {Jia}}\btxandcomma {} \btxandshort {.}\ \btxnamefont
  {H.~\btxlastnamefont {Song}}\btxauthorcolon\ \btxjtitlefont {\btxifchangecase
  {Learning latent odes with graph rnn for multi-channel time series
  forecasting}{Learning Latent ODEs With Graph RNN for Multi-Channel Time
  Series Forecasting}}.
\newblock \btxjournalfont {IEEE Signal Processing Letters}, 30:1432 – 1436,
  2023.
\newblock {\latintext
  \btxurlfont{https://www.scopus.com/inward/record.uri?eid=2-s2.0-85173313670&doi=10.1109

\bibitem {DBN}
\btxnamefont {C.~\btxlastnamefont {Zhang}}, \btxnamefont {P.~\btxlastnamefont
  {Lim}}, \btxnamefont {A.\btxfnamespaceshort K. \btxlastnamefont
  {Qin}}\btxandcomma {} \btxandshort {.}\ \btxnamefont {K.\btxfnamespaceshort
  C. \btxlastnamefont {Tan}}\btxauthorcolon\ \btxjtitlefont {\btxifchangecase
  {Multiobjective deep belief networks ensemble for remaining useful life
  estimation in prognostics}{Multiobjective Deep Belief Networks Ensemble for
  Remaining Useful Life Estimation in Prognostics}}.
\newblock \btxjournalfont {IEEE Transactions on Neural Networks and Learning
  Systems}, 28(10):2306--2318, 2017.

\bibitem {art_679}
\btxnamefont {H.~\btxlastnamefont {Zhang}}, \btxnamefont {L.~\btxlastnamefont
  {Yu}}, \btxnamefont {X.~\btxlastnamefont {Xiao}}, \btxnamefont
  {Q.~\btxlastnamefont {Li}}, \btxnamefont {F.~\btxlastnamefont {Mercaldo}},
  \btxnamefont {X.~\btxlastnamefont {Luo}}\btxandcomma {} \btxandshort {.}\
  \btxnamefont {Q.~\btxlastnamefont {Liu}}\btxauthorcolon\ \btxtitlefont
  {\btxifchangecase {{TFE-GNN}: A temporal fusion encoder using graph neural
  networks for fine-grained encrypted traffic classification}{{TFE-GNN}: A
  temporal fusion encoder using graph neural networks for fine-grained
  encrypted traffic classification}}.
\newblock \Btxinshort {.}\ \btxtitlefont {Proceedings of the {ACM} Web
  Conference 2023}, \btxpagesshort {.}\ 2066--2075, New York, NY, USA,
  \btxprintmonthyear{.}{4}{2023}{short}. \btxpublisherfont {ACM}.

\bibitem {art_487}
\btxnamefont {J.~\btxlastnamefont {Zhang}}, \btxnamefont {Y.~\btxlastnamefont
  {Cheng}}\btxandcomma {} \btxandshort {.}\ \btxnamefont {X.~\btxlastnamefont
  {He}}\btxauthorcolon\ \btxjtitlefont {\btxifchangecase {Fault diagnosis of
  energy networks based on improved spatial--temporal graph neural network with
  massive missing data}{Fault diagnosis of energy networks based on improved
  spatial--temporal graph neural network with massive missing data}}.
\newblock \btxjournalfont {IEEE Trans. Autom. Sci. Eng.}, 21(3):3576--3587,
  \btxprintmonthyear{.}{7}{2024}{short}.

\bibitem {art_339}
\btxnamefont {J.~\btxlastnamefont {Zhang}}, \btxnamefont {X.~\btxlastnamefont
  {Hu}}, \btxnamefont {C.~\btxlastnamefont {Ren}}, \btxnamefont
  {Z.~\btxlastnamefont {Zhan}}, \btxnamefont {T.~\btxlastnamefont
  {Wang}}\btxandcomma {} \btxandshort {.}\ \btxnamefont {D.~\btxlastnamefont
  {Ma}}\btxauthorcolon\ \btxtitlefont {\btxifchangecase {Short-term load
  forecasting based on spatial-temporal correlation for virtual power
  plant}{Short-term load forecasting based on spatial-temporal correlation for
  virtual power plant}}.
\newblock \Btxinshort {.}\ \btxtitlefont {2024 3rd International Conference on
  Power Systems and Electrical Technology ({PSET})}, \btxpagesshort {.}\
  791--796. \btxpublisherfont {IEEE}, \btxprintmonthyear{.}{8}{2024}{short}.

\bibitem {IMDSSN}
\btxnamefont {J.~\btxlastnamefont {Zhang}}, \btxnamefont {X.~\btxlastnamefont
  {Li}}, \btxnamefont {J.~\btxlastnamefont {Tian}}, \btxnamefont
  {H.~\btxlastnamefont {Luo}}\btxandcomma {} \btxandshort {.}\ \btxnamefont
  {S.~\btxlastnamefont {Yin}}\btxauthorcolon\ \btxjtitlefont {\btxifchangecase
  {An integrated multi-head dual sparse self-attention network for remaining
  useful life prediction}{An integrated multi-head dual sparse self-attention
  network for remaining useful life prediction}}.
\newblock \btxjournalfont {Reliability Engineering \& System Safety},
  233:109096, 2023\ifbtxprintISSN {, \mbox{\btxISSN~\btxISSNfont {0951-8320}}}.
\newblock {\latintext
  \btxurlfont{https://www.sciencedirect.com/science/article/pii/S095183202300011X}}.

\bibitem {GaAN}
\btxnamefont {J.~\btxlastnamefont {Zhang}}, \btxnamefont {X.~\btxlastnamefont
  {Shi}}, \btxnamefont {J.~\btxlastnamefont {Xie}}, \btxnamefont
  {H.~\btxlastnamefont {Ma}}, \btxnamefont {I.~\btxlastnamefont
  {King}}\btxandcomma {} \btxandshort {.}\ \btxnamefont {D.~\btxlastnamefont
  {Yeung}}\btxauthorcolon\ \btxtitlefont {\btxifchangecase {Gaan: Gated
  attention networks for learning on large and spatiotemporal graphs}{GaAN:
  Gated Attention Networks for Learning on Large and Spatiotemporal Graphs}}.
\newblock \Btxinshort {.}\ \btxtitlefont {Proceedings of the Thirty-Fourth
  Conference on Uncertainty in Artificial Intelligence}, \btxpagesshort {.}\
  339--349, 2018.

\bibitem {ST-ResNet}
\btxnamefont {J.~\btxlastnamefont {Zhang}}, \btxnamefont {Y.~\btxlastnamefont
  {Zheng}}\btxandcomma {} \btxandshort {.}\ \btxnamefont {D.~\btxlastnamefont
  {Qi}}\btxauthorcolon\ \btxjtitlefont {\btxifchangecase {Deep spatio-temporal
  residual networks for citywide crowd flows prediction}{Deep Spatio-Temporal
  Residual Networks for Citywide Crowd Flows Prediction}}.
\newblock \btxjournalfont {Proceedings of the AAAI Conference on Artificial
  Intelligence}, 31(1), \btxprintmonthyear{.}{Feb.}{2017}{short}.
\newblock {\latintext
  \btxurlfont{https://ojs.aaai.org/index.php/AAAI/article/view/10735}}.

\bibitem {DeepST}
\btxnamefont {J.~\btxlastnamefont {Zhang}}, \btxnamefont {Y.~\btxlastnamefont
  {Zheng}}, \btxnamefont {D.~\btxlastnamefont {Qi}}, \btxnamefont
  {R.~\btxlastnamefont {Li}}\btxandcomma {} \btxandshort {.}\ \btxnamefont
  {X.~\btxlastnamefont {Yi}}\btxauthorcolon\ \btxtitlefont {\btxifchangecase
  {Dnn-based prediction model for spatio-temporal data}{DNN-based prediction
  model for spatio-temporal data}}.
\newblock \Btxinshort {.}\ \btxtitlefont {Proceedings of the 24th ACM
  SIGSPATIAL International Conference on Advances in Geographic Information
  Systems}, SIGSPACIAL '16, New York, NY, USA, 2016. \btxpublisherfont
  {Association for Computing Machinery}\ifbtxprintISBN {,
  \mbox{\btxISBN~\btxISBNfont {9781450345897}}}.
\newblock {\latintext \btxurlfont{https://doi.org/10.1145/2996913.2997016}}.

\bibitem {MDL}
\btxnamefont {J.~\btxlastnamefont {Zhang}}, \btxnamefont {Y.~\btxlastnamefont
  {Zheng}}, \btxnamefont {J.~\btxlastnamefont {Sun}}\btxandcomma {}
  \btxandshort {.}\ \btxnamefont {D.~\btxlastnamefont {Qi}}\btxauthorcolon\
  \btxjtitlefont {\btxifchangecase {Flow prediction in spatio-temporal networks
  based on multitask deep learning}{Flow Prediction in Spatio-Temporal Networks
  Based on Multitask Deep Learning}}.
\newblock \btxjournalfont {IEEE Transactions on Knowledge and Data
  Engineering}, 32(3):468--478, 2020.

\bibitem {art_48}
\btxnamefont {J.~\btxlastnamefont {Zhang}}, \btxnamefont {P.~\btxlastnamefont
  {Zhou}}, \btxnamefont {Y.~\btxlastnamefont {Zheng}}\btxandcomma {}
  \btxandshort {.}\ \btxnamefont {H.~\btxlastnamefont {Wu}}\btxauthorcolon\
  \btxjtitlefont {\btxifchangecase {Predicting influenza with
  pandemic-awareness via dynamic virtual graph significance
  networks}{Predicting influenza with pandemic-awareness via Dynamic Virtual
  Graph Significance Networks}}.
\newblock \btxjournalfont {Computers in Biology and Medicine}, 158, 2023.
\newblock {\latintext
  \btxurlfont{https://www.scopus.com/inward/record.uri?eid=2-s2.0-85151297427&doi=10.1016

\bibitem {SFM}
\btxnamefont {L.~\btxlastnamefont {Zhang}}, \btxnamefont {C.~\btxlastnamefont
  {Aggarwal}}\btxandcomma {} \btxandshort {.}\ \btxnamefont
  {G.\btxfnamespaceshort J. \btxlastnamefont {Qi}}\btxauthorcolon\
  \btxtitlefont {\btxifchangecase {Stock price prediction via discovering
  multi-frequency trading patterns}{Stock Price Prediction via Discovering
  Multi-Frequency Trading Patterns}}.
\newblock \Btxinshort {.}\ \btxtitlefont {Proceedings of the 23rd ACM SIGKDD
  International Conference on Knowledge Discovery and Data Mining}, KDD '17,
  \btxpageshort {.}\ 2141–2149, New York, NY, USA, 2017. \btxpublisherfont
  {Association for Computing Machinery}\ifbtxprintISBN {,
  \mbox{\btxISBN~\btxISBNfont {9781450348874}}}.
\newblock {\latintext \btxurlfont{https://doi.org/10.1145/3097983.3098117}}.

\bibitem {VAR-MLP}
\btxnamefont {P.~\btxlastnamefont {Zhang}}\btxauthorcolon\ \btxjtitlefont
  {\btxifchangecase {Zhang, g.p.: Time series forecasting using a hybrid arima
  and neural network model. neurocomputing 50, 159-175}{Zhang, G.P.: Time
  Series Forecasting Using a Hybrid ARIMA and Neural Network Model.
  Neurocomputing 50, 159-175}}.
\newblock \btxjournalfont {Neurocomputing}, 50:159--175,
  \btxprintmonthyear{.}{01}{2003}{short}.

\bibitem {art_16}
\btxnamefont {Q.~\btxlastnamefont {Zhang}}, \btxnamefont {J.~\btxlastnamefont
  {Chen}}, \btxnamefont {G.~\btxlastnamefont {Xiao}}, \btxnamefont
  {S.~\btxlastnamefont {He}}\btxandcomma {} \btxandshort {.}\ \btxnamefont
  {K.~\btxlastnamefont {Deng}}\btxauthorcolon\ \btxjtitlefont {\btxifchangecase
  {Transformgraph: A novel short-term electricity net load forecasting
  model}{TransformGraph: A novel short-term electricity net load forecasting
  model}}.
\newblock \btxjournalfont {Energy Reports}, 9:2705 – 2717, 2023.
\newblock {\latintext
  \btxurlfont{https://www.scopus.com/inward/record.uri?eid=2-s2.0-85147090957&doi=10.1016

\bibitem {art_459}
\btxnamefont {Q.~\btxlastnamefont {Zhang}}, \btxnamefont {H.~\btxlastnamefont
  {Ji}}, \btxnamefont {L.~\btxlastnamefont {Li}}\btxandcomma {} \btxandshort
  {.}\ \btxnamefont {Z.~\btxlastnamefont {Zhu}}\btxauthorcolon\ \btxjtitlefont
  {\btxifchangecase {Automatic modulation recognition of unknown interference
  signals based on graph model}{Automatic modulation recognition of unknown
  interference signals based on graph model}}.
\newblock \btxjournalfont {IEEE Wirel. Commun. Lett.}, 13(9):2317--2321,
  \btxprintmonthyear{.}{9}{2024}{short}.

\bibitem {art_478}
\btxnamefont {Q.~\btxlastnamefont {Zhang}}, \btxnamefont {P.~\btxlastnamefont
  {Yang}}\btxandcomma {} \btxandshort {.}\ \btxnamefont {Q.~\btxlastnamefont
  {Liu}}\btxauthorcolon\ \btxjtitlefont {\btxifchangecase {A dual-stream
  spatio-temporal fusion network with multi-sensor signals for remaining useful
  life prediction}{A dual-stream spatio-temporal fusion network with
  multi-sensor signals for remaining useful life prediction}}.
\newblock \btxjournalfont {J. Manuf. Syst.}, 76:43--58,
  \btxprintmonthyear{.}{10}{2024}{short}.

\bibitem {art_104}
\btxnamefont {Q.~\btxlastnamefont {Zhang}}, \btxnamefont {K.~\btxlastnamefont
  {Yu}}, \btxnamefont {Z.~\btxlastnamefont {Guo}}, \btxnamefont
  {S.~\btxlastnamefont {Garg}}, \btxnamefont {J.\btxfnamespaceshort
  J.\btxfnamespaceshort P.\btxfnamespaceshort C. \btxlastnamefont {Rodrigues}},
  \btxnamefont {M.\btxfnamespaceshort M. \btxlastnamefont {Hassan}}\btxandcomma
  {} \btxandshort {.}\ \btxnamefont {M.~\btxlastnamefont
  {Guizani}}\btxauthorcolon\ \btxjtitlefont {\btxifchangecase {Graph neural
  network-driven traffic forecasting for the connected internet of
  vehicles}{Graph Neural Network-Driven Traffic Forecasting for the Connected
  Internet of Vehicles}}.
\newblock \btxjournalfont {IEEE Transactions on Network Science and
  Engineering}, 9(5):3015--3027, 2022.

\bibitem {art_120}
\btxnamefont {T.~\btxlastnamefont {Zhang}} \btxandshort {.}\ \btxnamefont
  {G.~\btxlastnamefont {Guo}}\btxauthorcolon\ \btxjtitlefont {\btxifchangecase
  {Graph attention lstm: A spatiotemporal approach for traffic flow
  forecasting}{Graph Attention LSTM: A Spatiotemporal Approach for Traffic Flow
  Forecasting}}.
\newblock \btxjournalfont {IEEE Intelligent Transportation Systems Magazine},
  14(2):190 – 196, 2022.
\newblock {\latintext
  \btxurlfont{https://www.scopus.com/inward/record.uri?eid=2-s2.0-85127914420&doi=10.1109

\bibitem {art_646}
\btxnamefont {W.~\btxlastnamefont {Zhang}}, \btxnamefont {C.~\btxlastnamefont
  {Yin}}, \btxnamefont {H.~\btxlastnamefont {Liu}}, \btxnamefont
  {X.~\btxlastnamefont {Zhou}}\btxandcomma {} \btxandshort {.}\ \btxnamefont
  {H.~\btxlastnamefont {Xiong}}\btxauthorcolon\ \btxtitlefont {\btxifchangecase
  {Irregular multivariate time series forecasting: A transformable patching
  graph neural networks approach}{Irregular Multivariate Time Series
  Forecasting: A Transformable Patching Graph Neural Networks Approach}}.
\newblock \Btxinshort {.}\ \btxtitlefont {Forty-first International Conference
  on Machine Learning}, 2024.
\newblock {\latintext \btxurlfont{https://openreview.net/forum?id=UZlMXUGI6e}}.

\bibitem {art_137}
\btxnamefont {W.~\btxlastnamefont {Zhang}}, \btxnamefont {Y.~\btxlastnamefont
  {Yin}}, \btxnamefont {J.~\btxlastnamefont {Tang}}\btxandcomma {} \btxandshort
  {.}\ \btxnamefont {B.~\btxlastnamefont {Yi}}\btxauthorcolon\ \btxjtitlefont
  {\btxifchangecase {A method for the spatiotemporal correlation prediction of
  the quality of multiple operational processes based on s-ggru}{A method for
  the spatiotemporal correlation prediction of the quality of multiple
  operational processes based on S-GGRU}}.
\newblock \btxjournalfont {Advanced Engineering Informatics}, 58, 2023.
\newblock {\latintext
  \btxurlfont{https://www.scopus.com/inward/record.uri?eid=2-s2.0-85174932585&doi=10.1016

\bibitem {art_666}
\btxnamefont {W.~\btxlastnamefont {Zhang}}, \btxnamefont {L.~\btxlastnamefont
  {Zhang}}, \btxnamefont {J.~\btxlastnamefont {Han}}, \btxnamefont
  {H.~\btxlastnamefont {Liu}}, \btxnamefont {Y.~\btxlastnamefont {Fu}},
  \btxnamefont {J.~\btxlastnamefont {Zhou}}, \btxnamefont {Y.~\btxlastnamefont
  {Mei}}\btxandcomma {} \btxandshort {.}\ \btxnamefont {H.~\btxlastnamefont
  {Xiong}}\btxauthorcolon\ \btxtitlefont {\btxifchangecase {Irregular traffic
  time series forecasting based on asynchronous spatio-temporal graph
  convolutional networks}{Irregular traffic time series forecasting based on
  asynchronous spatio-temporal graph convolutional networks}}.
\newblock \Btxinshort {.}\ \btxtitlefont {Proceedings of the 30th {ACM}
  {SIGKDD} Conference on Knowledge Discovery and Data Mining}, \btxvolumeshort
  {.}~\btxvolumefont {36}, \btxpagesshort {.}\ 4302--4313, New York, NY, USA,
  \btxprintmonthyear{.}{8}{2024}{short}. \btxpublisherfont {ACM}.

\bibitem {LLM_review}
\btxnamefont {X.~\btxlastnamefont {Zhang}}, \btxnamefont {R.\btxfnamespaceshort
  R. \btxlastnamefont {Chowdhury}}, \btxnamefont {R.\btxfnamespaceshort K.
  \btxlastnamefont {Gupta}}\btxandcomma {} \btxandshort {.}\ \btxnamefont
  {J.~\btxlastnamefont {Shang}}\btxauthorcolon\ \btxtitlefont {\btxifchangecase
  {Large language models for time series: A survey}{Large Language Models for
  Time Series: A Survey}}.
\newblock \Btxinshort {.}\ \btxnamefont {K.~\btxlastnamefont {Larson}}\
  (\btxeditorshort {.}): \btxtitlefont {Proceedings of the Thirty-Third
  International Joint Conference on Artificial Intelligence, {IJCAI-24}},
  \btxpagesshort {.}\ 8335--8343. \btxpublisherfont {International Joint
  Conferences on Artificial Intelligence Organization},
  \btxprintmonthyear{.}{8}{2024}{short}.
\newblock {\latintext \btxurlfont{https://doi.org/10.24963/ijcai.2024/921}},
  Survey Track.

\bibitem {TapNet}
\btxnamefont {X.~\btxlastnamefont {Zhang}}, \btxnamefont {Y.~\btxlastnamefont
  {Gao}}, \btxnamefont {J.~\btxlastnamefont {Lin}}\btxandcomma {} \btxandshort
  {.}\ \btxnamefont {C.\btxfnamespaceshort T. \btxlastnamefont
  {Lu}}\btxauthorcolon\ \btxjtitlefont {\btxifchangecase {Tapnet: Multivariate
  time series classification with attentional prototypical network}{TapNet:
  Multivariate Time Series Classification with Attentional Prototypical
  Network}}.
\newblock \btxjournalfont {Proceedings of the AAAI Conference on Artificial
  Intelligence}, 34(04):6845--6852, \btxprintmonthyear{.}{Apr.}{2020}{short}.
\newblock {\latintext
  \btxurlfont{https://ojs.aaai.org/index.php/AAAI/article/view/6165}}.

\bibitem {art_734}
\btxnamefont {X.~\btxlastnamefont {Zhang}}, \btxnamefont {C.~\btxlastnamefont
  {Huang}}, \btxnamefont {Y.~\btxlastnamefont {Xu}}\btxandcomma {} \btxandshort
  {.}\ \btxnamefont {L.~\btxlastnamefont {Xia}}\btxauthorcolon\ \btxtitlefont
  {\btxifchangecase {Spatial-temporal convolutional graph attention networks
  for citywide traffic flow forecasting}{Spatial-temporal convolutional graph
  attention networks for citywide traffic flow forecasting}}.
\newblock \Btxinshort {.}\ \btxtitlefont {Proceedings of the 29th {ACM}
  International Conference on Information \& Knowledge Management}, New York,
  NY, USA, \btxprintmonthyear{.}{10}{2020}{short}. \btxpublisherfont {ACM}.

\bibitem {art_58}
\btxnamefont {X.~\btxlastnamefont {Zhang}}, \btxnamefont {Z.~\btxlastnamefont
  {Long}}, \btxnamefont {J.~\btxlastnamefont {Peng}}, \btxnamefont
  {G.~\btxlastnamefont {Wu}}, \btxnamefont {H.~\btxlastnamefont {Hu}},
  \btxnamefont {M.~\btxlastnamefont {Lyu}}, \btxnamefont {G.~\btxlastnamefont
  {Qin}}\btxandcomma {} \btxandshort {.}\ \btxnamefont {D.~\btxlastnamefont
  {Song}}\btxauthorcolon\ \btxjtitlefont {\btxifchangecase {Fault prediction
  for electromechanical equipment based on spatial-temporal graph
  information}{Fault Prediction for Electromechanical Equipment Based on
  Spatial-Temporal Graph Information}}.
\newblock \btxjournalfont {IEEE Transactions on Industrial Informatics},
  19(2):1413 – 1424, 2023.
\newblock {\latintext
  \btxurlfont{https://www.scopus.com/inward/record.uri?eid=2-s2.0-85130824180&doi=10.1109

\bibitem {art_92}
\btxnamefont {X.~\btxlastnamefont {Zhang}}, \btxnamefont {Y.~\btxlastnamefont
  {Wang}}, \btxnamefont {L.~\btxlastnamefont {Zhang}}, \btxnamefont
  {B.~\btxlastnamefont {Jin}}\btxandcomma {} \btxandshort {.}\ \btxnamefont
  {H.~\btxlastnamefont {Zhang}}\btxauthorcolon\ \btxjtitlefont
  {\btxifchangecase {Exploring unsupervised multivariate time series
  representation learning for chronic disease diagnosis}{Exploring unsupervised
  multivariate time series representation learning for chronic disease
  diagnosis}}.
\newblock \btxjournalfont {International Journal of Data Science and
  Analytics}, 15(2):173 – 186, 2023.
\newblock {\latintext
  \btxurlfont{https://www.scopus.com/inward/record.uri?eid=2-s2.0-85118890785&doi=10.1007

\bibitem {art_645}
\btxnamefont {X.~\btxlastnamefont {Zhang}}, \btxnamefont {M.~\btxlastnamefont
  {Zeman}}, \btxnamefont {T.~\btxlastnamefont {Tsiligkaridis}}\btxandcomma {}
  \btxandshort {.}\ \btxnamefont {M.~\btxlastnamefont {Zitnik}}\btxauthorcolon\
  \btxtitlefont {\btxifchangecase {Graph-guided network for irregularly sampled
  multivariate time series}{Graph-Guided Network for Irregularly Sampled
  Multivariate Time Series}}.
\newblock
  \url{https://zitniklab.hms.harvard.edu/publications/papers/raindrop-iclr22.pdf},
  2022.

\bibitem {art_350}
\btxnamefont {Y.~\btxlastnamefont {Zhang}}, \btxnamefont {C.~\btxlastnamefont
  {Liu}}, \btxnamefont {X.~\btxlastnamefont {Rao}}, \btxnamefont
  {X.~\btxlastnamefont {Zhang}}\btxandcomma {} \btxandshort {.}\ \btxnamefont
  {Y.~\btxlastnamefont {Zhou}}\btxauthorcolon\ \btxjtitlefont {\btxifchangecase
  {Spatial-temporal load forecasting of electric vehicle charging stations
  based on graph neural network}{Spatial-temporal load forecasting of electric
  vehicle charging stations based on graph neural network}}.
\newblock \btxjournalfont {J. Intell. Fuzzy Syst.}, \btxpagesshort {.}\ 1--16,
  \btxprintmonthyear{.}{11}{2023}{short}.

\bibitem {art_147}
\btxnamefont {Y.~\btxlastnamefont {Zhang}}, \btxnamefont {J.~\btxlastnamefont
  {Liu}}, \btxnamefont {B.~\btxlastnamefont {Guo}}, \btxnamefont
  {Z.~\btxlastnamefont {Wang}}, \btxnamefont {Y.~\btxlastnamefont
  {Liang}}\btxandcomma {} \btxandshort {.}\ \btxnamefont {Z.~\btxlastnamefont
  {Yu}}\btxauthorcolon\ \btxjtitlefont {\btxifchangecase {App popularity
  prediction by incorporating time-varying hierarchical interactions}{App
  Popularity Prediction by Incorporating Time-Varying Hierarchical
  Interactions}}.
\newblock \btxjournalfont {IEEE Transactions on Mobile Computing},
  21(5):1566--1579, 2022.

\bibitem {art_32}
\btxnamefont {Y.~\btxlastnamefont {Zhang}}, \btxnamefont {L.~\btxlastnamefont
  {Xu}}\btxandcomma {} \btxandshort {.}\ \btxnamefont {J.~\btxlastnamefont
  {Yu}}\btxauthorcolon\ \btxjtitlefont {\btxifchangecase {Significant wave
  height prediction based on dynamic graph neural network with fusion of ocean
  characteristics}{Significant wave height prediction based on dynamic graph
  neural network with fusion of ocean characteristics}}.
\newblock \btxjournalfont {Dynamics of Atmospheres and Oceans}, 103, 2023.
\newblock {\latintext
  \btxurlfont{https://www.scopus.com/inward/record.uri?eid=2-s2.0-85168558949&doi=10.1016

\bibitem {Crossformer}
\btxnamefont {Y.~\btxlastnamefont {Zhang}} \btxandshort {.}\ \btxnamefont
  {J.~\btxlastnamefont {Yan}}\btxauthorcolon\ \btxtitlefont {\btxifchangecase
  {Crossformer: Transformer utilizing cross-dimension dependency for
  multivariate time series forecasting}{Crossformer: Transformer Utilizing
  Cross-Dimension Dependency for Multivariate Time Series Forecasting}}.
\newblock \Btxinshort {.}\ \btxtitlefont {The Eleventh International Conference
  on Learning Representations}, 2023.
\newblock {\latintext
  \btxurlfont{https://openreview.net/forum?id=vSVLM2j9eie}}.

\bibitem {art_76}
\btxnamefont {Z.~\btxlastnamefont {Zhang}}, \btxnamefont {Y.~\btxlastnamefont
  {Han}}, \btxnamefont {B.~\btxlastnamefont {Ma}}, \btxnamefont
  {M.~\btxlastnamefont {Liu}}\btxandcomma {} \btxandshort {.}\ \btxnamefont
  {Z.~\btxlastnamefont {Geng}}\btxauthorcolon\ \btxjtitlefont {\btxifchangecase
  {Temporal chain network with intuitive attention mechanism for long-term
  series forecasting}{Temporal Chain Network With Intuitive Attention Mechanism
  for Long-Term Series Forecasting}}.
\newblock \btxjournalfont {IEEE Transactions on Instrumentation and
  Measurement}, 72, 2023.
\newblock {\latintext
  \btxurlfont{https://www.scopus.com/inward/record.uri?eid=2-s2.0-85174801615&doi=10.1109

\bibitem {art_703}
\btxnamefont {Z.~\btxlastnamefont {Zhang}}, \btxnamefont {W.~\btxlastnamefont
  {Li}}\btxandcomma {} \btxandshort {.}\ \btxnamefont {H.~\btxlastnamefont
  {Liu}}\btxauthorcolon\ \btxtitlefont {\btxifchangecase {Multivariate time
  series forecasting by graph attention networks with theoretical
  guarantees}{Multivariate Time Series Forecasting By Graph Attention Networks
  With Theoretical Guarantees}}, 2023.
\newblock {\latintext \btxurlfont{https://openreview.net/forum?id=qg2XdQ773R}}.

\bibitem {art_121}
\btxnamefont {C.~\btxlastnamefont {Zhao}}, \btxnamefont {X.~\btxlastnamefont
  {Li}}, \btxnamefont {Z.~\btxlastnamefont {Shao}}, \btxnamefont
  {H.~\btxlastnamefont {Yang}}\btxandcomma {} \btxandshort {.}\ \btxnamefont
  {F.~\btxlastnamefont {Wang}}\btxauthorcolon\ \btxjtitlefont {\btxifchangecase
  {Multi-featured spatial-temporal and dynamic multi-graph convolutional
  network for metro passenger flow prediction}{Multi-featured spatial-temporal
  and dynamic multi-graph convolutional network for metro passenger flow
  prediction}}.
\newblock \btxjournalfont {Connection Science}, 34(1):1252 – 1272, 2022.
\newblock {\latintext
  \btxurlfont{https://www.scopus.com/inward/record.uri?eid=2-s2.0-85128853908&doi=10.1080

\bibitem {MTAD-GAT}
\btxnamefont {H.~\btxlastnamefont {Zhao}}, \btxnamefont {Y.~\btxlastnamefont
  {Wang}}, \btxnamefont {J.~\btxlastnamefont {Duan}}, \btxnamefont
  {C.~\btxlastnamefont {Huang}}, \btxnamefont {D.~\btxlastnamefont {Cao}},
  \btxnamefont {Y.~\btxlastnamefont {Tong}}, \btxnamefont {B.~\btxlastnamefont
  {Xu}}, \btxnamefont {J.~\btxlastnamefont {Bai}}, \btxnamefont
  {J.~\btxlastnamefont {Tong}}\btxandcomma {} \btxandshort {.}\ \btxnamefont
  {Q.~\btxlastnamefont {Zhang}}\btxauthorcolon\ \btxtitlefont {\btxifchangecase
  {Multivariate time-series anomaly detection via graph attention
  network}{Multivariate Time-Series Anomaly Detection via Graph Attention
  Network}}.
\newblock \Btxinshort {.}\ \btxtitlefont {2020 IEEE International Conference on
  Data Mining (ICDM)}, \btxpagesshort {.}\ 841--850, 2020.

\bibitem {T-GCN}
\btxnamefont {L.~\btxlastnamefont {Zhao}}, \btxnamefont {Y.~\btxlastnamefont
  {Song}}, \btxnamefont {C.~\btxlastnamefont {Zhang}}, \btxnamefont
  {Y.~\btxlastnamefont {Liu}}, \btxnamefont {P.~\btxlastnamefont {Wang}},
  \btxnamefont {T.~\btxlastnamefont {Lin}}, \btxnamefont {M.~\btxlastnamefont
  {Deng}}\btxandcomma {} \btxandshort {.}\ \btxnamefont {H.~\btxlastnamefont
  {Li}}\btxauthorcolon\ \btxjtitlefont {\btxifchangecase {T-gcn: A temporal
  graph convolutional network for traffic prediction}{T-GCN: A Temporal Graph
  Convolutional Network for Traffic Prediction}}.
\newblock \btxjournalfont {IEEE Transactions on Intelligent Transportation
  Systems}, 21(9):3848--3858, 2020.

\bibitem {art_23}
\btxnamefont {T.~\btxlastnamefont {Zhao}}, \btxnamefont {M.~\btxlastnamefont
  {Yue}}\btxandcomma {} \btxandshort {.}\ \btxnamefont {J.~\btxlastnamefont
  {Wang}}\btxauthorcolon\ \btxjtitlefont {\btxifchangecase {Structure-informed
  graph learning of networked dependencies for online prediction of power
  system transient dynamics}{Structure-Informed Graph Learning of Networked
  Dependencies for Online Prediction of Power System Transient Dynamics}}.
\newblock \btxjournalfont {IEEE Transactions on Power Systems}, 37(6):4885 –
  4895, 2022.
\newblock {\latintext
  \btxurlfont{https://www.scopus.com/inward/record.uri?eid=2-s2.0-85125352651&doi=10.1109

\bibitem {art_435}
\btxnamefont {W.~\btxlastnamefont {Zhao}}, \btxnamefont {G.~\btxlastnamefont
  {Yuan}}, \btxnamefont {R.~\btxlastnamefont {Bing}}, \btxnamefont
  {R.~\btxlastnamefont {Lu}}\btxandcomma {} \btxandshort {.}\ \btxnamefont
  {Y.~\btxlastnamefont {Shen}}\btxauthorcolon\ \btxjtitlefont {\btxifchangecase
  {Periodicity aware spatial-temporal adaptive hypergraph neural network for
  traffic forecasting}{Periodicity aware spatial-temporal adaptive hypergraph
  neural network for traffic forecasting}}.
\newblock \btxjournalfont {Geoinformatica},
  \btxprintmonthyear{.}{8}{2024}{short}.

\bibitem {art_125}
\btxnamefont {W.~\btxlastnamefont {Zhao}}, \btxnamefont {S.~\btxlastnamefont
  {Zhang}}, \btxnamefont {B.~\btxlastnamefont {Wang}}\btxandcomma {}
  \btxandshort {.}\ \btxnamefont {B.~\btxlastnamefont {Zhou}}\btxauthorcolon\
  \btxjtitlefont {\btxifchangecase {Spatio-temporal causal graph attention
  network for traffic flow prediction in intelligent transportation
  systems}{Spatio-temporal causal graph attention network for traffic flow
  prediction in intelligent transportation systems}}.
\newblock \btxjournalfont {PeerJ Computer Science}, 9, 2023.
\newblock {\latintext
  \btxurlfont{https://www.scopus.com/inward/record.uri?eid=2-s2.0-85168804113&doi=10.7717

\bibitem {art_484}
\btxnamefont {B.~\btxlastnamefont {Zheng}}, \btxnamefont {L.~\btxlastnamefont
  {Ming}}, \btxnamefont {K.~\btxlastnamefont {Zeng}}, \btxnamefont
  {M.~\btxlastnamefont {Zhou}}, \btxnamefont {X.~\btxlastnamefont {Zhang}},
  \btxnamefont {T.~\btxlastnamefont {Ye}}, \btxnamefont {B.~\btxlastnamefont
  {Yang}}, \btxnamefont {X.~\btxlastnamefont {Zhou}}\btxandcomma {}
  \btxandshort {.}\ \btxnamefont {C.\btxfnamespaceshort S. \btxlastnamefont
  {Jensen}}\btxauthorcolon\ \btxjtitlefont {\btxifchangecase {Adversarial graph
  neural network for multivariate time series anomaly detection}{Adversarial
  graph neural network for multivariate time series anomaly detection}}.
\newblock \btxjournalfont {IEEE Trans. Knowl. Data Eng.}, 36(12):7612--7626,
  \btxprintmonthyear{.}{12}{2024}{short}.

\bibitem {GMAN}
\btxnamefont {C.~\btxlastnamefont {Zheng}}, \btxnamefont {X.~\btxlastnamefont
  {Fan}}, \btxnamefont {C.~\btxlastnamefont {Wang}}\btxandcomma {} \btxandshort
  {.}\ \btxnamefont {J.~\btxlastnamefont {Qi}}\btxauthorcolon\ \btxjtitlefont
  {\btxifchangecase {Gman: A graph multi-attention network for traffic
  prediction}{GMAN: A Graph Multi-Attention Network for Traffic Prediction}}.
\newblock \btxjournalfont {Proceedings of the AAAI Conference on Artificial
  Intelligence}, 34(01):1234--1241, \btxprintmonthyear{.}{Apr.}{2020}{short}.
\newblock {\latintext
  \btxurlfont{https://ojs.aaai.org/index.php/AAAI/article/view/5477}}.

\bibitem {art_714}
\btxnamefont {S.~\btxlastnamefont {Zheng}}, \btxnamefont {Z.~\btxlastnamefont
  {Gao}}, \btxnamefont {W.~\btxlastnamefont {Cao}}, \btxnamefont
  {J.~\btxlastnamefont {Bian}}\btxandcomma {} \btxandshort {.}\ \btxnamefont
  {T.\btxfnamespaceshort Y. \btxlastnamefont {Liu}}\btxauthorcolon\
  \btxtitlefont {\btxifchangecase {Hierst: A unified hierarchical
  spatial-temporal framework for covid-19 trend forecasting}{HierST: A Unified
  Hierarchical Spatial-temporal Framework for COVID-19 Trend Forecasting}}.
\newblock \Btxinshort {.}\ \btxtitlefont {Proceedings of the 30th ACM
  International Conference on Information \& Knowledge Management}, CIKM '21,
  \btxpageshort {.}\ 4383–4392, New York, NY, USA, 2021. \btxpublisherfont
  {Association for Computing Machinery}\ifbtxprintISBN {,
  \mbox{\btxISBN~\btxISBNfont {9781450384469}}}.
\newblock {\latintext \btxurlfont{https://doi.org/10.1145/3459637.3481927}}.

\bibitem {art_477}
\btxnamefont {Y.~\btxlastnamefont {Zheng}}, \btxnamefont {H.\btxfnamespaceshort
  Y. \btxlastnamefont {Koh}}, \btxnamefont {M.~\btxlastnamefont {Jin}},
  \btxnamefont {L.~\btxlastnamefont {Chi}}, \btxnamefont {H.~\btxlastnamefont
  {Wang}}, \btxnamefont {K.\btxfnamespaceshort T. \btxlastnamefont {Phan}},
  \btxnamefont {Y.\btxfnamespaceshort P.\btxfnamespaceshort P. \btxlastnamefont
  {Chen}}, \btxnamefont {S.~\btxlastnamefont {Pan}}\btxandcomma {} \btxandshort
  {.}\ \btxnamefont {W.~\btxlastnamefont {Xiang}}\btxauthorcolon\
  \btxjtitlefont {\btxifchangecase {Graph spatiotemporal process for
  multivariate time series anomaly detection with missing values}{Graph
  spatiotemporal process for multivariate time series anomaly detection with
  missing values}}.
\newblock \btxjournalfont {Inf. Fusion}, 106(102255):102255,
  \btxprintmonthyear{.}{6}{2024}{short}.

\bibitem {Zheng2024}
\btxnamefont {Y.~\btxlastnamefont {Zheng}}, \btxnamefont {L.~\btxlastnamefont
  {Yi}}\btxandcomma {} \btxandshort {.}\ \btxnamefont {Z.~\btxlastnamefont
  {Wei}}\btxauthorcolon\ \btxjtitlefont {\btxifchangecase {A survey of dynamic
  graph neural networks}{A survey of dynamic graph neural networks}}.
\newblock \btxjournalfont {Frontiers of Computer Science}, 19(6),
  \btxprintmonthyear{.}{12}{2024}{short}\ifbtxprintISSN {,
  \mbox{\btxISSN~\btxISSNfont {2095-2236}}}.
\newblock {\latintext
  \btxurlfont{http://dx.doi.org/10.1007/s11704-024-3853-2}}.

\bibitem {RGNN}
\btxnamefont {P.~\btxlastnamefont {Zhong}}, \btxnamefont {D.~\btxlastnamefont
  {Wang}}\btxandcomma {} \btxandshort {.}\ \btxnamefont {C.~\btxlastnamefont
  {Miao}}\btxauthorcolon\ \btxjtitlefont {\btxifchangecase {Eeg-based emotion
  recognition using regularized graph neural networks}{EEG-Based Emotion
  Recognition Using Regularized Graph Neural Networks}}.
\newblock \btxjournalfont {IEEE Transactions on Affective Computing},
  13(3):1290--1301, 2022.

\bibitem {BeatGAN}
\btxnamefont {B.~\btxlastnamefont {Zhou}}, \btxnamefont {S.~\btxlastnamefont
  {Liu}}, \btxnamefont {B.~\btxlastnamefont {Hooi}}, \btxnamefont
  {X.~\btxlastnamefont {Cheng}}\btxandcomma {} \btxandshort {.}\ \btxnamefont
  {J.~\btxlastnamefont {Ye}}\btxauthorcolon\ \btxtitlefont {\btxifchangecase
  {Beatgan: Anomalous rhythm detection using adversarially generated time
  series}{BeatGAN: Anomalous Rhythm Detection using Adversarially Generated
  Time Series}}.
\newblock \Btxinshort {.}\ \btxtitlefont {Proceedings of the Twenty-Eighth
  International Joint Conference on Artificial Intelligence, {IJCAI-19}},
  \btxpagesshort {.}\ 4433--4439. \btxpublisherfont {International Joint
  Conferences on Artificial Intelligence Organization},
  \btxprintmonthyear{.}{7}{2019}{short}.
\newblock {\latintext \btxurlfont{https://doi.org/10.24963/ijcai.2019/616}}.

\bibitem {art_465}
\btxnamefont {B.~\btxlastnamefont {Zhou}}, \btxnamefont {H.~\btxlastnamefont
  {Zhou}}, \btxnamefont {W.~\btxlastnamefont {Wang}}, \btxnamefont
  {L.~\btxlastnamefont {Chen}}, \btxnamefont {J.~\btxlastnamefont
  {Ma}}\btxandcomma {} \btxandshort {.}\ \btxnamefont {Z.~\btxlastnamefont
  {Zheng}}\btxauthorcolon\ \btxjtitlefont {\btxifchangecase {{HDM-GNN}: A
  heterogeneous dynamic multi-view graph neural network for crime
  prediction}{{HDM-GNN}: A Heterogeneous Dynamic Multi-view Graph Neural
  Network for crime prediction}}.
\newblock \btxjournalfont {ACM Trans. Sens. Netw.},
  \btxprintmonthyear{.}{5}{2024}{short}.

\bibitem {Informer}
\btxnamefont {H.~\btxlastnamefont {Zhou}}, \btxnamefont {S.~\btxlastnamefont
  {Zhang}}, \btxnamefont {J.~\btxlastnamefont {Peng}}, \btxnamefont
  {S.~\btxlastnamefont {Zhang}}, \btxnamefont {J.~\btxlastnamefont {Li}},
  \btxnamefont {H.~\btxlastnamefont {Xiong}}\btxandcomma {} \btxandshort {.}\
  \btxnamefont {W.~\btxlastnamefont {Zhang}}\btxauthorcolon\ \btxjtitlefont
  {\btxifchangecase {Informer: Beyond efficient transformer for long sequence
  time-series forecasting}{Informer: Beyond Efficient Transformer for Long
  Sequence Time-Series Forecasting}}.
\newblock \btxjournalfont {Proceedings of the AAAI Conference on Artificial
  Intelligence}, 35(12):11106–11115,
  \btxprintmonthyear{.}{5}{2021}{short}\ifbtxprintISSN {,
  \mbox{\btxISSN~\btxISSNfont {2159-5399}}}.
\newblock {\latintext
  \btxurlfont{http://dx.doi.org/10.1609/aaai.v35i12.17325}}.

\bibitem {review_traffico_con_codici}
\btxnamefont {J.~\btxlastnamefont {Zhou}}, \btxnamefont {G.~\btxlastnamefont
  {Cui}}, \btxnamefont {S.~\btxlastnamefont {Hu}}, \btxnamefont
  {Z.~\btxlastnamefont {Zhang}}, \btxnamefont {C.~\btxlastnamefont {Yang}},
  \btxnamefont {Z.~\btxlastnamefont {Liu}}, \btxnamefont {L.~\btxlastnamefont
  {Wang}}, \btxnamefont {C.~\btxlastnamefont {Li}}\btxandcomma {} \btxandshort
  {.}\ \btxnamefont {M.~\btxlastnamefont {Sun}}\btxauthorcolon\ \btxjtitlefont
  {\btxifchangecase {Graph neural networks: A review of methods and
  applications}{Graph neural networks: A review of methods and applications}}.
\newblock \btxjournalfont {AI Open}, 1:57--81, 2020\ifbtxprintISSN {,
  \mbox{\btxISSN~\btxISSNfont {2666-6510}}}.
\newblock {\latintext
  \btxurlfont{https://www.sciencedirect.com/science/article/pii/S2666651021000012}}.

\bibitem {FEDformer}
\btxnamefont {T.~\btxlastnamefont {Zhou}}, \btxnamefont {Z.~\btxlastnamefont
  {Ma}}, \btxnamefont {Q.~\btxlastnamefont {Wen}}, \btxnamefont
  {X.~\btxlastnamefont {Wang}}, \btxnamefont {L.~\btxlastnamefont
  {Sun}}\btxandcomma {} \btxandshort {.}\ \btxnamefont {R.~\btxlastnamefont
  {Jin}}\btxauthorcolon\ \btxtitlefont {\btxifchangecase {{FED}former:
  Frequency enhanced decomposed transformer for long-term series
  forecasting}{{FED}former: Frequency Enhanced Decomposed Transformer for
  Long-term Series Forecasting}}.
\newblock \Btxinshort {.}\ \btxnamefont {K.~\btxlastnamefont {Chaudhuri}},
  \btxnamefont {S.~\btxlastnamefont {Jegelka}}, \btxnamefont
  {L.~\btxlastnamefont {Song}}, \btxnamefont {C.~\btxlastnamefont
  {Szepesvari}}, \btxnamefont {G.~\btxlastnamefont {Niu}}\btxandcomma {}
  \btxandshort {.}\ \btxnamefont {S.~\btxlastnamefont {Sabato}}\
  (\btxeditorsshort {.}): \btxtitlefont {Proceedings of the 39th International
  Conference on Machine Learning}, \btxvolumeshort {.}\ \btxvolumefont {162}
  \btxofseriesshort {.}\ \btxtitlefont {Proceedings of Machine Learning
  Research}, \btxpagesshort {.}\ 27268--27286. \btxpublisherfont {PMLR},
  \btxprintmonthyear{.}{17--23 Jul}{2022}{short}.
\newblock {\latintext
  \btxurlfont{https://proceedings.mlr.press/v162/zhou22g.html}}.

\bibitem {art_723}
\btxnamefont {Z.~\btxlastnamefont {Zhou}}, \btxnamefont {J.~\btxlastnamefont
  {Shi}}, \btxnamefont {H.~\btxlastnamefont {Zhang}}, \btxnamefont
  {Q.~\btxlastnamefont {Chen}}, \btxnamefont {X.~\btxlastnamefont {Wang}},
  \btxnamefont {H.~\btxlastnamefont {Chen}}\btxandcomma {} \btxandshort {.}\
  \btxnamefont {Y.~\btxlastnamefont {Wang}}\btxauthorcolon\ \btxtitlefont
  {\btxifchangecase {{CreST}: A credible spatiotemporal learning framework for
  uncertainty-aware traffic forecasting}{{CreST}: A credible spatiotemporal
  learning framework for uncertainty-aware traffic forecasting}}.
\newblock \Btxinshort {.}\ \btxtitlefont {Proceedings of the 17th {ACM}
  International Conference on Web Search and Data Mining}, \btxpagesshort {.}\
  985--993, New York, NY, USA, \btxprintmonthyear{.}{3}{2024}{short}.
  \btxpublisherfont {ACM}.

\bibitem {art_673}
\btxnamefont {G.~\btxlastnamefont {Zhu}}, \btxnamefont {H.~\btxlastnamefont
  {Hou}}, \btxnamefont {P.~\btxlastnamefont {Wang}}, \btxnamefont
  {C.~\btxlastnamefont {Yuan}}\btxandcomma {} \btxandshort {.}\ \btxnamefont
  {Y.~\btxlastnamefont {Huang}}\btxauthorcolon\ \btxtitlefont {\btxifchangecase
  {{STSD}: Modeling spatial temporal staticity and dynamicity in traffic
  forecasting}{{STSD}: Modeling spatial temporal staticity and dynamicity in
  traffic forecasting}}.
\newblock \Btxinshort {.}\ \btxtitlefont {2023 {IEEE} International Conference
  on Data Mining ({ICDM})}. \btxpublisherfont {IEEE},
  \btxprintmonthyear{.}{12}{2023}{short}.

\bibitem {PDCNN}
\btxnamefont {Q.~\btxlastnamefont {Zhu}}, \btxnamefont {J.~\btxlastnamefont
  {Chen}}, \btxnamefont {L.~\btxlastnamefont {Zhu}}, \btxnamefont
  {X.~\btxlastnamefont {Duan}}\btxandcomma {} \btxandshort {.}\ \btxnamefont
  {Y.~\btxlastnamefont {Liu}}\btxauthorcolon\ \btxjtitlefont {\btxifchangecase
  {Wind speed prediction with spatio–temporal correlation: A deep learning
  approach}{Wind Speed Prediction with Spatio–Temporal Correlation: A Deep
  Learning Approach}}.
\newblock \btxjournalfont {Energies}, 11(4):705,
  \btxprintmonthyear{.}{3}{2018}{short}\ifbtxprintISSN {,
  \mbox{\btxISSN~\btxISSNfont {1996-1073}}}.
\newblock {\latintext \btxurlfont{http://dx.doi.org/10.3390/en11040705}}.

\bibitem {AGCNN}
\btxnamefont {Q.~\btxlastnamefont {Zhu}}, \btxnamefont {Q.~\btxlastnamefont
  {Xiong}}, \btxnamefont {Z.~\btxlastnamefont {Yang}}\btxandcomma {}
  \btxandshort {.}\ \btxnamefont {Y.~\btxlastnamefont {Yu}}\btxauthorcolon\
  \btxjtitlefont {\btxifchangecase {A novel feature-fusion-based end-to-end
  approach for remaining useful life prediction}{A novel feature-fusion-based
  end-to-end approach for remaining useful life prediction}}.
\newblock \btxjournalfont {J. Intell. Manuf.}, 34(8):3495–3505,
  \btxprintmonthyear{.}{sep}{2022}{short}\ifbtxprintISSN {,
  \mbox{\btxISSN~\btxISSNfont {0956-5515}}}.
\newblock {\latintext \btxurlfont{https://doi.org/10.1007/s10845-022-02015-x}}.

\bibitem {art_411}
\btxnamefont {X.~\btxlastnamefont {Zhu}}, \btxnamefont {C.~\btxlastnamefont
  {Liu}}, \btxnamefont {L.~\btxlastnamefont {Zhao}}\btxandcomma {} \btxandshort
  {.}\ \btxnamefont {S.~\btxlastnamefont {Wang}}\btxauthorcolon\ \btxjtitlefont
  {\btxifchangecase {{EEG} emotion recognition network based on attention and
  spatiotemporal convolution}{{EEG} emotion recognition network based on
  attention and spatiotemporal convolution}}.
\newblock \btxjournalfont {Sensors (Basel)}, 24(11):3464,
  \btxprintmonthyear{.}{5}{2024}{short}.

\bibitem {art_414}
\btxnamefont {X.~\btxlastnamefont {Zhu}}, \btxnamefont {Y.~\btxlastnamefont
  {Zhang}}, \btxnamefont {H.~\btxlastnamefont {Ying}}, \btxnamefont
  {H.~\btxlastnamefont {Chi}}, \btxnamefont {G.~\btxlastnamefont
  {Sun}}\btxandcomma {} \btxandshort {.}\ \btxnamefont {L.~\btxlastnamefont
  {Zeng}}\btxauthorcolon\ \btxjtitlefont {\btxifchangecase {Modeling epidemic
  dynamics using graph attention based spatial temporal networks}{Modeling
  epidemic dynamics using Graph Attention based Spatial Temporal networks}}.
\newblock \btxjournalfont {PLoS One}, 19(7):e0307159,
  \btxprintmonthyear{.}{7}{2024}{short}.

\bibitem {DAGMM}
\btxnamefont {B.~\btxlastnamefont {Zong}}, \btxnamefont {Q.~\btxlastnamefont
  {Song}}, \btxnamefont {M.\btxfnamespaceshort R. \btxlastnamefont {Min}},
  \btxnamefont {W.~\btxlastnamefont {Cheng}}, \btxnamefont {C.~\btxlastnamefont
  {Lumezanu}}, \btxnamefont {D.~\btxlastnamefont {Cho}}\btxandcomma {}
  \btxandshort {.}\ \btxnamefont {H.~\btxlastnamefont {Chen}}\btxauthorcolon\
  \btxtitlefont {\btxifchangecase {Deep autoencoding gaussian mixture model for
  unsupervised anomaly detection}{Deep Autoencoding Gaussian Mixture Model for
  Unsupervised Anomaly Detection}}.
\newblock \Btxinshort {.}\ \btxtitlefont {International Conference on Learning
  Representations}, 2018.
\newblock {\latintext \btxurlfont{https://openreview.net/forum?id=BJJLHbb0-}}.

\bibitem {art_716}
\btxnamefont {D.~\btxlastnamefont {Zou}}, \btxnamefont {S.~\btxlastnamefont
  {Wang}}, \btxnamefont {X.~\btxlastnamefont {Li}}, \btxnamefont
  {H.~\btxlastnamefont {Peng}}, \btxnamefont {Y.~\btxlastnamefont {Wang}},
  \btxnamefont {C.~\btxlastnamefont {Liu}}, \btxnamefont {K.~\btxlastnamefont
  {Sheng}}\btxandcomma {} \btxandshort {.}\ \btxnamefont {B.~\btxlastnamefont
  {Zhang}}\btxauthorcolon\ \btxtitlefont {\btxifchangecase {{MultiSPANS}: A
  multi-range spatial-temporal transformer network for traffic forecast via
  structural entropy optimization}{{MultiSPANS}: A multi-range spatial-temporal
  transformer network for traffic forecast via structural entropy
  optimization}}.
\newblock \Btxinshort {.}\ \btxtitlefont {Proceedings of the 17th {ACM}
  International Conference on Web Search and Data Mining}, \btxpagesshort {.}\
  1032--1041, New York, NY, USA, \btxprintmonthyear{.}{3}{2024}{short}.
  \btxpublisherfont {ACM}.

\bibitem {art_334}
\btxnamefont {M.~\btxlastnamefont {Zou}}, \btxnamefont {W.~\btxlastnamefont
  {Huang}}, \btxnamefont {J.~\btxlastnamefont {Jin}}, \btxnamefont
  {B.~\btxlastnamefont {Hu}}\btxandcomma {} \btxandshort {.}\ \btxnamefont
  {Z.~\btxlastnamefont {Liu}}\btxauthorcolon\ \btxjtitlefont {\btxifchangecase
  {Deep spatio-temporal feature fusion learning for multi-step building cooling
  load forecasting}{Deep spatio-temporal feature fusion learning for multi-step
  building cooling load forecasting}}.
\newblock \btxjournalfont {Energy Build.}, 322(114735):114735,
  \btxprintmonthyear{.}{11}{2024}{short}.

\end{thebibliography}

\end{document}